\documentclass[twoside,12pt]{article}
\usepackage{times,mathptm}
\usepackage{amssymb}

\mathchardef\Gammaoffont="7000
\mathchardef\Gamma="0100
\mathchardef\Deltaoffont="7001
\mathchardef\Delta="0101
\mathchardef\Thetaoffont="7002
\mathchardef\Theta="0102
\mathchardef\Lambdaoffont="7003
\mathchardef\Lambda="0103
\mathchardef\Xioffont="7004
\mathchardef\Xi="0104
\mathchardef\Pioffont="7005
\mathchardef\Pi="0105
\mathchardef\Sigmaoffont="7006
\mathchardef\Sigma="0106
\mathchardef\Upsilonoffont="7007
\mathchardef\Upsilon="0107
\mathchardef\Phioffont="7008
\mathchardef\Phi="0108
\mathchardef\Psioffont="7009
\mathchardef\Psi="0109
\mathchardef\Omegaoffont="700A
\mathchardef\Omega="010A
\mathchardef\itype="017B

\catcode`\@=11

\gdef\allowhyphens{\penalty\@M \hskip\z@skip}

\gdef\set@low@box#1{\setbox\tw@\hbox{,}\setbox\z@\hbox{#1}\dimen\z@\ht\z@
     \advance\dimen\z@ -\ht\tw@
     \setbox\z@\hbox{\lower\dimen\z@ \box\z@}\ht\z@\ht\tw@ \dp\z@\dp\tw@ }
\gdef\set@low@boxsingle#1{\setbox\tw@\hbox{\rm,}\setbox\z@\hbox{#1}\dimen\z@\ht\z@
     \advance\dimen\z@ -\ht\tw@
     \setbox\z@\hbox{\lower\dimen\z@ \box\z@}\ht\z@\ht\tw@ \dp\z@\dp\tw@ }
\gdef\@glqq{%
\ifhmode\edef\@SF{\spacefactor\the\spacefactor}%
\else\let\@SF\empty
\fi
\CheckFamily\font\fraknomath\ifSameFamily ``\relax
\else\CheckFamily\font\swab\ifSameFamily ``\relax
\else\leavevmode\set@low@box{''}\box\z@\kern-.04em\allowhyphens\@SF\relax
\fi\fi}
\gdef\glqq{\protect\@glqq\kern+.07em}
\gdef\@grqq{%
\ifhmode\edef\@SF{\spacefactor\the\spacefactor}%
\else\let\@SF\empty 
\fi 
\CheckFamily\font\fraknomath\ifSameFamily ''\relax
\else\CheckFamily\font\swab\ifSameFamily ''\relax
\else\kern+.07em``\kern.07em\@SF\relax
\fi\fi}
\gdef\grqq{\protect\@grqq}
\gdef\@glq{{\ifhmode \edef\@SF{\spacefactor\the\spacefactor}\else
     \let\@SF\empty \fi \leavevmode
     \set@low@boxsingle{'\/}\box\z@\kern-.04em\allowhyphens\@SF\relax}}
\gdef\glq{\protect\@glq\kern+.07em}
\gdef\@grq{\ifhmode \edef\@SF{\spacefactor\the\spacefactor}\else
     \let\@SF\empty \fi \kern-.0125em`\kern.07em\@SF\relax}
\gdef\grq{\protect\@grq}

\catcode`\@=12

\makeatletter
\if \@ptsize 0
   \newfont{\scriptscriptscriptgoth}{ygoth scaled 760}
   \newfont{\scriptscriptgoth}{ygoth scaled 833}
   \newfont{\scriptgoth}{ygoth scaled 912}
   \newfont{\gothnomath}{ygoth}
   \newfont{\Goth}{ygoth scaled \magstephalf}
   \newfont{\GOth}{ygoth scaled \magstep1}
   \newfont{\GOTh}{ygoth scaled \magstep2}
   \newfont{\GOTH}{ygoth scaled \magstep3}

   \newfont{\scriptscriptscriptswab}{yswab scaled 760}
   \newfont{\scriptscriptswab}{yswab scaled 833}
   \newfont{\scriptswab}{yswab scaled 912}
   \newfont{\swab}{yswab}
   \newfont{\Swab}{yswab scaled \magstephalf}
   \newfont{\SWab}{yswab scaled \magstep1}
   \newfont{\SWAb}{yswab scaled \magstep2}
   \newfont{\SWAB}{yswab scaled \magstep3}

   \newfont{\scriptscriptscriptfrak}{yfrak scaled 760}
   \newfont{\scriptscriptfrak}{yfrak scaled 833}
   \newfont{\scriptfrak}{yfrak scaled 912}
   \newfont{\fraknomath}{yfrak}
   \newfont{\Frak}{yfrak scaled \magstephalf}
   \newfont{\FRak}{yfrak scaled \magstep1}
   \newfont{\FRAk}{yfrak scaled \magstep2}
   \newfont{\FRAK}{yfrak scaled \magstep3}

   \newfont{\init}{yinit}
   \newfont{\Init}{yinit scaled \magstephalf}
   \newfont{\INit}{yinit scaled \magstep1}
   \newfont{\INIt}{yinit scaled \magstep2}
   \newfont{\INIT}{yinit scaled \magstep3}
\fi
\if \@ptsize 1
   \newfont{\scriptscriptscriptgoth}{ygoth scaled 833}
   \newfont{\scriptscriptgoth}{ygoth scaled 912}
   \newfont{\scriptgoth}{ygoth}
   \newfont{\gothnomath}{ygoth scaled \magstephalf}
   \newfont{\Goth}{ygoth scaled \magstep1}
   \newfont{\GOth}{ygoth scaled \magstep2}
   \newfont{\GOTh}{ygoth scaled \magstep3}
   \newfont{\GOTH}{ygoth scaled \magstep4}

   \newfont{\scriptscriptscriptswab}{yswab scaled 833}
   \newfont{\scriptscriptswab}{yswab scaled 912}
   \newfont{\scriptswab}{yswab}
   \newfont{\swab}{yswab scaled \magstephalf}
   \newfont{\Swab}{yswab scaled \magstep1}
   \newfont{\SWab}{yswab scaled \magstep2}
   \newfont{\SWAb}{yswab scaled \magstep3}
   \newfont{\SWAB}{yswab scaled \magstep4}

   \newfont{\scriptscriptscriptfrak}{yfrak scaled 833}
   \newfont{\scriptscriptfrak}{yfrak scaled 912}
   \newfont{\scriptfrak}{yfrak}
   \newfont{\fraknomath}{yfrak scaled \magstephalf}
   \newfont{\Frak}{yfrak scaled \magstep1}
   \newfont{\FRak}{yfrak scaled \magstep2}
   \newfont{\FRAk}{yfrak scaled \magstep3}
   \newfont{\FRAK}{yfrak scaled \magstep4}

   \newfont{\init}{yinit scaled \magstephalf}
   \newfont{\Init}{yinit scaled \magstep1}
   \newfont{\INit}{yinit scaled \magstep2}
   \newfont{\INIt}{yinit scaled \magstep3}
   \newfont{\INIT}{yinit scaled \magstep4}
\fi
\if \@ptsize 2
   \newfont{\scriptscriptscriptgoth}{ygoth scaled 912}
   \newfont{\scriptscriptgoth}{ygoth}
   \newfont{\scriptgoth}{ygoth scaled \magstephalf}
   \newfont{\gothnomath}{ygoth scaled \magstep1}
   \newfont{\Goth}{ygoth scaled \magstep2}
   \newfont{\GOth}{ygoth scaled \magstep3}
   \newfont{\GOTh}{ygoth scaled \magstep4}
   \newfont{\GOTH}{ygoth scaled \magstep5}

   \newfont{\scriptscriptscriptswab}{yswab scaled 912}
   \newfont{\scriptscriptswab}{yswab}
   \newfont{\scriptswab}{yswab scaled \magstephalf}
   \newfont{\swab}{yswab scaled \magstep1}
   \newfont{\Swab}{yswab scaled \magstep2}
   \newfont{\SWab}{yswab scaled \magstep3}
   \newfont{\SWAb}{yswab scaled \magstep4}
   \newfont{\SWAB}{yswab scaled \magstep5}

   \newfont{\scriptscriptscriptfrak}{yfrak scaled 912}
   \newfont{\scriptscriptfrak}{yfrak}
   \newfont{\scriptfrak}{yfrak scaled \magstephalf}
   \newfont{\fraknomath}{yfrak scaled \magstep1}
   \newfont{\Frak}{yfrak scaled \magstep2}
   \newfont{\FRak}{yfrak scaled \magstep3}
   \newfont{\FRAk}{yfrak scaled \magstep4}
   \newfont{\FRAK}{yfrak scaled \magstep5}

   \newfont{\init}{yinit scaled \magstep1}
   \newfont{\Init}{yinit scaled \magstep2}
   \newfont{\INit}{yinit scaled \magstep3}
   \newfont{\INIt}{yinit scaled \magstep4}
   \newfont{\INIT}{yinit scaled \magstep5}
\fi

\makeatother
\newif\ifSameFamily
\def\CheckFamily#1#2{\GetFamilyName{#1}\ArgOne
        \GetFamilyName{#2}\ArgTwo
        \ifx\ArgOne\ArgTwo\SameFamilytrue\else\SameFamilyfalse\fi}
\def\GetFamilyName#1{\edef\Tempa{#1}\def\Tempb{#1}\ifx\Tempa\Tempb
        \edef\Tempa{\fontname#1}\fi
        \edef\Tempa{\Tempa\space}%
        \expandafter\iGetFamilyName\Tempa\\}
\def\iGetFamilyName#1 #2\\#3{\def#3{#1}}
\def\DefFontName#1#2{{\escapechar-1\expandafter\expandafter\expandafter
        \iDefFontName\expandafter{\csname#2\endcsname}%
        \xdef#1{\expandafter\string\Tempa}}}
\def\iDefFontName{\def\Tempa}

\newcommand\unprotectedae
{\CheckFamily\font\fraknomath\ifSameFamily *a\else
 \CheckFamily\font\swab\ifSameFamily\char'212\else\"a\fi\fi}

\newcommand\unprotectedue
{\CheckFamily\font\fraknomath\ifSameFamily 
*u\else\CheckFamily\font\swab\ifSameFamily\char'237\else\"u\fi\fi}

\DefFontName\eccclarge{eccc1200}
\DefFontName\eccc{eccc1000}
\DefFontName\ecccsmall{eccc0900}
\DefFontName\ecccfootnotesize{eccc0800}

\newcommand\unprotectedes
{\CheckFamily\font\fraknomath\ifSameFamily\char'215\else
\CheckFamily\font\swab\ifSameFamily\char'215\else  
s\fi\fi}
\newcommand\unprotectedesi
{\CheckFamily\font\fraknomath\ifSameFamily\char'215\else
\CheckFamily\font\swab\ifSameFamily\char'215\else  
\mbox{s\hskip.05em}\fi\fi
\-\ignorespaces}

\newcommand\unprotectedmyparagraphsymbol
{\CheckFamily\font\fraknomath\ifSameFamily 
\char'244\else\CheckFamily\font\swab\ifSameFamily
\char'244\else\S\fi\fi}

\renewcommand\ae{\protect\unprotectedae}

\newcommand\ue  {\protect\unprotectedue}

\newcommand\es  {\protect\unprotectedes}

\newcommand\esi {\protect\unprotectedesi}  %wortinterne Version
\newcommand\myparagraphsymbol{\protect\unprotectedmyparagraphsymbol}

\newcommand\namefont{}

\newcommand\footroom{\raisebox{-1.5ex}{\rule{0ex}{.5ex}}}

\newcommand\smallfootroom{\raisebox{-1.0ex}{\rule{0ex}{3ex}}}

\newcommand\headroom{\rule{0ex}{2.8ex}}

\newcommand\bernhard{Bern\-hard}

\newcommand\carlos  {Car\-lo\es}

\newcommand\christoph{Christoph}

\newcommand\claus   {Clau\es}
\newcommand\david   {Da\-vid}

\newcommand\jacques {Jacque\es}

\newcommand\jan     {Jan}

\newcommand\juergen {J\ue r\-gen}
\newcommand\jean    {Jean}

\newcommand\klaus   {Klau\es}

\newcommand\peter   {Peter}
\newcommand\pierre  {Pierre}

\newcommand\ruediger{R\ue\-di\-ger}

\newcommand\wolfgang{Wolf\-gang}
\newcommand\ulrich  {Ul\-rich}

\newcommand\avenhaus        {Aven\-hau\es}
\newcommand\avenhausname    {\juergen\ \avenhaus}

\newcommand\becker          {Becker}
\newcommand\beckername      {\klaus\ \becker}

\newcommand\bergstra        {{\namefont Bergstra}}
\newcommand\bergstraname    {{\namefont\jan\ A. \bergstra}}

\newcommand\church          {{\namefont Church}}

\newcommand\dershowitz      {Dersho\-witz}
\newcommand\dershowitzname  {Nachum \dershowitz}

\newcommand\gabbay          {Gabbay}
\newcommand\gabbayname      {Dov M. \gabbay}

\newcommand\geser           {Geser}
\newcommand\gesername       {Alfons \geser}

\newcommand\gramlich        {Gram\-lich}
\newcommand\gramlichname    {\bernhard\ \gramlich}

\newcommand\hamoen          {Hamoen}
\newcommand\hamoenname      {Erik \hamoen}

\newcommand\huet            {Huet}
\newcommand\huetname        {G\'erard \huet}

\newcommand\jouannaud       {Jouannaud}
\newcommand\jouannaudname   {\jean-\pierre\ \jouannaud}

\newcommand\kaplan          {Kaplan}
\newcommand\kaplanname      {St\'ephane \kaplan}
\newcommand\kapur           {Kapur}
\newcommand\kapurname       {Deepak \kapur}

\newcommand\klop            {Klop}
\newcommand\klopname        {\jan\ Willem \klop}

\newcommand\kuechlin        {K\ue ch\-lin}
\newcommand\kuechlinname    {\wolfgang\ \kuechlin}
\newcommand\kuehler         {K\ue h\-ler}
\newcommand\kuehlername     {\ulrich\ \kuehler}

\newcommand\levyname        {\jean-\jacques\ L\'evy}

\newcommand\loria           {Lor\'\i a-S\'aenz}
\newcommand\lorianame       {\carlos\ A. \loria}

\newcommand\lunde           {Lunde}
\newcommand\lundename       {\ruediger\ \lunde}

\newcommand\madlener        {Mad\-lener}
\newcommand\madlenername    {\klaus\ \madlener}

\newcommand\middeldorp      {Middel\-dorp}
\newcommand\middeldorpname  {Aart \middeldorp}

\newcommand\musser          {Musser}
\newcommand\mussername      {\david\ R. \musser}
\newcommand\narendran       {Narendran}
\newcommand\narendranname   {Paliath \narendran}

\newcommand\newman          {Newman}

\newcommand\newmanlemma     {\newman\ Lemma}

\newcommand\okadaname       {Mitsuhiro Okada}

\newcommand\oostrom         {{\namefont Oostrom}}
\newcommand\oostromname     {{\namefont Vincent van \oostrom}}

\newcommand\rosser          {Rosser}

\newcommand\sivakumarname   {G. Sivakumar}

\newcommand\suzuki          {Suzuki}
\newcommand\suzukiname      {Taro \suzuki}

\newcommand\toyama          {{\namefont Toyama}}
\newcommand\toyamaname      {{\namefont Yoshihito \toyama}}

\newcommand\walther         {Walther} 
\newcommand\walthername     {\christoph\ \walther}

\newcommand\wirth           {{\namefont Wirth}}
\newcommand\wirthname       {{\namefont\claus-\peter\ \wirth}}

\newcommand\FB   {FB}
\newcommand\FBinf{Fachbereich Informatik}
\newcommand\FBinfshort{\FB\ Informatik}

\newcommand\qedhelp[1]{Q.e.d.~({#1})}
\newcommand\getittotheright[1]  
{\hfill\mbox{}\penalty 100\mbox{\ \,}\nolinebreak
\hfill\hfill\hfill\nolinebreak\mbox{#1}\ignorespaces}
\newcommand\Qed      [1]{\underline{\qedhelp{#1}}}
\newcommand\Qeddouble[1]{\underline{\underline{\qedhelp{#1}}}}
\newcommand\Qedtriple[1]{\underline{\underline{\underline{\qedhelp{#1}}}}}
\newcommand\Qedbf    [1]{\mbox{\bf\qedhelp{#1}}}
\newcommand\QED      [1]{\getittotheright{\Qed      {#1}}}
\newcommand\QEDdouble[1]{\getittotheright{\Qeddouble{#1}}}
\newcommand\QEDtriple[1]{\getittotheright{\Qedtriple{#1}}}
\newcommand\QEDbf    [1]{\getittotheright{\Qedbf    {#1}}}

\newcommand\uni  {Uni\-ver\-si\-t\ae t}
\newcommand\Univ {Univ.}

\newcommand\Cf   {Cf.}
\newcommand\cf   {cf.}

\newcommand\CS   {Computer \Sci}

\newcommand\eds  {eds.}
\newcommand\eg   {e.g.}
\newcommand\Eg   {E.g.}

\newcommand\esp  {esp.}
\newcommand\etalabbrev{\&al.}
\newcommand\etc  {\&c.}

\newcommand\signatureenlarge     {enrich}

\newcommand\signatureenlarged    {enriched}

\newcommand\ie   {i.e.}

\newcommand\uiff {\ iff\ }
\newcommand\udiff{\ if\ }

\newcommand\pp   {pp.}
\newcommand\PP[2]{\pp\,\ignorespaces#1--\ignorespaces#2}
\newcommand\resp {resp.}

\newcommand\sect {\myparagraphsymbol} 
\newcommand\Sci  {Sci.}

\newcommand\unary{singulary} % Following Quine's Mathematical Logic, p. 13:
\newcommand\wrog {w.l.o.g.} % My favorite misspelling. Moreover \wlog is
\newcommand\Wrog {W.l.o.g.} % already a latex command
\newcommand\wrt  {w.r.t.}
\newcommand\Wrt  {W.r.t.}

\newcommand\Examplename{Ex\-am\-ple}

\newcommand\nthpositioner[2]
{#1\raisebox{0.52ex}{\scriptsize\hspace{0.07em}#2}}
\newcommand\nth[1]{\nthtinypositioner{#1}{\nthstring{#1}}}
\newcommand\nthtinypositioner[2]{#1\raisebox{0.52ex}{\tiny\hspace{0.07em}#2}}

\newcommand\mthpositioner[2]
{\math{#1}\raisebox{0.52ex}{\scriptsize\hspace{0.07em}#2}}
\newcommand\modulointocountzero[2]
{\count1=#1
 \count2=#2
 \count0=\count1
 \divide  \count0 by \count2
 \multiply\count0 by-\count2
 \advance \count0 by \count1}
\newcommand\absolutevalueintocountzero[1]
{\count0=#1
 \ifnum\count0<0\multiply\count0 by -1\fi}
\newcommand\nthstring[1]
{\def\myargone{#1}\ifcat a\myargone th\else\nthstringnochar{#1}\fi}
\newcommand\nthstringnochar[1]
{\absolutevalueintocountzero{#1}%
 \modulointocountzero{\count0}{100} %we need the blank here!
 \ifnum\count0>9\ifnum\count0<20 th\else\stupidnthstring\fi
                                   \else\stupidnthstring\fi}
\newcommand\stupidnthstring
{\modulointocountzero{\count0}{10}
 \ifnum\count0=1 \hskip-0.2em st\else
 \ifnum\count0=2 nd\else
 \ifnum\count0=3 rd\else 
                 th\fi\fi\fi}

\newcommand\writeascents
[1]{\count4=#1
\ifnum\count4<0 
-\multiply\count4 by -1\fi
\modulointocountzero{\count4}{10}
\divide\count4 by 10
\count3=\the\count0
\modulointocountzero{\count4}{10}
\divide\count4 by 10
\the\count4
.\the\count0
\the\count3
}

\newcommand\frenchnthstring[1]
{\def\myargone{#1}\ifcat a\myargone th\else\frenchnthstringnochar{#1}\fi}
\newcommand\frenchnthstringnochar[1]
{\absolutevalueintocountzero{#1}%
 \modulointocountzero{\count0}{100} %we need the blank here!
 \ifnum\count0>9\ifnum\count0<20 th\else\frenchstupidnthstring\fi
                                   \else\frenchstupidnthstring\fi}
\newcommand\frenchstupidnthstring
{\modulointocountzero{\count0}{10}
 \ifnum\count0=1 \hskip-0.2em re\else
 \ifnum\count0=2 me\else
 \ifnum\count0=3 rd\else 
                 th\fi\fi\fi}

\newcommand\CLAM      {{\rm CL\kern-.36em\raise.39ex\hbox{\sc a}\kern-.15emM}}

\newcommand\TEXMACS   {{\sc T\kern-.1667em\lower.5ex\hbox{E}\kern-.125emX\kern-.1em\lower.5ex\hbox{\textsc{m\kern-.05ema\kern-.125emc\kern-.05ems}}}}

\newcommand\KL             {Kai\-ser\esi lau\-tern}

\newcommand\uniKL{\uni\ \KL}

\newcommand\uniKLshort{\Univ\ \KL}

\newcommand\addressuniKL{\FBinf, \uniKL}
\newcommand\addressuniKLshort{\FBinfshort, \uniKLshort}

\newcommand\LNCSvol[1]
{\mbox{LNCS}\,\ignorespaces#1}

\newcommand\LNAIvol[1]
{\mbox{LNAI}\,\ignorespaces#1}

\newcommand\academicpress{Academic Press (\elsevier)}
\newcommand\ACM{ACM}
\newcommand\acmpress{\ACM\ Press}

\newcommand\clarendonpress{Clarendon Press%, \Oxford
}

\newcommand\elsevier{Elsevier}

\newcommand\springerverlag{Sprin\-ger}

\newcommand\sekireportname{SEKI-Report}

\newcommand\sekireportnoaddress[2]{\sekireportname\ \mbox{SR--#1--#2}}
\newcommand\sekireport   [2]{\sekireportnoaddress{#1}{#2}, \addressuniKLshort}

\newcommand\sekiworkingpapername{SEKI-Working-Paper}

\newcommand\sekiworkingpapernoaddress[2]
{\sekiworkingpapername\ \mbox{SWP--#1--#2}}
\newcommand\sekiworkingpaper[2]{\sekiworkingpapernoaddress
{#1}{#2}, \addressuniKLshort}

\newcommand\lncsconf[6]{\nth{#2}\,#1\,#3, #4, \PP{#5}{#6}, \springerverlag}

\newcommand\fourthAlgLogPninetyfour
{\lncsconf{Algebraic and Logic Programming}{4}{1994}{\LNCSvol{850}}}

\newcommand\CADEshort{CADE}

\newcommand\twelvethCADEninetyfour               
{\lncsconf\CADEshort{12}{1994}{\LNAIvol{ 814}}}

\newcommand\CTRSshort{CTRS}

\newcommand\firstCTRSeightyseven   {\lncsconf\CTRSshort{1}{1987}{\LNCSvol{308}}}

\newcommand\thirdCTRSninetytwo     {\lncsconf\CTRSshort{3}{1992}{\LNCSvol{656}}}

\newcommand\fourthCTRSninetyfour   {\lncsconf\CTRSshort{4}{1994}{\LNCSvol{968}}}

\newcommand\fifthLPARninetyfour {\lncsconf{LPAR} {5}{1994}{\LNAIvol {822}}}

\newcommand\fourthRTAninetyone       {\lncsconf{RTA}{ 4}{1991}{\LNCSvol{ 488}}}

\newcommand\sixthRTAninetyfive       {\lncsconf{RTA}{ 6}{1995}{\LNCSvol{ 914}}}

\newcommand\eleventhSTACSninetyfour{\lncsconf{STACS}{11}{1994}{\LNCSvol{775}}}

\newcommand\TAPSOFTninetythree[2]
{TAPSOFT 1993, \LNCSvol{668}, \PP{#1}{#2}, \springerverlag}

\newcommand\newspaperreference[5]
{\def\nameofjournalpress{#2}#1, #4 #5, #3\if?\nameofjournalpress
 \else, #2\fi}

\newcommand\dateinjournal[1]{}

\newcommand\journalreference[6]
{\def\nameofjournalpress{#2}#1\nolinebreak\hskip.2em%
 \dateinjournal{(#3) }{\mbox{\bf #4}}, \PP{#5}{#6}\if?\nameofjournalpress
 \else, #2\fi}

\newcommand\journalreferenceprintyear[6]
{\def\nameofjournalpress{#2}#1 
 {(#3) }{\bf #4}, \PP{#5}{#6}\if?\nameofjournalpress
 \else, #2\fi}

\newcommand\journalreferenceprintyearaspartofnumber[6]
{\def\nameofjournalpress{#2}#1 
 {#4/#3}, \PP{#5}{#6}\if?\nameofjournalpress
 \else, #2\fi}

\newcommand\informationandcomputation
{\journalreference{Information and Computation}\academicpress}

\newcommand\informationandcontrol
{\journalreference{Information and Control}?}

\newcommand\aaeccname
{J. Applicable Algebra in Engineering, Communication and Computing (AAECC)}
\newcommand\aaecc
{\journalreference\aaeccname\springerverlag}

\newcommand\jacm
{\journalreference{J. \ACM}\acmpress}

\newcommand\jcssname{J. Computer and System \Sci}
\newcommand\jcss
{\journalreference{\jcssname}\academicpress}

\newcommand\jscname
{J. Symbolic Computation}

\newcommand\jscprintyear
{\journalreferenceprintyear{\jscname}\academicpress}

\newcommand\tcsname{Theoretical \CS}
\newcommand\tcsjournal
{\journalreference\tcsname\elsevier}
\newcommand\tcsjournalprintyear
{\journalreferenceprintyear\tcsname\elsevier}

\input dina4.sty
\input headerhot

\newcommand\Proofof{Proof of}
\renewenvironment{proof}[1]{\yestop\begin{sloppypar}\noindent
{\bf\Proofof\ {#1}}\\}{\end{sloppypar}}
\renewenvironment{proofqed}[1]
{\begin{sloppypar}\def\fooqed{#1}\noindent{\bf\Proofof\ \fooqed}}
{\QEDbf\fooqed\end{sloppypar}}
\renewenvironment{proofparsep}[1]{\parindent=0pt\begin
{sloppypar}\noindent{\bf\Proofof\ {#1}}\nopagebreak\par}
{\end{sloppypar}}
\renewenvironment{proofparsepqed}[1]{\parindent=0pt\begin
{sloppypar}\def\fooqed{#1}\noindent{\bf\Proofof\ \fooqed}\nopagebreak\par}
{\nopagebreak\QEDbf\fooqed\end{sloppypar}}
\renewenvironment{proofshort}[1]{\begin{sloppypar}\noindent
{\bf\Proofof\ {#1}:}}{\end{sloppypar}}

\newcommand\arr[1]{\mbox{$\begin{array}#1\end{array}$}}

{\end{enumerate}\renewcommand{\theenumi}{\arabic{enumi}}}

\mathcommand\myfootnotemark[1]{^{#1}}
\newcommand\repeatfootnotemark[1]{\myfootnotemark{\ref{#1}}}

\newcommand\repname{{\rm set}}
\mathapplycommand\rep\repname
\mathcommand\repr[1]{{\repname[{#1}]}}
\mathcommand\msa{\langle}
\mathcommand\mse{\rangle}
\mathcommand\msu{\,\sqcup\,}
\mathcommand\msin{{\rm\;in\;}}
\mathcommand\mssetminus{\setminus\!\!\setminus}
\mathcommand\tightmssubseteq{\sqsubseteq}
\mathcommand\mssubseteq{\ \tightmssubseteq\ }
\mathcommand\approxapprox{\approx\:\!\!\approx}
\mathcommand\quasilquasil{\,\lesssim\!\lesssim\,}
\mathcommand\quasibquasib{\,\gtrsim\!\gtrsim\,}
\mathcommand\fmul[1]{{\rm FMul}(#1)}
\mathcommand\smul[1]{{\rm SMul}(#1)}
\mathcommand\multisetwith [2]{\msa\ {#1}\ |\ {#2}\ \mse}
\mathcommand\multisetwithq[3]{\msa\ {#2}\ |_{#1}\ {#3}\ \mse}
\newcommand\superseteq     {\supseteq}

\mathcommand\quasilhd
{\mathop{\mbox{\raisebox{0.31ex}{\math\lhd}\hspace{-0.75em}\raisebox
{-0.6ex}{\math\sim}}}}
\newcommand\quasirhd{\mbox{\raisebox{0.31ex}{$\rhd$}\hspace{-0.75em}\raisebox{-0.6ex}{$\sim$}}}

\mathcommand\rhdrhd{\rhd$\hspace{-0.35em}$\rhd}
\mathcommand\lhdlhd{\lhd$\hspace{-0.21em}$\lhd}

\mathcommand\quasilhdquasilhd{\quasilhd$\hspace{-0.13em}$\quasilhd}

\newcommand\hiddensubSS{_{_{\rm SS}}}
\mathcommand\antisubsum     {\rhd\hiddensubSS}
\mathcommand\notantisubsum  {\ntriangleright\hiddensubSS}
\mathcommand\subsum         {\lhd\hiddensubSS}
\mathcommand\notsubsum      {\ntriangleleft\hiddensubSS}
\mathcommand\antisubsumeq   {\trianglerighteq\hiddensubSS}
\mathcommand\subsumeq       {\trianglelefteq\hiddensubSS}
\mathcommand\quasisubsum    {\,\quasilhd\raisebox{0.1ex}{$\hiddensubSS$}}
\mathcommand\antiquasisubsum{\,\quasirhd\raisebox{0.1ex}{$\hiddensubSS$}}
\mathcommand\quasiquasisubsum{\quasisubsum\!\!\quasisubsum}
\mathcommand\antiquasiquasisubsum{\antiquasisubsum\!\!\!\antiquasisubsum}

\newcommand\hiddensubH{_{_{\rm H}}}
\newcommand\hiddensubCONS{_{_\CONS}}
\mathcommand\hql   {\,\lesssim\hiddensubH}
\mathcommand\consql{\,\lesssim\hiddensubCONS}
\mathcommand\hl    {\,<       \hiddensubH}
\mathcommand\hleq  {\,\leq    \hiddensubH}
\mathcommand\consl {\,<       \hiddensubCONS}
\mathcommand\conseq{\,\approx \hiddensubCONS}

\newcommand\cons {{\rm cons}}
\mathcommand\sigconsV{\sig/\cons/\V}
\mathcommand\sigconsR{\sig/\cons/\R}
\mathcommand\primesigconsV{\sig'\!/\cons'\!/\V'}
\mathcommand\primesigconsR{\sig'\!/\cons'\!/\R'}
\newcommand\dunno{{\rm sc}}
\newcommand\DUNNO{{\rm SC}}
\mathcommand\SIGCONS   {\{\SIG,\CONS\}}
\mathcommand\sigsortstimes{\SIGCONS\tight\times\sigsorts}

\mathcommand\TERMSSYM
{\mathcal{T\hspace{-.20em}E\hspace{-.08em}R\hspace{-.16em}M\hspace{-.10em}S}}
\mathapplycommand\condterms{\TERMSSYM}
\mathcommand\kurzregel{((l,r),C)}
\mathcommand\kurzregelprime{((l',r'),C')}
\mathcommand\kurzregelindex[1]{((l_{#1},r_{#1}),C_{#1})}
\mathapplycommand\lhs{\rm lhs}

\mathcommand\red{\redsimple} %conflicts with ps tricks

\mathcommand\lemms{L}
\mathcommand\hypos{H}
\mathcommand\goals{G}
\mathcommand\lemmsprime{\lemms'}
\mathcommand\hyposprime{\hypos'}
\mathcommand\goalsprime{\goals'}
\mathcommand\lemmsprimeprime{\lemms''}
\mathcommand\hyposprimeprime{\hypos''}
\mathcommand\goalsprimeprime{\goals''}
\mathcommand\oldtriple            {(\lemms   ,\hypos   ,\goals  )}
\mathcommand\inittriple        {(\emptyset,\emptyset,\goals  )}
\mathcommand\triplehelp[1]     {(\lemms#1,\hypos#1 ,\goals#1)}
\mathcommand\tripleprime       {\triplehelp'}
\mathcommand\triplenogoalsprime{(\lemmsprime,\hyposprime,\emptyset  )}
\mathcommand\tripleprimeprime  {\triplehelp{''}}
\mathcommand\tripleindex[1]    {\triplehelp{_{#1}}}

\mathcommand\constcong[1]{\,\,\sim_{\!_{#1}}\,}

\mathapplycommand\avail{\rm\Av ail}

\def\emph#1{\/ {\itshape#1}\/}

\input headertheorem
\mathcommand\notconflu{\mathchoice
             {{\hskip1.5pt\nmid\hskip-4.697545pt\downarrow}}%   display-style
             {{\hskip1.5pt\nmid\hskip-4.65pt\downarrow}}   %   text-style
             {{\hskip1pt\nmid\hskip-3.494pt\downarrow\hskip1pt}}  
             {{\hskip1pt\nmid\hskip-3.01pt\downarrow\hskip0.5pt}}   
}
\mathcommand\redpara{\mathchoice
           {{\redsimple\hskip-16pt  \shortparallel}\hskip8.5pt}%display-style
           {{\redsimple\hskip-16pt  \shortparallel}\hskip8.5pt}%text-style
           {{\redsimple\hskip-8.5pt \shortparallel}\hskip6pt}%scriptstyle
           {{\redsimple\hskip-7.5pt \shortparallel}\hskip5pt}%scriptscriptstyle
}
\mathcommand\antiredpara{\mathchoice
           {{\antired\hskip-14.6pt  \shortparallel}\hskip7pt}%display-style
           {{\antired\hskip-14.6pt  \shortparallel}\hskip7pt}%text-style
           {{\antired\hskip-8.pt \shortparallel}\hskip5pt}%scriptstyle
           {{\antired\hskip-7.pt \shortparallel}\hskip5pt}%scriptscriptstyle
}
\mathcommand\revpara{\mathchoice
           {{\redsimple\hskip-16pt  \infty}\hskip4.8pt}%display-style
           {{\redsimple\hskip-16pt  \infty}\hskip4.8pt}%text-style
           {{\redsimple\hskip-11.5pt\infty}\hskip4pt}%scriptstyle
           {{\redsimple\hskip-9.9pt \infty}\hskip3pt}%scriptscriptstyle
}
\mathcommand\antirevpara{\mathchoice
           {{\antired\hskip-15.4pt\infty}\hskip4pt}%display-style
           {{\antired\hskip-15.4pt\infty}\hskip4pt}%text-style
           {{\antired\hskip-10.8pt\infty}\hskip3pt}%scriptstyle
           {{\antired\hskip-9.5pt \infty}\hskip3pt}%scriptscriptstyle
}
\mathcommand\simpara{\mathchoice%this needs more fixing
           {{\redsimple\hskip-13pt  \circ}\hskip7pt}%display-style
           {{\redsimple\hskip-13pt  \circ}\hskip7pt}%text-style
           {{\redsimple\hskip-11.5pt\circ}\hskip4pt}%scriptstyle
           {{\redsimple\hskip-9.9pt \circ}\hskip3pt}%scriptscriptstyle
}

\mathcommand\ident[1]{\mathsf{#1}}
\newcommand\plussymbol  {\ident{+}}
\newcommand\minussymbol {\ident{-}}
\newcommand\dividesymbol{\ident{/}}
\newcommand\timessymbol {\ident{*}}

\newcommand\nat     {\ident{nat}}

\newcommand\lists   {\ident{list}}
\newcommand\set     {\ident{set}}

\newcommand\bool    {\ident{bool}}
\newcommand\intsort {\ident{int}}

\newcommand\naturalssymbol{\ident{naturals}}
\newcommand\gensymsymbol{\ident{gensym}}
\mathcommand\mbpsymbol{\ident{m\hspace{-0.055em}b\hspace{-0.045em}p}}

\newcommand\csymbol     {\ident c}
\newcommand\esymbol     {\ident e}
\newcommand\fsymbol     {\ident f}
\newcommand\gsymbol     {\ident g}
\newcommand\hsymbol     {\ident h}
\newcommand\ksymbol     {\ident k}
\newcommand\psymbol     {\ident p}
\newcommand\ssymbol     {\ident s}
\newcommand\Everysymbol {\ident{Every}}
\newcommand\Permsymbol {\ident{Perm}}
\newcommand\RExistssymbol{\ident{Rexists}}
\newcommand\invertsymbol{\ident{invert}}
\newcommand\invsymbol{\ident{inv}}
\newcommand\abssymbol   {\ident{abs}}
\newcommand\cnssymbol   {\ident{cons}}
\mathcommand\cnsindexsymbol[1]{\ident{cons}_{#1}}

\newcommand\lengthsymbol{\ident{length}}
\newcommand\dlsymbol    {\ident{dl}}
\newcommand\dloncesymbol{\ident{delonce}}
\newcommand\rcsymbol    {\ident{rc}}
\newcommand\brsymbol    {\ident{br}}
\newcommand\revtailsymbol{\ident{revtail}}
\newcommand\revsymbol{\ident{rev}}
\newcommand\appendsymbol {\ident{append}}
\newcommand\zeropredicatesymbol{\ident{zerop}}
\newcommand\eqsymbol        {\ident{eq}}
\newcommand\ifthensymbol    {\mbox{\ident{If{}Then}}}
\newcommand\ifthenelsesymbol{\mbox{\ident{If{}ThenElse}}}
\mathcommand\eqindexsymbol        [1]{\eqsymbol        _{#1}}
\mathcommand\ifthenindexsymbol    [1]{\ifthensymbol    _{#1}}
\mathcommand\ifthenelseindexsymbol[1]{\ifthenelsesymbol_{#1}}
\newcommand\orsymbol    {\ident{or}}
\newcommand\andsymbol   {\ident{and}}
\newcommand\leqsymbol   {\ident{leq}}
\newcommand\lessymbol   {\ident{less}}
\newcommand\lexsymbol   {\ident{lex}}
\newcommand\acksymbol   {\ident{ack}}
\newcommand\switchsymbol{\ident{switch}}
\newcommand\swatchsymbol{\ident{swatch}}
\newcommand\diveinssymbol{\ident{div1}}
\newcommand\divzweisymbol{\ident{div2}}
\newcommand\divrestsymbol{\ident{divrest}}
\newcommand\diveinstailsymbol{\ident{div1tail}}
\newcommand\divzweitailsymbol{\ident{div2tail}}

\newcommand\turingmachinesymbol{\ident T}
\newcommand\terminatespsymbol  {\ident{terminatesp}}
\newcommand\statesymbol        {\ident{state}}
\newcommand\cmdsymbol          {\ident{cmd}}
\newcommand\nthsymbol          {\ident{nth}}
\newcommand\doublesymbol       {\ident{double}}

\newcommand\ppsymbol           {\ident{p}}
\newcommand\qpsymbol           {\ident{q}}
\newcommand\Epsymbol           {\ident{E}}
\newcommand\Ppsymbol           {\ident{P}}
\newcommand\Qpsymbol           {\ident{Q}}
\newcommand\Marriessymbol      {\ident{Marries}}
\newcommand\Lovessymbol        {\ident{Loves}}
\newcommand\StolenBysymbol     {\ident{StolenBy}}
\newcommand\Humansymbol        {\ident{Human}}
\newcommand\Evensymbol         {\ident{Even}}
\newcommand\Oddsymbol          {\ident{Odd}}
\newcommand\Primesymbol        {\ident{Prime}}
\newcommand\EveryPairsymbol   {\ident{EveryPair}}
\newcommand\Givesymbol         {\ident{Give}}
\newcommand\Fathersymbol       {\ident{Father}}
\newcommand\Elephantpsymbol    {\ident{Elephant}}
\newcommand\Flowerpsymbol    {\ident{Flower}}
\newcommand\Germanpsymbol      {\ident{German}}
\newcommand\Bicyclepsymbol     {\ident{Bicycle}}
\newcommand\Hugepsymbol        {\ident{Huge}}
\newcommand\Animalpsymbol      {\ident{Animal}}
\newcommand\Malepsymbol        {\ident{Male}}
\newcommand\Boypsymbol        {\ident{Boy}}
\newcommand\Girlpsymbol        {\ident{Girl}}
\newcommand\Femalepsymbol      {\ident{Female}}
\newcommand\Roundpsymbol       {\ident{Round}}
\newcommand\Quadrangularpsymbol{\ident{Quadrangular}}
\newcommand\Metpsymbol         {\ident{Met}}
\newcommand\Bishopsymbol       {\ident{Bishop}}
\newcommand\mindexsymbol[1]{\existsvari w{#1}}

\newcommand\nonnegpsymbol      {\ident{nonnegp}}
\newcommand\wellsymbol         {\ident{well}}
\newcommand\welltailsymbol     {\ident{welltail}}
\newcommand\varsymbol          {\ident{var}}
\newcommand\aritysymbol        {\ident{arity}}

\newcommand\whilesymbol        {\ident{while}}

\newcommand\nullsymbol         {\ident{null}}
\newcommand\hdsymbol           {\ident{hd}}
\newcommand\tlsymbol           {\ident{tl}}
\newcommand\insymbol           {\ident{in}}
\newcommand\applysymbol        {\ident{app}}
\newcommand\termsymbol         {\ident{term}}
\mathcommand\tightim{\longrightarrow}
\mathcommand\im{\ \tightim\ }
\mathcommand\rs{\:\rulesugar\:\:}
\mathcommand\rulesugar{\longleftarrow}

\mathcommand\doublepp[1]      {\doublesymbol   \beginargs{#1}\allargs}
\mathcommand\aritypp[1]      {\aritysymbol   \beginargs{#1}\allargs}
\mathcommand\lengthpp[1]      {\lengthsymbol   \beginargs{#1}\allargs}
\mathcommand\wellpp[1]      {\wellsymbol   \beginargs{#1}\allargs}
\mathcommand\welltailpp[1]      {\welltailsymbol   \beginargs{#1}\allargs}
\mathcommand\varpp[1]      {\varsymbol   \beginargs{#1}\allargs}
\mathcommand\divrestpp[2]    {\divrestsymbol\beginargs{#1}\separgs{#2}\allargs}
\mathcommand\diveinspp[2]    {\diveinssymbol\beginargs{#1}\separgs{#2}\allargs}
\mathcommand\divzweipp[3]    {\divzweisymbol\beginargs{#1}\separgs{#2}
\separgs{#3}\allargs}
\mathcommand\diveinstailpp[4]    {\diveinstailsymbol\beginargs{#1}\separgs{#2}
\separgs{#3}\separgs{#4}\allargs}
\mathcommand\divzweitailpp[6]    {\divzweitailsymbol\beginargs{#1}\separgs{#2}
\separgs{#3}\separgs{#4}\separgs{#5}\separgs{#6}\allargs}
\mathcommand\mbppp[2]         {\mbpsymbol   \beginargs{#1}\separgs{#2}\allargs}
\mathcommand\revpp[1]     
{\revsymbol\beginargs{#1}\allargs}
\mathcommand\revppi[2]     
{\revsymbol^{#1}\beginargs{#2}\allargs}
\mathcommand\revtailpp[2]     
{\revtailsymbol\beginargs{#1}\separgs{#2}\allargs}
\mathcommand\revtailppi[3]
{\revtailsymbol^{#1}\beginargs{#2}\separgs{#3}\allargs}
\mathcommand\Permpp[2]     
{\Permsymbol\beginargs{#1}\separgs{#2}\allargs}
\mathcommand\Permppi[3]
{\Permsymbol^{#1}\beginargs{#2}\separgs{#3}\allargs}
\mathcommand\appendpp[2]      
{\appendsymbol \beginargs{#1}\separgs{#2}\allargs}
\mathcommand\appendppi[3]      
{\appendsymbol^{#1}\beginargs{#2}\separgs{#3}\allargs}
\mathcommand\Everypp[2]      
{\Everysymbol \beginargs{#1}\separgs{#2}\allargs}
\mathcommand\RExistspp[1]      
{\RExistssymbol \beginargs{#1}\allargs}
\mathcommand\appendlongpp[2]      
{\appendsymbol\left(\begin{array}{@{}l@{}}{#1}\separgs\\{#2}\end{array}\right)}
\mathcommand\cnspp[2]         {\cnssymbol   \beginargs{#1}\separgs{#2}\allargs}
\mathcommand\cnsppi[3]       {\cnssymbol^{#1}\beginargs{#2}\separgs{#3}\allargs}
\mathcommand\cnsindexpp[3]
{\cnsindexsymbol{#1}\beginargs{#2}\separgs{#3}\allargs}
\mathcommand\dlpp[2]          {\dlsymbol    \beginargs{#1}\separgs{#2}\allargs}
\mathcommand\dloncepp[2]      {\dloncesymbol\beginargs{#1}\separgs{#2}\allargs}
\mathcommand\dlonceppi[3]{\dloncesymbol^{#1}\beginargs{#2}\separgs{#3}\allargs}
\mathcommand\rcpp[2]          {\rcsymbol    \beginargs{#1}\separgs{#2}\allargs}
\mathcommand\brpp[2]          {\brsymbol    \beginargs{#1}\separgs{#2}\allargs}
\mathcommand\orpp[2]          {\orsymbol    \beginargs{#1}\separgs{#2}\allargs}
\mathcommand\andpp[2]         {\andsymbol   \beginargs{#1}\separgs{#2}\allargs}
\mathcommand\shortcnspp[2]    {\csymbol     \beginargs{#1}\separgs{#2}\allargs}
\mathcommand\tightshortcnspp[2]
{\csymbol\beginargs{#1}\tightsepargs{#2}\allargs}
\mathcommand\spp[1]           {\ssymbol     \beginargs{#1}\allargs}
\mathcommand\sppiterated[2]   {\ssymbol^{#1}\beginargs{#2}\allargs}
\mathcommand\ppp[1]           {\psymbol     \beginargs{#1}\allargs}
\mathcommand\pppiterated[2]   {\psymbol^{#1}\beginargs{#2}\allargs}
\mathcommand\zeropp           {\ident 0}
\mathcommand\Julietpp         {\ident{Juliet}}
\mathcommand\Romeopp          {\ident{Romeo}}
\mathcommand\Ipp              {\ident I}
\mathcommand\onepp            {\ident1}
\mathcommand\twopp            {\ident2}
\mathcommand\threepp          {\ident3}
\mathcommand\invertpp[1]      {\invertsymbol\beginargs{#1}\allargs}
\mathcommand\invpp[1]         {\invsymbol\beginargs{#1}\allargs}
\mathcommand\abspp[1]         {\abssymbol\beginargs{#1}\allargs}
\mathcommand\naturalspp[1]    {\naturalssymbol\beginargs{#1}\allargs}
\mathcommand\gensympp[1]      {\gensymsymbol\beginargs{#1}\allargs}
\mathcommand\nilpp            {\ident{nil}}
\mathcommand\falsepp          {\ident{false}}
\mathcommand\truepp           {\ident{true}}
\mathcommand\FALSEpp          {\ident{FALSE}}
\mathcommand\TRUEpp           {\ident{TRUE}}
\mathcommand\weirdppp         {\ident{weirdp}}
\mathcommand\ambigppp         {\ident{ambigp}}
\mathcommand\zeropredicatepp[1]{\zeropredicatesymbol\beginargs{#1}\allargs}
\mathcommand\cppeins       [1]{\csymbol     \beginargs{#1}\allargs}
\mathcommand\cppzwei       [2]{\csymbol\beginargs{#1}\separgs{#2}\allargs}
\mathcommand\eppeins       [1]{\esymbol     \beginargs{#1}\allargs}
\mathcommand\fppeins       [1]{\fsymbol     \beginargs{#1}\allargs}
\mathcommand\fppeinsindex  [2]{\fsymbol_{#1}\beginargs{#2}\allargs}
\mathcommand\fppeinsiterated[2]{\fsymbol^{#1}\beginargs{#2}\allargs}
\mathcommand\gppeins       [1]{\gsymbol     \beginargs{#1}\allargs}
\mathcommand\gppzwei       [2]{\gsymbol     \beginargs{#1}\separgs{#2}\allargs}
\mathcommand\hppeins       [1]{\hsymbol     \beginargs{#1}\allargs}
\mathcommand\kppeins       [1]{\ksymbol     \beginargs{#1}\allargs}
\mathcommand\appzero          {\ident a}
\mathcommand\bppzero          {\ident b}
\mathcommand\cppzero          {\ident c}
\mathcommand\dppzero          {\ident d}
\mathcommand\eppzero          {\ident e}
\mathcommand\eqindexpp[3]{\eqindexsymbol{#1}\beginargs{#2}\separgs{#3}\allargs}
\mathcommand\ifthenindexpp
[3]{\ifthenindexsymbol{#1}\beginargs{#2}\separgs{#3}\allargs}
\mathcommand\ifthenelseindexpp
[4]{\ifthenelseindexsymbol{#1}\beginargs{#2}\separgs{#3}\separgs{#4}\allargs}
\mathcommand\eqpp[2]{\eqsymbol\beginargs{#1}\separgs{#2}\allargs}
\mathcommand\leqpp[2]{\leqsymbol\beginargs{#1}\separgs{#2}\allargs}
\mathcommand\lespp[2]{\lessymbol\beginargs{#1}\separgs{#2}\allargs}
\mathcommand\lexpp[3]{\lexsymbol\beginargs{#1}\separgs{#2}\separgs{#3}\allargs}
\mathcommand\ackpp[2]{\acksymbol\beginargs{#1}\separgs{#2}\allargs}
\mathcommand\switchpp[1]{\switchsymbol\beginargs{#1}\allargs}
\mathcommand\swatchpp[1]{\swatchsymbol\beginargs{#1}\allargs}
\mathcommand\whilepp[2]{\whilesymbol\beginargs{#1}\separgs{#2}\allargs}
\mathcommand\nullpp[1]{\nullsymbol\beginargs{#1}\allargs}
\mathcommand\nullppiterated[2]{\nullsymbol^{#1}\beginargs{#2}\allargs}
\mathcommand\hdpp[1]{\hdsymbol\beginargs{#1}\allargs}
\mathcommand\hdppiterated[2]{\hdsymbol^{#1}\beginargs{#2}\allargs}
\mathcommand\tlpp[1]{\tlsymbol\beginargs{#1}\allargs}
\mathcommand\tlppiterated[2]{\tlsymbol^{#1}\beginargs{#2}\allargs}
\mathcommand\inpp[2]{\insymbol\beginargs{#1}\separgs{#2}\allargs}
\mathcommand\inppiterated[3]{\insymbol^{#1}\beginargs{#2}\separgs{#3}\allargs}
\mathcommand\applypp[2]{\applysymbol\beginargs{#1}\separgs{#2}\allargs}
\mathcommand\applyppiterated
[3]{\applysymbol^{#1}\beginargs{#2}\separgs{#3}\allargs}
\mathcommand\termpp[2]{\termsymbol\beginargs{#1}\separgs{#2}\allargs}
\mathcommand\setpp[1]{\set\beginargs{#1}\allargs}

\mathcommand\Tpp[6]{\turingmachinesymbol\beginargs{#1}\separgs{#2}\separgs
{#3}\separgs{#4}\separgs{#5}\separgs{#6}\allargs}
\mathcommand\Tppseven[7]{\turingmachinesymbol\beginargs{#1}\separgs{#2}\separgs
{#3}\separgs{#4}\separgs{#5}\separgs{#6}\separgs{#7}\allargs}
\mathcommand\foreverppp[6]{\ident{foreverp}\beginargs{#1}\separgs{#2}\separgs
{#3}\separgs{#4}\separgs{#5}\separgs{#6}\allargs}
\mathcommand\terminatesppp[6]{\terminatespsymbol\beginargs{#1}\separgs
{#2}\separgs{#3}\separgs{#4}\separgs{#5}\separgs{#6}\allargs}
\mathcommand\terminatespppone[1]{\terminatespsymbol \beginargs{#1}\allargs}
\mathcommand\statepp
[3]{\statesymbol\beginargs{#1}\separgs{#2}\separgs{#3}\allargs}
\mathcommand\tightstatepp
[3]{\statesymbol\beginargs{#1}\tightsepargs{#2}\tightsepargs{#3}\allargs}
\mathcommand\cmdpp
[3]{\cmdsymbol  \beginargs{#1}\separgs{#2}\separgs{#3}\allargs}
\mathcommand\tightcmdpp
[3]{\cmdsymbol  \beginargs{#1}\tightsepargs{#2}\tightsepargs{#3}\allargs}
\mathcommand\stoppp           {\ident{stop}}
\mathcommand\leftpp           {\ident{left}}
\mathcommand\rightpp          {\ident{right}}
\mathcommand\nthpp         [2]{\nthsymbol  \beginargs{#1}\separgs{#2}\allargs}
\mathcommand\pppp          [1]{\ppsymbol\beginargs{#1}            \allargs}
\mathcommand\qppp          [2]{\qpsymbol\beginargs{#1}\separgs{#2}\allargs}
\mathcommand\Eppp          [1]{\Epsymbol\beginargs{#1}            \allargs}
\mathcommand\Epppzwei      [2]{\Epsymbol\beginargs{#1}\separgs{#2}\allargs}
\mathcommand\Pppp          [1]{\Ppsymbol\beginargs{#1}            \allargs}
\mathcommand\Ppppdrei      
[3]{\Ppsymbol\beginargs{#1}\separgs{#2}\separgs{#3}\allargs}
\mathcommand\Ppppvier
[4]{\Ppsymbol\beginargs{#1}\separgs{#2}\separgs{#3}\separgs{#4}\allargs}
\mathcommand\Qppp          [2]{\Qpsymbol\beginargs{#1}\separgs{#2}\allargs}
\mathcommand\Qpppeins      [1]{\Qpsymbol\beginargs{#1}\allargs}
\mathcommand\Qpppdrei      
[3]{\Qpsymbol\beginargs{#1}\separgs{#2}\separgs{#3}\allargs}
\mathcommand\Fatherpp      [2]{\Fathersymbol\beginargs{#1}\separgs{#2}\allargs}
\mathcommand\Marriespp     [2]{\Marriessymbol\beginargs{#1}\separgs{#2}\allargs}
\mathcommand\Lovespp       [2]{\Lovessymbol\beginargs{#1}\separgs{#2}\allargs}
\mathcommand\StolenBypp    [2]
{\StolenBysymbol\beginargs{#1}\separgs{#2}\allargs}
\mathcommand\Humanpp       [1]{\Humansymbol\beginargs{#1}\allargs}
\mathcommand\Evenpp        [1]{\Evensymbol\beginargs{#1}\allargs}
\mathcommand\Evenppi       [2]{\Evensymbol^{#1}\beginargs{#2}\allargs}
\mathcommand\Oddpp         [1]{\Oddsymbol\beginargs{#1}\allargs}
\mathcommand\Primepp       [1]{\Primesymbol\beginargs{#1}\allargs}
\mathcommand\EveryPairpp  [2]{\EveryPairsymbol\beginargs{#1}\separgs
{#2}\allargs}
\mathcommand\mindexppeins  [2]{\mindexsymbol{#1}\beginargs{#2}\allargs}
\mathcommand\Givepp        [3]{\Givesymbol
\beginargs{#1}\separgs{#2}\separgs{#3}\allargs}
\mathcommand\mindexppzwei  [3]{\mindexsymbol
{#1}\beginargs{#2}\separgs{#3}\allargs}
\mathcommand\mindexppdrei  [4]{\mindexsymbol
{#1}\beginargs{#2}\separgs{#3}\separgs{#4}\allargs}

\mathcommand\nonnegppp     [1]{\nonnegpsymbol\beginargs{#1}\allargs}

\mathcommand\anonymouscsymbol{c}
\mathcommand\anonymouscindexsymbol[1]{\anonymouscsymbol_{#1}}
\mathcommand\anonymousfsymbol{f}
\mathcommand\anonymouscpp
[2]{\anonymouscsymbol\beginargs{#1}\separgs\ldots\separgs{#2}\allargs}
\mathcommand\anonymouscindexpp
[3]{\anonymouscindexsymbol{#1}\beginargs{#2}\separgs\ldots\separgs{#3}\allargs}
\mathcommand\anonymousfpp
[2]{\anonymousfsymbol\beginargs{#1}\separgs\ldots\separgs{#2}\allargs}
\mathcommand\coerceindexpp[3]{[#3]_{#1}^{#2}}

\mathcommand\Elephantppp    [1]{\Elephantpsymbol\beginargs{#1}\allargs}
\mathcommand\Flowerppp      [1]{\Flowerpsymbol  \beginargs{#1}\allargs}
\mathcommand\Bicycleppp     [1]{\Bicyclepsymbol \beginargs{#1}\allargs}
\mathcommand\Germanppp      [1]{\Germanpsymbol  \beginargs{#1}\allargs}
\mathcommand\Hugeppp        [1]{\Hugepsymbol    \beginargs{#1}\allargs}
\mathcommand\Animalppp      [1]{\Animalpsymbol  \beginargs{#1}\allargs}
\mathcommand\Maleppp        [1]{\Malepsymbol    \beginargs{#1}\allargs}
\mathcommand\Boyppp         [1]{\Boypsymbol     \beginargs{#1}\allargs}
\mathcommand\Girlppp        [1]{\Girlpsymbol    \beginargs{#1}\allargs}
\mathcommand\Femaleppp      [1]{\Femalepsymbol  \beginargs{#1}\allargs}
\mathcommand\Roundppp       [1]{\Roundpsymbol   \beginargs{#1}\allargs}
\mathcommand\Bishoppp       [1]{\Bishopsymbol   \beginargs{#1}\allargs}
\mathcommand\Quadrangularppp[1]{\Quadrangularpsymbol  \beginargs{#1}\allargs}
\mathcommand\Metppp[2]{\Metpsymbol     \beginargs{#1}\separgs{#2}\allargs}

\newcommand\bound     {{\rm bound}}
\newcommand\free      {{\rm free}}

\mathcommand\Vtripleindex[3]{\V\!_{{#1},\,{#2},\,{#3}}}
\mathcommand\Vdoubleindex[2]{\V\!_{{#1},\,{#2}}}
\mathcommand\Vsingleindex[1]{\V\!_{{#1}}}

\mathcommand\Erel[1]{\Gammaoffont\!_{#1}}
\mathcommand\Urel[1]{\Deltaoffont_{#1}}

\mathcommand\theRprimefromstrongtoweak{
  \inparenthesesinlinetight{
     \domres\id{\Vwall\cup\Vsome\setminus\RAN\varsigma}
     \nottight{\nottight\uplus}
     \reverserelation\varsigma
  }
  \nottight{\circ}
  \ranres
    {\transclosureinline R}
    {\Vwall\cup\Vsome\setminus\RAN\varsigma}
  \nottight{\nottight{\nottight{\uplus}}}
  \Vsome\tighttimes\Vsall
}

\mathcommand\deltaminus{\delta^-}
\mathcommand\deltaplus{\delta^+}
\mathcommand\deltaplusplus{\delta^{+^+}}
\mathcommand\deltastar{\delta^*}
\mathcommand\deltastarstar{\delta^{*^*}}

\mathcommand\Vall     {\Vsingleindex\indexdelta         }
\mathcommand\Vwall    {\Vsingleindex\indexdeltaminu     }
\mathcommand\Vsall    {\Vsingleindex\indexdeltaplus     }
\mathcommand\Vgsome   {\Vsingleindex\indexgammaplus     }
\mathcommand\Vsome    {\Vsingleindex\indexgamma         }
\mathcommand\Vfree    {\Vsingleindex\indexfree          }
\mathcommand\Vbound   {\Vsingleindex\indexbound         }
\mathcommand\Vsomesall{\Vsingleindex\indexgammadeltaplus}

\mathapplycommand\VARall      {\VARsingleindex\indexdelta         }
\mathapplycommand\VARwall     {\VARsingleindex\indexdeltaminu     }
\mathapplycommand\VARsall     {\VARsingleindex\indexdeltaplus     }
\mathapplycommand\VARgsome    {\VARsingleindex\indexgammaplus     }
\mathapplycommand\VARsome     {\VARsingleindex\indexgamma         }
\mathapplycommand\VARfree     {\VARsingleindex\indexfree          }
\mathapplycommand\VARbound    {\VARsingleindex\indexbound         }
\mathapplycommand\VARsomesall {\VARsingleindex\indexgammadeltaplus}
\mathcommand\displayVARsall[1]{\VARsingleindex\indexdeltaplus
\!\!\!\:\left(\begin{array}{@{}c@{}}#1\end{array}\right)}

\mathcommand\rigidvari     [2]{#1_{#2}^\indexgammadeltaplus}
\mathcommand\existsvari    [2]{#1_{#2}^\indexgamma    }
\mathcommand\forallvari    [2]{#1_{#2}^\indexdelta    }
\mathcommand\freevari      [2]{#1_{#2}^\indexfree     }
\mathcommand\wforallvari   [2]{#1_{#2}^\indexdeltaminu}
\mathcommand\sforallvari   [2]{#1_{#2}^\indexdeltaplus}
\mathcommand\gexistsvari   [2]{#1_{#2}^\indexgammaplus}
\mathcommand\boundvari     [2]{#1_{#2}}
\mathcommand\vari          [2]{#1_{#2}}
\mathcommand\wforallvarilow[2]{#1_{#2}^
{\raisebox{-.82ex}{\math\indexdeltaminu}}}

\newcommand\indexhelper[1]{{\scriptscriptstyle#1\:\!\!}}
\newcommand\indexdeltaplus
{\indexhelper{\delta^{\raisebox{-.17ex}{\fvesf\hskip-0.14em +}}}}
\newcommand\indexdeltaminu
{\indexhelper{\delta^{\mbox{\fvesf\hskip-0.14em\rule[.2ex]{.7em}{.15ex}}}}}
\newcommand\indexgammaplus
{\indexhelper{\gamma^{\mbox{\fvesf\hskip-0.14em +}}}}
\newcommand\indexgammadeltaplus
{\indexhelper{\gamma\delta^{\raisebox{-.17ex}{\fvesf\hskip-0.14em +}}}}

\newcommand\indexdelta{\indexhelper\delta}
\newcommand\indexgamma{\indexhelper\gamma}
\newcommand\indexfree
{{\scriptscriptstyle\free}}
\newcommand\indexbound
{{\scriptscriptstyle\bound}}

\newcommand\Wellfsymb{\ident{Wellf}}
\mathapplycommand\Wellfpp{\Wellfsymb}
\mathcommand\beginargs{(}
\mathcommand\allargs  {)}
\mathcommand\separgs  {,\,}
\mathcommand\tightsepargs{,}

\mathcommand\minusppnoparentheses  [2]{{#1}\,\minussymbol\,{#2}}
\mathcommand\tightminusppnoparentheses  [2]{{#1}\minussymbol{#2}}
\mathcommand\divideppnoparentheses [2]{{#1}\,\dividesymbol\,{#2}}
\mathcommand\plusppnoparentheses   [2]{{#1}\,\plussymbol \,{#2}}
\mathcommand\plusppnoparenthesesi  [3]{{#2}\,\plussymbol^{#1}\,{#3}}
\mathcommand\tightplusppnoparentheses   [2]{{#1}\plussymbol{#2}}
\mathcommand\timesppnoparentheses  [2]{{#1}\,\timessymbol\,{#2}}
\mathcommand\undppnoparentheses    [2]{{#1}\und            {#2}}
\mathcommand\oderppnoparentheses   [2]{{#1}\oder           {#2}}
\mathcommand\impliesppnoparentheses[2]{{#1}\implies        {#2}}
\mathcommand\leqinfixppnoparentheses[2]{{#1}\,\tight\leq\,{#2}}
\mathcommand\geqinfixppnoparentheses[2]{{#1}\,\tight\geq\,{#2}}
\mathcommand\dividepp [2]{(\divideppnoparentheses {#1}{#2})}
\mathcommand\minuspp  [2]{(\minusppnoparentheses  {#1}{#2})}
\mathcommand\pluspp   [2]{(\plusppnoparentheses   {#1}{#2})}
\mathcommand\timespp  [2]{(\timesppnoparentheses  {#1}{#2})}
\mathcommand\undpp    [2]{(\undppnoparentheses    {#1}{#2})}
\mathcommand\oderpp   [2]{(\oderppnoparentheses   {#1}{#2})}
\mathcommand\impliespp[2]{(\impliesppnoparentheses{#1}{#2})}

\newcommand\citebossessurvey      {\avenhaus\ \& \madlener\ (1989)}
\newcommand\citeSR                {\avenhaus\ \& \becker\ (1992)}
\newcommand\citebeckerstacs       {\avenhaus\ \& \becker\ (1994)}
\newcommand\citeloriaextravariables{\avenhaus\ \& \loria\ (1994)}
\newcommand\citebeckertapsoft     {\becker\ (1993)}
\newcommand\citebeckerdiss        {\becker\ (1994)}
\newcommand\citebergstraklop      {\bergstra\ \& \klop\ (1986)}
\newcommand\citedershowitz        {\dershowitz\ (1987)}
\newcommand\citeder               {\dershowitz\ \etalabbrev\ (\citederdate)}
\newcommand\citederdate           {1988}
\newcommand\citeopenproblems     {\dershowitz\ \etalabbrev\ (1991)}
\newcommand\citegeserterminationpair{\geser\ (1994)}
\newcommand\citegramlichmodularconditional{\gramlich\ (1994)}
\newcommand\citegramlichelkuzdate  {1995a}
\newcommand\citegramlicheiteldate  {1995b}
\newcommand\citegramlichelkuz      {\gramlich\ (\citegramlichelkuzdate)}
\newcommand\citegramlicheitel      {\gramlich\ (\citegramlicheiteldate)}
\newcommand\citehuet              {\huet\ (1980)}
\newcommand\citekaplanugly        {\kaplan\ (1987)}
\newcommand\citekaplan            {\kaplan\ (1988)}
\newcommand\citekapurcp           {\kapur\ \etalabbrev\ (1988)}
\newcommand\citeklophandbook       {\klop\ (1992)}
\newcommand\citeklopdiss           {\klop\ (1980)}
\newcommand\citesubconnected      {\kuechlin\ (1985)}
\newcommand\citemiddeldorpcond    {\middeldorp\ (1993)}
\newcommand\citemiddeldorpbasicnarrowing{\middeldorp\ \& \hamoen\ (1994)}
\newcommand\citeoostromdissdate   {1994a}
\newcommand\citeoostromcolordate  {1994b}
\newcommand\citeoostromdiss       {\oostrom\ (\citeoostromdissdate)}
\newcommand\citeoostromcolor      {\oostrom\ (\citeoostromcolordate)}
\newcommand\citeplaisted          {Plaisted (1985)}
\newcommand\citetoyama            {\toyama\ (1988)}
\newcommand\citetoyamadiss        {\toyama\ (1990)}
\newcommand\citesuzuki             {\suzuki\ \etalabbrev\ (1995)}
\newcommand\citewaltherhandbook   {\walther\ (1994)}
\newcommand\citewgctrslncs        {\wirth\ \& \gramlich\ (1993)}
\newcommand\citewgkp              {\wirth\ \etalabbrev\ (1993)}
\newcommand\citewgcade            {\wirth\ \& \gramlich\ (\citewgcadeyear)}
\newcommand\citewgcadeyear        {1994b}
\newcommand\citewgjsc             {\wirth\ \& \gramlich\ (\citewgjscyear)}
\newcommand\citewgjscyear         {1994a}
\newcommand\citewirthlunde        {\wirth\ \& Lunde (1994)}

\input diagrams
\diagramstyle[%
PostScript=dvips,%
height=2em,%
width=3.0em,%
tight,%
midshaft,%
balance,%
dpi=80,%
heads=vee,labelstyle=\scriptstyle,%
]

\def\paramH{\hbox to 0pt{\hss$\parallel$\hss}}
\def\paramV{\smash{=\!\!=}}
\newarrowmiddle{++}\paramH\paramH\paramV\paramV

\newarrow{red}---->
\newarrow{redexists}{}{dash}{}{dash}>
\newarrow{redpara}--{++}->
\newarrow{equal}=====
\newarrow{antired}<----

\def\rredindex#1{\rred_{#1%\littleheadroom
}}
\def\dredindex#1{\dred_{#1}}
\def\rantiredindex#1{\rantired_{#1}}
\def\dantiredindex#1{\dantired_{#1}}

\def\rrefltransindex#1{\rred^{\textstyle\ast}_{#1%\littleheadroom
}}
\def\drefltransindex#1{\dred^{\textstyle\ast}_{#1}}
\def\rrefltrans{\rred^{\textstyle\ast}}
\def\drefltrans{\dred^{\textstyle\ast}}
\def\rantirefltransindex#1{\rantired^{\textstyle\ast}_{#1}}

\def\rantirefltrans{\rantired^{\textstyle\ast}}

\def\rredparaindex#1{\rredpara_{#1}}
\def\dredparaindex#1{\dredpara_{~~#1}}

\def\ronlyonceindex#1{\rredindex{#1}^{%{\scriptscriptstyle\leq}1
=}}
\def\donlyonceindex#1{\dredindex{#1}^{%{\scriptscriptstyle\leq}1
=}}

\def\rredindexn#1#2{\rredindex{#2}^{#1}}
\def\dredindexn#1#2{\dredindex{#2}^{#1}}
\def\rantiredindexn#1#2{\rantiredindex{#2}^{#1}}

\input headerjscbib
\setjscbibstyle
\newlength{\contentsandreferencesfootroom} %room under line ``Contents''
\newdimen\rotdimen
\def\vspec#1{\special{ps:#1}}%  passes #1 verbatim to the output
\def\rotstart#1{\vspec{gsave currentpoint currentpoint translate
   #1 neg exch neg exch translate}}% #1 can be any origin-fixing transformation
\def\rotfinish{\vspec{currentpoint grestore moveto}}% gets back in synch
\def\rotr#1{\rotdimen=\ht#1\advance\rotdimen by\dp#1%
   \hbox to\rotdimen{\hskip\ht#1\vbox to\wd#1{\rotstart{90 rotate}%
   \box#1\vss}\hss}\rotfinish}
\def\rotl#1{\rotdimen=\ht#1\advance\rotdimen by\dp#1%
   \hbox to\rotdimen{\vbox to\wd#1{\vskip\wd#1\rotstart{270 rotate}%
   \box#1\vss}\hss}\rotfinish}%
\def\rotu#1{\rotdimen=\ht#1\advance\rotdimen by\dp#1%
   \hbox to\wd#1{\hskip\wd#1\vbox to\rotdimen{\vskip\rotdimen
   \rotstart{-1 dup scale}\box#1\vss}\hss}\rotfinish}%
\def\rotf#1{\hbox to\wd#1{\hskip\wd#1\rotstart{-1 1 scale}%
   \box#1\hss}\rotfinish}%

\mathcommand\defshallowjoinablecondition
{
       \forall i\prec 2\stopq
       \inparentheses{
         \omega\tight* n_i\preceq\gamma_i\ \ \ \ \wedge\ 
         \\
         D_i\varphi\mbox{ fulfilled \wrt\ }\redindex{\RX,\gamma_i}
       }
}

\mathcommand\criticalpeak
{ \ 
  ((t_0,
    D_0,
    n_0),\ 
   (t_1,
    D_1,
    n_1),\ 
  \hat t,\ 
  p)
  \ 
}

\mathcommand\newcriticalpeak
{\ ((t_0,D_0,\Lambda_0),\ (t_1,D_1,\Lambda_1),\ \hat t,\ p)\ }

\mathcommand\criticalpeaklongform
{ ((t_0,
    D_0,
    \Lambda_0),\ 
   (t_1,
    D_1,
    \Lambda_1),\ 
  \hat t,\ 
  \sigma,\ 
  p)
}

\mathcommand\tightpreclhd{\prec\!\!\lhd}
\mathcommand\preclhd     {\,\,\tightpreclhd\,\,}
\mathcommand\preclhdeq   {\,\,\underline{\math{\:\!\!\tightpreclhd\!}}\,\,}
\mathcommand\tightsuccrhd{\succ\!\!\!\rhd}
\mathcommand\succrhd     {\,\,\tightsuccrhd\,\,}

\mathapplycommand\uplim{\rm uplim}
\mathapplycommand\lowlim{\rm lowlim}

\mathcommand\maxoftwo[2]{{\rm max}\{{#1},{#2}\}}
\mathcommand\maxofthree[3]{{\rm max}\{{#1},{#2},{#3}\}}

\newcommand\lemmamonotonicinbeta{Lemma~\ref{lemma monotonic wrt ordinals}}
\newcommand\lemmaconskeeping{Lemma~\ref{lemma about conskeeping}}
\newcommand\lemmaaboutconflu{Lemma~\ref{lemma about conflu}}

\mathcommand\plusomega{+_{\!\!_{\omega}}}
\mathcommand\plusalpha{+_{\!\!_{\alpha}}}
\mathcommand\pluszero {+_{\!\!_{0     }}}
\begin{document}
%%%%%%%%%%%%%%%%%%%%%%%%%%%%%%%%%%%%%%%%%%%%%%%%%%%%%%%%%%%%%%%%%%%%%%%%%%%%%%%
{\flushbottom
\thispagestyle{empty}
\setcounter{page}{1}

\noindent
\begin{minipage}{\textwidth}
\mbox{}

\vspace{10.0em}
\begin{center}
{\LARGE\bf
Syntactic
Confluence 
Criteria
for
\\
%Constructor-Based
%\\
Positive/Negative-Conditional
\\
Term Rewriting Systems
\\\mbox{}
}
{
\\
Claus-Peter Wirth
\\\mbox{}
\\\mbox{}
\\\mbox{}
\\
}
{\small
SEKI Report SR--95--09
\\
\mbox{}Searchable Online Edition
\\\mbox{}\\
July~23, 1995
\\
Revised March~6, 1996
\\
(``noetherian'' replaced with ``terminating''.
Example~\ref{ex quasi over} added.) 
\\
\mbox{}\\
Revised January and October 2005
\\
\mbox{}\\
Universit\"at Kaiserslautern
\\
Fachbereich Informatik
\\
D-67663 Kaiserslautern
\\\mbox{}
\\
}

\end{center}
\end{minipage}

\vspace{\fill}
{\footnotesize

\noindent
{\bf Abstract:}
\renewcommand{\baselinestretch}{0.8}
We study 
the combination of the following already known ideas for showing confluence 
of unconditional or 
conditional term rewriting systems into practically more useful confluence
criteria for conditional systems:
Our syntactic separation into constructor and non-constructor symbols,
\huet's introduction 
and \toyama's generalization
of parallel closedness for non-terminating unconditional systems,
the use of shallow confluence for proving confluence 
of terminating and non-terminating conditional systems, 
the idea that certain kinds of limited confluence can be assumed for
checking the fulfilledness or infeasibility of the conditions of 
conditional critical pairs,
and
the idea that (when termination is given)
only prime superpositions have to be considered
and certain normalization restrictions can be applied
for the substitutions fulfilling the conditions of conditional critical pairs.
Besides combining and improving already known methods, 
we present the following new ideas and results:
We strengthen the criterion for overlay joinable terminating systems, and,
by using the expressiveness of our syntactic separation
into constructor and non-constructor symbols, we are able 
to present criteria for level confluence that are 
not criteria for shallow confluence actually
and also able
to weaken the
severe requirement of normality (stiffened with left-linearity) in the 
criteria for shallow confluence 
of terminating and non-terminating conditional systems
to the easily satisfied requirement of quasi-normality.
Finally, the whole paper also gives a practically useful
overview of the syntactic means for
showing confluence of conditional term rewriting systems.

}

\noindent
\footnoterule
\noindent
{\footnotesize 
This research was supported by the Deutsche
Forschungsgemeinschaft, SFB 314 (D4-Projekt)

}}

\pagebreak

\setcounter{tocdepth}{1}
\pagestyle{empty}
\setcounter{page}{2}
\samepage{
%\vspace*{\fill}
\tableofcontents\nopagebreak
%\contentsline{section}{References}{\pageref{bibliography}}
\vfill\vfill
 
}%samepage

\pagebreak

%%%%%%%%%%%%%%%%%%%%%%%%%%%%%%%%%%%%%%%%%%%%%%%%%%%%%%%%%%%%%%%%%%%%%%%%%%%%%%%
%\input{headerdate} % for working versions
\pagestyle{myheadings}% % for final version
\pagenumbering{arabic}
\setcounter{page}{1}
%%%%%%%%%%%%%%%%%%%%%%%%%%%%%%%%%%%%%%%%%%%%%%%%%%%%%%%%%%%%%%%%%%%%%%%%%%%%%%%
\section{Introduction and Overview}
\label{section motivation}

%\yestop
%\noindent
While\footnote{Please do try not to read the footnotes for a first reading!}
powerful confluence criteria for conditional term rewriting systems%
\footnote{\headroom For an introduction to the subject 
\cf\ \citebossessurvey\ or \citeklophandbook.}
are in great demand and 
while there are interesting new results
for\emph{unconditional} systems\footnote{\headroom\Cf\ 
\citeoostromdiss\ and \citeoostromcolor.
Note that the lemmas 5.1 and 5.3 of \citeoostromcolor\ do not apply
for conditional systems because 
they are not subsumed by the notion of 
``patterm rewriting systems'' of \citeoostromcolor.},
hardly any new results on confluence of\emph{conditional} 
term rewriting systems
(besides some on modularity\footnote{\headroom
\Cf\ \citemiddeldorpcond, \citegramlichmodularconditional.}
and on the treatment of extra-variables in conditions\footnote{\headroom\Cf\ 
\citeloriaextravariables\ 
for the case of decreasing systems
and \citesuzuki\ for the case of orthogonal systems.})
have been published since \citeder,
\citetoyama, and \citebergstraklop, and not even a common generalization
(as given by our theorems
 \ref{theorem parallel closed} and \ref{theorem parallel closed zero})
of 
the main confluence
theorems of the latter two papers 
(\ie\ something like confluence of parallel closed conditional systems)
has to our knowledge been published.
We guess that this is due to the following problems:

\begin{enumerate}

\item
A proper treatment is very tedious and technically
most complicated, especially in the case of non-terminating
reduction relations.%
\footnote{\headroom
The technique we apply
for proving our confluence criteria for non-terminating
reduction relations is in essence to show strong confluence of relations 
whose reflexive \& transitive closures are equal to that
of the reduction relation.

In \citebergstraklop\ another technique is used. Instead of an actual 
presentation of the proof there is only a pointer to \citeklopdiss.
It would be worthwhile to reformulate this proof in modern notions
(including path orderings) and notations.
While we did not do this, we just try to describe here
the abstract global idea of this proof: 
 
The field of the 
reduction relation is changed from terms to terms with licenses in such
a way that the projection to terms just yields the original reduction relation
again.
The transformed reduction relation becomes terminating since it consumes
and inherits licenses in a wellfounded manner; thus 
its confluence is implied by its local confluence that is to be shown.
Finally, each diverging peak of the 
original reduction relation is a projection of a diverging peak in the 
transformed reduction relation when one only provides enough licenses.

We did not apply this global proof idea since (while we were able to 
generalize it for allowing parallel closed critical pairs as in 
the corollary on page 815 in \citehuet) we were not able to generalize
it for proving  Corollary~3.2 of \citetoyama\ 
(which generalizes this corollary of \citehuet).%
} 

\item
There is a big gap between the known criteria and those 
criteria that are supposed to be true, even for unconditional systems.%
\footnote{\headroom\Cf\ \eg\ Problem~13 of \citeopenproblems.}

\item
The usual framework for conditional term
rewriting systems does not allow us to model
some simple and straightforward applications naturally 
in such a way that the
resulting reduction relation is known to be confluent, unless some
sophisticated semantic
or termination knowledge is postulated a priori.
\yestop

\pagebreak
\item\label{item infinite number of substitutions}
For conditional rule systems 
there is another big gap between the known criteria and those 
criteria that are required for 
practical purposes. 
This results from the difficulty to capture (with effective means)
the infinite number
of substitutions that must be tested for fulfilling the conditions of critical
pairs.

\end{enumerate}

\vfill

\noindent
While we are not able to contribute too much regarding the first two problems,
we are able to present some progress with the latter two.

\yestop
\noindent
Our positive/negative-conditional rule systems including a syntactic
separation between constructor and non-constructor symbols as presented in 
\citewgjsc\ offer more expressive power than the standard positive conditional
rule systems and therefore allow us to model more applications 
naturally in such a way that their confluence is given by the 
new confluence criteria presented in this paper.
Using the separation into constructor and non-constructor rules
(generated by the syntactic separation into constructor and non-constructor
function symbols)
it is possible to divide the problem of showing confluence of the whole
rule system into three smaller sub-problems, namely
confluence of the constructor rules,
confluence of the non-constructor rules,
and their commutation.
The important advantage of this modularization is not only the division 
into smaller problems, but is due to the possibility to tackle the sub-problems
with different confluence criteria. 
\Eg, when confluence of the constructor rules is not trivial
then its confluence often can only be shown by sophisticated semantical
considerations or by criteria that are applicable to terminating systems only.
For the whole rule system, however, neither semantic confluence criteria
nor confluence criteria requiring termination of the reduction relation
are practically feasible in general. This is because, 
on the one hand, an effective application
of semantic confluence criteria requires that the specification given by the
whole rule system has actually been modeled before in some formalism.
On the other hand, 
termination of the whole rule system may not be given or difficult
to be shown without some confluence assumptions.\footnote{When termination is 
assumed, there are approaches to prove confluence automatically, 
\cf\ \citebeckertapsoft\ and \citebeckerdiss.}
Fortunately, 
without requiring termination of the whole rule system
the syntactic confluence criteria%
\footnote{\Cf\ our theorems
\ref{theorem complementary},
\ref{theorem weakly complementary},
and
\ref{theorem complementary zero}.}
presented in this paper guarantee confluence of the non-constructor rules
of a class of rule systems that is sufficient for practical specification.
This class of rule systems generalizes 
the function specification style used
in the framework of classic inductive theorem proving%
\footnote{\Cf\ \citewaltherhandbook.
Note that we can even keep the notation style similar to 
this function specification style, \cf\ \citewirthlunde.}
by allowing of partial functions resulting from incomplete specification
as well as from non-termination.
Together with the notions of inductive validity presented in \citewgcade\
this extends the area of semantically clearly understood inductive 
specification considerably.

\yestop
\noindent
Regarding the last problem of the above problem list 
(occuring in case of conditional rule systems),
by carefully including the invariants of the proofs for the 
confluence criteria into the conditions of the joinability tests for the 
conditional critical pairs we allow of more reasoning on those substitutions
that fulfill the condition of a critical pair. 
\Eg\ consider the following example:

\vfill

\pagebreak

\notop\halftop
\begin{example}
Let\/ \R:
\bigmath{  \begin{array}[t]{lllll}
  \fppeins{\spp{\spp x}}
 &\boldequal
 &\spp\zeropp
 &\rulesugar
 &\fppeins x\boldequal\zeropp
\\\fppeins{\spp{\spp x}}
 &\boldequal
 &\zeropp
 &\rulesugar
 &\fppeins x\boldequal\spp\zeropp
\\\fppeins{\spp\zeropp}
 &\boldequal
 &\spp\zeropp
\\\fppeins\zeropp
 &\boldequal
 &\zeropp
\\\end{array}}

\noindent
Assume \zeropp\ and \spp\zeropp\ to be irreducible.
\end{example}

\noindent
The experts may notice that the part of \R\ we are given in this example 
is rather well-behaved: It is left-linear and normal;
it may be decreasing; and the only critical pair is an overlay.
Now, for showing the critical pair between the first two rules to be
joinable, one has to show that it is impossible that both conditions hold 
simultaneously for a substitution $\{x\!\mapsto\! t\}$.
One could argue the following way:
If both conditions were fulfilled, then \fppeins t would reduce to \zeropp\ as 
well as to \spp\zeropp, which contradicts confluence below \fppeins t.
But, as our aim is to establish confluence, 
it is not all clear that we are allowed to assume confluence for the
joinability test here.
None of the theorems in \citeder\ or \citebergstraklop\ provides us with
such a confluence assumption, 
even if their proofs could do so with little additional effort.
For practical purposes, however, it is important that 
the joinability test allows us to assume a sufficient kind of confluence
for the condition terms.
Therefore, all our joinability notions provide us with sufficient assumptions
that allow us to easily establish the infeasibility of the condition of a
critical pair,
without knowing the proofs for the confluence criteria by heart.
This applies for example, when
two rules with same left-hand side are meant to express
a case distinction that is established by the condition of the one
containing a condition literal 
``\math{p\boldequal\truepp}'' or ``\math{u\boldequal v}''
and the condition 
of the other containing the condition literal
``\math{p\boldequal\falsepp}'' or ``\math{u\boldunequal v}''.\footnote{In 
Definition~4.4 of \citeloriaextravariables\ the critical pair resulting
from such two rules is called ``infeasible'' (in the case with
``\math{p\boldequal\truepp}'' and ``\math{p\boldequal\falsepp}''). 
We will call it 
``complementary'' instead (in both cases), 
\cf\ Theorem~\ref{theorem complementary}.}

For terminating reduction relations we carefully investigate whether
the joinability test can be restricted by certain irreducibility 
requirements, \eg\ whether the substitutions which must be tested for
fulfilling the conditions of critical pairs can be required to be 
normalized, \cf\ \sect~\ref{section now termination}, \esp\
Example~\ref{example integers}.
The restrictions on the infinite number of substitutions for which
the condition of a critical pair must be tested for fulfilledness
may be a great help in practice.
However, they do not solve the principle problem
that the number of substitutions is still infinite.

\yestop
\noindent
Another important point is that we weaken the severe restriction
imposed on 
terminating systems by Theorem~2 of \citeder\
and on 
non-terminating systems by Theorem~3.5 of \citebergstraklop,
namely normality, which in our framework
can be considerably weakened to the so-called {\em quasi-normality},
\cf\ our theorems \ref{theorem parallel closed} and \ref{theorem quasi-free}.

Moreover, besides these two criteria for shallow confluence,
we present to our knowledge the first criteria
for level confluence that are
not criteria for shallow confluence actually%
\footnote{as is the case with \citesuzuki.},
\cf\ our theorems \ref{theorem level parallel closed} and
\ref{theorem level one}.

Finally, we considerably improve the notion of 
``quasi overlay joinability'' of \citewgjsc, generalizing the notion
of    ``overlay joinability'' of \citeder.
This results in a stronger criterion with a simpler proof,
\cf\ \sect~\ref{sect quasi overlay joinability}
and Theorem~\ref{theorem quasi overlay joinable}.

\vfill

\pagebreak

\yestop
\noindent
Since our main interest is in positive/negative-conditional rule systems 
with two kinds of variables and two kinds of function symbols
as presented in \citewgkp\ and \citewgjsc, 
the whole paper is based on this framework.
We know that this is problematic because 
the paper may also be of interest for readers interested in 
positive conditional rule systems with one kind of variables and function
symbols only:
With the exception of our generalization of normality to quasi-normality
and our criteria for level confluence,
our results also have interesting implications for this special case
(which is subsumed by our approach).
Nevertheless we prefer our more expressive framework for this presentation
because it provides us with more power for most of our confluence criteria
which is lost when restricting them to the standard framework.
Therefore in the following section we are going to repeat
those results of \citewgjsc\ which are essential for this paper.
Those readers who are only interested in the implications of this paper
for standard positive conditional rule systems with one kind of variables 
and function symbols should try to read only the theorems presented or 
pointed at in 
\sect~\ref{section constructor confluence}, which have been supplied with
independent proofs for allowing a direct understanding.
The contents of the other sections are explained by their titles.
For a first reading sections \ref{section sophisticated shallow} and
\ref{section sophisticated level} should only be skimmed and its definitions
looked up by need. Due to their enormous length,
most of the proofs have been put into \ref{sect proofs}.

\yestop
\noindent
We conclude this section with a list on where in this paper to find 
generalizations of known theorems:
\begin{description}

\item[Parallel Closed + Left-Linear + Unconditional:]\mbox{}\\
The corollary on page 815 in \citehuet\ as well as
Corollary~3.2 in \citetoyama\
are generalized by our theorems 
\ref{theorem parallel closed}(I),
\ref{theorem parallel closed}(III),
\ref{theorem parallel closed}(IV),
\ref{theorem level parallel closed}(I),
\ref{theorem level parallel closed}(III),
\ref{theorem level parallel closed}(IV),
and
\ref{theorem parallel closed zero}(I).

\item[No Critical Pairs + Left-Linear + Normal:]\mbox{}\\
Theorem~3.5 in \citebergstraklop\ 
as well as 
Theorem~1 in \citeder\
are generalized by our theorems
\ref{theorem complementary},
\ref{theorem weakly complementary},
\ref{theorem parallel closed},
\ref{theorem parallel closed zero},
and
\ref{theorem complementary zero}.

\item[Strongly Joinable + Strong Variable Restriction:]\mbox{}\\
Lemma~3.2 of \citehuet\ as well as
the translation of Theorem~5.2 in \citebeckerstacs\
into our framework is generalized by our theorems
\ref{theorem parallel closed}(II)
and
\ref{theorem level parallel closed}(II).

\item[Shallow Joinable + Left-Linear + Normal + Terminating:]\mbox{}\\
Theorem~2 in \citeder\ is generalized by our theorems
\ref{theorem quasi-free} and
\ref{theorem quasi-free zero}.

\item[Overlay Joinable + Terminating:]\mbox{}\\
Theorem~4 in \citeder\ as well as
Theorem~6.3 in \citewgjsc\ are generalized by our theorem
\ref{theorem quasi overlay joinable}.

\item[Joinable + Variable Restriction + Terminating:]\mbox{}\\
Theorem~7.18 in \citewgjsc\ 
is generalized by our theorem \ref{theorem quasi-free three}.

\item[Joinable + Decreasing:]\mbox{}\\
Theorem~3.3 in \citekaplanugly,
Theorem~4.2 in \citekaplan,
Theorem~3 in \citeder,
as well as
Theorem~7.17 in \citewgjsc\ 
are generalized by our theorems
\ref{theoremconfluence}
and
\ref{theorem quasi-free three}.

\pagebreak
\end{description}

\section{Positive/Negative-Conditional Rule Systems}

\begin{sloppypar}
\yestop
\noindent
We use `\math\uplus' for the union of disjoint classes and `\id' for the
identity function.
`\N' denotes the set of natural numbers and we define
\bigmath{\N_+:=\setwith{n\tightin\N}{0\tightnotequal n}.}
For 
%\linebreak
classes $A,B$ we define:
\bigmath{\DOM A:=\setwith{\!a}{\exists b\stopq          (a,b)\tightin A\!};}
\bigmath{\RAN A:=\setwith{\!b}{\exists a\stopq          (a,b)\tightin A\!};}
\bigmath{  B[A]:=\setwith{\!b}{\exists a\tightin A\stopq(a,b)\tightin B\!}.}
This use of ``[\ldots]'' should not be confused with our habit of stating two
definitions, lemmas, or theorems (and their proofs \etc)\ in one, 
where the parts between `[' and `]' are
optional and are meant to be all included or all omitted.
Furthermore, we use `\math\emptyset' to denote the empty set as well as the
empty function or empty word.
%We define the set of nonempty words over \math A as \math{A^+:=\setwith{a}
%{\exists n\tightin\N\stopq\nottight{\FUNDEF{a}{\{0,\ldots,n\}}A}}}; the set of
%words over \math A as \math{A^\ast:=\setwith{a}{\exists n\tightin\N\stopq
%\nottight{\FUNDEF{a}{\{0,\ldots,n\tight-1\}}A}}}; and the set of monadic words
%by \bigmath{ A^\times:=\setwith{a}{\nottight{a\tightequal\emptyset}\oder
%\nottight{\FUNDEF{a}{\{0\}}A}}}. Thus \bigmath{A^\ast\tightequal A^+\cup 
%A^\times.}
\end{sloppypar}

\yestop
\yestop
\subsection{Terms and Substitutions}
\label{section terms}

\yestop
\noindent
Since our approach is based on the consequent syntactic
distinction of constructors,
we have to be quite explicit about terms and substitutions.

\yestop
\noindent
We will consider terms of fixed arity over many-sorted signatures.
A\emph{signature}
\bigmath{
  \sig =(\sigfunsym,\sigsorts,\sigarity)
}
consists of 
an
enumerable
set of function symbols 
\sigfunsym,
a finite set of sorts 
\sigsorts\
(disjoint from \sigfunsym),
and a
computable
arity-function
\bigmath{\FUNDEF{\sigarity}{\sigfunsym}{\sigsorts^{+}}.}
For
\math{f\in\sigfunsym\stopq}\ 
$\sigarity(f)$
is the list of argument sorts augmented by the sort of the result of
$f$;
to ease reading we will sometimes insert a 
`\aritysugarsymbol' between
a nonempty list of argument sorts and the result sort.
A\emph{constructor sub-signature of the signature 
\math{(\sigfunsym,\sigsorts,\sigarity)}}
is a signature 
\bigmath{
  \cons = (\consfunsym,\sigsorts,\domres\sigarity\consfunsym)
}
such that the set \consfunsym\ is a decidable subset of \,\sigfunsym\@.
\ \ \consfunsym\ is called the set of\emph{constructor symbols}; the complement
\bigmath{\deffunsym = \sigfunsym \setminus \consfunsym}
is called the set of {\em non-constructor symbols}.

\begin{example}[Signature with Constructor Sub-Signature]\mbox{}\\\label{exa}
 \math{
  \begin{array}[t]{rclcl}
    \consfunsym &
    = &
    \multicolumn{3}{l}{\{\zeropp,
                         \ssymbol,
                         \falsepp,
                         \truepp,
                         \nilpp,
                         \cnssymbol\}}\\
    \deffunsym &
    = &
    \multicolumn{3}{l}{\{\minussymbol,
                         \mbpsymbol
                         \}}\\
    \sigsorts &
    = &
    \multicolumn{3}{l}{\{\nat,
                         \bool,
                         \lists\}}
    \\
    \\
    \sigarity(\zeropp) &
    = &
    &
    &
    \nat\\
    \sigarity(\ssymbol) &
    = &
    \nat &
    \aritysugar &
    \nat\\
  \end{array}\hfill
  \begin{array}[t]{|@{~~~}rclcl@{~~~~}}
    \sigarity(\falsepp) &
    = &
    &
    &
    \bool\\
    \sigarity(\truepp) &
    = &
    &
    &
    \bool\\
    \sigarity(\nilpp) &
    = &
    &
    &
    \lists\\
    \sigarity(\cnssymbol) &
    = &
    \nat\ \lists &
    \aritysugar &
    \lists\\
    \sigarity(\minussymbol) &
    = &
    \nat\ \nat &
    \aritysugar &
    \nat\\
    \sigarity(\mbpsymbol) &
    = &
    \nat\ \lists &
    \aritysugar &
    \bool\\
  \end{array}
 }
\end{example}

\noindent
To reduce declaration effort, in all 
examples 
(unless stated otherwise)
in this and the following sections we will have only one sort;
`\appzero', `\bppzero', `\cppzero', `\dppzero', `\eppzero', 
and `\zeropp'
will always be constants;
`\ssymbol', `\psymbol',
`\fsymbol', 
`\gsymbol', 
and 
`\hsymbol' 
will always denote functions with one argument;
`\plussymbol' and `\minussymbol' take two arguments in infix notation;
`\math W', `\math X', `\math Y', `\math Z' are variables from \Vsig\
(\cf\ below).
%`\math z' is a variable from \Vcons.

\begin{sloppypar}
\yestop
\noindent
A\emph{variable-system for a signature \math{(\sigfunsym,\sigsorts,\sigarity)}}
is an \sigsorts-sorted family of decidable
sets of variable symbols
which are mutually disjoint and disjoint from \sigfunsym.
By abuse of notation we will use the symbol `$X$' for an 
\sigsorts-sorted family 
to denote not only the family \bigmath{X=(X_{s})_{s\in\sigsorts}} itself,
but also the union of its {\em ranges\/}:
\bigmath{\bigcup_{s\in\sigsorts}X_{s}.}
As the basis for our terms
throughout the whole 
paper 
we assume two fixed disjoint variable-systems
\bigmath{\Vsig} of {\em general variables}\/ 
and \bigmath{\Vcons} of {\em constructor variables}\/ 
such that \Vsigindex{s} as well as 
\Vconsindex{s} contain infinitely many elements for each \math{s\in\sigsorts}.

\yestop
\noindent
\bigmath{\tss} denotes the \sigsorts-sorted family of all 
well-sorted {\em (variable-mixed) terms}\/ over `\sig/\sigV',
while
\bigmath{\tgsig} denotes the \sigsorts-sorted family of all 
well-sorted {\em ground terms}\/ over `\sig'.
Similarly,
\bigmath{\tcs} denotes the \sigsorts-sorted family of all 
{\em (variable-mixed) constructor terms},
\bigmath{\tcc} denotes the \sigsorts-sorted family of all 
{\em pure constructor terms},
while
\bigmath{\tgcons}
denotes the \sigsorts-sorted family of all 
{\em constructor ground terms}.
To avoid problems with empty sorts,
we assume \tgcons\ to have nonempty ranges only.

\yestop
\noindent
We define
\bigmath{\V:=(\Vvarsigmaindex{s})_{(\varsigma,s)\in\sigsortstimes}}
and call it a\emph{variable-system for a signature 
\math{(\sigfunsym,\sigsorts,\sigarity)} with constructor sub-signature}.
We use \VAR A to denote the \sigsortstimes-sorted family 
of variables occurring in a structure \math A
(\eg\ a term or a set or list of terms).
Let \math{\X\subseteq\V} be a variable-system.
We define 
\bigmath{\tX=(\tX_{\varsigma,s})_{(\varsigma,s)\in\sigsortstimes}}
by
(\math{s\in\sigsorts}):
\bigmath{\tX_{\SIG ,s}:=\tsigX      _s}
and
\bigmath{\tX_{\CONS,s}:=\tconsVconsX_s.}
To avoid confusion: Note that 
\bigmath{
  \tX_{\CONS,s}
  \subseteq
  \tX_{\SIG ,s}
}
for
\math{s\in\sigsorts},
whereas 
\bigmath{
  \Vconsindex{s}\cap\Vsigindex{s}=\emptyset
.}
Furthermore we write \gt\ for \temptyset\ as well as
\vt\ for \tV\@.
Our custom of reusing the symbol of a family for the union of its ranges
now allows us to write \bigmath{\vt} as a shorthand for \bigmath{\tss.}

\yestop
\noindent
For a term \math{t\in\vt} 
we denote by \TPOS t the\emph{set of its positions}
(which are lists of positive natural numbers),
by $t/p$ the subterm of \math t at position \math p,
and by \repl{t}{p}{t'} the result of replacing $t/p$ with $t'$ 
at position \math p in \math t.
We write \bigmath{\neitherprefix p q} 
to express that neither p is a prefix of q, nor q a prefix of p.
For \bigmath{\Pi\subseteq\TPOS t}
with
\bigmath{
  \forall p,q\tightin\Pi\stopq
     (p\tightequal q\oder\neitherprefix p q)
}
we denote
by \replpar{t}{p}{t_{p}'}{p\tightin\Pi} the result of replacing, 
for each \math{p\in\Pi}, 
the subterm at position \math p in the term \math t with the term \math{t_p'}. 
~~\math t is\emph{linear}
\udiff\ 
\bigmath{\forall p,q\tightin\TPOS t\stopq
 (\mbox{$t/p\!=\!t/q\!\in\!\V$}\implies p\!=\!q)\ 
.}

\yestop
\noindent
The set of\emph{substitutions}\/ from \V\ to a 
\sigsortstimes-sorted family of sets 
\bigmath{T=(T_{\varsigma,s})_{(\varsigma,s)\in\sigsortstimes}}
is defined to be
\\\linemath{\footroom\headroom
  \SUBST{\V}{T}
  :=
  \setwith{\FUNDEF\sigma\V T}
          {\forall(\varsigma,s)\tightin\sigsortstimes  \stopq
           \forall x           \tightin\Vvarsigmaindex{s}\stopq
              \sigma(x)\tightin T_{\varsigma,s}}
.}
Note that
\bigmath{\forall\sigma\tightin\SUBST{\V}{\vt}      \stopq
         \forall(\varsigma,s)\tightin\sigsortstimes\stopq
         \forall t\tightin\vtvarsigmaindex{s}      \stopq
            \ t\sigma\tightin\vtvarsigmaindex{s}
.}

\end{sloppypar}

Let \math E be a finite set of equations and \X\ a finite subset of \V\@.
A substitution \math{\sigma}
\linebreak
\math{\in\SUBST\V\vt} is 
called\emph{a unifier for \math E} \udiff\ \bigmath{E\sigma\subseteq\id.}
Such a unifier is called\emph{most general on \X} \udiff\
for each unifier \math\mu\ for \math E there is some \math{\tau\in\SUBST\V\vt}
such that \bigmath{\domres{\inpit{\sigma\tau}}\X=\domres\mu\X.}
If \math E has a unifier, then it also has
a most general unifier% 
\footnote{For 
    this most general unifier \math\sigma\ we could, as usual, even require
    \bigmath{\sigma\sigma=\sigma} 
    but
    \emph{not}
    \bigmath{\VAR{\sigma[\VAR{E}]}\ \subseteq\ \VAR{E}.}%
}
on \X, denoted by \minmgu E\X.
\nopagebreak

\vfill

\pagebreak
%%%%%%%%%%%%%%%%%%%%%%%%%%%%%%%%%%%%%%%%%%%%%%%%%%%%%%%%%%%%%%%%%%%%%%%%%%%%%%%
\subsection{Relations}
\label{sect relations}

\yestop
\noindent
Let \math{\X\tightsubseteq\V}\@. 
Let \math{{\rm T}\subseteq\vt}\@. 
A relation $R$ on \vt\ is called:\footroom 
\begin{quote}
\notop
  {\em sort-invariant}
    \udiff\ 
    $\forall(t,t')\tightin R       \stopq
     \exists s    \tightin\sigsorts\stopq 
        t,t'\tightin\vtsigindex{s}$

  {\em \stable\ (\wrt\ substitution)}
    \udiff\ 
    $\forall(t_0,\ldots,t_{n-1})\tightin R     \stopq
     \forall\sigma              \tightin\Xsubst\stopq$
    \\\LINEmath{(t_0\sigma,\ldots,t_{n-1}\sigma)\in R}

  {\em \math{\rm T}-monotonic} \udiff\ 
    \math{
     \forall(t',t'')\tightin R    \stopq
     \forall t      \tightin\vt   \stopq
     \forall p      \tightin\TPOS t\stopq
    }\\\LINEmath{
      \inparentheses{
         \inparenthesesoplist{
             \exists s\tightin\sigsorts\stopq
                t/p,t',t''\tightin\vtsigindex s
           \oplistund
             \repl t p{t'}\in{\rm T}
         }
       \implies
         \inparenthesesoplist{
             (\repl t p{t'},\repl t p{t''})\in R
           \oplistund
             \repl t p{t''}\in{\rm T}
         }
      }
    }

\end{quote}

\yestop
\noindent
%A {\em reduction ordering on \vt} is a \Vmonotonic, \Vstable, 
%and wellfounded ordering.
The\emph{subterm ordering\/ \subterm\ on \vt}
is the \Vstable\ and wellfounded ordering defined by:
\math{t\subtermeq t'} \udiff\ 
\math{\exists p\tightin\TPOS{t'}\stopq t\tightequal t'/p}.
%A {\em simplification ordering on \vt}\/ is a 
%reduction ordering on \vt\ containing \subterm\@.
A\emph{termination-pair} over\/ \sig/\/\V\ is a pair
\math{(>,\rhd)}
of \Vstable, wellfounded orderings on \vt\ such that
\math > is \Vmonotonic,
\bigmath{\tight >\subseteq\tight\rhd,} and 
\bigmath{\tight\superterm\subseteq\tight\rhd.}
\Cf\ \citewgjsc\ for further theoretical aspects of termination-pairs, 
and \citegeserterminationpair\ for interesting practical examples.
For further details on orderings \cf\ \citedershowitz\@.

\yestop
\noindent
The reflexive, symmetric, transitive, and reflexive \& transitive 
closure of a relation 
\redsimple\ 
will be denoted by \onlyonce, \sym,
\trans, and \refltrans, \resp.%
\footnote{Note that this is actually an abuse of notation
since \math{A^+} now denotes the transitive closure of \math A
as well as the set of nonempty words over \math A
and 
since \math{A^\ast} now denotes the reflexive \& transitive closure of \math A
as well as the set of words over \math A. In our former papers we prefered to
denote different things different but now we have found back to this standard
abuse of notion for the sake of convenient readability, because 
the reader will easily find out what is meant with any application
with the exception of those in the proof of Lemma~\ref{lemmaa}.}
Two terms \math v, \math w are called
\emph{joinable \wrt\ \redsimple} 
\udiff\ \math{v\tight\downarrow w}, 
\ie\ \udiff\ \math{v\refltrans\circ\antirefltrans w}\@. 
They are\emph{strongly joinable \wrt\ \redsimple} 
\udiff\ \math{v\downdownarrow w},
\ie\ \udiff\ 
\math{v\onlyonce\circ\antirefltrans w\onlyonce\circ\antirefltrans v}\@. 
\ \redsimple\ is called \emph{terminating below \math u} \udiff\ there is no
\FUNDEF{s}{\N}{\DOM\redsimple} such that
\bigmath{
  u\tightequal s_0
  \und
  \forall i\tightin\N\stopq
    s_i\redsimple s_{i+1}
.}  
\mycomment{
It is\emph{terminating} \udiff\ it is terminating below all \math u.
\ \redsimple~is said to be\emph{contained 
below \math u
in a relation \math R 
}
\udiff\
\bigmath{
  \forall s,t\stopq
  \inparenthesesinline{
      u\refltrans s\redsimple t
    \ \implies\ 
      (s,t)\tightin R
  }
.}
}

\vfill

\pagebreak

\subsection{The Reduction Relation}

In the definition below we restrict our constructor rules to contain
no non-constructor function symbols, to be extra-variable free,
and to contain no negative literals.
This is important for our approach (\cf\ \lemmaconskeeping,
\lemmaaboutconflu, and \lemmamonotonicinbeta)
and should always be kept in mind when reading the following sections.

\begin{sloppypar}
\begin{definition}[Syntax of CRS]\label{defrules}\label{defrulescontinu}
\\
% The set of\emph{directed equations} is defined by
% \\\linemath{
%   \direqvsig
%   :=
%   \setwith
%      {(t,t')}
%      {\exists s\tightin\sigsorts\stopq
%         t,t'\tightin\tss_{s}
%      }
% .}
 \condlit\ is the set of\emph{condition literals} over 
 the following
 predicate
 symbols on terms from \tss: 
 `\boldequal',~`\boldunequal'~(binary, symmetric, sort-invariant),
 and `\/\Def'~(\unary).
 The terms\/\footnote{To avoid misunderstanding: For a condition list, 
   say ``\bigmath{s\boldequal t,\ u\boldunequal v,\ \Def\,w}'',
   we mean the top level terms \bigmath{s,t,u,v,w\in\tss,} 
   but neither their proper subterms nor the literals 
   ``\math{s\boldequal t}'',
   ``\math{u\boldunequal v}'',
   ``\math{\Def\,w}''
   themselves.} 
 of a list \math C of condition literals 
 are called\emph{condition terms} and their set is denoted by \condterms{C}.
 A  (positive/negative-)\emph{conditional rule system} 
 (\/{\em CRS}\/)\/ \R\ over \sig/\cons/\/\V\ is
 a finite subset of the\emph{set of rules} over\/ \sig/\cons/\/\V,
 which is defined by
\math{\left\{\left.\ \inparentheses{(l,r),\ C}\ \right|\right.}
\\\mbox{}\hfill\math{
\left.\inparenthesesoplist{
    \exists s\tightin\sigsorts\stopq
      l,r\tightin\tss_s
  \oplistund
    C\in(\condlit)^\ast
  \oplistund
    \inparentheses{
        l\in\tcs
      \ \implies\\
        \inparenthesesoplist{
          \{r\} \cup \condterms{C}\nottight{\nottight\subseteq}\tcs 
        \oplistund
          \VAR{\{r\}\cup\condterms{C}}\nottight{\nottight\subseteq}\VAR{l}
        \oplistund
          \forall L\mbox{ in }C\stopq
          \forall u,v          \stopq
            \ L\not=(u\boldunequal v) 
        }
    }
}\right\}
.}
\\
 A rule \bigmath{((l,r),\emptyset)} with an empty condition 
 will be written\/  $l\boldequal r$.
 Note that $l\boldequal r$ differs from $r\boldequal l$ whenever the 
 equation is used as a reduction rule.
 A rule \kurzregel\ with condition $C$ will be written \sugarregel.
 We call $l$ the {\em left-hand side} and 
 $r$ the {\em right-hand side} of the rule \sugarregel.
 A rule is said to be\emph{left-linear} (or else\emph{right-linear}) 
 \udiff\ its left-hand (or else right-hand) side is a linear term.
 A rule \sugarregel\ is said 
 to be\/ {\em extra-variable free} \udiff\ 
 \bigmath{\VAR{\{r\}\cup\condterms{C}}\nottight{\nottight\subseteq}\VAR{l}.}
 The whole CRS\/ \R\ is said to have one of these properties \udiff\ 
 each of its rules has it.
 A rule \sugarregel\ is called a\emph{constructor rule} \udiff\
 its left-hand side is a constructor term, \ie\ \bigmath{l\tightin\tcs.}
\end{definition}
\end{sloppypar}

\yestop
\noindent
In the following example we define the subtraction operation `\minussymbol'
partially (due to a non-complete defining case distinction),
whereas we define a member-predicate `\mbpsymbol' totally on the 
constructor ground terms.
\begin{example}\label{exb}
{\em (continuing Example~\ref{exa})}
\\
Let \math{x,y\in\Vconsindex\nat} and \math{l\in\Vconsindex\lists}.
\\
\math{\R_{\,\rm\ref{exb}}}: \mbox{}
\math{
  \begin{array}[t]{l@{\ \ }c@{\ \ }lcl}
   \minusppnoparentheses x\zeropp&\boldequal&x\\
   \minusppnoparentheses{\spp x}{\spp y}&\boldequal&\minusppnoparentheses x y\\
  \end{array}~~~~
  \begin{array}[t]{|l@{\ \ }c@{\ \ }lcl}
   \mbppp x\nilpp      &\boldequal&\falsepp\\
   \mbppp x{\cnspp y l}&\boldequal&\truepp   &\rs&x\boldequal   y\\
   \mbppp x{\cnspp y l}&\boldequal&\mbppp x l&\rs&x\boldunequal y\\
  \end{array}
}
\end{example}

\pagebreak

\begin{definition}
[Fulfilledness]
\label{def fulfilledness}
\\
A list \math{D\in\Xcondlit^\ast} of condition literals is said to 
be\emph{fulfilled \wrt\ some relation \redsimple} \udiff
\\\LINEmath{\forall u,v\in\vt\stopq
\inparentheses{
\begin{array}{l
              @{\ \ \mbox{\bf(}\ \mbox{(}(}
              c
              @{)\mbox{ in }D\mbox{)}\ \ \Rightarrow\ \ }
              r
              l
              }
    &u\boldequal v  
    &                                 
    &u\tight\downarrow v\ \ \mbox{\bf)}     
    \\\wedge
    &\DEF u        
    &\exists\hat{u}\in\tgcons\stopq        
    &u\refltrans\hat{u}\ \mbox{\bf)}  
    \\\wedge
    &u\boldunequal v
    &\exists\hat{u},\hat{v}\in\tgcons\stopq
    &u\refltrans\hat u\notconflu\hat v\antirefltrans v\ \mbox{\bf)}
    \\
\end{array}
}
.}
\end{definition}

\yestop
\yestop
\yestop
\noindent
To avoid a non-monotonic behaviour of our negative conditions,
we define our reduction relation \redsub\ via a double closure:
First we define \redindex{\RX,\omega} by using the constructor rules only.
Then we define \redindex{\RX,\omega+\omega} via a second closure
including all rules.

\begin{definition}[\redsub]
\label{defred}\mbox{}
\\
Let\/ \R\ be a CRS over \sig/\cons/\/\V\@.
Let\/ \math{\X\tightsubseteq\V}. 
Let\/ \math\prec\ denote the ordering on the ordinal numbers.
For \math{\beta\preceq\omega\tight+\omega} 
and \math{p\in\N_+^\ast} 
the reduction relations
\redindex{\RX,\beta} 
and 
\redindex{\RX,\beta,p}
on \tsigX\ are inductively defined as follows:
For 
\math{ 
  s,t\in\tsigX
}:
\\
\bigmath{s\redindex{\RX,\beta}t}
\udiff\
\bigmath{
  \exists p\tightin\TPOS s\stopq
  s\redindex{\RX,\beta,p}t
.}
\\
For \math{p\in\N_+^\ast}: 
\bigmath{\redindex{\RX,0,p}:=\emptyset.}~~~For 
\math{i\in\N};
\math{ 
  s,t\in\tsigX
}:
\\
\bigmath{s\redindex{\RX,i+1,p}t} \udiff\ 
\math{
  \ 
  \exists 
  \left\langle\arr{{l}
      \kurzregel\tightin\R     
    \\\sigma    \tightin\Xsubst
  }\right\rangle
  \stopq
  \inparenthesesoplist{
    l\in\tcs
    \oplistund
    s/p=l\sigma
    \oplistund
    t=\repl s p{r\sigma}
    \oplistund
    C\sigma\mbox{ is fulfilled \wrt\ }\redindex{\RX,i}
  }
.}\vspace{+0.5em}\\
\bigmath{
  \redindex{\RX,\omega,p}:=\bigcup_{_{i\in\N}}\redindex{\RX,i,p}
.}~~~For
\math{i\in\N};
\math{s,t\in\tsigX}:
\ 
\bigmath{s\redindex{\RX,\omega+i+1,p}t} 
\udiff\\
\math{
  s\redindex{\RX,\omega,p}t  
  \ \oder\ 
  \exists
  \left\langle\arr{{l}
      \kurzregel\tightin\R     
    \\\sigma    \tightin\Xsubst
  }\right\rangle
  \stopq
  \inparenthesesoplist{
    s/p=l\sigma
    \oplistund
    t=\repl s p{r\sigma} 
    \oplistund
    C\sigma\mbox{ is fulfilled \wrt\ }\redindex{\R,\X,\omega+i}   
  }
.}\\
\bigmath{
  \redindex{\RX,\omega+\omega,p}
  :=
  \bigcup_{_{i\in\N}}\redindex{\R,\X,\omega+i,p}
;}
\bigmath{\redsub:=
         \redindex{\RX,\omega+\omega}
}\@.
\end{definition}
%%%%%%%%%%%%%%%%%%%%%%%%%%%%%%%%%%%%%%%%%%%%%%%%%%%%%%%%%%%%%%%%%%%%%%%%%%%%%%

\noindent
We will drop ``\RX'' in \redsub\ and \redindex{\RX,\beta} \etc\ 
when referring to some fixed \RX\@.
%%%%%%%%%%%%%%%%%%%%%%%%%%%%%%%%%%%%%%%%%%%%%%%%%%%%%%%%%%%%%%%%%%%%%%%%%%%%%%

\yestop
\yestop
\yestop
\begin{corollary}
\label{corollary redsubomega is minimum}
\\
\redindex{\RX,\omega}
is the minimum 
(\wrt\ set-inclusion)
of all relations
\bigmath\rightsquigarrow\ 
on \vt\ 
satisfying
for all \math{s,t\in\tsigX}:
\bigmath{s\rightsquigarrow t} if
\mbox{}\hfill\math{ 
  \exists
  \left\langle\arr{{l}
       p        \tightin\TPOS{s}
    \\
      \kurzregel\tightin\R     
    \\
      \sigma    \tightin\Xsubst
    \\
  }\right\rangle
  \stopq
  \inparenthesesoplist{
    l\tightin\tcs
    \oplistund
    s/p\!=\!l\sigma
    \oplistund
    t\!=\!\repl s p{r\sigma}
    \oplistund
    C\sigma\mbox{ is fulfilled \wrt\/ }\rightsquigarrow
  }
.}\nopagebreak
\end{corollary}

\yestop
\begin{lemma}\label{lemma red is minimal}\mbox{}
% Let\/ \R\ be a CRS over \sig/\cons/\/\V\@.
Let\/ \bigmath{S_{_\RX}} be the set of all relations\/ 
\bigmath\rightsquigarrow\ 
on \vt\ 
satisfying
\begin{enumerate}

\noitem
\item
\bigmath{
  (\,\,\rightsquigarrow\ \cap\ (\tgcons\tight\times\vt)\,)
  \ \subseteq\ 
  \redindex{\RX,\omega}
}
\ 
as well as

\item
for all \math{s,t\in\tsigX}:
\\\mbox{}\hfill
\bigmath{s\rightsquigarrow t} if\/ 
\bigmath{ 
  \exists
  \left\langle\arr{{l}
       p        \tightin\TPOS{s}
    \\
      \kurzregel\tightin\R     
    \\
      \sigma    \tightin\Xsubst
    \\
  }\right\rangle
  \stopq
  \inparenthesesoplist{
    s/p\!=\!l\sigma
    \oplistund
    t\!=\!\repl s p{r\sigma}
    \oplistund
    C\sigma\mbox{ is fulfilled \wrt\/ }\rightsquigarrow
  }
.}\nopagebreak

\notop
\end{enumerate}
\headroom
Now \redsub\ is the minimum 
(\wrt\ set-inclusion)
in
\math{S_{_\RX}},
and
\math{S_{_\RX}} is closed under nonempty intersection.
\end{lemma}

\pagebreak

\yestop
\yestop
\yestop
\yestop
\yestop
\begin{corollary}[Monotonicity of \redsimple\ \wrt\ Replacement]
\label{corollary monotonic}\mbox{}\\
\redindex{\R,\X,\beta} (for $\beta\preceq\omega\tightplus\omega$)
and \redsub\ are \monotonic\ as well as\/ 
\math{\refltranssub{[{\rm T}]}}-monotonic for each 
\math{{\rm T}\subseteq\tsigX}.
\notop\halftop
\end{corollary}
\begin{corollary}[Stability of \redsimple]\label{corollary stable}\mbox{}\\ 
\redindex{\R,\X,\beta} (for $\beta\preceq\omega\tightplus\omega$), 
\redsub, and their respective fulfilledness-predicates are \stable.
\end{corollary}

\begin{lemma}%
\label{lemma about conskeeping}%
\label{lemma about sortkeeping}%
\label{lemma about groundconskeeping}%
\mbox{}
For\/ 
\math{\X\subseteq\Y\subseteq\V}:
\\\linemath{
  \forall n\tightin\N     \stopq
  \forall s\tightin\tconsX\stopq
  \forall t               \stopq
  \inparentheses{\!\!
    s\redindexn{n}{\R,\Y}t\ \implies\ 
    (s\redindexn{n}{\R,\Y,\omega}t\in\tconsX
)
  \!\!}\!\!
}
\end{lemma}

\notop\notop\notop\halftop
\begin{lemma}\label{lemma about conflu}\mbox{}
\bigmath{
  \downarrow\cap\ (\,\tcs\times\tcs\,)
  \ \ \subseteq\ \ 
  \downarrowindex{\omega}
}
\end{lemma}

%%%%%%%%%%%%%%%%%%%%%%%%%%%%%%%%%%%%%%%%%%%%%%%%%%%%%%%%%%%%%%%%%%%%%%%%%%%%%%
\begin{lemma}[Monotonicity of \redindex{\beta} 
              and of Fulfilledness \wrt\ \redindex{\beta} in $\beta$]
\label{lemma monotonic wrt ordinals}
\\
For\/ $\beta\preceq\gamma\preceq\omega\!+\!\omega$:
\bigmath{
  \redindex{\beta}
  \ \subseteq\ 
  \redindex{\gamma}
  \ \subseteq\ 
  \redsimple
};~  
and
if\/ $C$ is fulfilled \wrt\ \redindex{\beta}
and 
\\
\bigmath{
         \omega\preceq\beta
         \ \vee\ 
         \forall u,v\stopq
            ((u\boldunequal v)\mbox{ is not in }C)
},~ 
then $C$ is fulfilled \wrt\ \redindex{\gamma} and \wrt\ \redsimple.
\end{lemma}

\noindent
Note that monotonicity of fulfilledness is not given in general for 
\bigmath{\beta\tightprec\omega} 
and a negative literal which may become invalid during the growth of the 
reduction relation on constructor terms.

\yestop
\noindent
For the proofs \cf\ \citewgjsc.

\yestop
\yestop
\yestop

\subsection{The Parallel Reduction Relation}

\yestop
\noindent
The following relation is essential for sophisticated joinability
notions as well as for most of our proofs:

\begin{definition}[Parallel Reduction]
\label{def parallel reduction}
\mbox{}\\
For\/ \math{\beta\preceq\omega\tight+\omega} we define 
the\emph{parallel reduction relation \redparaindex{\RX,\beta}} on 
\tsigX\footroom:
\\
\bigmath{s\redparaindex{\RX,\beta} t}
\udiff\ 
\bigmath{
  \exists\,\Pi\subseteq\TPOS s\stopq
    s\redparaindex{\RX,\beta,\Pi}t
,}
where
\\
\bigmath{s\redparaindex{\RX,\beta,\Pi}t}
\udiff\ 
\bigmath{
    \inparenthesesoplist{
      \forall p,q\tightin\Pi\stopq
        \inparentheses{
          p\tightequal q\ \oder\ \neitherprefix p q
        }
      \oplistund 
      t\tightequal\replpar s p{t/p}{p\tightin \Pi}
      \oplistund 
      \forall p\tightin \Pi\stopq
           s/p
           \redindex{\RX,\beta}
           t/p
    }
.}
\end{definition}

\begin{corollary}
\label{corollary parallel one}
\mbox{}
\bigmath{
  \forall\beta\tightpreceq\omega\tight+\omega\stopq
    \redindex{\RX,\beta} 
    \subseteq
    \redparaindex{\RX,\beta} 
    \subseteq
    \refltransindex{\RX,\beta} 
.}
\end{corollary}

\mycomment{%%%%%%%%%%%%%%%%%%%%%%%%%%%%%%%%%%%%%%%%%%%%%%%%%%
\noindent
The following is a corollary of \lemmamonotonicinbeta:

\notop
\begin{corollary}
\label{corollary parallel two}
\mbox{}
\bigmath{
  \forall\gamma\preceq\omega\tight+\omega\stopq
  \forall\beta\preceq\gamma\stopq
  \ \redparaindex{\RX,\beta}\subseteq\redparaindex{\RX,\gamma} 
.}
\end{corollary}
}%comment%%%%%%%%%%%%%%%%%%%%%%%%%%%%%%%%%%%%%%%%%%%%%%%%%%

%%%%%%%%%%%%%%%%%%%%%%%%%%%%%%%%%%%%%%%%%%%%%%%%%%%%%%%%%%%%%%%%%%%%%%%%%%%%%%%
\vfill

\pagebreak

\section{Confluence}

The following notions and lemmas have become folklore, 
\cf\ \eg\ \citeklopdiss\ or \citehuet\ for more information.

\begin{definition}[Commutation and Confluence]

\noindent
Two relations \redindex0 and \redindex1 are \emph{commuting} \udiff\
\\\linemath{
  \forall s,t_0,t_1\stopq
  \inparentheses{
    t_0\antirefltransindex0 s\refltransindex1 t_1
    \ \implies\ \ 
    t_0\refltransindex1\circ\antirefltransindex0 t_1
  }
.}
\redindex0 and \redindex1 are \emph{locally commuting} \udiff\
\\\LINEmath{
  \forall s,t_0,t_1\stopq
  \inparentheses{
    t_0\antiredindex0 s\redindex1 t_1
    \ \implies\ \ 
    t_0\refltransindex1\circ\antirefltransindex0 t_1
  }
.}
\\
\redindex1\emph{strongly commutes over} \redindex0 \udiff\
\\\LINEmath{
  \forall s,t_0,t_1\stopq
  \inparentheses{
    t_0\antiredindex0 s\redindex1 t_1
    \ \implies\ 
    t_0\onlyonceindex 1\circ\antirefltransindex0 t_1
  }
.}
\mycomment{%%%%%%%%%%%
\\
\redindex0 and \redindex1 are \emph{very strongly commuting} \udiff\
\\\LINEmath{
  \forall s,t_0,t_1\stopq
  \inparentheses{
    t_0\antiredindex0 s\redindex1 t_1
    \ \implies\ \ 
    t_0\onlyonceindex1\circ\antionlyonceindex0 t_1
  }
.}
}%%%%%%%%%%%

\yestop
\noindent
\begin{diagram}
  s&\rrefltransindex1&t_1
&&s&\rredindex      1&t_1
&&s&\rredindex      1&t_1
\\
  \drefltransindex0&&\drefltransindex0
&&\dredindex      0&&\drefltransindex0
&&\dredindex      0&&\drefltransindex0
\\
  t_0&\rrefltransindex1&\circ
&&t_0&\rrefltransindex1&\circ
&&t_0&\ronlyonceindex1 &\circ
\\
  &\mbox{\begin{tabular}[t]{c}
           \redindex0 and \redindex1 are
           \\
           commuting
         \end{tabular}}&
&&&\mbox{\begin{tabular}[t]{c}
           \redindex0 and \redindex1 are
           \\
           locally commuting
         \end{tabular}}&
&&&\mbox{\begin{tabular}[t]{c}
           \redindex1 strongly com-
           \\
           mutes over \redindex0
         \end{tabular}}&
\end{diagram}

\yestop
\noindent
A single relation \redsimple\ is called\emph{{[\math{\;\!}locally]} confluent}
\udiff\ \redsimple\ and \redsimple\ are {[locally]} commuting.
It is called\emph{strongly confluent} 
\udiff\ \redsimple\ strongly commutes over \redsimple\@.
It is called\emph{confluent below $u$} \udiff\ 
\bigmath{
  \forall v,w\stopq
     \inparenthesesinline{
       v\antirefltrans u\refltrans w
       \ \implies\ 
       v\tight\downarrow w
     }
.}
\end{definition}

\yestop
\begin{lemma}[Generalized \newmanlemma]\label{lemma commutation copy}\\
If\/ \redindex0 and \redindex1 are commuting, then
they are locally commuting, too.
\\
Furthermore, if \bigmath{\redindex0\cup\redindex1} is terminating
or if \redindex0 or \redindex1 is transitive, then 
also the converse is true, \ie\/ \redindex0 and \redindex1 are commuting
\uiff\ they are locally commuting. 
\end{lemma}

\yestop
\begin{lemma}
\label{lemma strong commutation one copy}
\\
The following three properties are logically equivalent:
\begin{enumerate}

\notop
\item
\redindex1 strongly commutes over \redindex0.

\noitem
\item
\redindex1 strongly commutes over \transindex0.

\noitem
\item
\redindex1 strongly commutes over \refltransindex0.

\notop
\end{enumerate}
Moreover, each of them implies that
\redindex0 and \redindex1 are commuting.
\end{lemma}

\yestop
\begin{lemma}[\church-\rosser]
\label{lemma church rosser}
\\
Assume that \redsimple\ is confluent. Now: 
\bigmath{\congru\subseteq\tight\downarrow.}
\end{lemma}

\vfill

\pagebreak

\yestop
\yestop
\yestop
\noindent
Besides strong confluence there are two other important versions of
strengthened confluence for conditional rule systems.
They are based on the 
depth of the reduction steps, \ie\ on the \math\beta\ of \redindex{\RX,\beta}.
Therefore they actually are properties of \RX\ instead of \redsub,
unless one considers \redsub\ to be the family 
\bigmath{(\redindex{\RX,\beta})_{\beta\preceq\omega+\omega}.}
These two strengthened versions of confluence are\emph{shallow confluence} 
and\emph{level confluence}.
Their generalizations to our generalized framework here are 
called%
\emph{\math 0-shallow confluence}
for the closure \wrt\ our constructor rules,
as well as%
\emph{\math\omega-shallow confluence}
and%
\emph{\math\omega-level confluence} 
for our second closure.
Shallow and level confluence are interesting:
On the on hand, they provide us with stronger induction hypotheses for the
proofs of our confluence criteria.
On the other hand, the stronger confluence properties may be essential for
certain kinds of reasoning with the specification of a rule system;
for level joinability \cf\ \citemiddeldorpbasicnarrowing.

\yestop
\noindent
Before we define our notions of shallow and level confluence 
we present some operations on ordinal numbers:

\begin{definition}[\pluszero, \plusomega, \monus]
\\
Let \math{\alpha\in\{0,\omega\}}.
Let `\math+' be the addition of ordinal numbers.
\\
Define `\/\pluszero', `\/\plusomega',  
and `\/\monus' for\/ 
\math{n_0,n_1\prec\omega}:
\\\LINEmath{
 \arr{[t]{lll}
  0{\plusalpha}n_1
  &:=&
  n_1
  \\
  n_0{\plusalpha}0
  &:=&
  n_0
  \\
  (n_0\tight+1){\plusalpha}(n_1\tight+1)
  &:=&
  \alpha+n_0\tight+1+n_1\tight+1
  \\ 
  \\ 
  (n_0\tight+n_1)\monus n_1
  &:=&
  n_0
  \\
  n_0\monus(n_0\tight+n_1)
  &:=&
  0
  \\
  \\
 }
}

\noindent
Note that the subscript of the operator `\plusomega' 
is chosen to remind that it adds an extra \math\omega\
to the left if both arguments are different from \math 0.
Moreover, note that 
\bigmath{
  \domres\pluszero{\N\times\N}
  \tightequal
  \domres{\tight+}{\N\times\N}
.}
\ `\monus'~is sometimes called\emph{monus}.
\end{definition}

\yestop
\noindent
Since we want to use shallow and level confluence also
for terminating reduction relations 
we have to parameterize them \wrt\ wellfounded orderings.
Let\/ `\math\succ' as before be the wellordering of the ordinal numbers.
Let\/ `\math\rhd' be some wellfounded ordering on \vt.
We denote the lexicographic combination of \math\succ\ and \math\rhd\ 
by `\succrhd',
its reverse by `\preclhd',
and the reflexive closure of the latter by
`\preclhdeq'.

\vfill

\pagebreak

\begin{definition}[ \math 0-Shallow Confluent / \math\omega-Shallow Confluent ]
\label{def omega shallow confluent no termination}
\\
Let\/ \math{\alpha\in\{0,\omega\}}.
Let\/ \math{\beta\preceq\omega\tight+\omega.}  
Let\/ \math{s\in\vt}.
\\
\RX\ is said to 
be\emph{\math\alpha-shallow confluent 
up to \math\beta\ and \math s in \math\lhd}
\udiff\
\\\LINEmath{
        \forall n_0,n_1\tightprec\omega\stopq
        \forall u,v,w\stopq
        \inparentheses{
          \inparenthesesoplist{
            \inparentheses{
              n_0\tight{\plusalpha}n_1
              ,\ 
              u
            }
            \preclhdeq
            \inparentheses{
              \beta
              ,\
              s
            }
            \oplistund 
              v
              \antirefltransindex{\RX,\alpha+n_0}
              u
              \refltransindex    {\RX,\alpha+n_1}
              w
          }
          \\\implies\
            v
              \refltransindex    {\RX,\alpha+n_1}
            \circ
              \antirefltransindex{\RX,\alpha+n_0}
            w
          }.
}\\
\headroom
\RX\ is said to be\emph{\math\alpha-shallow confluent up to \math\beta} \udiff\
\\\linenomath{
  \RX\ is \math\alpha-shallow confluent 
  up to \math\beta\ and\footnote{\label{footnote lhd does not matter}Note
  that reference to a special \math\lhd\ becomes irrelevant here} \math s 
  for all \math{s\in\vt}.
}
\RX\ is said to be\emph{\math\alpha-shallow confluent} \udiff\
  \RX\ is \math\alpha-shallow confluent up to \math{\omega\tight+\alpha}.
\end{definition}

\begin{definition}[\math\omega-Level Confluent]%
\label{def omega level confluent no termination}
\\
Let\/ \math{\beta\preceq\omega}.  
Let\/ \math{s\in\vt}.  
\RX\ is said to 
be\emph{\math\omega-level confluent up to \math\beta\ and \math s in \math\lhd}
\udiff\
\\\LINEmath{
        \forall n_0,n_1\tightprec\omega\stopq
        \forall u,v,w\stopq
        \inparentheses{
          \inparenthesesoplist{
            \inparentheses{
              \max\{n_0,n_1\} 
              ,\ 
              u
            }  
            \preclhdeq
            \inparentheses{
              \beta
              ,\ 
              s 
            }  
            \oplistund 
              v
              \antirefltransindex{\RX,\omega+n_0}
              u
              \refltransindex    {\RX,\omega+n_1}
              w
          }
          \\\implies\
            v
              \downarrowindex{\RX,\omega+\max\{n_0,n_1\}}
            w
        }.
}\\
\headroom
\RX\ is said to be\emph{\math\omega-level confluent up to \math\beta} \udiff\
\\\linenomath{
  \RX\ is \math\omega-level confluent up to \math\beta\ 
  and\repeatfootnotemark{footnote lhd does not matter} \math s
    for all \math{s\in\vt}.
}
\RX\ is said to be\emph{\math\omega-level confluent} \udiff\
  \RX\ is \math\omega-level confluent up to \math{\omega}.
\end{definition}

\noindent
Note that \math\omega-level and \math\omega-shallow confluence 
specialize to the standard definitions 
of level and shallow confluence, \resp, for the
case that all symbols are considered to be non-constructor symbols 
(where \math n becomes the standard depth of \redindex{\RX,\omega+n});
and that \math 0-shallow confluence specializes to the standard definition 
of shallow confluence for the
case that all symbols are considered to be constructor symbols.

\begin{corollary}
[ \math\omega-Shallow Confluent 
  \mbox\implies\ 
  \math\omega-Level Confluent
  \mbox\implies\ 
  Confluent
]
\label{corollary omega shallow confluent implies confluent no termination}
\mbox{}\\
If\/ \RX\ is \math\omega-shallow confluent,
then\/ \RX\ is \math\omega-level confluent.
\\
If\/ \RX\ is \math\omega-level confluent,
then\/ \redsub\ is confluent.
\end{corollary}

\begin{corollary}
\label{corollary omega shallow confluent up to omega no termination}
\mbox{}\\
\RX\ is \math\omega-shallow confluent up to \math{0} \uiff\\ 
\RX\ is \math\omega-level confluent up to \math{0} \uiff\\ 
\redindex{\RX,\omega} is confluent.
\end{corollary}

\vfill

\pagebreak

\section{Critical Peaks}

\noindent
Critical peaks describe those possible sources of non-confluence 
that directly arise from the syntax of the given rule system. 
While the so-called\emph{variable overlaps} 
can hardly be approached via syntactic means, 
the critical peaks describe
the non-variable overlaps resulting from an instantiated left-hand side
being subterm of an instantiated left-hand side at a non-variable 
position.
Our critical peaks
capture more information than the standard\emph{critical pairs}:
Besides the pair, they contain the peak term and its overlap position.
Furthermore, each element of the pair is augmented with the condition that must
be fulfilled for enabling the reduction step down from the peak term, and with
a bit indicating whether the rule applied was a non-constructor rule or not.

\begin{definition}[Critical Peak]
\\
If the left-hand side of a rule
\bigmath{l_0\boldequal r_0\rulesugar C_0}
and 
\\
the subterm at non-variable (\ie\ \math{l_1/p\notin\V}) 
position \math{p\in\TPOS{l_1}}
\\
of the left-hand side of a rule 
\bigmath{l_1\boldequal r_1\rulesugar C_1}
\\
(assuming 
\bigmath{\VAR{\sugarregelindex0}\cap\VAR{\sugarregelindex1}=\emptyset} \wrog
\footnote{To achieve this, let \math{\xi\in\SUBST\V\V} be a bijection with 
\bigmath{
  \xi[\VAR{\sugarregelindex0}]\cap\VAR{\sugarregelindex1}
  =
  \emptyset
} and then replace \sugarregelindex0 with \bigmath{(\sugarregelindex0)\xi.}
})
are unifiable by 
\\\linemath{
    \headroom\footroom
    \sigma
    =
    \minmgu{\ \{(l_0,l_1/p)\}}
           {\ \VAR{l_0\boldequal r_0\rulesugar C_0,
                   l_1\boldequal r_1\rulesugar C_1}},
}
if 
(for \math{i\tightprec2}) 
\bigmath{
  \ \ 
  \Lambda_i
  =
  \left\{\arr{{ll}0&\mbox{ if }l_i\in\tcs\\1&\mbox{ otherwise}\\}\right\},
} 
\\
and if the resulting critical pair is non-trivial
(\ie\ \math{\repl{l_1}p{r_0}\sigma\not=r_1\sigma}),
then 
\\\linemath{\headroom\footroom
  \inparentheses{ 
   (\repl{l_1}p{r_0},\ 
    C_0,\ 
    \Lambda_0),\ \ 
   (r_1,\ 
    C_1,\ 
    \Lambda_1),\ \ 
   l_1,\ \  
   \sigma,\ \  
   p
  }
}
is a 
(non-trivial)\emph{critical peak (of the form \math{(\Lambda_0,\Lambda_1)})}
consisting of the
conditional critical pair, 
its peak term \math{l_1}, the most general unifier \math{\sigma}, 
and the overlap position \math p.

\noindent
For convenience we usually identify this critical peak with its instantiated 
version
\\\linemath{\headroom\footroom
  \inparentheses{ 
   (\repl{l_1}p{r_0}\sigma,\ 
    C_0\sigma,\ 
    \Lambda_0),\ \ 
   (r_1\sigma,\ 
    C_1\sigma,\ 
    \Lambda_1),\ \ 
   l_1\sigma,\ \  
   p
  }
}
which should not lead to confusion because the tuple is shorter.

\noindent
The set of all critical peaks of a CRS\/ \R\ is denoted by \math{{\rm CP}(\R)}.
\end{definition}

\begin{example}
\label{ex cp}
{\em (continuing Example~\ref{exb})}
\\
\math{{\rm CP}(\R_{\,\rm\ref{exb}})} contains two critical peaks,
namely (in the instantiated version)
\\\inparentheses{
  (\truepp     ,(x\boldequal   y),1),\
  (\mbppp x l,(x\boldunequal y),1),\
  \mbppp x{\cnspp y l}            ,\
  \emptyset
} 
and
\\\inparentheses{
  (\mbppp x l,(x\boldunequal y),1),\
  (\truepp     ,(x\boldequal   y),1),\
  \mbppp x{\cnspp y l}            ,\
  \emptyset
}
\\
which we would (partially) display as
\begin{diagram}
\mbppp x{\cnspp y l}&&\rred&\mbppp x l
&&&&
\mbppp x{\cnspp y l}&&\rred&\truepp
\\
\dredindex{\dots,\,\emptyset}
&&&
&&&&\dredindex{\dots,\,\emptyset}
\\
\truepp 
&&&
&&&&\mbppp x l
\\
\end{diagram}
Note that we omit the position at the arrow to the right because it is always
\math\emptyset.
Furthermore, 
note that the two critical peaks are different although they look similar.
Namely, the one is the symmetric overlay (\cf\ below) of the other.
\end{example}

\pagebreak

\section{Basic Forms of Joinability of Critical Peaks}
\label{sect basic joinability}

%\noindent
%A rule system \R\ is said to be\emph{overlapping} 
%\udiff\ 
%\math{{\rm CP}(\R)\tightnotequal\emptyset}.
%
%\yestop
\noindent
A critical peak 
\footroom\\\linemath{\headroom\footroom\criticalpeaklongform}\headroom
is\emph{joinable \wrt\ \RX}
(for
\math{\X\tightsubseteq\V}) 
\udiff\ \bigmath{\forall\varphi\tightin\Xsubst\stopq}
\\\LINEmath{
  \inparentheses{
          ((D_0 D_1)\sigma\varphi\mbox{ fulfilled \wrt\ }\redsub)
          \implies t_{0}\sigma\varphi\downarrowsub t_{1}\sigma\varphi
  }
.}

\noindent
It is an\emph{overlay} \udiff\ \math{p\tightequal\emptyset}. \mbox{}
It is a\emph{non-overlay} \udiff\ \math{p\tightnotequal\emptyset}.
\\
It is\emph{overlay joinable \wrt\ \RX} \udiff\ it is joinable \wrt\ \RX\ 
and is an overlay.

\noindent
In the following two definitions `\truepp' and `\falsepp' denote two 
arbitrary irreducible ground terms. Their special names have only been
chosen to make clear the intuition behind.

\noindent 
The above critical peak is\emph{complementary \wrt\ \RX} \udiff\
\\\LINEmath{
  \inparenthesesoplist{ 
      \exists u,v\tightin\vt\stopq
      \exists i\tightprec 2\stopq       
      \inparenthesesoplist{ 
          (u\boldequal   v)\mbox{ occurs in }D_{i  }\sigma
        \oplistund
          (u\boldunequal v)\mbox{ occurs in }D_{1-i}\sigma
      }
    \oplistoder
      \exists p\tightin\vt\stopq
      \exists\truepp,\falsepp\tightin\gt\tightsetminus\DOM\redsub\stopq
      \exists i\tightprec 2\stopq       
      \inparenthesesoplist{ 
          (p\boldequal\truepp )\mbox{ occurs in }D_{i  }\sigma
        \oplistund
          (p\boldequal\falsepp)\mbox{ occurs in }D_{1-i}\sigma
        \oplistund
          \truepp\tightnotequal\falsepp
      }
  }
.}
\\
It is\emph{weakly complementary \wrt\ \RX} \udiff\
\\\LINEmath{
  \inparenthesesoplist{ 
      \exists u,v\tightin\vt\stopq
      \inparentheses{ 
          (u\boldequal   v)\mbox{ and }
        \\
          (u\boldunequal v)\mbox{ occur in }(D_0 D_1)\sigma
      }
    \oplistoder
      \exists p\tightin\vt\stopq
      \exists\truepp,\falsepp\tightin\gt\tightsetminus\DOM\redsub\stopq
      \inparenthesesoplist{ 
          (p\boldequal\truepp )\mbox{ and }
        \oplistnl
          (p\boldequal\falsepp)\mbox{ occur in }(D_0 D_1)\sigma
        \oplistund
          \truepp\tightnotequal\falsepp
      }
  }
.}

\noindent
It is\emph{strongly joinable \wrt\ \RX} \udiff\ 
\bigmath{\forall\varphi\tightin\Xsubst\stopq}
\\\LINEmath{
  \inparentheses{
          ((D_0 D_1)\sigma\varphi\mbox{ fulfilled \wrt\ }\redsub)
          \implies 
          t_{0}\sigma\varphi\downdownarrowsub t_{1}\sigma\varphi
  }
.}
%\\
%It is\emph{$\rhd$-weakly joinable \wrt\ \RX} \udiff\
%\bigmath{\forall\varphi\tightin\Xsubst\stopq}
%\\\linemath{
%     \inparentheses{
%          \inparenthesesoplist{
%             (D_0 D_1)\sigma\varphi\mbox{ fulfilled \wrt\ }\redsub
%            \oplistund
%             \forall u\stopq
%             \inparentheses{
%                u\lhd\hat{t}\sigma\varphi
%                \ \implies\ 
%                \redsub\mbox{ is confluent below }u
%             }
%          }
%        \implies\  
%        t_{0}\sigma\varphi\downarrowsub t_{1}\sigma\varphi
%        \\
%     }
%.}

\noindent
In the following definition `\math A' is an arbitrary function from
positions to sets of terms.
%, which is intended to serve as an 
%optional interface that is to be specified by the confluence criteria
%using it.

\noindent
The above critical peak 
is\emph{$\rhd$-weakly joinable \wrt\ \RX\ {[besides \math A]}}
\udiff\
\mbox{\math{
  \ 
  \forall\varphi\tightin\Xsubst\stopq
}}\\\LINEmath{
     \inparentheses{\!\!
          \inparenthesesoplist{
              (D_0 D_1)\sigma\varphi\mbox{ fulfilled \wrt\ }\redsub
            \oplistund
              \forall u\stopq
              \inparentheses{
                u\lhd\hat{t}\sigma\varphi
                \ \implies\ 
                \redsub\mbox{ is confluent below }u
              }
            \oplistund
              \forall x\tightin\V\stopq
                x\varphi\tightnotin\DOM\redsub
            \oplistund
              \inparentheses{
                  p\tightnotequal\emptyset
                \ \implies\ 
                  \forall x\tightin\VAR{\hat t}\stopq
                    x\sigma\varphi\tightnotin\DOM\redsub
              }
            \\
            \multicolumn{2}{@{}l@{}}{
            \left[\begin{array}{@{\wedge\ \ \ }l@{\ \ \ \ }}
               \hat t\sigma\varphi
               \tightnotin
               A(p)
            \end{array}\right]}
          }
        \implies\ 
        t_{0}\sigma\varphi\downarrowsub t_{1}\sigma\varphi
        \\
     }
.}

\yestop
\noindent
Note that \math\rhd-weak joinability adds several useful features to the
single condition of joinability, forming a conjunctive condition list.
The first new feature allows to assume confluence below all terms that
are strictly smaller than the peak term. The following features allow us to 
assume some irreducibilities for the joinability test, where the optional one 
is an interface that is to be specified by the confluence criteria using it, 
\cf\ our theorems \ref{theoremconfluence} and \ref{theorem quasi-free three}.
For a demonstration of the usefulness of these additional features \cf\
Example~\ref{example integers}.

\yestop
\begin{lemma}[Joinability of Critical Peaks is Necessary for Confluence]
\label{lemma necessary}
\\
If\/ \redsub\ is confluent, then all critical peaks in 
${\rm CP}(\R)$ are joinable \wrt\ \RX\@.
\end{lemma}
\yestop

\pagebreak

\section{Basic Forms of Shallow and Level Joinability}

\noindent
Just like confluence and strong confluence, also level and shallow
confluence have their corresponding joinability notion. Sorry to say,
they are pretty complicated, however.

\begin{definition}[ 
\math     0-Shallow Joinable /
\math\omega-Shallow Joinable ]
\label{def shallow joinable no termination}\mbox{}
\\
Let\/ \math{\alpha\in\{0,\omega\}}.
Let\/ \math{\beta\preceq\omega\tight+\alpha}.
Let\/ \math{s\in\vt}.
A critical peak \criticalpeaklongform\
\\ 
is\emph
{\math\alpha-shallow joinable up to \math\beta\ and \math s 
 \wrt\ \RX\ and \math\lhd\ {[besides \math A]}} 
\udiff\\
\math{
  \forall\varphi\tightin\Xsubst\stopq
  \forall n_0,n_1\prec\omega\stopq
}
\\\LINEmath{
  \inparenthesesoplist{
    \inparenthesesoplist{
         \inparentheses{
           n_0\plusalpha n_1
           ,\ 
           \hat t\sigma\varphi
         }
         \preclhdeq
         \inparentheses{
           \beta
           ,\ 
           s
         }
       \oplistund
         \forall i\prec 2\stopq
         \inparenthesesoplist{
             \inparentheses{
                 \alpha\tightequal 0
               \ \implies\ 
                 \Lambda_i\tightequal 0\tightprec n_i
             }
           \oplistund
             \inparentheses{
                 \alpha\tightequal\omega
               \ \implies\ 
                 \Lambda_i\tightpreceq n_i
             }
           \oplistund
             D_i\sigma\varphi\mbox{ fulfilled \wrt\ }
             \redindex{\RX,\alpha+(n_i\monus 1)}
         }
       \oplistund
         \forall
           \inparentheses{\delta,\ s'}
           \tightpreclhd 
           \inparentheses{n_0\plusalpha n_1,\ \hat t\sigma\varphi}
         \stopq
           \inparentheses{\begin{array}{@{}l@{}}
             \RX\mbox{ is \math\alpha-shallow confluent}
              \\\mbox{up to }\delta
                \mbox{ and }s' 
                \mbox{ in }\lhd\\\end{array}
           }
       \oplistund
         \forall x\tightin\V\stopq
           x\varphi\tightnotin\DOM{\redindex{\RX,\alpha+\min\{n_0,n_1\}}}
       \oplistund
         \inparentheses{
             p\tightnotequal\emptyset
           \ \implies\ 
             \forall x\tightin\VAR{\hat t}\stopq
               x\sigma\varphi
               \tightnotin
               \DOM{\redindex{\RX,\alpha+\min\{n_0,n_1\}}}
         }
       \\
       \multicolumn{2}{@{}l@{}}{
       \left[\begin{array}{@{\wedge\ \ \ }l@{\ \ \ \ }}
          \hat t\sigma\varphi
          \tightnotin
          A(p,\min\{n_0,n_1\})
       \end{array}\right]}
    }
    \oplistimplies
    \inparentheses{
             t_0\sigma\varphi
             \refltransindex{\RX,\alpha+n_1}
             \circ
             \antirefltransindex{\RX,\alpha+n_0}
             t_1\sigma\varphi
    }
  }
.}
\\\headroom
It is 
called\emph{\math\alpha-shallow joinable up to \math\beta\ \wrt\ \RX\ 
and \math\lhd\ {[besides \math A]}}  
\udiff\\
\LINEnomath{
  it is \math\alpha-shallow joinable up to  
  \math\beta\ and \math s \wrt\ \RX\ and \math\lhd\ 
  {[besides \math A]}
  for all \math{s\in\vt}.
}
\\
It is called\emph{\math\alpha-shallow joinable \wrt\ \RX\ and 
\math\lhd\ {[besides \math A]}}  
\udiff\\
\LINEnomath{
  it is \math\alpha-shallow joinable
  up to  \math{\omega\tight+\alpha} \wrt\ \RX\ and \math\lhd\ 
  {[besides \math A]}.
}
\\
When \math\lhd\ is not specified, we tacitly assume it to be \subterm.
\end{definition}

\begin{definition}[\math\omega-Level Joinable]%
\label{def level joinable}\mbox{}
\\
Let\/ \math{\beta\preceq\omega}.
Let\/ \math{s\in\vt}.
A critical peak \criticalpeaklongform\
\\ 
is\emph
{\math\omega-level joinable up to \math\beta\ and \math s 
 \wrt\ \RX\ and \math\lhd\ {[besides \math A]}} 
\udiff\\
\math{
  \forall\varphi\tightin\Xsubst\stopq
  \forall n_0,n_1\prec\omega\stopq
}
\\\LINEmath{
  \inparenthesesoplist{
    \inparenthesesoplist{
         \inparentheses{
           \max\{n_0,n_1\}
           ,\ 
           \hat t\sigma\varphi
         }
         \preclhdeq
         \inparentheses{
           \beta
           ,\ 
           s
         }
       \oplistund
         \forall i\prec 2\stopq
         \inparenthesesoplist{
             \Lambda_i\preceq n_i
           \oplistund
             D_i\sigma\varphi\mbox{ fulfilled \wrt\ }
             \redindex{\RX,\omega+(n_i\monus 1)}
         }
       \oplistund
         \forall
           \inparentheses{\delta,\ s'}
           \tightpreclhd 
           \inparentheses{\max\{n_0,n_1\},\ \hat t\sigma\varphi}
         \stopq
           \inparentheses{\begin{array}{@{}l@{}}
             \RX\mbox{ is \math\omega-level confluent}
              \\\mbox{up to }\delta
                \mbox{ and }s' 
                \mbox{ in }\lhd\end{array}
           }
       \oplistund
         \forall x\tightin\V\stopq
           x\varphi\tightnotin\DOM{\redindex{\RX,\omega+\max\{n_0,n_1\}}}
       \oplistund
         \inparentheses{
             p\tightnotequal\emptyset
           \ \implies\ 
             \forall x\tightin\VAR{\hat t}\stopq
               x\sigma\varphi
               \tightnotin
               \DOM{\redindex{\RX,\omega+\max\{n_0,n_1\}}}
         }
       \\
       \multicolumn{2}{@{}l@{}}{
       \left[\begin{array}{@{\wedge\ \ \ }l@{\ \ \ \ }}
          \hat t\sigma\varphi
          \tightnotin
          A(p,\max\{n_0,n_1\})
       \end{array}\right]}
    }
    \oplistimplies 
    \inparentheses{
             t_0\sigma\varphi
             \downarrowindex{\RX,\omega+\max\{n_0,n_1\}}
             t_1\sigma\varphi
    }
  }
.}
\\\headroom
It is 
called\emph{\math\omega-level joinable up to \math\beta\ \wrt\ \RX\ 
and \math\lhd\ {[besides \math A]}}  
\udiff\\
\LINEnomath{
  it is \math\omega-level joinable up to  
  \math\beta\ and \math s \wrt\ \RX\ and \math\lhd\ 
  {[besides \math A]} 
  for all \math{s\in\vt}.
}
\\
It is called\emph{\math\omega-level joinable \wrt\ \RX\ and \math\lhd\ 
{[besides \math A]}}  
\udiff\\
\LINEnomath{
  it is \math\omega-level joinable
  up to  \math{\omega} \wrt\ \RX\ and \math\lhd\ {[besides \math A]}.
}
\\
When \math\lhd\ is not specified, we tacitly assume it to be \subterm.
\end{definition}

\pagebreak

\begin{sloppypar}
\noindent
Please notice the generic structure of these and the following
definitions that makes them
actually less complicated than they look like.
While the conclusions of their implications should be clear, the elements
of their conjunctive condition lists have the following purposes:
The first just parameterizes the notion in \math\beta\ and \math s.
The second requires the appropriate fulfilledness of the conditions of the
critical peak, where \bigmath{\Lambda_i\tightpreceq n_i} allows us
to assume
\bigmath{1\tightpreceq n_i} when the term \math{t_i}
is generated by a non-constructor rule
which is important since otherwise the conclusion is very unlikely
to be fulfilled, \cf\ also below.
The third allows us to assume a certain confluence property
which can be applied when checking the fulfilledness of the conditions.
\Eg, this condition sometimes implies that the fulfilledness assumptions
of the second element for 
``\math{i\tightequal 0}''
and 
``\math{i\tightequal 1}''
are contradictory.
An example for this are the critical peaks of Example~\ref{ex cp}
which are 
both \math\omega-level and \math\omega-shallow confluent since the condition
list can never be fulfilled. But how do we know that? 
Suppose that
\bigmath{
  (x\boldequal y)\varphi
}
is fulfilled \wrt\ \redindex{\RX,\omega+(n_0\monus 1)}
and that
\bigmath{
  (x\boldunequal y)\varphi
}
is fulfilled \wrt\ \redindex{\RX,\omega+(n_1\monus 1)}.
Then there are \math{\hat u,\hat v\in\tgcons} such that 
\bigmath{
  x\varphi
  \downarrowindex{\RX,\omega+(n_0\monus 1)}
  y\varphi
}
and 
\bigmath{
  x\varphi
  \refltransindex{\RX,\omega+(n_1\monus 1)} 
  \hat u
  \notconfluindex{\RX,\omega+(n_1\monus 1)}
  \hat v
  \antirefltransindex{\RX,\omega+(n_1\monus 1)} 
  y\varphi
.}
By \bigmath{x,y\tightin\Vcons} we get 
\bigmath{x\varphi,y\varphi\tightin\tcc} and thus by \lemmaconskeeping\
we get 
\bigmath{
  x\varphi
  \downarrowindex{\RX,\omega}
  y\varphi
}
and 
\bigmath{
  x\varphi
  \refltransindex{\RX,\omega}
  \hat u
  \notconfluindex{\RX,\omega}
  \hat v
  \antirefltransindex{\RX,\omega}
  y\varphi
.}
This contradicts confluence of \redindex{\RX,\omega} and then by
Corollary~\ref{corollary omega shallow confluent up to omega no termination}
it also contradicts \math\omega-level and \math\omega-shallow confluence up
to \math 0.
However, we are allowed to assume this
since 
we know 
\bigmath{
  0
  \prec  
  \max\{n_0,n_1\}
}
and
\bigmath{
  0
  \prec  
   n_0         \plusomega n_1
}
due to
\bigmath{\Lambda_0\tightequal\Lambda_1\tightequal 1}
(and \math{\Lambda_i\tightpreceq n_i}).
A more general argumentation of this kind proves 
theorems \ref{theorem complementary}, \ref{theorem weakly complementary},
and \ref{theorem complementary zero}, which are confluence criteria for 
rule systems with complementary critical peaks.
Finally, the following items in the conjunctive condition lists allow us to 
assume some irreducibilities similar to those for \math\rhd-weak joinability
but less powerful.
\end{sloppypar}

\begin{lemma}
[\math\alpha-Shallow Joinability is Necessary for 
 \math\alpha-Shallow Confluence]%
\label{lemma omega shallow joinablility necessary}
\\
Let\/ \math{\alpha\in\{0,\omega\}}.
If\/ \RX\ is \math\alpha-shallow confluent 
[up to \math\beta\ [and \math s in \math\lhd]], 
then
\\
all critical peaks in ${\rm CP}(\R)$ 
are \math\alpha-shallow joinable 
[up to \math\beta\ [and \math s]] \wrt\ \RX\ [[and \math\lhd]]\@. 
\end{lemma}

\begin{lemma}
[\math\omega-Level Joinability is Necessary for 
 \math\omega-Level Confluence]%
\label{lemma omega level joinablility necessary}
\\
If\/ \RX\ is \math\omega-level confluent 
[up to \math\beta\ [and \math s in \math\lhd]], 
then
\\
all critical peaks in ${\rm CP}(\R)$ 
are \math\omega-level joinable 
[up to \math\beta\ [and \math s]] \wrt\ \RX\ [[and \math\lhd]]\@. 
\end{lemma}

\vfill

\pagebreak

\section{Sophisticated Forms of Shallow Joinability}
\label{section sophisticated shallow}

For a first reading this section should only be skimmed and its definitions
looked up by need. 
At least \sect~\ref{sect closed systems} should be read before.%
\footnote{We put this section here 
because we do not want to scatter our later
discussion with a big definition section 
and 
because we do not want to use the (for a first reading not essential) 
joinability labels in the 
boxes of the examples in the following sections before defining them.}

The \math\omega-shallow joinability notions of 
this section are only necessary for understanding the sophisticated 
Theorem~\ref{theorem parallel closed} and its interrelation with the
examples in the following sections, but not for the important practical
consequence of this theorem, namely Theorem~\ref{theorem complementary}, 
which is easy to understand and sufficient for many practical applications.
The \math 0-shallow joinability notions are needed for 
Theorem~\ref{theorem parallel closed zero} only.

\yestop
\noindent
The following notion will be applied for non-overlays of the forms
\math{(1,0)} 
and 
\math{(1,1)} 
for ``\mbox{\math{\alpha\tightequal\omega}}'' and of the form \math{(0,0)} 
for ``\mbox{\math{\alpha\tightequal 0     }}'':

\notop
\halftop
\begin{definition}[
\math     0-Shallow Parallel Closed
/
\math\omega-Shallow Parallel Closed
]%
\label{def parallel closed}\mbox{}
\\
Let\/ \math{\alpha\in\{0,\omega\}}.
Let\/ \math{\beta\preceq\omega\tight+\alpha}.
A critical peak \newcriticalpeak\
is\emph
{\math\alpha-shallow parallel closed up to \math\beta\ \wrt\ \RX} 
\udiff\
\bigmath{
  \forall\varphi\tightin\Xsubst\stopq
  \forall n_0,n_1\prec\omega\stopq
}
\\\LINEmath{
  \inparenthesesoplist{
    \inparenthesesoplist{
         0\tightprec n_0\tightsucceq n_1
       \oplistund
         n_0\plusalpha n_1
         \preceq
         \beta
       \oplistund
         \forall i\prec 2\stopq
         \inparenthesesoplist{
             \inparentheses{
                 \alpha\tightequal 0
               \ \implies\ 
                 \Lambda_i\tightequal 0\tightprec n_i
             }
           \oplistund
             \inparentheses{
                 \alpha\tightequal\omega
               \ \implies\ 
                 \Lambda_i\tightpreceq n_i
             }
           \oplistund
             D_i\varphi\mbox{ fulfilled \wrt\ }
             \redindex{\RX,\alpha+(n_i\monus 1)}
         }
       \oplistund
         \forall\delta\tightprec n_0\plusalpha n_1\stopq
           \RX\mbox{ is \math\alpha-shallow confluent up to }\delta
    }
    \oplistimplies
    \inparenthesesoplist{
        \inparentheses{
            n_1
            \tightequal
            0
          \ \implies\   
            t_0\varphi
            \redparaindex{\RX,\alpha}
            t_1\varphi
        }
      \oplistund
             t_0\varphi
             \redparaindex{\RX,\alpha+n_1}
             \tight\circ
             \refltransindex{\RX,\alpha+(n_1\monus1)}
             \circ
             \antirefltransindex{\RX,\alpha}
             t_1\varphi
    }
  }
.}
\\\headroom
It is called\emph{\math\alpha-shallow parallel closed \wrt\ \RX} \udiff\\
\LINEnomath{
  it is \math\alpha-shallow parallel closed 
  up to  \math{\omega\tight+\alpha} \wrt\ \RX\@.
}
\end{definition}

\yestop
\noindent
The following notion will be applied for critical peaks of the forms
\math{(0,1)} and \math{(1,1)} 
for ``\mbox{\math{\alpha\tightequal\omega}}'' and of the form \math{(0,0)} 
for ``\mbox{\math{\alpha\tightequal 0     }}'':

\notop
\halftop
\begin{definition}[
\math     0-Shallow 
/
\math\omega-Shallow 
{[Noisy]} Parallel Joinable]%
\label{def parallel joinable}\mbox{}
\\
Let\/ \math{\alpha\in\{0,\omega\}}.
Let\/ \math{\beta\preceq\omega\tight+\alpha}.
A critical peak \newcriticalpeak\
is\emph
{\math\alpha-shallow {\rm[}noisy\/{\rm]} parallel joinable up to \math\beta\ \wrt\ \RX} 
\udiff\
\bigmath{
  \forall\varphi\tightin\Xsubst\stopq
  \forall n_0,n_1\prec\omega\stopq
}
\\\LINEmath{
  \inparenthesesoplist{
    \inparenthesesoplist{
         n_0\tightpreceq n_1\tightsucc0
       \oplistund
         n_0\plusalpha n_1
         \preceq
         \beta
       \oplistund
         \forall i\prec 2\stopq
         \inparenthesesoplist{
             \inparentheses{
                 \alpha\tightequal 0
               \ \implies\ 
                 \Lambda_i\tightequal 0\tightprec n_i
             }
           \oplistund
             \inparentheses{
                 \alpha\tightequal\omega
               \ \implies\ 
                 \Lambda_i\tightpreceq n_i
             }
           \oplistund
             D_i\varphi\mbox{ fulfilled \wrt\ }
             \redindex{\RX,\alpha+(n_i\monus 1)}
         }
       \oplistund
         \forall\delta\tightprec n_0\plusalpha n_1\stopq
           \RX\mbox{ is \math\alpha-shallow confluent up to }\delta
    }
    \oplistimplies
      t_0\varphi
      \redparaindex{\RX,\alpha+n_1}
      \circ
      \refltransindex    {\RX,\alpha[+(n_1\monus1)]}
      \circ
      \antirefltransindex{\RX,\alpha+n_0}
      t_1\varphi
  }
.}
\\\headroom
It is called\emph{\math\alpha-shallow {\rm[}noisy\/{\rm]} parallel 
joinable \wrt\ \RX} \udiff\\
\LINEnomath{
  it is \math\alpha-shallow {[noisy]} parallel joinable 
  up to  \math{\omega\tight+\alpha} \wrt\ \RX\@.
}
\end{definition}

\pagebreak

\noindent
Note that \math\alpha-shallow parallel closedness
specializes to the standard definition of parallel closedness of
\citehuet\ 
for the
case that all symbols are considered to be non-constructor symbols
in case of \bigmath{\alpha\tightequal\omega} 
(or else constructor symbols in case of \math{\alpha\tightequal0})
and the rule system is unconditional
(since then
\math{\redindex{\RX,\alpha}\tightequal\emptyset}
and \math{\redindex{\RX,\alpha+1}\tightequal\redsub}).
Similarly, \math\alpha-shallow 
parallel joinability specializes for these cases to
the joinability required for overlays in \citetoyama.
Moreover, 
note that the notions whose names end with ``closed'' are always restricted 
to ``\math{0\tightprec n_0\tightsucceq n_1}'', whereas those whose names end
with ``joinable'' are always restricted to 
``\math{n_0\tightpreceq n_1\tightsucc 0}''.
Finally, note that some notions have ``noisy'' variants which are weaker 
since they allow some ``noise'', 
\ie\ some reduction on a smaller depth than the preceding reduction step.%
\footnote{The name for the notion was inspired by \citeoostromdiss.}

\yestop
\noindent
The following notion will be applied for non-overlays of the forms 
\math{(1,0)} and \math{(1,1)}
for ``\mbox{\math{\alpha\tightequal\omega}}'' and of the form \math{(0,0)} 
for ``\mbox{\math{\alpha\tightequal 0     }}'':

\notop
\halftop
\begin{definition}[
\math     0-Shallow 
/
\math\omega-Shallow {[Noisy]} Anti-Closed
]%
\label{def anti closed}\mbox{}
\\
Let\/ \math{\alpha\in\{0,\omega\}}.
Let\/ \math{\beta\preceq\omega\tight+\alpha}.
A critical peak \newcriticalpeak\
is\emph
{\math\alpha-shallow {\rm[}noisy\/{\rm]} anti-closed up to \math\beta\ \wrt\ \RX} 
\udiff\
\bigmath{
  \forall\varphi\tightin\Xsubst\stopq
  \forall n_0,n_1\prec\omega\stopq
}
\\\LINEmath{
  \inparenthesesoplist{
    \inparenthesesoplist{
         0\tightprec n_0\tightsucceq n_1
       \oplistund
         n_0\plusalpha n_1
         \preceq
         \beta
       \oplistund
         \forall i\prec 2\stopq
         \inparenthesesoplist{
             \inparentheses{
                 \alpha\tightequal 0
               \ \implies\ 
                 \Lambda_i\tightequal 0\tightprec n_i
             }
           \oplistund
             \inparentheses{
                 \alpha\tightequal\omega
               \ \implies\ 
                 \Lambda_i\tightpreceq n_i
             }
           \oplistund
             D_i\varphi\mbox{ fulfilled \wrt\ }
             \redindex{\RX,\alpha+(n_i\monus 1)}
         }
       \oplistund
         \forall\delta\tightprec n_0\plusalpha n_1\stopq
           \RX\mbox{ is \math\alpha-shallow confluent up to }\delta
    }
    \oplistimplies
    \inparenthesesoplist{
        \inparentheses{
            n_1\tightequal 0
          \ \implies\ 
            t_0\varphi
            \refltransindex{\RX,\alpha}
            \circ
            \antirefltransindex{\RX,\alpha[+(n_0\monus1)]}
            \tight\circ
            \antionlyonceindex{\RX,\alpha+n_0}
            t_1\varphi
        }
      \oplistund
             t_0\varphi
             \refltransindex{\RX,\alpha+n_1}
             \circ
             \antirefltransindex{\RX,\alpha{[+(n_0\monus1)]}}
             \tight\circ
             \antionlyonceindex{\RX,\alpha+n_0}
             \tight\circ
             \antirefltransindex{\RX,\alpha}
             t_1\varphi
    }
  }
.}
\\\headroom
It is called\emph{\math\alpha-shallow {\rm[}noisy\/{\rm]} 
anti-closed \wrt\ \RX} \udiff\\
\LINEnomath{
  it is \math\alpha-shallow {[noisy]} anti-closed 
  up to  \math{\omega\tight+\alpha} \wrt\ \RX\@.
}
\end{definition}

\yestop
\noindent
The following notion will be applied for critical peaks of the form
\math{(0,1)} and \math{(1,1)}
for ``\mbox{\math{\alpha\tightequal\omega}}'' and of the form \math{(0,0)} 
for ``\mbox{\math{\alpha\tightequal 0     }}'':

\notop
\halftop
\begin{definition}[
\math     0-Shallow 
/
\math\omega-Shallow {[Noisy]} Strongly Joinable
]%
\label{def strongly joinable}\mbox{}
\\
Let\/ \math{\alpha\in\{0,\omega\}}.
Let\/ \math{\beta\preceq\omega\tight+\alpha}.
A critical peak \newcriticalpeak\
is\emph
{\math\alpha-shallow {\rm[}noisy\/{\rm]} 
strongly joinable up to \math\beta\ \wrt\ \RX} 
\udiff\
\bigmath{
  \forall\varphi\tightin\Xsubst\stopq
  \forall n_0,n_1\prec\omega\stopq
}
\\\LINEmath{
  \inparenthesesoplist{
    \inparenthesesoplist{
         n_0\tightpreceq n_1\tightsucc0
       \oplistund
         n_0\plusalpha n_1
         \preceq
         \beta
       \oplistund
         \forall i\prec 2\stopq
         \inparenthesesoplist{
             \inparentheses{
                 \alpha\tightequal 0
               \ \implies\ 
                 \Lambda_i\tightequal 0\tightprec n_i
             }
           \oplistund
             \inparentheses{
                 \alpha\tightequal\omega
               \ \implies\ 
                 \Lambda_i\tightpreceq n_i
             }
           \oplistund
             D_i\varphi\mbox{ fulfilled \wrt\ }
             \redindex{\RX,\alpha+(n_i\monus 1)}
         }
       \oplistund
         \forall\delta\tightprec n_0\plusalpha n_1\stopq
           \RX\mbox{ is \math\alpha-shallow confluent up to }\delta
    }
    \oplistimplies
    \inparenthesesoplist{  
        \inparentheses{
            n_0\tightequal 0
          \ \implies\ 
            t_0\varphi
            \onlyonceindex{\RX,\alpha+n_1}
            \circ
            \refltransindex    {\RX,\alpha[+(n_1\monus1)]}
            \circ
            \antirefltransindex{\RX,\alpha%+n_0
            }
            t_1\varphi
        }
      \oplistund
            t_0\varphi
            \refltransindex    {\RX,\alpha}
            \circ
            \onlyonceindex{\RX,\alpha+n_1}
            \circ
            \refltransindex    {\RX,\alpha{[+(n_1\monus1)]}}
            \circ
            \antirefltransindex{\RX,\alpha+n_0}
            t_1\varphi
    }
  }
.}
\\\headroom
It is called\emph{\math\alpha-shallow {\rm[}noisy\/{\rm]} 
strongly joinable \wrt\ \RX} \udiff\\
\LINEnomath{
  it is \math\alpha-shallow {[noisy]} strongly joinable 
  up to  \math{\omega\tight+\alpha} \wrt\ \RX\@.
}
\end{definition}

\pagebreak

\yestop
\noindent
The following notion will be applied for non-overlays of the forms
\math{(1,0)} and
\math{(1,1)}:

\begin{definition}[\math\omega-Shallow Closed]%
\label{def closed}\mbox{}
\\
Let\/ \math{\beta\preceq\omega\tight+\omega}.
A critical peak \newcriticalpeak\
is\emph
{\math\omega-shallow closed up to \math\beta\ \wrt\ \RX} 
\udiff\
\bigmath{
  \forall\varphi\tightin\Xsubst\stopq
  \forall n_0,n_1\prec\omega\stopq
}
\\\LINEmath{
  \inparenthesesoplist{
    \inparenthesesoplist{
         0\tightprec n_0\tightsucceq n_1
       \oplistund
         n_0\plusomega n_1
         \preceq
         \beta
       \oplistund
         \forall i\prec 2\stopq
         \inparenthesesoplist{
             \Lambda_i\preceq n_i
           \oplistund
             D_i\varphi\mbox{ fulfilled \wrt\ }
             \redindex{\RX,\omega+(n_i\monus 1)}
         }
       \oplistund
         \forall\delta\tightprec n_0\plusomega n_1\stopq
           \RX\mbox{ is \math\omega-shallow confluent up to }\delta
    }
    \oplistimplies
    \inparenthesesoplist{
        \inparentheses{
            n_1\tightequal0 
          \ \implies\ 
            t_0\varphi
            \onlyonceindex{\RX,\omega}
            \circ
            \antirefltransindex{\RX,\omega}
            t_1\varphi
        }
      \oplistund
             t_0\varphi
             \onlyonceindex{\RX,\omega+n_1}
             \tight\circ
             \refltransindex{\RX,\omega+(n_1\monus1)}
             \circ
             \antirefltransindex{\RX,\omega}
             t_1\varphi
    }
  }
.}
\\\headroom
It is called\emph{\math\omega-shallow closed \wrt\ \RX} \udiff\\
\LINEnomath{
  it is \math\omega-shallow closed 
  up to  \math{\omega\tight+\omega} \wrt\ \RX\@.
}
\end{definition}

\yestop
\noindent
The following notion will be applied for critical peaks of the forms
\math{(0,1)} and
\math{(1,1)}:

\begin{definition}[\math\omega-Shallow {[Noisy]} Weak Parallel Joinable]%
\label{def parallel weak joinable}\mbox{}
\\
Let\/ \math{\beta\preceq\omega\tight+\omega}.
A critical peak \newcriticalpeak\
is\emph
{\math\omega-shallow {\rm[}noisy\/{\rm]} 
weak parallel joinable up to \math\beta\ \wrt\ \RX} 
\udiff\
\bigmath{
  \forall\varphi\tightin\Xsubst\stopq
  \forall n_0,n_1\prec\omega\stopq
}
\\\LINEmath{
  \inparenthesesoplist{
    \inparenthesesoplist{
         n_0\tightpreceq n_1\tightsucc0
       \oplistund
         n_0\plusomega n_1
         \preceq
         \beta
       \oplistund
         \forall i\prec 2\stopq
         \inparenthesesoplist{
             \Lambda_i\preceq n_i
           \oplistund
             D_i\varphi\mbox{ fulfilled \wrt\ }
             \redindex{\RX,\omega+(n_i\monus 1)}
         }
       \oplistund
         \forall\delta\tightprec n_0\plusomega n_1\stopq
           \RX\mbox{ is \math\omega-shallow confluent up to }\delta
    }
    \oplistimplies
      t_0\varphi
      \refltransindex{\RX,\omega}
      \circ
      \redparaindex{\RX,\omega+n_1}
      \circ
      \refltransindex    {\RX,\omega[+(n_1\monus1)]}
      \circ
      \antirefltransindex{\RX,\omega+n_0}
      t_1\varphi
  }
.}
\\\headroom
It is called\emph{\math\omega-shallow {\rm[}noisy\/{\rm]} 
weak parallel joinable \wrt\ \RX} \udiff\\
\LINEnomath{
  it is \math\omega-shallow {[noisy]} weak parallel joinable 
  up to  \math{\omega\tight+\omega} \wrt\ \RX\@.
}
\end{definition}

\noindent
The following are corollaries of Corollary~\ref{corollary parallel one}:
\begin{corollary}\label{corollary joinabilities}
Let\/ \math{\alpha\in\{0,\omega\}}.
Now \wrt\ \RX\ the following holds:\\
If a critical peak is 
\math\omega-shallow {\rm[}noisy\/{\rm]} parallel joinable up to 
\math{\beta\preceq\omega\tight+\omega},
\\
then it is
\math\omega-shallow {\rm[}noisy\/{\rm]} 
weak parallel joinable up to \math\beta.
\\
If a critical peak is 
\math\omega-shallow {\rm[}noisy\/{\rm]} strongly joinable up to 
\math{\beta\preceq\omega\tight+\omega},\\
then it is
\math\omega-shallow {\rm[}noisy\/{\rm]} weak 
parallel joinable up to \math\beta.
\\
If a critical peak is 
\math\alpha-shallow {\rm[}noisy\/{\rm]} strongly joinable up to 
\math{\beta\preceq\omega},\\
then it is
\math\alpha-shallow {\rm[}noisy\/{\rm]} parallel joinable up to \math\beta\@.
\end{corollary}

\begin{corollary}\label{corollary closednesses one}
Let\/ \math{\alpha\in\{0,\omega\}}.
Let\/ \math{\beta\preceq\omega\tight+\alpha}.
Now \wrt\ \RX\ the following holds:\\
If a critical peak is 
\math\alpha-shallow parallel closed or 
(for \math{\alpha\tightequal\omega})
\math\alpha-shallow closed 
up to \math\beta,
then it is
\math\alpha-shallow {\rm[}noisy\/{\rm]} anti-closed up to \math\beta.
\end{corollary}

\pagebreak
\setbox0=\hbox to \textheight{\vbox{\noindent
{\large\bf
Overview over sophisticated forms of \math\omega-Shallow \ldots\ 
of \newcriticalpeak
}
\\
\noindent
{Generally assumed condition for 
\math{\varphi\in\Xsubst}; \math{n_0,n_1\prec\omega}:
\   \inparenthesesoplist{
         \mbox{``Property~1''}
       \ \und\ 
         n_0\plusomega n_1
         \preceq
         \beta
       \oplistund
         \forall i\prec 2\stopq
         \inparentheses{
             \Lambda_i\preceq n_i
           \ \und\
             D_i\varphi\mbox{ fulfilled \wrt\ }
             \redindex{\RX,\omega+(n_i\monus 1)}
         }
       \oplistund
         \forall\delta\tightprec n_0\plusomega n_1\stopq
           \RX\mbox{ is \math\omega-shallow confluent up to }\delta
    }
}
\\
\hbox{%
Required conclusion (P
:=
Parallel;
C
:=
Closed;
N
:=
Noisy;
J
:=
Joinable;
W
:=
Weak;
A
:=
Anti-;
S
:=
Strongly):
}

\noindent
\begin{tabular}{@{}c@{\vline\hskip1pt\vline~~}c|c@{\vline\hskip1pt\vline}c@{}}
Property~1 := \ldots
&\multicolumn{2}{@{}c@{\vline\hskip1pt\vline}}{
\math{0\prec n_0\succeq n_1}
\mbox{}
\mbox{}
\mbox{}
\mbox{}
\mbox{}
\mbox{}
\mbox{}
\mbox{}
\mbox{}
\mbox{}
\mbox{}
\mbox{}
}
&\math{n_0\preceq n_1\succ0}
\\
In case of \ldots
&\math{n_1=0}
&\math{n_1\succ0}
&
\\\hline
&
\begin{diagram}
&&t_1\varphi\headroom
\\
&\mbox{PC}
&\dequal
\\
t_0\varphi&\rredparaindex{\omega}&\circ
\end{diagram}
\mbox{}
\mbox{}
\mbox{}
\mbox{}
\mbox{}
\mbox{}
\mbox{}
\mbox{}
&
\begin{diagram}
&&&&t_1\varphi\headroom
\\
&&\mbox{PC}
&&\drefltransindex{\omega}
\\
t_0\varphi&\rredparaindex{\omega+n_1}&\circ
&\rrefltransindex{\omega+(n_1\monus1)}&\circ
\end{diagram}
\mbox{}
\mbox{}
\mbox{}
\mbox{}
&
\begin{diagram}
&&&&t_1\varphi\headroom
\\
&&\mbox{{[N]}PJ}&&\drefltransindex{\omega+n_0}
\\
t_0\varphi&\rredparaindex{\omega+n_1}&\circ
&\rrefltransindex{\omega[+(n_1\monus1)]~}
&\circ
\end{diagram}
\raisebox{-10ex}{\rule{0ex}{.5ex}}
\mbox{}
\mbox{}
\mbox{}
\mbox{}
\mbox{}
\mbox{}
\\\hline
&
\begin{diagram}
&&t_1\varphi\headroom
\\
&\mbox{C}&\drefltransindex{\omega}
\\
t_0\varphi&\ronlyonceindex{\omega}&\circ
\end{diagram}
\mbox{}
\mbox{}
\mbox{}
\mbox{}
\mbox{}
\mbox{}
\mbox{}
\mbox{}
&
\begin{diagram}
&&&&t_1\varphi\headroom
\\
&&\mbox{C}
&&\drefltransindex{\omega}
\\
t_0\varphi&\ronlyonceindex{\omega+n_1}
&\circ&\rrefltransindex{\omega+(n_1\monus1)}&\circ
\end{diagram}
\mbox{}
\mbox{}
\mbox{}
\mbox{}
&
\mbox{}
\mbox{}
\mbox{}
\begin{diagram}
&&&&&&t_1\varphi\headroom
\\
&&&\mbox{{[N]}WPJ}
&&&\drefltransindex{\omega+n_0}
\\
t_0\varphi&\rrefltransindex{\omega}&\circ
&\rredparaindex{\omega+n_1}&\circ&\rrefltransindex{\omega[+(n_1\monus1)]~}
&\circ
\end{diagram}
\raisebox{-10ex}{\rule{0ex}{.5ex}}
\mbox{}
\mbox{}
\mbox{}
\mbox{}
\mbox{}
\mbox{}
\\\hline
&
%\multicolumn{2}{@{}c@{\vline\hskip1pt\vline}}{
\begin{diagram}
&&t_1\varphi\headroom
\\
&&\donlyonceindex{\omega+n_0}
\\
&\mbox{{[N]}AC}
&\circ
\\
&&\drefltransindex{\omega[+(n_0\monus1)]}
\\
t_0\varphi&\rrefltransindex{\omega}&\circ
\end{diagram}
\mbox{}
\mbox{}
\mbox{}
\mbox{}
\mbox{}
\mbox{}
\mbox{}
\mbox{}
&
\begin{diagram}
&&t_1\varphi\headroom
\\
&&\drefltransindex\omega
\\
&&\circ
\\
&\mbox{{[N]}AC}
&\donlyonceindex{\omega+n_0}
\\
&&\circ
\\
&&\drefltransindex{\omega[+(n_0\monus1)]}
\\
t_0\varphi&\rrefltransindex{\omega+n_1}&\circ
\end{diagram}
\mbox{}
\mbox{}
\mbox{}
\mbox{}
%}
&
\mbox{}
\mbox{}
\mbox{}
\begin{diagram}
&&&&&&t_1\varphi\headroom
\\
&&&\mbox{{[N]}SJ}
&&&\drefltransindex{\omega+n_0}
\\
t_0\varphi
&\rrefltransindex{(n_0\neq0)\omega}&\circ
&\ronlyonceindex{\omega+n_1}
&\circ&\rrefltransindex{\omega[+(n_1\monus1)]~}
&\circ
\end{diagram}
\raisebox{-10ex}{\rule{0ex}{.5ex}}
\mbox{}
\mbox{}
\mbox{}
\mbox{}
\mbox{}
\mbox{}
\end{tabular}
}}
%\copy0
%\pagebreak
\rotl0

\pagebreak

\section{Sophisticated Forms of Level Joinability}
\label{section sophisticated level}

For a first reading this section should only be skimmed and its definitions
looked up by need. 
At least \sect~\ref{section sophisticated shallow} should be read before.

\yestop
\noindent
This section is only necessary for understanding the sophisticated 
Theorem~\ref{theorem level parallel closed} and its interrelation with the
examples in the following sections, but not for the easy to understand
consequence of this theorem, namely Theorem~\ref{theorem weakly complementary}.

\yestop
\noindent
Having completed our special notions for shallow confluence, we now present
some for level confluence.

\yestop
\noindent
The following notion will be applied for non-overlays of the form \math{(1,1)}:

\begin{definition}[\math\omega-Level Parallel Closed]%
\label{def level parallel closed}\mbox{}
\\
Let\/ \math{\beta\preceq\omega}.
A critical peak \newcriticalpeak\
is
\\\emph
{\math\omega-level parallel closed up to \math\beta\ \wrt\ \RX} 
\udiff\\
\bigmath{
  \forall\varphi\tightin\Xsubst\stopq
  \forall n\prec\omega\stopq
}
\\\LINEmath{
  \inparenthesesoplist{
    \inparenthesesoplist{
         0\tightprec n
       \oplistund
         n
         \preceq
         \beta
       \oplistund
         \forall i\prec 2\stopq
         \inparenthesesoplist{
             \Lambda_i\preceq n
           \oplistund
             D_i\varphi\mbox{ fulfilled \wrt\ }
             \redindex{\RX,\omega+(n\monus 1)}
         }
       \oplistund
         \forall\delta\tightprec n\stopq
           \RX\mbox{ is \math\omega-level confluent up to }\delta
       \oplistund
         \RX\mbox{ is \math\omega-shallow confluent up to }\omega
    }
    \oplistimplies 
             t_0\varphi
             \redparaindex{\RX,\omega+n}
             \tight\circ
             \refltransindex{\RX,\omega}
             \circ
             \antirefltransindex{\RX,\omega}
             t_1\varphi
  }
.}
\\\headroom
It is called\emph{\math\omega-level parallel closed \wrt\ \RX} \udiff\\
\LINEnomath{
  it is \math\omega-level parallel closed 
  up to  \math{\omega} \wrt\ \RX\@.
}
\end{definition}

\yestop
\noindent
The following notion will be applied for critical peaks 
of the form \math{(1,1)}:

\begin{definition}[\math\omega-Level Parallel Joinable]%
\label{def level parallel joinable}\mbox{}
\\
Let\/ \math{\beta\preceq\omega}.
A critical peak \newcriticalpeak\
is
\\\emph
{\math\omega-level parallel joinable up to \math\beta\ \wrt\ \RX} 
\udiff\\
\math{
  \forall\varphi\tightin\Xsubst\stopq
  \forall n\prec\omega\stopq
}
\\\LINEmath{
  \inparenthesesoplist{
    \inparenthesesoplist{
         n\tightsucc0
       \oplistund
         n
         \preceq
         \beta
       \oplistund
         \forall i\prec 2\stopq
         \inparenthesesoplist{
             \Lambda_i\preceq n
           \oplistund
             D_i\varphi\mbox{ fulfilled \wrt\ }
             \redindex{\RX,\omega+(n\monus 1)}
         }
       \oplistund
         \forall\delta\tightprec n\stopq
           \RX\mbox{ is \math\omega-level confluent up to }\delta
       \oplistund
         \RX\mbox{ is \math\omega-shallow confluent up to }\omega
    }
    \oplistimplies 
      t_0\varphi
      \redparaindex{\RX,\omega+n}
      \circ
      \refltransindex    {\RX,\omega}
      \circ
      \antirefltransindex{\RX,\omega+n}
      t_1\varphi
  }
.}
\\\headroom
It is called\emph{\math\omega-level parallel joinable \wrt\ \RX} \udiff\\
\LINEnomath{
  it is \math\omega-level parallel joinable 
  up to  \math{\omega} \wrt\ \RX\@.
}
\end{definition}

\vfill

\pagebreak

\yestop
\noindent
The following notion will be applied for non-overlays of the form \math{(1,1)}:

\begin{definition}[\math\omega-Level Anti-Closed]%
\label{def level anti-closed}\mbox{}
\\
Let\/ \math{\beta\preceq\omega}.
A critical peak \newcriticalpeak\
\\ 
is\emph
{\math\omega-level anti-closed up to \math\beta\ \wrt\ \RX} 
\udiff\\
\math{
  \forall\varphi\tightin\Xsubst\stopq
  \forall n\prec\omega\stopq
}
\\\LINEmath{
  \inparenthesesoplist{
    \inparenthesesoplist{
         0\tightprec n
       \oplistund
         n
         \preceq
         \beta
       \oplistund
         \forall i\prec 2\stopq
         \inparenthesesoplist{
             \Lambda_i\preceq n
           \oplistund
             D_i\varphi\mbox{ fulfilled \wrt\ }
             \redindex{\RX,\omega+(n\monus 1)}
         }
       \oplistund
         \forall\delta\tightprec n\stopq
           \RX\mbox{ is \math\omega-level confluent up to }\delta
       \oplistund
         \RX\mbox{ is \math\omega-shallow confluent up to }\omega
    }
    \oplistimplies 
             t_0\varphi
             \refltransindex{\RX,\omega+n}
             \circ
             \antirefltransindex{\RX,\omega}
             \tight\circ
             \antionlyonceindex{\RX,\omega+n}
             \tight\circ
             \antirefltransindex{\RX,\omega}
             t_1\varphi
  }
.}
\\\headroom
It is called\emph{\math\omega-level anti-closed \wrt\ \RX} \udiff\\
\LINEnomath{
  it is \math\omega-level anti-closed 
  up to  \math{\omega} \wrt\ \RX\@.
}
\end{definition}

\yestop
\noindent
The following notion will be applied for critical peaks 
of the form \math{(1,1)}:

\begin{definition}[\math\omega-Level Strongly Joinable]%
\label{def level strongly joinable}\mbox{}
\\
Let\/ \math{\beta\preceq\omega}.
A critical peak \newcriticalpeak\
\\ 
is\emph
{\math\omega-level strongly joinable up to \math\beta\ \wrt\ \RX} 
\udiff\\
\math{
  \forall\varphi\tightin\Xsubst\stopq
  \forall n\prec\omega\stopq
}
\\\LINEmath{
  \inparenthesesoplist{
    \inparenthesesoplist{
         n\tightsucc0
       \oplistund
         n
         \preceq
         \beta
       \oplistund
         \forall i\prec 2\stopq
         \inparenthesesoplist{
             \Lambda_i\preceq n
           \oplistund
             D_i\varphi\mbox{ fulfilled \wrt\ }
             \redindex{\RX,\omega+(n\monus 1)}
         }
       \oplistund
         \forall\delta\tightprec n\stopq
           \RX\mbox{ is \math\omega-level confluent up to }\delta
       \oplistund
         \RX\mbox{ is \math\omega-shallow confluent up to }\omega
    }
    \oplistimplies 
      t_0\varphi
      \refltransindex{\RX,\omega}
      \tight\circ
      \onlyonceindex{\RX,\omega+n}
      \tight\circ
      \refltransindex    {\RX,\omega}
      \circ
      \antirefltransindex{\RX,\omega+n}
      t_1\varphi
  }
.}
\\\headroom
It is called\emph{\math\omega-level strongly joinable \wrt\ \RX} \udiff\\
\LINEnomath{
  it is \math\omega-level strongly joinable 
  up to  \math{\omega} \wrt\ \RX\@.
}
\end{definition}

\vfill

\pagebreak

\yestop
\noindent
The following notion will be applied for non-overlays of the form \math{(1,1)}:

\begin{definition}[\math\omega-Level Closed]%
\label{def level closed}\mbox{}
\\
Let\/ \math{\beta\preceq\omega}.
A critical peak \newcriticalpeak\
\\ 
is\emph
{\math\omega-level closed up to \math\beta\ \wrt\ \RX} 
\udiff\\
\math{
  \forall\varphi\tightin\Xsubst\stopq
  \forall n\prec\omega\stopq
}
\\\LINEmath{
  \inparenthesesoplist{
    \inparenthesesoplist{
         0\tightprec n
       \oplistund
         n
         \preceq
         \beta
       \oplistund
         \forall i\prec 2\stopq
         \inparenthesesoplist{
             \Lambda_i\preceq n
           \oplistund
             D_i\varphi\mbox{ fulfilled \wrt\ }
             \redindex{\RX,\omega+(n\monus 1)}
         }
       \oplistund
         \forall\delta\tightprec n\stopq
           \RX\mbox{ is \math\omega-level confluent up to }\delta
       \oplistund
         \RX\mbox{ is \math\omega-shallow confluent up to }\omega
    }
    \oplistimplies 
             t_0\varphi
             \onlyonceindex{\RX,\omega+n}
             \tight\circ
             \refltransindex{\RX,\omega}
             \circ
             \antirefltransindex{\RX,\omega}
             t_1\varphi
  }
.}
\\\headroom
It is called\emph{\math\omega-level closed \wrt\ \RX} \udiff\\
\LINEnomath{
  it is \math\omega-level closed 
  up to  \math{\omega} \wrt\ \RX\@.
}
\end{definition}

\yestop
\noindent
The following notion will be applied for critical peaks 
of the form \math{(1,1)}:

\begin{definition}[\math\omega-Level Weak Parallel Joinable]%
\label{def level parallel weak joinable}\mbox{}
\\
Let\/ \math{\beta\preceq\omega}.
A critical peak \newcriticalpeak\
\\ 
is\emph
{\math\omega-level weak parallel joinable up to \math\beta\ \wrt\ \RX} 
\udiff\\
\math{
  \forall\varphi\tightin\Xsubst\stopq
  \forall n\prec\omega\stopq
}
\\\LINEmath{
  \inparenthesesoplist{
    \inparenthesesoplist{
         n\tightsucc0
       \oplistund
         n
         \preceq
         \beta
       \oplistund
         \forall i\prec 2\stopq
         \inparenthesesoplist{
             \Lambda_i\preceq n
           \oplistund
             D_i\varphi\mbox{ fulfilled \wrt\ }
             \redindex{\RX,\omega+(n\monus 1)}
         }
       \oplistund
         \forall\delta\tightprec n\stopq
           \RX\mbox{ is \math\omega-level confluent up to }\delta
       \oplistund
         \RX\mbox{ is \math\omega-shallow confluent up to }\omega
    }
    \oplistimplies 
      t_0\varphi
      \refltransindex{\RX,\omega}
      \circ
      \redparaindex{\RX,\omega+n}
      \circ
      \refltransindex    {\RX,\omega}
      \circ
      \antirefltransindex{\RX,\omega+n}
      t_1\varphi
  }
.}
\\\headroom
It is called\emph{\math\omega-level weak parallel joinable \wrt\ \RX} \udiff\\
\LINEnomath{
  it is \math\omega-level weak parallel joinable 
  up to  \math{\omega} \wrt\ \RX\@.
}
\end{definition}

\vfill

\pagebreak
\setbox0=\hbox to \textheight{\vbox{\noindent
{\large\bf
Overview over sophisticated forms of \math\omega-Level \ldots\ 
of \newcriticalpeak
}

\yestop
\noindent
{Generally assumed condition for 
\math{\varphi\in\Xsubst}; \math{n\prec\omega}:
\    \inparenthesesoplist{
         0\tightprec n
       \ \und\
         n
         \preceq
         \beta
       \oplistund
         \forall i\prec 2\stopq
         \inparentheses{
             \Lambda_i\preceq n
           \ \und\
             D_i\varphi\mbox{ fulfilled \wrt\ }
             \redindex{\RX,\omega+(n\monus 1)}
         }
       \oplistund
         \forall\delta\tightprec n\stopq
           \RX\mbox{ is \math\omega-level confluent up to }\delta
       \oplistund
         \RX\mbox{ is \math\omega-shallow confluent up to }\omega
    }
}

\noindent
\hbox{%
Required conclusion (P
:=
Parallel;
C
:=
Closed;
J
:=
Joinable;
W
:=
Weak;
A
:=
Anti-;
S
:=
Strongly):
}

\noindent
\begin{tabular}{@{}c@{\vline\hskip1pt\vline~~}c@{\vline\hskip1pt\vline}c@{}}
\hline
&
\begin{diagram}
&&&&t_1\varphi\headroom
\\
&&\mbox{PC}
&&\drefltransindex{\omega}
\\
t_0\varphi&\rredparaindex{\omega+n}&\circ
&\rrefltransindex{\omega}&\circ
\end{diagram}
\mbox{}
\mbox{}
\mbox{}
\mbox{}
&
\begin{diagram}
&&&&t_1\varphi\headroom
\\
&&\mbox{PJ}&&\drefltransindex{\omega+n}
\\
t_0\varphi&\rredparaindex{\omega+n}&\circ
&\rrefltransindex{\omega}
&\circ
\end{diagram}
\raisebox{-10ex}{\rule{0ex}{.5ex}}
\mbox{}
\mbox{}
\mbox{}
\mbox{}
\mbox{}
\mbox{}
\\\hline
&
\begin{diagram}
&&&&t_1\varphi\headroom
\\
&&\mbox{C}
&&\drefltransindex{\omega}
\\
t_0\varphi&\ronlyonceindex{\omega+n}
&\circ&\rrefltransindex{\omega}&\circ
\end{diagram}
\mbox{}
\mbox{}
\mbox{}
\mbox{}
&
\mbox{}
\mbox{}
\mbox{}
\begin{diagram}
&&&&&&t_1\varphi\headroom
\\
&&&\mbox{WPJ}
&&&\drefltransindex{\omega+n}
\\
t_0\varphi&\rrefltransindex{\omega}&\circ
&\rredparaindex{\omega+n}&\circ&\rrefltransindex{\omega}
&\circ
\end{diagram}
\raisebox{-10ex}{\rule{0ex}{.5ex}}
\mbox{}
\mbox{}
\mbox{}
\mbox{}
\mbox{}
\mbox{}
\\\hline
&
\begin{diagram}
&&t_1\varphi\headroom
\\
&&\drefltransindex{\omega}
\\
&&\circ
\\
&\mbox{AC}&\donlyonceindex{\omega+n}
\\
&&\circ
\\
&&\drefltransindex{\omega}
\\
t_0\varphi&\rrefltransindex{\omega+n}&\circ
\end{diagram}
\mbox{}
\mbox{}
\mbox{}
\mbox{}
&
\mbox{}
\mbox{}
\mbox{}
\begin{diagram}
&&&&&&t_1\varphi\headroom
\\
&&&\mbox{SJ}
&&&\drefltransindex{\omega+n}
\\
t_0\varphi&\rrefltransindex{\omega}&\circ
&\ronlyonceindex{\omega+n}&\circ&\rrefltransindex{\omega}
&\circ
\end{diagram}
\raisebox{-10ex}{\rule{0ex}{.5ex}}
\mbox{}
\mbox{}
\mbox{}
\mbox{}
\mbox{}
\mbox{}
\end{tabular}
}}
%\copy0
%\pagebreak
\rotl0

\pagebreak

\section{Quasi Overlay Joinability}
\label{sect quasi overlay joinability}

According to Theorem~4 of \citeder, a terminating
positive conditional rule system
is confluent if it is overlay joinable. 
The remainder of this section is only relevant for 
Theorem~\ref{theorem quasi overlay joinable} and even this can be applied
without knowing about \math\rhd-quasi 
overlay joinability when one just knows:
\begin{lemma}[
              Overlay Joinable
              \implies\
              {\rm\tight\rhd}-Quasi Overlay Joinable
]
\label{lemma quasi overlay joinable}
\\
\Wrt\ \RX\ the following holds for each critical peak:
\\
If it
is overlay joinable, then it is \math\rhd-quasi overlay joinable.
\end{lemma}

\yestop
\noindent
In \citewgjsc\ we introduced the following definition:
\\
A critical peak \criticalpeaklongform\ 
is\emph{quasi overlay joinable \wrt\ \RX} \udiff\\ 
\LINEmath{
  \forall\varphi\tightin\Xsubst\stopq
   \inparenthesesoplist{
       \inparentheses{
            (D_0 D_1)\sigma\varphi\mbox{ fulfilled \wrt\ }\redsub
       }
       \oplistimplies
       \inparenthesesoplist{
           t_1\sigma\varphi
           \tightequal
           \repl
             {t_0\sigma\varphi}
             {p}
             {t_1\sigma\varphi/p}
         \oplistund
           (t_0/p)\sigma\varphi
           \,\,\downarrowsub\,\,
           t_1\sigma\varphi/p
           \,\,\,\transclosureinline{(\antiredsub\cup\subterm)}\,\,\, 
           (\hat t/p)\sigma\varphi
       }
   }
.}

\noindent
This notion of quasi overlay joinability, however, 
has turned out to produce a wondrous effect in case
that for some critical peak,
\wrog\ say
\\\linemath{
   ((\repl{l_1}{p}{r_0},
     C_0,
     \Lambda_0),\ 
    (r_1,
     C_1,
     \Lambda_1),\ 
    l_1,\ 
    \sigma,\ 
    p\ 
   )
}
generated by two rules 
\kurzregelindex0,
\kurzregelindex1
(with \wrog\ no variables in common)
due to 
\bigmath{
  \sigma
  \tightequal
  \minmgu
    {\{(l_0,l_1/p)\}}
    {\Y}
}
for
\bigmath{
  \Y
  :=
  \VAR{\kurzregelindex0,\kurzregelindex1}
,}
and for some
\math{
  \varphi
  \in
  \Xsubst
}
with 
\bigmath{
  (C_0\,C_1)\sigma\varphi
}
fulfilled \wrt\ \redsub,
there 
is some \math{p'\in\TPOS{l_1}\tightsetminus\{p\}} with 
\bigmath{l_1/p'\tightnotin\V}
and
\bigmath{
  l_0\sigma\varphi
  \tightequal
  (l_1/p')\sigma\varphi
;}
\ie\ the left-hand side of the rule
\kurzregelindex 0 occurs a second time in the instantiated peak term
(or superposition term) at a non-variable position \math{p'}. 
In this case due to 
\bigmath{
  l_0\sigma\varphi
  \tightequal
  (l_1/p')\sigma\varphi
}
there are
\bigmath{
  \sigma'
  \tightequal
  \minmgu
    {\{(l_0,l_1/p')\}}
    {\Y}
}
and 
\math{
  \varphi'
  \in
  \Xsubst
}
with 
\bigmath{
  \domres{\inpit{\sigma'\varphi'}}\Y
  \tightequal
  \domres{\inpit{\sigma\varphi}}\Y
}
and then 
(unless \math{\repl{l_1}{p'}{r_0}\sigma'\tightequal r_1\sigma'})
we get another critical peak
\\\linemath{
   ((\repl{l_1}{p'}{r_0},
     C_0,
     \Lambda_0),\ 
    (r_1,
     C_1,
     \Lambda_1),\ 
    l_1,\ 
    \sigma',\ 
    p'\ 
   )
.}
Now 
(since 
\bigmath{
  (C_0\,C_1)\sigma'\varphi'
  =
  (C_0\,C_1)\sigma \varphi
}
is fulfilled \wrt\ \redsub),
if both critical peaks are quasi overlay joinable,
then we get by the first conclusion in the above definition:
\\\linemath{
  \begin{array}[t]{lll@{}l@{}l@{}ll}
    r_1\sigma\varphi
    &\tightequal
    &l_1
    &\replsuffix{p}{r_0}
    &\sigma\varphi
    &\replsuffix{p}{r_1\sigma\varphi/p}
    &;
    \\
    r_1\sigma'\varphi'
    &\tightequal
    &l_1
    &\replsuffix{p'}{r_0}
    &\sigma'\varphi'
    &\replsuffix{p'}{r_1\sigma'\varphi'/p'}
    &
    \\
  \end{array}
}
(unless 
\math{\repl{l_1}{p'}{r_0}\sigma'\varphi'\tightequal r_1\sigma'\varphi'}).
Simplified, this means:
\\\linemath{
  \begin{array}[t]{llll}
    r_1\sigma\varphi
    &\tightequal
    &l_1\sigma\varphi\replsuffix{p}{r_1\sigma\varphi/p}
    &;
    \\
    r_1\sigma\varphi
    &\tightequal
    &l_1\sigma\varphi\replsuffix{p'}{r_1\sigma\varphi/p'}
    &
    \\
  \end{array}
}
(unless 
\math{
  r_1\sigma\varphi
  \tightequal
  \repl{l_1\sigma\varphi}{p'}{r_0\sigma\varphi}
}).
Thus, in any case, we get
\\\linemath{
  \repl{l_1\sigma\varphi}{p }{\ldots}
  \tightequal
  r_1\sigma\varphi
  \tightequal
  \repl{l_1\sigma\varphi}{p'}{\ldots}
.}
Since 
(due to \bigmath{p\tightnotequal p'} and 
\math{
  (l_1/p')\sigma\varphi
  \tightequal
  l_0\sigma\varphi
  \tightequal
  (l_1/p )\sigma\varphi
})
we have 
\bigmath{\neitherprefix{p'}p,}
this has the wondrous result
\\\LINEmath{\headroom\footroom
  l_1\sigma\varphi
  \tightequal
  r_1\sigma\varphi
.}(!)
\\
Using the second conclusion of the quasi overlay joinability we get
\\\bigmath{
  l_1\sigma\varphi/p
  \tightequal
  r_1\sigma\varphi/p
  \;\transclosureinline{\inparenthesesinlinetight{\antired\cup\subterm}}\;
  (l_1/p)\sigma\varphi
}
which implies
\\\LINEmath{\headroom\footroom
  l_0\sigma\varphi
  \,\,\transclosureinline{\inparenthesesinlinetight{\antired\cup\subterm}}\,\,
  l_0\sigma\varphi
.}(!!)
\\
\noindent
Since both results (!)\ and (!!)\ are absurd for a property which is only
to be used for a noethe-\linebreak rian reduction relation \redsub,
we now generalize our notion of quasi overlay joinability.

\pagebreak

\begin{definition}[\mbox{\rm\tight\rhd}-Quasi Overlay Joinability]
\label{def quasi overlay joinability}
\\
A critical peak \criticalpeaklongform\ 
is\/ {\em \math\rhd-quasi overlay joinable \wrt\ \RX} \udiff\ 
\\\math{
  \forall\varphi\tightin\Xsubst\stopq
  \forall\Delta\stopq
}
\\\LINEmath{
    \inparenthesesoplist{
          \inparenthesesoplist{
              (D_0 D_1)\sigma\varphi\mbox{ fulfilled \wrt\ }\redsub
            \oplistund
              \Delta
              \tightequal
              \setwith
                {p'\tightin\TPOS{\hat t}\tightsetminus\{p\}}
                {  \hat t/p'\tightnotin\V
                 \und
                   (\hat t/p')\sigma\varphi
                   \tightequal
                   (\hat t/p )\sigma\varphi
                }
            \oplistund
              \forall w\,\transclosureinline{(\antiredsub\tightcup\,\lhd)}\,\,
                      (\hat t/p )\sigma\varphi
              \stopq
                \ \redsub\mbox{ is confluent below }w
            \oplistund
              \forall p''
                \tightin
                \TPOS{(\hat t/p )\sigma\varphi}\tightsetminus\{\emptyset\}
                \stopq
                (\hat t/p )\sigma\varphi/p''\tightnotin\DOM\redsub
          }
          \oplistimplies
              \exists \bar n\tightin\N\stopq
              \exists \bar p          \stopq
              \exists \bar u          \stopq
              \inparenthesesoplist{
                  \FUNDEF{\bar p}{\{0,\ldots,\bar n\tight-1\}}{\N^\ast}
                \oplistund
                  \FUNDEF{\bar u}{\{0,\ldots,\bar n\}}        \vt      
                \oplistund
                  \replpar
                    {t_0}
                    {p'}
                    {t_0/p}
                    {p'\tightin\Delta}
                  \sigma\varphi
                  \refltranssub
                  \bar u_{\bar n}
                \oplistund
                  \forall i\tightprec\bar n\stopq
                  \inparenthesesoplist{
                      \bar u_{i+1}
                      \tightequal           
                      \repl{\bar u_i}{\bar p_i}{\bar u_{i+1}/\bar p_i}
                    \oplistund
                      \bar u_{i+1}/\bar p_i
                      \antirefltranssub
                      \bar u_i/\bar p_i
                      \;\transclosureinline{(\antiredsub\tightcup\,\lhd)}\;
                      (\hat t/p)\sigma\varphi
                  }
                \oplistund
                  \bar u_0
                  \tightequal
                  t_1\sigma\varphi
              }
    }
.}
\end{definition}

\yestop
\noindent
For \math{\varphi\in\Xsubst} and 
\math{\Delta\subseteq\TPOS{\hat t}\tightsetminus\{\emptyset\}}
with \math{(D_0 D_1)\sigma\varphi} fulfilled \wrt\ \redsub\
and 
\bigmath{
              \forall p'\tightin\Delta\stopq
              \inparenthesesinline{
                  \hat t/p'\tightnotin\V
                \ \und\ 
                  (\hat t/p')\sigma\varphi
                  \tightequal
                  (\hat t/p )\sigma\varphi
              }
}
the critical peak, the further reduction of its left part, and the 
required joinability after this reduction can be depicted as follows:%
\footnote{It should be noted 
that the fact that the parallel reduction can be restricted
not only to non-variable positions of \math{\hat t} but also to the same
identical redex \math{(\hat t/p )\sigma\varphi} 
(and the necessity of the analogous 
restriction in the proof)
was especially brought to our attention by 
\gramlichname\ (\cf\ \citegramlicheitel) who already had similar 
but less general ideas on the weakening of overlay joinability.}

\indent
\begin{diagram}[inline]
\hat t\sigma\varphi
&\rred&&t_1\sigma\varphi
\\
&&&\dequal
\\
\dredindex{\omega+\omega,\,p}
&&&\bar u_0
&&&\bar u_0/\bar p_0
&&\transclosureinline{(\antired\cup\tight\lhd)}
&&(\hat t/p)\sigma\varphi
\\
&&&\drefltrans
&&&\drefltrans
\\
t_0\sigma\varphi
&&&\repl{\bar u_0}{\bar p_0}{\bar u_1/\bar p_0}
&&&{\bar u_1/\bar p_0}
\\
\dequal
&&&\dequal
\\
\replpar{t_0}{p'}{\hat t/p}{p'\tightin\Delta}\sigma\varphi
&&&\bar u_1
&&&{\bar u_1/\bar p_1}
&&\transclosureinline{(\antired\cup\tight\lhd)}
&&(\hat t/p)\sigma\varphi
\\
&&&\vdots
&&&\vdots
&&\vdots&&\vdots
\\
&&&\bar u_{\bar n-1}
&&&{\bar u_{\bar n-1}/\bar p_{\bar n-1}}
&&\transclosureinline{(\antired\cup\tight\lhd)}
&&(\hat t/p)\sigma\varphi
\\
\dredparaindex{\omega+\omega,\,\Delta}
&&&\drefltrans
&&&\drefltrans
\\
&&&\repl
     {\bar u_{\bar n-1}}
     {\bar p_{\bar n-1}}
     {\bar u_{\bar n}/\bar p_{\bar n-1}}
&&&{\bar u_{\bar n}/\bar p_{\bar n-1}}
\\
&&&\dequal
\\
\replpar{t_0}{p'}{t_0/p   }{p'\tightin\Delta}\sigma\varphi
&\rrefltrans&&\bar u_{\bar n}
\\
\end{diagram}

\pagebreak

\yestop
\noindent
It is rather easy to see that \superterm-quasi overlay joinability
of a critical peak generalizes the old notion of quasi overlay
joinability: 

In case that \bigmath{\Delta\tightequal\emptyset:}
For 
quasi overlay joinability of \criticalpeaklongform, \ie\ for
\bigmath{
           t_1\sigma\varphi
           \tightequal
           \repl
             {t_0\sigma\varphi}
             {p}
             {t_1\sigma\varphi/p}
;}
\bigmath{
           (t_0/p)\sigma\varphi
           \refltranssub
           w
           \antirefltranssub
           t_1\sigma\varphi/p
           \,\transclosureinline{(\antiredsub\cup\subterm)}\,
           (\hat t/p)\sigma\varphi
;}
we simply choose
\bigmath{
  \bar n
  :=
  1
;}
\bigmath{
  \bar u_0
  :=
  t_1\sigma\varphi
;}
\bigmath{
  \bar u_1
  :=
  \repl{\bar u_0}{p}{w}
;}
and get
\\\linemath{
  t_0\sigma\varphi
  \tightequal
  \repl{t_1\sigma\varphi}{p}{t_0\sigma\varphi/p}
  \penalty-1
  \refltranssub
  \penalty-1
  \repl{t_1\sigma\varphi}{p}{w}
  \tightequal
  \bar u_1
}
and
\LINEmath{
  \bar u_1/p
  \tightequal
  w
  \antirefltranssub
  t_1\sigma\varphi/p
  \tightequal
  \bar u_0/p
  \tightequal
  t_1\sigma\varphi/p
  \,\transclosureinline{(\antiredsub\cup\subterm)}\,
  (\hat t/p)\sigma\varphi
.}

In case that \bigmath{\Delta\tightnotequal\emptyset:}
\superterm-quasi overlay joinability 
of some critical peak,
\wrog\ say
\bigmath{
   ((\repl{l_1}{p}{r_0},
     C_0,
     \Lambda_0),\ 
    (r_1,
     C_1,
     \Lambda_1),\ 
    l_1,\ 
    \sigma,\ 
    p\ 
   )
}
generated by two rules 
\kurzregelindex0,
\kurzregelindex1
(with \wrog\ no variables in common)
due to 
\bigmath{
  \sigma
  \tightequal
  \minmgu
    {\{(l_0,l_1/p)\}}
    {\Y}
}
for
\bigmath{
  \Y
  :=
  \VAR{\kurzregelindex0,\kurzregelindex1}
,}
generalizes quasi overlay joinability
of the critical peaks resulting from overlapping
\kurzregelindex0 into \kurzregelindex1.
While we are not going to discuss the (then obvious) general case 
in detail here,
the case of \bigmath{\Delta\tightequal\{p'\}}
was just discussed before the definition above and we complete this
discussion now as follows:
Defining
\bigmath{
  \hat t
  :=
  l_1
;}
\bigmath{
  t_0
  :=
  \repl{l_1}{p}{r_0}
;}
\bigmath{
  t_1
  :=
  r_1
;}
\bigmath{
  \bar n
  :=
  2
;}
\bigmath{
  \bar u_0
  :=
  t_1\sigma\varphi
;}
\bigmath{
  \bar u_1
  :=
  \repl{\bar u_0}{p}{r_0\sigma\varphi}
;}
\bigmath{
  \bar u_2
  :=
  \repl{\bar u_1}{p'}{r_0\sigma\varphi}
;}
due to (!)\ 
we have
\\\linemath{
                  \replpar
                    {t_0}
                    {p'}
                    {t_0/p}
                    {p'\tightin\Delta}
                  \sigma\varphi
  \tightequal         
  \repl
    {\repl
       {l_1}
       {p}
       {r_0}
    }
    {p'}
    {r_0}
  \sigma\varphi
  \tightequal
  \repl
    {\repl
       {r_1\sigma\varphi}
       {p}
       {r_0\sigma\varphi}
    }
    {p'}
    {r_0\sigma\varphi}
  \tightequal
  \bar u_2
}
and due to (!!)\
we have
\bigmath{
  \bar u_2/p'
  \tightequal
  \bar u_1/p
  \tightequal
  r_0\sigma\varphi
  \antiredsub
  l_0\sigma\varphi
  \,\transclosureinline{\inparenthesesinlinetight{\antired\cup\subterm}}\,
  l_0\sigma\varphi
  \tightequal
  (\hat t/p)\sigma\varphi
}
where by~(!)\
\linemath{
  l_0\sigma\varphi
  \tightequal
  (l_1/p)\sigma\varphi
  \tightequal
  l_1\sigma\varphi/p
  \tightequal
  t_1\sigma\varphi/p
  \tightequal
  \bar u_0/p
}
and
\LINEmath{
  l_0\sigma\varphi
  \tightequal
  (l_1/p)\sigma\varphi
  \tightequal
  (l_1/p')\sigma\varphi
  \tightequal
  l_1\sigma\varphi/p'
  \tightequal
  t_1\sigma\varphi/p'
  \tightequal
  \bar u_0/p'
  \tightequal
  \bar u_1/p'
.}
\\
In the case of an arbitrary \bigmath{\Delta\tightnotequal\emptyset,} 
quasi overlay joinability of 
any two of the 
critical peaks involved implies that the diagram from above then 
looks the following way 
(where \math{\bar n:=\CARD{\{p\}\tightcup\Delta}}):
\begin{diagram}
\hat t\sigma\varphi&&\requal&t_1\sigma\varphi
\\
\dredparaindex{\omega+\omega,\,\{p\}\cup\Delta}
&&&
\dredparaindex{\omega+\omega,\,\{p\}\cup\Delta}
\\
\replpar{t_0}{p'}{t_0/p   }{p'\tightin\Delta}\sigma\varphi
&&\requal&\bar u_{\bar n}
\end{diagram}

\vfill

\pagebreak

\yestop
\noindent
That the wondrous results of quasi overlay joinability 
in the above reported case
can be overcome 
with the new notion of \math\rhd-quasi overlay joinability can be seen
from the following example: 
\begin{example}
\label{ex overlay problems overcome}
%\\
\arr{[t]{lll}
 \consfunsym
&:=
&\{\fsymbol,\gsymbol,\appzero,\cppzero\}
\\\deffunsym
&:=
&\{\plussymbol\}
\\\R_{\,\rm\ref{ex overlay problems overcome}}
&:
&\arr{[t]{r@{\,}l@{\,}ll}
   {\fppeins X}                            &=&{\gppeins X}
   \\
   \appzero                            &=&\cppzero
   \\
   \plusppnoparentheses{\fppeins X}{\fppeins X}&=&\appzero
   \\
   \plusppnoparentheses{\gppeins X}{\gppeins X}&=&\cppzero
   &\rulesugar\ 
   \plusppnoparentheses{\fppeins X}{\fppeins X}\;\boldequal\;\cppzero
}
}

\noindent
Now the unconditional version of\/
\math{\R_{\,\rm\ref{ex overlay problems overcome}}}
is compatible with the lexicographic path ordering \math\rhd\ resulting from
the following precedence on function symbols (in decreasing order):
\fsymbol, 
\gsymbol, 
\appzero, 
\cppzero.
The critical peak 
\bigmath{
  ((\plusppnoparentheses{\gppeins X}{\fppeins X},
    \emptyset,
    0),\ 
   (\appzero,
    \emptyset,
    1),\
   \plusppnoparentheses{\fppeins X}{\fppeins X},\ 
   \emptyset,\ 
   1\ )
}
cannot be quasi overlay joinable because \bigmath{\appzero/1} is undefined.
It is, however, \math\rhd-quasi overlay joinable:
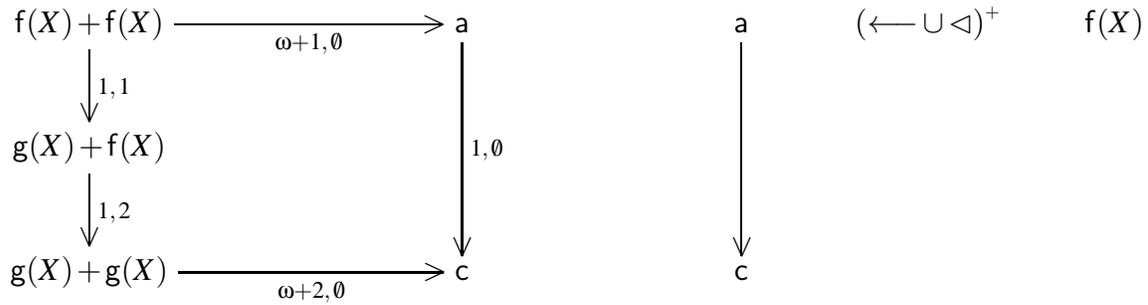
\mycomment{
 because we have
\bigmath{
  \plusppnoparentheses{\gppeins X}{\fppeins X}
  \ \redindex{\R_{\,\rm\ref{ex overlay problems overcome}}}\
  \plusppnoparentheses{\gppeins X}{\gppeins X}
  \ \redindex{\R_{\,\rm\ref{ex overlay problems overcome}}}\
  \appzero
.}}
\begin{diagram}
\plusppnoparentheses{\fppeins X}{\fppeins X}&&\rredindex{\omega+1,\,\emptyset}
&&\appzero
&&&\appzero
&&\transclosureinline{(\antired\cup\tight\lhd)}&&{\fppeins X}
\\
\dredindex{1,\,1}&&
\\
\plusppnoparentheses{\gppeins X}{\fppeins X}&&&&\dredindex{1,\,\emptyset}
&&&\dred
\\
\dredindex{1,\,2}&&
\\
\plusppnoparentheses{\gppeins X}{\gppeins X}&&\rredindex{\omega+2,\,\emptyset}
&&\cppzero
&&&\cppzero
\\
\end{diagram}
\end{example}

\yestop
\yestop
\yestop
\yestop
\yestop
\noindent
That the \math\Delta\ in the notion of \tight\superterm-quasi overlay 
joinability cannot be restricted to be empty can be seen from 
Example~\ref{ex toll}.

\vfill

\pagebreak

\section{Some Unconditional Examples}
\label{sect unconditional examples}

Our main goal in this and 
the following sections is to find confluence criteria that do not depend
on termination arguments but on the structure of the joinability of critical
peaks only. 
Finally in \sect~\ref{section now termination} we will investigate
how termination can strengthen our criteria.
Up to then, however, we are not going to use termination arguments.
Instead, we are looking for confluence criteria of the form
``If all critical peaks of a \ldots\ 
(\eg\ normal, left-linear, \etc)
rule system are joinable according to the pattern \ldots\ 
(\eg\ shallow joinable, parallel closed, \etc)
then the reduction relation is confluent.''

First we want to make clear that this approach has its limits. We do this 
by giving some examples. 
To distinguish confluent from non-confluent examples the rule systems
of the latter ones are displayed in a box 
at the right margin while in a connected box to the left
we list the example's crucial properties, concerning
joinability structure of their critical peaks,
variable occurrence,  
condition properties,
\etc.
The reader should not try to understand the sophisticated joinability labels
in the boxes at a first reading. 
This is not necessary for understanding the examples. 
The sophisticated joinability labels are only needed for 
\sect~\ref{sect non-terminating confluence}.

In this section we start with some unconditional examples.
The first one shows that left-linearity is essential%
\footnote{Since this counterexample for confluence
is unconditional it must be non-terminating of course. For conditional
systems, however, left-linearity is essential also for terminating systems
for joinability of critical peaks to imply confluence, \cf\ the transformation
described in \sect~\ref{sect transformation} applied to 
Example~\ref{ex cpw not normal} as described in 
\sect~\ref{sect transformation}.%
}:

\begin{example}[\citehuet]
\label{ex not left linear}
\\
\fbox{\bf
\begin{tabular}[t]{@{}l}
No Critical Peaks
\\\hline 
{\em Not Left-Linear}
\\\hline 
Unconditional
\\\hline 
{\em Not Terminating}
\end{tabular}
}%
\hrulefill
\fbox{
\arr{[t]{lll}
\consfunsym
&:=
&\{\zeropp, \ssymbol, \cppzero, \dppzero\}
\\\deffunsym
&:=
&\{\plussymbol\}
\\
}
\arr{[t]{|lll}
\R_{\,\rm\ref{ex not left linear}}
&:
&\arr{[t]{r@{\,}l@{\,}l}
    \zeropp
   &=
   &\spp\zeropp
   \\\plusppnoparentheses X X
   &=
   &\cppzero
   \\\plusppnoparentheses X{\spp X}
   &=
   &\dppzero
}
}
}

\noindent
There are no critical peaks.
Nevertheless, \redindex{\R_{\,\rm\ref{ex not left linear}},\emptyset} is not confluent:
\begin{diagram}
  \cppzero&&\rantiredindex{\omega+1}&&\plusppnoparentheses\zeropp\zeropp&&&&
\\&&&&\dredindex1&&&&
\\\dequal&&&&\plusppnoparentheses\zeropp{\spp\zeropp}
&&\rredindex{\omega+1}&&\dppzero
\\&&&&\dredindex1&&&&
\\\cppzero&&\rantiredindex{\omega+1}
&&\plusppnoparentheses{\spp\zeropp}{\spp\zeropp}
&&&&\dequal
\\
\dequal&&&&\dredindex1&&&&
\\
\\
\vdots&&\vdots&&\vdots&&\vdots&&\vdots
\end{diagram}
\yestop
\yestop
\end{example}

\pagebreak

\begin{example}
\label{ex a}
\\
\fbox{\bf
\begin{tabular}[t]{@{}l}
\begin{diagram}[height=1em,width=1.5em,inline,top]
   \circ&\rredindex{\omega+1}&\circ
 \\&&\dredindex{\omega+1}
 \\
 \\\dredindex1&&\circ
 \\&&\dredindex1
 \\
 \\\circ&\requal&\circ
\end{diagram}~~~
\\
\\
\math{\,\omega}-Level {[{[Weak]} Parallel]} Joinable
\\
\math{\,\omega}-Level Strongly Joinable
\\
{\em Not \math\omega-Shallow {[{[Noisy]} Parallel]} Joinable up to \math\omega}
\\\hline
Ground
\\\hline
Unconditional
\\\hline
{\em Not Terminating}
\end{tabular}
}%
\hrulefill
\fbox{
\arr{[t]{lll}
 \consfunsym
&:=
&\{\appzero, \bppzero, \cppzero, \dppzero\}
\\\deffunsym
&:=
&\{\fsymbol\}
\\\R_{\,\rm\ref{ex a}}
&:
&\arr{[t]{r@{\,}l@{\,}l}
     \appzero
   &=
   &\cppzero
   \\\bppzero
   &=
   &\dppzero
   \\\fppeins\appzero
   &=
   &\fppeins\bppzero
   \\\fppeins\bppzero
   &=
   &\fppeins\appzero
}
}
}

\noindent
The critical peaks are all of the form \math{(0,1)} and can be closed as
follows:
\begin{diagram}
\fppeins\appzero&&\rredindex{\omega+1}&&\fppeins\bppzero
&&\fppeins\bppzero&&\rredindex{\omega+1}&&\fppeins\appzero
\\
&&&&\dredindex{\omega+1,\,\emptyset}
&&&&&&\dredindex{\omega+1,\,\emptyset}
\\
\dredindex{1,\,1}&&&&\fppeins\appzero
&&\dredindex{1,\,1}&&&&\fppeins\bppzero
\\
&&&&\dredindex{1,\,1}
&&&&&&\dredindex{1,\,1}
\\
\fppeins\cppzero&&\requal&&\fppeins\cppzero
&&\fppeins\dppzero&&\requal&&\fppeins\dppzero
\end{diagram}
However, \redindex{\R_{\,\rm\ref{ex a}},\emptyset} is not confluent:
\begin{diagram}
\fppeins\cppzero&&\rantiredindex{1,\,1}&&\fppeins\appzero
&&\pile{\rredindex{\omega+1,\,\emptyset}\\\rantiredindex{\omega+1,\,\emptyset}}
&&\fppeins\bppzero&&\rredindex{1,\,1}&&\fppeins\dppzero
\end{diagram}
\yestop
\end{example}

\vfill

\pagebreak

\begin{example}
\label{ex b}
%\\
\arr{[t]{lll}
 \consfunsym
&:=
&\{\zeropp, \ssymbol,\psymbol\}
\\\deffunsym
&:=
&\{\plussymbol\}
\\
}
\arr{[t]{|lll}
\R_{\,\rm\ref{ex b}}
&:
&\arr{[t]{r@{\,}l@{\,}l}
     \spp{\ppp X}&=&X
   \\\ppp{\spp X}&=&X
   \\\plusppnoparentheses\zeropp Y&=&Y
   \\\plusppnoparentheses{\spp X}Y&=&\spp{\plusppnoparentheses X Y}
   \\\plusppnoparentheses{\ppp X}Y&=&\ppp{\plusppnoparentheses X Y}
}
}

\noindent
The critical peaks are all of the form \math{(0,1)} and can be closed as
follows:
\begin{diagram}
\plusppnoparentheses{\spp{\ppp X}}Y&&\rredindex{\omega+1}
&&\spp{\plusppnoparentheses{\ppp X}Y}
&&&\plusppnoparentheses{\ppp{\spp X}}Y&&\rredindex{\omega+1}
&&\ppp{\plusppnoparentheses{\spp X}Y}
\\
&&&&\dredindex{\omega+1,\,1}
&&&&&&&\dredindex{\omega+1,\,1}
\\
\dredindex{1,\,1}&&&&\spp{\ppp{\plusppnoparentheses X Y}}
&&&\dredindex{1,\,1}&&&&\ppp{\spp{\plusppnoparentheses X Y}}
\\
&&&&\dredindex{1,\,\emptyset}
&&&&&&&\dredindex{1,\,\emptyset}
\\
\plusppnoparentheses X Y&&\requal&&\plusppnoparentheses X Y
&&&\plusppnoparentheses X Y&&\requal&&\plusppnoparentheses X Y
\end{diagram}
Since the reduction relation is terminating, we have confluence here.
However, note that the structure of the joinability of the critical
peaks is identical to that of Example~\ref{ex a} (with the exception of the 
positions). 
Thus, argumentation on the
joinability structure of critical peaks must fail to infer confluence for
this example
(at least if we do not take positions into account).
\end{example}

\vfill

\pagebreak

\yestop
\noindent
The following example results from Example~\ref{ex a} 
just by changing `\appzero' and `\bppzero' into non-constructors.
While Example~\ref{ex a} was able to discourage generalizations of 
Theorem~\ref{theorem level parallel closed}, by the slight change the
following example is able to discourage generalizations of 
Theorem~\ref{theorem parallel closed} regarding the required
\math\omega-shallow parallel closedness 
(for part (I) of Theorem~\ref{theorem parallel closed}),
\math\omega-shallow noisy anti-closedness
(for part (II)),
or 
\math\omega-shallow closedness 
(for parts (III) and (IV))
of the non-overlays of the form \math{(1,1)}.

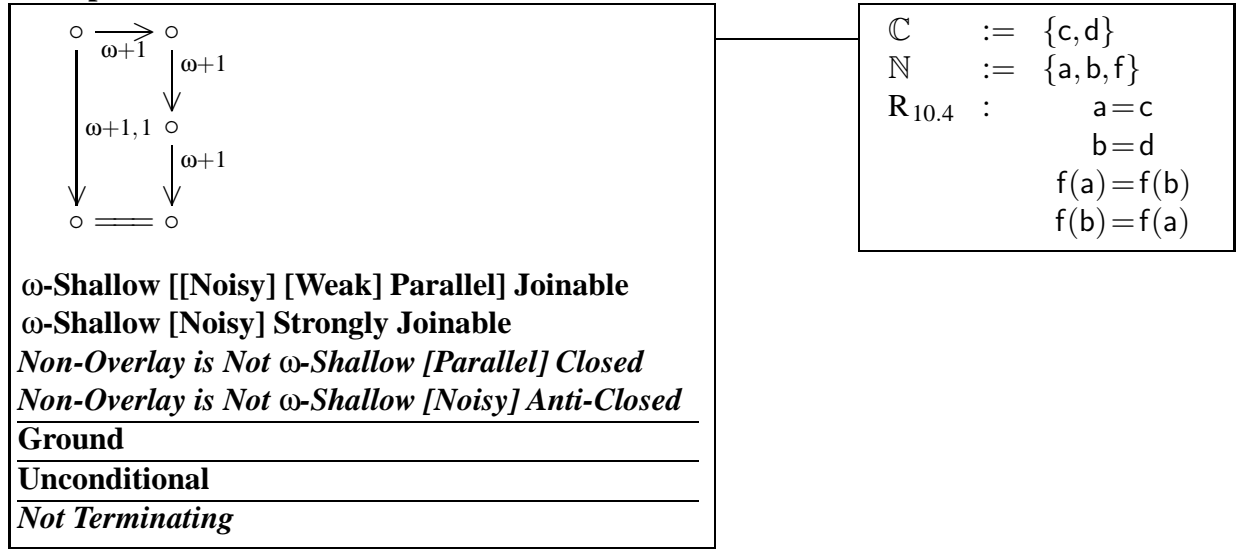
\begin{example}
\label{ex a modified}
\\
\fbox{\bf
\begin{tabular}[t]{@{}l}
\begin{diagram}[height=1em,width=1.5em,inline,top]
   \circ&\rredindex{\omega+1}&\circ
 \\&&\dredindex{\omega+1}
 \\
 \\\dredindex{\omega+1,\,1}&&\circ
 \\&&\dredindex{\omega+1}
 \\
 \\\circ&\requal&\circ
\end{diagram}~~~
\\
\\
\math{\,\omega}-Shallow {[{[Noisy]} {[Weak]} Parallel]} Joinable
\\
\math{\,\omega}-Shallow {[Noisy]} Strongly Joinable
\\
{\em Non-Overlay is Not \math\omega-Shallow {[Parallel]} Closed}
\\
{\em Non-Overlay is Not \math\omega-Shallow {[Noisy]} Anti-Closed}
\\\hline
Ground
\\\hline
Unconditional
\\\hline
{\em Not Terminating}
\end{tabular}
}%
\hrulefill
\fbox{
\arr{[t]{lll}
 \consfunsym
&:=
&\{\cppzero, \dppzero\}
\\\deffunsym
&:=
&\{\appzero, \bppzero, \fsymbol\}
\\\R_{\,\rm\ref{ex a modified}}
&:
&\arr{[t]{r@{\,}l@{\,}l}
     \appzero
   &=
   &\cppzero
   \\\bppzero
   &=
   &\dppzero
   \\\fppeins\appzero
   &=
   &\fppeins\bppzero
   \\\fppeins\bppzero
   &=
   &\fppeins\appzero
}
}
}

\noindent
The critical peaks are all of the form \math{(1,1)} now and can be closed as
follows:
\notop\halftop
\begin{diagram}
\fppeins\appzero&&\rredindex{\omega+1}&&\fppeins\bppzero
&&\fppeins\bppzero&&\rredindex{\omega+1}&&\fppeins\appzero
\\
&&&&\dredindex{\omega+1,\,\emptyset}
&&&&&&\dredindex{\omega+1,\,\emptyset}
\\
\dredindex{\omega+1,\,1}&&&&\fppeins\appzero
&&\dredindex{\omega+1,\,1}&&&&\fppeins\bppzero
\\
&&&&\dredindex{\omega+1,\,1}
&&&&&&\dredindex{\omega+1,\,1}
\\
\fppeins\cppzero&&\requal&&\fppeins\cppzero
&&\fppeins\dppzero&&\requal&&\fppeins\dppzero
\end{diagram}
However, \redindex{\R_{\,\rm\ref{ex a modified}},\emptyset} is not confluent:
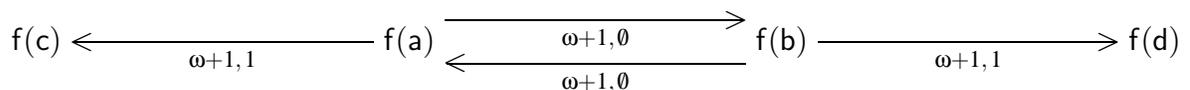
\notop
\begin{diagram}
\fppeins\cppzero&&\rantiredindex{\omega+1,\,1}&&\fppeins\appzero
&&\pile{\rredindex{\omega+1,\,\emptyset}\\\rantiredindex{\omega+1,\,\emptyset}}
&&\fppeins\bppzero&&\rredindex{\omega+1,\,1}&&\fppeins\dppzero
\end{diagram}
\yestop
\end{example}

\pagebreak

\begin{example}
\label{ex c}
\\
\fbox{\bf
\begin{tabular}[t]{@{}l}
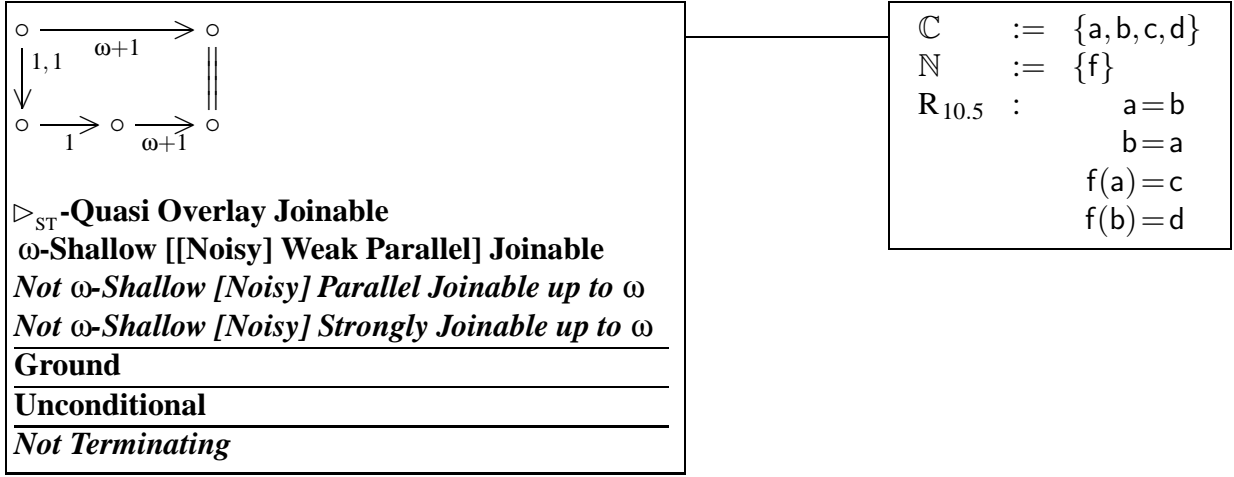
\begin{diagram}[height=1em,width=1.5em,inline,top]
\circ&&\rredindex{\omega+1}&&\circ
\\\dredindex{1,\,1}&&&&\dequal
\\
\\\circ&\rredindex{1}&\circ&\rredindex{\omega+1}&\circ
\end{diagram}
\\
\\
\math{\!\rm\superterm}-Quasi Overlay Joinable
\\
\math{\,\omega}-Shallow {[{[Noisy]} Weak Parallel]} Joinable
\\
{\em Not \math\omega-Shallow {[Noisy]} Parallel Joinable up to \math\omega}
\\
{\em Not \math\omega-Shallow {[Noisy]} Strongly Joinable up to \math\omega}
\\\hline
Ground
\\\hline
Unconditional
\\\hline
{\em Not Terminating}
\end{tabular}
}%
\hrulefill
\fbox{
\arr{[t]{lll}
 \consfunsym
&:=
&\{\appzero, \bppzero, \cppzero, \dppzero\}
\\\deffunsym
&:=
&\{\fsymbol\}
\\\R_{\,\rm\ref{ex c}}
&:
&\arr{[t]{r@{\,}l@{\,}l}
     \appzero
   &=
   &\bppzero
   \\\bppzero
   &=
   &\appzero
   \\\fppeins\appzero
   &=
   &\cppzero
   \\\fppeins\bppzero
   &=
   &\dppzero
}
}
}

\noindent
The critical peaks are all of the form \math{(0,1)} and can be closed as
follows:
\begin{diagram}
  \fppeins\appzero&&\rredindex{\omega+1}&&\cppzero
&&\fppeins\bppzero&&\rredindex{\omega+1}&&\dppzero
\\
\\\dredindex{1,\,1}&&&&\dequal
&&\dredindex{1,\,1}&&&&\dequal
\\
\\
\fppeins\bppzero&\rredindex{1,\,1}&\fppeins\appzero
&\rredindex{\omega+1,\,\emptyset}
&\cppzero
&&\fppeins\appzero&\rredindex{1,\,1}&\fppeins\bppzero
&\rredindex{\omega+1,\,\emptyset}
&\dppzero
\end{diagram}
However, \redindex{\R_{\,\rm\ref{ex c}},\emptyset} is not confluent:
\begin{diagram}
\cppzero&&\rantiredindex{\omega+1,\,\emptyset}
&&\fppeins\appzero
&&\pile{\rredindex{1,\,1}\\\rantiredindex{1,\,1}}
&&\fppeins\bppzero
&&\rredindex{\omega+1,\,\emptyset}
&&\dppzero
\end{diagram}
\end{example}

\begin{example}
\label{ex d}
%\\
\arr{[t]{lll|}
 \consfunsym
&:=
&\{\zeropp, \ssymbol,\psymbol\}
\\\deffunsym
&:=
&\{\plussymbol\}
\\
}
\arr{[t]{lll}
\R_{\,\rm\ref{ex d}}
&:
&\R_{\,\rm\ref{ex b}}
\mbox{~~}+\mbox{~~}
%\\&&
\arr{[t]{r@{\,}l@{\,}l}
     X&=&\spp{\ppp X}
   \\X&=&\ppp{\spp X}
}
}
\\
Note that we have added two rules to the system from Example~\ref{ex b}:
The critical peaks of the form \math{(0,1)} 
of Example~\ref{ex b} still exist but can now be closed in different way;
\eg, the first one can be closed as follows:
\begin{diagram}
\plusppnoparentheses{\spp{\ppp X}}Y&&&&\rredindex{\omega+1}
&&&&\spp{\plusppnoparentheses{\ppp X}Y}
\\
\\
\dredindex{1,\,1}&&&&&&&&\dequal
\\
\\
\plusppnoparentheses X Y&&\rredindex{1,\,1}
&&\plusppnoparentheses{\spp{\ppp X}}Y
&&\rredindex{\omega+1,\,\emptyset}&&\spp{\plusppnoparentheses{\ppp X}Y}
\end{diagram}
Since \redindex{\R_{\,\rm\ref{ex b}},\emptyset} is confluent and
\bigmath{
  \redindex{\R_{\,\rm\ref{ex b}},\emptyset}\subseteq
  \redindex{\R_{\,\rm\ref{ex d}},\emptyset}\subseteq
  \congruindex{\R_{\,\rm\ref{ex b}},\emptyset}
,}
\redindex{\R_{\,\rm\ref{ex d}},\emptyset} is confluent, too
(\cf\ Lemma~\ref{lemma church rosser}). 
However, note that the structure of the joinability of the critical
peaks is identical to that of Example~\ref{ex c}. 
Thus, argumentation on the
joinability structure of critical peaks must fail to infer confluence for
this example.
\end{example}

\pagebreak

\yestop
\yestop
\yestop
\noindent
According to Lemma~3.2 of \citehuet, unconditional left- and right-linear
rule systems with strongly joinable critical peaks are [strongly] confluent.
That the severe restriction of right-linearity is essential here can be seen 
from the following example:

\begin{example}[\levyname\ as cited in \citehuet]
\label{ex levy a}
\\
\begin{tabular}[t]{@{}c@{}}
\fbox{\bf
\begin{tabular}[t]{@{}l}
\begin{diagram}[height=1em,width=1.5em,inline,top]
\circ&&&\rredindex{\omega+1}&&&\circ
\\\dredindex{1}&&&&&&\dredindex{\omega+1}
\\
\\\circ&&&\rredindex{\omega+1}&&&\circ
\end{diagram}
\\
\\
\math\omega-Level 
{[Parallel]} 
Joinable
\\
{[\math\omega-Level]}
{[Strongly]} 
Joinable
\\
{\em Not \math\omega-Shallow {[Noisy]} Parallel Joinable up to \math\omega}
\\
{\em Not \math\omega-Shallow {[Noisy]} Strongly Joinable up to \math\omega}
\\\hline
Left-Linear 
\\
Right-Linear Constructor Rules
\\
{\em Not Right-Linear}
\\\hline
Unconditional
\\\hline
{\em Not Terminating}
\end{tabular}
}
\\\math\mid
\\
\fbox{\bf
\begin{tabular}[t]{@{}l}
\begin{diagram}[height=1em,width=1.5em,inline,top]
\circ&&&\rredindex{\omega+1}&&&\circ
\\\dredindex{1}&&&&&&\dequal
\\
\\\circ&\rredindex{\omega+1,\,\emptyset}
&&\circ&\rredindex{\omega+1,\,\emptyset}&&\circ
\end{diagram}
\\
\\
{[\math\omega-Shallow]} 
Joinable
\\
{\em Not \math\omega-Shallow {[Noisy]} Parallel Joinable up to \math\omega}
\\
{\em Not \math\omega-Shallow {[Noisy]} Strongly Joinable up to \math\omega}
\\\hline
Left-Linear 
\\
Right-Linear Constructor Rules
\\
{\em Not Right-Linear}
\\\hline
Unconditional
\\\hline
{\em Not Terminating}
\end{tabular}
}
\\
\end{tabular}%
\hrulefill
\fbox{
\arr{[t]{lll}
 \consfunsym
&:=
&\{\appzero, \bppzero, \cppzero, \dppzero\}
\\\deffunsym
&:=
&\{\plussymbol,\minussymbol\}
\\\R_{\,\rm\ref{ex levy a}}
&:
&\arr{[t]{r@{\,}l@{\,}l}
     \appzero&=&\cppzero
   \\\bppzero&=&\dppzero
   \\\plusppnoparentheses\appzero\appzero&=
    &\minusppnoparentheses\bppzero\bppzero
   \\\plusppnoparentheses\cppzero X&=
    &\plusppnoparentheses X X
   \\\plusppnoparentheses X\cppzero&=
    &\plusppnoparentheses X X
   \\\minusppnoparentheses\bppzero\bppzero&=
    &\plusppnoparentheses\appzero\appzero
   \\\minusppnoparentheses\dppzero X&=
    &\minusppnoparentheses X X
   \\\minusppnoparentheses X\dppzero&=
    &\minusppnoparentheses X X
}
}
}

\noindent
There are only four critical peaks and they are all of the form \math{(0,1)}.
Using the symmetry of \plussymbol\ in its arguments as well the symmetry of
\appzero, \cppzero, \plussymbol\ with
\bppzero, \dppzero, \minussymbol, all other 
critical peaks are symmetric to the 
following one, which can be closed in the following two different ways:
\begin{diagram}
\plusppnoparentheses\appzero\appzero
&&&\rredindex{\omega+1}
&&&\minusppnoparentheses\bppzero\bppzero
\\\dredindex{1,\,1}&&&&&&\dredindex{\omega+1,\,\emptyset}
\\\plusppnoparentheses\cppzero\appzero
&&&\rredindex{\omega+1,\,\emptyset}
&&&\plusppnoparentheses\appzero\appzero
\end{diagram}
\begin{diagram}
\plusppnoparentheses\appzero\appzero
&&&\rredindex{\omega+1}
&&&\minusppnoparentheses\bppzero\bppzero
\\\dredindex{1,\,1}&&&&&&\dequal
\\\plusppnoparentheses\cppzero\appzero
&\rredindex{\omega+1,\,\emptyset}
&&\plusppnoparentheses\appzero\appzero
&\rredindex{\omega+1,\,\emptyset}
&&\minusppnoparentheses\bppzero\bppzero
\end{diagram}

\pagebreak

\noindent
Nevertheless, \redindex{\R_{\,\rm\ref{ex levy a}},\emptyset} is not confluent:
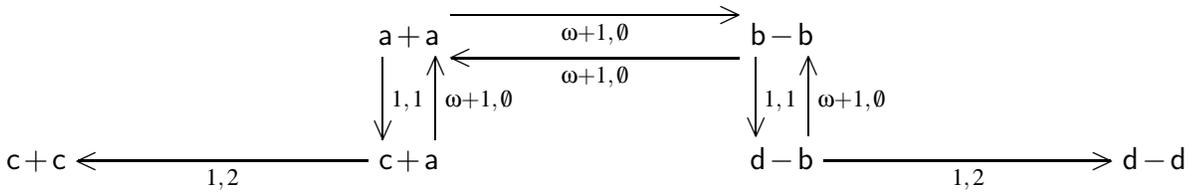
\notop
\begin{diagram}
&&&&\plusppnoparentheses\appzero\appzero
&&\pile{\rredindex{\omega+1,\,\emptyset}\\\rantiredindex{\omega+1,\,\emptyset}}
&&\minusppnoparentheses\bppzero\bppzero
&&&&
\\&&&&\dredindex{1,\,1~}\dantiredindex{\omega+1,\,\emptyset}
&&&&\dredindex{1,\,1~}\dantiredindex{\omega+1,\,\emptyset}
&&&&
\\\plusppnoparentheses\cppzero\cppzero
&&\rantiredindex{1,\,2}&&\plusppnoparentheses\cppzero\appzero
&&&&\minusppnoparentheses\dppzero\bppzero
&&\rredindex{1,\,2}&&\minusppnoparentheses\dppzero\dppzero
\end{diagram}

\yestop
\yestop
\yestop
\noindent
We now use the same \math{\R_{\,\rm\ref{ex levy a}}} to show that even another
structure of joinability is insufficient for confluence. We do this
by changing the separation into constructors and non-constructors:

\yestop
\noindent
\begin{tabular}[t]{@{}l@{}}
\fbox{\bf
\begin{tabular}[t]{@{}l}
\begin{diagram}[height=1em,width=1.5em,inline,top]
\circ&&&\rredindex{\omega+1}&&&\circ
\\\dredindex{\omega+1,\,1}&&&&&&\dredindex{\omega+1}
\\
\\\circ&&&\rredindex{1}&&&\circ
\end{diagram}
\\
\\
\math\omega-Shallow {[{[Noisy]} {[Weak]} Parallel]} Joinable
\\
{\em Non-Overlay is Not \math\omega-Shallow {[Parallel]} Closed 
 \mbox{} \mbox{}}
\\
{[\math\omega-Shallow]} Strongly Joinable
\\
\math\omega-Shallow Anti-Closed
\\\hline
Left-Linear 
\\
{\em {[Constructor Rules]} Not Right-Linear}
\\
\hline
Unconditional
\\\hline
{\em Not Terminating}
\end{tabular}
}
\\\mbox{}\hfill\math\mid\hfill\mbox{}
\\
\fbox{\bf
\begin{tabular}[t]{@{}l}
\begin{diagram}[height=1em,width=1.5em,inline,top]
\circ&&&\rredindex{\omega+1}&&&\circ
\\\dredindex{\omega+1,\,1}&&&&&&\dequal
\\
\\\circ&\rredindex{1,\,\emptyset}
&&\circ&\rredindex{\omega+1,\,\emptyset}&&\circ
\end{diagram}
\\
\\
\math\omega-Shallow {[{[Noisy]} Weak Parallel]} Joinable
\\
{\em Non-Overlay is Not \math\omega-Shallow {[Parallel]} Closed}
\\
\math\omega-Shallow Strongly Joinable
\\
\math\omega-Shallow Anti-Closed
\\\hline
Left-Linear 
\\
{\em {[Constructor Rules]} Not Right-Linear}
\\\hline
Unconditional
\\\hline
{\em Not Terminating}
\end{tabular}
}
\\
\end{tabular}%
\hrulefill
\fbox{
\arr{[t]{lll}
 \consfunsym
&:=
&\{\cppzero, \dppzero, \plussymbol,\minussymbol\}
\\\deffunsym
&:=
&\{\appzero, \bppzero\}
\\\R_{\,\rm\ref{ex levy a}}
&:
&\arr{[t]{r@{\,}l@{\,}l}
     \plusppnoparentheses\cppzero X&=
    &\plusppnoparentheses X X
   \\\plusppnoparentheses X\cppzero&=
    &\plusppnoparentheses X X
   \\\minusppnoparentheses\dppzero X&=
    &\minusppnoparentheses X X
   \\\minusppnoparentheses X\dppzero&=
    &\minusppnoparentheses X X
   \\\appzero&=&\cppzero
   \\\bppzero&=&\dppzero
   \\\plusppnoparentheses\appzero\appzero&=
    &\minusppnoparentheses\bppzero\bppzero
   \\\minusppnoparentheses\bppzero\bppzero&=
    &\plusppnoparentheses\appzero\appzero
}
}
}

\pagebreak

\noindent
Note that the rule system is not changed, but only reordered to have
the constructor rules precede the non-constructor rules.
The rewrite relation \redindex{\R_{\,\rm\ref{ex levy a}},\emptyset} 
is not changed by this constructor re-declaration. 
(Note \math{X\tightin\Vsig}.)
The critical peaks 
only have changed their form from
\math{(0,1)} to \math{(1,1)} and
are still all symmetric to the following one that closes 
in the two following ways:
%\notop
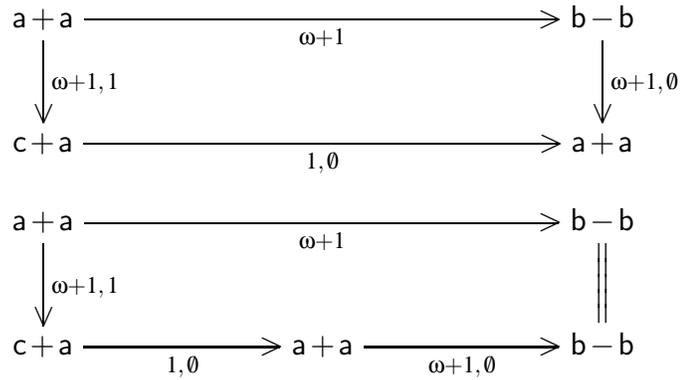
\begin{diagram}
\plusppnoparentheses\appzero\appzero
&&&\rredindex{\omega+1}
&&&\minusppnoparentheses\bppzero\bppzero
\\
\dredindex{\omega+1,\,1}&&&&&&\dredindex{\omega+1,\,\emptyset}
\\
\plusppnoparentheses\cppzero\appzero
&&&\rredindex{1,\,\emptyset}
&&&\plusppnoparentheses\appzero\appzero
\end{diagram}
\begin{diagram}
\plusppnoparentheses\appzero\appzero
&&&\rredindex{\omega+1}
&&&\minusppnoparentheses\bppzero\bppzero
\\
\dredindex{\omega+1,\,1}&&&&&&\dequal
\\
\plusppnoparentheses\cppzero\appzero
&\rredindex{1,\,\emptyset}
&&\plusppnoparentheses\appzero\appzero
&\rredindex{\omega+1,\,\emptyset}
&&\minusppnoparentheses\bppzero\bppzero
\end{diagram}

\noindent
Finally, the divergence looks the following way now
(Please note that now \redindex{\R_{\,\rm\ref{ex levy a}},\emptyset,\omega} and
\redindex{\R_{\,\rm\ref{ex levy a}},\emptyset} are commuting, 
which was not the case before.):
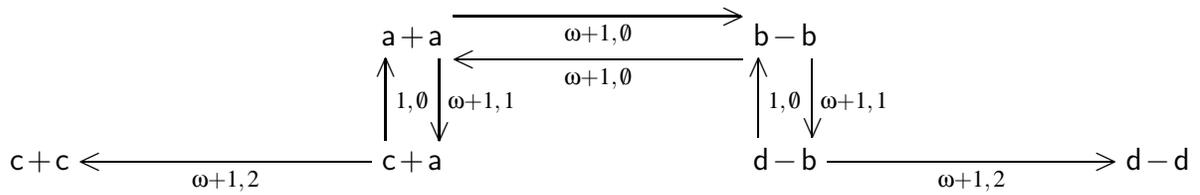
\notop\halftop
\begin{diagram}
&&&&\plusppnoparentheses\appzero\appzero
&&\pile{\rredindex{\omega+1,\,\emptyset}\\\rantiredindex{\omega+1,\,\emptyset}}
&&\minusppnoparentheses\bppzero\bppzero&&&&
\\
&&&&\dantiredindex{1,\,\emptyset~}\dredindex{\omega+1,\,1}
&&&&\dantiredindex{1,\,\emptyset~}\dredindex{\omega+1,\,1}&&&&
\\
\plusppnoparentheses\cppzero\cppzero
&&\rantiredindex{\omega+1,\,2}&&\plusppnoparentheses\cppzero\appzero
&&&&\minusppnoparentheses\dppzero\bppzero
&&\rredindex{\omega+1,\,2}&&\minusppnoparentheses\dppzero\dppzero
\end{diagram}
\end{example}

\vfill

\pagebreak

\yestop
\noindent
The following example is a slight variation of Example~\ref{ex levy a}
which is interesting \wrt\ Example~\ref{ex asso}.

\begin{example}
\label{ex for asso}

\noindent
\begin{tabular}[t]{@{}l@{}}
\fbox{\bf
\begin{tabular}[t]{@{}l}
\begin{diagram}[height=1em,width=1.5em,inline,top]
\circ&&&\rredindex{\omega+1}&&&\circ
\\\dredindex{\omega+1,\,1}&&&&&&\dredindex{\omega+1,\,\emptyset}
\\
\\\circ&\rredindex{1,\,\emptyset}&&\circ&\rredindex{1,\,2}&&\circ
\end{diagram}
\\
\\
\math\omega-Shallow {[{[Noisy]} Parallel]} Joinable
\\
{\em Non-Overlay is Not \math\omega-Shallow {[Parallel]} Closed}
\\
\math\omega-Shallow Strongly Joinable
\\
\math\omega-Shallow Anti-Closed
\\
\hline
Left-Linear 
\\
{\em {[Constructor Rules]} Not Right-Linear}
\\\hline
Unconditional
\\\hline
{\em Not Terminating}
\end{tabular}
}
\\\mbox{}\hfill\math\mid\hfill\mbox{}
\\
\fbox{\bf
\begin{tabular}[t]{@{}l}
\begin{diagram}[height=1em,width=1.5em,inline,top]
\circ&&&\rredindex{\omega+1}&&&\circ
\\\dredindex{\omega+1,\,1}&&&&&&\dequal
\\
\\\circ&\rredindex{1,\,\emptyset}
&\circ&\rredindex{1,\,2}&\circ&\rredindex{\omega+1,\,\emptyset}&\circ
\end{diagram}
\\
\\
\math\omega-Shallow {[{[Noisy]} Weak Parallel]} Joinable
\\
{\em Non-Overlay is Not \math\omega-Shallow {[Parallel]} Closed}
\\
\math\omega-Shallow Strongly Joinable
\\
\math\omega-Shallow Anti-Closed
\\\hline
Left-Linear 
\\
{\em {[Constructor Rules]} Not Right-Linear}
\\\hline
Unconditional
\\\hline
{\em Not Terminating}
\end{tabular}
}
\\
\end{tabular}%
\hrulefill
\fbox{
\arr{[t]{lll}
 \consfunsym
&:=
&\{\cppzero, \dppzero, \plussymbol,\minussymbol,\fsymbol,\gsymbol\}
\\\deffunsym
&:=
&\{\appzero, \bppzero\}
\\\R_{\,\rm\ref{ex for asso}}
&:
&\arr{[t]{r@{\,}l@{\,}l}
     \plusppnoparentheses\cppzero X &=
    &\plusppnoparentheses X{\fppeins X}
   \\\plusppnoparentheses X\cppzero &=
    &\plusppnoparentheses X{\fppeins X}
   \\\fppeins X                     &=
    &X
   \\\minusppnoparentheses\dppzero X&=
    &\minusppnoparentheses X{\gppeins X}
   \\\minusppnoparentheses X\dppzero&=
    &\minusppnoparentheses X{\gppeins X}
   \\\gppeins X                     &=
    &X
   \\\appzero&=&\cppzero
   \\\bppzero&=&\dppzero
   \\\plusppnoparentheses\appzero\appzero&=
    &\minusppnoparentheses\bppzero\bppzero
   \\\minusppnoparentheses\bppzero\bppzero&=
    &\plusppnoparentheses\appzero\appzero
}
}
}

\noindent
There are only four critical peaks and they are all of the form \math{(1,1)}.
Using the symmetry of \plussymbol\ in its relevant
arguments as well the symmetry of
\appzero, \cppzero, \plussymbol, \fsymbol\ with
\bppzero, \dppzero, \minussymbol, \gsymbol,
all other 
critical peaks are symmetric to the 
following one, which can be closed in the following two different ways:
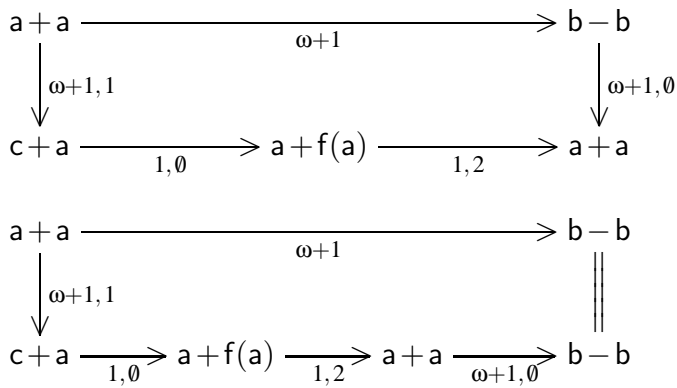
\begin{diagram}
\plusppnoparentheses\appzero\appzero
&&&\rredindex{\omega+1}
&&&\minusppnoparentheses\bppzero\bppzero
\\\dredindex{\omega+1,\,1}&&&&&&\dredindex{\omega+1,\,\emptyset}
\\\plusppnoparentheses\cppzero\appzero
&&\rredindex{1,\,\emptyset}
&\plusppnoparentheses\appzero{\fppeins\appzero}
&\rredindex{1,\,2}
&&\plusppnoparentheses\appzero\appzero
\end{diagram}
\begin{diagram}
\plusppnoparentheses\appzero\appzero
&&&\rredindex{\omega+1}
&&&\minusppnoparentheses\bppzero\bppzero
\\\dredindex{\omega+1,\,1}&&&&&&\dequal
\\\plusppnoparentheses\cppzero\appzero
&\rredindex{1,\,\emptyset}
&\plusppnoparentheses\appzero{\fppeins\appzero}
&\rredindex{1,\,2}
&\plusppnoparentheses\appzero\appzero
&\rredindex{\omega+1,\,\emptyset}
&\minusppnoparentheses\bppzero\bppzero
\end{diagram}

\pagebreak

\noindent
Finally, the divergence looks the following way now:
\notop\halftop
\begin{diagram}
&&\plusppnoparentheses\appzero\appzero
&\requal&\plusppnoparentheses\appzero\appzero
&&\pile{\rredindex{\omega+1,\,\emptyset}\\\rantiredindex{\omega+1,\,\emptyset}}
&&\minusppnoparentheses\bppzero\bppzero
&\requal&\minusppnoparentheses\bppzero\bppzero
&&
\\
&&\dredindex{\omega+1,\,1}
&&\dantiredindex{1,\,2}
&&
&&\dantiredindex{1,\,2}
&&\dredindex{\omega+1,\,1}
&&
\\
\plusppnoparentheses\cppzero\cppzero
&\rantiredindex{\omega+1,\,2}&\plusppnoparentheses\cppzero\appzero
&\rredindex{1,\,\emptyset}
&\plusppnoparentheses\appzero{\fppeins\appzero}
&&&&
\minusppnoparentheses\bppzero{\gppeins\bppzero}&\rantiredindex{1,\,\emptyset}
&\minusppnoparentheses\dppzero\bppzero
&\rredindex{\omega+1,\,2}&\minusppnoparentheses\dppzero\dppzero
\end{diagram}
\end{example}

\yestop
\yestop
\yestop
\begin{example}
\label{ex asso}
%\\
\arr{[t]{lll}
 \consfunsym
&:=
&\{\zeropp\}
\\\deffunsym
&:=
&\{\plussymbol\}
\\\R_{\,\rm\ref{ex asso}}
&:
&\arr{[t]{r@{\,}l@{\,}l}
 \plusppnoparentheses{\pluspp X Y}Z&=&\plusppnoparentheses X{\pluspp Y Z}
}
}
\\
There is only one critical peak. It is of the form \math{(1,1)} 
and can be closed as follows:
\begin{diagram}
  \plusppnoparentheses{\pluspp{\pluspp W X}Y}Z  
&&&&&\rredindex{\omega+1}
&&&&&\plusppnoparentheses{\pluspp W X}{\pluspp Y Z}
\\\dredindex{\omega+1,\,1}
&&&&&&&&&&\dredindex{\omega+1,\,\emptyset}
\\\plusppnoparentheses{\pluspp W{\pluspp X Y}}Z  
&&\rredindex{\omega+1,\,\emptyset}
&&&\plusppnoparentheses W{\pluspp{\pluspp X Y}Z} 
&&\rredindex{\omega+1,\,2}
&&&\plusppnoparentheses W{\pluspp X{\pluspp Y Z}}
\end{diagram}
However, note that the structure of the joinability of the critical
peak is weaker than the first alternative of Example~\ref{ex for asso}. 
Thus, argumentation on the
joinability structure of critical peaks must fail to infer confluence for
this example.
\end{example}

\vfill

\pagebreak

\section{Normality}

When we now start to consider conditional besides unconditional rule systems,
the first to notice is that we have to impose some normality restriction,
as can be seen from Example~\ref{ex bergstra klop} below.

A rule system is called\emph{normal} 
\udiff\ 
for all equations 
``\math{u_0\boldequal u_1}'' in the condition lists of the rules,
at least one of \math{u_0}, \math{u_1} is an irreducible ground term.

Normality is no serious restriction unless left-linearity is required, too.
This is because each non-normal system can be transformed into a normal but
then not left-linear system without changing the reduction relation on the old
sorts: 

\begin{sloppypar}
\label{sect transformation}
One just adds for each old sort \math s a new constructor
function symbol 
\eqindexsymbol s with 
arity
\bigmath{
%  \sigarity(\eqindexsymbol s)\tightequal 
  s\, s\aritysugar s_{\rm new}
}
(where \math{s_{\rm new}} is a new sort)
and 
a new constructor constant symbol \math\bot\ of the sort \math{s_{\rm new}}.
Then in each condition of each rule one transforms each equation of the form
``\math{u\boldequal v}'' with \bigmath{u,v\tightin\tss_s} 
into ``\math{\eqindexpp s u v\boldequal\bot}'' and adds for each
old sort \math s the rule \bigmath{\eqindexpp s{X_s}{X_s}\tightequal\bot}
(where \math{X_s\tightin\Vsigindex s}).
Furthermore one adds the condition 
``\math{\eqindexpp s \appzero\appzero\boldequal\bot}'' to each
unconditional rule for some arbitrary constant \appzero\ of an arbitrary 
old sort \math s.
\end{sloppypar}

The only change this transformation brings for the old sorts is that 
exactly those reductions which were possible with \redindex{\RX,n} 
(for \math{n\prec\omega}) 
become exactly those reductions which are possible
with \redindex{\RX,n+1} after the transformation.
\redindex{\RX,\omega+n}, however, is not changed
by the transformation. \Eg\ for the rule system of
Example~\ref{ex cpw not normal} 
the transformation yields a \math\omega-shallow {[parallel]} joinable, 
terminating system
that is normal now but not left-linear anymore.
 
\yestop
\noindent
Now we return to the question whether joinability implies confluence.
While Lemma~\ref{lemma necessary} states the converse, 
actually little is known about the other direction 
unless the rule system is decreasing.
Theorems 1 
(which is taken from \citebergstraklop)
and 2 
of \citeder\ state that left-linear and normal rule systems are 
confluent if they have no critical pairs or 
are both shallow joinable and terminating.
That normality is essential to imply confluence
of systems with no critical pairs 
can be seen from Example~\ref{ex bergstra klop}.
That normality is also essential to imply confluence
of shallow joinable and terminating systems
can be seen from Example~\ref{ex cpw not normal}.
That left-linearity too is essential in both cases follows from the 
transformation described above.

\vfill

\pagebreak

\yestop
\noindent
In our framework, normality can
be generalized and weakened to quasi-normality, which is a major
result of this paper.

\begin{definition}[Quasi-Normal]
\label{def quasi-normal}

\noindent
Let\/ \math{\alpha\in\{0,\omega\}}.

\noindent
A rule \sugarregel\ is said to be\emph{\math\alpha-quasi-normal \wrt} 
\RX\ \udiff\\
\math{
  \forall\tau\tightin\Xsubst\stopq
}
\\\mbox{}\hfill\inparenthesesoplist{ 
      \inparentheses{
        C\tau\mbox{ fulfilled \wrt\ }\redindex{\RX,\omega+\alpha}
      }
    \oplistimplies
        \forall (u_0\boldequal u_1)\mbox{ in }C\stopq
        \inparenthesesoplist{
            \inparenthesesoplist{
                \alpha\tightequal\omega
              \oplistund 
                \VAR{u_0,u_1}\subseteq\Vcons
            }
          \oplistoder
             \VAR{u_0,u_1}\subseteq\emptyset
          \oplistoder
            \exists i\tight\prec 2\stopq
             \inparenthesesoplist{
                  u_i\tau\tightnotin\DOM{\redindex{\RX,\omega+\alpha}}
                  \oplistoder
                  \inparenthesesoplist{
                      \alpha\tightequal\omega
                    \oplistund
                      (\DEF u_i\tau)\mbox{ occurs in }C\tau
                  }
             }             
        }
}.

\noindent
\RX\ is said to be\emph{\math\omega-quasi-normal}
\udiff\ 
\\\LINEnomath{all rules in \R\ are \math\omega-quasi-normal \wrt\ \RX\@.}
\\
\RX\ is said to be\emph{\math 0-quasi-normal}
\udiff\ 
\\\LINEnomath{
  all constructor rules in \R\ 
  are \math 0-quasi-normal \wrt\ \RX\@.
}

\noindent
Since the case of ``\math{\alpha\tightequal\omega}'' is more important than 
the case of ``\math{\alpha\tightequal 0}'', we use
``{\em quasi-normal}\/'' as an abbreviation for 
``\math\omega-quasi-normal''.
\end{definition}

\yestop
\yestop
\yestop
\noindent
First note that we have added a condition that may reduce the instantiations
of a rule we have to consider. While this may be useless in practice most of
the time, it may allow of further theoretical treatment. 

Also the fact that we have given up the requirement that the irreducible term 
has to be ground may be of minor importance: In practice this usually 
allows only for constructor variables or variables of sorts having only 
irreducible terms. 

Important, however, 
is the fact that equations containing only constructor variables
are not restricted by quasi-normality anymore. \Eg, the rule system of 
Example~\ref{exb} is quasi-normal but not normal. 

Besides this, it is important that quasi-normality
also allows to make any system quasi-normal 
simply by replacing any equation
``\math{u\boldequal v}'' in a condition with 
``\math{u\boldequal v\comma\DEF v}''.

Furthermore, note that no restrictions are 
imposed on \Def- and \boldunequal-literals.

\vfill

\pagebreak

\yestop
\begin{example}[\citebergstraklop]
\label{ex bergstra klop}
\\
\fbox{\bf
\begin{tabular}[t]{@{}l}
No Critical Peaks
\\\hline 
Left- \& Right-Linear
\\\hline 
{\em  Not {[Quasi-]} Normal}
\\\hline 
{\em  Not Terminating}
\end{tabular}
}%
\hrulefill
\fbox{
\arr{[t]{lll}
 \consfunsym
&:=
&\{\dppzero\}
\\\deffunsym
&:=
&\{\bppzero, \gsymbol\}
\\\R_{\,\rm\ref{ex bergstra klop}}
&:
&\arr{[t]{r@{\,}l@{\,}l@{}l}
     \bppzero&=&\gppeins\bppzero
   \\\gppeins X&=&\dppzero&\rs\gppeins X\tightequal X
}
}
}

\noindent
There are no critical peaks.
Nevertheless, \redindex{\R_{\,\rm\ref{ex bergstra klop}},\emptyset} 
is not confluent:
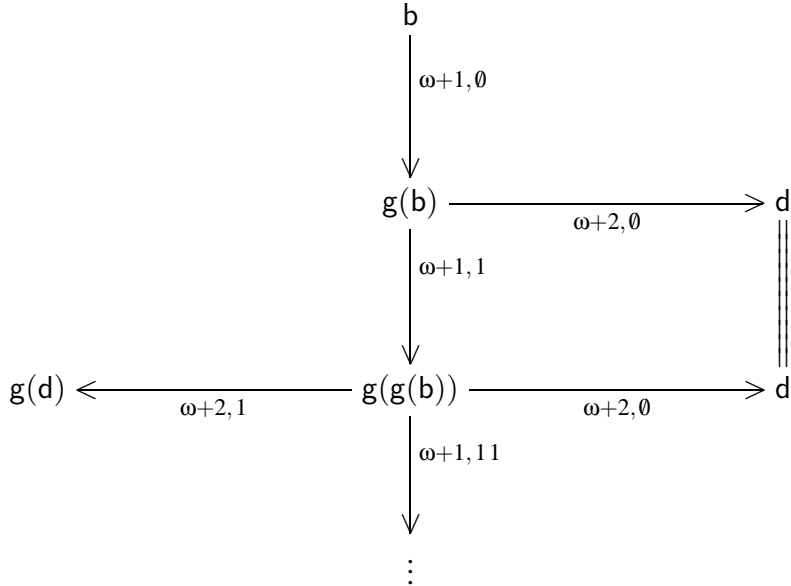
\begin{diagram}
&&&&\bppzero&&&&
\\
&&&&\dredindex{\omega+1,\,\emptyset}&&&&
\\
\\
&&&&\gppeins\bppzero&&\rredindex{\omega+2,\,\emptyset}&&\dppzero
\\
&&&&\dredindex{\omega+1,\,1}&&&&\dequal
\\
\\
\gppeins\dppzero&&\rantiredindex{\omega+2,\,1}&&\gppeins{\gppeins\bppzero}
&&\rredindex{\omega+2,\,\emptyset}&&\dppzero
\\
&&&&\dredindex{\omega+1,\,1\,1}&&&&
\\
\\
&&&&\vdots&&&&
\end{diagram}

\yestop
\yestop
\end{example}

\yestop
\noindent
The following example shows that normality is also required for 
terminating systems. Note that this was already shown by Example~C of
\citeder\ which, however, is more complicated because it has there additional
critical peaks.

\begin{example}
\label{ex cpw not normal}
\\
\fbox{\bf
\begin{tabular}[t]{@{}l}
\begin{diagram}[height=1em,width=1.5em,inline,top]
  \circ&&\rredindex{\omega+1}&&\circ
\\\dredindex{\omega+1,\,1}&&&&\dredindex{\omega+1}^3
\\\\\circ&&\rredindex{\omega+1}&&\circ
\end{diagram}
~~~~~
\begin{diagram}[height=1em,width=1.5em,inline,top]
\circ&&\rredindex{\omega+2}&&\circ
\\\dredindex{\omega+1,\,\emptyset}&&&&\dequal
\\
\\\circ&&\rredindex{\omega+2}&&\circ
\end{diagram}
\\
\\
\math\omega-Shallow {[{[Noisy]} Parallel]} Joinable
\\
\math\omega-Shallow {[Noisy]} Strongly Joinable
\\
{\em Non-Overlay is}
\\
{\em \mbox{~~~} Neither \math\omega-Shallow {[Parallel]} Closed}
\\
{\em \mbox{~~~} Nor \math\omega-Shallow {[Noisy]} Anti-Closed}
\\
{\em Not {[\superterm-Quasi]} Overlay Joinable}
\\\hline 
Left- \& Right-Linear
\\\hline 
{\em  Not {[Quasi-]} Normal}
\\
\hline 
Terminating
\end{tabular}
}%
\hrulefill
\fbox{
\arr{[t]{lll}
 \consfunsym
&:=
&\{\cppzero,\dppzero,\eppzero\}
\\\deffunsym
&:=
&\{\appzero,\bppzero,\fsymbol, \gsymbol, \hsymbol\}
\\\R_{\,\rm\ref{ex cpw not normal}}
&:
&\arr{[t]{r@{\,}l@{\,}l@{}l}
     \appzero&=&\cppzero
   \\\bppzero&=&\dppzero
   \\\fppeins\appzero&=&\gppeins\bppzero
   \\\fppeins\cppzero&=&\hppeins\cppzero
   \\\gppeins\dppzero&=&\hppeins\appzero
   \\\gppeins X      &=&\eppzero 
     &\rs X\boldequal\bppzero
   \\\hppeins X      &=&\eppzero
     &\rs \fppeins X\boldequal\eppzero
}
}
}

\pagebreak

\noindent
There are three critical peaks and they are all of the form \math{(1,1)}.
Since the third is the symmetric overlay of the second, we do not depict
it. The first and the second are joinable as follows:
\begin{diagram}
\fppeins\appzero&&\rredindex{\omega+1}&&\gppeins\bppzero
\\
&&&&\dredindex{\omega+1,\,1}
\\
&&&&\gppeins\dppzero
&&\gppeins\dppzero&&\rredindex{\omega+2}&&\eppzero
\\
\dredindex{\omega+1,\,1}&&&&\dredindex{\omega+1,\,\emptyset}
&&\dredindex{\omega+1,\,\emptyset}&&&&\dequal
\\
&&&&\hppeins\appzero
&&\hppeins\appzero&&\rredindex{\omega+2,\,\emptyset}&&\eppzero
\\
&&&&\dredindex{\omega+1,\,1}
\\
\fppeins\cppzero&&\rredindex{\omega+1,\,\emptyset}&&\hppeins\cppzero
\end{diagram}
Nevertheless, \redindex{\R_{\,\rm\ref{ex cpw not normal}},\emptyset} 
is not confluent:
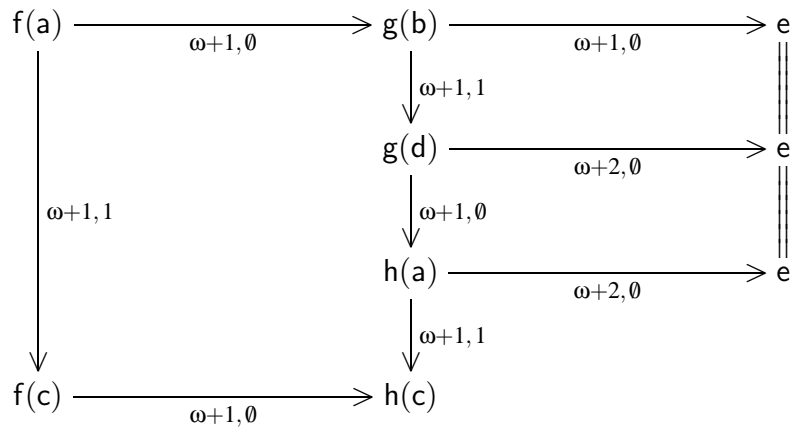
\begin{diagram}
\fppeins\appzero&&\rredindex{\omega+1,\,\emptyset}&&\gppeins\bppzero
&&\rredindex{\omega+1,\,\emptyset}&&\eppzero
\\
&&&&\dredindex{\omega+1,\,1}
&&&&\dequal
\\
&&&&\gppeins\dppzero
&&\rredindex{\omega+2,\,\emptyset}&&\eppzero
\\
\dredindex{\omega+1,\,1}&&&&\dredindex{\omega+1,\,\emptyset}
&&&&\dequal
\\
&&&&\hppeins\appzero
&&\rredindex{\omega+2,\,\emptyset}&&\eppzero
\\
&&&&\dredindex{\omega+1,\,1}
\\
\fppeins\cppzero&&\rredindex{\omega+1,\,\emptyset}&&\hppeins\cppzero
\end{diagram}

\noindent
Note that the overlay would lose its shallow joinability 
if we made
the system normal 
(or else quasi-normal)
by writing the condition of the one but last rule in the
form ``\math{X\boldequal\dppzero}''
(or else in the form 
``\math{X\boldequal\bppzero\comma\DEF\bppzero}'' and declaring 
\bppzero\ to be a constructor), since then we would have 
\bigmath{\gppeins\dppzero\redindex{\omega+1}\eppzero.}
Similarly, the overlay would lose its shallow joinability if we made
the system quasi-normal by writing the condition of the one but last 
rule in the
form ``\math{X\boldequal\bppzero\comma\DEF\bppzero}''
or by substituting \math X with a variable from \Vcons, 
since then we would have 
\bigmath{\hppeins\appzero\redindex{\omega+3}\eppzero}
only 
(since \math{\gppeins\bppzero\notredindex{\omega+1}\eppzero}).
\notop
\end{example}

\pagebreak

\section{Counterexamples for Closed Systems}
\label{sect closed systems}

\yestop
\yestop
\noindent
From the examples of the previous sections 
we can draw the following conclusions:
\begin{enumerate}

\item
For being able to apply syntactic confluence criteria to
non-terminating conditional rule systems,
some kind of [quasi-] normality must be required.

\item
Syntactic
confluence criteria based solely on the joinability structure of the 
critical peaks
must fail 
on some rather simple and common joinability structures.

\end{enumerate}

\yestop
\yestop
\noindent
Therefore, it is now the time 
to have a look at the two most simple non-trivial 
joinability structures under the requirement of normality.

\yestop
\yestop
\yestop
\noindent
These two most simple joinability structures of critical peaks 
are\emph{closedness} and\emph{anti-closedness}, \cf\ below.
Regarding the names of notions below, ``parallel closed'' is taken
from \citehuet, 
``closed'' and ``anti-closed'' have been derived from ``parallel closed''
in an obvious manner, and ``parallel joinable'' was the simplest name%
\footnote{The only obvious wrong intuitions
it could rise are either 
meaningless (since the transitive closures of reduction and
parallel reduction are always identical) or an unnecessary sharpening of 
our notion.}
we found for the last important variant.

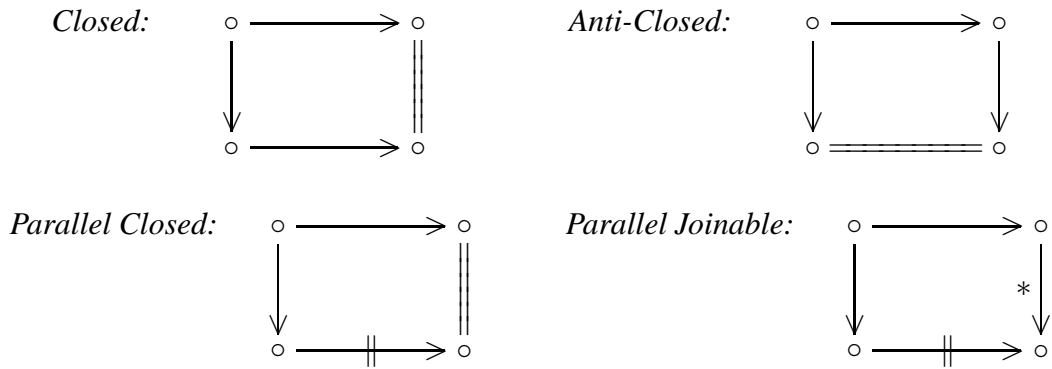
\yestop
\noindent
\LINEnomath{
\hfill
  \emph{Closed:} 
\hfill
\begin{diagram}[%height=1em,width=1.5em,
inline,top]
  \circ&\rred&\circ
\\\dred&&\dequal
\\\circ&\rred&\circ
\end{diagram}
\hfill
\hfill
\emph{Anti-Closed:} 
\hfill
\begin{diagram}[%height=1em,width=1.5em,
inline,top]
  \circ&\rred&\circ
\\\dred&&\dred
\\\circ&\requal&\circ
\end{diagram}%
\hfill
}

\yestop
\yestop
\noindent
\LINEnomath{
\hfill
\emph{Parallel Closed:} 
\hfill
\begin{diagram}[%height=1em,width=1.5em,
inline,top]
  \circ&\rred&\circ
\\\dred&&\dequal
\\\circ&\rredpara&\circ
\end{diagram}%
\hfill
\hfill
\emph{Parallel Joinable:} 
\hfill
\begin{diagram}[%height=1em,width=1.5em,
inline,top]
  \circ&\rred&\circ
\\\dred&&\drefltrans
\\\circ&\rredpara&\circ
\end{diagram}%
\hfill
}

\yestop
\yestop
\yestop
\noindent
It may seem to be surprising that the question whether anti-closedness of  
critical peaks
implies confluence 
for left-linear, non-right-linear, unconditional systems
was listed as Problem~13
in the list of open problems of \citeopenproblems\ and still seems to be open.

For the question whether closedness of critical peaks,
a positive answer follows from the corollary on page
815 in \citehuet\ which says that a left-linear and unconditional system is
confluent if all its critical pairs are parallel closed. The condition of
parallel closedness was weakened in Corollary~3.2 of \citetoyama\ for the
overlays which are required to be 
only parallel joinable instead of parallel closed.

\yestop
\yestop
\yestop
\noindent
For conditional systems, however, neither closedness nor anti-closedness
implies confluence. And this situation does not change when we additionally
require the rule systems to be terminating and normal, as can be seen from the
following examples:

\vfill

\pagebreak

\begin{example}[\middeldorpname, modified by \gramlichname]
\label{ex gramlich}
\\
\fbox{\bf
\begin{tabular}[t]{@{}l}
\begin{diagram}[height=1em,width=1.5em,inline,top]
\circ&&\rredindex{\omega+1}&&\circ
\\\dredindex{1,\,1}&&&&\dredindex{\omega+2}
\\
\\\circ&&\requal&&\circ
\end{diagram}
\\
\\
Anti-Closed
\\
Strongly Joinable 
\\
{\em Not \math\omega-Level Joinable}
\\
{\em Not \math\omega-Shallow Joinable}
\\
{\em Not {[{\rm\superterm}-Quasi]} Overlay Joinable}
\\\hline 
Left-Linear \& Right-Linear
\\\hline 
{[Quasi-]} Normal
\\\hline
Terminating
\\
{\em Not Decreasing}
\end{tabular}
}%
\hrulefill
\fbox{
\arr{[t]{lll}
 \consfunsym
&:=
&\{\appzero, \cppzero \}
\\\deffunsym
&:=
&\{\fsymbol,\gsymbol\}
\\\R_{{\,\rm\ref{ex gramlich}}}
&:
&\arr{[t]{r@{\,}l@{\,}l@{}l}
     \appzero&=&\cppzero
   \\\fppeins\appzero&=&\gppeins\appzero
   \\\gppeins X      &=&\fppeins\cppzero
    &\rs\fppeins X\tightequal\gppeins\cppzero
}
}
}

\noindent
There is only the following critical peak and is of the form
\math{(0,1)}:
\begin{diagram}
\fppeins\appzero&&\rredindex{\omega+1}&&\gppeins\appzero
\\
\dredindex{1,\,1}&&&&\dredindex{\omega+2,\,\emptyset}
\\
\fppeins\cppzero&&\requal&&\fppeins\cppzero
\end{diagram}
Nevertheless, 
\redindex{\R_{{\,\rm\ref{ex gramlich}}},\emptyset} is not confluent:
\begin{diagram}
&&&&\fppeins\appzero&&\rredindex{1,\,1}&&\fppeins\cppzero
\\
&&&&\dredindex{\omega+1,\,\emptyset}&&&&\dequal
\\
\gppeins\cppzero&&\rantiredindex{1,\,1}&&\gppeins\appzero
&&\rredindex{\omega+2,\,\emptyset}&&\fppeins\cppzero
\end{diagram}

\noindent
Since all critical peaks are joinable, \math{\R_{{\,\rm\ref{ex gramlich}}}}
is necessarily non-decreasing
and not compatible with a termination-pair.%
\footnote{\Cf\ Definition~\ref{defcompat} and Theorem~\ref{theoremconfluence}} 
Nevertheless, it is obviously terminating,
since \bigmath{\{X\mapsto\appzero\}} is the only solution for the condition
of the last equation.
Furthermore,
\math{\R_{\,\rm\ref{ex gramlich}}} 
is left-linear, right-linear, and normal\/%
\footnote{even if some authors would not call it ``normal'' since the left-hand
side of the last rule matches the right-hand side of the equation of its
condition}.
Thus (since it is not confluent),
it can be neither overlay joinable nor \math\omega-shallow joinable.%
\footnote{\Cf\ 
theorems \ref{theorem quasi overlay joinable} and \ref{theorem quasi-free}}
It is, however, not \math\omega-level joinable and we did not find a 
\math\omega-level anti-closed but non-confluent system, 
though we spent some time searching for such an example.
\end{example}

\pagebreak

\yestop
\yestop
\begin{example}
\label{ex toll}
\\
\fbox{\bf
\begin{tabular}[t]{@{}l}
\begin{diagram}[height=1em,width=1.5em,inline,top]
\circ&&\rredindex{\omega+1}&&\circ
\\\dredindex{\omega+2,\,1}&&&&\dequal
\\
\\\circ&&\rredindex{\omega+2}&&\circ
\end{diagram}
\mbox{~~~}
\begin{diagram}[height=1em,width=1.5em,inline,top]
\circ&&\rredindex{\omega+1}&&\circ
\\\dredindex{\omega+2,\,2}&&&&\dequal
\\
\\\circ&&\rredindex{\omega+2}&&\circ
\end{diagram}
\\
\\
{[\math\omega-Level {[Parallel]}]} 
Closed
\\
\math\omega-Level Anti-Closed
\\
{[\math\omega-Level]} {[Strongly]}
Joinable 
\\
\math\omega-Level {[Weak]} Parallel Joinable 
\\
{\em Not \math\omega-Shallow Joinable}
\\
{\em Not {[{\rm\superterm}-Quasi]} Overlay Joinable}
\\\hline 
Left-Linear \& Right-Linear
\\
{\em Conditions contain General variables}
\\\hline 
{[Quasi-]} Normal
\\\hline
Terminating
\\
{\em Not Decreasing}
\end{tabular}
}%
\hrulefill
\fbox{
\arr{[t]{lll}
 \consfunsym
&:=
&\{\cppzero, \dppzero \}
\\\deffunsym
&:=
&\{\appzero, \bppzero, \plussymbol\}
\\\R_{{\,\rm\ref{ex toll}}}
&:
&\arr{[t]{r@{\,}l@{\,}l@{}l}
     \appzero&=&\cppzero&\rs\bppzero\tightequal\dppzero
   \\\bppzero&=&\dppzero
   \\\plusppnoparentheses\appzero\appzero&=
    &\dppzero
   \\\plusppnoparentheses\cppzero X&=
    &\dppzero&\rs\plusppnoparentheses X X\tightequal\dppzero
   \\\plusppnoparentheses X\cppzero&=
    &\dppzero&\rs\plusppnoparentheses X X\tightequal\dppzero
}
}
}

\noindent
There are only two critical peaks and they are of the form
\math{(1,1)}.
Using the symmetry of \plussymbol\ in its arguments, 
the other critical peak is symmetric to the 
following one.
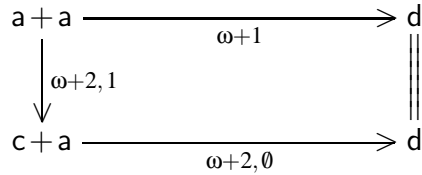
\begin{diagram}
\plusppnoparentheses\appzero\appzero
&&\rredindex{\omega+1}
&&\dppzero
\\
\dredindex{\omega+2,\,1}&&&&\dequal
\\
\plusppnoparentheses\cppzero\appzero
&&\rredindex{\omega+2,\,\emptyset}
&&\dppzero
\end{diagram}
Nevertheless, \redindex{\R_{{\,\rm\ref{ex toll}}},\emptyset} is not confluent:
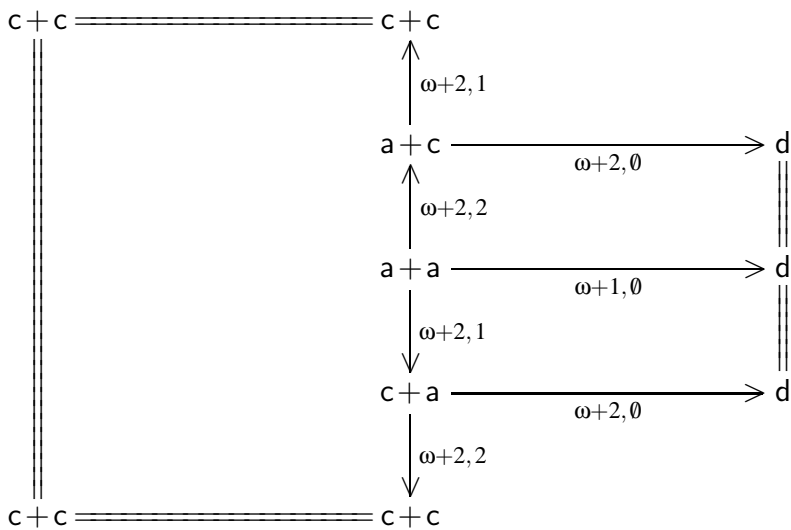
\begin{diagram}
\plusppnoparentheses\cppzero\cppzero&&\requal
&&\plusppnoparentheses\cppzero\cppzero
&&&&
\\
&&&&\dantiredindex{\omega+2,\,1}
&&&&
\\
&&
&&\plusppnoparentheses\appzero\cppzero
&&\rredindex{\omega+2,\,\emptyset}&&\dppzero
\\
&&
&&\dantiredindex{\omega+2,\,2}
&&&&\dequal
\\
\dequal&&
&&\plusppnoparentheses\appzero\appzero
&&\rredindex{\omega+1,\,\emptyset}&&\dppzero
\\
&&
&&\dredindex{\omega+2,\,1}
&&&&\dequal
\\
&&
&&\plusppnoparentheses\cppzero\appzero
&&\rredindex{\omega+2,\,\emptyset}&&\dppzero
\\
&&&&\dredindex{\omega+2,\,2}
&&&&
\\
\plusppnoparentheses\cppzero\cppzero&&\requal
&&\plusppnoparentheses\cppzero\cppzero
&&&&
\end{diagram}

\noindent
Since all critical peaks are joinable, our system is necessarily 
non-decreasing, \cf\ Theorem~\ref{theoremconfluence}.
Nevertheless, it is obviously terminating,
left-linear, right-linear, and normal.
Thus (since it is not confluent),
it can be neither overlay joinable nor \math\omega-shallow joinable, \cf\ 
theorems \ref{theorem quasi overlay joinable} and \ref{theorem quasi-free}.
Due to the given forms of \math\omega-level joinability, the occurrence of 
general variables in the conditions is essential for this example, \cf\ 
theorems \ref{theorem level parallel closed} and \ref{theorem level one}.
\mycomment{
Finally note that we can do the following two variations:
First, when we rearrange our function symbols by 
\bigmath{
  \deffunsym\tightnotni
  \appzero,\bppzero
  \tightin\consfunsym
,}
then all vertical arrows get the subscript \math{2} instead of 
\math{\omega\tight+2}
as well as the critical peaks get the form \math{(0,1)}.
Second, when additionally rearrange our function symbols by
\bigmath{
  \deffunsym\tightnotni
  \plussymbol
  \tightin\consfunsym
,}
\ie\ by declaring all function symbols to be constructor symbols,
then ``\/\math{\omega+}'' is removed from all subscripts
and the critical peaks get the form \math{(0,0)}.
}%comment
\end{example}

\pagebreak

\section{Criteria for Confluence}
\label{sect non-terminating confluence}

\yestop
\yestop
\noindent
Most of the theorems we present in this and the following section
assume the constructor sub-system \redindex{\RX,\omega} to be confluent
and then suggest how to find out that the whole system \redsub\ is 
confluent, too. How to find out that \redindex{\RX,\omega} is 
confluent will be discussed in \sect~\ref{section constructor confluence}.

\yestop
\yestop
\yestop
\noindent
In this section we present confluence criteria that do not rely on
termination. They are, of course, also applicable to terminating systems,
which might be very attractive if one does not know how to show termination
or if the correctness of the 
technique for proving termination requires confluence.

\yestop
\yestop
\yestop
\noindent
Before we state our main theorems 
it is convenient to introduce some further
syntactic restriction. By disallowing non-constructor variables 
in conditions of constructor equations 
we disentangle the 
fulfilledness of conditions of constructor equations from the 
influence of non-constructor rules.

\begin{definition}[Conservative Constructors]
\label{def conservative constructors}
\mbox{}\\
\R\ is said to have\emph{conservative constructors}
\udiff\\
\LINEmath{
  \forall\kurzregel\tightin\R\stopq
  \inparentheses{ 
    l\tightin\tcs
    \ \implies\ 
    \VAR C\subseteq\Vcons
  }   
.}
\end{definition}

\noindent
Let us consider a rule system with conservative constructors.
Together with our global restrictions on constructor rules 
(\cf\ Definition~\ref{defrulescontinu}) this means that 
the condition terms of constructor rules are\emph{pure} constructor terms. 
This has the 
advantage that (contrary to the general case) the 
condition terms of constructor rules still are constructor
terms after they have been instantiated with some substitution.
By \lemmaconskeeping\ this means that the reducibility with 
constructor rules does not depend on the new possibilities 
which could be added by the non-constructor rules later on, \ie\
that the constructor rules are conservative \wrt\ their decision
not to reduce a given term because 
non-constructor rules cannot generate 
additional solutions for their conditions.%
\footnote{Since ``conservative constructors'' is actually a property not
of the constructors (\ie\ constructor function symbols) but of the 
constructor\emph{rules}, the notion should actually be called
``conservative constructor rules''. But 
the commonplace notion of ``free constructors'' is just the same.}

The condition of conservative constructors is very natural and not
very restrictive. (Note that even now constructor rules may have general 
variables in their left- and right-hand sides.) That conservative constructors
make the construction of confluence criteria much easier can be seen from the
following lemma which can treat a special case of 
possible divergence, namely a sub-case of the ``variable overlap case''. 
In this case it is 
important that a reduction with a certain rule can still be done after
the instantiating substitution has been reduced.

\pagebreak

\begin{lemma}
\label{lemma invariance of fulfilledness two}
\mbox{}\\
Let\/ \math{\mu,\nu\in\Xsubst}.
Let\/ \math{\kurzregel\in\R} with \bigmath{l\tightin\tcs.}
\\
Assume that 
\inparenthesesoplist{
    \R\mbox{ has conservative constructors}
  \oplistoder
    \VAR C\tightsubseteq\Vcons
  \oplistoder
    \condterms{C\mu}\tightsubseteq\tcs
}.
\\
Assume \refltransindex{\RX,\omega} to be confluent.
\\
Now, if\/ 
\math{C\mu} is fulfilled \wrt\ \redsub\ 
and\/
\bigmath{
  \forall x\tightin\V\stopq 
    x\mu\refltranssub x\nu
,}
\\
then\/
\math{C\nu} is fulfilled \wrt\ \redindex{\RX,\omega}
and
\bigmath{l\nu\redindex{\RX,\omega}r\nu.}
\notop
\halftop
\end{lemma}

\vfill

\yestop
\yestop
\yestop
\yestop
\yestop
\yestop
\noindent
While the conditions of our main theorems of this section, Theorem~\ref
{theorem parallel closed} and Theorem~\ref{theorem level parallel closed}, 
are rather complicated and difficult to check, 
they are always satisfied for a certain class of rule systems captured
by Theorem~\ref{theorem complementary} 
(being a consequence of Theorem~\ref{theorem parallel closed}) 
and Theorem~\ref{theorem weakly complementary} 
(being a consequence of Theorem~\ref{theorem level parallel closed})
below.

This class consists of left-linear rule systems 
with conservative constructors that achieve 
quasi-normality just by requiring the presence of a \Def-literal
for each equation not containing an irreducible ground term
in a condition of a rule, 
and satisfy the joinability requirements due to the critical peaks being
complementary, \ie\ having complementary literals in their condition lists,
\cf\ \sect~\ref{sect basic joinability}.
Furthermore, rule systems of this class are quite useful in practice.
It generalizes 
the function specification style 
that is usually required in the framework of
classic inductive theorem proving (\cf\ \eg\ \citewaltherhandbook)
by allowing for partial functions resulting from non-complete 
defining case distinctions as well as resulting from non-termination.

\yestop
\yestop
\begin{theorem}[Syntactic Confluence Criterion]
\label{theorem complementary}
\\
Let\/ \R\ be a left-linear CRS over \sig/\cons/\/\V\
with conservative constructors.
\\
Assume 
\bigmath{
  \forall\kurzregel\tightin\R\stopq
     \forall(u_0\boldequal u_1)\mbox{ in }C\stopq
     \exists i\tightprec2\stopq
        \inparenthesesoplist{
             (\DEF u_i)\mbox{ occurs in }C
           \oplistoder
             u_i\in\gt\tightsetminus\DOM\redsub
        }
.}
\\
Assume that \redindex{\RX,\omega} is confluent.
Now:

\noindent
If each critical peak in\/ \math{{\rm CP}(\R)} 
of the form\/ \math{(0,1)}, \math{(1,0)}, or\/ \math{(1,1)} is complementary,
then \redsub\ is confluent.
\end{theorem}

\yestop
\yestop
\begin{theorem}[Syntactic Confluence Criterion]
\label{theorem weakly complementary}
\\
Let\/ \R\ be a left-linear CRS over \sig/\cons/\/\V\ with 
\bigmath{
  \forall\kurzregel\tightin\R\stopq
    \VAR C\tightsubseteq\Vcons
.}
\\
Assume 
\bigmath{
  \forall\kurzregel\tightin\R\stopq
     \forall(u_0\boldequal u_1)\mbox{ in }C\stopq
     \exists i\tightprec2\stopq
        \inparenthesesoplist{
             (\DEF u_i)\mbox{ occurs in }C
           \oplistoder
             u_i\in\gt\tightsetminus\DOM\redsub
        }
.}
\\
Assume that \redindex{\RX,\omega} is confluent.
Now:

\noindent
If each critical peak in\/ \math{{\rm CP}(\R)} 
of the form\/ \math{(0,1)} or\/ \math{(1,0)} is complementary
and each critical peak in\/ \math{{\rm CP}(\R)} of the form\/ \math{(1,1)} 
is weakly complementary, then \redsub\ is confluent.
\end{theorem}

\pagebreak

\yestop
\yestop
\noindent
Note that both theorems are applicable%
\footnote{The careful reader may have noticed that the last two rules
of \math{\R_{{\,\rm\ref{exb}}}} actually are lacking the required
\Def-literals. For practical specification, however, this \Def-literal
can be omitted here because it is tautological for 
\redsub\ if \bigmath{\X\tightsubseteq\Vsig}. 
Note that in practice of specification  
one is only interested
in \redindex{\R,\emptyset} and \redindex{\R,\Vsig}
\cf\ \citewgjsc\ and \citewgcade.
(This, however, does not mean that we do not need
formulas containing \Vcons\ for inductive theorem proving.)}
to the rule system of 
Example~\ref{exb} where the subtraction on natural numbers
is defined via a non-complete syntactic case distinction 
that does not yield critical peaks at all and where the member-predicate
is defined by a syntactic case distinction followed 
(for the case of a nonempty list)
by a semantic case distinction via condition literals
which yields only critical peaks with complementary equations.
To illustrate the possibility of partiality due to non-termination as well as
the possibility of critical peaks with complementary predicate literals,
here is another toy example to which we can apply 
Theorem~\ref{theorem complementary} (but not
Theorem~\ref{theorem weakly complementary}).

\begin{example}
\label{ex while}
(continuing Example~\ref{exb})

\noindent
\arr{[t]{lll}
 \consfunsym
&:=
&\{\zeropp, \ssymbol, \truepp, \falsepp, \nilpp, \cnssymbol\}
\\\deffunsym
&:=
&\{\minussymbol, \mbpsymbol, \whilesymbol\}
\\\sigsorts
&:=
&\{\nat, \bool, \lists\}
\\
}
\\
\arr{[t]{lll}
\R_{\,\rm\ref{ex while}}
&:
&\R_{\,\rm\ref{exb}}
\\&&
\vdots
\\&&
\arr{[t]{@{}r@{\,}l@{\,}ll}
     \whilepp X Y&=&Y
     &\rulesugar\ X\boldequal\falsepp
     \\
     \whilepp X Y&=&\whilepp\ldots\ldots
     &\rulesugar\ X\boldequal\truepp,\ \ldots
}
\\&&\vdots
}

\noindent
We have added two rules to the system from Example~\ref{exb}
for a function `\/\whilesymbol' with arity 
``\/\mbox{\math{\ \bool\;\nat\aritysugar\nat\ }}''
where \math X is meant to be a variable from \Vsigindex{\bool}
and \math Y from \Vsigindex{\nat}.  
The two resulting critical peaks 
are of the form \math{(1,1)} and complementary.
Furthermore, we assume that there are no rules with 
\truepp, \falsepp, or a variable of the sort \bool\  as left-hand sides, 
such that we have
\bigmath{
  \truepp, \falsepp 
  \in 
  \gt
  \tightsetminus
  \DOM{
    \redindex{
      \R_{\,\rm\ref{ex while}},
      \X
    }
  }
.}
\end{example}

\vfill

\yestop
\yestop
\yestop
\yestop
\noindent
The main part of the following theorem is part (I).
Parts (III) and (IV) only weaken the required 
\math\omega-shallow noisy parallel joinability for critical peaks of the form
\math{(1,1)} to \math\omega-shallow noisy\emph{weak} parallel joinability but 
have to pay a considerable price for it. 
It would be of practical importance (\cf\ Example~\ref{ex d})
to achieve this weakening for critical peaks of the form \math{(0,1)}, 
but this is not possible, \cf\ Example~\ref{ex c}.
Furthermore, the difference between (III) and (IV) is marginal since 
non-overlays of the form \math{(1,0)} are pathological%
\footnote{A critical peak of the form \math{(1,0)} requires a
non-constructor rule whose left-hand side has a constructor function symbol
as top symbol, and also requires a constructor rule
with a general variable in its left-hand side.}
anyway.
(II) is rather interesting for the cases where it is possible to restrict
the right-hand sides to be linear \wrt\ general variables; 
this severe restriction is necessary, however; 
\cf\ the second version of Example~\ref{ex levy a}
or \cf\ Example~\ref{ex for asso}.
Besides these examples, also Example~\ref{ex a modified} 
may be able to discourage
the search for a further generalization of the theorem.
Finally note that the `\math i' and `\math j' in the theorem range over
\math{\{0,1\}}.

\pagebreak

\yestop
\yestop
\begin{theorem}[Syntactic Criterion for \math\omega-Shallow Confluence]
\label{theorem parallel closed}
\sloppy\mbox{}

\noindent
Let\/    \R\ be a CRS over \sig/\cons/\/\V\@.
Let\/    \bigmath{\X\tightsubseteq\V.}
\\
Assume\/ \R\ to have conservative constructors,
\RX\ to be quasi-normal,
and
the following weak kind of left-linearity:
\\\math{
  \forall\kurzregel\tightin\R\stopq
  \forall p,q\tightin\TPOS l\stopq
  \forall x  \tightin\V    \stopq
}
\\
\LINEmath{
          \inparentheses{
            \inparenthesesoplist{
                l/p\tightequal x\tightequal l/q
              \oplistund
                p\tightnotequal q
            }
            \implies\
            \inparenthesesoplist{
                    l\tightin\tcs
                  \oplistund
                    x\tightin\Vcons
            }
    }
.}
\\
Furthermore, assume that
\redindex{\RX,\omega} 
is confluent.
\begin{enumerate}

\item[\bf(I)]
Now if 
each critical peak in \math{{\rm CP}(\R)} 
of the form 
\math{(i,1)}
is \math\omega-shallow noisy parallel joinable up to 
\math{
  \omega
  \tight+
  i
  \tight*
  \omega
} 
\wrt\ \RX, 
and
each non-overlay in \math{{\rm CP}(\R)}
of the form \math{(1,j)}
is \math\omega-shallow parallel closed  up to 
\math{
  \omega
  \tight+
  j 
  \tight*
  \omega
}
\wrt\ \RX, 
then
\RX\ is \math\omega-shallow confluent.

\item[\bf(II)]
If we have the following kind of right-linearity \wrt\ general variables
\\\linemath{
  \forall\kurzregel\tightin\R    \stopq
  \forall x        \tightin\Vsig \stopq
  \forall p,q      \tightin\TPOS r\stopq
  \inparentheses{
      r/p\tightequal x\tightequal r/q
    \ \implies\   
      p\tightequal q
  }
,}
and if
each critical peak in \math{{\rm CP}(\R)} 
of the form 
\math{(i,1)}
is \math\omega-shallow noisy strongly joinable up to 
\math{
  \omega
  \tight+
  i
  \tight*
  \omega
} 
\wrt\ \RX, 
and
each non-overlay in \math{{\rm CP}(\R)}
of the form \math{(1,j)}
is \math\omega-shallow noisy anti-closed  up to 
\math{
  \omega
  \tight+
  j 
  \tight*
  \omega
}
\wrt\ \RX, 
then
\RX\ is \math\omega-shallow confluent.

\end{enumerate}
Now additionally assume the following very weak kind of right-linearity
of constructor rules:
\\\math{
  \forall\kurzregel\tightin\R    \stopq
  \forall x        \tightin\Vsig \stopq
  \forall p,q      \tightin\TPOS r\stopq
          \!\!\inparentheses{\!\!
               \inparenthesesoplist{
                  l\tightin\tcs\!\!
                \oplistund
                  r/p\tightequal x\tightequal r/q
              }
              \ \implies\
                  p\tightequal q
          }
.}
\\
Furthermore, additionally assume that 
each critical peak in \math{{\rm CP}(\R)} 
of the form 
\math{(0,1)}
is \math\omega-shallow noisy strongly joinable up to 
\math{\omega},
that each critical peak in \math{{\rm CP}(\R)} 
of the form 
\math{(1,1)}
is \math\omega-shallow noisy weak parallel joinable 
\wrt\ \RX,
and that
each non-overlay in \math{{\rm CP}(\R)}
of the form \math{(1,1)}
is \math\omega-shallow closed
\wrt\ \RX. 
\begin{enumerate}

\item[\bf(III)]
Now if 
each non-overlay in \math{{\rm CP}(\R)}
of the form \math{(1,0)}
is \math\omega-shallow parallel closed  up to 
\math{
  \omega
}
\wrt\ \RX, 
then
\RX\ is \math\omega-shallow confluent.

\end{enumerate}
Now additionally assume that \redindex{\RX,\omega} is strongly confluent.
\begin{enumerate}
\item[\bf(IV)]
Now if 
each non-overlay in \math{{\rm CP}(\R)}
of the form \math{(1,0)}
is \math\omega-shallow closed  
%and \math\omega-shallow noisy anti-closed 
up to \math\omega\ \wrt\ \RX, 
then
\RX\ is \math\omega-shallow confluent.

\end{enumerate}

\yestop
\yestop
\yestop
\yestop
\end{theorem}

\noindent
If we consider all symbols to be non-constructor symbols, 
then each of the parts (I), (III), and (IV) of 
Theorem~\ref{theorem parallel closed} is strong enough to imply
Theorem~1 of \citeder\ (which is taken from \citebergstraklop).
If we, moreover, restrict to unconditional rule systems,
then Theorem~\ref{theorem parallel closed}(I)
specializes to Corollary~3.2 of \citetoyama\ 
(which is stronger than the more restrictive corollary on page
815 in \citehuet\ which says that a left-linear and unconditional system is
confluent if all its critical pairs are parallel closed). 
Moreover, Theorem~\ref{theorem parallel closed}(II) is a generalization
of Theorem~5.2 of \citebeckerstacs\ translated into our framework.

\yestop
\yestop
\yestop
\yestop
\noindent
The proof of Theorem~\ref{theorem parallel closed} is similar to 
that of Corollary~3.2 of \citetoyama\ for unconditional systems, 
but with a global induction
loop on the depth of reduction for using the shallow joinability to get
along with the conditions of the rules, and this whole proof twice
due to our separation into constructors and non-constructors, and this 
again for each part of the theorem.
Since it is very long, tedious,
and uninteresting we have put most its lemmas into \ref{sect furter lemmas} 
and the proofs into \ref{sect proofs}. The only lemmas we consider to be 
interesting are those which make clear why it is possible to generalize
from normal to quasi-normal rule systems.
The problematic case is always the variable-overlap case since it is not
covered by critical peaks. The hard step in this case is to show that
an equation ``\math{u_0\boldequal u_1}'' 
which had been joinable when instantiated
with substitution \math\mu\ is still joinable after the instantiations
for its variables have been reduced, yielding a new substitution \math\nu. 
Thus one has to show that for two natural numbers \math{n_0} and \math{n_1} 
with \bigmath{u_0\mu\downarrowindex{\RX,\omega+n_1}u_1\mu}
and \bigmath{\forall x\tightin\V\stopq x\mu\refltransindex{\RX,\omega+n_0}x\nu}
we always have \bigmath{u_0\nu\downarrowindex{\RX,\omega+n_1}u_1\nu}.
This means that the fulfilledness of the instantiated equation 
``\math{u_0\boldequal u_1}'' is not changed by the reduction of its
instantiating substitution.
For showing this we may use the global induction hypothesis implying that
\RX\ is \math\omega-shallow confluent up to \math{n_0\plusomega n_1}.
The reader may verify that we do not seem to have a chance 
for being successful here
unless we require some kind of normality. 
Lemma~\ref{lemma simpler cases}(\ref{lemma simpler cases two item two})
depicts the situation we are in 
(matching its \math{s_i} to \math{u_i\mu} and its \math{s_i'} to \math{u_i\nu})
and shows that irreducibility of \math{u_1\nu} 
(roughly speaking \ie\ normality)
is just as helpful as
some literal ``\math{\DEF u_1\mu}'' in the condition list
(\ie\ an alternative allowed by quasi-normality)
(because the latter implies the existence of some \math{t_1\in\tgcons}
with \bigmath{u_1\mu\refltransindex{\RX,\omega+n_1}t_1}).
Finally, Lemma~\ref{lemma invariance of fulfilledness} states that 
the other alternative given by quasi-normality 
(\ie\ that the equation contains no non-constructor variables)
is no problem either, and that \Def- and \boldunequal-literals 
do not make any problems and therefore need not at all be restricted 
by normality requirements.

Since we consider the proofs of the following two lemmas to be interesting, 
we did not put them into the appendix but included them here.
The form of presentation is very general.
This enables the proof to present the idea of quasi-normality in its 
essential form and also enables more than a dozen of applications of 
Lemma~\ref{lemma invariance of fulfilledness}  in the proofs of the theorems
in this and the following sections.
When reading the lemmas please note that the optional parts are 
only necessary for reusing
the lemmas in the proofs of the theorems of the following sections where
termination arguments will be included into the confluence criteria.
Moreover for a first reading only the second cases of their initial
disjunctive assumptions should be considered. The others are uninteresting
special cases.

\vfill

\pagebreak

\yestop
\yestop
\begin{lemma}
\label{lemma simpler cases}

\noindent
{{[}Let \math\rhd\ be a wellfounded ordering.]}
Let \math{n_0,n_1\prec\omega}.
Let \math{\alpha\in\{0,\omega\}}.
Assume that
\\ 
\bigmath{
  \forall i\tightprec 2\stopq
    \inparenthesesoplist{
        s_i\tightequal s_i'
%        s_i\tightnotin\DOM{\redindex{\RX,\omega+\alpha}}
      \oplistoder  
        \mbox{\RX\ is \math\alpha-shallow confluent up to }
        n_0\plusalpha n_1 
        \mbox{ [and }s_i\mbox{ in }\lhd\mbox{]}
    }
.}
\noindent
Now:
\begin{enumerate}

\noitem
\item\label{lemma simpler cases item one and a half}
\bigmath{n_0\tightpreceq n_1}
and\/
\bigmath{
  s_0'
  \antirefltransindex{\RX,\alpha+n_0}
  s_0
  \refltransindex    {\RX,\alpha+n_1}
  t_0
}
implies\/
\bigmath{
  s_0'
  \downarrowindex{\RX,\alpha+n_1}
  t_0
.}

\item\label{lemma simpler cases item two}
\bigmath{n_0\tightpreceq n_1}
and\/
\bigmath{
  s_0'
  \antirefltransindex{\RX,\alpha+n_0}
  s_0
  \refltransindex{\RX,\alpha+n_1}
  t_0
  \antirefltransindex{\RX,\alpha+n_1}
  s_1
  \refltransindex{\RX,\alpha+n_0}
  s_1'
}
implies\/
\\\mbox{}
\bigmath{
  s_0'
  \downarrowindex{\RX,\alpha+n_1}
  s_1'.}

\item\label{lemma simpler cases two item one}
\bigmath{
  s_0'
  \antirefltransindex{\RX,\alpha+n_0}
  s_0
  \refltransindex{\RX,\alpha+n_1}
  t_2
  \tightin
  \tgcons
}
implies\/
\\\mbox{}
\bigmath{
  \exists t_3\tightin\tgcons\stopq
  s_0'
  \refltransindex{\RX,\alpha+n_1}
  t_3
  \antirefltransindex{\RX,\omega+n_1}
  t_2
.}

\item\label{lemma simpler cases two item two}
\bigmath{
  s_0'
  \antirefltransindex{\RX,\alpha+n_0}
  s_0
  \refltransindex{\RX,\alpha+n_1}
  t_0
  \antirefltransindex{\RX,\alpha+n_1}
  s_1
  \refltransindex{\RX,\alpha+n_0}
  s_1'
}
together with
either
\\\mbox{}
\bigmath{
  s_1\tightnotin\DOM{\redindex{\RX,\omega+\alpha}}
}
or
\\\mbox{}
\bigmath{
\inparenthesesoplist{
    \alpha\tightequal\omega
  \oplistund
    s_1
    \refltransindex{\RX,\omega+n_1}
    t_1
    \tightin
    \tgcons
  \oplistund
    \forall\delta\tightprec n_0\plusomega n_1\stopq
      \RX\mbox{ is \math\omega-shallow confluent up to }
      \delta\mbox{ [and }
      s_1\mbox{ in }
      \tight\lhd\mbox]
}
}
implies\/
\\\mbox{}
\bigmath{s_0'\downarrowindex{\RX,\alpha+n_1}s_1'.}

\end{enumerate}
\end{lemma}

\yestop
\yestop
\yestop
\begin{proofqed}{Lemma~\ref{lemma simpler cases}}
\underline{\underline{\ref{lemma simpler cases item one and a half}:}}
Consider the peak
\bigmath{
  s_0'
  \antirefltransindex{\RX,\alpha+n_0}
  s_0
  \refltransindex{\RX,\alpha+n_1}
  t_0
.}
{If 
\bigmath{
  s_0\tightequal s_0'
%  s_0\tightnotin\DOM{\redindex{\RX,\omega+\alpha}}
,} 
then we are finished due to
\bigmath{s_0'\tightequal s_0\refltransindex{\RX,\alpha+n_1}t_0.} Otherwise:}
We have assumed
that \RX\ is \math\alpha-shallow confluent up to \math{n_0\plusalpha n_1}
{[and \math{s_0} in \math\lhd]}.
Thus we get 
\bigmath{
  s_0'
  \refltransindex    {\RX,\alpha+n_1}
  \circ
  \antirefltransindex{\RX,\alpha+n_0}
  t_0
}
and then due to \bigmath{n_0\tightpreceq n_1} and 
\lemmamonotonicinbeta\ we get \bigmath{s_0'\downarrowindex{\RX,\alpha+n_1}t_0}.
\\
\underline{\underline{\ref{lemma simpler cases item two}:}}
By (\ref{lemma simpler cases item one and a half}) we get
\bigmath{
  s_0'
  \refltransindex{\RX,\alpha+n_1}
  t_1
  \antirefltransindex{\RX,\alpha+n_1}
  t_0}
for some \math{t_1}.
Finally, consider the peak
\bigmath{
  t_1
  \antirefltransindex{\RX,\alpha+n_1}
  s_1
  \refltransindex{\RX,\alpha+n_0}
  s_1'
.}
By (\ref{lemma simpler cases item one and a half}) again we get
\bigmath{
  s_0'
  \refltransindex{\RX,\alpha+n_1}
  t_1
  \downarrowindex{\RX,\alpha+n_1}
  s_1'
}
as desired.
\\
\underline{\underline{\ref{lemma simpler cases two item one}:}}
Consider the peak
\bigmath{
  s_0'
  \antirefltransindex{\RX,\alpha+n_0}
  s_0
  \refltransindex{\RX,\alpha+n_1}
  t_2
.}
{If 
\bigmath{
  s_0\tightequal s_0'
%  s_0\tightnotin\DOM{\redindex{\RX,\omega+\alpha}}
,} 
then we are finished due to
\bigmath{s_0'\tightequal s_0\refltransindex{\RX,\alpha+n_1}t_2.} Otherwise:}
By \math\alpha-shallow confluence up to \math{n_0\plusalpha n_1} 
{[and \math{s_0} in \math\lhd]}
we get
\bigmath{
  s_0'
  \refltransindex{\RX,\alpha+n_1}
  t_3
  \antirefltransindex{\RX,\alpha+n_0}
  t_2
}
for some \math{t_3}.
By 
\bigmath{t_2\tightin\tgcons}
and \lemmaconskeeping\ 
we get
\bigmath{
  \tgcons
  \tightni
  t_3
  \antirefltransindex{\RX,\omega}
  t_2
.}
Thus
we have
\bigmath{
  s_0'
  \refltransindex{\RX,\alpha+n_1}
  t_3
  \antirefltransindex{\RX,\omega+n_1}
  t_2
}
as desired.
\\
\underline{\underline{\ref{lemma simpler cases two item two}:}}
\underline{\math{
  s_1\tightnotin\DOM{\redindex{\RX,\omega+\alpha}}
}:}
{If 
\bigmath{
  s_0\tightequal s_0'
%  s_0\tightnotin\DOM{\redindex{\RX,\omega+\alpha}}
,} 
then we are finished due to
\bigmath{
  s_0'
  \tightequal
  s_0
  \refltransindex{\RX,\alpha+n_1}
  t_0
  \tightequal
  s_1
  \tightequal 
  s_1'
.} 
Otherwise:}
Consider the peak
\bigmath{
  s_0'
  \antirefltransindex{\RX,\alpha+n_0}
  s_0
  \refltransindex{\RX,\alpha+n_1}
  t_0
.}
By \math\alpha-shallow confluence 
up to \math{n_0\plusalpha n_1}
{[and \math{s_0} in \math\lhd]}
we get
\bigmath{
  s_0'
  \refltransindex{\RX,\alpha+n_1}
  t_2
  \antirefltransindex{\RX,\alpha+n_0}
  t_0
}
for some \math{t_2}.
Since \bigmath{s_1\tightnotin\DOM{\redindex{\RX,\omega+\alpha}}} 
this finishes the proof in this case
due to 
\bigmath{t_2\tightequal t_0\tightequal s_1\tightequal s_1'.}
\pagebreak
\\
\underline{\math{
  s_1\tightin\DOM{\redindex{\RX,\omega+\alpha}}
}:}
Then we have
\bigmath{
  \alpha
  \tightequal
  \omega
,}
\bigmath{
  s_1
  \refltransindex{\RX,\omega+n_1}
  t_1
  \tightin
  \tgcons
,}
and\\
\bigmath{
    \forall\delta\tightprec n_0\plusomega n_1\stopq
      \RX\mbox{ is \math\omega-shallow confluent up to }
      \delta\mbox{ [and }
      s_1\mbox{ in }
      \tight\lhd\mbox]
,}
\cf\ the diagram below.
Consider the peak
\bigmath{
  t_0
  \antirefltransindex{\RX,\omega+n_1}
  s_1
  \refltransindex    {\RX,\omega+n_1}
  t_1
.}
We may assume \bigmath{n_1\tightprec n_0} because in
case of \bigmath{n_0\tightpreceq n_1} the proof is finished due to
(\ref{lemma simpler cases item two}).
Then we have
\bigmath{
  n_1\plusomega n_1
  \prec
  n_0\plusomega n_1
.}
Thus by
\math\omega-shallow confluence up to 
\math{n_1\plusomega n_1}
{[and \math{s_1} in \math\lhd]}
we get
\bigmath{
  t_0
  \refltransindex    {\RX,\omega+n_1}
  t_2
  \antirefltransindex{\RX,\omega+n_1}
  t_1
}
for some \math{t_2}.
By
\bigmath{t_1\tightin\tgcons}
and \lemmaconskeeping\ 
we get
\bigmath{
  \tgcons
  \tightni
  t_2
.}
Consider the peak
\bigmath{
  s_0'
  \antirefltransindex{\RX,\omega+n_0}
  s_0
  \refltransindex{\RX,\omega+n_1}
  t_2
.}
Due to 
\bigmath{
  t_2\tightin\tgcons
} 
and (\ref{lemma simpler cases two item one}) 
there is some
\math{t_3\in\tgcons}
with
\bigmath{
  s_0'
  \refltransindex{\RX,\omega+n_1}
  t_3
  \antirefltransindex{\RX,\omega+n_1}
  t_2
.}
By (\ref{lemma simpler cases two item one}) again,
the peak
\bigmath{
  t_3
  \antirefltransindex{\RX,\omega+n_1}
  s_1
  \refltransindex{\RX,\omega+n_0}
  s_1'
}
implies
\bigmath{
%  s_0'
%  \refltransindex{\RX,\omega+n_1}
  t_3
  \downarrowindex{\RX,\omega+n_1}
  s_1'
}
as desired.
\begin{diagram}
 s_0'
&\rantirefltransindex{\omega+n_0}
&s_0
&\rrefltransindex{\omega+n_1}
&t_0
&\rantirefltransindex{\omega+n_1}
&s_1
&\rrefltransindex{\omega+n_0}
&s_1'
\\
 \drefltransindex{\omega+n_1}
&&&&\drefltransindex{\omega+n_1}
&&\drefltransindex{\omega+n_1}
&&
\\
 t_3
&
&\rantirefltransindex{\omega+n_1}
&
&t_2
&\rantirefltransindex{\omega+n_1}
&t_1
&
&\drefltransindex{\omega+n_1}
\\
 \drefltransindex{\omega+n_1}
&&&&&&&&
\\
 \circ
&&&&\requal
&&&&\circ
\\
\end{diagram}
\end{proofqed}\notop

\yestop
\yestop
\yestop
\yestop
\yestop
\yestop
\begin{lemma}
\label{lemma invariance of fulfilledness}
\\
{{[}Let \math\rhd\ be a wellfounded ordering.]}
\\
Let\/ \math{\alpha\in\{0,\omega\}}.
Let\/ \math{n_0,n_1\prec\omega}.
Let\/ \math{\mu,\nu\in\Xsubst}.
\\
Let\/ \math{\kurzregel\in\R} 
with\/ \bigmath{\alpha\tightequal 0\implies l\tightin\tcs.}
\\
Assume that\/ \bigmath{n_0\tightpreceq n_1}
or that\/ \kurzregel\ is \math\alpha-quasi-normal \wrt\ \RX\@.
Assume that
\\\math{
  \forall L\mbox{ in }C\stopq
  \forall u\tightin\condterms L\stopq  
}\\\mbox{}\hfill\inparenthesesoplist{
    u\mu\tightnotin\DOM{\redindex{\RX,\omega+\alpha}}
  \oplistoder
    \RX\mbox{ is \math\alpha-shallow confluent up to\/ }
    n_0\plusalpha n_1
    \mbox{ [and }
    u\mu
    \mbox{ in }
    \lhd\mbox{]}
  \oplistoder
    \inparenthesesoplist{
        \forall x\tightin\VAR u\stopq x\mu\tightequal x\nu
      \oplistund
        \inparenthesesoplist{
            \alpha\tightequal0
          \oplistoder
            \forall v\stopq L\tightnotin\{(u\boldequal v),(v\boldequal u)\}
          \oplistoder
            \forall x\tightin\VAR L\stopq x\mu\tightequal x\nu
          \oplistoder
            \forall\delta\tightprec n_0\plusalpha n_1\stopq 
          \oplistnl
            \RX\mbox{ is \math\alpha-shallow confluent up to\/ }
            \delta\mbox{ [and }
            u\mu\mbox{ in }\lhd\mbox{]}
        }
    }
}.

\noindent
Now, if\/ 
\math{C\mu} is fulfilled \wrt\ \redindex{\RX,\alpha+n_1}
and\/
\bigmath{
  \forall x\tightin\V\stopq 
    x\mu\refltransindex{\RX,\alpha+n_0}x\nu
,}
\\
then\/
\math{C\nu} is fulfilled \wrt\ \redindex{\RX,\alpha+n_1}
and
\bigmath{l\nu\redindex{\RX,\alpha+n_1+1}r\nu.}
\end{lemma}

\pagebreak

\begin{proofqed}{Lemma~\ref{lemma invariance of fulfilledness}}
Since \bigmath{\alpha\tightequal 0\implies l\tightin\tcs,}
it suffices to show that for each literal \math L in \math C: 
\math{L\nu} is fulfilled \wrt\ \redindex{\RX,\alpha+n_1}.
Note that we already know that 
\math{L\mu} is fulfilled \wrt\ \redindex{\RX,\alpha+n_1}.
In case of \bigmath{u\mu\tightnotin\DOM{\redindex{\RX,\omega+\alpha}}}
we get \bigmath{u\mu\tightequal u\nu} due to 
\bigmath{u\mu\refltransindex{\RX,\alpha+n_0} u\nu.}
In case of \bigmath{\forall x\tightin\VAR u\stopq x\mu\tightequal x\nu}
we get \bigmath{u\mu\tightequal u\nu} again.
Thus we may assume
\bigmath{
              \forall u\tightin\condterms L\stopq
                \inparenthesesinline{
                     u\mu\tightequal u\nu
                   \ \oder\ 
                     \RX
                     \mbox{ is \math\alpha-shallow confluent up to }
                     n_0\plusalpha n_1
                     \mbox{ [and }
                     u\mu
                     \mbox{ in }
                     \lhd\mbox{]}
                     }
.}
\\
\underline{\math{L=(s_0\boldequal s_1)}:}
We have
\bigmath{
  s_0\nu
  \antirefltransindex{\RX,\alpha+n_0}
  s_0\mu
  \refltransindex    {\RX,\alpha+n_1}\penalty-1
  t_0
  \antirefltransindex{\RX,\alpha+n_1}\penalty-1
  s_1\mu
  \refltransindex    {\RX,\alpha+n_0}
  s_1\nu
}
for some \math{t_0.}
In case of \bigmath{n_0\tightpreceq n_1} we get the desired
\bigmath{s_0\nu\downarrowindex{\RX,\alpha+n_1}s_1\nu}
by Lemma~\ref{lemma simpler cases}(\ref{lemma simpler cases item two}).
Otherwise, by assumption of the lemma, 
\kurzregel\ must be \math\alpha-quasi-normal.
Since \math{C\mu} is fulfilled \wrt\ \redindex{\RX,\omega+\alpha},
according to the definition of \math\alpha-quasi-normality 
and the disjunctive assumption of the lemma 
we have two distinguish several cases here.
First we treat the case in which 
\bigmath{
  \exists i\tightprec 2\stopq
  s_i\mu\tightnotin\DOM{\redindex{\RX,\omega+\alpha}}
.} 
\Wrog\ say
\bigmath{s_1\mu\tightnotin\DOM{\redindex{\RX,\omega+\alpha}}.} 
By Lemma~\ref{lemma simpler cases}(\ref{lemma simpler cases two item two})
we get the desired \bigmath{s_0\nu\downarrowindex{\RX,\alpha+n_1}s_1\nu.}
Second, in case of
\bigmath{
  \forall x\tightin\VAR L\stopq
    x\mu\tightequal x\nu
}
we know that \bigmath{L\nu\tightequal L\mu} which is fulfilled \wrt\ 
\redindex{\RX,\alpha+n_1}.
Note that now we may assume that \bigmath{\alpha\tightequal\omega}
because the second case includes the only case left for
\math0-quasi-normality, namely
\bigmath{\VAR{s_0,s_1}\tightsubseteq\emptyset.}
Third, in case of \bigmath{\VAR{s_0,s_1}\tightsubseteq\Vcons}
we have for all \math{x\in\VAR{s_0,s_1}}:
\bigmath{x\mu\tightin\tcc;}
and then
\bigmath{x\mu\refltransindex{\RX,\omega}x\nu}
by \lemmaconskeeping.
This means \bigmath{s_i\mu\refltransindex{\RX,\omega}s_i\nu.}
By Lemma~\ref{lemma simpler cases}(\ref{lemma simpler cases item two})
(matching its \math{n_0} to \math{0})
due to \bigmath{0\plusomega n_1\preceq n_0\plusomega n_1}
we get the desired \bigmath{s_0\nu\downarrowindex{\RX,\omega+n_1}s_1\nu.}
Finally we come to the fourth case where \wrog\ 
\math{(\DEF s_1\mu)} occurs in \math{C\mu}.
Then there is some \math{t_1\in\tgcons} with
\bigmath{
  s_1\mu
  \refltransindex{\RX,\omega+n_1}
  t_1
.}
Since we may assume that we are not in any of the previous cases,
the disjunctive assumption of the lemma now states that
\bigmath{
  \forall\delta\tightprec n_0\plusomega n_1\stopq
  \RX\mbox{ is \math\omega-shallow confluent up to }
  \delta\mbox{ [and \math{u\mu} in \tight\lhd]}
.}
By Lemma~\ref{lemma simpler cases}(\ref{lemma simpler cases two item two})
we get the desired \bigmath{s_0\nu\downarrowindex{\RX,\omega+n_1}s_1\nu.}
\\\underline{$L=(\DEF s)$:}
We know the existence of
\math{t\in\tgcons}
with
\bigmath{
  s\nu
  \antirefltransindex{\RX,\alpha+n_0}
  s\mu
  \refltransindex{\RX,\alpha+n_1}
  t
.}
By Lemma~\ref{lemma simpler cases}(\ref{lemma simpler cases two item one})
there is some
\math{t'\in\tgcons} with
\bigmath{
  s\nu
  \refltransindex{\RX,\alpha+n_1}
  t'
  \antirefltransindex{\RX,\omega+n_1}
  t
.}
\\
\underline{$L=(s_0\boldunequal s_1)$:}
There exist some \math{t_0,t_1\in\tgcons}
with
\bigmath{
  \forall i\tightprec2\stopq
    s_i\nu
    \antirefltransindex{\RX,\alpha+n_0}
    s_i\mu
    \refltransindex{\RX,\alpha+n_1}
    t_i
}
and 
\bigmath{
  t_0
  \notconfluindex{\RX,\alpha+n_1}  
  t_1
.}
Just like above
we get \math{t_0',\ t_1'\in\tgcons} with
\bigmath{
  \forall i\tightprec2\stopq
    s_i\nu
    \refltransindex{\RX,\alpha+n_1}
    \penalty-1
    t_i'
    \penalty-1
    \antirefltransindex{\RX,\omega+n_1}
    t_i
.}
Finally 
\bigmath{
  t_0'
  \antirefltransindex{\RX,\omega+n_1}
  t_0
  \notconfluindex{\RX,\alpha+n_1}  
  t_1
  \refltransindex{\RX,\omega+n_1}
  t_1'
}
implies
\bigmath{
  t_0'
  \notconfluindex{\RX,\omega+n_1}  
  t_1'
}
since we have \bigmath{\alpha\tightequal\omega} due to 
\bigmath{l\tightnotin\tcs}
in this case of a negative literal.
\end{proofqed}

\pagebreak

\yestop
\yestop
\yestop
\yestop
\noindent
We do not have to discuss the following theorem in detail here,
because it is very similar to Theorem~\ref{theorem parallel closed},
but weakens the required \math\omega-shallow joinabilities
to \math\omega-level joinabilities wherever possible.
Note that from Example~\ref{ex a} we can conclude that
the \math\omega-shallow joinabilities required for critical
peaks of the form \math{(0,1)} cannot be weakened to \math\omega-level
joinabilities in any of the four parts of the theorem.%
\footnote{Note that with the exception of part (II) of the theorem
we could also use the first version of Example~\ref{ex levy a} for this
conclusion.}
However, the price we have to pay for
weakening shallow to level joinability is to extend our requirement
that the conditions contain constructor variables only, from constructor
rules (``conservative constructors'') to all rules! 
That this restriction is necessary indeed can be seen from
Example~\ref{ex toll}.
On the other hand, this restriction gives\emph{quasi}-normality for free.

We prefer to discuss and 
apply Theorem~\ref{theorem parallel closed} wherever possible
because 
contrary to Theorem~\ref{theorem level parallel closed}
it has interesting implications for the standard framework without
the separation into constructor and non-constructor symbols where 
``only constructor variables in conditions'' means 
``no variables in conditions'' which again can (in general not effectively) 
be reduced to 
``no conditions'' by removing the fulfilled conditions and the rules with
non-fulfilled conditions.

The main part of the following theorem is part (I).
Parts (III) and (IV) only weaken the required 
\math\omega-level parallel joinability for critical peaks of the form
\math{(1,1)} to \math\omega-level weak parallel joinability but have to 
pay a considerable price for it. 
Furthermore, the difference between (III) and (IV) is marginal since 
non-overlays of the form \math{(1,0)} are pathological anyway.
(II) is rather interesting for the cases where it is possible to restrict
the right-hand sides to be linear \wrt\ general variables; 
this severe restriction is necessary, however; 
\cf\ the second version of Example~\ref{ex levy a}
or \cf\ Example~\ref{ex for asso}.

\pagebreak

\yestop
\begin{theorem}[Syntactic Criterion for \math\omega-Level Confluence]
\label{theorem level parallel closed}
\sloppy\mbox{}

\noindent
Let\/    \R\ be a CRS over \sig/\cons/\/\V\@.
Let\/    \bigmath{\X\tightsubseteq\V.}
\\
Assume the following important restriction on variables in conditions to hold:
\\\linemath{\forall\kurzregel\tightin\R\stopq\VAR C\tightsubseteq\Vcons.}
Moreover, assume the following weak kind of left-linearity:
\\\math{
  \forall\kurzregel\tightin\R\stopq
  \forall p,q\tightin\TPOS l\stopq
  \forall x  \tightin\V    \stopq
}
\\
\LINEmath{
          \inparentheses{
            \inparenthesesoplist{
                l/p\tightequal x\tightequal l/q
              \oplistund
                p\tightnotequal q
            }
            \implies\
            \inparenthesesoplist{
                    l\tightin\tcs
                  \oplistund
                    x\tightin\Vcons
            }
    }
.}
\\
Furthermore, assume that
\redindex{\RX,\omega} 
is confluent.
\begin{enumerate}
\item[\bf(I)]
Now if 
each critical peak in \math{{\rm CP}(\R)} 
of the form 
\math{(0,1)}
is \math\omega-shallow parallel joinable up to 
\math{
  \omega
} 
\wrt\ \RX, 
each non-overlay in \math{{\rm CP}(\R)}
of the form \math{(1,0)}
is \math\omega-shallow parallel closed  up to 
\math{
  \omega
}
\wrt\ \RX, 
each critical peak in \math{{\rm CP}(\R)} 
of the form 
\math{(1,1)}
is \math\omega-level parallel joinable \wrt\ \RX, 
and
each non-overlay in \math{{\rm CP}(\R)}
of the form \math{(1,1)}
is \math\omega-level parallel closed \wrt\ \RX, 
then
\RX\ is \math\omega-level confluent.

\item[\bf(II)]
If we have the following kind of right-linearity \wrt\ general variables
\\\linemath{
  \forall\kurzregel\tightin\R    \stopq
  \forall x        \tightin\Vsig \stopq
  \forall p,q      \tightin\TPOS r\stopq
  \inparentheses{
      r/p\tightequal x\tightequal r/q
    \ \implies\   
      p\tightequal q
  }
,}
and if
each critical peak in \math{{\rm CP}(\R)} 
of the form 
\math{(0,1)}
is \math\omega-shallow strongly joinable up to 
\math\omega\ \wrt\ \RX, 
each non-overlay in \math{{\rm CP}(\R)} of the form \math{(1,0)}
is \math\omega-shallow anti-closed  up to 
\math\omega\ \wrt\ \RX, 
each critical peak in \math{{\rm CP}(\R)} 
of the form 
\math{(1,1)}
is \math\omega-level strongly joinable \wrt\ \RX,
and 
each non-overlay in \math{{\rm CP}(\R)} of the form \math{(1,1)}
is \math\omega-level anti-closed \wrt\ \RX, 
then \RX\ is \math\omega-level confluent.

\end{enumerate}
Now additionally assume the following very weak kind of right-linearity
of constructor rules:
\\\math{
  \forall\kurzregel\tightin\R    \stopq
  \forall x        \tightin\Vsig \stopq
  \forall p,q      \tightin\TPOS r\stopq\!\!
          \inparentheses{\!\!
            \inparenthesesoplist{
                l\tightin\tcs\!\!
                \oplistund
                r/p\tightequal x\tightequal r/q
            }
            \ \implies\
                p\tightequal q
           }
.}
\\
Furthermore, additionally assume that 
each critical peak in \math{{\rm CP}(\R)} 
of the form 
\math{(0,1)}
is \math\omega-shallow strongly joinable up to 
\math{\omega},
that each critical peak in \math{{\rm CP}(\R)} 
of the form 
\math{(1,1)}
is \math\omega-level weak parallel joinable 
\wrt\ \RX,
and that
each non-overlay in \math{{\rm CP}(\R)}
of the form \math{(1,1)}
is \math\omega-level closed
\wrt\ \RX. 
\begin{enumerate}

\item[\bf(III)]
Now if 
each non-overlay in \math{{\rm CP}(\R)}
of the form \math{(1,0)}
is \math\omega-shallow parallel closed  up to 
\math{
  \omega
}
\wrt\ \RX, 
then
\RX\ is \math\omega-level confluent.

\end{enumerate}
Now additionally assume that \redindex{\RX,\omega} is strongly confluent.
\begin{enumerate}
\item[\bf(IV)]
Now if 
each non-overlay in \math{{\rm CP}(\R)}
of the form \math{(1,0)}
is \math\omega-shallow closed  
%and \math\omega-shallow anti-closed 
up to \math\omega\ \wrt\ \RX, 
then
\RX\ is \math\omega-level confluent.

\pagebreak

\end{enumerate}
\end{theorem}

\section{Criteria for Confluence of Terminating Systems}
\label{section now termination}

\yestop
\yestop
In this section we examine how we can relax our joinability requirements
when we additionally require termination for our reduction relation.
Note that in confluence criteria whose proof is by induction on an
extension of the reduction relation the joinability requirement can be
weakened to a\emph{sub-connectedness} requirement, \cf\ \citesubconnected.
We here, however, present the simpler versions only, where
the connectedness is required to have the form of a single ``valley''.

\yestop
\yestop
\yestop
\noindent
Due to its fundamental importance, 
we first repeat Theorem~7.17 of \citewgjsc\ here, 
which generalizes Theorem~3 of \citeder\ by weakening 
decreasingness to compatibility with a termination-pair
(defined in \sect~\ref{sect relations})
as well as joinability to \math\rhd-weak joinability
(defined in \sect~\ref{sect basic joinability})
which provides us with some confluence assumption 
when checking the fulfilledness of the condition of a critical peak.

\yestop
\yestop
\begin{definition}[Compatibility with a Termination-Pair]
\label{defcompat}
\\
A rule\/ \math{\kurzregel} is is\emph{\RXcompatible} 
with a termination-pair\/ \math{(>,\rhd)}
over\/ \sig/\/\V\ 
\udiff\\  
\bigmath{
  \forall\tau      \tightin\Xsubst\stopq
}
\\\LINEmath{
  \inparentheses{
      C\tau
      \mbox{ fulfilled \wrt\ \redsub}
    \ \implies\ 
      \inparenthesesoplist{
           l\tau> r\tau
         \oplistund
           \forall u\tightin\condterms C\stopq
             l\tau\rhd u\tau
      }
  }
.\footnote{We could require the weaker 
\bigmath{
  \forall u\tightin\condterms C\stopq
  \inparenthesesoplist{
      u\tau\tightnotin\DOM\redsub
    \oplistoder
      l\tau\rhd u\tau
  }
}
instead of 
\bigmath{
  \forall u\tightin\condterms C\stopq
      l\tau\rhd u\tau
}
here.
Theorem~\ref{theoremconfluence} would still be true since its proof need not
be modified.
We did not do this because we did not see an interesting application
that would justify the change of the notion already introduced in 
\citewgctrslncs,
\citewgkp,
and
\citewgjsc.}}
\\
A CRS\/ \R\ over\/ 
\sig/\cons/\/\V\ is\emph{\compatible} 
with a termination-pair\/ \math{(>,\rhd)}
over\/ \sig/\/\V\ 
\udiff\ 
\bigmath{
  \forall\kurzregel\tightin\R     \stopq
  \kurzregel\mbox{ is \RX-compatible with }(>,\rhd)
.}
\end{definition}

\vfill

\begin{theorem}[Syntactic Test for Confluence]
\label{theoremconfluence}\mbox{}\\
Let\/ \R\ be a CRS over\/ \sig/\cons/\/\V\ 
and
\bigmath{\X\tightsubseteq\V.}
\\
Assume that \R\ is \compatible\  with a termi\-nation-pair\/
\math{(>,\rhd)} over\/ \sig/\/\V\@.
\\
{[For each \math{t\in\tsigX} assume \math{\lll_t} to be a wellfounded
ordering on \TPOS t.
Define (\math{p\tightin\N_+^\ast})
\bigmath{
  A(p)
  :=
  \setwith
    {t\tightin\DOM{\redindex{\RX,\omega+\omega,q}}}
    {\emptyset\tightnotequal q\lll_t p}
.}]}
\\
The following two are logically equivalent:
\begin{enumerate}

\notop
\item
Each critical peak in \math{{\rm CP}(\R)}
is \math\rhd-weakly
joinable \wrt\ \RX\ {[besides \math A]}\@.

\noitem
\item
\redsub\ is confluent.

\yestop
\yestop
\yestop
\yestop
\end{enumerate}
\end{theorem}

\vfill

\pagebreak

\noindent
Due to a weakening of the notion of \math\rhd-weak joinability,
Theorem~\ref{theoremconfluence} actually differs from
Theorem~7.17 of \citewgjsc\ in that it provides several irreducibility
assumptions intended to restrict the number of substitutions \math\varphi\
for which for a critical peak 
\\\linemath{ 
  \inparentheses{
    \inparenthesesinlinetight{
      \repl
        {l_1}
        {p}
        {r_0},
      C_0,
      \ldots
    },\ 
    \inparenthesesinlinetight{
      r_1,
      C_1,
      \ldots
    },\
    l_1,\ 
    \sigma,\
    p 
  }
}
resulting from two rules
\sugarregelindex0 and \sugarregelindex1 (with no variables in common)
we have to show 
\bigmath{
  \repl
    {l_1}
    {p}
    {r_0}
  \sigma\varphi
  \downarrowsub
  r_1\sigma\varphi
}
in case of \bigmath{(C_0 C_1)\sigma\varphi} being fulfilled.
This means that 
Theorem~\ref{theoremconfluence} provides further means to tackle
problem~\ref{item infinite number of substitutions} of our 
\sect~\ref{section motivation}.

The first assumption allowed
is that the substitution \math\varphi\ itself
is normalized: 
\bigmath{
  \forall x\in\V\stopq
    x\varphi\tightnotin\DOM\redsub
.}

The second allows to assume that for non-overlays 
(\ie\ for \math{p\tightnotequal\emptyset})
even \math{\sigma\varphi} is 
normalized on all variables occurring in the left-hand side \math{l_1}.

Moreover, by weakening ``\math\rhd-weak joinability'' to 
``\math\rhd-weak joinability besides \math A'' 
with \math A defined as in the theorem via some family 
\math{\ggg}
=
\math{(\ggg_t)_{t\in\tsigX}}
of arbitrary wellfounded orderings \math{\ggg_t} on \TPOS t,
we have added a new feature which allows to assume the 
instantiated peak term (or superposition term) \math{l_1\sigma\varphi} 
to be irreducible at all nonempty positions which are 
\math{\lll_{l_1\sigma\varphi}}-smaller than the 
overlap position \math p. 
Generally, 
beyond our first two assumptions,
we may use \math\lll\ to further reduce the number of instantiations
for which the joinability test must succeed in the following way:
If we can choose \math{\lll_{l_1\sigma\varphi}} such that
\\\LINEmath{
  \inparentheses{
  p\tightequal\emptyset
  \ \implies\ 
  \forall x\tightin\VAR{l_1}\stopq
  \inparenthesesoplist{
  x\sigma\tightnotequal x
  \oplistimplies
  \exists q\tightin\TPOS{l_1}\stopq
    \inparenthesesoplist{
        l_1/q\tightequal x
      \oplistund
        \forall q'\tightin\TPOS{x\sigma\varphi}\stopq
           q q'\lll_{l_1\sigma\varphi} p
    }
  }
  }
}
\\
as well as
\LINEmath{
  \forall x\tightin\VAR{l_0}\stopq
  \inparenthesesoplist{
  x\sigma\tightnotequal x
  \oplistimplies
  \exists q\tightin\TPOS{l_0}\stopq
    \inparenthesesoplist{
        l_0/q\tightequal x
      \oplistund
        \forall q'\tightin\TPOS{x\sigma\varphi}\stopq
           p q q'\lll_{l_1\sigma\varphi} p
    }
  }
,}
\\
then we may assume \math{\sigma\varphi} to be normalized:
\bigmath{
  \forall x\tightin\V\stopq
    x\sigma\varphi\tightnotin\DOM\redsub
.}
This can be a considerable help for showing that 
\math{(C_0 C_1)\sigma\varphi} is not fulfilled 
when we have a certain knowledge on the normal forms of the terms of the
sorts of the variables occurring in \math{C_0 C_1}. 
\Eg, when we define the depth of a term \math{t\in\vt} by
\bigmath{
  \depth{t}
  :=
  \max\setwith{\CARD{p'}}{p'\tightin\TPOS t}
}
and then define (\math{p,q\tightin\TPOS t})
\bigmath{q \lll_t p}
\udiff\
\bigmath{
  \depth t-\CARD q
  \prec
  \depth t-\CARD p
,}
then we can forget about all critical peaks which are
called ``composite'' in \sect~2.3 of \citekapurcp\ 
--- and even some more, namely all those whose peak term 
is reducible at some position 
that is longer than the overlap position of the critical peak. 
\citekapurcp\ already states in Corollary~5 that 
(unless \math{l_0\tightin\V}, which some authors generally disallow)
the irreducibility of these positions implies the 
irreducibility of all terms introduced by the unifying substitution 
\math\sigma; more precisely, the joinability test may assume:
\bigmath{
  \forall x\tightin\V\stopq
  \inparentheses{
       x\sigma\tightnotequal x
     \ \implies\ 
       x\sigma\varphi\tightnotin\DOM\redsub
  }
,}
which, by our first irreducibility assumption can be simplified to 
\bigmath{
  \forall x\tightin\V\stopq
       x\sigma\varphi\tightnotin\DOM\redsub
.}
If we, however, revert \math{\lll} by defining
\bigmath{q \lll_t p}
\udiff\
\bigmath{
  \CARD q
  \prec
  \CARD p
,}
then we can forget about all critical peaks which are
called ``composite'' in \sect~4.1 of \citekapurcp\ 
--- and even some more, 
namely all those whose peak term 
is reducible at some nonempty position that is shorter than
the overlap position of the critical peak. 

\pagebreak

\yestop
\noindent
The power of the combination of the two weakenings of the
joinability requirement,
\ie\ the confluence and the irreducibility assumptions,
is demonstrated by the following 
simple but non-trivial example whose predicate `\nonnegpsymbol'
checks whether an integer number is non-negative:

\yestop
\yestop
\begin{example}\label{example integers}
\sloppy
\arr{[t]{lll}
 \consfunsym
&:=
&\{\zeropp, \ssymbol,\psymbol,\truepp,\falsepp\}
\\\deffunsym
&:=
&\{\nonnegpsymbol\}
\\
\R_{\,\rm\ref{example integers}}
&:
&\begin{array}[t]{llll}
     \spp{\ppp y}            &=&y       \\
     \ppp{\spp y}            &=&y       \\
     \nonnegppp\zeropp       &=&\truepp \\
     \nonnegppp{\spp x}      &=&\truepp &\rs\nonnegppp x=\truepp\\
     \nonnegppp{\ppp\zeropp} &=&\falsepp\\
     \nonnegppp{\ppp x}      &=&\falsepp&\rs\nonnegppp x=\falsepp\\
 \end{array}
}

\noindent
Let\/ \zeropp, \ssymbol, \psymbol\ be constructor symbols of the sort
\intsort\
and\/ \truepp, \falsepp\ constructor symbols of the sort\/ \bool.
Let\/ \nonnegpsymbol\ be a non-constructor predicate with arity
``\/\mbox{\math{\ \intsort\aritysugar\bool\ }}''. 
Let \math x, \math y be constructor variables of the sort\/ \intsort.

Obviously, \math{\R_{\,\rm\ref{example integers}},\V} 
is \Vcompatible\
with the termination-pair \math{(\tight\rhd,\tight\rhd)} where
\tight\rhd\ is the lexicographic path ordering generated by 
\nonnegpsymbol\ being bigger than \truepp\ and \falsepp.

There are only the following two critical peaks 
which are both of the form \math{(0,1)}:
\begin{diagram}
&\nonnegppp{\spp{x}}\sigma&&\rred&&\truepp
&&&\nonnegppp{\ppp{x}}\sigma'&&\rred&&\falsepp
\\
&\dredindex{1,\,1}&&&&
&&&\dredindex{1,\,1}&&&&
\\
&\nonnegppp y&&&&
&&&\nonnegppp y&&&&
\\
\end{diagram}
where 
\math{\sigma:=\{x\mapsto\ppp y\}}
and
\math{\sigma':=\{x\mapsto\spp y\}}.
Their respective condition lists are the following two lists containing
each one literal only:
\begin{diagram}
&&&\nonnegppp{x}\sigma\tightequal\truepp&&
&&&&&\nonnegppp{x}\sigma'\tightequal\falsepp&&
\end{diagram}

\noindent 
Now the following is easy to show:
The irreducible constructor terms of the sort \intsort\ are exactly 
the terms of the form \sppiterated n z or \pppiterated{n+1}z
with \math{n\tightin\N} and 
\math{z\tightin\Vconsindex\intsort\tightcup\{\zeropp\}}.
The irreducible constructor terms of the sort \bool\ are 
\bigmath{\Vconsindex\bool\tightcup\{\truepp,\falsepp\}}.
Furthermore, by induction on \math{n\tightin\N} one easily shows
\bigmath{
  \nonnegppp{\sppiterated n\zeropp}
  \refltransindex{\R_{\,\rm\ref{example integers}},\emptyset}
  \truepp
}
and
\bigmath{
  \nonnegppp{\pppiterated{n+1}\zeropp}
  \refltransindex{\R_{\,\rm\ref{example integers}},\emptyset}
  \falsepp
.}
Finally
by induction on \math{n\tightin\N} one easily
shows that
\bigmath{
  \nonnegppp t
  \refltransindex{\R_{\,\rm\ref{example integers}},\V,\omega+n}
  \truepp
\ \oder\ 
  \nonnegppp t
  \refltransindex{\R_{\,\rm\ref{example integers}},\V,\omega+n}
  \falsepp
}
implies
\bigmath{
  \VAR t\tightequal\emptyset
,}
which we only need to show confluence besides ground confluence.

Define \math\lll\ via (\math{p,q\tightin\TPOS t}):
\bigmath{q \lll_t p}
\udiff\
\bigmath{
  \depth t-\CARD q
  \prec
  \depth t-\CARD p
.}
Now the new combined weakening of joinability to 
\math\rhd-weak joinability \wrt\ \math{\R_{\,\rm\ref{example integers}},\V} 
besides \math A (with \math A defined as in the theorem)
allows us to show joinability of the above critical peaks very easily.
Since the second critical peak can be treated analogous to the first,
we explain how to treat the first only:
By the new additional feature for assuming irreducibility,
our weakened joinability allows to assume
that \bigmath{x\sigma\varphi} 
is irreducible for the first critical peak,
which can be seen in two different ways:
First, since the critical peak is a non-overlay and \math x occurs in the 
peak term \nonnegppp{\spp x}.
Second, since the overlap position is \bigmath{1,}
\bigmath{
  \nonnegppp{\spp x}/1\ 1
  =
  x
}
and 
\bigmath{
  \forall q'\tightin\TPOS{x\sigma\varphi}\stopq
    \ 1\ 1\ q'
    \lll_{\nonnegppp{\spp x}\sigma\varphi}
    1
.}
Furthermore, we are allowed to assume that the condition of the critical
peak is fulfilled,
\ie\ that 
\bigmath{
  \nonnegppp{x}\sigma\varphi
  \refltransindex{\R_{\,\rm\ref{example integers}},\V}
  \truepp
.}
Together with the irreducibility of 
\math{x\sigma\varphi\tightequal\ppp y\varphi} this implies
that \math{y\varphi} is of the form 
\pppiterated n\zeropp.
This again implies
\bigmath{
  \nonnegppp{x}\sigma\varphi
  \refltransindex{\R_{\,\rm\ref{example integers}},\V}
  \falsepp
.}
But since we may assume confluence below the condition term
\bigmath{\nonnegppp{x}\sigma\varphi}
we get 
\bigmath{
  \truepp
  \downarrowindex{\R_{\,\rm\ref{example integers}},\V}
  \falsepp
,} 
which is impossible.
Thus the properties that weak joinability allows us to assume for the
joinability test are inconsistent and the critical pair need not be joined
at all.

All in all, Theorem~\ref{theoremconfluence} implies confluence of 
\redindex{\R_{\,\rm\ref{example integers}},\V} without solving the task
of showing that for each arbitrary 
(not necessarily normalized)
substitution \math\varphi\ 
either
\bigmath{
  \nonnegppp{\ppp y}\varphi
  \refltransindex{\R_{\,\rm\ref{example integers}},\V} 
  \truepp
}
does not hold or\/  
\bigmath{
  \nonnegppp{y}\varphi
  \refltransindex{\R_{\,\rm\ref{example integers}},\V} 
  \truepp
}
holds, 
which is more difficult to show than 
our simple properties above.
\end{example}

\vfill

\pagebreak

\yestop
\yestop
\yestop
\yestop
\noindent
The following theorem is a generalization of Theorem~7.18 in \citewgjsc. 
In comparison with Theorem~\ref{theoremconfluence} 
it offers for each condition term \math u of a rule \sugarregel\
the possibility to replace 
the requirement
\bigmath{l\tau\rhd u\tau}
(roughly speaking \ie\ decreasingness)
with 
\bigmath{\VAR u\tightsubseteq\Vcons}
(\ie\ the absence of general variables).
The basic idea of its proof is first to show \math\omega-shallow confluence
up to \math{\omega} 
(\ie\ commutation of \redindex{\RX,\omega} and \redsub)
with the usual argumentation on quasi-normality, left-linearity, termination
and \math\omega-shallow joinability
(\cf\ Theorem~\ref{theorem quasi-free}), 
and then to use decreasingness argumentation for the confluence
of \redsub.

\yestop
\yestop
\yestop
\begin{theorem}[Syntactic Test for Confluence]
\label{theorem quasi-free three}
\sloppy

\noindent
Let\/    \R\ be a CRS over \sig/\cons/\/\V\@.
Let\/    \bigmath{\X\tightsubseteq\V.}
\\
Assume
the following very weak kind of left-linearity of constructor rules
\wrt\ general variables:
\\\math{
  \forall\kurzregel\tightin\R\stopq
  \forall x  \tightin\Vsig   \stopq
  \forall p,q\tightin\TPOS l  \stopq
}
\\\LINEmath{
          \inparentheses{
            \inparenthesesoplist{
                l\tightin\tcs
              \oplistund
                l/p\tightequal x\tightequal l/q
            }
            \ \implies\
                p\tightequal q
          }
.}

\noindent
Furthermore, assume that constructor rules are quasi-normal \wrt\ \RX:
\\\math{
  \forall\kurzregel\tightin\R\stopq
  \forall\tau\tightin\Xsubst\stopq
}
\\\LINEmath{
  \inparentheses{
       \inparenthesesoplist{
           l\tightin\tcs
         \oplistund
           C\tau\mbox{ fulfilled \wrt\ }\redindex{\RX,\omega}
       }
     \ \implies\ 
       \kurzregel
       \mbox{ is quasi-normal \wrt\ \RX}
  }
.}

\noindent
Moreover, assume the following compatibility property for 
a termination-pair \math{(>,\rhd)} over \sig/\/\V:
\\\math{
  \forall\kurzregel\tightin\R\stopq
  \forall\tau\tightin\Xsubst\stopq
}\\\LINEmath{
  \inparentheses{ 
        C\tau\mbox{ fulfilled \wrt\ }\redindex{\RX}
    \ \implies
    \inparenthesesoplist{
        l\tau> r\tau
      \oplistund
        \forall u\in\condterms C\stopq
          \inparenthesesoplist{
              l\tau\rhd u\tau
            \oplistoder 
              u\tau\tightnotin\DOM\redsub
            \oplistoder
               \VAR u\subseteq\Vcons
          }\!\!
    }\!\!
  }   
.}

\noindent
Assume\/ \redindex{\RX,\omega} to be confluent.

\noindent
Assume that each critical peak 
\bigmath{\criticalpeaklongform\in{\rm CP}(\R)}
\\
(with \bigmath{(\Lambda_0,\Lambda_1)\tightnotequal(1,1)}
and 
\inparenthesesinline{
    (\Lambda_0,\Lambda_1)\tightnotequal(0,0)
  \ \oder\ 
    \condterms{D_0\sigma\,D_1\sigma}\tightnotsubseteq\tcc
})
\\
is \math\omega-shallow joinable  up to \math{\omega} \wrt\ \RX\ and \math\lhd.

\noindent
{[For each \math{t\in\tsigX} assume \math{\lll_t} to be a wellfounded
ordering on \TPOS t.
Define (\math{p\tightin\N_+^\ast})
\bigmath{
  A(p)
  :=
  \setwith
    {t\tightin\DOM{\redindex{\RX,\omega+\omega,q}}}
    {\emptyset\tightnotequal q\lll_t p}
  \nottight\cup 
  \DOM{\redindex{\RX,\omega}}
.}]}

\yestop
\noindent
Now the following two are logically equivalent:
\begin{enumerate}

\noitem
\item
Each critical peak 
\bigmath{\criticalpeaklongform\in{\rm CP}(\R)}
\\
(with 
\math{
  \forall k\tightprec2\stopq
  \inparenthesesinline{
      \Lambda_k\tightequal1
    \ \oder\ 
      \condterms{D_k\sigma}\tightnotsubseteq\tcc
  }
})
\\
is \math\rhd-weakly joinable \wrt\ \RX\ {[besides \math A]}\@.

\noitem
\item
\redsub\ is confluent.

\end{enumerate}
\end{theorem}

\vfill

\pagebreak

\yestop
\yestop
\noindent
The following theorem generalizes Theorem~2 in \citeder\
by weakening normality to quasi-normality.

\yestop
\yestop
\yestop
\begin{theorem}[Syntactic Test for \math\omega-Shallow Confluence]
\label{theorem quasi-free}
\sloppy\mbox{}

\noindent
Let\/    \R\ be a CRS over \sig/\cons/\/\V\@.
Let\/    \bigmath{\X\tightsubseteq\V.}
\\
Assume
the following weak kind of left-linearity \wrt\ general variables:
\\\math{
  \forall\kurzregel\tightin\R    \stopq
  \forall x        \tightin\Vsig \stopq
  \forall p,q      \tightin\TPOS l\stopq
          \inparentheses{
                l/p\tightequal x\tightequal l/q
            \ \implies\
                p\tightequal q
          }
.}

\noindent
Furthermore, assume \RX\ to be quasi-normal.

\noindent
Let\/ \math{(\tight>,\tight\rhd)} be a termination-pair over \sig/\/\V\ 
such that the following compatibility property for constructor rules holds
(which is always satisfied when \R\ has conservative constructors):
\\\math{
  \forall\kurzregel\tightin\R\stopq
  \forall\tau\tightin\Xsubst\stopq
}
\\
\LINEmath{
  \inparentheses{ 
    \inparenthesesoplist{
        l\in\tcs
      \oplistund
        C\tau\mbox{ fulfilled \wrt\ }\redindex{\RX,\omega}
    }
    \implies\
        \forall u\tightin\condterms C\stopq
          \inparenthesesoplist{
              l\tau\rhd u\tau
            \oplistoder 
              u\tau\tightnotin\DOM\redsub
            \oplistoder
               \VAR u\subseteq\Vcons
          }
  }   
.}

\noindent
Furthermore, assume that the system is terminating:
\\\LINEmath{
  \forall\kurzregel\tightin\R\stopq
  \forall\tau\tightin\Xsubst\stopq
  \inparentheses{ 
      \inparentheses{
        C\tau\mbox{ fulfilled \wrt\ }\redindex{\RX}
      }
    \ \implies\ 
        l\tau> r\tau
  }   
.}

\noindent
{[For each \math{t\in\tsigX} assume \math{\lll_t} to be a wellfounded
ordering on \TPOS t.
Define (\math{p\tightin\N_+^\ast}, \math{n\tightprec\omega})
\bigmath{
  A(p,n)
  :=
  \setwith
    {t\tightin\DOM{\redindex{\RX,\omega+n,q}}}
    {\emptyset\tightnotequal q\lll_t p}
.}]}

\yestop
\noindent
Now the following two are logically equivalent:
\begin{enumerate}

\notop
\item
\redindex{\RX,\omega} is confluent and 
\\
each critical peak 
\bigmath{\criticalpeaklongform\in{\rm CP}(\R)}
\\
(with 
\inparenthesesinline{
    (\Lambda_0,\Lambda_1)\tightnotequal(0,0)
  \ \oder\ 
    \condterms{D_0\sigma\,D_1\sigma}\tightnotsubseteq\tcc
})
\\
is \math\omega-shallow joinable \wrt\ \RX\ and \math\lhd\ 
{[besides \math A]}\@.

\noitem
\item
\RX\ is \math\omega-shallow confluent.
\end{enumerate}
\end{theorem}

\vfill

\pagebreak

\yestop
\yestop
\noindent
The following theorem weakens the \math\omega-shallow joinability
requirement to that of \math\omega-level joinability, but disallows
general variables in conditions of rules.
That this restriction is necessary indeed can be seen from
Example~\ref{ex toll}.

\yestop
\yestop
\yestop
\begin{theorem}[Syntactic Test for \math\omega-Level Confluence]
\label{theorem level one}
\sloppy

\noindent
Let\/    \R\ be a CRS over \sig/\cons/\/\V\@.
Let\/    \bigmath{\X\tightsubseteq\V.}

\noindent
Assume
\bigmath{
  \forall\kurzregel\tightin\R\stopq
      \VAR C\tightsubseteq\Vcons
.}

\noindent
Let\/ \math{(\tight>,\tight\rhd)} be a termination-pair over \sig/\/\V\@.
Assume that the system is terminating:
\\\mbox{}\hfill\math{
  \forall\kurzregel\tightin\R\stopq
  \forall\tau\tightin\Xsubst\stopq
  \inparentheses{ 
    \inparentheses{
      C\tau\mbox{ fulfilled \wrt\ }\redsub
    }
    \implies\
      l\tau> r\tau
  }   
.}

\noindent
{[For each \math{t\in\tsigX} assume \math{\lll_t} to be a wellfounded
ordering on \TPOS t.
Define (\math{p\tightin\N_+^\ast}, \math{n\tightprec\omega})
\bigmath{
  A(p,n)
  :=
  \setwith
    {t\tightin\DOM{\redindex{\RX,\omega+n,q}}}
    {\emptyset\tightnotequal q\lll_t p}
.}]}

\yestop
\noindent
Now the following two are logically equivalent:
\begin{enumerate}

\notop
\item
\redindex{\RX,\omega} is confluent and 
each critical peak in \math{{\rm CP}(\R)}
\\
of the forms \math{(0,1)},  \math{(1,0)}, or \math{(1,1)}  
\\
is \math\omega-level joinable \wrt\ \RX\ and \math\rhd\ {[besides \math A]}\@.

\noitem
\item
\RX\ is \math\omega-level confluent.

\end{enumerate}
\end{theorem}

\vfill

\pagebreak

\yestop
\yestop
\yestop
\yestop
\yestop
\yestop
\yestop
\yestop
\yestop
\yestop
\noindent
The following theorem generalizes Theorem~4 in \citeder\
and Theorem~6.3 in \citewgjsc\
by weakening overlay joinability to \math\rhd-quasi overlay joinability.
For a discussion of the 
notion of \math\rhd-quasi overlay joinability
\cf\ \sect~\ref{sect quasi overlay joinability}.
The 
proof is discussed above 
the key 
lemma~\ref{lemma for theorem quasi overlay joinable}.

\yestop
\yestop
\yestop
\yestop
\begin{theorem}[Syntactic Confluence Criterion]
\label{theorem quasi overlay joinable}

\noindent
Let\/ \R\ be a CRS over \sig/\cons/\/\V\ and\/ \math{\X\tightsubseteq\V}.

\noindent
Assume either that\/ \redsub\ is terminating\/%
\footnote{Actually innermost termination is enough here when we require
overlay joinability instead of \math\rhd-quasi overlay joinability,
\cf\ \citegramlichelkuz.}
and\/
\bigmath{\tight\rhd\tightequal\superterm}
or that\/ 
\bigmath{\redsub\subseteq\tight\rhd,}
\bigmath{\superterm\subseteq\tight\rhd,}
and\/ \math\rhd\ is a wellfounded ordering on \vt.
 
\noindent
Now, if all critical peaks in ${\rm CP}(\R)$ are
\math\rhd-quasi overlay joinable \wrt\ \RX,
\\
then \redsub\ is confluent.

\end{theorem}

\yestop
\yestop
\yestop
\yestop
\begin{example}
\label{ex quasi over}

\yestop
\noindent
Let\/ \math{\X\tightsubseteq\V}.
The following system 
is neither decreasing, nor left-linear, nor overlay
joinable; but it is terminating and \superterm-quasi overlay joinable
\wrt\ \math{\R_{\,\rm\ref{ex quasi over}},\X}.
Thus Theorem~\ref{theorem quasi overlay joinable} is the only one that
implies confluence of\/ \redindex{\R_{\,\rm\ref{ex quasi over}},\X}.
Note that Theorem~\ref{theorem quasi-free three} becomes applicable
when we replace the non-constructor variable in (\/\ppsymbol1) with
a constructor variable. Moreover, if we additionally do the same with
(\/\ppsymbol2), then Theorem~\ref{theorem level one} becomes applicable, too.

\yestop
\noindent
Even though it is irrelevant for Theorem~\ref{theorem quasi overlay joinable},
let \math{X,Y\in\Vsig}, 
\math{\zeropp,\ssymbol,\appzero,\truepp,\falsepp\in\consfunsym},
and\/ \math{\lessymbol,\ppsymbol,\fsymbol,\gsymbol\in\sigfunsym}.
Note that \math{\zeropp,\ssymbol,\appzero,\lessymbol} 
model the ordinal number \math{\omega\tight+1}.

\yestop
\noindent
\math{\R_{\,\rm\ref{ex quasi over}}}: \mbox{}
\math{
  \begin{array}[t]{@{}l@{\mbox{~~~~~~~~~}}l@{\ =\ }ll}
    \\
    \mbox{(\/\ssymbol1)}
    &\spp\appzero
    &\appzero
    &
    \\
    \mbox{(\/\lessymbol1)}
    &\lespp{\spp X}{\spp Y}
    &\lespp X Y
    &
    \\
    \mbox{(\/\lessymbol2)}
    &\lespp X X
    &\falsepp
    \\
    \mbox{(\/\lessymbol3)}
    &\lespp\zeropp{\spp Y}
    &\truepp
    \\
    \mbox{(\/\lessymbol4)}
    &\lespp X\zeropp
    &\falsepp
    \\
    \mbox{(\/\lessymbol5)}
    &\lespp\zeropp\appzero
    &\truepp
    \\
    \mbox{(\/\lessymbol6)}
    &\lespp\appzero{\spp Y}
    &\lespp\appzero Y
    &
    \\
    \mbox{(\/\lessymbol7)}
    &\lespp{\spp X}\appzero
    &\lespp X\appzero
    &
    \\
    \mbox{(\/\ppsymbol1)}
    &\pppp X
    &\truepp
    &\rs\pppp{\spp X}\boldequal\truepp
    \\
    \mbox{(\/\ppsymbol2)}
    &\pppp X
    &\truepp
    &\rs\lespp{\fppeins X}{\gppeins X}\boldequal\truepp
    \\
    \mbox{(\/\fsymbol\math i)}
    &\fppeins X
    &\ldots
    &
    \\
    \mbox{(\/\gsymbol\math i)}
    &\gppeins X
    &\ldots
    &
    \\
  \end{array}
}

\pagebreak

\yestop
\noindent
The critical peaks are the following:

\noindent
From (\/\ssymbol1) into (\/\lessymbol1) we get:
\begin{diagram}
\lespp{\spp\appzero}{\spp Y}&&\rredindex{\omega+1}&&\lespp\appzero Y
\\
\dredindex{1,\,1}&&&&\dequal
\\
\lespp\appzero{\spp Y}&&\rredindex{\omega+1,\,\emptyset}&&\lespp\appzero Y
\\
\\
\lespp{\spp X}{\spp\appzero}&&\rredindex{\omega+1}&&\lespp X\appzero
\\
\dredindex{1,\,2}&&&&\dequal
\\
\lespp{\spp X}\appzero&&\rredindex{\omega+1,\,\emptyset}&&\lespp X\appzero
\\
\\
\lespp{\spp\appzero}{\spp\appzero}&&\rredindex{\omega+1}
&&\lespp\appzero\appzero
\\
\dredparaindex{1,\,\{1,2\}}&&
&&\dequal
\\
\lespp\appzero\appzero&&\requal
&&\lespp\appzero\appzero
\\
\end{diagram}
From (\/\ssymbol1) into (\/\lessymbol3) we get:
\begin{diagram}
\lespp\zeropp{\spp\appzero}&&\rredindex{\omega+1}&&\truepp
\\
\dredindex{1,\,2}&&&&\dequal
\\
\lespp\zeropp\appzero&&\rredindex{\omega+1,\,\emptyset}&&\truepp
\\
\end{diagram}
The criticial peaks resulting from (\/\ssymbol1) into (\/\lessymbol6) 
and (\/\lessymbol7) are trivial.
\\
From (\/\lessymbol1) into (\/\lessymbol2) we get:
\begin{diagram}
  \lespp{\spp X}{\spp X}&&\rredindex{\omega+1}&&\falsepp
\\
\dredindex{\omega+1,\,\emptyset}&&&&\dequal
\\
\lespp X X&&\rredindex{\omega+1,\,\emptyset}&&\falsepp
\\
\end{diagram}
From (\/\lessymbol2) into (\/\lessymbol1) we get:
\begin{diagram}
  \lespp{\spp X}{\spp X}&&\rredindex{\omega+1}&&\lespp X X
\\
\dredindex{\omega+1,\,\emptyset}&&&&\dredindex{\omega+1,\,\emptyset}
\\
\falsepp&&\requal&&\falsepp
\\
\end{diagram}
The criticial peaks resulting from 
(\/\lessymbol2) into (\/\lessymbol4),
(\/\lessymbol4) into (\/\lessymbol2),
(\/\ppsymbol 1) into (\/\ppsymbol 2),
and
(\/\ppsymbol 2) into (\/\ppsymbol 1)
are trivial.
\end{example}

\vfill\pagebreak

\section{Criteria for Confluence of the Constructor Sub-System}
\label{section constructor confluence}

Define the\emph{constructor sub-system} of a rule system \R\ to be\/
\\\linemath{\R_\CONS:=\setwith{\kurzregel\tightin\R}{l\tightin\tcs},}
\ie\ the system of the constructor rules of \R\@.
In this section we discuss the problem how to find out that
\bigmath{
  \redindex{\RX,\omega}
  =
  \redindex{\R_\CONS,\X,\omega}
} 
is confluent.
Note that this is a necessary ingredient for achieving confluence via
any of the theorems
\ref{theorem complementary},
\ref{theorem weakly complementary},
\ref{theorem parallel closed},
\ref{theorem level parallel closed},
\ref{theorem quasi-free three},
\ref{theorem quasi-free},
and
\ref{theorem level one}.

The easiest way to achieve confluence of \redindex{\RX,\omega}
is to have no constructor rules at all, 
\ie\ \bigmath{\R_\CONS\tightequal\emptyset.} 
While it is rather restrictive,
this case of\emph{free constructors} is very important in practice
since a lot of data structures can be specified this way.
Moreover, it is economic to restrict to this case because 
non-free constructors make a lot of trouble when working with the 
specification, \eg, most techniques for proving inductive validity get
into tremendous trouble with non-free constructors --- if they are able
to handle them at all.

The second case where confluence of \redindex{\RX,\omega} is immediate
is when for each rule \sugarregel\ in \RCONS\ 
also
\bigmath{
  r\boldequal l\rulesugar C
}
is an instance of a rule of \R,
and then also of \RCONS\ due to the restriction on the constructor rule
\sugarregel\ given by Definition~\ref{defrulescontinu}.
An example for this is the commutativity rule which is equal to a renamed
version of the reverse of itself.
In this case it may be worthwhile to consider reduction modulo a 
constructor congruence as described in \citeSR\ and \citebeckerstacs.

A third way to achieve confluence of \redindex{\RX,\omega}
is to use semantic confluence criteria in the style of 
\citeplaisted, \cf\ also Theorem~6.5 in \citewgjsc.
While this semantic argumentation is very powerful when one has sufficient
knowledge about the constructor domain, it is, however, not at all obvious
how to formalize or even automate such semantic considerations.
Above that, these semantic confluence criteria 
are based on the existence of normal forms and therefore require termination
of the constructor sub-system (at least in some weak form). 

Termination of the constructor sub-system, of course, does not mean termination
of the whole rule system. We may, \eg, apply Theorem~\ref{theoremconfluence}
to infer confluence of a terminating constructor sub-system containing the 
associativity rule of Example~\ref{ex for asso} 
(whose confluence can hardly be inferred without termination)
and then infer the 
confluence of the whole non-terminating rule system by some of the theorems
of \sect~\ref{sect non-terminating confluence}.
This case where a terminating constructor sub-system is part of a non-terminating
rule system seems to be important in practice since confluence of 
non-free constructors often can hardly be inferred without termination 
whereas termination is usually not needed for then inferring confluence 
of the whole system
because the non-constructor rules can be chosen in such a way that their
critical peaks are complementary, \cf\ Theorem~\ref{theorem complementary}. 
Moreover note that the reverse case, \ie\ that of a non-terminating constructor
sub-system of a terminating rule system, is impossible in our framework
but not in the abovementioned one of \citeSR\ and \citebeckerstacs\
where the notion of reduction
is different, namely reduction via \bigmath{\R\tightsetminus\R_\CONS} modulo
\math{\R_\CONS}.

In the rest of this section we will present syntactic criteria for confluence
of \redindex{\RX,\omega}.

\vfill

\pagebreak

\yestop
\noindent
First note that the theorems \ref{theoremconfluence} and 
\ref{theorem quasi overlay joinable} can directly be applied to infer
confluence of \redindex{\RX,\omega} simply by instantiating the `\R'\ of these
theorems with \math{\R_\CONS}.

The other theorems we will present in the following are nothing but 
informal corollaries of other theorems of the 
sections \ref{sect non-terminating confluence} 
and \ref{section now termination}.
To apply the latter theorems to our special case here, 
it is not sufficient only to throw away the non-constructor rules,
but we also have to transform the constructor function symbols of the
constructor rules into non-constructor function symbols.
For consistency we then also have to rename their constructor variables
with general variables. 
Then the constructor sub-system of the transformed system 
is empty and therefore trivially confluent, 
such that these theorems can be applied.
If the constructor rules contain general variables or
\Def-literals, 
then, however, this transformation
brings us beyond the two layered framework presented in this paper:
As we translate 
constructor  variables (level~\math 0) 
into general variables (level~\math 1),
then, for consistency, since \redindex{\RX,\omega} is a relation on the
terms of the whole signature,
we also have to translate general variables (level~\math 1)
into some kind of variables of level~\math 2,
and non-constructor function symbols (level~\math 1)
into some kind of function symbols of level~\math 2.
Symbols of level~\math 2, however, are not present in the framework
presented in this paper. Moreover we have to translate our \Def-literals
(which test for reducibility to a ground term of level~\math 0) into predicate
literals that test for reducibility to a ground term of level~\math 1, which
are also not present in our framework. While it would be possible 
and beautiful to present our confluence criteria of the sections
\ref{sect non-terminating confluence} and \ref{section now termination}
in a framework with a special signature and variable-system for the level
of each natural number, we have decided not to do so for the following reasons:
First, it would make the paper even more technically and conceptually
difficult as it is. 
Second, the infinitely layered framework may be of little importance 
(since its only useful application so far is this section). 
Third, the step of level~\math 0 we want to
treat here may in principle allow of more powerful criteria 
than an arbitrary level~\math i
and therefore it does not seem to be a good idea to achieve its confluence
criteria as corollaries of the theorems for an arbitrary level.
Fourth, by proving the theorems of this section separately, 
we provide the reader interested only 
in the standard positive conditional rule systems without
constructor sub-signature and constructor sub-system 
with a direct approach to this special
case. This can clearly be seen when one translates a system of the standard 
positive conditional framework into our framework by simply saying that all
its symbols are constructor symbols.

\yestop
\yestop
\yestop
\noindent
For all the following theorems let \math{\R_\CONS} 
be the constructor sub-system of a CRS \R\ over \sig/\cons/\/\V\
as defined above,
and let \bigmath{\X\tightsubseteq\V.}
Note that the critical peaks in \math{{\rm CP}(\R_\CONS)} are exactly the
critical peaks of the form \math{(0,0)} in \math{{\rm CP}(\R)}.

\vfill

\mycomment{%%%%%%%%%%%%%%%%%%%%%%%%%%%%%%%%%%%%%%%%%%%%%%%%%%%%%%%%%%%%%%%%%%%%%%
\yestop
\begin{definition}[\math 0-Shallow Parallel Closed]%
\label{def parallel closed zero}\mbox{}
\\
Let\/ \math{\beta\preceq\omega}.
A critical peak \newcriticalpeak\
\\ 
is\emph
{\math 0-shallow parallel closed up to \math\beta\ \wrt\ \RX} 
\udiff\\
\math{
  \forall\varphi\tightin\Xsubst\stopq
  \forall n_0,n_1\prec\omega\stopq
}
\\\LINEmath{
  \inparentheses{
    \inparenthesesoplist{
         n_0\tightsucceq n_1
       \oplistund
         n_0\tight+ n_1
         \preceq
         \beta
       \oplistund
         \forall i\prec 2\stopq
         \inparenthesesoplist{
             1\preceq n_i
           \oplistund
             D_i\varphi\mbox{ fulfilled \wrt\ }
             \redindex{\RX,n_i\monus 1}
         }
       \oplistund
         \forall\delta\tightprec n_0\tight+ n_1\stopq
           \RX\mbox{ is \math 0-shallow confluent up to }\delta
    }
    \\\implies\ 
             t_0\varphi
             \redparaindex{\RX,n_1}
             \tight\circ
             \refltransindex{\RX,n_1\monus1}
             t_1\varphi
  }
.}
\\\headroom
It is called\emph{\math 0-shallow parallel closed \wrt\ \RX} \udiff\\
\LINEnomath{
  it is \math 0-shallow parallel closed 
  up to  \math{\omega} \wrt\ \RX\@.
}
\end{definition}

\pagebreak

\begin{definition}[\math 0-Shallow {[Noisy]} Parallel Joinable]%
\label{def parallel joinable zero}\mbox{}
\\
Let\/ \math{\beta\preceq\omega}.
A critical peak \newcriticalpeak\
\\ 
is\emph
{\math 0-shallow {\rm[}noisy\/{\rm]} 
parallel joinable up to \math\beta\ \wrt\ \RX} 
\udiff\\
\math{
  \forall\varphi\tightin\Xsubst\stopq
  \forall n_0,n_1\prec\omega\stopq
}
\\\LINEmath{
  \inparentheses{
    \inparenthesesoplist{
         n_0\tightpreceq n_1
       \oplistund
         n_0\tight+ n_1
         \preceq
         \beta
       \oplistund
         \forall i\prec 2\stopq
         \inparenthesesoplist{
             1\preceq n_i
           \oplistund
             D_i\varphi\mbox{ fulfilled \wrt\ }
             \redindex{\RX,n_i\monus 1}
         }
       \oplistund
         \forall\delta\tightprec n_0\tight+ n_1\stopq
           \RX\mbox{ is \math 0-shallow confluent up to }\delta
    }
    \\\implies\ 
      t_0\varphi
      \redparaindex{\RX,n_1}
      \mbox{[}\tight\circ\ 
      \refltransindex    {\RX,n_1\monus1}
      \mbox{]}
      \circ
      \antirefltransindex{\RX,n_0}
      t_1\varphi
  }
.}
\\\headroom
It is called\emph{\math 0-shallow {\rm[}noisy\/{\rm]} parallel 
joinable \wrt\ \RX} \udiff\\
\LINEnomath{
  it is \math 0-shallow {[noisy]} parallel joinable 
  up to  \math{\omega} \wrt\ \RX\@.
}
\end{definition}
Note that \math 0-shallow parallel closedness
specializes to the standard definition of parallel closedness of
\citehuet\ 
for the
case that all symbols are considered to be constructor symbols 
and the rule system is unconditional
(since then 
\math{\redindex{\RX,1}\tightequal\redsub}
and
\math{\redindex{\RX,0}\tightequal\emptyset}).
Similarly, \math 0-shallow parallel joinability specializes for this case to
the joinability required for overlays in \citetoyama.
}%comment%%%%%%%%%%%%%%%%%%%%%%%%%%%%%%%%%%%%%%%%%%%%%%%%%%%%%%%%%%%%%%%%%%%%%%

\pagebreak

\yestop
\noindent
The following is the analogue of parts (I) and (II) of 
Theorem~\ref{theorem parallel closed}. Note that we do not present the 
analogues of parts (III) and (IV) because they are subsumed%
\footnote{This is because the 
notion of \math     0-shallow {[noisy]}\emph{weak} parallel joinability 
(when defined analogous to the 
notion of \math\omega-shallow {[noisy]}\emph{weak} parallel joinability)
is identical to the 
notion of \math     0-shallow {[noisy]} parallel joinability.}
by the analogue  of part  (I).

\begin{theorem}[Syntactic Criterion for \math 0-Shallow Confluence]
\label{theorem parallel closed zero}
\sloppy
\\
Assume \math{\RX} to be \math 0-quasi-normal 
and \RCONS\ to be left-linear.
\begin{enumerate}

\item[{\bf(I)}]
Now if each critical peak in \math{{\rm CP}(\R)}
of the form \math{(0,0)} 
is\/ \math 0-shallow noisy parallel joinable \wrt\ \RX, 
and
each non-overlay in \math{{\rm CP}(\R)}
of the form \math{(0,0)} 
is\/ \math 0-shallow parallel closed \wrt\ \RX, 
then
\RX\ is\/ \math 0-shallow confluent.

\item[{\bf(IIa)}]
If\/ \RCONS\ is right-linear
and if
each critical peak in \math{{\rm CP}(\R)} 
of the form 
\math{(0,0)}
is\/ \math 0-shallow noisy strongly joinable \wrt\ \RX, 
and each non-overlay in \math{{\rm CP}(\R)}
of the form \math{(0,0)}
is\/ \math 0-shallow noisy anti-closed \wrt\ \RX, 
then
\RX\ is\/ \math 0-shallow confluent.

\item[{\bf(IIb)}]
If\/ \RCONS\ is right-linear
and if
each critical peak in \math{{\rm CP}(\R)} 
of the form 
\math{(0,0)}
is\/ \math 0-shallow strongly joinable \wrt\ \RX, 
and each non-overlay in \math{{\rm CP}(\R)}
of the form \math{(0,0)}
is\/ \math 0-shallow anti-closed \wrt\ \RX, 
then \redindex{\RX,\omega} is strongly confluent.

\yestop
\yestop
\yestop
\end{enumerate}
\end{theorem}

\begin{corollary}
\label{corollary zero shallow confluent implies confluent no termination}
If\/   \RX\ is\/ \math 0-shallow confluent,
then\/ \redindex{\RX,\omega}\ is confluent.
\end{corollary}

\yestop
\yestop
\yestop
\noindent
We omit the analogue of Theorem~\ref{theorem level parallel closed}
here because it requires that the conditions of the constructor rules 
do not contain any variables. In this case \RCONS\ can 
(in general not effectively)
be transformed into
an unconditional system with identical reduction relation (with possibly
different depths) to which we can then apply 
Theorem~\ref{theorem parallel closed zero} instead.

\yestop
\yestop
\yestop
\noindent
The following is the analogue of Theorem~\ref{theorem complementary}.

\begin{theorem}[Syntactic Confluence Criterion]
\label{theorem complementary zero}
\\
If\/ \RCONS\ is left-linear and normal and all critical peaks of\/ \RCONS\
are complementary, then \redindex{\RX,\omega} is confluent.
\end{theorem}

\mycomment{%%%%%%%%%%%%%%%%%%%%%%%%%%%%%%%%%%%%%%%%%%%%%%%%%%%%%%%%%%%%%%%%%%%%%
\yestop
\begin{definition}[\math 0-Shallow Joinable]%
\label{def shallow joinable no termination zero}\mbox{}
\\
Let\/ \math{\beta\preceq\omega}.
Let\/ \math{s\in\vt}.
A critical peak \criticalpeaklongform\
\\ 
is\emph
{\math0-shallow joinable up to \math\beta\ and \math s 
 \wrt\ \RX\ and \math\lhd\ {[besides \math A]}} 
\udiff\\
\math{
  \forall\varphi\tightin\Xsubst\stopq
  \forall n_0,n_1\prec\omega\stopq
}
\\\LINEmath{
  \inparentheses{
    \inparenthesesoplist{
         \inparentheses{
           n_0\tight+ n_1
           ,\ 
           \hat t\sigma\varphi
         }
         \preclhdeq
         \inparentheses{
           \beta
           ,\ 
           s
         }
       \oplistund
         \forall i\prec 2\stopq
         \inparenthesesoplist{
             1\preceq n_i
           \oplistund
             D_i\sigma\varphi\mbox{ fulfilled \wrt\ }
             \redindex{\RX,n_i\monus 1}
         }
       \oplistund
         \forall
           \inparentheses{\delta,\ s'}
           \tightpreclhd 
           \inparentheses{n_0\tight+ n_1,\ \hat t\sigma\varphi}
         \stopq
           \inparentheses{
             \RX\mbox{ is \math 0-shallow confluent}
              \\\mbox{up to }\delta
                \mbox{ and }s' 
                \mbox{ in }\lhd
           }
       \oplistund
         \forall x\tightin\V\stopq
           x\varphi\tightnotin\DOM{\redindex{\RX,\min\{n_0,n_1\}}}
       \oplistund
         \inparentheses{
             p\tightnotequal\emptyset
           \ \implies\ 
             \forall x\tightin\VAR{\hat t}\stopq
               x\sigma\varphi
               \tightnotin
               \DOM{\redindex{\RX,\min\{n_0,n_1\}}}
         }
       \\
       \multicolumn{2}{@{}l@{}}{
       \left[\begin{array}{@{\wedge\ \ \ }l@{\ \ \ \ }}
          \hat t\sigma\varphi
          \tightnotin
          A(p,\min\{n_0,n_1\})
       \end{array}\right]}
    }
    \\\implies\ 
    \inparentheses{
             t_0\sigma\varphi
             \refltransindex{\RX,n_1}
             \circ
             \antirefltransindex{\RX,n_0}
             t_1\sigma\varphi
    }
  }
.}
\\\headroom
It is 
called\emph{\math0-shallow joinable up to \math\beta\ \wrt\ \RX\ 
and \math\lhd\ {[besides \math A]}}  
\udiff\\
\LINEnomath{
  it is \math0-shallow joinable up to  
  \math\beta\ and \math s \wrt\ \RX\ and \math\lhd\ 
  {[besides \math A]}
  for all \math{s\in\vt}.
}
\\
It is called\emph{\math0-shallow joinable \wrt\ \RX\ and 
\math\lhd\ {[besides \math A]}}  
\udiff\\
\LINEnomath{
  it is \math0-shallow joinable
  up to  \math{\omega} \wrt\ \RX\ and \math\lhd\ 
  {[besides \math A]}.
}
\\
When \math\lhd\ is not specified, we tacitly assume it to be \subterm.
\end{definition}
}%comment%%%%%%%%%%%%%%%%%%%%%%%%%%%%%%%%%%%%%%%%%%%%%%%%%%%%%%%%%%%%%%%%%%%%%%

\yestop
\yestop
\yestop
\noindent
The analogue of theorems
\ref{theoremconfluence} and \ref{theorem quasi-free three} is just
Theorem~\ref{theoremconfluence} with `\R' instantiated with \RCONS.

\yestop
\yestop
\mbox{}

\vfill

\pagebreak

\yestop
\noindent
The following is the analogue of Theorem~\ref{theorem quasi-free}.

\begin{theorem}[Syntactic Test for \math 0-Shallow Confluence]
\label{theorem quasi-free zero}
\\
Let\/ \math{(\tight>,\tight\rhd)} be a termination-pair over \sig/\/\V\@.
\\Assume \math{\RX} to be \math 0-quasi-normal and \RCONS\ to be left-linear.
\\Furthermore, assume that \redindex{\RX,\omega} is terminating:
\\\math{
  \forall\kurzregel\tightin\R\stopq
  \forall\tau\tightin\Xsubst\stopq
  \inparentheses{ 
      \inparenthesesoplist{
          l\tightin\tcs
        \oplistund
          C\tau\mbox{ fulfilled \wrt\ }\redindex{\RX,\omega}
      }
    \ \implies\ 
        l\tau> r\tau
  }   
.}

\noindent
{[For each \math{t\in\tsigX} assume \math{\lll_t} to be a wellfounded
ordering on \TPOS t.
Define (\math{p\tightin\N_+^\ast}, \math{n\tightprec\omega})
\bigmath{
  A(p,n)
  :=
  \setwith
    {t\tightin\DOM{\redindex{\RX,n,q}}}
    {\emptyset\tightnotequal q\lll_t p}
.}]}

\noindent
Now the following two are logically equivalent:
\begin{enumerate}

\notop
\item
Each critical peak in \math{{\rm CP}(\R)} of the form \math{(0,0)}
\\
is\/ 
\math 0-shallow joinable \wrt\ \RX\ and \math\lhd\ {[besides \math A]}.

\noitem
\item
\RX\ is\/ \math 0-shallow confluent.

\yestop
\end{enumerate}
\end{theorem}

\yestop
\noindent
We omit the analogue of Theorem~\ref{theorem level one}
here because it requires that the conditions of the constructor rules 
do not contain any variables. In this case \RCONS\ can be transformed into
an unconditional system with identical reduction relation 
to which we can then apply 
Theorem~\ref{theoremconfluence} with `\R' instantiated with \RCONS.

\yestop
\noindent
The analogue of Theorem~\ref{theorem quasi overlay joinable}
is just Theorem~\ref{theorem quasi overlay joinable} 
with `\R' instantiated with \RCONS.

\mbox{}

\yestop

\noindent{\bf Acknowledgements:} I would like to thank \gramlichname\
for many fruitful discussions and \avenhausname,
Roland Fettig, \madlenername, Birgit Reinert, 
and Andrea Sattler-Klein for some useful hints.
I also would like to thank Thomas Dei\ss\ for providing me with a 
\TeX-version with huge semantic stack size and Paul Taylor
for his support with his diagram typesetting \TeX-package.

\pagebreak

\appendix

\section{Further Lemmas for 
Section~\protect{\ref{sect non-terminating confluence}}}
\label{sect furter lemmas}

\begin{lemma}\label{lemma parallel closed first level two}
\sloppy
Let\/    \R\ be a CRS over \sig/\cons/\/\V\@.
Let\/    \bigmath{\X\tightsubseteq\V.}
\\
Assume\/ \R\ to have conservative constructors,
\RX\ to be quasi-normal,
and the following weak kind of left-linearity:
\\\math{
  \forall\kurzregel\tightin\R\stopq
  \forall p,q\tightin\TPOS l\stopq
  \forall x  \tightin\V    \stopq
}
\\
\LINEmath{
    \inparentheses{
            \inparenthesesoplist{
                l/p\tightequal x\tightequal l/q
              \oplistund
                p\tightnotequal q
            }
            \ \implies\
            \inparenthesesoplist{
                    l\tightin\tcs
                  \oplistund
                    x\tightin\Vcons
            }
    }
.}
\\
Furthermore, assume that
\redindex{\RX,\omega} 
is confluent, 
that
each 
critical peak from \math{{\rm CP}(\R)} 
of the form 
\math{(0,1)}
is \math\omega-shallow {\rm[}noisy\/{\rm]} 
parallel joinable up to \math\omega\ \wrt\ \RX, 
and
that
each  
non-overlay from \math{{\rm CP}(\R)} 
of the form 
\math{(1,0)}
is \math\omega-shallow parallel closed  up to \math\omega\ \wrt\ \RX\@. 

\noindent
Now for each \math{n\prec\omega}:
\bigmath{
  \redparaindex{\RX,\omega+n}
  \tight\circ
  \refltransindex{\RX,\omega[+(n\monus1)]}
}
strongly commutes over \refltransindex{\RX,\omega}.
\\
A fortiori\/
\,
\RX\ is \math\omega-shallow confluent up to \math\omega.
\end{lemma}

\begin{lemma}\label{lemma parallel closed second level two}
\sloppy
Let\/    \R\ be a CRS over \sig/\cons/\/\V\@.
Let\/    \bigmath{\X\tightsubseteq\V.}
\\
Assume\/ \R\ to have conservative constructors,
\RX\ to be quasi-normal,
and the following very weak kind of left-linearity:
\\\math{
  \forall\kurzregel\tightin\R\stopq
  \forall p,q\tightin\TPOS l\stopq
  \forall x  \tightin\V    \stopq
}
\\
\LINEmath{
          \inparentheses{
            \inparenthesesoplist{
                l/p\tightequal x\tightequal l/q
              \oplistund
                p\tightnotequal q
            }
            \implies\
            \inparenthesesoplist{
                 l\tightin\tcs
              \oplistoder
                 x\tightin\Vcons
            }
          }
.}
\\
Furthermore, assume that
for each \math{n\prec\omega}:
\\\LINEnomath{
  \bigmath{
    \redparaindex{\RX,\omega+n}
    \tight\circ
    \refltransindex{\RX,\omega+(n\monus1)}
  }
  strongly commutes over \refltransindex{\RX,\omega}.
}
\\
Moreover, assume 
that each  
critical peak from \math{{\rm CP}(\R)} 
of the form 
\math{(1,1)}
is \math\omega-shallow noisy parallel joinable \wrt\ \RX, 
and
that
each non-overlay from \math{{\rm CP}(\R)} 
of the form 
\math{(1,1)}
is \math\omega-shallow parallel closed \wrt\ \RX\@. 

\noindent
Now for all \math{n_0\preceq n_1\prec\omega}:
\\\LINEnomath{
  \bigmath{
    \refltransindex{\RX,\omega}
    \tight\circ
    \redparaindex{\RX,\omega+n_1}
    \tight\circ
    \refltransindex{\RX,\omega+(n_1\monus1)}
  }
  strongly commutes over \refltransindex{\RX,\omega+n_0}.
}

\noindent
A fortiori\/
\,
\RX\ is \math\omega-shallow confluent.

\end{lemma}

\begin{lemma}\label{lemma parallel closed first level three}
\sloppy
Let\/    \R\ be a CRS over \sig/\cons/\/\V\@.
Let\/    \bigmath{\X\tightsubseteq\V.}
\\
Assume\/ \R\ to have conservative constructors,
\RX\ to be quasi-normal,
and the following very weak kind of left-linearity:
\\\math{
  \forall\kurzregel\tightin\R\stopq
  \forall p,q\tightin\TPOS l\stopq
  \forall x  \tightin\V    \stopq
}
\\\LINEmath{
          \inparentheses{
            \inparenthesesoplist{
               l\tightin\tcs
              \oplistund
                 l/p\tightequal x\tightequal l/q
              \oplistund
                p\tightnotequal q
            }
            \ \implies\
                    x\tightin\Vcons
          }
.}
\\
Furthermore, assume that
\redindex{\RX,\omega} 
is strongly confluent, 
that
each  
critical peak from \math{{\rm CP}(\R)} 
of the form 
\math{(0,1)}
is \math\omega-shallow 
{\rm[}noisy\/{\rm]} 
weak parallel joinable up to \math\omega\ \wrt\ \RX, 
and
that
each 
non-overlay from \math{{\rm CP}(\R)} 
of the form 
\math{(1,0)}
is \math\omega-shallow closed  up to \math\omega\ \wrt\ \RX\@. 

\noindent
Now for each \math{n\prec\omega}:
\\\LINEnomath{
  \bigmath{
    \refltransindex{\RX,\omega}
    \tight\circ
    \redparaindex{\RX,\omega+n}
    \tight\circ
    \refltransindex{\RX,\omega[+(n\monus1)]}}
  strongly commutes over \refltransindex{\RX,\omega}.
}

\noindent
A fortiori\/
\,
\RX\ is \math\omega-shallow confluent up to \math\omega.
\end{lemma}

\pagebreak

\begin{lemma}\label{lemma closed first level two}
\sloppy
Let\/    \R\ be a CRS over \sig/\cons/\/\V\@.
Let\/    \bigmath{\X\tightsubseteq\V.}
\\
Assume\/ \R\ to have conservative constructors,
\RX\ to be quasi-normal,
and
the following weak kinds of left- and right-linearity:
\\\math{
  \forall\kurzregel\tightin\R\stopq
  \forall x  \tightin\V    \stopq
}
\\
\LINEmath{
  \inparenthesesoplist{ 
        \forall p,q\tightin\TPOS l\stopq
          \inparentheses{
            \inparenthesesoplist{
                l/p\tightequal x\tightequal l/q
              \oplistund
                p\tightnotequal q
            }
            \ \implies\
            \inparenthesesoplist{
                    l\tightin\tcs
              \oplistund
                    x\tightin\Vcons
            }
          }
     \oplistund
       \forall p,q\tightin\TPOS r\stopq
          \inparentheses{
            \inparenthesesoplist{
                l\tightin\tcs
                \oplistund
                r/p\tightequal x\tightequal r/q
              \oplistund
                p\tightnotequal q
            }
            \ \implies\
                    x\tightin\Vcons
           }
  }   
.}
\\
Furthermore, assume that
\redindex{\RX,\omega} 
is confluent, 
that
each  
critical peak from \math{{\rm CP}(\R)} 
of the form 
\math{(0,1)}
is \math\omega-shallow {\rm[}noisy\/{\rm]} 
strongly joinable up to \math\omega\ \wrt\ \RX, 
and
that
each 
non-overlay from \math{{\rm CP}(\R)} 
of the form 
\math{(1,0)}
is \math\omega-shallow {\rm[}noisy\/{\rm]} 
anti-closed  up to \math\omega\ \wrt\ \RX\@. 
\\
\noindent
Now for each \math{n\prec\omega}:
\bigmath{
  \redindex{\RX,\omega+n}\tight\circ\refltransindex{\RX,\omega[+(n\monus1)]}}
strongly commutes over \refltransindex{\RX,\omega}.
\\
A fortiori\/
\,
\RX\ is \math\omega-shallow confluent up to \math\omega.
\end{lemma}

\begin{lemma}\label{lemma closed second level two}
\sloppy
Let\/    \R\ be a CRS over \sig/\cons/\/\V\@.
Let\/    \bigmath{\X\tightsubseteq\V.}
\\
Assume\/ \R\ to have conservative constructors,
\RX\ to be quasi-normal,
and the following very weak kind of left-linearity\/ 
\\\math{
  \forall\kurzregel\tightin\R\stopq
        \forall p,q\tightin\TPOS l\stopq
        \forall x  \tightin\V    \stopq
}
\\
\LINEmath{
          \inparentheses{
            \inparenthesesoplist{
                l/p\tightequal x\tightequal l/q
              \oplistund
                p\tightnotequal q
            }
            \ \implies\
            \inparenthesesoplist{
                 l\tightin\tcs
              \oplistoder
                 x\tightin\Vcons
            }
          }
.}
\\
Furthermore, assume that
for each \math{n\prec\omega}:
\\\LINEnomath{
  \bigmath{
    \redindex{\RX,\omega+n}
    \tight\circ
    \refltransindex{\RX,\omega+(n\monus1)}
  }
  strongly commutes over \refltransindex{\RX,\omega}.
}
\\\LINEmath{
      \antirefltransindex{\RX,\omega}
      \circ
      \redparaindex{\RX,\omega+n}
      \tight\circ
      \refltransindex{\RX,\omega+(n\monus1)}
      \ \subseteq\ 
      \refltransindex{\RX,\omega}
      \tight\circ
      \redparaindex{\RX,\omega+n}
      \tight\circ
      \refltransindex{\RX,\omega+(n\monus1)}
      \circ
      \antirefltransindex{\RX,\omega}
.}
\\
Moreover, assume that
that
each  
critical peak from \math{{\rm CP}(\R)} 
of the form 
\math{(1,1)}
is \math\omega-shallow noisy weak parallel joinable \wrt\ \RX, 
and
that
each non-overlay from \math{{\rm CP}(\R)}  
of the form 
\math{(1,1)}
is \math\omega-shallow closed \wrt\ \RX\@. 
\\
\noindent
Now for all \math{n_0\preceq n_1\prec\omega}:
\\\LINEnomath{
  \bigmath{
    \refltransindex{\RX,\omega}
    \tight\circ
    \redparaindex{\RX,\omega+n_1}
    \tight\circ
    \refltransindex{\RX,\omega+(n_1\monus1)}
  }
  strongly commutes over \refltransindex{\RX,\omega+n_0}.
}
\\
\noindent
A fortiori\/
\,
\RX\ is \math\omega-shallow confluent.
\end{lemma}

\begin{lemma}\label{lemma strongly closed second level two}
\sloppy
Let\/    \R\ be a CRS over \sig/\cons/\/\V\@.
Let\/    \bigmath{\X\tightsubseteq\V.}
\\
Assume\/ \R\ to have conservative constructors,
\RX\ to be quasi-normal,
and the following very weak kind of left- and right-linearity:
\\\math{
  \forall\kurzregel\tightin\R\stopq
  \forall p,q\stopq
  \forall x  \tightin\V    \stopq
}
\\
\LINEmath{
          \inparentheses{
            \inparenthesesoplist{
                    l/p\tightequal x\tightequal l/q
                  \oplistoder
                    r/p\tightequal x\tightequal r/q
            }
            \implies\
            \inparenthesesoplist{
                 p\tightequal q
              \oplistoder
                 l\tightin\tcs
              \oplistoder
                 x\tightin\Vcons
            }
          }
.}
\\
Furthermore, assume that
for each \math{n\prec\omega}:
\\\LINEnomath{
  \bigmath{
    \redindex{\RX,\omega+n}
    \tight\circ
    \refltransindex{\RX,\omega+(n\monus1)}
  }
  strongly commutes over \refltransindex{\RX,\omega}.
}
\\
Moreover, assume 
that each  
critical peak from \math{{\rm CP}(\R)} 
of the form 
\math{(1,1)}
is \math\omega-shallow noisy  strongly joinable \wrt\ \RX, 
and
that
each non-overlay from \math{{\rm CP}(\R)} 
of the form 
\math{(1,1)}
is \math\omega-shallow noisy  anti-closed \wrt\ \RX\@. 
\\
\noindent
Now for all \math{n_0\preceq n_1\prec\omega}:
\\\LINEnomath{
  \bigmath{
    \refltransindex{\RX,\omega}
    \tight\circ
    \redindex{\RX,\omega+n_1}
    \tight\circ
    \refltransindex{\RX,\omega+(n_1\monus1)}
  }
  strongly commutes over \refltransindex{\RX,\omega+n_0}.
}
\\
\noindent
A fortiori\/
\,
\RX\ is \math\omega-shallow confluent.
\notop
\end{lemma}

\pagebreak

\yestop
\begin{lemma}
\label{lemma invariance of fulfilledness level}
\\
Let\/ \math{n_0,n_1\prec\omega}.
Let\/ \math{\mu,\nu\in\Xsubst}.
Let\/ \math{\kurzregel\in\R}.
\\
Assume that\/ \bigmath{n_0\tightpreceq n_1}
or that\/ \bigmath{\VAR C\tightsubseteq\Vcons.}
Assume that
\\
\RX\ is \math\omega-level confluent up to\/ \math{n_1}.

\noindent
Now, if\/ 
\math{C\mu} is fulfilled \wrt\ \redindex{\RX,\omega+n_1}
and\/
\bigmath{
  \forall x\tightin\V\stopq 
    x\mu\refltransindex{\RX,\omega+n_0}x\nu
,}
\\
then\/
\math{C\nu} is fulfilled \wrt\ \redindex{\RX,\omega+n_1}
and
\bigmath{l\nu\redindex{\RX,\omega+n_1+1}r\nu.}
\end{lemma}

\begin{lemma}\label{lemma level parallel closed second level two}
\sloppy
Let\/    \R\ be a CRS over \sig/\cons/\/\V\@.
Let\/    \bigmath{\X\tightsubseteq\V.}
\\
Assume\/ 
\bigmath{
  \forall\kurzregel\tightin\R\stopq
    \VAR C\tightsubseteq\Vcons
}
and the following very weak kind of left-linearity:
\\\math{
  \forall\kurzregel\tightin\R\stopq
  \forall p,q\tightin\TPOS l\stopq
  \forall x  \tightin\V    \stopq
}
\\
\LINEmath{
          \inparentheses{
            \inparenthesesoplist{
                l/p\tightequal x\tightequal l/q
              \oplistund
                p\tightnotequal q
            }
            \implies\
            \inparenthesesoplist{
                 l\tightin\tcs
              \oplistoder
                 x\tightin\Vcons
            }
          }
.}
\\
Furthermore, assume%
\footnote{Contrary to analogous lemma for shallow joinability
(\ie\ Lemma~\ref{lemma parallel closed second level two}),
this strong commutation assumption is not really essential
for this lemma if we are confident with the result that
{\normalsize\bigmath{
    \redparaindex  {\RX,\omega+n}
    \tight\circ
    \refltransindex{\RX,\omega}
}}
(instead of 
{\normalsize\math{
    \refltransindex{\RX,\omega}
    \tight\circ
    \redparaindex  {\RX,\omega+n}
    \tight\circ
    \refltransindex{\RX,\omega}
}})
strongly commutes over {\normalsize\refltransindex{\RX,\omega+n}}
(which directly allows to get rid of the application of the 
strong commutation assumption in the proof of Claim~2).
Then it is sufficient to assume that 
\RX\ is \math\omega-shallow confluent up to \math\omega\
(which means that Claim~0 of the proof holds directly),
that
{\normalsize\bigmath{
  \antiredparaindex{\omega}
  \circ
  \redparaindex{\omega+n}
  \nottight{\nottight\subseteq}
  \redparaindex{\omega+n}
  \tight\circ
  \refltransindex{\omega}
  \circ
  \antirefltransindex{\omega+n}
}}
(which replaces the application of the 
strong commutation assumption in the proof of Claim~5),
and that the non-overlays of the form \math{(1,1)} satisfy
{\notop\normalsize\begin{diagram}[height=1.6em,width=2.4em]
&&&&t_1\varphi\headroom
\\
&&\mbox{PC'}
&&\dequal
\\
t_0\varphi&\rredparaindex{\omega+n}&\circ
&\rredparaindex{\omega}&\circ
\end{diagram}}

\notop
\noindent
instead of \math\omega-level parallel closedness
(which allows to replace the application of the strong commutation assumption
at the end of ``The critical peak case'').}
that
for each \math{n\prec\omega}:
\\\LINEnomath{
  \bigmath{
    \redparaindex{\RX,\omega+n}
    \tight\circ
    \refltransindex{\RX,\omega}
  }
  strongly commutes over \refltransindex{\RX,\omega}.
}
\\
Moreover, assume that each critical peak from \math{{\rm CP}(\R)} 
of the form \math{(1,1)}
is \math\omega-level parallel joinable \wrt\ \RX, 
and
that
each non-overlay from \math{{\rm CP}(\R)} 
of the form 
\math{(1,1)}
is \math\omega-level parallel closed \wrt\ \RX\@. 

\noindent
Now for all \math{n\prec\omega}:
\\\LINEnomath{
  \bigmath{
    \refltransindex{\RX,\omega}
    \tight\circ
    \redparaindex  {\RX,\omega+n}
    \tight\circ
    \refltransindex{\RX,\omega}
  }
  strongly commutes over \refltransindex{\RX,\omega+n}.
}

\noindent
A fortiori\/
\,
\RX\ is \math\omega-level confluent.
\end{lemma}

\begin{lemma}\label{lemma level parallel closed second level three}
\sloppy
Let\/    \R\ be a CRS over \sig/\cons/\/\V\@.
Let\/    \bigmath{\X\tightsubseteq\V.}
\\
Assume\/ 
\bigmath{
  \forall\kurzregel\tightin\R\stopq\VAR C\tightsubseteq\Vcons
,}
and the following very weak kind of left-linearity\/ 
\\\math{
  \forall\kurzregel\tightin\R\stopq
        \forall p,q\tightin\TPOS l\stopq
        \forall x  \tightin\V    \stopq
}
\\
\LINEmath{
          \inparentheses{
            \inparenthesesoplist{
                l/p\tightequal x\tightequal l/q
              \oplistund
                p\tightnotequal q
            }
            \ \implies\
            \inparenthesesoplist{
                 l\tightin\tcs
              \oplistoder
                 x\tightin\Vcons
            }
          }
.}
\\
Furthermore, assume that
for each \math{n\prec\omega}:
\\\LINEnomath{
  \bigmath{
    \redindex{\RX,\omega+n}
    \tight\circ
    \refltransindex{\RX,\omega}
  }
  strongly commutes over \refltransindex{\RX,\omega}.
}
\\\LINEmath{
      \antirefltransindex{\RX,\omega}
      \circ
      \redparaindex{\RX,\omega+n}
      \tight\circ
      \refltransindex{\RX,\omega}
      \ \subseteq\ 
      \refltransindex{\RX,\omega}
      \tight\circ
      \redparaindex{\RX,\omega+n}
      \tight\circ
      \refltransindex{\RX,\omega}
      \circ
      \antirefltransindex{\RX,\omega}
.}
\\
\pagebreak
\\
Moreover, assume that
that
each  
\\
critical peak from \math{{\rm CP}(\R)} 
of the form 
\math{(1,1)}
is \math\omega-level weak parallel joinable \wrt\ \RX, 
\\
and
that
each non-overlay from \math{{\rm CP}(\R)}  
of the form 
\math{(1,1)}
is \math\omega-level closed \wrt\ \RX\@. 

\noindent
Now for all \math{n\prec\omega}:
\\\LINEnomath{
  \bigmath{
    \refltransindex{\RX,\omega}
    \tight\circ
    \redparaindex{\RX,\omega+n}
    \tight\circ
    \refltransindex{\RX,\omega}
  }
  strongly commutes over \refltransindex{\RX,\omega+n}.
}

\noindent
A fortiori\/
\,
\RX\ is \math\omega-level confluent.
\end{lemma}

\vfill
\begin{lemma}\label{lemma level strongly closed second level two}
\sloppy
Let\/    \R\ be a CRS over \sig/\cons/\/\V\@.
Let\/    \bigmath{\X\tightsubseteq\V.}
\\
Assume\/
\bigmath{
  \forall\kurzregel\tightin\R\stopq
  \VAR C\tightsubseteq\Vcons
}
and the following very weak kind of left- and right- linearity:
\\\math{
  \forall\kurzregel\tightin\R\stopq
  \forall p,q\stopq
  \forall x  \tightin\V    \stopq
}
\\
\LINEmath{
          \inparentheses{
            \inparenthesesoplist{
                    l/p\tightequal x\tightequal l/q
                  \oplistoder
                    r/p\tightequal x\tightequal r/q
            }
            \implies\
            \inparenthesesoplist{
                 p\tightequal q
              \oplistoder
                 l\tightin\tcs
              \oplistoder
                 x\tightin\Vcons
            }
          }
.}
\\
Furthermore, assume%
\footnote{Contrary to analogous lemma for shallow joinability
(\ie\ Lemma~\ref{lemma strongly closed second level two}),
this strong commutation assumption is not really essential
for this lemma if we are confident with the result that
{\normalsize\bigmath{
    \redindex  {\RX,\omega+n}
    \tight\circ
    \refltransindex{\RX,\omega}
}}
(instead of 
{\normalsize\math{
    \refltransindex{\RX,\omega}
    \tight\circ
    \redindex  {\RX,\omega+n}
    \tight\circ
    \refltransindex{\RX,\omega}
}})
strongly commutes over {\normalsize\refltransindex{\RX,\omega+n}}
(which directly allows to get rid of the application of the 
strong commutation assumption in the proof of Claim~2).
Then it is sufficient to assume that 
\RX\ is \math\omega-shallow confluent up to \math\omega\
(which means that Claim~0 of the proof holds directly),
that
{\normalsize\bigmath{
  \antiredindex{\omega}
  \circ
  \redindex{\omega+n}
  \nottight{\nottight\subseteq}
  \onlyonceindex{\omega+n}
  \tight\circ
  \refltransindex{\omega}
  \circ
  \antirefltransindex{\omega+n}
}}
(which replaces the application of the 
strong commutation assumption in the proof of Claim~5),
that the critical peaks of the form \math{(1,1)} satisfy
{\notop\normalsize\begin{diagram}[height=1.6em,width=2.4em]
&&&&t_1\varphi\headroom
\\
&&\mbox{SJ'}
&&\drefltransindex{\omega+n}
\\
t_0\varphi&\ronlyonceindex{\omega+n}&\circ&\rrefltransindex{\omega}
&\circ
\end{diagram}}

\notop
\noindent
instead of \math\omega-level strong joinability
(which allows to complete ``The second critical peak case''
for the new induction hypothesis),
that the non-overlays of the form \math{(1,1)} satisfy
{\notop\normalsize\begin{diagram}[height=1.6em,width=2.4em]
&&t_1\varphi\headroom
\\
&&\donlyonceindex{\omega+n}
\\
&\mbox{AC'}&\circ
\\
&&\drefltransindex{\omega}
\\
t_0\varphi&\rrefltransindex{\omega+n}&\circ
\end{diagram}}

\notop
\noindent
instead of \math\omega-level anti-closedness
(which allows to complete ``The critical peak case''
for the new induction hypothesis).}
that for each \math{n\prec\omega}:
\\\LINEnomath{
  \bigmath{
    \redindex{\RX,\omega+n}
    \tight\circ
    \refltransindex{\RX,\omega}
  }
  strongly commutes over \refltransindex{\RX,\omega}.
}
\\
Moreover, assume 
that each  
critical peak from \math{{\rm CP}(\R)} 
of the form 
\math{(1,1)}
is \math\omega-level strongly joinable \wrt\ \RX, 
and
that
each non-overlay from \math{{\rm CP}(\R)} 
of the form 
\math{(1,1)}
is \math\omega-level anti-closed \wrt\ \RX\@. 
\\
\noindent
Now for all \math{n\prec\omega}:
\\\LINEnomath{
  \bigmath{
    \refltransindex{\RX,\omega}
    \tight\circ
    \redindex{\RX,\omega+n}
    \tight\circ
    \refltransindex{\RX,\omega}
  }
  strongly commutes over \refltransindex{\RX,\omega+n}.
}
\\
\noindent
A fortiori\/
\,
\RX\ is \math\omega-level confluent.
\vfill
\end{lemma}

\pagebreak

\section{Further Lemmas for Section~\protect{\ref{section now termination}}}

\vfill

\begin{lemma}\label{lemma terminating reduction relation}
Let\/  \R\ be a CRS over \sig/\cons/\/\V\@.
Let\/  \math{\X\tightsubseteq\V}.
Let\/    \math{\alpha\in\{0,\omega\}}.
\\
Let\/ \math{(>,\rhd)}
be a termination-pair over \sig/\/\V\@.
\\
If\/
\bigmath{
  \forall\kurzregel\tightin\R\stopq
  \forall\tau\tightin\Xsubst\stopq
}
\\\LINEmath{
  \inparentheses{ 
      \inparenthesesoplist{
          C\tau\mbox{ fulfilled \wrt\ }\redindex{\RX,\omega+\alpha}
        \oplistund
          \inparentheses{
              \alpha\tightequal 0
            \ \implies
              l\tightin\tcs
          }
      }
    \implies\
        l\tau> r\tau
  }   
,}
\\
then \bigmath{\redindex{\RX,\omega+\alpha}\subseteq\tight\rhd.} 
\end{lemma}

\vfill

\begin{lemma}\label{lemma invariance of fulfilledness compatible}
\\
Let\/ \math{\mu,\nu\in\Xsubst}.
Let\/ \math{\kurzregel\in\R}.
\\
Let\/ \math{(\tight>,\tight\rhd)} be a termination-pair over \sig/\/\V\ 
such that:
\\\math{
  \forall\tau\tightin\Xsubst\stopq
}
\\
\LINEmath{
  \inparentheses{ 
        C\tau\mbox{ fulfilled \wrt\ }\redindex{\RX}
    \ \implies
    \inparenthesesoplist{
        l\tau> r\tau
      \oplistund
        \forall u\in\condterms C\stopq
          \inparenthesesoplist{
              l\tau\rhd u\tau
            \oplistoder 
              u\tau\tightnotin\DOM\redsub
            \\
            \multicolumn{2}{@{}l@{}}{
            \left[\begin{array}{@{\vee\ \ }l@{\ }}
               \VAR u\subseteq\Vcons
            \end{array}\right]}
          }\!\!
    }\!\!
  }   
.}
\\
Assume that 
\bigmath{
  \forall u\tight\lhd l\mu\stopq
    \redsub\mbox{ is confluent below }u
.}
\\
{[Assume that 
\bigmath{
  \antirefltransindex{\RX,\omega}\circ\refltranssub
  \subseteq
  \downarrowsub
.}]}

\noindent
Now, if\/ 
\math{C\mu} is fulfilled \wrt\ \redindex{\RX}
and\/
\bigmath{
  \forall x\tightin\V\stopq 
    x\mu\refltransindex{\RX}x\nu
,}
\\
then\/
\math{C\nu} is fulfilled \wrt\ \redindex{\RX}
and
\bigmath{l\nu\redindex{\RX}r\nu.}
\end{lemma}

\vfill

\begin{lemma}\label{lemma quasi-free three}
\sloppy
\\
Let\/    \R\ be a CRS over \sig/\cons/\/\V\@.
Let\/    \bigmath{\X\tightsubseteq\V.}
\\
Let\/ \math{(\tight>,\tight\rhd)} be a termination-pair over \sig/\/\V\ 
such that:
\\\math{
  \forall\kurzregel\tightin\R\stopq
  \forall\tau\tightin\Xsubst\stopq
}
\\
\LINEmath{
  \inparentheses{ 
        C\tau\mbox{ fulfilled \wrt\ }\redindex{\RX}
    \ \implies
    \inparenthesesoplist{
        l\tau> r\tau
      \oplistund
        \forall u\in\condterms C\stopq
          \inparenthesesoplist{
              l\tau\rhd u\tau
            \oplistoder 
              u\tau\tightnotin\DOM\redsub
            \\
            \multicolumn{2}{@{}l@{}}{
            \left[\begin{array}{@{\vee\ \ }l@{\ }}
               \VAR u\subseteq\Vcons
            \end{array}\right]}
          }\!\!
    }\!\!
  }   
.}
\\
For each \math{t\in\tsigX} assume \math{\lll_t} to be a wellfounded
ordering on \TPOS t.
Define (\math{p\tightin\N_+^\ast})
\bigmath{
  A(p)
  :=
  \setwith
    {t\tightin\DOM{\redindex{\RX,\omega+\omega,q}}}
    {\emptyset\tightnotequal q\lll_t p}
  \nottight{[\nottight\cup \DOM{\redindex{\RX,\omega}}]}
.}
\\
{[Assume that 
\bigmath{
  \antirefltransindex{\RX,\omega}\circ\refltranssub
  \subseteq
  \downarrowsub
.}]}
\\
Assume that each critical peak 
\bigmath{\criticalpeaklongform\in{\rm CP}(\R)}
\\
{[with 
\bigmath{
  \forall k\tightprec2\stopq
  \inparentheses{
      \Lambda_k\tightequal1
    \ \oder\ 
      \condterms{D_k\sigma}\tightnotsubseteq\tcc
  }
}]}
\\
is \math\rhd-weakly joinable \wrt\ \RX\ besides \math A\@.

\noindent
Now:
\redsub\ is confluent.
\end{lemma}

\vfill

\pagebreak

\begin{lemma}\label{lemma quasi-free}
\sloppy\mbox{}
\\
Let\/  \R\ be a CRS over \sig/\cons/\/\V\@.
Let\/  \math{\X\tightsubseteq\V}.
Let\/  \math{\beta\preceq\omega}. \ 
Let\/  \math{\hat s\tightin\vt}.
\\
Assume
the following very weak kind of left-linearity:
\\\math{
  \forall\kurzregel\tightin\R    \stopq
  \forall x        \tightin\Vsig \stopq
  \forall p,q      \tightin\TPOS l\stopq
}
\\\LINEmath{
          \inparentheses{
            \inparenthesesoplist{
                l\tightin\tcs
              \oplistund
                l/p\tightequal x\tightequal l/q
            }
            \ \implies\
                p\tightequal q
          }
.}

\noindent
Furthermore, assume the following compatibility property for 
a termination-pair \math{(>,\rhd)} over \sig/\/\V:
\\\math{
  \forall\kurzregel\tightin\R\stopq
  \forall\tau\tightin\Xsubst\stopq
}
\\
\LINEmath{
  \inparenthesesoplist{ 
    \inparenthesesoplist{
        l\in\tcs
      \oplistund
        C\tau\mbox{ fulfilled \wrt\ }\redindex{\RX,\omega}
    }
    \oplistimplies
    \inparenthesesoplist{
        \kurzregel\mbox{ is quasi-normal \wrt\ \RX}
      \oplistund
        \forall u\in\condterms C\stopq
          \inparenthesesoplist{
              l\tau\rhd u\tau
            \oplistoder 
              u\tau\tightnotin\DOM\redsub
            \oplistoder
               \VAR u\subseteq\Vcons
          }
    }
  }   
}
\\
and
\\\math{
  \forall\kurzregel\tightin\R\stopq
  \forall\tau\tightin\Xsubst\stopq
  \inparentheses{ 
      \inparentheses{
        C\tau\mbox{ fulfilled \wrt\ }\redindex{\RX}
      }
    \ \implies\ 
        l\tau> r\tau
  }   
.}
\\
{[For each \math{t\in\tsigX} assume \math{\lll_t} to be a wellfounded
ordering on \TPOS t.
Define (\math{p\tightin\N_+^\ast}, \math{n\tightprec\omega})
\bigmath{
  A(p,n)
  :=
  \setwith
    {t\tightin\DOM{\redindex{\RX,\omega+n,q}}}
    {\emptyset\tightnotequal q\lll_t p}
.}]}

\noindent
Assume\/ \redindex{\RX,\omega} to be confluent.
\\
Assume that each critical peak 
\bigmath{\criticalpeaklongform\in{\rm CP}(\R)}
with \bigmath{(\Lambda_0,\Lambda_1)\tightnotequal(1,1)}
and 
\inparentheses{
    (\Lambda_0,\Lambda_1)\tightnotequal(0,0)
  \ \oder\ 
    \condterms{D_0\sigma\,D_1\sigma}\tightnotsubseteq\tcc
}
is \math\omega-shallow joinable 
up to \math\beta\ and \math{\hat s} \wrt\ \RX\ and \math\lhd\ 
{[besides \math A]}\@.

\noindent
Now:
\RX\ is \math\omega-shallow confluent 
up to \math\beta\ and \math{\hat s} in \math\lhd.
\end{lemma}

\begin{lemma}\label{lemma quasi-free two and zero}
\sloppy\mbox{}
\\
Let\/    \R\ be a CRS over \sig/\cons/\/\V\@.
Let\/    \bigmath{\X\tightsubseteq\V.}
Let\/    \math{\alpha\in\{0,\omega\}}.
Let\/    \math{\beta\preceq\omega\tight+\alpha}. \ 
Let\/    \bigmath{\hat s\tightin\vt.}
\\
Assume
the following weak kind of left-linearity:
\bigmath{
  \forall\kurzregel\tightin\R    \stopq
  \forall x        \tightin\V    \stopq
  \forall p,q      \tightin\TPOS l\stopq
}
\\\LINEmath{
          \inparentheses{
              \inparenthesesoplist{
                  l/p\tightequal x\tightequal l/q
                \oplistund
                  \inparentheses{ 
                      \alpha\tightequal 0
                    \ \implies\
                      l\tightin\tcs
                  }
                \oplistund
                  \inparentheses{ 
                      \alpha\tightequal\omega
                    \ \implies\
                      x\tightin\Vsig
                  } 
              }
            \ \implies\
                p\tightequal q
          }
.}
\\
Furthermore, assume \RX\ to be \math\alpha-quasi-normal.
\\
Let\/ \math{(\tight>,\tight\rhd)} be a termination-pair over \sig/\/\V\ 
such that the following compatibility property holds:
\bigmath{
  \forall\kurzregel\tightin\R\stopq
  \forall\tau\tightin\Xsubst\stopq
}
\\\LINEmath{
  \inparentheses{ 
      \inparenthesesoplist{
          C\tau\mbox{ fulfilled \wrt\ }\redindex{\RX,\omega+\alpha}
        \oplistund
          \inparentheses{
              \alpha\tightequal 0
            \ \implies
              l\tightin\tcs
          }
      }
    \implies\
        l\tau> r\tau
  }   
}
\\
{[For each \math{t\in\tsigX} assume \math{\lll_t} to be a wellfounded
ordering on \TPOS t.
Define (\math{p\tightin\N_+^\ast}, \math{n\tightprec\omega})
\bigmath{
  A(p,n)
  :=
  \setwith
    {t\tightin\DOM{\redindex{\RX,\alpha+n,q}}}
    {\emptyset\tightnotequal q\lll_t p}
.}]}

\noindent
Assume\/ \RX\ to be \math\alpha-shallow confluent up to \math{\alpha}.
\\
Assume that each critical peak 
\bigmath{\criticalpeaklongform\in{\rm CP}(\R)}
\\
with 
\bigmath{
  \forall k\tightprec2\stopq
     \inparenthesesoplist{
         \inparentheses{
             \alpha\tightequal 0
           \ \implies\ 
             \Lambda_k\tightequal 0
         }
       \oplistund 
         \inparentheses{
             \alpha\tightequal\omega
           \ \implies\ 
             \inparentheses{
                 \Lambda_k\tightequal1
               \ \oder\ 
                 \condterms{D_k\sigma}\tightnotsubseteq\tcc
              }
         }
     }
}
\\
is \math\alpha-shallow joinable up to \math\beta\ and \math{\hat s}
\wrt\ \RX\ and \math\lhd\ {[besides \math A]}\@.

\noindent
Now:
\RX\ is \math\alpha-shallow confluent 
up to \math\beta\ and \math{\hat s} in \math\lhd.
\end{lemma}

\pagebreak

\yestop
\begin{lemma}\label{lemma level one}
\sloppy\mbox{}
\\
Let\/    \R\ be a CRS over \sig/\cons/\/\V\@.
Let\/    \bigmath{\X\tightsubseteq\V.}
Let\/ \ \mbox{\math{\beta\preceq\omega}.} \ 
Let\/    \bigmath{\hat s\tightin\vt.}
\\
Assume
\bigmath{
  \forall\kurzregel\tightin\R\stopq
      \VAR C\tightsubseteq\Vcons
.}
\\
Let\/ \math{(\tight>,\tight\rhd)} be a termination-pair over \sig/\/\V\ 
such that the following compatibility property holds:
\\\mbox{}\hfill\math{
  \forall\kurzregel\tightin\R\stopq
  \forall\tau\tightin\Xsubst\stopq
  \inparentheses{ 
    \inparentheses{
      C\tau\mbox{ fulfilled \wrt\ }\redsub
    }
    \implies\
      l\tau> r\tau
  }   
.}
\\
{[For each \math{t\in\tsigX} assume \math{\lll_t} to be a wellfounded
ordering on \TPOS t.
Define (\math{p\tightin\N_+^\ast}, \math{n\tightprec\omega})
\bigmath{
  A(p,n)
  :=
  \setwith
    {t\tightin\DOM{\redindex{\RX,\omega+n,q}}}
    {\emptyset\tightnotequal q\lll_t p}
.}]}

\noindent
Assume\/ \redindex{\RX,\omega} to be confluent.
Assume that each critical peak in \math{{\rm CP}(\R)}
of the forms \math{(0,1)},  \math{(1,0)}, or \math{(1,1)}  
is \math\omega-level joinable up to \math\beta\ and \math{\hat s} 
\wrt\ \RX\ and \math\lhd\ {[besides \math A]}\@.
\\
Now:
\RX\ is \math\omega-level confluent 
up to \math\beta\ and \math{\hat s} in \math\lhd.
\end{lemma}

\vfill

\yestop
\noindent
The following lemma generalizes Lemma~7.6 of \citewgjsc\
by requiring \math\rightrightarrows\ to be terminating
only below a restricted set of terms {\rm T}:

\begin{lemma}\label{lemmaa}
\\
Let\/ \bigmath{{\rm T}\subseteq\vt.}
Let\/ \bigmath{\supertermeq{[{\rm T}]}} 
denote the set of subterms of\/ {\rm T}\@.
Let\/ \math\rightrightarrows\ be a sort-invariant
\linebreak
(This can always be achieved by identifying all sorts.)
and\/ {\rm T}-monotonic relation on \vt\@.
Define
\bigmath{
    \tight\rhd
    \nottight{\nottight{\nottight{:=}}}
    \tight{\domres\id{\supertermeq{[{\rm T}]}}}
    \nottight\circ
    \transclosureinline{(\rightrightarrows\cup\superterm)}
.}
Now:
\begin{enumerate}

\noitem
\item\label{lemmaa iteme}%
\bigmath{
  \begin{array}[t]{@{}r@{\nottight{\nottight{\nottight{\nottight=}}}}cl}
    \tight{\domres\id{\supertermeq{[{\rm T}]}}}
    \nottight\circ
    \tight\rightrightarrows
    &
    \tight{\domres\id{\supertermeq{[{\rm T}]}}}
    \nottight\circ
    \tight\rightrightarrows
    \nottight\circ
    \tight{\domres\id{\supertermeq{[{\rm T}]}}}
    &;
    \\
    \tight{\domres\id{{\rm T}}}
    \nottight\circ
    \tight\rightrightarrows
    &
    \tight{\domres\id{{\rm T}}}
    \nottight\circ
    \tight\rightrightarrows
    \nottight\circ
    \tight{\domres\id{{\rm T}}}
    &.
    \\
  \end{array}
}

\item\label{lemmaa itema}%
\bigmath{ 
  \domres\id{{\rm T}}
  \nottight\circ
  \superterm
  \nottight\circ
  \tight\rightrightarrows
  \nottight{\nottight{\nottight{\subseteq}}}
  \domres\id{{\rm T}}
  \nottight\circ
  \tight\rightrightarrows
  \nottight\circ
  \domres\id{{\rm T}}
  \nottight\circ
  \superterm
.}
\\\mbox{}\ 
Moreover, for\/ \math{{\rm T}\tightequal\vt}: 
$\ \ \superterm\circ\rightrightarrows
 \ \ \subseteq\ \ \rightrightarrows\circ\superterm$.

\item\label{lemmaa itemb}%
\bigmath{
  \begin{array}[t]{@{}r@{}c@{}ll}
    \rhd
    &\nottight{\nottight{\nottight\subseteq}}
    &\tight\subtermeq
    \nottight\circ
    \tight{\domres\id{{\rm T}}}
    \nottight\circ
    \transclosureinline{(\rightrightarrows\cup\superterm)}
    &;
    \\
    \rhd
    &\nottight{\nottight{\nottight=}}
    &\transclosureinline{
       \inparentheses{
         \inparenthesesinlinetight{
            \tight{\domres\id{\supertermeq{[{\rm T}]}}}
            \nottight\circ
            \rightrightarrows
         }
         \nottight\cup
         \inparenthesesinlinetight{
            \tight{\domres\id{\supertermeq{[{\rm T}]}}}
            \nottight\circ
            \tight\superterm
         }
       }
     }
    \nottight\circ
    \tight{\domres\id{\supertermeq{[{\rm T}]}}}
    &;
    \\
    \tight{\domres\id{{\rm T}}}
    \nottight\circ
    \transclosureinline{(\rightrightarrows\cup\superterm)}
    &\nottight{\nottight{\nottight=}}
    &\inparentheses{
      \tight{\domres\id{\rm T}}
      \nottight\circ
      \tight\superterm
    }
    \nottight{\nottight\cup}
    \inparentheses{
      \transclosureinline{
        \inparenthesesinline{
          \tight{\domres\id{\rm T}}
          \nottight\circ
          \tight\rightrightarrows
        }
      }
      \nottight\circ
      \tight{\domres\id{{\rm T}}}
      \nottight\circ
      \tight\supertermeq
    }
    &.
    \\
  \end{array}
}
\\\mbox{}\ 
Moreover, for\/ \math{{\rm T}\tightequal\vt}: 
\bigmath{
  \nottight{
    \tight\rhd
    \nottight{\nottight{\nottight=}}
    \superterm
    \nottight\cup
    (\transclosureinline\rightrightarrows\circ\supertermeq)
  }
.}

\item\label{lemmaa itemc}\headroom\label{item does not hold in general if}%
If\/ $\rightrightarrows$ is terminating (below all \math{t\in{\rm T}}) 
{[and\/ \math\rightrightarrows\ and\/ {\rm T} are \stable]}, 
then \math\rhd\ is a wellfounded 
[and \stable] 
ordering on \bigmath{\supertermeq{[{\rm T}]}}
(which does not need to be sort-invariant or\/ {\rm T}-monotonic).

\item\label{lemmaa itemd}%
(\ref{item does not hold in general if})
\sloppy 
does not hold in general if one of the two conditions
\mbox{``\/\math\rightrightarrows\ sort-invariant''}
or  
\mbox{``\/\math\rightrightarrows\ {\rm T}-monotonic''}
is removed.
Moreover, 
(\ref{item does not hold in general if})
does not hold in general for 
\transclosureinline{
  \inparenthesesinlinetight{
    \rightrightarrows
    \cup
    \superterm
  }
}
instead of\/ \math\rhd.

\yestop
\end{enumerate}
\end{lemma}

\vfill

\yestop
\noindent
The proof of the following lemma and its far more restrictive predecessors
has an interesting history. 
After its first occurrence in \citeder\ 
for overlay joinable positive conditional systems, 
in our proof for quasi overlay joinable positive/negative-conditional systems
in \citewgjsc\ we changed the third component of the induction ordering
from \transsub\ to \math\succ, the ordering of the ordinals. 
This change was done 
because it allowed us to check for generalizations more easily
but did not result in a stronger criterion at first.
Later, however, this change of the induction ordering turned out to 
be essential for Theorem~21 of \citegramlichelkuz\ saying that an 
innermost terminating overlay joinable positive conditional rule system
is terminating and confluent:
Due to the mutual dependency of the termination and the
confluence proof, when proving confluence
it was not possible to assume global termination
but local termination only. 
And it was especially impossible to assume 
termination for that part of \redsub\ which was necessary for 
the third component of the induction ordering. 
The following lemma (just like Theorem~7 of \citegramlichelkuz)
requires local termination instead of global termination, which is not
really necessary for proving Theorem~\ref{theorem quasi overlay joinable} but
again allows us to check for future generalizations more easily.
Moreover, note that the form of the proof has been considerably improved
compared to any previous publication: Claim~0 of the proof
does not only provide us
with the new irreducibility assumptions we have included into the
notion of \math\rhd-quasi overlay joinability but also subsumes the whole
second case of the global case distinction of the proof
(as presented in \citedershowitz\ as well as presented in \citewgjsc).
As a consequence, in the whole new proof now the
second and the third component of the induction ordering are used only once.

\begin{lemma}[Syntactic Confluence Criterion]
\label{lemma for theorem quasi overlay joinable}\mbox{}\\
Let\/ \R\ be a CRS over \sig/\cons/\/\V\ and \math{\X\tightsubseteq\V}.
Let\/ \math{\hat s\in\tsigX}.
Define 
\bigmath{
  {\rm T}
  :=
  \refltranssub{[\{\hat s\}]}
.}
\\
Assume either that\/ 
\bigmath{
  \domres\redsub{{\rm T}}
}
is terminating 
and \bigmath{\tight\rhd=\superterm}
\\
or that
%\/\footnote{Note, however, that 
%\bigmath{
%    \domres\redsub{{\rm T}}
%    {\nottight\subseteq}
%    \math\rhd
%}
%instead of
%\bigmath{
%    \domres\redsub{\trianglerighteq{[{\rm T}]}}
%    {\nottight\subseteq}
%    \math\rhd
%}
%is not sufficient!}
\bigmath{
    \domres\redsub{\trianglerighteq{[{\rm T}]}}
    {\nottight\subseteq}
    \math\rhd
,}
\bigmath{
    \superterm
    {\nottight\subseteq}
    \math\rhd
,}
and \tight\rhd\ is a wellfounded ordering on \vt.
\\
Now, if 
all critical peaks in ${\rm CP}(\R)$ are
\tight\rhd-quasi overlay joinable \wrt\ \RX,
\\
then 
\bigmath{\domres\redsub{\trianglerighteq{[{\rm T}]}}} is confluent.
\end{lemma}

\vfill

\section{\math\omega-Coarse Level Joinability}

Using the following notions for \math\omega-coarse level joinability one
can work out a whole analogue of Theorem~\ref{theorem level parallel closed}.
We did not do so because this analogue does not allow of a corollary theorem
analogous to Theorem~\ref{theorem weakly complementary} because the information
on confluence provided by the joinability notion for testing the conditions 
of critical peaks is to poor for practically applicable reasoning.
To those who are interested in this notion, however, we present here
the analogues of Definition~\ref{def level parallel closed},
Definition~\ref{def level parallel joinable}, 
Lemma~\ref{lemma invariance of fulfilledness level},
and
Lemma~\ref{lemma level parallel closed second level two},
for which we also have included the proofs.

\begin{definition}[\math\omega-Coarse Level Parallel Closed]%
\label{def coarse level parallel closed}
\\
A critical peak \newcriticalpeak\
\\ 
is\emph
{\math\omega-coarse level parallel closed \wrt\ \RX} 
\udiff
\\\LINEmath{
  \forall\varphi\tightin\Xsubst\stopq
  \inparenthesesoplist{
    \inparenthesesoplist{
         \forall i\prec 2\stopq
             D_i\varphi\mbox{ fulfilled \wrt\ }
             \redindex{\RX}
       \oplistund
         \redsub\mbox{ and }\redindex{\RX,\omega}\mbox{ are commuting}
    }
    \oplistimplies 
             t_0\varphi
             \redparaindex{\RX}
             \tight\circ
             \refltransindex{\RX,\omega}
             \circ
             \antirefltransindex{\RX,\omega}
             t_1\varphi
  }
.}
\end{definition}

\begin{definition}[\math\omega-Coarse Level Parallel Joinable]%
\label{def coarse level parallel joinable}\mbox{}
\\
A critical peak \newcriticalpeak\
\\ 
is\emph
{\math\omega-coarse level parallel joinable \wrt\ \RX} 
\udiff
\\\LINEmath{
  \forall\varphi\tightin\Xsubst\stopq
  \inparenthesesoplist{
    \inparenthesesoplist{
         \forall i\prec 2\stopq
             D_i\varphi\mbox{ fulfilled \wrt\ }
             \redindex{\RX}
       \oplistund
         \redsub\mbox{ and }\redindex{\RX,\omega}\mbox{ are commuting}
    }
    \oplistimplies 
      t_0\varphi
      \redparaindex{\RX}
      \circ
      \refltransindex    {\RX,\omega}
      \circ
      \antirefltransindex{\RX}
      t_1\varphi
  }
.}
\end{definition}

\begin{lemma}\label{lemma level parallel closed second coarse level two}
\sloppy
Let\/    \R\ be a CRS over \sig/\cons/\/\V\@.
Let\/    \bigmath{\X\tightsubseteq\V.}
\\
Assume\/ 
\bigmath{
  \forall\kurzregel\tightin\R\stopq
    \VAR C\tightsubseteq\Vcons
}
and the following very weak kind of left-linearity:
\\\math{
  \forall\kurzregel\tightin\R\stopq
  \forall p,q\tightin\TPOS l\stopq
  \forall x  \tightin\V    \stopq
}
\\
\LINEmath{
          \inparentheses{
            \inparenthesesoplist{
                l/p\tightequal x\tightequal l/q
              \oplistund
                p\tightnotequal q
            }
            \implies\
            \inparenthesesoplist{
                 l\tightin\tcs
              \oplistoder
                 x\tightin\Vcons
            }
          }
.}
\\
Furthermore, assume that
\redindex{\RX,\omega} is confluent and that
\\\LINEnomath{
  \bigmath{
    \redparaindex{\RX}
    \tight\circ
    \refltransindex{\RX,\omega}
  }
  strongly commutes over \refltransindex{\RX,\omega}.
}
\\
Moreover, assume 
that each  
critical peak from \math{{\rm CP}(\R)} 
of the form 
\math{(1,1)}
is \math\omega-coarse level parallel joinable \wrt\ \RX, 
and
that
each non-overlay from \math{{\rm CP}(\R)} 
of the form 
\math{(1,1)}
is \math\omega-coarse level parallel closed \wrt\ \RX\@. 

\noindent
Now:
\\\LINEnomath{
  \bigmath{
    \refltransindex{\RX,\omega}
    \tight\circ
    \redparaindex  {\RX}
    \tight\circ
    \refltransindex{\RX,\omega}
  }
  strongly commutes over \refltransindex{\RX}.
}

\noindent
A fortiori\/
\,
\redsub\ is confluent.

\end{lemma}

\begin{lemma}
\label{lemma invariance of fulfilledness coarse level}
\\
Let\/ \math{\mu,\nu\in\Xsubst}.
Let\/ \math{\kurzregel\in\R}.
\\
Assume that\/ 
\bigmath{\VAR C\tightsubseteq\Vcons.}
\\
Assume\/ 
\bigmath{
  \antirefltranssub\circ\refltransindex{\RX,\omega}
  \subseteq
  \downarrowsub
.}

\noindent
Now, if\/ 
\math{C\mu} is fulfilled \wrt\ \redsub\
and\/
\bigmath{
  \forall x\tightin\V\stopq 
    x\mu\refltranssub x\nu
,}
\\
then\/
\math{C\nu} is fulfilled \wrt\ \redsub\
and
\bigmath{l\nu\redsub r\nu.}
\end{lemma}

\raggedbottom
\section{The Proofs}
\label{sect proofs}

\begin{proof}{Lemma~\ref{lemma commutation copy}}
\headroom
Assume 
\redindex0 and \redindex1 to be locally commuting.
\\\headroom
For the first claim we assume that \bigmath{\redindex0\cup\redindex1} 
is terminating.
We show commutation by induction over the wellfounded ordering
\bigmath{\transclosureinline{\redindex0\cup\redindex1}.}
Suppose 
\bigmath{
  t_0'
  \antirefltransindex0 
  s
  \refltransindex1
  t_1'
.}
We have to show
\bigmath{
  t_0'
  \refltransindex1
  \circ
  \antirefltransindex0 
  t_1'
.}
In case there is some \math{i\prec2} with
\bigmath{t_i'\tightequal s}
the proof is finished due to
\bigmath{
  t_i'\tightequal s\refltransindex{1-i}t_{1-i}'\antirefltransindex{i}t_{1-i}'
.}
Otherwise
\bigmath{
  t_0'
  \antirefltransindex0 
  t_0
  \antiredindex0 
  s
  \redindex1 
  t_1
  \refltransindex1
  t_1'
}
for some \math{t_0}, \math{t_1}
(\cf\ diagram below).
By local commutation there is some \math{s'} with
\bigmath{
  t_0
  \refltransindex1 
  s'
  \antirefltransindex0 
  t_1
.}
Due to \bigmath{s\;\transclosureinline{\redindex0\cup\redindex1}\;t_0,}
by induction hypothesis we get some \math{s''} with
\bigmath{
  t_0'
  \refltransindex1
  s''
  \antirefltransindex0
  s'
.}
Due to \bigmath{s\;\transclosureinline{\redindex0\cup\redindex1}\;t_1,}
by induction hypothesis we get 
\bigmath{
  s''
  \refltransindex1
  \circ
  \antirefltransindex0
  t_1'
.}
\begin{diagram}
s&\rredindex1&t_1&\rrefltransindex1&t_1'
\\\dredindex0&&\drefltransindex0&&
\\t_0&\rrefltransindex1&s'&&\drefltransindex0
\\\drefltransindex0&&\drefltransindex0&&
\\t_0'&\rrefltransindex1&s''&\rrefltransindex1&\circ
\end{diagram}
\noindent
For the second claim we now assume that \redindex0 or \redindex1 is transitive.
\Wrog\ (due to symmetry in \math0 and \math1) say \redindex0 is transitive.
It is sufficient to show
\\\linemath{
  \forall n\tightin\N\stopq
  \forall s,t_0,t_1\stopq
    (
     t_0\antirefltransindex0 s\redindexn n 1 t_1
     \ \implies\ 
     t_0\refltransindex1\circ\antirefltransindex0 t_1
    )
.}
\underline{\math{n\tightequal0}:}
\bigmath{
  t_0\refltransindex1 t_0\antirefltransindex0 s\tightequal t_1
.}
\\\underline{\math{n\implies(n\tight+1)}:}
Assume 
\bigmath{
  t_0\antirefltransindex0 s\redindexn n 1 t'\redindex1 t_1
}
(\cf\ diagram below).
By induction hypothesis there is some \math w with 
\bigmath{
  t_0\refltransindex1 w\antirefltransindex0 t'
.}
In case of \bigmath{w\tightequal t'} the proof is finished by
\bigmath{
  t_0\refltransindex1 w\tightequal t'\redindex1 t_1\antirefltransindex0 t_1
.}
Otherwise, since \redindex0 is transitive, we have
\bigmath{
  w\antiredindex0 t'\redindex1 t_1
.}
By the local commutation of \redindex0 and \redindex1 this implies
\bigmath{
%  t_0\refltransindex1 
  w\refltransindex1\circ\antirefltransindex0 t_0
.}
\begin{diagram}
s&\rredindexn{n}1&t'&\rredindex1&t_1
\\\drefltransindex0&&\drefltransindex0&&\drefltransindex0
\\t_0&\rrefltransindex1&w&\rrefltransindex1&\circ
\end{diagram}
\end{proof}

\pagebreak

\begin{proof}{Lemma~\ref{lemma strong commutation one copy}}
That (3) (or else (2)) implies (1) is trivial. For (1) implying (2) and (3)
it is sufficient to show under the assumption of (1) that
\\\LINEmath{\footroom
  \forall n\tightin\N\stopq
  \forall s,t_0,t_1\stopq
  (
  t_0\antiredindexn n 0 s\redindex1 t_1
  \ \implies\ 
  t_0\onlyonceindex1\circ\antirefltransindex0 t_1
  )
.}
\\\underline{\math{n\tightequal0}:}
\bigmath{t_0\tightequal s\redindex1 t_1\antirefltransindex0 t_1.}
\\\underline{\math{n\implies(n\tight+1)}:}
Suppose
\bigmath{
  t_0\antiredindex0 t'\antiredindexn n 0 s\redindex1 t_1
}
(\cf\ diagram below).
By induction hypothesis there is some \math w with
\bigmath{
  t'\onlyonceindex1 w\antirefltransindex0 t_1
.}
In case of \bigmath{t'\tightequal w} the proof is finished due to
\bigmath{
  t_0\onlyonceindex1 t_0\antiredindex0 t'\tightequal w\antirefltransindex0 t_1.
}
Otherwise we have
\bigmath{t_0\antiredindex0 t'\redindex1 w}
and get by the assumed strong commutation
\bigmath{
  t_0
  \onlyonceindex1
  \circ
  \antirefltransindex0 
  w
%  \antirefltransindex0 
%  t_1
.}
\begin{diagram}
s&\rredindex1&t_1
\\\dredindexn{n}0&&\drefltransindex0
\\t'&\ronlyonceindex1&w
\\\dredindex0&&\drefltransindex0
\\t_0&\ronlyonceindex1&\circ
\end{diagram}

\noindent
For proving the final implication of the lemma,
we may assume that \redindex1 strongly commutes over
\transindex0.
A fortiori \transindex0 and \redindex1 are locally commuting. 
By Lemma~\ref{lemma commutation copy} they are commuting. Therefore
\redindex0 and \redindex1 are commuting, too.
\end{proof}

\begin{proof}{Lemma~\ref{lemma church rosser}}
It is trivial to show 
\bigmath{
  \forall n\tightin\N\stopq
    \symindexn n{}\subseteq \tight\downarrow
}
by induction on \math n.
\end{proof}

\pagebreak

\begin{proof}{Lemma~\ref{lemma necessary}}
Just like the proof of Lemma~\ref{lemma omega shallow joinablility necessary}
when the depth considerations are omitted.
\end{proof}

\begin{proof}{Lemma~\ref{lemma omega shallow joinablility necessary}}
For \math{\newcriticalpeak\tightin{\rm CP}(\R)}
there are two rules \sugarregelindex0 and \sugarregelindex1 in \R\
(assuming 
\bigmath{\VAR{\sugarregelindex0}\cap\VAR{\sugarregelindex1}=\emptyset} 
\wrog)
and 
\bigmath{\sigma\in\Vsubst}
with
\bigmath{l_{0}\sigma=l_{1}\sigma/p;}
\bigmath{
 \ 
 (t_0,\ D_0,\ t_1,\ D_1,\ \hat t)
 =
 (\repl{l_1}p{r_0},\ 
  C_0,\ 
  r_1,\ 
  C_1,\ 
  l_1
  )
 \sigma
}
and
\bigmath{
  \Lambda_i
  =
  \left\{\arr{{ll}0&\mbox{ if }l_i\in\tcs\\1&\mbox{ otherwise}\\}\right\}.
}
Let \bigmath{\varphi\tightin\Xsubst;}
\bigmath{n_0,n_1\prec\omega;} 
and assume
[\math{
 (n_0\plusalpha n_1
 ,\ 
  \hat t\varphi
 )
 \preclhdeq
 (\beta,\ s)
\ 
} 
and]\/
for all \math{i\prec  2}:
\bigmath{
  \inparenthesesinline{
      \alpha\tightequal 0    \implies\Lambda_i\tightequal 0\tightprec n_i
  }
;}
\bigmath{
  \inparenthesesinline{
      \alpha\tightequal\omega\implies\Lambda_i\tightpreceq n_i
  }
;}
\math{D_i      \varphi} fulfilled \wrt\ \redindex{\RX,\alpha+(n_i\monus 1)};
\ie\
\math{C_i\sigma\varphi} fulfilled \wrt\ \redindex{\RX,\alpha+(n_i\monus 1)}.
In case of \bigmath{n_i\tightequal 0} we have 
\bigmath{\Lambda_i\tightequal 0}
and
\bigmath{\alpha\tightequal\omega}
and therefore by Corollary~\ref{corollary redsubomega is minimum}
\bigmath{l_i\sigma\varphi\redindex{\RX,\alpha+n_i}r_i\sigma\varphi.}
In case of \bigmath{n_i\tightsucc 0} we have 
\bigmath{n_i\tightequal(n_i\monus 1)\tight+1}
and therefore 
\bigmath{l_i\sigma\varphi\redindex{\RX,\alpha+n_i}r_i\sigma\varphi}
again
due to \bigmath{\alpha\tightequal 0\implies\Lambda_i\tightequal 0.}
Then
\\\linemath{
  t_{0}\varphi
  =
  \repl{l_{1}\sigma\varphi}
       {p}
       {r_{0}\sigma\varphi}
  \antiredindex{\RX,\alpha+n_0}
  l_{1}\sigma\varphi
  \redindex    {\RX,\alpha+n_1}
  r_{1}\sigma\varphi
  =
  t_{1}\varphi
.}
By \math\alpha-shallow confluence 
{[up to \math\beta\ [and \math s in \math\lhd]]}
we have
\bigmath{
  t_0\varphi
  \refltransindex    {\RX,\alpha+n_1}
  \circ
  \antirefltransindex{\RX,\alpha+n_0}
  t_1\varphi
}\@.
\end{proof}

\begin{proof}{Lemma~\ref{lemma omega level joinablility necessary}}
The proof is analogous to the proof of 
Lemma~\ref{lemma omega shallow joinablility necessary}.
\end{proof}

\begin{proof}{Lemma~\ref{lemma quasi overlay joinable}}
In case of
\bigmath{
  (\hat t/p'       )\sigma\varphi
  \tightequal
  (\hat t/\emptyset)\sigma\varphi
}
we get \bigmath{p'\tightequal\emptyset.}
Thus
\bigmath{
  \Delta\subseteq\TPOS{\hat t}\tightsetminus\{\emptyset\}
}
together with
\bigmath{
              \forall p'\tightin\Delta\stopq
                (\hat t/p'       )\sigma\varphi
                \tightequal
                (\hat t/\emptyset)\sigma\varphi
}
implies 
\bigmath{\Delta\tightequal\emptyset.}
If there is some \math{\bar u_1}
with
\bigmath{
  t_0\sigma\mu
  \refltrans
  \bar u_1
  \penalty-1
  \antirefltrans
  t_1\sigma\mu
;}
define
\math{\bar n:=1};
\math{\bar u_0:=t_1\sigma\mu};
\math{\bar p_0:=\emptyset};
and note that 
\bigmath{
  t_1\sigma\varphi
  \antired
  \hat t\sigma\varphi
}
when
\math{D_1\sigma\varphi} is fulfilled.
\end{proof}

\begin{proof}{Lemma~\ref{lemma invariance of fulfilledness two}}
If \R\ has conservative constructors we get 
\bigmath{\VAR C\tightsubseteq\Vcons}
(since \math{l\tightin\tcs}).
If \bigmath{\VAR C\subseteq\Vcons,}
\\
then 
\bigmath{
  \condterms{C\mu}\tightsubseteq\tcs
}
(since 
\math{l\tightin\tcs}).
\\
Thus we can always assume
\bigmath{
  \condterms{C\mu}\tightsubseteq\tcs
.}
Then we have 
\bigmath{
  \forall x\tightin\VAR C\stopq 
    x\mu\tightin\tcs
}
and thus
\bigmath{
  \forall x\tightin\VAR C\stopq 
    x\mu\refltransindex{\RX,\omega} x\nu
}
by \lemmaconskeeping.
Moreover \math{C\mu} is fulfilled \wrt\ \redindex{\RX,\omega}
by \lemmaconskeeping.
By confluence of \redindex{\RX,\omega}
and \lemmaconskeeping\ 
\math{C\nu} is fulfilled \wrt\ \redindex{\RX,\omega}.
By Corollary~\ref{corollary redsubomega is minimum}
we finally get \bigmath{l\nu\redindex{\RX,\omega}r\nu.}
\end{proof}

\pagebreak

\begin{proofparsepqed}
{Theorem~\ref{theorem complementary}
and Theorem~\ref{theorem weakly complementary}}
Due to Corollary~\ref
{corollary omega shallow confluent implies confluent no termination},
it suffices to show that the conditions of 
Theorem~\ref{theorem parallel closed}(I) 
or else (in case of Theorem~\ref{theorem weakly complementary})
Theorem~\ref{theorem level parallel closed}(I) 
are satisfied.
The only non-trivial part are the joinability requirements for the 
critical pairs. We just have to show that the conjunctive
condition lists of the
joinability notions are never satisfied.
Assume \newcriticalpeak\ to be a critical peak.

We first treat the critical peaks of the form 
\math{(0,1)} or \math{(1,0)}, 
and, in case of Theorem~\ref{theorem complementary}, 
also of the form \math{(1,1)}.
For these we have to show \math\omega-shallow parallel joinability or else
\math\omega-shallow parallel closedness.
Thus, assume 
\math{\varphi\in\Xsubst} and
\math{n_0,n_1\prec\omega} such
that
\bigmath{
  \forall i\tightprec 2\stopq
    \inparenthesesinline{
         D_i\varphi\mbox{ fulfilled \wrt\ } \redindex{\RX,\omega+(n_i\monus 1)}
    }
}
and
\bigmath{
    \forall\delta\tightprec n_0\plusomega n_1\stopq
    \inparenthesesinline{
           \RX\mbox{ is \math\omega-shallow confluent up to }\delta
    }
.}
By the assumed complementarity there must be complementary 
equation literals in \math{D_0} and \math{D_1}. Due to our symmetry in
\math 0 and \math 1 so far, we may \wrog\ assume that 
\math{(u\boldequal   v)} occurs in \math{D_0} and 
\math{(u\boldunequal v)} occurs in \math{D_1}
or else that 
\math{(p\boldequal\truepp )} occurs in \math{D_0} and 
\math{(p\boldequal\falsepp)} occurs in \math{D_1}.
We treat the first case first.
Then there are \math{\hat u,\hat v\in\tgcons} with
\bigmath{
  \hat u
  \antirefltransindex{\omega+(n_1\monus 1)}
  u\varphi
  \downarrowindex{\omega+(n_0\monus 1)}
  v\varphi
  \refltransindex    {\omega+(n_1\monus 1)}
  \hat v
}
and
\bigmath{
  \hat u
  \notconfluindex    {\omega}
  \hat v
.}
In case of \bigmath{n_0,n_1\tightpreceq 1} this contradicts the required
confluence of \redindex{\omega}, \cf\ Lemma~\ref{lemma church rosser}.
Otherwise, in case of \bigmath{n_0\tightsucceq 1} we have 
\bigmath{
  (n_0\monus 1)\plusomega(n_1\monus 1)
  \prec
  n_0\plusomega n_1
}
and thus by our above assumption
\RX\ is \math\omega-shallow confluent up to 
\math{(n_0\monus 1)\plusomega(n_1\monus 1)}.
Due to the assumption of the theorem at least one of \math{u\varphi}, 
\math{v\varphi},
\wrog\ say \math{v\varphi}, must be either irreducible or have a 
\math{v'\in\tgcons} with 
\bigmath{
  v\varphi
  \refltransindex{\omega+(n_0\monus 1)}
  v'
.}
Now
Lemma~\ref{lemma simpler cases}(\ref{lemma simpler cases two item two})
implies 
\bigmath{
  \hat u
  \downarrowindex{\omega+(n_0\monus 1)}
  \hat v
,}
and then \lemmaaboutconflu\
implies the contradicting
\bigmath{
  \hat u
  \downarrowindex{\omega}
  \hat v
.}
Now we treat the case that
that 
\math{(p\boldequal\truepp )} occurs in \math{D_0} and 
\math{(p\boldequal\falsepp)} occurs in \math{D_1}.
Due to the definition of complementarity,
\truepp\ and \falsepp\ are distinct irreducible ground terms.
Thus we have
\bigmath{
  p\varphi
  \refltransindex{\omega+(n_0\monus 1)}
  \truepp
}
and
\bigmath{
  p\varphi
  \refltransindex{\omega+(n_1\monus 1)}
  \falsepp
.}
In case of \bigmath{n_0,n_1\tightpreceq 1} this contradicts the required
confluence of \redindex{\omega}.
Otherwise, in case of \bigmath{n_0\tightsucceq 1} we have 
\bigmath{
  (n_0\monus 1)\plusomega(n_1\monus 1)
  \prec
  n_0\plusomega n_1
}
and thus by our above assumption
\RX\ is \math\omega-shallow confluent up to 
\math{(n_0\monus 1)\plusomega(n_1\monus 1)}.
This again implies the contradicting
\bigmath{
  \truepp
  \downarrow
  \falsepp
.}

Finally we treat the critical peaks of the form \math{(1,1)}
in case of Theorem~\ref{theorem weakly complementary}.
For these we have to show \math\omega-level parallel joinability or else
\math\omega-level parallel closedness.
Thus, assume 
\math{\varphi\in\Xsubst} and
\math{n\prec\omega} with \bigmath{0\tightprec n} 
such that
\bigmath{
  \forall i\tightprec 2\stopq
    \inparenthesesinline{
         D_i\varphi\mbox{ fulfilled \wrt\ } \redindex{\RX,\omega+(n\monus 1)}
    }
}
and
\bigmath{
    \forall\delta\tightprec n\stopq
    \inparenthesesinline{
           \RX\mbox{ is \math\omega-level confluent up to }\delta
    }
.}
Due to \bigmath{0\tightprec n} 
we have 
\bigmath{
  n\monus 1
  \tightprec
  n
}
and thus \RX\ is \math\omega-level confluent up to \math{n\monus 1}.
By the assumed weak complementarity there must be complementary 
equation literals in \math{D_0 D_1}. 
First we treat the case that
\math{(u\boldequal   v)} and
\math{(u\boldunequal v)} occur in \math{D_0 D_1}.
Then there are \math{\hat u,\hat v\in\tgcons} and \math{v'\in\tsigX} with
\bigmath{
  \hat u
  \antirefltransindex{\omega+(n\monus 1)}
  u\varphi
  \refltransindex    {\omega+(n\monus 1)}
  v'
  \antirefltransindex{\omega+(n\monus 1)}
  v\varphi
  \refltransindex    {\omega+(n\monus 1)}
  \hat v
}
and
\bigmath{
  \hat u
  \notconfluindex    {\omega}
  \hat v
.}
Now, by \math\omega-level confluence up to \math{n\monus 1}, there is some
\math{u'} with
\bigmath{
  \hat u
  \refltransindex    {\omega+(n\monus 1)}
  u'
  \antirefltransindex{\omega+(n\monus 1)}
  v'
}
and then by \math\omega-level confluence up to \math{n\monus 1} again
\bigmath{
  u'
  \downarrowindex{\omega+(n\monus 1)}
  \hat v
,}
and then \lemmaaboutconflu\
implies the contradicting
\bigmath{
  \hat u
  \downarrowindex{\omega}
  \hat v
.}
Now we treat the case that
that 
\math{(p\boldequal\truepp )} and 
\math{(p\boldequal\falsepp)} occur in \math{D_0 D_1}.
Due to the definition of weak complementarity,
\truepp\ and \falsepp\ are distinct irreducible ground terms.
Thus we have
\bigmath{
  \truepp
  \antirefltransindex{\omega+(n\monus 1)}
  p\varphi
  \refltransindex    {\omega+(n\monus 1)}
  \falsepp
.}
By \math\omega-level confluence up to \math{n\monus 1} 
this again implies the contradicting
\bigmath{
  \truepp
  \downarrow
  \falsepp
.}
\end{proofparsepqed}

\pagebreak

\begin{proof}{Theorem~\ref{theorem parallel closed}}
(I) follows from the lemmas 
\ref{lemma parallel closed first level two}
and
\ref{lemma parallel closed second level two}.
\\
(II) follows from the lemmas 
\ref{lemma closed first level two}
and
\ref{lemma strongly closed second level two}.
\\
(III) follows from the lemmas
\ref{lemma parallel closed first level two},
\ref{lemma closed first level two},
and
\ref{lemma closed second level two}, 
since for critical peaks of the form \math{(0,1)}
\math\omega-shallow noisy strong joinability up to \math\omega\ implies 
\math\omega-shallow noisy parallel joinability up to \math\omega\ 
(\cf\ Corollary~\ref{corollary joinabilities}) 
and for non-overlays of the form \math{(1,0)} 
\math\omega-shallow parallel closedness up to \math\omega\ implies
\math\omega-shallow noisy anti-closedness up to \math\omega\ 
(\cf\ Corollary~\ref{corollary closednesses one}).
\\
(IV) follows from the lemmas
\ref{lemma parallel closed first level three},
\ref{lemma closed first level two},
and
\ref{lemma closed second level two}, 
since for critical peaks of the form \math{(0,1)}
\math\omega-shallow noisy strong joinability up to \math\omega\ implies 
\math\omega-shallow noisy weak parallel joinability up to \math\omega\ 
(\cf\ Corollary~\ref{corollary joinabilities})
and for critical peaks of  the form \math{(1,0)}
\math\omega-shallow closedness up to \math\omega\ implies
\math\omega-shallow anti-closedness up to \math\omega\
(\cf\ Corollary~\ref{corollary closednesses one}).
\end{proof}

\begin{proof}{Theorem~\ref{theorem level parallel closed}}
(I) follows from the lemmas 
\ref{lemma parallel closed first level two}
and
\ref{lemma level parallel closed second level two}.
\\
(II) follows from the lemmas 
\ref{lemma closed first level two}
and
\ref{lemma level strongly closed second level two}
\\
(III) follows from the lemmas
\ref{lemma parallel closed first level two},
\ref{lemma closed first level two},
and
\ref{lemma level parallel closed second level three}, 
since for critical peaks of the form \math{(0,1)}
\math\omega-shallow strong joinability up to \math\omega\ implies 
\math\omega-shallow parallel joinability up to \math\omega\ 
(\cf\ Corollary~\ref{corollary joinabilities}) 
and for non-overlays of the form \math{(1,0)} 
\math\omega-shallow parallel closedness up to \math\omega\ implies
\math\omega-shallow anti-closedness up to \math\omega\ 
(\cf\ Corollary~\ref{corollary closednesses one}).
\\
(IV) follows from the lemmas
\ref{lemma parallel closed first level three},
\ref{lemma closed first level two},
and
\ref{lemma level parallel closed second level three}, 
since for critical peaks of the form \math{(0,1)}
\math\omega-shallow strong joinability up to \math\omega\ implies 
\math\omega-shallow weak parallel joinability up to \math\omega\ 
(\cf\ Corollary~\ref{corollary joinabilities})
and for critical peaks of  the form \math{(1,0)}
\math\omega-shallow closedness up to \math\omega\ implies
\math\omega-shallow anti-closedness up to \math\omega\
(\cf\ Corollary~\ref{corollary closednesses one}).
\end{proof}

\begin{proof}{Theorem~\ref{theoremconfluence}}
\underline{1\implies 2:}
By Lemma~\ref{lemma quasi-free three}.
\underline{2\implies 1:}
By Lemma~\ref{lemma necessary}.
\end{proof}

\begin{proof}{Theorem~\ref{theorem quasi-free three}}
\underline{1\implies 2:}
Directly by the lemmas
\ref{lemma quasi-free} and \ref{lemma quasi-free three}.
\underline{2\implies 1:}
By Lemma~\ref{lemma necessary}.
\end{proof}

\begin{proof}{Theorem~\ref{theorem quasi-free}}
\underline{1\implies 2:}
Directly by the lemmas
\ref{lemma quasi-free} and \ref{lemma quasi-free two and zero}.
\underline{2\implies 1:}
By Corollary~\ref{corollary omega shallow confluent up to omega no termination}
and Lemma~\ref{lemma omega shallow joinablility necessary}.
\end{proof}

\begin{proof}{Theorem~\ref{theorem level one}}
\underline{1\implies 2:}
Directly by Lemma~\ref{lemma level one}.
\underline{2\implies 1:}
By Corollary~\ref{corollary omega shallow confluent up to omega no termination}
and Lemma~\ref{lemma omega level joinablility necessary}.
\end{proof}

\begin{proof}{Theorem~\ref{theorem quasi overlay joinable}}
Directly by Lemma~\ref{lemma for theorem quasi overlay joinable}.
\end{proof}

\pagebreak

\begin{proofparsepqed}{Theorem~\ref{theorem parallel closed zero}(I)}
\underline{Claim~1:}
If 
\bigmath{
    \redparaindex{n_1}
    \tight\circ
    \refltransindex{n_1\monus1}
}
strongly commutes over
\refltransindex{n_0},
then
\redindex{n_1} and \redindex{n_0} are commuting.
\\
\underline{Proof of Claim~1:}
\bigmath{
    \redparaindex{n_1}
    \tight\circ
    \refltransindex{n_1\monus1}
}
and
\refltransindex{n_0}
are commuting
by Lemma~\ref{lemma strong commutation one copy}.
Since by Corollary~\ref{corollary parallel one}
and \lemmamonotonicinbeta\
we have
\bigmath{
  \redindex{n_1}
  \subseteq
  \redparaindex{n_1}
  \tight\circ
  \refltransindex{n_1\monus1}
  \subseteq
  \refltransindex{n_1}
,}
now \redindex{n_1} 
and \redindex{n_0} are commuting, too.\QED{Claim~1}

\yestop
\yestop
\noindent
For \math{n_0\preceq n_1\prec\omega} 
we are going to show by induction on \math{n_0\tight+ n_1}
the following property\footroom:
\\\LINEmath{
    w_0
    \antiredparaindex{n_0}
    u
    \redparaindex{n_1}
    w_1
  \quad\implies\quad
    w_0
    \redparaindex{n_1}
    \tight\circ
    \refltransindex{n_1\monus1}
    \circ
    \antirefltransindex{n_0}
    w_1
.}
\begin{diagram}
u&&\rredparaindex{n_1}&&w_1
\\\dredparaindex{n_0}&&&&\drefltransindex{n_0}
\\w_0&\rredparaindex{n_1}&\circ&\rrefltransindex{n_1\monus1}&\circ
\end{diagram}

\noindent
\underline{Claim~2:}
Let \math{\delta\prec\omega}.
If
\\
\linemath{
  \forall n_0,n_1\tightprec\omega\stopq
  \inparenthesesoplist{ 
    \inparenthesesoplist{
        n_0\tightpreceq n_1      
      \oplistund
        n_0\tight+ n_1\tightpreceq\delta
    }
    \oplistimplies
    \forall w_0,w_1,u\stopq
    \inparenthesesoplist{
        w_0
        \antiredparaindex{n_0}
        u
        \redparaindex{n_1}
        w_1
      \oplistimplies 
        w_0
        \redparaindex{n_1}
        \tight\circ
        \refltransindex{n_1\monus1}
        \circ
        \antirefltransindex{n_0}
        w_1
    }
  }
,}
then
\\
\linemath{
  \forall n_0,n_1\tightprec\omega\stopq
  \inparenthesesoplist{ 
    \inparenthesesoplist{
        n_0\tightpreceq n_1      
      \oplistund
        n_0\tight+ n_1\tightpreceq\delta
    }
    \oplistimplies 
            \redparaindex{n_1}
            \tight\circ
            \refltransindex{n_1\monus1}
      \mbox{ strongly commutes over }
            \refltransindex{n_0}
  }
,}
and 
\RX\ is \math 0-shallow confluent up to \math{\delta}.
\\
\underline{Proof of Claim~2:}
By induction on \math{\delta} in \tightprec.
First we show the strong commutation.
Assume \math{n_0\preceq n_1\prec\omega} with
\math{n_0\tight+ n_1\tightpreceq\delta}. 
By Lemma~\ref{lemma strong commutation one copy} it suffices to show that
\bigmath{
    \redparaindex{n_1}
    \tight\circ
    \refltransindex{n_1\monus1}
}
strongly commutes over
\redindex{n_0}.
Assume
\bigmath{
    w_0
    \antiredindex{n_0}
    u
    \redparaindex{n_1}
    w_1
    \refltransindex{n_1\monus1}
    w_2
}
(\cf\ diagram below).
By the above property there is some \math{w_1'}
with
\bigmath{
    w_0
    \redparaindex{n_1}
    \tight\circ
    \refltransindex{n_1\monus1}
    w_1'
    \antirefltransindex{n_0}
    w_1
.}
Next we show that we can close the peak
\bigmath{ 
    w_1'
    \antirefltransindex{n_0}
    w_1
    \refltransindex{n_1\monus1}
    w_2
}
according to 
\bigmath{
    w_1'
    \refltransindex{n_1\monus1}
    w_2'
    \antirefltransindex{n_0}
    w_2
}
for some \math{w_2'}.
In case of \bigmath{n_1\tightequal0}
this is possible due \bigmath{w_1\tightequal w_2.}
Otherwise we have 
\bigmath{
  n_0\tight+(n_1\monus1)
  \tightprec 
  n_0\tight+ n_1
  \tightpreceq
  \delta
} 
and due to our induction hypothesis
(saying that \RX\ is \math 0-shallow confluent up to all 
\math{\delta'\prec\delta
})
this is possible again.
\begin{diagram}
u
&&\rredparaindex{n_1}&&
w_1&\rrefltransindex{n_1\monus1}&w_2
\\
\dredindex{n_0}
&&&
&\drefltransindex{n_0}&&\drefltransindex{n_0}
\\
w_0
&\rredparaindex{n_1}&\circ&\rrefltransindex{n_1\monus1}
&w_1'&\rrefltransindex{n_1\monus1}&w_2'
\end{diagram}
Finally we show \math 0-shallow confluence up to \math\delta.
Assume
\bigmath{n_0\tight+ n_1\tightpreceq\delta}
and
\bigmath{
    w_0
    \antirefltransindex{n_0}
    u
    \refltransindex    {n_1}
    w_1
.}
Due to symmetry in \math{n_0} and \math{n_1} we may assume
\bigmath{n_0\tightpreceq n_1.}
Above we have shown that 
\bigmath{
    \redparaindex{n_1}
    \tight\circ
    \refltransindex{n_1\monus1}
}
strongly commutes over
\refltransindex{n_0}. 
By Claim~1 we finally get
\bigmath{
  w_0\refltransindex{n_1}\circ\antirefltransindex{n_0} w_1
}
as desired.%
\QED{Claim~2}

\pagebreak

\yestop
\yestop
\noindent
Note that 
for \bigmath{n_0\tightequal0} 
our property follows 
from 
\bigmath{\antiredparaindex0\subseteq\id.}

The benefit of 
Claim~2 is twofold: First, it says that our theorem is valid
if the above property
holds for all \math{n_0\preceq n_1\prec\omega}.
Second, it strengthens the property when used as induction hypothesis. Thus 
(writing \math{n_i\tight+1} instead of \math{n_i} since we may assume
 \math{0\tightprec n_0\tightpreceq n_1})
it
now suffices to show
for
\math{n_0\preceq n_1\prec\omega}
that 
\\\linemath{
    w_0
    \antiredparaindex{n_0+1,\Pi_0}
    u
    \redparaindex    {n_1+1,\Pi_1}
    w_1
}
together with our induction hypotheses 
that\headroom
\\\linemath{\headroom
  \forall\delta\tightprec(n_0\tight+1)\tight+(n_1\tight+1)\stopq
  \mbox{\RX\ is \math 0-shallow confluent up to }\delta
}
\headroom
and (due to \math{n_0\tightpreceq n_1\tight+1}
and 
\math{
  n_0\tight+(n_1\tight+1)\tightprec(n_0\tight+1)\tight+(n_1\tight+1)
})
\\\linenomath{\headroom
  \bigmath{ 
    \redparaindex{n_1+1}
    \tight\circ
    \refltransindex{n_1}
  }
  strongly commutes over
  \refltransindex{n_0}
}
\headroom
implies
\LINEmath{
    w_0
    \redparaindex{n_1+1}
    \tight\circ
    \refltransindex{n_1}
    \circ
    \antirefltransindex{n_0+1}
    w_1
.\ \ }
\begin{diagram}
u&&\rredparaindex{n_1+1,\,\Pi_1}
&&w_1
\\\dredparaindex{n_0+1,\,\Pi_0}&&
&&\drefltransindex{n_0+1}
\\w_0&\rredparaindex{n_1+1}
&\circ&\rrefltransindex{n_1}&\circ
\end{diagram}

\noindent
{Note that for the availability of our second induction hypothesis  
it is important that we have imposed the restriction
``\math{n_0\tightpreceq n_1}'' in opposition to the restriction
``\math{n_0\tightsucceq n_1}''.
In the latter case the availability of our second induction hypothesis would
require 
\bigmath{
  n_0\tight+1\tightsucceq n_1\tight+1
  \implies
  n_0\tightsucceq n_1\tight+1
}
which is not true for \bigmath{n_0\tightequal n_1.}
The additional hypothesis  
\\\linenomath{
  \bigmath{ 
    \redparaindex{n_1}
    \tight\circ
    \refltransindex{n_1\monus1}
  }
  strongly commutes over
  \refltransindex{n_0+1}
}
of the latter restriction is useless for our proof.}

\noindent
\Wrog\ let the positions of \math{\Pi_i} be maximal
in the sense that for any \math{p\in\Pi_i} 
and \math{\Xi\subseteq \TPOS u\tightcap(p\N^+)}
we do not have 
\bigmath{
  u
  \redparaindex{n_i+1,(\Pi_i\setminus\{p\})\cup\Xi}
  w_i
}
anymore.
Then for each \math{i\prec2} and
\math{p\in\Pi_i} there are
\bigmath{\kurzregelindex{i,p}\in\R}
and
\bigmath{\mu_{i,p}\in\Xsubst}
with
\bigmath{l_{i,p}\tightin\tcs,}
\bigmath{u/p\tightequal l_{i,p}\mu_{i,p},}
\bigmath{r_{i,p}\mu_{i,p}\tightequal w_i/p,}
\math{C_{i,p}\mu_{i,p}} fulfilled \wrt\ \redindex{n_i}.
Finally, for each \math{i\prec2}:
\bigmath{
  w_i\tightequal\replpar{u}{p}{r_{i,p}\mu_{i,p}}{p\tightin\Pi_i}
.}

\noindent
Define
the set of inner overlapping positions by
\\\linemath{
  \displaystyle
  \Omega(\Pi_0,\Pi_1)
  :=
  \bigcup_{i\prec2}
    \setwith
      {p\tightin\Pi_{1-i}}
      {\exists q\tightin\Pi_i\stopq\exists q'\tightin\N^\ast\stopq
        p\tightequal q q'
      }
,}
and the length of a term by
\bigmath{\lambda(\anonymousfpp{t_0}{t_{m-1}}):=1+\sum_{j\prec m}\lambda(t_j).}

\noindent
Now we start a second level of induction on
\bigmath{  \displaystyle
  \sum_{p'\in\Omega(\Pi_0,\Pi_1)}\lambda(u/p')
}
in \tightprec.

\noindent
Define the set of top positions by
\\\linemath{
  \displaystyle
  \Theta
  :=
      \setwith
      {p\tightin\Pi_0\tightcup\Pi_1}
      {\neg\exists q\tightin\Pi_0\tightcup\Pi_1\stopq
           \exists q'\tightin\N^+\stopq
             p\tightequal q q'
      }
.}
Since the prefix ordering is wellfounded we have
\bigmath{
  \forall i\tightprec2\stopq
  \forall p\tightin\Pi_i\stopq
  \exists q\tightin\Theta\stopq
  \exists q'\tightin\N^\ast\stopq
    p\tightequal q q'
.}
Then
\bigmath{
  \forall i\tightprec2\stopq
  w_i
  \tightequal
  \replpar{w_i}{q}{w_i/q}{q\tightin\Theta}
  \tightequal
  \replpar{\replpar{u}{p}{r_{i,p}\mu_{i,p}}{p\tightin\Pi_i}}
          {q}{w_i/q}{q\tightin\Theta}
  \tightequal
  \replpar{u}{q}{w_i/q}{q\tightin\Theta}
.}
Thus, it now suffices to show for all \math{q\in\Theta}
\\\linemath{\headroom\footroom
    w_0/q
    \redparaindex{n_1+1}
    \tight\circ
    \refltransindex{n_1}
    \circ
    \antirefltransindex{n_0+1}
    w_1/q
}
because then we have 
\\\LINEmath{
  w_0
  \tightequal
  \replpar{u}{q}{w_0/q}{q\tightin\Theta}
    \redparaindex{n_1+1}
    \tight\circ
    \refltransindex{n_1}
    \circ
    \antirefltransindex{n_0+1}
  \replpar{u}{q}{w_1/q}{q\tightin\Theta}
  \tightequal
  w_1
.}

\noindent
Therefore we are left with the following two cases for \math{q\in\Theta}:

\pagebreak

\yestop
\yestop
\noindent
\underline{\underline{\underline{\math{q\tightnotin\Pi_1}:}}}
Then \bigmath{q\tightin\Pi_0.}
Define \math{\Pi_1':=\setwith{p}{q p\tightin\Pi_1}}.
\noindent
We have two cases:

\yestop
\noindent
\underline{\underline{``The variable overlap (if any) case'':
\math{
  \forall p\tightin\Pi_1'\tightcap\TPOS{l_{0,q}}\stopq
    l_{0,q}/p\tightin\V
}:}}
\begin{diagram}
l_{0,q}\mu_{0,q}&&\rredparaindex{n_1+1,\,\Pi_1'}&&&&w_1/q
\\&&&&&&\dequal
\\\dredindex{n_0+1,\,\emptyset}&&&&&&l_{0,q}\nu
\\&&&&&&\dredindex{n_0+1}
\\w_0/q&\requal&r_{0,q}\mu_{0,q}&&\rredparaindex{n_1+1}&&r_{0,q}\nu
\end{diagram}
\noindent
Define a function \math\Gamma\ on \V\ by (\math{x\tightin\V}):
\bigmath{
  \Gamma(x):=
  \setwith{(p',p'')}
          {l_{0,q}/p'\tightequal x\ \wedge\ p' p''\in\Pi_1'}
.}

\noindent
\underline{Claim~7:}
There is some \math{\nu\in\Xsubst} with
\\\LINEmath{
  \forall x\in\V\stopq
    \inparenthesesoplist{
       x\mu_{0,q}
       \redparaindex{n_1+1}
       x\nu
    \oplistund
       \forall p'\tightin\DOM{\Gamma(x)}\stopq
         x\nu
         \tightequal
         \replpar
           {x\mu_{0,q}}
           {p''}
           {r_{1,q p' p''}\mu_{1,q p' p''}}
           {(p',p'')\tightin\Gamma(x)}
    }
.}
\\
\underline{Proof of Claim~7:}
\\
In case of \bigmath{\DOM{\Gamma(x)}\tightequal\emptyset} we define
\bigmath{x\nu:=x\mu_{0,q}.}
If there is some \math{p'} such that 
\bigmath{\DOM{\Gamma(x)}\tightequal\{p'\}}
we define 
\bigmath{
  x\nu
  :=
  \replpar
    {x\mu_{0,q}}
    {p''}
    {r_{1,q p' p''}\mu_{1,q p' p''}}
    {(p',p'')\tightin\Gamma(x)}
.}
This is appropriate since due to 
\bigmath{
  \forall(p',p'')\tightin\Gamma(x)\stopq
    x\mu_{0,q}/p''
    \tightequal 
    l_{0,q}\mu_{0,q}/p' p''
    \tightequal 
    u/q p' p''
    \tightequal 
    l_{1,q p' p''}\mu_{1,q p' p''}
}
we have
\\\LINEmath{
  \begin{array}{l@{}l@{}l}
  x\mu_{0,q}&
  \tightequal&
  \replpar
    {x\mu_{0,q}}
    {p''}
    {l_{1,q p' p''}\mu_{1,q p' p''}}
    {(p',p'')\tightin\Gamma(x)}
  \redparaindex{n_1+1}\\&&
  \replpar
     {x\mu_{0,q}}
     {p''}
     {r_{1,q p' p''}\mu_{1,q p' p''}}
     {(p',p'')\tightin\Gamma(x)}
  \tightequal
  x\nu.  
  \end{array}
}
\\
Finally, in case of \bigmath{\CARD{\DOM{\Gamma(x)}}\succ1,} \math{l_{0,q}} is
not linear in \math x, which contradicts the left-linearity assumption of the
theorem.\QED{Claim~7}

\noindent
\underline{Claim~8:}
\bigmath{
  l_{0,q}\nu
  \tightequal 
  w_1/q
.}
\\
\underline{Proof of Claim~8:}
\\
By Claim~7 we get
\bigmath{
  w_1/q
  \tightequal
  \replpar 
    {u/q}
    {p' p''}
    {r_{1,q p' p''}\mu_{1,q p' p''}}
    {\exists x\tightin\V\stopq(p',p'')\tightin\Gamma(x)}
  \tightequal\\
  \replpar
    {\replpar
       {l_{0,q}}
       {p'}
       {x\mu_{0,q}}
       {l_{0,q}/p'\tightequal x\tightin\V}
    }
    {p' p''}
    {r_{1,q p' p''}\mu_{1,q p' p''}}
    {\exists x\tightin\V\stopq(p',p'')\tightin\Gamma(x)}
  \tightequal\\
  \replpar
    {l_{0,q}}
    {p'}
    {\replpar
       {x\mu_{0,q}}
       {p''}
       {r_{1,q p' p''}\mu_{1,q p' p''}}
       {(p',p'')\tightin\Gamma(x)}}
    {l_{0,q}/p'\tightequal x\tightin\V}
  \tightequal\\
  \replpar
    {l_{0,q}}
    {p'}
    {x\nu}
    {l_{0,q}/p'\tightequal x\tightin\V}
  \tightequal
  l_{0,q}\nu
.}\QED{Claim~8}

\noindent
\underline{Claim~9:}
\bigmath{
  w_0/q
  \redparaindex{n_1+1}
  r_{0,q}\nu
.}
\\
\underline{Proof of Claim~9:} 
Since 
\bigmath{
  w_0/q
  \tightequal
  r_{0,q}\mu_{0,q}
,} 
this follows directly from Claim~7.%
\QED{Claim~9}

\noindent
By claims 8 and 9 it now suffices to show
\bigmath{
  l_{0,q}\nu
  \redindex{n_0+1}
  r_{0,q}\nu
,}
which again follows from 
Lemma~\ref{lemma invariance of fulfilledness}
since 
\kurzregelindex{0,q} is \math 0-quasi-normal \wrt\ \RX\
(due to \bigmath{l_{0,q}\tightin\tcs} and the assumption of our theorem),
since
\RX\ is \math 0-shallow confluent up to
\bigmath{(n_1\tight+1)\tight+ n_0}
(by our induction hypothesis),
and since
\bigmath{\forall x\tightin\V\stopq x\mu_{0,q}\refltransindex{n_1+1}x\nu}
by Claim~7 and Corollary~\ref{corollary parallel one}.%
\\\Qeddouble{``The variable overlap (if any) case''}

\pagebreak

\yestop
\noindent
\underline{\underline{``The critical peak case'':
There is some \math{p\in \Pi_1'\tightcap\TPOS{l_{0,q}}}
with \math{l_{0,q}/p\tightnotin\V}:}}
\begin{diagram}
l_{0,q}\mu_{0,q}&\rredindex{n_1+1,\,p}&u'
&&\rredparaindex{n_1+1,\,\Pi_1'\setminus\{p\}}
&&w_1/q
\\
&&\dredparaindex{n_0+1}
&&
&&\drefltransindex{n_0+1}
\\
\dredindex{n_0+1,\,\emptyset}&&v_1
&\rredparaindex{n_1+1}
&\circ&\rrefltransindex{n_1}&v_1'
\\
&&\drefltransindex{n_0}
&&
&&\drefltransindex{n_0}
\\
w_0/q
&\requal&w_0/q
&\rredparaindex{n_1+1}
&\circ&\rrefltransindex{n_1}&\circ
\end{diagram}
\underline{Claim~10:}
\bigmath{p\tightnotequal\emptyset.}
\\
\underline{Proof of Claim~10:}
If \bigmath{p\tightequal\emptyset,} then
\bigmath{\emptyset\tightin\Pi_1',} then
\bigmath{q\tightin\Pi_1,} which contradicts our global case assumption.%
\QED{Claim~10}

\noindent
Let \math{\xi\in\SUBST\V\V} be a bijection with 
\bigmath{
  \xi[\VAR{\kurzregelindex{1,q p}}]\cap\VAR{\kurzregelindex{0,q}}
  =
  \emptyset
.}
\\
Define
\bigmath{
  \Y
  :=
  \xi[\VAR{\kurzregelindex{1,q p}}]\cup\VAR{\kurzregelindex{0,q}}
.}
\\
Let \math{\varrho\in\Xsubst} be given by
$\ x\varrho=
\left\{\begin{array}{@{}l@{}l@{}}
  x\mu_{0,q}        &\mbox{ if }x\in\VAR{\kurzregelindex{0,q}}\\
  x\xi^{-1}\mu_{1,q p}&\mbox{ else}\\
\end{array}\right\}
\:(x\tightin\V)$.
\\
By
\math{
  l_{1,q p}\xi\varrho
  \tightequal 
  l_{1,q p}\xi\xi^{-1}\mu_{1,q p}
  \tightequal 
  u/q p
  \tightequal 
  l_{0,q}\mu_{0,q}/p  
  \tightequal 
  l_{0,q}\varrho/p
  \tightequal 
  (l_{0,q}/p)\varrho
}
\\
let
\math{
  \sigma:=\minmgu{\{(l_{1,q p}\xi}{l_{0,q}/p)\},\Y}
}
and
\math{\varphi\in\Xsubst}
with
\math{
  \domres{\inpit{\sigma\varphi}}\Y
  \tightequal
  \domres\varrho\Y
}.
\\
Define 
\math{
  u':=  
  \repl{l_{0,q}\mu_{0,q}}
       {p}
       {r_{1,q p}\mu_{1,q p}}
}.
We get
\\\LINEmath{
  \arr{{l@{}l}
    u'\tightequal
    &
    \repl
      {\replpar
         {u/q}
         {p'}
         {l_{1,q p'}\mu_{1,q p'}}
         {p'\tightin\Pi_1'\tightsetminus\{p\}}}
      {p}
      {r_{1,q p}\mu_{1,q p}}
    \redparaindex{n_1+1,\Pi_1'\setminus\{p\}}
    \\&
    \replpar{u/q}{p'}{r_{1,q p'}\mu_{1,q p'}}{p'\tightin\Pi_1'}    
    \tightequal
    w_1/q
  .
  }
}
\\
If 
\bigmath{
  \repl{l_{0,q}}{p}{r_{1,q p}\xi}\sigma
  \tightequal
  r_{0,q}\sigma
,}
then the proof is finished due to 
\\\LINEmath{
  w_0/q
  \tightequal
  r_{0,q}\mu_{0,q}
  \tightequal
  r_{0,q}\sigma\varphi
  \tightequal
  \repl{l_{0,q}}{p}{r_{1,q p}\xi}\sigma\varphi
  \tightequal
  u'
  \redparaindex{n_1+1,\Pi_1'\setminus\{p\}}
  w_1/q
.}
\\
Otherwise 
we have
\bigmath{
  (\,
   (\repl{l_{0,q}}{p}{r_{1,q p}\xi},
    C_{1,q p}\xi,    
    0),\penalty-1\,
   (r_{0,q},
    C_{0,q},
    0),\penalty-1\,
    l_{0,q},\,
    \sigma,\penalty-1\,
    p\,)
  \in{\rm CP}(\R)
;}
\bigmath{p\tightnotequal\emptyset}
(due to Claim~10);
\bigmath{C_{1,q p}\xi\sigma\varphi=C_{1,q p}\mu_{1,q p}}
is fulfilled \wrt\ \redindex{n_1};
\bigmath{C_{0,q}\sigma\varphi=C_{0,q}\mu_{0,q}}
is fulfilled \wrt\ \redindex{n_0}.
Since 
\bigmath{
  \forall\delta\tightprec(n_1\tight+1)\tight+(n_0\tight+1)\stopq
  \mbox{\RX\ is \math 0-shallow confluent up to }\delta
}
(by our induction hypothesis) 
due to our assumed \math 0-shallow parallel closedness 
(matching the definition's \math{n_0} to our \math{n_1\tight+1}
                   and its \math{n_1} to our \math{n_0\tight+1})
we have
\bigmath{
  u'
  \tightequal
  \repl{l_{0,q}}{p}{r_{1,q p}\xi}\sigma\varphi
  \penalty-1
  \redparaindex{n_0+1}
  \penalty-1
  v_1
  \penalty-1
  \refltransindex{n_0}
  r_{0,q}\sigma\varphi
  \tightequal
  r_{0,q}\mu_{0,q}
  \tightequal
  w_0/q
}
for some \math{v_1}.
We then have
\bigmath{
  v_1
  \antiredparaindex{n_0+1,\Pi''} 
  u'
  \redparaindex{n_1+1,\Pi_1'\setminus\{p\}}
  w_1/q
}
for some \math{\Pi''}.
By 
\\\bigmath{
  \displaystyle
  \sum_{p''\in\Omega(\Pi'',\Pi_1'\setminus\{p\})}
  \lambda(u'/p'')
  \ \ \preceq
  \sum_{p''\in\Pi_1'\setminus\{p\}}
  \lambda(u'/p'')
  \ \ =
  \sum_{p''\in\Pi_1'\setminus\{p\}}
  \lambda(u/q p'')
  \ \ \prec
  \sum_{p''\in\Pi_1'}
  \lambda(u/q p'')
  \ \ =
}\\\bigmath{\displaystyle
  \sum_{p'\in q\Pi_1'}
  \lambda(u/p')
  \ \ =
  \sum_{p'\in\Omega(\{q\},\Pi_1)}
  \lambda(u/p')
  \ \ \preceq
  \sum_{p'\in\Omega(\Pi_0,\Pi_1)}
  \lambda(u/p')
,}
due to our second induction level 
we get some \math{v_1'} with 
\bigmath{
  v_1
  \redparaindex{n_1+1}
  \tight\circ
  \refltransindex{n_1}
  v_1'
  \antirefltransindex{n_0+1}
  w_1/q
.}
Finally by our induction hypothesis that
  \bigmath{ 
    \redparaindex{n_1+1}
    \tight\circ
    \refltransindex{n_1}
  }
  strongly commutes over
  \refltransindex{n_0}
the peak at \math{v_1} can be closed according to 
\bigmath{
  w_0/q
  \redparaindex{n_1}
  \tight\circ
  \refltransindex{n_1}
  \circ
  \antirefltransindex{n_0}
  v_1'
.}%
\\
\Qeddouble{``The critical peak case''}\QEDtriple{``\math{q\tightnotin\Pi_1}''}

\pagebreak

\noindent
\underline{\underline{\underline{\math{q\tightin\Pi_1}:}}}
Define \math{\Pi_0':=\setwith{p}{q p\tightin\Pi_0}}.
We have two cases:

\yestop
\noindent
\underline{\underline{``The second variable overlap (if any) case'':
\math{
  \forall p\tightin\Pi_0'\tightcap\TPOS{l_{1,q}}\stopq
    l_{1,q}/p\tightin\V
}:}}
\begin{diagram}
l_{1,q}\mu_{1,q}&&&\rredindex{n_1+1,\,\emptyset}&&&w_1/q
\\&&&&&&\dequal
\\\dredparaindex{n_0+1,\,\Pi_0'}&&&&&&r_{1,q}\mu_{1,q}
\\&&&&&&\dredparaindex{n_0+1}
\\w_0/q&\requal&l_{1,q}\nu&&\rredindex{n_1+1}&&r_{1,q}\nu
\end{diagram}
\noindent
Define a function \math\Gamma\ on \V\ by (\math{x\tightin\V}):
\bigmath{
  \Gamma(x):=
  \setwith{(p',p'')}
          {l_{1,q}/p'\tightequal x\ \wedge\ p' p''\in\Pi_0'}
.}

\noindent
\underline{Claim~11:}
There is some \math{\nu\in\Xsubst} with
\\\LINEmath{
  \forall x\in\V\stopq
    \inparenthesesoplist{
       x\nu
       \antiredparaindex{n_0+1}
       x\mu_{1,q}
    \oplistund
       \forall p'\tightin\DOM{\Gamma(x)}\stopq
         \replpar
           {x\mu_{1,q}}
           {p''}
           {r_{0,q p' p''}\mu_{0,q p' p''}}
           {(p',p'')\tightin\Gamma(x)}
         \tightequal
         x\nu
    }
.}
\\
\underline{Proof of Claim~11:}
\\
In case of \bigmath{\DOM{\Gamma(x)}\tightequal\emptyset} we define
\bigmath{x\nu:=x\mu_{1,q}.}
If there is some \math{p'} such that 
\bigmath{\DOM{\Gamma(x)}\tightequal\{p'\}}
we define 
\bigmath{
  x\nu
  :=
  \replpar
    {x\mu_{1,q}}{p''}{r_{0,q p' p''}\mu_{0,q p' p''}}{(p',p'')\tightin\Gamma(x)}
.}
This is appropriate since due to 
\bigmath{
  \forall(p',p'')\tightin\Gamma(x)\stopq
    x\mu_{1,q}/p''
    \tightequal 
    l_{1,q}\mu_{1,q}/p' p''
    \tightequal 
    u/q p' p''
    \tightequal 
    l_{0,q p' p''}\mu_{0,q p' p''}
}
we have
\\\LINEmath{
  \begin{array}{l@{}l@{}l}
  x\mu_{1,q}&
  \tightequal&
  \replpar
    {x\mu_{1,q}}
    {p''}
    {l_{0,q p' p''}\mu_{0,q p' p''}}
    {(p',p'')\tightin\Gamma(x)}
  \redparaindex{n_0+1}\\&&
  \replpar
     {x\mu_{1,q}}
     {p''}
     {r_{0,q p' p''}\mu_{0,q p' p''}}
     {(p',p'')\tightin\Gamma(x)}
  \tightequal
  x\nu.  
  \end{array}
}
\\
Finally, in case of \bigmath{\CARD{\DOM{\Gamma(x)}}\succ1,} \math{l_{1,q}} is
not linear in \math x, which contradicts the left-linearity assumption of the
theorem.\QED{Claim~11}

\noindent
\underline{Claim~12:}
\bigmath{w_0/q\tightequal l_{1,q}\nu.}
\\
\underline{Proof of Claim~12:}
\\
By Claim~11 we get
\bigmath{
  w_0/q
  \tightequal
  \replpar 
    {u/q}
    {p' p''}
    {r_{0,q p' p''}\mu_{0,q p' p''}}
    {\exists x\tightin\V\stopq(p',p'')\tightin\Gamma(x)}
  \tightequal\\
  \replpar
    {\replpar
       {l_{1,q}}
       {p'}
       {x\mu_{1,q}}
       {l_{1,q}/p'\tightequal x\tightin\V}
    }
    {p' p''}
    {r_{0,q p' p''}\mu_{0,q p' p''}}
    {\exists x\tightin\V\stopq(p',p'')\tightin\Gamma(x)}
  \tightequal\\
  \replpar
    {l_{1,q}}
    {p'}
    {\replpar
       {x\mu_{1,q}}
       {p''}
       {r_{0,q p' p''}\mu_{0,q p' p''}}
       {(p',p'')\tightin\Gamma(x)}}
    {l_{1,q}/p'\tightequal x\tightin\V}
  \tightequal\\
  \replpar
    {l_{1,q}}
    {p'}
    {x\nu}
    {l_{1,q}/p'\tightequal x\tightin\V}
  \tightequal
  l_{1,q}\nu
.}\QED{Claim~12}

\noindent
\underline{Claim~13:}
\bigmath{
  r_{1,q}\nu
  \antiredparaindex{n_0+1}
  w_1/q
.}
\\
\underline{Proof of Claim~13:} 
Since \bigmath{r_{1,q}\mu_{1,q}\tightequal w_1/q,} 
this follows directly from Claim~11.%
\QED{Claim~13}

\noindent
By claims 12 and 13 
using Corollary~\ref{corollary parallel one} 
it now suffices to show
\bigmath{
  l_{1,q}\nu
  \redindex{n_1+1}
  r_{1,q}\nu
,}
which again follows from 
Claim~11,
Corollary~\ref{corollary parallel one},
Lemma~\ref{lemma invariance of fulfilledness}
(matching 
 its \math{n_0} to our \math{n_0\tight+1} and 
 its \math{n_1} to our \math{n_1}),
and our induction hypothesis that \RX\ is \math 0-shallow confluent up to
\bigmath{
  (n_0\tight+1)\tight+ n_1
.}%
\\\Qeddouble{``The second variable overlap (if any) case''}

\pagebreak

\yestop
\noindent
\underline{\underline{``The second critical peak case'':
There is some \math{p\in \Pi_0'\tightcap\TPOS{l_{1,q}}}
with \math{l_{1,q}/p\tightnotin\V}:}}
\begin{diagram}
l_{1,q}\mu_{1,q}
&&&\rredindex{n_1+1,\,\emptyset}&&&w_1/q
\\\dredindex{n_0+1,\,p}
&&
&&
&&\drefltransindex{n_0+1}
\\u'
&&\rredparaindex{n_1+1}
&&v_1
&\rrefltransindex{n_1}&v_2
\\
\dredparaindex{n_0+1,\,\Pi_0'\setminus\{p\}}
&&
&&\drefltransindex{n_0+1}
&&\drefltransindex{n_0+1}
\\
w_0/q
&\rredparaindex{n_1+1}
&\circ&\rrefltransindex{n_1}&v_1'
&\rrefltransindex{n_1}&\circ
\end{diagram}
Let \math{\xi\in\SUBST\V\V} be a bijection with 
\bigmath{
  \xi[\VAR{\kurzregelindex{0,q p}}]\cap\VAR{\kurzregelindex{1,q}}
  =
  \emptyset
.}
\\
Define
\bigmath{
  \Y
  :=
  \xi[\VAR{\kurzregelindex{0,q p}}]\cup\VAR{\kurzregelindex{1,q}}
.}
\\
Let \math{\varrho\in\Xsubst} be given by
$\ x\varrho=
\left\{\begin{array}{@{}l@{}l@{}}
  x\mu_{1,q}        &\mbox{ if }x\in\VAR{\kurzregelindex{1,q}}\\
  x\xi^{-1}\mu_{0,q p}&\mbox{ else}\\
\end{array}\right\}
\:(x\tightin\V)$.
\\
By
\math{
  l_{0,q p}\xi\varrho
  \tightequal 
  l_{0,q p}\xi\xi^{-1}\mu_{0,q p}
  \tightequal 
  u/q p
  \tightequal 
  l_{1,q}\mu_{1,q}/p  
  \tightequal 
  l_{1,q}\varrho/p
  \tightequal 
  (l_{1,q}/p)\varrho
}
\\
let
\math{
  \sigma:=\minmgu{\{(l_{0,q p}\xi}{l_{1,q}/p)\},\Y}
}
and
\math{\varphi\in\Xsubst}
with
\math{
  \domres{\inpit{\sigma\varphi}}\Y
  \tightequal
  \domres\varrho\Y
}.
\\
Define 
\math{
  u':=  
  \repl{l_{1,q}\mu_{1,q}}
       {p}
       {r_{0,q p}\mu_{0,q p}}
}. 
We get
\\\LINEmath{
  \begin{array}{l@{}l@{}l}
  w_0/q&
  \tightequal&
  \replpar{u/q}{p'}{r_{0,q p'}\mu_{0,q p'}}{p'\tightin\Pi_0'}
  \antiredparaindex{n_0+1,\Pi_0'\setminus\{p\}}
  \\&&
  \repl
    {\replpar
       {u/q}{p'}{l_{0,q p'}\mu_{0,q p'}}{p'\tightin\Pi_0'\tightsetminus\{p\}}}
    {p}
    {r_{0,q p}\mu_{0,q p}}
  \tightequal
  u'
  .  
  \end{array}
}  
\\
If 
\bigmath{
  \repl{l_{1,q}}{p}{r_{0,q p}\xi}\sigma
  \tightequal
  r_{1,q}\sigma
,}
then the proof is finished due to 
\\\linemath{
  w_0/q
  \antiredparaindex{n_0+1,\Pi_0'\setminus\{p\}}
  u'
  \tightequal
  \repl{l_{1,q}}{p}{r_{0,q p}\xi}\sigma\varphi
  \tightequal
  r_{1,q}\sigma\varphi
  \tightequal
  r_{1,q}\mu_{1,q}
  \tightequal
  w_1/q.
}
Otherwise 
we have
\bigmath{
  (\,
   (\repl{l_{1,q}}{p}{r_{0,q p}\xi},
    C_{0,q p}\xi,    
    0),\penalty-1\,
   (r_{1,q},
    C_{1,q},
    0),\penalty-1\,
    l_{1,q},\,
    \sigma,\penalty-1\,
    p\,)
  \in{\rm CP}(\R)
;}
\bigmath{C_{0,q p}\xi\sigma\varphi=C_{0,q p}\mu_{0,q p}}
is fulfilled \wrt\ \redindex{n_0};
\bigmath{C_{1,q}\sigma\varphi=C_{1,q}\mu_{1,q}}
is fulfilled \wrt\ \redindex{n_1}.
Since 
\bigmath{
  \forall\delta\tightprec(n_0\tight+1)\tight+(n_1\tight+1)\stopq
  \mbox{\RX\ is \math 0-shallow confluent up to }\delta
}
(by our induction hypothesis) 
due to our assumed \math 0-shallow noisy parallel joinability 
(matching the definition's \math{n_0} to our \math{n_0\tight+1}
                   and its \math{n_1} to our \math{n_1\tight+1}
)
we have
\bigmath{
  u'
  \tightequal
  \repl{l_{1,q}}{p}{r_{0,q p}\xi}\sigma\varphi
  \redparaindex{n_1+1}
  \penalty-1
  v_1
  \refltransindex{n_1}
  \penalty-1
  v_2
  \antirefltransindex{n_0+1}
  \penalty-1
  r_{1,q}\sigma\varphi
  \tightequal
  r_{1,q}\mu_{1,q}
  \tightequal
  w_1/q
}
for some \math{v_1}, \math{v_2}.
We then have
\bigmath{
  w_0/q
  \antiredparaindex{n_0+1,\Pi_0'\setminus\{p\}}
  u'
  \redparaindex{n_1+1,\Pi''} 
  v_1
}
for some \math{\Pi''}.
Since
\bigmath{\displaystyle
  \sum_{p''\in\Omega(
    \Pi_0'\setminus\{p\}
    ,
    \Pi''
    )}
  \lambda(u'/p'')
  \ \ \preceq
  \sum_{p''\in\Pi_0'\setminus\{p\}}
  \lambda(u'/p'')
  \ \ =
  \sum_{p''\in\Pi_0'\setminus\{p\}}
  \lambda(u/q p'')
  \ \ \prec
  \sum_{p''\in\Pi_0'}
  \lambda(u/q p'')
  \ \ =
  \sum_{p'\in q\Pi_0'}
  \lambda(u/p')
  \ \ =
  \sum_{p'\in\Omega(
    \Pi_0
    ,
    \{q\}
    )}
  \lambda(u/p')
  \ \ \preceq
  \sum_{p'\in\Omega(\Pi_0,\Pi_1)}
  \lambda(u/p')
}
due to our second induction level 
we get some \math{v_1'} with 
\bigmath{
  w_0/q
  \redparaindex{n_1+1}
  \tight\circ
  \refltransindex{n_1}
  v_1'
  \antirefltransindex{n_0+1}
  v_1
.} 
Finally the peak at \math{v_1} can be closed according to
\bigmath{
  v_1'
  \refltransindex{n_1}
  \circ
  \antirefltransindex{n_0+1}
  v_2
}
by our induction hypothesis saying that \RX\ is
\math 0-shallow confluent up to \math{(n_0\tight+1)\tight+ n_1}.%
\\\Qeddouble{``The second critical peak case''}
\end{proofparsepqed}

\pagebreak

\begin{proofparsepqed}{Theorem~\ref{theorem parallel closed zero}(II)}
The parts in the following proof which are only for 
{Theorem~\ref{theorem parallel closed zero}(IIa)}
are in optional brackets.

\yestop
\noindent
\underline{Claim~1:}
If 
\bigmath{
    \redindex{n_1}
    \tight\circ
    \refltransindex{0{[+(n_1\monus1)]}}
}
strongly commutes over
\refltransindex{n_0},
then
\redindex{n_1} and \redindex{n_0} are commuting.
\\
\underline{Proof of Claim~1:}
\bigmath{
    \redindex{n_1}
    \tight\circ
    \refltransindex{0{[+(n_1\monus1)]}}
}
and
\refltransindex{n_0}
are commuting
by Lemma~\ref{lemma strong commutation one copy}.
Since by 
\lemmamonotonicinbeta\
we have
\bigmath{
  \redindex{n_1}
  \subseteq
  \redindex{n_1}
  \tight\circ
  \refltransindex{0{[+(n_1\monus1)]}}
  \subseteq
  \refltransindex{n_1}
,}
now \redindex{n_1} 
and \redindex{n_0} are 
commuting, too.%
\hfill\Qed{Claim~1}

\yestop
\yestop
\noindent
For \math{n_0\preceq n_1\prec\omega} 
we are going to show by induction on \math{n_0\tight+n_1}
the following property\footroom:
\\\LINEmath{
    w_0
    \antiredindex{n_0}
    u
    \redindex{n_1}
    w_1
  \quad\implies\quad
    w_0
    \onlyonceindex{n_1}
    \tight\circ
    \refltransindex{0{[+(n_1\monus1)]}}
    \circ
    \antirefltransindex{n_0}
    w_1
.}
\begin{diagram}
u&&\rredindex{n_1}&&w_1
\\\dredindex{n_0}&&&&\drefltransindex{n_0}
\\w_0&\ronlyonceindex{n_1}
&\circ&\rrefltransindex{0{[+(n_1\monus1)]}}&\circ
\end{diagram}

\yestop
\noindent
\underline{Claim~2:}
Let \math{\delta\prec\omega}.
If
\\
\linemath{
  \forall n_0,n_1\tightprec\omega\stopq
  \inparenthesesoplist{ 
    \inparenthesesoplist{
        n_0\tightpreceq n_1      
      \oplistund
        n_0\tight+ n_1\tightpreceq\delta
    }
    \oplistimplies
    \forall w_0,w_1,u\stopq
    \inparenthesesoplist{
        w_0
        \antiredindex{n_0}
        u
        \redindex{n_1}
        w_1
      \oplistimplies 
        w_0
        \onlyonceindex{n_1}
        \tight\circ
        \refltransindex{0{[+(n_1\monus1)]}}
        \circ
        \antirefltransindex{n_0}
        w_1
    }
  }
,}
then
\\
\linemath{
  \forall n_0,n_1\tightprec\omega\stopq
  \inparenthesesoplist{ 
    \inparenthesesoplist{
        n_0\tightpreceq n_1      
      \oplistund
        n_0\tight+ n_1\tightpreceq\delta
    }
    \oplistimplies 
            \redindex{n_1}
            \tight\circ
            \refltransindex{0{[+(n_1\monus1)]}}
      \mbox{ strongly commutes over }
            \refltransindex{n_0}
  }
,}
and 
\RX\ is \math 0-shallow confluent up to \math{\delta}.
\\
\underline{Proof of Claim~2:}
By induction on \math{\delta} in \tightprec.
First we show the strong commutation.
Assume \math{n_0\preceq n_1\prec\omega} with
\math{n_0\tight+ n_1\tightpreceq\delta}. 
By Lemma~\ref{lemma strong commutation one copy} it suffices to show that
\bigmath{
    \redindex{n_1}
    \tight\circ
    \refltransindex{0{[+(n_1\monus1)]}}
}
strongly commutes over
\redindex{n_0}.
Assume
\bigmath{
    w_0
    \antiredindex{n_0}
    u
    \redindex{n_1}
    w_1
    \refltransindex{0{[+(n_1\monus1)]}}
    w_2
}
(\cf\ diagram below).
By the above property there is some \math{w_1'}
with
\bigmath{
    w_0
    \onlyonceindex{n_1}
    \tight\circ
    \refltransindex{0{[+(n_1\monus1)]}}
    w_1'
    \antirefltransindex{n_0}
    w_1
.}
Next we show that we can close the peak
\bigmath{ 
    w_1'
    \antirefltransindex{n_0}
    w_1
    \refltransindex{0{[+(n_1\monus1)]}}
    w_2
}
according to 
\bigmath{
    w_1'
    \refltransindex{0{[+(n_1\monus1)]}}
    w_2'
    \antirefltransindex{n_0}
    w_2
}
for some \math{w_2'}.
In case of \bigmath{n_1\tightequal0}
this is possible due to \bigmath{w_1\tightequal w_2.}
Otherwise we have 
\bigmath{
  n_0\tight+(0{[\tight+(n_1\monus1)]})
  \tightprec 
  n_0\tight+ n_1
  \tightpreceq
  \delta
} 
and due to our induction hypothesis
(saying that \RX\ is \math 0-shallow confluent up to all 
\math{\delta'\prec\delta
})
this is possible again.
\begin{diagram}
u&&\rredindex{n_1}&&w_1&\rrefltransindex{0{[+(n_1\monus1)]}}&w_2
\\
\dredindex{n_0}&&&&\drefltransindex{n_0}&&\drefltransindex{n_0}
\\
w_0&\ronlyonceindex{n_1}&\circ&\rrefltransindex{0{[+(n_1\monus1)]}}
&w_1'&\rrefltransindex{0{[+(n_1\monus1)]}}&w_2'
\end{diagram}
Finally we show \math 0-shallow confluence up to \math\delta.
Assume
\bigmath{n_0\tight+ n_1\tightpreceq\delta}
and
\bigmath{
    w_0
    \antirefltransindex{n_0}
    u
    \refltransindex    {n_1}
    w_1
.}
Due to symmetry in \math{n_0} and \math{n_1} we may assume
\bigmath{n_0\tightpreceq n_1.}
Above we have shown that 
\bigmath{
    \redindex{n_1}
    \tight\circ
    \refltransindex{0{[+(n_1\monus1)]}}
}
strongly commutes over
\refltransindex{n_0}. 
By Claim~1 we finally get
\bigmath{
  w_0\refltransindex{n_1}\circ\antirefltransindex{n_0} w_1
}
as desired.%
\QED{Claim~2}

\pagebreak

\yestop
\yestop
\noindent
Note that 
for \bigmath{n_0\tightequal0} 
our property follows 
from \bigmath{\antiredindex{n_0}\subseteq\id.}

The benefit of 
Claim~2 is twofold: First, it says that our theorem is valid
if the above property
holds for all \math{n_0\preceq n_1\prec\omega}.
For part (IIb) this is because then
by Lemma~\ref{lemma strong commutation one copy}
\redindex{n_1} strongly commutes over \redindex{n_0} for
all \math{n_0\preceq n_1\prec\omega},
\ie\ \redindex{\omega} strongly commutes over \redindex{n_0},
\ie\ \redindex{\omega} strongly commutes over \redindex{\omega},
\ie\ \redindex{\omega} is strongly confluent.
Second, it strengthens the property when used as induction hypothesis. Thus 
(writing \math{n_i\tight+1} instead of \math{n_i} since we may assume
 \math{0\tightprec n_0\tightpreceq n_1})
it
now suffices to show
for
\math{n_0\preceq n_1\prec\omega}
that 
\\\linemath{
    w_0
    \antiredindex{n_0+1,\bar p_0}
    u
    \redindex    {n_1+1,\bar p_1}
    w_1
}
together with our induction hypotheses 
that\headroom
\\\linemath{\headroom
  \forall\delta\tightprec(n_0\tight+1)\tight+(n_1\tight+1)\stopq
  \mbox{\RX\ is \math 0-shallow confluent up to }\delta
}
\headroom
implies
\\\LINEmath{
    w_0
    \onlyonceindex{n_1+1}
    \tight\circ
    \refltransindex{0{[+n_1]}}
    \circ
    \antirefltransindex{n_0+1}
    w_1
.}
\begin{diagram}
u&&\rredindex{n_1+1,\,\bar p_1}&&w_1
\\
\dredindex{n_0+1,\,\bar p_0}&&&&\drefltransindex{n_0+1}
\\
w_0&\ronlyonceindex{n_1+1}&\circ&\rrefltransindex{0{[+n_1]}}&\circ
\end{diagram}

\yestop
\noindent
Now for each \math{i\prec2} there are
\bigmath{\kurzregelindex{i}\in\R}
and
\bigmath{\mu_{i}\in\Xsubst}
with
\bigmath{u/\bar p_i\tightequal l_{i}\mu_{i},}
\bigmath{
  w_i\tightequal\repl{u}{\bar p_i}{r_{i}\mu_{i}}
,}
\math{C_{i}\mu_{i}} fulfilled \wrt\ \redindex{n_i},
and 
\bigmath{
    l_{i}\tightin\tcs
.}

\yestop
\yestop
\noindent
In case of \bigmath{\neitherprefix{\bar p_0}{\bar p_1}} 
we have 
\bigmath{
  w_{i}/\bar p_{1-i}
  \tightequal
  \repl u{\bar p_{i}}{r_{i}\mu_{i}}/\bar p_{1-i}
  \tightequal
  u/\bar p_{1-i}
  \tightequal
  l_{1-i}\mu_{1-i}
} 
and
therefore
\bigmath{
  w_{i}
  \redindex{n_{i}+1}
  \replpar u{\bar p_{k}}{r_{k}\mu_{k}}{k\tightprec2}
,}
\ie\ our proof is finished.
Thus, according to whether \math{\bar p_0} is a prefix of \math{\bar p_1}
or vice versa, we have the following two cases left:
\pagebreak

\yestop
\yestop
\noindent
\underline{\underline{\underline{%
There is some \math{\bar p_1'} with
\bigmath{\bar p_0\bar p_1'\tightequal\bar p_1} and
\bigmath{\bar p_1'\tightnotequal\emptyset}:%
}}}

\noindent
We have two cases:

\noindent
\underline{\underline{``The variable overlap case'':}}
\\\LINEnomath{\underline{\underline{%
There are \math{x\in\V} and \math{p'}, \math{p''} such that
\math{
  l_{0}/p'\tightequal x
  \und
  p' p''\tightequal\bar p_1'
}:}}}
\begin{diagram}
l_{0}\mu_{0}&&&\rredindex{n_1+1,\,\bar p_1'}&&&w_1/\bar p_0
\\
&&&&&&\dequal
\\
\dredindex{n_0+1,\,\emptyset}&&&&&&l_{0}\nu
\\
&&&&&&\dredindex{n_0+1}
\\
w_0/\bar p_0&\requal&r_{0}\mu_{0}&&\ronlyonceindex{n_1+1}
&&r_{0}\nu
\end{diagram}
\noindent
\underline{Claim~6:}
We have 
\bigmath{
  x\mu_{0}/p''\tightequal l_{1}\mu_{1}
.}
\\\underline{Proof of Claim~6:}
We have 
\bigmath{
  x\mu_{0}/p''
  \tightequal
  l_{0}\mu_{0}/p' p''
  \tightequal
  u/\bar p_0 p' p''
  \tightequal
  u/\bar p_0\bar p_1'
  \tightequal
  u/\bar p_1
  \tightequal
  l_{1}\mu_{1}
.}\QED{Claim~6}

\noindent
\underline{Claim~7:}
We can define \math{\nu\in\Xsubst} by
\bigmath{
  x\nu
  \tightequal
  \repl
    {x\mu_{0}}
    {p''}
    {r_{1}\mu_{1}}
}
and
\bigmath{
  \forall y\tightin\V\tightsetminus\{x\}\stopq 
    y\nu\tightequal y\mu_{0}
.}
Then we have 
\bigmath{
  x\mu_{0}
  \redindex{n_1+1}
  x\nu
.}
\\
\underline{Proof of Claim~7:}
This follows directly from Claim~6.%
\QED{Claim~7}

\noindent
\underline{Claim~8:}
\bigmath{
  l_{0}\nu
  \tightequal
  w_1/\bar p_0
.}
\\
\underline{Proof of Claim~8:}
By the left-linearity assumption of our theorem we may assume
\bigmath{
  \setwith
    {p'''}
    {l_{0}/p'''\tightequal x}
  =
  \{p'\}
.}
Thus, by Claim~7 we get
\bigmath{
  w_1/\bar p_0
  \tightequal
  \repl 
    {u/\bar p_0}
    {\bar p_1'}
    {r_{1}\mu_{1}}
  \tightequal\\
  \repl
    {\replpar
       {l_{0}}
       {p'''}
       {y\mu_{0}}
       {l_{0}/p'''\tightequal y\tightin\V}
    }
    {\bar p_1'}
    {r_{1}\mu_{1}}
  \tightequal\\
  \repl
    {\repl
       {\replpar
          {l_{0}}
          {p'''}
          {y\mu_{0}}
          {l_{0}/p'''\tightequal y\tightin\V\und y\tightnotequal x}
       }
       {p'}
       {x\mu_{0}}
    }
    {p' p''}
    {r_{1}\mu_{1}}
  \tightequal\\
  \repl
    {\replpar
       {l_{0}}
       {p'''}
       {y\nu}
       {l_{0}/p'''\tightequal y\tightin\V\und y\tightnotequal x}
    }
    {p'}
    {\repl
       {x\mu_{0}}
       {p''}
       {r_{1}\mu_{1}}
    }
  \tightequal\\
  \replpar
    {l_{0}}
    {p'''}
    {y\nu}
    {l_{0}/p'''\tightequal y\tightin\V}
  \tightequal
  l_{0}\nu
.}\QED{Claim~8}

\noindent
\underline{Claim~9:}
\bigmath{
  w_0/\bar p_0
  \onlyonceindex{n_1+1}
  r_{0}\nu
.}
\\
\underline{Proof of Claim~9:} 
By the right-linearity assumption of our theorem we may assume 
\bigmath{\CARD{\setwith{p'''}{r_{0}/p'''\tightequal x}}\tightpreceq1.}
Thus by Claim~7 we get:
\bigmath{
  w_0/\bar p_0
  \tightequal 
  r_{0}\mu_{0}
  \tightequal
%  \replpar
%    {r_{0}}
%    {p'''}
%    {y\mu_{0}}
%    {r_{0}/p'''\tightequal y\tightin\V}
%  \tightequal
  \\
  \replpar
    {\replpar
       {r_{0}}
       {p'''}
       {y\mu_{0}}
       {r_{0}/p'''\tightequal y\tightin\V\tightsetminus\{x\}}
    }
    {p'''}
    {x\mu_{0}}
    {r_{0}/p'''\tightequal x}
  \onlyonceindex{n_1+1}
  \\
  \replpar
    {\replpar
       {r_{0}}
       {p'''}
       {y\mu_{0}}
       {r_{0}/p'''\tightequal y\tightin\V\tightsetminus\{x\}}
    }
    {p'''}
    {x\nu}
    {r_{0}/p'''\tightequal x}
  \tightequal
  \\
  \replpar
    {\replpar
       {r_{0}}
       {p'''}
       {y\nu}
       {r_{0}/p'''\tightequal y\tightin\V\tightsetminus\{x\}}
    }
    {p'''}
    {x\nu}
    {r_{0}/p'''\tightequal x}
  \tightequal
  r_{0}\nu
.}%
\QED{Claim~9}

\noindent
By claims 8 and 9 it now suffices to show
\bigmath{
  l_{0}\nu
  \redindex{n_0+1}
  r_{0}\nu
,}
which again follows from 
Lemma~\ref{lemma invariance of fulfilledness}
(matching its \math{n_0} to our \math{n_1\tight+1}
      and its \math{n_1} to our \math{n_0})
since \RX\ is \math 0-quasi-normal and \math 0-shallow confluent up to
\bigmath{(n_1\tight+1)\tight+ n_0}
by our induction hypothesis,
and since
\bigmath{
  \forall y\tightin\V\stopq 
  y\mu_{0}
  \refltransindex{n_1+1}
  y\nu
}
by Claim~7.\QEDdouble{``The variable overlap case''}

\pagebreak

\yestop
\noindent
\underline{\underline{``The critical peak case'':
\math{
  \bar p_1'\tightin\TPOS{l_{0}}
  \und
  l_{0}/\bar p_1'\tightnotin\V
}:}}
\begin{diagram}
l_{0}\mu_{0}&&\rredindex{n_1+1,\,\bar p_1'}&&w_1/\bar p_0
\\
\dredindex{n_0+1,\,\emptyset}&&&&\drefltransindex{n_0+1}
\\
w_0/\bar p_0&\ronlyonceindex{n_1+1}&\circ&\rrefltransindex{0{[+n_1]}}&\circ
\end{diagram}
Let \math{\xi\in\SUBST\V\V} be a bijection with 
\bigmath{
  \xi[\VAR{\kurzregelindex{1}}]\cap\VAR{\kurzregelindex{0}}
  =
  \emptyset
.}
\\
Define
\bigmath{
  \Y
  :=
  \xi[\VAR{\kurzregelindex{1}}]\cup\VAR{\kurzregelindex{0}}
.}
\\
Let \math{\varrho\in\Xsubst} be given by
$\ x\varrho=
\left\{\begin{array}{@{}l@{}l@{}}
  x\mu_{0}        &\mbox{ if }x\in\VAR{\kurzregelindex{0}}\\
  x\xi^{-1}\mu_{1}&\mbox{ else}\\
\end{array}\right\}
\:(x\tightin\V)$.
\\
By
\math{
  l_{1}\xi\varrho
  \tightequal 
  l_{1}\xi\xi^{-1}\mu_{1}
  \tightequal 
  u/\bar p_1
  \tightequal 
  u/\bar p_0\bar p_1'
  \tightequal 
  l_{0}\mu_{0}/\bar p_1'  
  \tightequal 
  l_{0}\varrho/\bar p_1'
  \tightequal 
  (l_{0}/\bar p_1')\varrho
}
\\
let
\math{
  \sigma:=\minmgu{\{(l_{1}\xi}{l_{0}/\bar p_1')\},\Y}
}
and
\math{\varphi\in\Xsubst}
with
\math{
  \domres{\inpit{\sigma\varphi}}\Y
  \tightequal
  \domres\varrho\Y
}.
\\
If 
\bigmath{
  \repl{l_{0}}{\bar p_1'}{r_{1}\xi}\sigma
  \tightequal
  r_{0}\sigma
,}
then the proof is finished due to 
\\\LINEmath{
  w_0/\bar p_0
  \tightequal
  r_{0}\mu_{0}
  \tightequal
  r_{0}\sigma\varphi
  \tightequal
  \repl{l_{0}}{\bar p_1'}{r_{1}\xi}\sigma\varphi
  \tightequal
  \repl{l_{0}\mu_{0}}{\bar p_1'}{r_{1}\mu_{1}}
  \tightequal
  w_1/\bar p_0
.}
\\
Otherwise 
we have
\bigmath{
  (\,
   (\repl{l_{0}}{\bar p_1'}{r_{1}\xi},
    C_{1}\xi,    
    0),\penalty-1\,
   (r_{0},
    C_{0},
    0),\penalty-1\,
    l_{0},\penalty-1\,
    \sigma,\penalty-1\,
    \bar p_1'\,)
  \in{\rm CP}(\R)
;}
\bigmath{\bar p_1'\tightnotequal\emptyset}
(due the global case assumption);
\bigmath{C_{1}\xi\sigma\varphi=C_{1}\mu_{1}}
is fulfilled \wrt\ \redindex{n_1};
\bigmath{C_{0}\sigma\varphi=C_{0}\mu_{0}}
is fulfilled \wrt\ \redindex{n_0}.
Since 
\bigmath{
  \forall\delta\tightprec(n_1\tight+1)\tight+(n_0\tight+1)\stopq
    \RX\mbox{ is \math 0-shallow confluent up to }\delta
}
(by our induction hypothesis),
due to our assumed \math 0-shallow {[noisy]}  anti-closedness
(matching the definition's \math{n_0} to our \math{n_1\tight+1} 
                  and its \math{n_1} to \math{n_0\tight+1})
we have
\bigmath{
  w_1/\bar p_0
  \tightequal
  \repl{l_{0}\mu_{0}}{\bar p_1'}{r_{1}\mu_{1}}
  \tightequal
  \repl{l_{0}}{\bar p_1'}{r_{1}\xi}\sigma\varphi
  \refltransindex{n_0+1}
  \circ
  \antirefltransindex{0{[+n_1]}}
  \tight\circ
  \antionlyonceindex{n_1+1}
  r_{0}\sigma\varphi
  \tightequal
  r_{0}\mu_{0}
  \tightequal
  w_0/\bar p_0.
}
\\
\Qeddouble{``The critical peak case''}%
\QEDtriple{``There is some \math{\bar p_1'} with
\bigmath{\bar p_0\bar p_1'\tightequal\bar p_1} and
\bigmath{\bar p_1'\tightnotequal\emptyset}''}

\pagebreak

\yestop
\yestop
\noindent
\underline{\underline{\underline{%
There is some \math{\bar p_0'} with
\bigmath{\bar p_1\bar p_0'\tightequal\bar p_0}:%
}}}

\noindent
We have two cases:

\noindent
\underline{\underline{%
``The second variable overlap case'':%
}}
\\\LINEnomath{\underline{\underline{%
There are \math{x\tightin\V} and \math{p'}, \math{p''} such that
\math{
    l_{1}/p'
    \tightequal
    x
  \und
    p'p''
    \tightequal
    \bar p_0'
}:%
}}}
\begin{diagram}
l_{1}\mu_{1}&&&\rredindex{n_1+1,\,\emptyset}&&&w_1/\bar p_1
\\&&&&&&\dequal
\\\dredindex{n_0+1,\,\bar p_0'}&&&&&&r_{1}\mu_{1}
\\&&&&&&\dredparaindex{n_0+1}
\\w_0/\bar p_1&\requal&l_{1}\nu&&\rredindex{n_1+1}&&r_{1}\nu
\end{diagram}
\noindent
\underline{Claim~11a:}
We have 
\bigmath{
  x\mu_{1}/p''\tightequal l_{0}\mu_{0}
.}
\\\underline{Proof of Claim~11a:}
We have 
\bigmath{
  x\mu_{1}/p''
  \tightequal
  l_{1}\mu_{1}/p' p''
  \tightequal
  u/\bar p_1 p' p''
  \tightequal
  u/\bar p_1\bar p_0'
  \tightequal
  u/\bar p_0
  \tightequal
  l_{0}\mu_{0}
.}\QED{Claim~11a}

\noindent
\underline{Claim~11b:}
We can define \math{\nu\in\Xsubst} by
\bigmath{
  x\nu
  \tightequal
  \repl
    {x\mu_{1}}
    {p''}
    {r_{0}\mu_{0}}
}
and
\bigmath{
  \forall y\tightin\V\tightsetminus\{x\}\stopq y\nu\tightequal y\mu_{1}
.}
Then we have 
\bigmath{
  x\mu_{1}\redindex{n_0+1}x\nu
.}
\\
\underline{Proof of Claim~11b:}
This follows directly from Claim~11a.%
\QED{Claim~11b}

\noindent
\underline{Claim~12:}
\bigmath{
  w_0/\bar p_1
  \tightequal
  l_{1}\nu
.}
\\
\underline{Proof of Claim~12:}
\\
By the left-linearity assumption of our theorem we may assume
\bigmath{
  \setwith
    {p'''}
    {l_{1}/p'''\tightequal x}
  =
  \{p'\}
.}
Thus, by Claim~11b we get
\bigmath{
  w_0/\bar p_1
  \tightequal
  \repl 
    {u/\bar p_1}
    {\bar p_0'}
    {r_{0}\mu_{0}}
  \tightequal\\
  \repl
    {\replpar
       {l_{1}}
       {p'''}
       {y\mu_{1}}
       {l_{1}/p'''\tightequal y\tightin\V}
    }
    {\bar p_0'}
    {r_{0}\mu_{0}}
  \tightequal\\
  \repl
    {\repl
       {\replpar
          {l_{1}}
          {p'''}
          {y\mu_{1}}
          {l_{1}/p'''\tightequal y\tightin\V\und y\tightnotequal x}
       }
       {p'}
       {x\mu_{1}}
    }
    {p' p''}
    {r_{0}\mu_{0}}
  \tightequal\\
  \repl
    {\replpar
       {l_{1}}
       {p'''}
       {y\nu}
       {l_{1}/p'''\tightequal y\tightin\V\und y\tightnotequal x}
    }
    {p'}
    {\repl
       {x\mu_{1}}
       {p''}
       {r_{0}\mu_{0}}
    }
  \tightequal\\
  \replpar
    {l_{1}}
    {p'''}
    {y\nu}
    {l_{1}/p'''\tightequal y\tightin\V}
  \tightequal
  l_{1}\nu
.}\QED{Claim~12}

\noindent
\underline{Claim~13:}
\bigmath{
  r_{1}\nu
  \antiredparaindex{n_0+1}
  w_1/\bar p_1
.}
\\
\underline{Proof of Claim~13:} 
Since \bigmath{r_{1}\mu_{1}\tightequal w_1/\bar p_1,} 
this follows directly from Claim~11b.%
\QED{Claim~13}

\noindent
By claims 12 and 13 
using Corollary~\ref{corollary parallel one} 
it now suffices to show
\bigmath{
  l_{1}\nu
  \redindex{n_1+1}
  r_{1}\nu
,}
which again follows from 
Claim~11b,
Lemma~\ref{lemma invariance of fulfilledness}
(matching 
 its \math{n_0} to our \math{n_0\tight+1} and 
 its \math{n_1} to our \math{n_1}),
and our induction hypothesis that \RX\ is \math 0-shallow confluent up to
\bigmath{
  (n_0\tight+1)\tight+ n_1
.}%
\\\Qeddouble{``The second variable overlap case''}

\pagebreak

\yestop
\noindent
\underline{\underline{``The second critical peak case'':
\math{
    \bar p_0'\tightin\TPOS{l_{1}}
  \und
    l_{1}/\bar p_0'\tightnotin\V
}:%
}}
\begin{diagram}
l_{1}\mu_{1}&&\rredindex{n_1+1,\,\emptyset}&&w_1/\bar p_1
\\
\dredindex{n_0+1,\,\bar p_0'}&&&&\drefltransindex{n_0+1}
\\
w_0/\bar p_1&\ronlyonceindex{n_1+1}&\circ&\rrefltransindex{0{[+n_1]}}&\circ
\end{diagram}
Let \math{\xi\in\SUBST\V\V} be a bijection with 
\bigmath{
  \xi[\VAR{\kurzregelindex{0}}]\cap\VAR{\kurzregelindex{1}}
  =
  \emptyset
.}
\\
Define
\bigmath{
  \Y
  :=
  \xi[\VAR{\kurzregelindex{0}}]\cup\VAR{\kurzregelindex{1}}
.}
\\
Let \math{\varrho\in\Xsubst} be given by
$\ x\varrho=
\left\{\begin{array}{@{}l@{}l@{}}
  x\mu_{1}        &\mbox{ if }x\in\VAR{\kurzregelindex{1}}\\
  x\xi^{-1}\mu_{0}&\mbox{ else}\\
\end{array}\right\}
\:(x\tightin\V)$.
\\
By
\math{
  l_{0}\xi\varrho
  \tightequal 
  l_{0}\xi\xi^{-1}\mu_{0}
  \tightequal 
  u/\bar p_0
  \tightequal 
  u/\bar p_1\bar p_0'
  \tightequal 
  l_{1}\mu_{1}/\bar p_0'  
  \tightequal 
  l_{1}\varrho/\bar p_0'
  \tightequal 
  (l_{1}/\bar p_0')\varrho
}
\\
let
\math{
  \sigma:=\minmgu{\{(l_{0}\xi}{l_{1}/\bar p_0')\},\Y}
}
and
\math{\varphi\in\Xsubst}
with
\math{
  \domres{\inpit{\sigma\varphi}}\Y
  \tightequal
  \domres\varrho\Y
}.
\\
If 
\bigmath{
  \repl{l_{1}}{\bar p_0'}{r_{0}\xi}\sigma
  \tightequal
  r_{1}\sigma
,}
then the proof is finished due to 
\\\linemath{
  w_0/\bar p_1
  \tightequal
  \repl{l_{1}\mu_{1}}{\bar p_0'}{r_{0}\mu_{0}}
  \tightequal
  \repl{l_{1}}{\bar p_0'}{r_{0}\xi}\sigma\varphi
  \tightequal
  r_{1}\sigma\varphi
  \tightequal
  r_{1}\mu_{1}
  \tightequal
  w_1/\bar p_1.
}
Otherwise 
we have
\bigmath{
  (\,
   (\repl{l_{1}}{\bar p_0'}{r_{0}\xi},
    C_{0}\xi,    
    0),\penalty-1\,
   (r_{1},
    C_{1},
    0),\penalty-1\,
    l_{1},\penalty-1\,
    \sigma,\penalty-1\,
    \bar p_0'\,)
  \in{\rm CP}(\R)
;}
\bigmath{C_{0}\xi\sigma\varphi=C_{0}\mu_{0}}
is fulfilled \wrt\ \redindex{n_0};
\bigmath{C_{1}\sigma\varphi=C_{1}\mu_{1}}
is fulfilled \wrt\ \redindex{n_1}.
Since 
\bigmath{
  \forall\delta\tightprec(n_0\tight+1)\tight+(n_1\tight+1)\stopq
  \mbox{\RX\ is \math 0-shallow confluent up to }\delta
}
(by our induction hypothesis) 
due to our assumed \math 0-shallow 
{[noisy]}  strong joinability 
(matching the definition's \math{n_0} to our \math{n_0\tight+1}
                   and its \math{n_1} to our \math{n_1\tight+1})
we have
\bigmath{
  w_0/\bar p_1
  \tightequal
  \repl{l_{1}\mu_{1}}{\bar p_0'}{r_{0}\mu_{0}}
  \tightequal
  \repl{l_{1}}{\bar p_0'}{r_{0}\xi}\sigma\varphi
  \onlyonceindex{n_1+1}
  \tight\circ
  \refltransindex{0{[+n_1]}}
  \circ
  \antirefltransindex{n_0+1}
  \penalty-1
  r_{1}\sigma\varphi
  \tightequal
  r_{1}\mu_{1}
  \tightequal
  w_1/\bar p_1
.}
\\\Qeddouble{``The second critical peak case''}
\end{proofparsepqed}

\begin{proofqed}{Theorem~\ref{theorem complementary zero}}
Due to Corollary~\ref
{corollary zero shallow confluent implies confluent no termination}
it suffices to show that the conditions of Theorem~\ref
{theorem parallel closed zero}
are satisfied.
Since \RCONS\ is normal, \RX\ is \math 0-quasi-normal.
Thus we only have to show that the conjunctive condition lists 
of the \math 0-shallow joinability notions are never satisfied
for critical peaks of the form \math{(0,0)}.
Thus, assume 
\math{\varphi\in\Xsubst} and
\math{n_0,n_1\prec\omega} such
that
\bigmath{
  \forall i\tightprec 2\stopq
    \inparenthesesinline{
         D_i\varphi\mbox{ fulfilled \wrt\ } \redindex{\RX,n_i\monus 1}
    }
}
and
\bigmath{
    \forall\delta\tightprec n_0\tight+ n_1\stopq
    \inparenthesesinline{
           \RX\mbox{ is \math 0-shallow confluent up to }\delta
    }
.}
By the assumed complementarity there must be complementary 
equation literals in \math{D_0} and \math{D_1}. Due to our symmetry in
\math 0 and \math 1 so far, we may \wrog\ assume that 
\math{(u\boldequal   v)} occurs in \math{D_0} and 
\math{(u\boldunequal v)} occurs in \math{D_1}
or else that 
\math{(p\boldequal\truepp )} occurs in \math{D_0} and 
\math{(p\boldequal\falsepp)} occurs in \math{D_1}.
Since negative conditions are not allowed for constructor rules 
we must be in the latter case here.
Due to the definition of complementarity,
\truepp\ and \falsepp\ are distinct irreducible ground terms.
Thus we have
\bigmath{
  p\varphi
  \refltransindex{n_0\monus 1}
  \truepp
}
and
\bigmath{
  p\varphi
  \refltransindex{n_1\monus 1}
  \falsepp
.}
In case of \bigmath{n_0,n_1\tightpreceq 1} this 
implies the contradicting
\bigmath{
  \truepp
  \tightequal
  p\varphi
  \tightequal
  \falsepp
.}
Otherwise, in case of \bigmath{n_0\tightsucceq 1} we have 
\bigmath{
  (n_0\monus 1)\tight+ (n_1\monus 1)
  \prec
  n_0\tight+ n_1
}
and thus by our above assumption
\RX\ is \math 0-shallow confluent up to 
\math{(n_0\monus 1)\tight+(n_1\monus 1)}.
This implies the contradicting
\bigmath{
  \truepp
  \downarrow
  \falsepp
.}
\end{proofqed}

\begin{proof}{Theorem~\ref{theorem quasi-free zero}}
\underline{1\implies 2:}
Directly by Lemma~\ref{lemma quasi-free two and zero}.
\underline{2\implies 1:}
Directly by Lemma~\ref{lemma omega shallow joinablility necessary}. 
\end{proof}

\pagebreak

\begin{proofparsepqed}{Lemma~\ref{lemma parallel closed first level two}}
For \math{n\prec\omega} we are going to show by induction on \math n
the following property\footroom:
\\\LINEmath{
    w_0
    \antiredparaindex\omega
    u
    \redparaindex{\omega+n}
    w_1
  \quad\implies\quad
    w_0
    \redparaindex{\omega+n}
    \tight\circ
    \refltransindex{\omega[+(n\monus 1)]}
    \circ
    \antirefltransindex\omega
    w_1
.}
\begin{diagram}
u&&\rredparaindex{\omega+n}&&w_1\\
\dredparaindex{\omega}&&&&\drefltransindex{\omega}\\
w_0&\rredparaindex{\omega+n}&\circ&
\rrefltransindex{\omega[+(n\monus 1)]}&\circ\\
\end{diagram}

\noindent
\underline{Claim~1:}
If the above property holds for a fixed \math{n\prec\omega}, 
and 
\\\bigmath{
  \forall k\tightprec n\stopq
     (\RX\mbox{ is \math\omega-shallow confluent up to }k)
,} then
\bigmath{
    \redparaindex{\omega+n}
    \tight\circ
    \refltransindex{\omega[+(n\monus 1)]}
}
strongly commutes over
\refltransindex\omega.
\\
\underline{Proof of Claim~1:}
By Lemma~\ref{lemma strong commutation one copy} it suffices to show that
\bigmath{
    \redparaindex{\omega+n}
    \tight\circ
    \refltransindex{\omega[+(n\monus 1)]}
}
strongly commutes over
\redindex{\omega}.
Assume
\bigmath{
    w_0
    \antiredindex\omega
    u
    \redparaindex{\omega+n}
    w_1
    \refltransindex{\omega[+(n\monus 1)]}
    w'
}
(\cf\ diagram below).
By the above property there is some \math{v'}
with
\bigmath{
    w_0
    \redparaindex{\omega+n}
    \tight\circ
    \refltransindex{\omega[+(n\monus 1)]}
    v'
    \antirefltransindex\omega
    w_1
.}
We only have to show that we can close the peak
\bigmath{ 
    v'
    \antirefltransindex\omega
    w_1
    \refltransindex{\omega[+(n\monus 1)]}
    w'
}
according to 
\bigmath{
    v'
    \refltransindex{\omega[+(n\monus 1)]}
    \circ
    \antirefltransindex\omega
    w'
.}
{[In case of \bigmath{n\tightequal0:}]}
This is possible due to confluence of \redindex\omega.
{[Otherwise we have \bigmath{n\monus1\tightprec n} and due to the 
assumed \math\omega-shallow confluence up to \math{n\monus1}
this is possible again.]}
\begin{diagram}
u&&\rredparaindex{\omega+n}&&w_1&\rrefltransindex{\omega[+(n\monus 1)]}&w'\\
\dredindex{\omega}&&&&\drefltransindex{\omega}
&&\drefltransindex{\omega}\\
w_0&\rredparaindex{\omega+n}&\circ&
\rrefltransindex{\omega[+(n\monus 1)]}&v'
&\rrefltransindex{\omega[+(n\monus 1)]}&\circ\\
\end{diagram}
\Qed{Claim~1}

\yestop
\noindent
\underline{Claim~2:}
If the above property holds for a fixed \math{n\prec\omega}, and 
\\\bigmath{
  \forall k\tightprec n\stopq
     (\RX\mbox{ is \math\omega-shallow confluent up to }k)
,} 
then
\redindex{\omega+n} and \redindex\omega\ are commuting.
\\
\underline{Proof of Claim~2:}
\bigmath{
    \redparaindex{\omega+n}
    \tight\circ
    \refltransindex{\omega[+(n\monus 1)]}
}
and
\refltransindex\omega\
are commuting
by Lemma~\ref{lemma strong commutation one copy}
and Claim~1.
Since by Corollary~\ref{corollary parallel one}
and \lemmamonotonicinbeta\
we have
\bigmath{
  \redindex{\omega+n}
  \subseteq
  \redparaindex{\omega+n}\tight\circ\refltransindex{\omega[+(n\monus 1)]}
  \subseteq
  \refltransindex{\omega+n}
,}
now \redindex{\omega+n} and \redindex\omega\ are commuting, too.\QED{Claim~2}

\yestop
\noindent
\underline{Claim~3:}
If the above property holds for all 
\math{n\preceq m} for some \math{m\prec\omega}, then
\RX\ is \math\omega-shallow confluent up to \math{m}.
\\
\underline{Proof of Claim~3:}
By induction on \math{m} in \tightprec.
Assume
\bigmath{i\plusomega n\tightpreceq m}
and
\bigmath{
    w_0
    \antirefltransindex{\omega+i}
    u
    \refltransindex    {\omega+n}
    w_1
.}
By definition of `\plusomega' and 
\bigmath{i\plusomega n\tightprec\omega}
\wrog\ we have 
\bigmath{i\tightequal0}
and
\bigmath{n\tightpreceq m.}
By Claim~2 and our induction hypothesis we finally get
\bigmath{
  w_0\refltransindex{\omega+n}\circ\antirefltransindex{\omega} w_1
}
as desired.\QED{Claim~3}

\pagebreak

\yestop
\yestop
\noindent
Note that our property for is trivial for \bigmath{n\tightequal0} since
then by Corollary~\ref{corollary parallel one} we have
\bigmath{
  \redparaindex{\omega+n}=\redparaindex{\omega}\subseteq\refltransindex\omega
}
and \redindex\omega\ is confluent.

The benefit of 
claims 1 and 3 is twofold: 
First, they say that our lemma is valid if the above property
holds for all \math{n\prec\omega}.
Second, they strengthen the property when used as induction hypothesis. Thus 
(writing \math{n\tight+1} instead of \math{n} since we may assume
 \math{0\tightprec n})
it
now suffices to show
for
\math{
  n\prec\omega
}
that 
\\\linemath{
    w_0
    \antiredparaindex{\omega,\Pi_0}
    u
    \redparaindex{\omega+n+1,\Pi_1}
    w_1
}
together with our induction hypothesis 
that\headroom
\\\linenomath{\headroom
  \RX\ is \math\omega-shallow confluent up to \math n
}
\headroom
implies
\\\LINEmath{
    w_0
    \redparaindex{\omega+n+1}
    \tight\circ
    \refltransindex{\omega[+n]}
    \circ
    \antirefltransindex\omega
    w_1
.}
\begin{diagram}
u&&\rredparaindex{\omega+n+1,\,\Pi_1}&&w_1
\\\dredparaindex{\omega,\,\Pi_0}&&&&\drefltransindex{\omega}
\\w_0&\rredparaindex{\omega+n+1}&\circ&\rrefltransindex{\omega[+n]}&\circ
\end{diagram}

\noindent
\Wrog\ let the positions of \math{\Pi_0} (and \math{\Pi_1}) be maximal
in the sense that for any \math{p\in\Pi_0} (or else \math{p\in\Pi_1})
and \math{\Xi\subseteq \TPOS u\tightcap(p\N^+)}
we do not have 
\bigmath{
  w_0\antiredparaindex{\omega,(\Pi_0\setminus\{p\})\cup\Xi}u
}
(or else \math{\ 
  u\redparaindex{\omega+n+1,(\Pi_1\setminus\{p\})\cup\Xi}w_1
})
anymore.
Then for each \math{i\prec2} and
\math{p\in\Pi_i} there are
\bigmath{\kurzregelindex{i,p}\in\R}
and
\bigmath{\mu_{i,p}\in\Xsubst}
with
\bigmath{u/p\tightequal l_{i,p}\mu_{i,p},}
\bigmath{r_{i,p}\mu_{i,p}\tightequal w_i/p.}
Moreover, for each \math{p\in\Pi_0}:
\bigmath{l_{0,p}\tightin\tcs} and
\math{\ C_{0,p}\mu_{0,p}} is fulfilled \wrt\ \redindex{\omega}.
Similarly,
for each \math{p\in\Pi_1}:
\math{\ C_{1,p}\mu_{1,p}} is fulfilled \wrt\ \redindex{\omega+n}.
Finally, for each \math{i\prec2}:
\bigmath{
  w_i\tightequal\replpar{u}{p}{r_{i,p}\mu_{i,p}}{p\tightin\Pi_i}
.}

\pagebreak

\yestop
\noindent
\underline{Claim~5:}
We may assume 
\bigmath{
  \forall p\tightin\Pi_1\stopq
    l_{1,p}\tightnotin\tcs
.}
\\
\underline{Proof of Claim~5:}
Define \math{\Xi:=\setwith{p\tightin\Pi_1}{l_{1,p}\tightin\tcs}}
and
\math{
  u'
  :=
  \replpar{u}{p}{r_{1,p}\mu_{1,p}}{p\tightin\Pi_1\tightsetminus\Xi}
}.
If we have succeeded with our proof under the assumption of Claim~5,
then we have shown 
\bigmath{
  w_0 
  \redparaindex{\omega+n+1}
  \tight\circ
  \refltransindex{\omega[+n]}
  v'
  \antirefltransindex\omega
  u'
}
for some \math{v'}
(\cf\ diagram below).
By Lemma~\ref{lemma invariance of fulfilledness two} 
(matching both its \math\mu\ and \math\nu\ to our \math{\mu_{1,p}})
we get
\bigmath{
  \forall p\tightin\Xi\stopq
    l_{1,p}\mu_{1,p}\redindex\omega r_{1,p}\mu_{1,p}
.}
Thus from
\bigmath{
  v'
  \antirefltransindex\omega
  u'
  \refltransindex\omega
  w_1
}
we get 
\bigmath{
%  w_0 
%  \redparaindex{\omega+n+1}
%  \tight\circ
%  \refltransindex{\omega[+n]}
  v'
  \refltransindex\omega
  \circ
  \antirefltransindex\omega
  w_1
}
by confluence of \redindex\omega.
\begin{diagram}
u&&\rredparaindex{\omega+n+1,\,\Pi_1\setminus\Xi}
&&u'
&\rrefltransindex{\omega}&w_1
\\\dredparaindex{\omega,\,\Pi_0}&&
&&\drefltransindex{\omega}
&&\drefltransindex{\omega}
\\w_0&\rredparaindex{\omega+n+1}&\circ
&\rrefltransindex{\omega[+n]}&v'
&\rrefltransindex{\omega}&\circ
\end{diagram}
\Qed{Claim~5}

\yestop
\noindent
Define
the set of inner overlapping positions by
\\\linemath{
  \displaystyle
  \Omega(\Pi_0,\Pi_1)
  :=
  \bigcup_{i\prec2}
    \setwith
      {p\tightin\Pi_{1-i}}
      {\exists q\tightin\Pi_i\stopq\exists q'\tightin\N^\ast\stopq
        p\tightequal q q'
      }
,}
and the length of a term by
\bigmath{\lambda(\anonymousfpp{t_0}{t_{m-1}}):=1+\sum_{j\prec m}\lambda(t_j).}

\yestop
\noindent
Now we start a second level of induction on
\bigmath{  \displaystyle
  \sum_{p'\in\Omega(\Pi_0,\Pi_1)}\lambda(u/p')
}
in \tightprec.

\yestop
\noindent
Define the set of top positions by
\\\linemath{
  \displaystyle
  \Theta
  :=
      \setwith
      {p\tightin\Pi_0\tightcup\Pi_1}
      {\neg\exists q\tightin\Pi_0\tightcup\Pi_1\stopq
           \exists q'\tightin\N^+\stopq
             p\tightequal q q'
      }
.}
Since the prefix ordering is wellfounded we have
\bigmath{
  \forall i\tightprec2\stopq
  \forall p\tightin\Pi_i\stopq
  \exists q\tightin\Theta\stopq
  \exists q'\tightin\N^\ast\stopq
    p\tightequal q q'
.}
Then
\bigmath{
  \forall i\tightprec2\stopq
  w_i
  \tightequal
  \replpar{w_i}{q}{w_i/q}{q\tightin\Theta}
  \tightequal
  \replpar{\replpar{u}{p}{r_{i,p}\mu_{i,p}}{p\tightin\Pi_i}}
          {q}{w_i/q}{q\tightin\Theta}
  \tightequal
  \replpar{u}{q}{w_i/q}{q\tightin\Theta}
.}
Thus, it now suffices to show for all \math{q\in\Theta}
\\\linemath{\headroom\footroom
    w_0/q
    \redparaindex{\omega+n+1}
    \tight\circ
    \refltransindex{\omega[+n]}
    \circ
    \antirefltransindex\omega
    w_1/q
}
because then we have 
\\\LINEmath{
  w_0
  \tightequal
  \replpar{u}{q}{w_0/q}{q\tightin\Theta}
    \redparaindex{\omega+n+1}
    \tight\circ
    \refltransindex{\omega[+n]}
    \circ
    \antirefltransindex\omega
  \replpar{u}{q}{w_1/q}{q\tightin\Theta}
  \tightequal
  w_1
.}

\noindent
Therefore we are left with the following two cases for \math{q\in\Theta}:

\pagebreak

\yestop
\yestop
\noindent
\underline{\underline{\underline{\math{q\tightnotin\Pi_1}:}}}
Then \bigmath{q\tightin\Pi_0.}
Define \math{\Pi_1':=\setwith{p}{q p\tightin\Pi_1}}.
We have two cases:

\noindent
\underline{\underline{``The variable overlap (if any) case'':
\math{
  \forall p\tightin\Pi_1'\tightcap\TPOS{l_{0,q}}\stopq
    l_{0,q}/p\tightin\V
}:}}
\begin{diagram}
l_{0,q}\mu_{0,q}&&&\rredparaindex{\omega+n+1}&&&w_1/q
\\&&&&&&\drefltransindex{\omega}
\\\dredindex{\omega,\,\emptyset}&&&&&&l_{0,q}\nu
\\&&&&&&\dredindex{\omega}
\\w_0/q&\requal&r_{0,q}\mu_{0,q}&\rredparaindex{\omega+n+1}&\circ
&\rrefltransindex{\omega}&r_{0,q}\nu
\end{diagram}
Define a function \math\Gamma\ on \V\ by (\math{x\tightin\V}):
\bigmath{
  \Gamma(x):=
  \setwith{(p',p'')}
          {l_{0,q}/p'\tightequal x\ \wedge\ p' p''\in\Pi_1'}
.}

\noindent
\underline{Claim~7:}
There is some \math{\nu\in\Xsubst} with
\\\LINEmath{
  \forall x\in\V\stopq
    \inparenthesesoplist{
       x\mu_{0,q}
       \redparaindex{\omega+n+1}
       \tight\circ
       \refltransindex\omega
       x\nu
    \oplistund
       \forall p'\tightin\DOM{\Gamma(x)}\stopq
         x\nu
         \antirefltransindex\omega
         \replpar
           {x\mu_{0,q}}
           {p''}
           {r_{1,q p' p''}\mu_{1,q p' p''}}
           {(p',p'')\tightin\Gamma(x)}
    }
.}
\\
\underline{Proof of Claim~7:}
\\
In case of \bigmath{\DOM{\Gamma(x)}\tightequal\emptyset} we define
\bigmath{x\nu:=x\mu_{0,q}.}
If there is some \math{p'} such that 
\bigmath{\DOM{\Gamma(x)}\tightequal\{p'\}}
we define 
\bigmath{
  x\nu
  :=
  \replpar
    {x\mu_{0,q}}{p''}{r_{1,q p' p''}\mu_{1,q p' p''}}{(p',p'')\tightin\Gamma(x)}
.}
This is appropriate since due to 
\bigmath{
  \forall(p',p'')\tightin\Gamma(x)\stopq
    x\mu_{0,q}/p''
    \tightequal 
    l_{0,q}\mu_{0,q}/p' p''
    \tightequal 
    u/q p' p''
    \tightequal 
    l_{1,q p' p''}\mu_{1,q p' p''}
}
we have
\\\LINEmath{
  \begin{array}{l@{}l@{}l}
  x\mu_{0,q}&
  \tightequal&
  \replpar
    {x\mu_{0,q}}
    {p''}
    {l_{1,q p' p''}\mu_{1,q p' p''}}
    {(p',p'')\tightin\Gamma(x)}
  \redparaindex{\omega+n+1}\\&&
  \replpar
     {x\mu_{0,q}}
     {p''}
     {r_{1,q p' p''}\mu_{1,q p' p''}}
     {(p',p'')\tightin\Gamma(x)}
  \tightequal
  x\nu.  
  \end{array}
}
\\
Finally, in case of \bigmath{\CARD{\DOM{\Gamma(x)}}\succ1,} \math{l_{0,q}} is
not linear in \math x. By the conditions of our lemma, this implies
\bigmath{x\tightin\Vcons.}
Therefore 
\bigmath{x\mu_{0,q}\tightin\tcc.}
Together with
\\ 
\bigmath{
  \forall p'\tightin\DOM{\Gamma(x)}\stopq
  x\mu_{0,q}
  \refltransindex{\omega+n+1}
  \replpar
    {x\mu_{0,q}}{p''}{r_{1,q p' p''}\mu_{1,q p' p''}}{(p',p'')\tightin\Gamma(x)}
}
this implies
\\
\bigmath{
  \forall p'\tightin\DOM{\Gamma(x)}\stopq
  x\mu_{0,q}
  \refltransindex{\omega}
  \replpar
    {x\mu_{0,q}}{p''}{r_{1,q p' p''}\mu_{1,q p' p''}}{(p',p'')\tightin\Gamma(x)}
  \in\tcc
}
\\
by \lemmaconskeeping.
By confluence of \redindex{\omega} 
and \lemmaconskeeping\ again, there is some \math{t\in\tcc} with
\\
\bigmath{
  \forall p'\tightin\DOM{\Gamma(x)}\stopq
  \replpar
    {x\mu_{0,q}}{p''}{r_{1,q p' p''}\mu_{1,q p' p''}}{(p',p'')\tightin\Gamma(x)}
  \refltransindex{\omega}
  t
.}
Therefore we can define \bigmath{x\nu:=t} in this case.
This is appropriate since by 
\bigmath{
  \exists p'\tightin\DOM{\Gamma(x)}\stopq
  x\mu_{0,q}
  \refltransindex{\omega}
  \penalty-1
  \replpar
    {x\mu_{0,q}}{p''}{r_{1,q p' p''}\mu_{1,q p' p''}}{(p',p'')\tightin\Gamma(x)}
  \refltransindex{\omega}
  x\nu
}
we have
\bigmath{
  x\mu_{0,q}
  \refltransindex{\omega}
  x\nu  
.}\QED{Claim~7}

\noindent
\underline{Claim~8:}
\bigmath{
  l_{0,q}\nu
  \antirefltransindex{\omega}
  w_1/q
.}
\\
\underline{Proof of Claim~8:}
\\
By Claim~7 we get
\bigmath{
  w_1/q
  \tightequal
  \replpar 
    {u/q}
    {p' p''}
    {r_{1,q p' p''}\mu_{1,q p' p''}}
    {\exists x\tightin\V\stopq(p',p'')\tightin\Gamma(x)}
  \tightequal\\
  \replpar
    {\replpar
       {l_{0,q}}
       {p'}
       {x\mu_{0,q}}
       {l_{0,q}/p'\tightequal x\tightin\V}
    }
    {p' p''}
    {r_{1,q p' p''}\mu_{1,q p' p''}}
    {\exists x\tightin\V\stopq(p',p'')\tightin\Gamma(x)}
  \tightequal\\
  \replpar
    {l_{0,q}}
    {p'}
    {\replpar
       {x\mu_{0,q}}
       {p''}
       {r_{1,q p' p''}\mu_{1,q p' p''}}
       {(p',p'')\tightin\Gamma(x)}}
    {l_{0,q}/p'\tightequal x\tightin\V}
  \refltransindex\omega\\
  \replpar
    {l_{0,q}}
    {p'}
    {x\nu}
    {l_{0,q}/p'\tightequal x\tightin\V}
  \tightequal
  l_{0,q}\nu
.}\QED{Claim~8}

\noindent
\underline{Claim~9:}
\bigmath{
  w_0/q
  \redparaindex{\omega+n+1}
  \tight\circ
  \refltransindex\omega
  r_{0,q}\nu
.}
\\
\underline{Proof of Claim~9:} 
Since 
\bigmath{
  w_0/q
  \tightequal 
  r_{0,q}\mu_{0,q}
,} 
this follows directly from Claim~7.%
\QED{Claim~9}

\noindent
By claims 8 and 9 it now suffices to show
\bigmath{
  r_{0,q}\nu
  \antiredindex{\omega}
  l_{0,q}\nu
,}
which again follows from 
Lemma~\ref{lemma invariance of fulfilledness two}
since 
\bigmath{\forall x\tightin\V\stopq x\mu_{0,q}\refltransindex{\omega+n+1}x\nu}
by Claim~7 and Corollary~\ref{corollary parallel one}.%
\QEDdouble{``The variable overlap (if any) case''}

\pagebreak

\yestop
\noindent
\underline{\underline{``The critical peak case'':
There is some \math{p\in \Pi_1'\tightcap\TPOS{l_{0,q}}}
with \math{l_{0,q}/p\tightnotin\V}:}}
\begin{diagram}
l_{0,q}\mu_{0,q}&\rredindex{\omega+n+1,\,p}&u'
&&\rredparaindex{\omega+n+1,\,\Pi_1'\setminus\{p\}}&&w_1/q
\\&&&&&&
\\\dredindex{\omega,\,\emptyset}&&\dredparaindex{\omega,\,\Pi''}
&&&&\drefltransindex{\omega}
\\&&&&&&
\\w_0/q&\requal&w_0/q&\rredparaindex{\omega+n+1}&\circ
&\rrefltransindex{\omega[+n]}&\circ
\end{diagram}
\underline{Claim~10:}
\bigmath{p\tightnotequal\emptyset.}
\\
\underline{Proof of Claim~10:}
If \bigmath{p\tightequal\emptyset,} then
\bigmath{\emptyset\tightin\Pi_1',} then
\bigmath{q\tightin\Pi_1,} which contradicts our global case assumption.%
\QED{Claim~10}

\noindent
Let \math{\xi\in\SUBST\V\V} be a bijection with 
\bigmath{
  \xi[\VAR{\kurzregelindex{1,q p}}]\cap\VAR{\kurzregelindex{0,q}}
  =
  \emptyset
.}
\\
Define
\bigmath{
  \Y
  :=
  \xi[\VAR{\kurzregelindex{1,q p}}]\cup\VAR{\kurzregelindex{0,q}}
.}
\\
Let \math{\varrho\in\Xsubst} be given by
$\ x\varrho=
\left\{\begin{array}{@{}l@{}l@{}}
  x\mu_{0,q}        &\mbox{ if }x\in\VAR{\kurzregelindex{0,q}}\\
  x\xi^{-1}\mu_{1,q p}&\mbox{ else}\\
\end{array}\right\}
\:(x\tightin\V)$.
\\
By
\math{
  l_{1,q p}\xi\varrho
  \tightequal 
  l_{1,q p}\xi\xi^{-1}\mu_{1,q p}
  \tightequal 
  u/q p
  \tightequal 
  l_{0,q}\mu_{0,q}/p  
  \tightequal 
  l_{0,q}\varrho/p
  \tightequal 
  (l_{0,q}/p)\varrho
}
\\
let
\math{
  \sigma:=\minmgu{\{(l_{1,q p}\xi}{l_{0,q}/p)\},\Y}
}
and
\math{\varphi\in\Xsubst}
with
\math{
  \domres{\inpit{\sigma\varphi}}\Y
  \tightequal
  \domres\varrho\Y
}.
\\
Define 
\math{
  u':=  
  \repl{l_{0,q}\mu_{0,q}}
       {p}
       {r_{1,q p}\mu_{1,q p}}
}.
We get
\\\LINEmath{
  \arr{{l@{}l}
  u'
  \tightequal&  
  \repl
    {\replpar
       {u/q}{p'}{l_{1,q p'}\mu_{1,q p'}}{p'\tightin\Pi_1'\tightsetminus\{p\}}}
    {p}
    {r_{1,q p}\mu_{1,q p}}
  \redparaindex{\omega+n+1,\Pi_1'\setminus\{p\}}
  \\&
  \replpar{u/q}{p'}{r_{1,q p'}\mu_{1,q p'}}{p'\tightin\Pi_1'}
  \tightequal  
  w_1/q
  .
  }
}
\\
If 
\bigmath{
  \repl{l_{0,q}}{p}{r_{1,q p}\xi}\sigma
  \tightequal
  r_{0,q}\sigma
,}
then the proof is finished due to 
\\\LINEmath{
  w_0/q
  \tightequal
  r_{0,q}\mu_{0,q}
  \tightequal
  r_{0,q}\sigma\varphi
  \tightequal
  \repl{l_{0,q}}{p}{r_{1,q p}\xi}\sigma\varphi
  \tightequal
  u'
  \redparaindex{\omega+n+1,\Pi_1'\setminus\{p\}}
  w_1/q
.}
\\
Otherwise 
we have
\bigmath{
  (\,
   (\repl{l_{0,q}}{p}{r_{1,q p}\xi}\sigma,
    C_{1,q p}\xi\sigma,    
    1),\penalty-1\,
   (r_{0,q}\sigma,
    C_{0,q}\sigma,
    0),\penalty-1\,
    l_{0,q}\sigma,\penalty-1\,
    p\,)
  \in{\rm CP}(\R)
}
(due to Claim~5);
\bigmath{p\tightnotequal\emptyset}
(due to Claim~10);
\bigmath{C_{1,q p}\xi\sigma\varphi=C_{1,q p}\mu_{1,q p}}
is fulfilled \wrt\ \redindex{\omega+n};
\bigmath{C_{0,q}\sigma\varphi=C_{0,q}\mu_{0,q}}
is fulfilled \wrt\ \redindex{\omega}.
Since \RX\ is
\math\omega-shallow confluent up to \math{n}
(by our induction hypothesis),
due to our assumed \math\omega-shallow parallel closedness up to \math\omega\ 
(matching the definition's \math{n_0} to our \math{n\tight+1} 
                  and its \math{n_1} to \math{0})
we have
\bigmath{
  u'
  \tightequal
  \repl{l_{0,q}}{p}{r_{1,q p}\xi}\sigma\varphi
  \redparaindex\omega
  r_{0,q}\sigma\varphi
  \tightequal
  r_{0,q}\mu_{0,q}
  \tightequal
  w_0/q.
}
We then have
\bigmath{
  w_0/q
  \antiredparaindex{\omega,\Pi''} 
  u'
  \redparaindex{\omega+n+1,\Pi_1'\setminus\{p\}}
  w_1/q
}
for some \math{\Pi''}.
We can finish the proof in this case due to our second induction level since
\\\bigmath{
  \displaystyle
  \sum_{p''\in\Omega(\Pi'',\Pi_1'\setminus\{p\})}
  \lambda(u'/p'')
  \ \ \preceq
  \sum_{p''\in\Pi_1'\setminus\{p\}}
  \lambda(u'/p'')
  \ \ =
  \sum_{p''\in\Pi_1'\setminus\{p\}}
  \lambda(u/q p'')
}\\\LINEmath{\displaystyle
  \ \ \prec
  \sum_{p''\in\Pi_1'}
  \lambda(u/q p'')
  \ \ =
  \sum_{p'\in q\Pi_1'}
  \lambda(u/p')
  \ \ =
  \sum_{p'\in\Omega(\{q\},\Pi_1)}
  \lambda(u/p')
  \ \ \preceq
  \sum_{p'\in\Omega(\Pi_0,\Pi_1)}
  \lambda(u/p')
.}
\\
\Qeddouble{``The critical peak case''}\QEDtriple{``\math{q\tightnotin\Pi_1}''}

\pagebreak

\noindent
\underline{\underline{\underline{\math{q\tightin\Pi_1}:}}}
Define \math{\Pi_0':=\setwith{p}{q p\tightin\Pi_0}}.
We have two cases:

\yestop
\noindent
\underline{\underline{``The second variable overlap (if any) case'':
\math{
  \forall p\tightin\Pi_0'\tightcap\TPOS{l_{1,q}}\stopq
    l_{1,q}/p\tightin\V
}:}}
\begin{diagram}
l_{1,q}\mu_{1,q}&&&\rredindex{\omega+n+1,\,\emptyset}&&&w_1/q
\\&&&&&&\dequal
\\\dredparaindex{\omega}&&&&&&r_{1,q}\mu_{1,q}
\\&&&&&&\dredparaindex{\omega}
\\w_0/q&\requal&l_{1,q}\nu&&\rredindex{\omega+n+1}&&r_{1,q}\nu
\end{diagram}
\noindent
Define a function \math\Gamma\ on \V\ by (\math{x\tightin\V}):
\bigmath{
  \Gamma(x):=
  \setwith{(p',p'')}
          {l_{1,q}/p'\tightequal x\ \wedge\ p' p''\in\Pi_0'}
.}

\noindent
\underline{Claim~11:}
There is some \math{\nu\in\Xsubst} with
\\\LINEmath{
  \forall x\in\V\stopq
    \inparenthesesoplist{
       x\nu
       \antiredparaindex\omega
       x\mu_{1,q}
    \oplistund
       \forall p'\tightin\DOM{\Gamma(x)}\stopq
         \replpar
           {x\mu_{1,q}}
           {p''}
           {r_{0,q p' p''}\mu_{0,q p' p''}}
           {(p',p'')\tightin\Gamma(x)}
         \tightequal
         x\nu
    }
.}
\\
\underline{Proof of Claim~11:}
\\
In case of \bigmath{\DOM{\Gamma(x)}\tightequal\emptyset} we define
\bigmath{x\nu:=x\mu_{1,q}.}
If there is some \math{p'} such that 
\bigmath{\DOM{\Gamma(x)}\tightequal\{p'\}}
we define 
\bigmath{
  x\nu
  :=
  \replpar
    {x\mu_{1,q}}{p''}{r_{0,q p' p''}\mu_{0,q p' p''}}{(p',p'')\tightin\Gamma(x)}
.}
This is appropriate since due to 
\bigmath{
  \forall(p',p'')\tightin\Gamma(x)\stopq
    x\mu_{1,q}/p''
    \tightequal 
    l_{1,q}\mu_{1,q}/p' p''
    \tightequal 
    u/q p' p''
    \tightequal 
    l_{0,q p' p''}\mu_{0,q p' p''}
}
we have
\\\LINEmath{
  \begin{array}{l@{}l@{}l}
  x\mu_{1,q}&
  \tightequal&
  \replpar
    {x\mu_{1,q}}
    {p''}
    {l_{0,q p' p''}\mu_{0,q p' p''}}
    {(p',p'')\tightin\Gamma(x)}
  \redparaindex{\omega}\\&&
  \replpar
     {x\mu_{1,q}}
     {p''}
     {r_{0,q p' p''}\mu_{0,q p' p''}}
     {(p',p'')\tightin\Gamma(x)}
  \tightequal
  x\nu.  
  \end{array}
}
\\
Finally, in case of \bigmath{\CARD{\DOM{\Gamma(x)}}\succ1,} \math{l_{1,q}} is
not linear in \math x. By the conditions of our lemma, this 
contradicts Claim~5.\QED{Claim~11}

\noindent
\underline{Claim~12:}
\bigmath{w_0/q\tightequal l_{1,q}\nu.}
\\
\underline{Proof of Claim~12:}
\\
By Claim~11 we get
\bigmath{
  w_0/q
  \tightequal
  \replpar 
    {u/q}
    {p' p''}
    {r_{0,q p' p''}\mu_{0,q p' p''}}
    {\exists x\tightin\V\stopq(p',p'')\tightin\Gamma(x)}
  \tightequal\\
  \replpar
    {\replpar
       {l_{1,q}}
       {p'}
       {x\mu_{1,q}}
       {l_{1,q}/p'\tightequal x\tightin\V}
    }
    {p' p''}
    {r_{0,q p' p''}\mu_{0,q p' p''}}
    {\exists x\tightin\V\stopq(p',p'')\tightin\Gamma(x)}
  \tightequal\\
  \replpar
    {l_{1,q}}
    {p'}
    {\replpar
       {x\mu_{1,q}}
       {p''}
       {r_{0,q p' p''}\mu_{0,q p' p''}}
       {(p',p'')\tightin\Gamma(x)}}
    {l_{1,q}/p'\tightequal x\tightin\V}
  \tightequal\\
  \replpar
    {l_{1,q}}
    {p'}
    {x\nu}
    {l_{1,q}/p'\tightequal x\tightin\V}
  \tightequal
  l_{1,q}\nu
.}\QED{Claim~12}

\noindent
\underline{Claim~13:}
\bigmath{
  r_{1,q}\nu
  \antiredparaindex\omega
  w_1/q
.}
\\
\underline{Proof of Claim~13:} 
Since \bigmath{r_{1,q}\mu_{1,q}\tightequal w_1/q,} 
this follows directly from Claim~11.%
\QED{Claim~13}

\noindent
By claims 12 and 13 
using Corollary~\ref{corollary parallel one} 
it now suffices to show
\bigmath{
  l_{1,q}\nu
  \redindex{\omega+n+1}
  r_{1,q}\nu
,}
which again follows from 
Claim~11,
Lemma~\ref{lemma invariance of fulfilledness}
(matching its \math{n_0} to \math{0} and its \math{n_1} to our \math{n})
and our induction hypothesis that \RX\ is \math\omega-shallow confluent up
to \math n.%
\QEDdouble{``The second variable overlap (if any) case''}

\pagebreak

\yestop
\noindent
\underline{\underline{``The second critical peak case'':
There is some \math{p\in \Pi_0'\tightcap\TPOS{l_{1,q}}}
with \math{l_{1,q}/p\tightnotin\V}:}}
\begin{diagram}
l_{1,q}\mu_{1,q}&&&\rredindex{\omega+n+1,\,\emptyset}&&&w_1/q
\\\dredindex{\omega,\,p}&&&&&&\drefltransindex{\omega}
\\u'&&\rredparaindex{\omega+n+1}&&v_1&\rrefltransindex{\omega[+n]}&v_2
\\\dredparaindex{\omega,\,\Pi_0'\setminus\{p\}}&&&&\drefltransindex{\omega}
&&\drefltransindex{\omega}
\\w_0/q&\rredparaindex{\omega+n+1}&\circ&\rrefltransindex{\omega[+n]}&v_1'
&\rrefltransindex{\omega[+n]}&\circ
\end{diagram}
Let \math{\xi\in\SUBST\V\V} be a bijection with 
\bigmath{
  \xi[\VAR{\kurzregelindex{0,q p}}]\cap\VAR{\kurzregelindex{1,q}}
  =
  \emptyset
.}
\\
Define
\bigmath{
  \Y
  :=
  \xi[\VAR{\kurzregelindex{0,q p}}]\cup\VAR{\kurzregelindex{1,q}}
.}
\\
Let \math{\varrho\in\Xsubst} be given by
$\ x\varrho=
\left\{\begin{array}{@{}l@{}l@{}}
  x\mu_{1,q}        &\mbox{ if }x\in\VAR{\kurzregelindex{1,q}}\\
  x\xi^{-1}\mu_{0,q p}&\mbox{ else}\\
\end{array}\right\}
\:(x\tightin\V)$.
\\
By
\math{
  l_{0,q p}\xi\varrho
  \tightequal 
  l_{0,q p}\xi\xi^{-1}\mu_{0,q p}
  \tightequal 
  u/q p
  \tightequal 
  l_{1,q}\mu_{1,q}/p  
  \tightequal 
  l_{1,q}\varrho/p
  \tightequal 
  (l_{1,q}/p)\varrho
}
\\
let
\math{
  \sigma:=\minmgu{\{(l_{0,q p}\xi}{l_{1,q}/p)\},\Y}
}
and
\math{\varphi\in\Xsubst}
with
\math{
  \domres{\inpit{\sigma\varphi}}\Y
  \tightequal
  \domres\varrho\Y
}.
\\
Define
\math{
  u':=  
  \repl{l_{1,q}\mu_{1,q}}
       {p}
       {r_{0,q p}\mu_{0,q p}}
.}
We get 
\\\LINEmath{
  \begin{array}{l@{}l@{}l}
  w_0/q&
  \tightequal&
  \replpar{u/q}{p'}{r_{0,q p'}\mu_{0,q p'}}{p'\tightin\Pi_0'}
  \antiredparaindex{\omega,\Pi_0'\setminus\{p\}}
  \\&&
  \repl
    {\replpar
       {u/q}{p'}{l_{0,q p'}\mu_{0,q p'}}{p'\tightin\Pi_0'\tightsetminus\{p\}}}
    {p}
    {r_{0,q p}\mu_{0,q p}}
  \tightequal
  u'
  .
  \end{array}
}
\\
If 
\bigmath{
  \repl{l_{1,q}}{p}{r_{0,q p}\xi}\sigma
  \tightequal
  r_{1,q}\sigma
,}
then the proof is finished due to 
\\\LINEmath{
  w_0/q  
  \antiredparaindex{\omega,\Pi_0'\setminus\{p\}}
  u'
  \tightequal
  \repl{l_{1,q}}{p}{r_{0,q p}\xi}\sigma\varphi
  \tightequal
  r_{1,q}\sigma\varphi
  \tightequal
  r_{1,q}\mu_{1,q}
  \tightequal
  w_1/q.
}
\\
Otherwise 
we have
\bigmath{
  (\,
   (\repl{l_{1,q}}{p}{r_{0,q p}\xi}\sigma,
    C_{0,q p}\xi\sigma,    
    0),\penalty-1\,
   (r_{1,q}\sigma,
    C_{1,q}\sigma,
    1),\penalty-1\,
    l_{1,q}\sigma,\penalty-1\,
    p\,)
  \in{\rm CP}(\R)
}
(due to Claim~5);
\bigmath{C_{0,q p}\xi\sigma\varphi=C_{0,q p}\mu_{0,q p}}
is fulfilled \wrt\ \redindex{\omega};
\bigmath{C_{1,q}\sigma\varphi=C_{1,q}\mu_{1,q}}
is fulfilled \wrt\ \redindex{\omega+n}.
Since \RX\  
\math\omega-shallow confluent up to \math{n}
(by our induction hypothesis),
due to our assumed \math\omega-shallow [noisy] 
parallel joinability up to \math\omega\ 
(matching the definition's \math{n_0} to \math 0
                   and its \math{n_1} to our \math{n\tight+1})
we have
\bigmath{
  u'
  \tightequal
  \repl{l_{1,q}}{p}{r_{0,q p}\xi}\sigma\varphi
  \redparaindex{\omega+n+1}
  v_1
  \refltransindex{\omega[+n]}
  v_2
  \antirefltransindex\omega
  r_{1,q}\sigma\varphi
  \tightequal
  r_{1,q}\mu_{1,q}
  \tightequal
  w_1/q
}
for some \math{v_1}, \math{v_2}.
We then have
\bigmath{
  w_0/q
  \antiredparaindex{\omega,\Pi_0'\setminus\{p\}}
  u'
  \redparaindex{\omega+n+1,\Pi''} 
  v_1
}
for some \math{\Pi''}.
Since
\bigmath{
  \displaystyle
  \sum_{p''\in\Omega(
    \Pi_0'\setminus\{p\}
    ,
    \Pi''
    )}
  \lambda(u'/p'')
  \ \ \preceq
  \sum_{p''\in\Pi_0'\setminus\{p\}}
  \lambda(u'/p'')
  \ \ =
  \sum_{p''\in\Pi_0'\setminus\{p\}}
  \lambda(u/q p'')
  \ \ \prec
  \sum_{p''\in\Pi_0'}
  \lambda(u/q p'')
  \ \ =
  \sum_{p'\in q\Pi_0'}
  \lambda(u/p')
  \ \ =
  \sum_{p'\in\Omega(
    \Pi_0
    ,
    \{q\}
    )}
  \lambda(u/p')
  \ \ \preceq
  \sum_{p'\in\Omega(\Pi_0,\Pi_1)}
  \lambda(u/p')
}
due to our second induction level 
we get some \math{v_1'} with 
\bigmath{
  w_0/q
  \redparaindex{\omega+n+1}
  \tight\circ
  \refltransindex{\omega[+n]}
  v_1'
  \antirefltransindex\omega
  v_1
.} 
From the peak
\bigmath{
  v_1'
  \antirefltransindex\omega
  v_1
  \refltransindex{\omega[+n]}
  v_2
} 
we finally get
\bigmath{
%  w_0/q
%  \redparaindex{\omega+n+1}
%  \tight\circ
%  \refltransindex{\omega[+n]}
  v_1'
  \refltransindex{\omega[+n]}
  \circ
  \antirefltransindex\omega
  v_2
%  \antirefltransindex\omega
%  w_1/q
}
by \math\omega-shallow confluence up to \math{0[+n]}.%
\\\Qeddouble{``The second critical peak case''}
\end{proofparsepqed}

\pagebreak

\begin{proofparsepqed}{Lemma~\ref{lemma parallel closed second level two}}
\underline{Claim~0:}
\RX\ is \math\omega-shallow confluent up to \math\omega.
\\\underline{Proof of Claim~0:}
Directly by the assumed strong commutation,
\cf\ the proofs of the claims 2 and 3 of the proof of 
Lemma~\ref{lemma parallel closed first level two}.\QED{Claim~0}

\yestop
\noindent
\underline{Claim~1:}
If 
\bigmath{
    \refltransindex{\omega}
    \tight\circ
    \redparaindex{\omega+n_1}
    \tight\circ
    \refltransindex{\omega+(n_1\monus1)}
}
strongly commutes over
\refltransindex{\omega+n_0},
then
\redindex{\omega+n_1} and \redindex{\omega+n_0} are commuting.
\\
\underline{Proof of Claim~1:}
\bigmath{
    \refltransindex{\omega}
    \tight\circ
    \redparaindex{\omega+n_1}
    \tight\circ
    \refltransindex{\omega+(n_1\monus1)}
}
and
\refltransindex{\omega+n_0}
are commuting
by Lemma~\ref{lemma strong commutation one copy}.
Since by Corollary~\ref{corollary parallel one}
and \lemmamonotonicinbeta\
we have
\bigmath{
  \redindex{\omega+n_1}
  \subseteq
  \refltransindex{\omega}
  \tight\circ
  \redparaindex{\omega+n_1}
  \tight\circ
  \refltransindex{\omega+(n_1\monus1)}
  \subseteq
  \refltransindex{\omega+n_1}
,}
now \redindex{\omega+n_1} 
and \redindex{\omega+n_0} are commuting, too.\QED{Claim~1}

\yestop
\yestop
\noindent
For \math{n_0\preceq n_1\prec\omega} 
we are going to show by induction on \math{n_0\plusomega n_1}
the following property\footroom:
\\\LINEmath{
    w_0
    \antiredparaindex{\omega+n_0}
    u
    \redparaindex{\omega+n_1}
    w_1
  \quad\implies\quad
    w_0
    \refltransindex{\omega}
    \tight\circ
    \redparaindex{\omega+n_1}
    \tight\circ
    \refltransindex{\omega+(n_1\monus1)}
    \circ
    \antirefltransindex{\omega+n_0}
    w_1
.}
\begin{diagram}
u&&&\rredparaindex{\omega+n_1}&&&w_1
\\\dredparaindex{\omega+n_0}&&&&&&\drefltransindex{\omega+n_0}
\\w_0&\rrefltransindex{\omega}&\circ&\rredparaindex{\omega+n_1}
&\circ&\rrefltransindex{\omega+(n_1\monus1)}&\circ
\end{diagram}

\pagebreak

\noindent
\underline{Claim~2:}
Let \math{\delta\prec\omega\tight+\omega}.
If
\\
\linemath{
  \forall n_0,n_1\tightprec\omega\stopq
  \inparenthesesoplist{ 
    \inparenthesesoplist{
        n_0\tightpreceq n_1      
      \oplistund
        n_0\plusomega n_1\tightpreceq\delta
    }
    \oplistimplies
    \forall w_0,w_1,u\stopq
    \inparenthesesoplist{
        w_0
        \antiredparaindex{\omega+n_0}
        u
        \redparaindex{\omega+n_1}
        w_1
      \oplistimplies 
        w_0
        \refltransindex{\omega}
        \tight\circ
        \redparaindex{\omega+n_1}
        \tight\circ
        \refltransindex{\omega+(n_1\monus1)}
        \circ
        \antirefltransindex{\omega+n_0}
        w_1
    }
  }
,}
then
\\
\linemath{
  \forall n_0,n_1\tightprec\omega\stopq
  \inparenthesesoplist{ 
    \inparenthesesoplist{
        n_0\tightpreceq n_1      
      \oplistund
        n_0\plusomega n_1\tightpreceq\delta
    }
    \oplistimplies 
            \refltransindex{\omega}
            \tight\circ
            \redparaindex{\omega+n_1}
            \tight\circ
            \refltransindex{\omega+(n_1\monus1)}
      \mbox{ strongly commutes over }
            \refltransindex{\omega+n_0}
  }
,}
and 
\RX\ is \math\omega-shallow confluent up to \math{\delta}.
\\
\underline{Proof of Claim~2:}
By induction on \math{\delta} in \tightprec.
First we show the strong commutation.
Assume \math{n_0\preceq n_1\prec\omega} with
\math{n_0\plusomega n_1\tightpreceq\delta}. 
By Lemma~\ref{lemma strong commutation one copy} it suffices to show that
\bigmath{
    \refltransindex{\omega}
    \tight\circ
    \redparaindex{\omega+n_1}
    \tight\circ
    \refltransindex{\omega+(n_1\monus1)}
}
strongly commutes over
\redindex{\omega+n_0}.
Assume
\bigmath{
    u''
    \antiredindex{\omega+n_0}
    u'
    \refltransindex{\omega}
    u
    \redparaindex{\omega+n_1}
    w_1
    \refltransindex{\omega+(n_1\monus1)}
    w_2
}
(\cf\ diagram below).
By the strong commutation assumed for our lemma
and Corollary~\ref{corollary parallel one},
there are \math{w_0} and \math{w_0'} with
\bigmath{
    u''
    \refltransindex{\omega}
    w_0'
    \antirefltransindex{\omega+(n_0\monus1)}
    w_0
    \antiredparaindex{\omega+n_0}
    u
.}
By the above property there are some \math{w_3}, \math{w_1'}
with
\bigmath{
    w_0
    \refltransindex{\omega}
    w_3
    \redparaindex{\omega+n_1}
    \tight\circ
    \refltransindex{\omega+(n_1\monus1)}
    w_1'
    \antirefltransindex{\omega+n_0}
    w_1
.}
Next we show that we can close the peak
\bigmath{ 
    w_1'
    \antirefltransindex{\omega+n_0}
    w_1
    \refltransindex{\omega+(n_1\monus1)}
    w_2
}
according to 
\bigmath{
    w_1'
    \refltransindex{\omega+(n_1\monus1)}
    w_2'
    \antirefltransindex{\omega+n_0}
    w_2
}
for some \math{w_2'}.
In case of \bigmath{n_1\tightequal0}
this is possible due to the
\math\omega-shallow confluence up to \math{\omega}
given by Claim~0.
Otherwise we have 
\bigmath{
  n_0\plusomega(n_1\monus1)
  \tightprec 
  n_0\plusomega n_1
  \tightpreceq
  \delta
} 
and due to our induction hypothesis
(saying that \RX\ is \math\omega-shallow confluent up to all 
\math{\delta'\prec\delta
})
this is possible again.
By Claim~0 again,
we can close the peak
\bigmath{
  w_0'
  \antirefltransindex{\omega+(n_0\monus1)}
  w_0
  \refltransindex{\omega}
  w_3
}
according to 
\bigmath{
  w_0'
  \refltransindex{\omega}
  w_3'
  \antirefltransindex{\omega+(n_0\monus1)}
  w_3
}
for some \math{w_3'}.
To close the whole diagram, we only have to show that we can close the peak
\bigmath{
    w_3'
    \antirefltransindex{\omega+(n_0\monus1)}
    w_3
    \redparaindex{\omega+n_1}
    \tight\circ
    \refltransindex{\omega+(n_1\monus1)}
    w_2'
}
according to 
\bigmath{
    w_3'
    \refltransindex{\omega}
    \tight\circ
    \redparaindex{\omega+n_1}
    \tight\circ
    \refltransindex{\omega+(n_1\monus1)}
    \circ
    \antirefltransindex{\omega+(n_0\monus1)}
    w_2'
.}
In case of \bigmath{n_0\tightequal0}
this is possible due to the strong commutation assumed for our lemma.
Otherwise we have 
\bigmath{
  n_0\monus1
  \tightprec 
  n_0
  \tightpreceq
  n_1
} 
and
\bigmath{
  (n_0\monus1)\plusomega n_1
  \tightprec 
  n_0\plusomega n_1
  \tightpreceq
  \delta
,} 
and then due to our
induction hypothesis
this is possible again.
\begin{diagram}
u'&\rrefltransindex{\omega}&u
&&
&\rredparaindex{\omega+n_1}&&
&w_1&\rrefltransindex{\omega+(n_1\monus1)}&w_2
\\&&\dredparaindex{\omega+n_0}
&&
&&&
&\drefltransindex{\omega+n_0}&&\drefltransindex{\omega+n_0}
\\\dredindex{\omega+n_0}&&w_0
&\rrefltransindex{\omega}&w_3
&\rredparaindex{\omega+n_1}&\circ&\rrefltransindex{\omega+(n_1\monus1)}
&w_1'&\rrefltransindex{\omega+(n_1\monus1)}&w_2'
\\&&\drefltransindex{\omega+(n_0\monus1)}
&&\drefltransindex{\omega+(n_0\monus1)}
&&&
&&&\drefltransindex{\omega+(n_0\monus1)}
\\u''&\rrefltransindex{\omega}&w_0'
&\rrefltransindex{\omega}&w_3'
&\rrefltransindex{\omega}&\circ&\rredparaindex{\omega+n_1}
&\circ&\rrefltransindex{\omega+(n_1\monus1)}&\circ
\end{diagram}
Finally we show \math\omega-shallow confluence up to \math\delta.
Assume
\bigmath{n_0\plusomega n_1\tightpreceq\delta}
and
\bigmath{
    w_0
    \antirefltransindex{\omega+n_0}
    u
    \refltransindex    {\omega+n_1}
    w_1
.}
Due to symmetry in \math{n_0} and \math{n_1} we may assume
\bigmath{n_0\tightpreceq n_1.}
Above we have shown that 
\bigmath{
    \refltransindex{\omega}
    \tight\circ
    \redparaindex{\omega+n_1}
    \tight\circ
    \refltransindex{\omega+(n_1\monus1)}
}
strongly commutes over
\refltransindex{\omega+n_0}. 
By Claim~1 we finally get
\bigmath{
  w_0\refltransindex{\omega+n_1}\circ\antirefltransindex{\omega+n_0} w_1
}
as desired.%
\QED{Claim~2}

\pagebreak

\yestop
\yestop
\noindent
Note that 
for \bigmath{n_0\tightequal0} 
our property follows 
from 
\bigmath{\antiredparaindex\omega\subseteq\antirefltransindex\omega}
(by Corollary~\ref{corollary parallel one})
and
the assumption of our lemma that
for each \math{n_1\prec\omega}:
\bigmath{
  \redparaindex{\RX,\omega+n_1}
  \tight\circ
  \refltransindex{\RX,\omega+(n_1\monus1)}
}
strongly commutes over \refltransindex{\RX,\omega}.

The benefit of 
Claim~2 is twofold: First, it says that our lemma is valid
if the above property
holds for all \math{n_0\preceq n_1\prec\omega}.
Second, it strengthens the property when used as induction hypothesis. Thus 
(writing \math{n_i\tight+1} instead of \math{n_i} since we may assume
 \math{0\tightprec n_0\tightpreceq n_1})
it
now suffices to show
for
\math{n_0\preceq n_1\prec\omega}
that 
\\\linemath{
    w_0
    \antiredparaindex{\omega+n_0+1,\Pi_0}
    u
    \redparaindex    {\omega+n_1+1,\Pi_1}
    w_1
}
together with our induction hypotheses 
that\headroom
\\\linemath{\headroom
  \forall\delta\tightprec(n_0\tight+1)\plusomega(n_1\tight+1)\stopq
  \mbox{\RX\ is \math\omega-shallow confluent up to }\delta
}
\headroom
and (due to \math{n_0\tightpreceq n_1\tight+1}
and 
\math{
  n_0\plusomega(n_1\tight+1)\tightprec(n_0\tight+1)\plusomega(n_1\tight+1)
})
\\\linenomath{\headroom
  \bigmath{ 
    \refltransindex{\omega}
    \tight\circ
    \redparaindex{\omega+n_1+1}
    \tight\circ
    \refltransindex{\omega+n_1}
  }
  strongly commutes over
  \refltransindex{\omega+n_0}
}
\headroom
implies
\\\LINEmath{
    w_0
    \refltransindex{\omega}
    \tight\circ
    \redparaindex{\omega+n_1+1}
    \tight\circ
    \refltransindex{\omega+n_1}
    \circ
    \antirefltransindex{\omega+n_0+1}
    w_1
.}
\begin{diagram}
u&&&\rredparaindex{\omega+n_1+1,\,\Pi_1}
&&&w_1
\\\dredparaindex{\omega+n_0+1,\,\Pi_0}&&&
&&&\drefltransindex{\omega+n_0+1}
\\w_0&\rrefltransindex{\omega}&\circ&\rredparaindex{\omega+n_1+1}
&\circ&\rrefltransindex{\omega+n_1}&\circ
\end{diagram}

\yestop
\noindent
{Note that for the availability of our second induction hypothesis  
it is important that we have imposed the restriction
``\math{n_0\tightpreceq n_1}'' in opposition to the restriction
``\math{n_0\tightsucceq n_1}''.
In the latter case the availability of our second induction hypothesis would
require 
\bigmath{
  n_0\tight+1\tightsucceq n_1\tight+1
  \implies
  n_0\tightsucceq n_1\tight+1
}
which is not true for \bigmath{n_0\tightequal n_1.}
The additional hypothesis  
\\\linenomath{
  \bigmath{ 
    \refltransindex{\omega}
    \tight\circ
    \redparaindex{\omega+n_1}
    \tight\circ
    \refltransindex{\omega+(n_1\monus1)}
  }
  strongly commutes over
  \refltransindex{\omega+n_0+1}
}
of the latter restriction is useless for our proof.}

\yestop
\noindent
\Wrog\ let the positions of \math{\Pi_i} be maximal
in the sense that for any \math{p\in\Pi_i} 
and \math{\Xi\subseteq \TPOS u\tightcap(p\N^+)}
we do not have 
\bigmath{
  u
  \redparaindex{\omega+n_i+1,(\Pi_i\setminus\{p\})\cup\Xi}
  w_i
}
anymore.
Then for each \math{i\prec2} and
\math{p\in\Pi_i} there are
\bigmath{\kurzregelindex{i,p}\in\R}
and
\bigmath{\mu_{i,p}\in\Xsubst}
with
\bigmath{u/p\tightequal l_{i,p}\mu_{i,p},}
\bigmath{r_{i,p}\mu_{i,p}\tightequal w_i/p,}
\math{C_{i,p}\mu_{i,p}} fulfilled \wrt\ \redindex{\omega+n_i}.
Finally, for each \math{i\prec2}:
\bigmath{
  w_i\tightequal\replpar{u}{p}{r_{i,p}\mu_{i,p}}{p\tightin\Pi_i}
.}

\pagebreak

\yestop
\noindent
\underline{Claim~5:}
We may assume 
\bigmath{
  \forall i\tightprec2\stopq
  \forall p\tightin\Pi_i\stopq
    l_{i,p}\tightnotin\tcs
.}
\\
\underline{Proof of Claim~5:}
Define \math{\Xi_i:=\setwith{p\tightin\Pi_i}{l_{i,p}\tightin\tcs}}
and 
\math{
  u_i':=\replpar{u}{p}{r_{i,p}\mu_{i,p}}{p\tightin\Pi_i\tightsetminus\Xi_i}
}.
If we have succeeded with our proof under the assumption of Claim~5,
then we have shown 
\bigmath{
  u_0'
  \refltransindex{\omega}
  v_0
  \redparaindex{\omega+n_1+1}
  \tight\circ
  \refltransindex{\omega+n_1}
  v_1
  \antirefltransindex{\omega+n_0+1}
  u_1'
}
for some \math{v_0}, \math{v_1}
(\cf\ diagram below).
By Lemma~\ref{lemma invariance of fulfilledness two} 
(matching both its \math\mu\ and \math\nu\ to our \math{\mu_{i,p}})
we get
\bigmath{
  \forall i\tightprec2\stopq
  \forall p\tightin\Xi_i\stopq
    l_{i,p}\mu_{i,p}\redindex\omega r_{i,p}\mu_{i,p}
}
and therefore
\bigmath{
  \forall i\tightprec2\stopq
  u_i'
  \refltransindex\omega
  w_i
.}
Thus from
\bigmath{
  v_1
  \penalty-1
  \antirefltransindex{\omega+n_0+1}
  \penalty-1
  u_1'
  \penalty-1
  \refltransindex\omega
  \penalty-1
  w_1
}
we get 
\bigmath{
  v_1
  \refltransindex\omega
  v_2
  \antirefltransindex{\omega+n_0+1}
  w_1
}
for some \math{v_2}
by \math\omega-shallow confluence up to \math{\omega} (\cf\ Claim~0).
For the same reason 
we can close the peak 
\bigmath{
  w_0
  \antirefltransindex{\omega}
  u_0'
  \refltransindex{\omega}
  v_0
}
according to 
\bigmath{
  w_0
  \refltransindex{\omega}
  v_0'
  \antirefltransindex{\omega}
  v_0
}
for some \math{v_0'}.
By the assumption of our lemma that
\bigmath{
  \redparaindex{\RX,\omega+n_1+1}
  \tight\circ
  \refltransindex{\RX,\omega+n_1}
}
strongly commutes over \refltransindex{\omega},
from
\bigmath{
  v_0'
  \antirefltransindex\omega
  v_0
  \redparaindex{\omega+n_1+1}
  \tight\circ
  \refltransindex{\omega+n_1}
  v_1
  \refltransindex{\omega+n_1}
  v_2
}
we can finally conclude
\bigmath{
  v_0'
  \redparaindex{\omega+n_1+1}
  \tight\circ
  \refltransindex{\omega+n_1}
  \circ
  \penalty-1
  \antirefltransindex{\omega}
  v_2
%  \antirefltransindex{\omega+n_0+1}
%  w_1
.}
\begin{diagram}
u&&&\rredparaindex{\omega+n_1+1,\,\Pi_1\setminus\Xi_1}
&&&u_1'&\rrefltransindex{\omega}&w_1
\\\dredparaindex{\omega+n_0+1,\,\Pi_0\setminus\Xi_0}&&&
&&&\drefltransindex{\omega+n_0+1}&&\drefltransindex{\omega+n_0+1}
\\u_0'&\rrefltransindex{\omega}&v_0&\rredparaindex{\omega+n_1+1}
&\circ&\rrefltransindex{\omega+n_1}&v_1&\rrefltransindex{\omega}&v_2
\\\drefltransindex{\omega}&&\drefltransindex{\omega}&
&&&&&\drefltransindex{\omega}
\\w_0&\rrefltransindex{\omega}&v_0'&\rredparaindex{\omega+n_1+1}
&\circ&&\rrefltransindex{\omega+n_1}&&\circ
\end{diagram}
\Qed{Claim~5}

\yestop
\noindent
Define
the set of inner overlapping positions by
\\\linemath{
  \displaystyle
  \Omega(\Pi_0,\Pi_1)
  :=
  \bigcup_{i\prec2}
    \setwith
      {p\tightin\Pi_{1-i}}
      {\exists q\tightin\Pi_i\stopq\exists q'\tightin\N^\ast\stopq
        p\tightequal q q'
      }
,}
and the length of a term by
\bigmath{\lambda(\anonymousfpp{t_0}{t_{m-1}}):=1+\sum_{j\prec m}\lambda(t_j).}

\yestop
\noindent
Now we start a second level of induction on
\bigmath{  \displaystyle
  \sum_{p'\in\Omega(\Pi_0,\Pi_1)}\lambda(u/p')
}
in \tightprec.

\yestop
\noindent
Define the set of top positions by
\\\linemath{
  \displaystyle
  \Theta
  :=
      \setwith
      {p\tightin\Pi_0\tightcup\Pi_1}
      {\neg\exists q\tightin\Pi_0\tightcup\Pi_1\stopq
           \exists q'\tightin\N^+\stopq
             p\tightequal q q'
      }
.}
Since the prefix ordering is wellfounded we have
\bigmath{
  \forall i\tightprec2\stopq
  \forall p\tightin\Pi_i\stopq
  \exists q\tightin\Theta\stopq
  \exists q'\tightin\N^\ast\stopq
    p\tightequal q q'
.}
Then
\bigmath{
  \forall i\tightprec2\stopq
  w_i
  \tightequal
  \replpar{w_i}{q}{w_i/q}{q\tightin\Theta}
  \tightequal
  \replpar{\replpar{u}{p}{r_{i,p}\mu_{i,p}}{p\tightin\Pi_i}}
          {q}{w_i/q}{q\tightin\Theta}
  \tightequal
  \replpar{u}{q}{w_i/q}{q\tightin\Theta}
.}
Thus, it now suffices to show for all \math{q\in\Theta}
\\\linemath{\headroom\footroom
    w_0/q
    \refltransindex{\omega}
    \tight\circ
    \redparaindex{\omega+n_1+1}
    \tight\circ
    \refltransindex{\omega+n_1}
    \circ
    \antirefltransindex{\omega+n_0+1}
    w_1/q
}
because then we have 
\\\LINEmath{
  w_0
  \tightequal
  \replpar{u}{q}{w_0/q}{q\tightin\Theta}
    \refltransindex{\omega}
    \tight\circ
    \redparaindex{\omega+n_1+1}
    \tight\circ
    \refltransindex{\omega+n_1}
    \circ
    \antirefltransindex{\omega+n_0+1}
  \replpar{u}{q}{w_1/q}{q\tightin\Theta}
  \tightequal
  w_1
.}

\noindent
Therefore we are left with the following two cases for \math{q\in\Theta}:

\pagebreak

\yestop
\yestop
\noindent
\underline{\underline{\underline{\math{q\tightnotin\Pi_1}:}}}
Then \bigmath{q\tightin\Pi_0.}
Define \math{\Pi_1':=\setwith{p}{q p\tightin\Pi_1}}.
\noindent
We have two cases:

\yestop
\noindent
\underline{\underline{``The variable overlap (if any) case'':
\math{
  \forall p\tightin\Pi_1'\tightcap\TPOS{l_{0,q}}\stopq
    l_{0,q}/p\tightin\V
}:}}
\begin{diagram}
l_{0,q}\mu_{0,q}&&\rredparaindex{\omega+n_1+1}&&&&w_1/q
\\&&&&&&\dequal
\\\dredindex{\omega+n_0+1,\,\emptyset}&&&&&&l_{0,q}\nu
\\&&&&&&\dredindex{\omega+n_0+1}
\\w_0/q&\requal&r_{0,q}\mu_{0,q}&&\rredparaindex{\omega+n_1+1}&&r_{0,q}\nu
\end{diagram}
\noindent
Define a function \math\Gamma\ on \V\ by (\math{x\tightin\V}):
\bigmath{
  \Gamma(x):=
  \setwith{(p',p'')}
          {l_{0,q}/p'\tightequal x\ \wedge\ p' p''\in\Pi_1'}
.}

\noindent
\underline{Claim~7:}
There is some \math{\nu\in\Xsubst} with
\\\LINEmath{
  \forall x\in\V\stopq
    \inparenthesesoplist{
       x\mu_{0,q}
       \redparaindex{\omega+n_1+1}
       x\nu
    \oplistund
       \forall p'\tightin\DOM{\Gamma(x)}\stopq
         x\nu
         \tightequal
         \replpar
           {x\mu_{0,q}}
           {p''}
           {r_{1,q p' p''}\mu_{1,q p' p''}}
           {(p',p'')\tightin\Gamma(x)}
    }
.}
\\
\underline{Proof of Claim~7:}
\\
In case of \bigmath{\DOM{\Gamma(x)}\tightequal\emptyset} we define
\bigmath{x\nu:=x\mu_{0,q}.}
If there is some \math{p'} such that 
\bigmath{\DOM{\Gamma(x)}\tightequal\{p'\}}
we define 
\bigmath{
  x\nu
  :=
  \replpar
    {x\mu_{0,q}}
    {p''}
    {r_{1,q p' p''}\mu_{1,q p' p''}}
    {(p',p'')\tightin\Gamma(x)}
.}
This is appropriate since due to 
\bigmath{
  \forall(p',p'')\tightin\Gamma(x)\stopq
    x\mu_{0,q}/p''
    \tightequal 
    l_{0,q}\mu_{0,q}/p' p''
    \tightequal 
    u/q p' p''
    \tightequal 
    l_{1,q p' p''}\mu_{1,q p' p''}
}
we have
\\\LINEmath{
  \begin{array}{l@{}l@{}l}
  x\mu_{0,q}&
  \tightequal&
  \replpar
    {x\mu_{0,q}}
    {p''}
    {l_{1,q p' p''}\mu_{1,q p' p''}}
    {(p',p'')\tightin\Gamma(x)}
  \redparaindex{\omega+n_1+1}\\&&
  \replpar
     {x\mu_{0,q}}
     {p''}
     {r_{1,q p' p''}\mu_{1,q p' p''}}
     {(p',p'')\tightin\Gamma(x)}
  \tightequal
  x\nu.  
  \end{array}
}
\\
Finally, in case of \bigmath{\CARD{\DOM{\Gamma(x)}}\succ1,} \math{l_{0,q}} is
not linear in \math x. By the conditions of our lemma and Claim~5 this implies
\bigmath{x\tightin\Vcons.}
Since there is some \math{(p',p'')\in\Gamma(x)} with
\bigmath{
    x\mu_{0,q}/p''
    \tightequal 
    l_{1,q p' p''}\mu_{1,q p' p''}
}
this implies 
\bigmath{
   l_{1,q p' p''}\mu_{1,q p' p''}\tightin\tcc
}
and then 
\bigmath{
   l_{1,q p' p''}\tightin\tcs
}
which contradicts Claim~5.%
\QED{Claim~7}

\noindent
\underline{Claim~8:}
\bigmath{
  l_{0,q}\nu
  \tightequal 
  w_1/q
.}
\\
\underline{Proof of Claim~8:}
\\
By Claim~7 we get
\bigmath{
  w_1/q
  \tightequal
  \replpar 
    {u/q}
    {p' p''}
    {r_{1,q p' p''}\mu_{1,q p' p''}}
    {\exists x\tightin\V\stopq(p',p'')\tightin\Gamma(x)}
  \tightequal\\
  \replpar
    {\replpar
       {l_{0,q}}
       {p'}
       {x\mu_{0,q}}
       {l_{0,q}/p'\tightequal x\tightin\V}
    }
    {p' p''}
    {r_{1,q p' p''}\mu_{1,q p' p''}}
    {\exists x\tightin\V\stopq(p',p'')\tightin\Gamma(x)}
  \tightequal\\
  \replpar
    {l_{0,q}}
    {p'}
    {\replpar
       {x\mu_{0,q}}
       {p''}
       {r_{1,q p' p''}\mu_{1,q p' p''}}
       {(p',p'')\tightin\Gamma(x)}}
    {l_{0,q}/p'\tightequal x\tightin\V}
  \tightequal\\
  \replpar
    {l_{0,q}}
    {p'}
    {x\nu}
    {l_{0,q}/p'\tightequal x\tightin\V}
  \tightequal
  l_{0,q}\nu
.}\QED{Claim~8}

\noindent
\underline{Claim~9:}
\bigmath{
  w_0/q
  \redparaindex{\omega+n_1+1}
  r_{0,q}\nu
.}
\\
\underline{Proof of Claim~9:} 
Since 
\bigmath{
  w_0/q
  \tightequal
  r_{0,q}\mu_{0,q}
,} 
this follows directly from Claim~7.%
\QED{Claim~9}

\noindent
By claims 8 and 9 it now suffices to show
\bigmath{
  l_{0,q}\nu
  \redindex{\omega+n_0+1}
  r_{0,q}\nu
,}
which again follows from 
Lemma~\ref{lemma invariance of fulfilledness}
since \RX\ is \math\omega-shallow confluent up to
\bigmath{(n_1\tight+1)\plusomega n_0}
by our induction hypothesis 
and since
\bigmath{\forall x\tightin\V\stopq x\mu_{0,q}\refltransindex{\omega+n_1+1}x\nu}
by Claim~7 and Corollary~\ref{corollary parallel one}.%
\\\Qeddouble{``The variable overlap (if any) case''}

\pagebreak

\yestop
\noindent
\underline{\underline{``The critical peak case'':
There is some \math{p\in \Pi_1'\tightcap\TPOS{l_{0,q}}}
with \math{l_{0,q}/p\tightnotin\V}:}}
\begin{diagram}
l_{0,q}\mu_{0,q}&\rredindex{\omega+n_1+1,\,p}&u'
&&&\rredparaindex{\omega+n_1+1,\,\Pi_1'\setminus\{p\}}
&&&w_1/q
\\&&\dredparaindex{\omega+n_0+1}
&&&
&&&\drefltransindex{\omega+n_0+1}
\\\dredindex{\omega+n_0+1,\,\emptyset}&&v_1
&\rrefltransindex{\omega}&\circ&\rredparaindex{\omega+n_1+1}
&\circ&\rrefltransindex{\omega+n_1}&v_1'
\\&&\drefltransindex{\omega+n_0}
&&&
&&&\drefltransindex{\omega+n_0}
\\w_0/q&\rrefltransindex{\omega}&v_2
&\rrefltransindex{\omega}&\circ&\rredparaindex{\omega+n_1+1}
&\circ&\rrefltransindex{\omega+n_1}&\circ
\end{diagram}
\underline{Claim~10:}
\bigmath{p\tightnotequal\emptyset.}
\\
\underline{Proof of Claim~10:}
If \bigmath{p\tightequal\emptyset,} then
\bigmath{\emptyset\tightin\Pi_1',} then
\bigmath{q\tightin\Pi_1,} which contradicts our global case assumption.%
\QED{Claim~10}

\noindent
Let \math{\xi\in\SUBST\V\V} be a bijection with 
\bigmath{
  \xi[\VAR{\kurzregelindex{1,q p}}]\cap\VAR{\kurzregelindex{0,q}}
  =
  \emptyset
.}
\\
Define
\bigmath{
  \Y
  :=
  \xi[\VAR{\kurzregelindex{1,q p}}]\cup\VAR{\kurzregelindex{0,q}}
.}
\\
Let \math{\varrho\in\Xsubst} be given by
$\ x\varrho=
\left\{\begin{array}{@{}l@{}l@{}}
  x\mu_{0,q}        &\mbox{ if }x\in\VAR{\kurzregelindex{0,q}}\\
  x\xi^{-1}\mu_{1,q p}&\mbox{ else}\\
\end{array}\right\}
\:(x\tightin\V)$.
\\
By
\math{
  l_{1,q p}\xi\varrho
  \tightequal 
  l_{1,q p}\xi\xi^{-1}\mu_{1,q p}
  \tightequal 
  u/q p
  \tightequal 
  l_{0,q}\mu_{0,q}/p  
  \tightequal 
  l_{0,q}\varrho/p
  \tightequal 
  (l_{0,q}/p)\varrho
}
\\
let
\math{
  \sigma:=\minmgu{\{(l_{1,q p}\xi}{l_{0,q}/p)\},\Y}
}
and
\math{\varphi\in\Xsubst}
with
\math{
  \domres{\inpit{\sigma\varphi}}\Y
  \tightequal
  \domres\varrho\Y
}.
\\
Define 
\math{
  u':=  
  \repl{l_{0,q}\mu_{0,q}}
       {p}
       {r_{1,q p}\mu_{1,q p}}
}.
We get
\\\LINEmath{
  \arr{{l@{}l}
    u'\tightequal
    &
    \repl
      {\replpar
         {u/q}
         {p'}
         {l_{1,q p'}\mu_{1,q p'}}
         {p'\tightin\Pi_1'\tightsetminus\{p\}}}
      {p}
      {r_{1,q p}\mu_{1,q p}}
    \redparaindex{\omega+n_1+1,\Pi_1'\setminus\{p\}}
    \\&
    \replpar{u/q}{p'}{r_{1,q p'}\mu_{1,q p'}}{p'\tightin\Pi_1'}    
    \tightequal
    w_1/q
  .
  }
}
\\
If 
\bigmath{
  \repl{l_{0,q}}{p}{r_{1,q p}\xi}\sigma
  \tightequal
  r_{0,q}\sigma
,}
then the proof is finished due to 
\\\LINEmath{
  w_0/q
  \tightequal
  r_{0,q}\mu_{0,q}
  \tightequal
  r_{0,q}\sigma\varphi
  \tightequal
  \repl{l_{0,q}}{p}{r_{1,q p}\xi}\sigma\varphi
  \tightequal
  u'
  \redparaindex{\omega+n_1+1,\Pi_1'\setminus\{p\}}
  w_1/q
.}
\\
Otherwise 
we have
\bigmath{
  (\,
   (\repl{l_{0,q}}{p}{r_{1,q p}\xi}\sigma,
    C_{1,q p}\xi\sigma,    
    1),\penalty-1\,
   (r_{0,q}\sigma,
    C_{0,q}\sigma,
    1),\penalty-1\,
    l_{0,q}\sigma,\penalty-1\,
    p\,)
  \in{\rm CP}(\R)
}
(due to Claim~5);
\bigmath{p\tightnotequal\emptyset}
(due to Claim~10);
\bigmath{C_{1,q p}\xi\sigma\varphi=C_{1,q p}\mu_{1,q p}}
is fulfilled \wrt\ \redindex{\omega+n_1};
\bigmath{C_{0,q}\sigma\varphi=C_{0,q}\mu_{0,q}}
is fulfilled \wrt\ \redindex{\omega+n_0}.
Since 
\bigmath{
  \forall\delta\tightprec(n_1\tight+1)\plusomega(n_0\tight+1)\stopq
  \mbox{\RX\ is \math\omega-shallow confluent up to }\delta
}
(by our induction hypothesis) 
due to our assumed \math\omega-shallow parallel closedness 
(matching the definition's \math{n_0} to our \math{n_1\tight+1}
                   and its \math{n_1} to our \math{n_0\tight+1})
we have
\bigmath{
  u'
  \tightequal
  \repl{l_{0,q}}{p}{r_{1,q p}\xi}\sigma\varphi
  \penalty-1
  \redparaindex{\omega+n_0+1}
  \penalty-1
  v_1
  \penalty-1
  \refltransindex{\omega+n_0}
  v_2
  \antirefltransindex{\omega}
  r_{0,q}\sigma\varphi
  \tightequal
  r_{0,q}\mu_{0,q}
  \tightequal
  w_0/q
}
for some \math{v_1}, \math{v_2}.
We then have
\bigmath{
  v_1
  \antiredparaindex{\omega+n_0+1,\Pi''} 
  u'
  \redparaindex{\omega+n_1+1,\Pi_1'\setminus\{p\}}
  w_1/q
}
for some \math{\Pi''}.
By 
\\\bigmath{
  \displaystyle
  \sum_{p''\in\Omega(\Pi'',\Pi_1'\setminus\{p\})}
  \lambda(u'/p'')
  \ \ \preceq
  \sum_{p''\in\Pi_1'\setminus\{p\}}
  \lambda(u'/p'')
  \ \ =
  \sum_{p''\in\Pi_1'\setminus\{p\}}
  \lambda(u/q p'')
  \ \ \prec
  \sum_{p''\in\Pi_1'}
  \lambda(u/q p'')
  \ \ =
}\\\bigmath{\displaystyle
  \sum_{p'\in q\Pi_1'}
  \lambda(u/p')
  \ \ =
  \sum_{p'\in\Omega(\{q\},\Pi_1)}
  \lambda(u/p')
  \ \ \preceq
  \sum_{p'\in\Omega(\Pi_0,\Pi_1)}
  \lambda(u/p')
,}
due to our second induction level 
we get some \math{v_1'} with 
\bigmath{
  v_1
  \refltransindex{\omega}
  \tight\circ
  \redparaindex{\omega+n_1+1}
  \tight\circ
  \refltransindex{\omega+n_1}
  v_1'
  \antirefltransindex{\omega+n_0+1}
  w_1/q
.}
Finally by our induction hypothesis that
  \bigmath{ 
    \refltransindex{\omega}
    \tight\circ
    \redparaindex{\omega+n_1+1}
    \tight\circ
    \refltransindex{\omega+n_1}
  }
  strongly commutes over
  \refltransindex{\omega+n_0}
the peak at \math{v_1} can be closed according to 
\bigmath{
  v_2
  \refltransindex{\omega}
  \tight\circ
  \redparaindex{\omega+n_1}
  \tight\circ
  \refltransindex{\omega+n_1}
  \circ
  \antirefltransindex{\omega+n_0}
  v_1'
.}%
\\
\Qeddouble{``The critical peak case''}\QEDtriple{``\math{q\tightnotin\Pi_1}''}

\pagebreak

\noindent
\underline{\underline{\underline{\math{q\tightin\Pi_1}:}}}
Define \math{\Pi_0':=\setwith{p}{q p\tightin\Pi_0}}.
We have two cases:

\yestop
\noindent
\underline{\underline{``The second variable overlap (if any) case'':
\math{
  \forall p\tightin\Pi_0'\tightcap\TPOS{l_{1,q}}\stopq
    l_{1,q}/p\tightin\V
}:}}
\begin{diagram}
l_{1,q}\mu_{1,q}&&&\rredindex{\omega+n_1+1,\,\emptyset}&&&w_1/q
\\&&&&&&\dequal
\\\dredparaindex{\omega+n_0+1}&&&&&&r_{1,q}\mu_{1,q}
\\&&&&&&\dredparaindex{\omega+n_0+1}
\\w_0/q&\requal&l_{1,q}\nu&&\rredindex{\omega+n_1+1}&&r_{1,q}\nu
\end{diagram}
\noindent
Define a function \math\Gamma\ on \V\ by (\math{x\tightin\V}):
\bigmath{
  \Gamma(x):=
  \setwith{(p',p'')}
          {l_{1,q}/p'\tightequal x\ \wedge\ p' p''\in\Pi_0'}
.}

\noindent
\underline{Claim~11:}
There is some \math{\nu\in\Xsubst} with
\\\LINEmath{
  \forall x\in\V\stopq
    \inparenthesesoplist{
       x\nu
       \antiredparaindex{\omega+n_0+1}
       x\mu_{1,q}
    \oplistund
       \forall p'\tightin\DOM{\Gamma(x)}\stopq
         \replpar
           {x\mu_{1,q}}
           {p''}
           {r_{0,q p' p''}\mu_{0,q p' p''}}
           {(p',p'')\tightin\Gamma(x)}
         \tightequal
         x\nu
    }
.}
\\
\underline{Proof of Claim~11:}
\\
In case of \bigmath{\DOM{\Gamma(x)}\tightequal\emptyset} we define
\bigmath{x\nu:=x\mu_{1,q}.}
If there is some \math{p'} such that 
\bigmath{\DOM{\Gamma(x)}\tightequal\{p'\}}
we define 
\bigmath{
  x\nu
  :=
  \replpar
    {x\mu_{1,q}}{p''}{r_{0,q p' p''}\mu_{0,q p' p''}}{(p',p'')\tightin\Gamma(x)}
.}
This is appropriate since due to 
\bigmath{
  \forall(p',p'')\tightin\Gamma(x)\stopq
    x\mu_{1,q}/p''
    \tightequal 
    l_{1,q}\mu_{1,q}/p' p''
    \tightequal 
    u/q p' p''
    \tightequal 
    l_{0,q p' p''}\mu_{0,q p' p''}
}
we have
\\\LINEmath{
  \begin{array}{l@{}l@{}l}
  x\mu_{1,q}&
  \tightequal&
  \replpar
    {x\mu_{1,q}}
    {p''}
    {l_{0,q p' p''}\mu_{0,q p' p''}}
    {(p',p'')\tightin\Gamma(x)}
  \redparaindex{\omega+n_0+1}\\&&
  \replpar
     {x\mu_{1,q}}
     {p''}
     {r_{0,q p' p''}\mu_{0,q p' p''}}
     {(p',p'')\tightin\Gamma(x)}
  \tightequal
  x\nu.  
  \end{array}
}
\\
Finally, in case of \bigmath{\CARD{\DOM{\Gamma(x)}}\succ1,} \math{l_{1,q}} is
not linear in \math x. 
By the conditions of our lemma and Claim~5 this implies
\bigmath{x\tightin\Vcons.}
Since there is some \math{(p',p'')\in\Gamma(x)} with
\bigmath{
    x\mu_{1,q}/p''
    \tightequal 
    l_{0,q p' p''}\mu_{0,q p' p''}
}
this implies 
\bigmath{
   l_{0,q p' p''}\mu_{0,q p' p''}\tightin\tcc
}
and then 
\bigmath{
   l_{0,q p' p''}\tightin\tcs
}
which contradicts Claim~5.%
\QED{Claim~11}

\noindent
\underline{Claim~12:}
\bigmath{w_0/q\tightequal l_{1,q}\nu.}
\\
\underline{Proof of Claim~12:}
\\
By Claim~11 we get
\bigmath{
  w_0/q
  \tightequal
  \replpar 
    {u/q}
    {p' p''}
    {r_{0,q p' p''}\mu_{0,q p' p''}}
    {\exists x\tightin\V\stopq(p',p'')\tightin\Gamma(x)}
  \tightequal\\
  \replpar
    {\replpar
       {l_{1,q}}
       {p'}
       {x\mu_{1,q}}
       {l_{1,q}/p'\tightequal x\tightin\V}
    }
    {p' p''}
    {r_{0,q p' p''}\mu_{0,q p' p''}}
    {\exists x\tightin\V\stopq(p',p'')\tightin\Gamma(x)}
  \tightequal\\
  \replpar
    {l_{1,q}}
    {p'}
    {\replpar
       {x\mu_{1,q}}
       {p''}
       {r_{0,q p' p''}\mu_{0,q p' p''}}
       {(p',p'')\tightin\Gamma(x)}}
    {l_{1,q}/p'\tightequal x\tightin\V}
  \tightequal\\
  \replpar
    {l_{1,q}}
    {p'}
    {x\nu}
    {l_{1,q}/p'\tightequal x\tightin\V}
  \tightequal
  l_{1,q}\nu
.}\QED{Claim~12}

\noindent
\underline{Claim~13:}
\bigmath{
  r_{1,q}\nu
  \antiredparaindex{\omega+n_0+1}
  w_1/q
.}
\\
\underline{Proof of Claim~13:} 
Since \bigmath{r_{1,q}\mu_{1,q}\tightequal w_1/q,} 
this follows directly from Claim~11.%
\QED{Claim~13}

\noindent
By claims 12 and 13 
using Corollary~\ref{corollary parallel one} 
it now suffices to show
\bigmath{
  l_{1,q}\nu
  \redindex{\omega+n_1+1}
  r_{1,q}\nu
,}
which again follows from 
Claim~11,
Corollary~\ref{corollary parallel one},
Lemma~\ref{lemma invariance of fulfilledness}
(matching 
 its \math{n_0} to our \math{n_0\tight+1} and 
 its \math{n_1} to our \math{n_1}),
and our induction hypothesis that \RX\ is \math\omega-shallow confluent up to
\bigmath{
  (n_0\tight+1)\plusomega n_1
.}%
\\\Qeddouble{``The second variable overlap (if any) case''}

\pagebreak

\yestop
\noindent
\underline{\underline{``The second critical peak case'':
There is some \math{p\in \Pi_0'\tightcap\TPOS{l_{1,q}}}
with \math{l_{1,q}/p\tightnotin\V}:}}
\begin{diagram}
l_{1,q}\mu_{1,q}
&&&
&\rredindex{\omega+n_1+1,\,\emptyset}&&
&&w_1/q
\\\dredindex{\omega+n_0+1,\,p}
&&&
&&&
&&\drefltransindex{\omega+n_0+1}
\\u'
&&&\rredparaindex{\omega+n_1+1}
&&&v_1
&\rrefltransindex{\omega+n_1}&v_2
\\\dredparaindex{\omega+n_0+1,\,\Pi_0'\setminus\{p\}}
&&&
&&&\drefltransindex{\omega+n_0+1}
&&\drefltransindex{\omega+n_0+1}
\\w_0/q
&\rrefltransindex{\omega}&\circ&\rredparaindex{\omega+n_1+1}
&\circ&\rrefltransindex{\omega+n_1}&v_1'
&\rrefltransindex{\omega+n_1}&\circ
\end{diagram}
Let \math{\xi\in\SUBST\V\V} be a bijection with 
\bigmath{
  \xi[\VAR{\kurzregelindex{0,q p}}]\cap\VAR{\kurzregelindex{1,q}}
  =
  \emptyset
.}
\\
Define
\bigmath{
  \Y
  :=
  \xi[\VAR{\kurzregelindex{0,q p}}]\cup\VAR{\kurzregelindex{1,q}}
.}
\\
Let \math{\varrho\in\Xsubst} be given by
$\ x\varrho=
\left\{\begin{array}{@{}l@{}l@{}}
  x\mu_{1,q}        &\mbox{ if }x\in\VAR{\kurzregelindex{1,q}}\\
  x\xi^{-1}\mu_{0,q p}&\mbox{ else}\\
\end{array}\right\}
\:(x\tightin\V)$.
\\
By
\math{
  l_{0,q p}\xi\varrho
  \tightequal 
  l_{0,q p}\xi\xi^{-1}\mu_{0,q p}
  \tightequal 
  u/q p
  \tightequal 
  l_{1,q}\mu_{1,q}/p  
  \tightequal 
  l_{1,q}\varrho/p
  \tightequal 
  (l_{1,q}/p)\varrho
}
\\
let
\math{
  \sigma:=\minmgu{\{(l_{0,q p}\xi}{l_{1,q}/p)\},\Y}
}
and
\math{\varphi\in\Xsubst}
with
\math{
  \domres{\inpit{\sigma\varphi}}\Y
  \tightequal
  \domres\varrho\Y
}.
\\
Define 
\math{
  u':=  
  \repl{l_{1,q}\mu_{1,q}}
       {p}
       {r_{0,q p}\mu_{0,q p}}
}. 
We get
\\\LINEmath{
  \begin{array}{l@{}l@{}l}
  w_0/q&
  \tightequal&
  \replpar{u/q}{p'}{r_{0,q p'}\mu_{0,q p'}}{p'\tightin\Pi_0'}
  \antiredparaindex{\omega+n_0+1,\Pi_0'\setminus\{p\}}
  \\&&
  \repl
    {\replpar
       {u/q}{p'}{l_{0,q p'}\mu_{0,q p'}}{p'\tightin\Pi_0'\tightsetminus\{p\}}}
    {p}
    {r_{0,q p}\mu_{0,q p}}
  \tightequal
  u'
  .  
  \end{array}
}  
\\
If 
\bigmath{
  \repl{l_{1,q}}{p}{r_{0,q p}\xi}\sigma
  \tightequal
  r_{1,q}\sigma
,}
then the proof is finished due to 
\\\linemath{
  w_0/q
  \antiredparaindex{\omega+n_0+1,\Pi_0'\setminus\{p\}}
  u'
  \tightequal
  \repl{l_{1,q}}{p}{r_{0,q p}\xi}\sigma\varphi
  \tightequal
  r_{1,q}\sigma\varphi
  \tightequal
  r_{1,q}\mu_{1,q}
  \tightequal
  w_1/q.
}
Otherwise 
we have
\bigmath{
  (\,
   (\repl{l_{1,q}}{p}{r_{0,q p}\xi}\sigma,
    C_{0,q p}\xi\sigma,    
    1),\penalty-1\,
   (r_{1,q}\sigma,
    C_{1,q}\sigma,
    1),\penalty-1\,
    l_{1,q}\sigma,\penalty-1\,
    p\,)
  \in{\rm CP}(\R)
}
(due to Claim~5);
\bigmath{C_{0,q p}\xi\sigma\varphi=C_{0,q p}\mu_{0,q p}}
is fulfilled \wrt\ \redindex{\omega+n_0};
\bigmath{C_{1,q}\sigma\varphi=C_{1,q}\mu_{1,q}}
is fulfilled \wrt\ \redindex{\omega+n_1}.
Since 
\bigmath{
  \forall\delta\tightprec(n_0\tight+1)\plusomega(n_1\tight+1)\stopq
  \mbox{\RX\ is \math\omega-shallow confluent up to }\delta
}
(by our induction hypothesis) 
due to our assumed \math\omega-shallow noisy parallel joinability 
(matching the definition's \math{n_0} to our \math{n_0\tight+1}
                   and its \math{n_1} to our \math{n_1\tight+1}
)
we have
\bigmath{
  u'
  \tightequal
  \repl{l_{1,q}}{p}{r_{0,q p}\xi}\sigma\varphi
  \redparaindex{\omega+n_1+1}
  \penalty-1
  v_1
  \refltransindex{\omega+n_1}
  \penalty-1
  v_2
  \antirefltransindex{\omega+n_0+1}
  \penalty-1
  r_{1,q}\sigma\varphi
  \tightequal
  r_{1,q}\mu_{1,q}
  \tightequal
  w_1/q
}
for some \math{v_1}, \math{v_2}.
We then have
\bigmath{
  w_0/q
  \antiredparaindex{\omega+n_0+1,\Pi_0'\setminus\{p\}}
  u'
  \redparaindex{\omega+n_1+1,\Pi''} 
  v_1
}
for some \math{\Pi''}.
Since
\bigmath{\displaystyle
  \sum_{p''\in\Omega(
    \Pi_0'\setminus\{p\}
    ,
    \Pi''
    )}
  \lambda(u'/p'')
  \ \ \preceq
  \sum_{p''\in\Pi_0'\setminus\{p\}}
  \lambda(u'/p'')
  \ \ =
  \sum_{p''\in\Pi_0'\setminus\{p\}}
  \lambda(u/q p'')
  \ \ \prec
  \sum_{p''\in\Pi_0'}
  \lambda(u/q p'')
  \ \ =
  \sum_{p'\in q\Pi_0'}
  \lambda(u/p')
  \ \ =
  \sum_{p'\in\Omega(
    \Pi_0
    ,
    \{q\}
    )}
  \lambda(u/p')
  \ \ \preceq
  \sum_{p'\in\Omega(\Pi_0,\Pi_1)}
  \lambda(u/p')
}
due to our second induction level 
we get some \math{v_1'} with 
\bigmath{
  w_0/q
  \refltransindex{\omega}
  \tight\circ
  \redparaindex{\omega+n_1+1}
  \tight\circ
  \refltransindex{\omega+n_1}
  v_1'
  \antirefltransindex{\omega+n_0+1}
  v_1
.} 
Finally the peak at \math{v_1} can be closed according to
\bigmath{
  v_1'
  \refltransindex{\omega+n_1}
  \circ
  \antirefltransindex{\omega+n_0+1}
  v_2
}
by our induction hypothesis saying that \RX\ is
\math\omega-shallow confluent up to \math{(n_0\tight+1)\plusomega n_1}.%
\\\Qeddouble{``The second critical peak case''}
\end{proofparsepqed}

\pagebreak

\begin{proofparsepqed}{Lemma~\ref{lemma parallel closed first level three}}
For \math{n\prec\omega} we are going to show by induction on \math n
the following property\footroom:
\\\LINEmath{
    w_0
    \antiredindex\omega
    u
    \redparaindex{\omega+n}
    w_1
  \quad\implies\quad
    w_0
    \refltransindex{\omega}
    \tight\circ
    \redparaindex{\omega+n}
    \tight\circ
    \refltransindex{\omega[+(n\monus1)]}
    \circ
    \antirefltransindex\omega
    w_1
.}
\begin{diagram}
u&&&\rredparaindex{\omega+n}&&&w_1
\\\dredindex{\omega}&&&&&&\drefltransindex{\omega}\\
w_0&\rrefltransindex{\omega}&\circ&\rredparaindex{\omega+n}&\circ&
\rrefltransindex{\omega[+(n\monus1)]}&\circ\\
\end{diagram}

\noindent
\underline{Claim~1:}
If the above property holds for a fixed \math{n\prec\omega}, 
and 
\\\bigmath{
  \forall k\tightprec n\stopq
     (\RX\mbox{ is \math\omega-shallow confluent up to }k)
,} then
\bigmath{
    \refltransindex{\omega}
    \tight\circ
    \redparaindex{\omega+n}
    \tight\circ
    \refltransindex{\omega[+(n\monus1)]}
}
strongly commutes over
\refltransindex\omega.
\\
\underline{Proof of Claim~1:}
By Lemma~\ref{lemma strong commutation one copy} it suffices to show that
\bigmath{
    \refltransindex{\omega}
    \tight\circ
    \redparaindex{\omega+n}
    \tight\circ
    \refltransindex{\omega[+(n\monus1)]}
}
strongly commutes over
\redindex{\omega}.
Assume
\bigmath{
    u''
    \antiredindex{\omega}    
    u'
    \refltransindex{\omega}
    u
    \redparaindex{\omega+n}
    w_1
    \refltransindex{\omega[+(n\monus1)]}
    w_2
}
(\cf\ diagram below).
By the strong confluence of \redindex{\omega} assumed for our lemma
we can close the peak
\bigmath{
    u''
    \antiredindex{\omega}    
    u'
    \refltransindex{\omega}
    u
}
according to    
\bigmath{
    u''
    \refltransindex{\omega}
    w_0
    \antionlyonceindex{\omega}    
    u
}
for some \math{w_0}.
By the above property there is some \math{w_1'}
with
\bigmath{
    w_0
    \refltransindex{\omega}
    \tight\circ
    \redparaindex{\omega+n}
    \tight\circ
    \refltransindex{\omega[+(n\monus1)]}
    w_1'
    \antirefltransindex\omega
    w_1
.}
We only have to show that we can close the peak
\bigmath{ 
    w_1'
    \antirefltransindex\omega
    w_1
    \refltransindex{\omega[+(n\monus1)]}
    w_2
}
according to 
\bigmath{
    w_1'
    \refltransindex{\omega[+(n\monus1)]}
    \circ
    \antirefltransindex\omega
    w_2
.}
{[In case of \bigmath{n\tightequal0:}]}
This is possible due to confluence of \redindex\omega.
{[Otherwise we have \bigmath{n\monus1\tightprec n} and due to the 
assumed \math\omega-shallow confluence up to \math{n\monus1}
this is possible again.]}
\begin{diagram}
u'&\rrefltransindex{\omega}&
u&&&\rredparaindex{\omega+n}&&&w_1&\rrefltransindex{\omega[+(n\monus1)]}&w_2
\\\dredindex{\omega}&&\donlyonceindex{\omega}&&&&&&\drefltransindex{\omega}
&&\drefltransindex{\omega}
\\u''&\rrefltransindex{\omega}&
w_0&\rrefltransindex{\omega}&\circ&\rredparaindex{\omega+n}&\circ&
\rrefltransindex{\omega[+(n\monus1)]}&w_1'
&\rrefltransindex{\omega[+(n\monus1)]}&\circ
\end{diagram}
\Qed{Claim~1}

\yestop
\noindent
\underline{Claim~2:}
If the above property holds for a fixed \math{n\prec\omega}, and 
\\\bigmath{
  \forall k\tightprec n\stopq
     (\RX\mbox{ is \math\omega-shallow confluent up to }k)
,} 
then
\redindex{\omega+n} and \redindex\omega\ are commuting.
\\
\underline{Proof of Claim~2:}
\bigmath{
    \refltransindex{\omega}
    \tight\circ
    \redparaindex{\omega+n}
    \tight\circ
    \refltransindex{\omega[+(n\monus1)]}
}
and
\refltransindex\omega\
are commuting
by Lemma~\ref{lemma strong commutation one copy}
and Claim~1.
Since by Corollary~\ref{corollary parallel one}
and \lemmamonotonicinbeta\
we have
\bigmath{
  \redindex{\omega+n}
  \subseteq
  \refltransindex{\omega}
  \tight\circ
  \redparaindex{\omega+n}
  \tight\circ
  \refltransindex{\omega[+(n\monus1)]}
  \subseteq
  \refltransindex{\omega+n}
,}
now \redindex{\omega+n} and \redindex\omega\ are commuting, too.\QED{Claim~2}

\yestop
\noindent
\underline{Claim~3:}
If the above property holds for all 
\math{n\preceq m} for some \math{m\prec\omega}, then
\RX\ is \math\omega-shallow confluent up to \math{m}.
\\
\underline{Proof of Claim~3:}
By induction on \math{m} in \tightprec.
Assume
\bigmath{i\plusomega n\tightpreceq m}
and
\bigmath{
    w_0
    \antirefltransindex{\omega+i}
    u
    \refltransindex    {\omega+n}
    w_1
.}
By definition of `\plusomega' and 
\bigmath{i\plusomega n\tightprec\omega}
\wrog\ we have 
\bigmath{i\tightequal0}
and
\bigmath{n\tightpreceq m.}
By Claim~2 and our induction hypothesis we finally get
\bigmath{
  w_0\refltransindex{\omega+n}\circ\antirefltransindex{\omega} w_1
}
as desired.\QED{Claim~3}

\pagebreak

\yestop
\yestop
\noindent
Note that our property for is trivial for \bigmath{n\tightequal0} since
then by Corollary~\ref{corollary parallel one} we have
\bigmath{
  \redparaindex{\omega+n}=\redparaindex{\omega}\subseteq\refltransindex\omega
}
and \redindex\omega\ is confluent.

The benefit of 
claims 1 and 3 is twofold: 
First, they say that our lemma is valid if the above property
holds for all \math{n\prec\omega}.
Second, they strengthen the property when used as induction hypothesis. Thus 
(writing \math{n\tight+1} instead of \math{n} since we may assume
 \math{0\tightprec n})
it
now suffices to show
for
\math{
  n\prec\omega
}
that 
\\\linemath{
    w_0
    \antiredparaindex{\omega,\bar p_0}
    u
    \redparaindex{\omega+n+1,\Pi_1}
    w_1
}
together with our induction hypothesis 
that\headroom
\\\linenomath{\headroom
  \RX\ is \math\omega-shallow confluent up to \math n
}
\headroom
implies
\\\LINEmath{
    w_0
    \refltransindex{\omega}
    \tight\circ
    \redparaindex{\omega+n+1}
    \tight\circ
    \refltransindex{\omega[+n]}
    \circ
    \antirefltransindex\omega
    w_1
.}
\begin{diagram}
u&&&\rredparaindex{\omega+n+1,\,\Pi_1}&&&w_1
\\\dredindex{\omega,\,\bar p_0}&&&&&&\drefltransindex{\omega}
\\w_0&\rrefltransindex{\omega}&\circ
&\rredparaindex{\omega+n+1}&\circ&\rrefltransindex{\omega[+n]}&\circ
\end{diagram}

\noindent
There are 
\bigmath{\kurzregelindex{0,\bar p_0}\in\R}
and
\bigmath{\mu_{0,\bar p_0}\in\Xsubst}
such that
\bigmath{l_{0,\bar p_0}\tightin\tcs,}
\bigmath{u/\bar p_0\tightequal l_{0,\bar p_0}\mu_{0,\bar p_0},}
\math{\ C_{0,\bar p_0}\mu_{0,\bar p_0}} is fulfilled \wrt\ \redindex{\omega},
and
\bigmath{
  w_0\tightequal\repl{u}{\bar p_0}{r_{0,\bar p_0}\mu_{0,\bar p_0}}
.}

\Wrog\ let the positions of \math{\Pi_1} be maximal
in the sense that for any \math{p\in\Pi_1}
and \math{\Xi\subseteq \TPOS u\tightcap(p\N^+)}
we do not have 
\math{\ 
  u\redparaindex{\omega+n+1,(\Pi_1\setminus\{p\})\cup\Xi}w_1
}
anymore.
Then for each 
\math{p\in\Pi_1} there are
\bigmath{\kurzregelindex{1,p}\in\R}
and
\bigmath{\mu_{1,p}\in\Xsubst}
such that
\bigmath{u/p\tightequal l_{1,p}\mu_{1,p},}
\bigmath{r_{1,p}\mu_{1,p}\tightequal w_1/p},
\math{\ C_{1,p}\mu_{1,p}} is fulfilled \wrt\ \redindex{\omega+n},
and
\bigmath{
  w_1\tightequal\replpar{u}{p}{r_{1,p}\mu_{1,p}}{p\tightin\Pi_1}
.}

\pagebreak

\yestop
\noindent
\underline{Claim~5:}
We may assume 
\bigmath{
  \forall p\tightin\Pi_1\stopq
    l_{1,p}\tightnotin\tcs
.}
\\
\underline{Proof of Claim~5:}
Define \math{\Xi:=\setwith{p\tightin\Pi_1}{l_{1,p}\tightin\tcs}}
and
\math{
  u'
  :=
  \replpar{u}{p}{r_{1,p}\mu_{1,p}}{p\tightin\Pi_1\tightsetminus\Xi}
}.
If we have succeeded with our proof under the assumption of Claim~5,
then we have shown 
\bigmath{
  w_0 
  \refltransindex{\omega}
  \tight\circ
  \redparaindex{\omega+n+1}
  \tight\circ
  \refltransindex{\omega[+n]}
  v'
  \antirefltransindex\omega
  u'
}
for some \math{v'}
(\cf\ diagram below).
By Lemma~\ref{lemma invariance of fulfilledness two} 
(matching both its \math\mu\ and \math\nu\ to our \math{\mu_{1,p}})
we get
\bigmath{
  \forall p\tightin\Xi\stopq
    l_{1,p}\mu_{1,p}\redindex\omega r_{1,p}\mu_{1,p}
.}
Thus from
\bigmath{
  v'
  \antirefltransindex\omega
  u'
  \refltransindex\omega
  w_1
}
we get 
\bigmath{
%  w_0 
%  \redparaindex{\omega+n+1}
%  \tight\circ
%  \refltransindex{\omega[+n]}
  v'
  \refltransindex\omega
  \circ
  \antirefltransindex\omega
  w_1
}
by confluence of \redindex\omega.
\begin{diagram}
u&&&\rredparaindex{\omega+n+1,\,\Pi_1\setminus\Xi}
&&&u'
&\rrefltransindex{\omega}&w_1
\\\dredindex{\omega}&&
&&&&\drefltransindex{\omega}
&&\drefltransindex{\omega}
\\w_0&\rrefltransindex{\omega}&\circ&\rredparaindex{\omega+n+1}&\circ
&\rrefltransindex{\omega[+n]}&v'
&\rrefltransindex{\omega}&\circ
\end{diagram}
\Qed{Claim~5}

\yestop
\noindent
Now we start a second level of induction on
\bigmath{
  \CARD{\Pi_1}
}
in \tightprec.

\noindent
Define the set of top positions by
\\\linemath{
  \displaystyle
  \Theta
  :=
      \setwith
      {p\tightin\{\bar p_0\}\tightcup\Pi_1}
      {\neg\exists q\tightin\{\bar p_0\}\tightcup\Pi_1\stopq
           \exists q'\tightin\N^+\stopq
             p\tightequal q q'
      }
.}
Since the prefix ordering is wellfounded we have
\bigmath{
  \forall p\tightin\{\bar p_0\}\tightcup\Pi_1\stopq
  \exists q\tightin\Theta\stopq
  \exists q'\tightin\N^\ast\stopq
    p\tightequal q q'
.}
It now suffices to show for all \math{q\in\Theta}
\\\linemath{\headroom\footroom
    w_0/q
    \refltransindex{\omega}
    \tight\circ
    \redparaindex{\omega+n+1}
    \tight\circ
    \refltransindex{\omega[+n]}
    \circ
    \antirefltransindex{\omega}
    w_1/q
}
because then we have 
\bigmath{
  w_0
  \tightequal
  \replpar{w_0}{q}{w_0/q}{q\tightin\Theta}
  \tightequal
  \replpar{\repl{u}{\bar p_0}{r_{0,\bar p_0}\mu_{0,\bar p_0}}}
          {q}{w_0/q}{q\tightin\Theta}
  \tightequal
  \replpar{u}{q}{w_0/q}{q\tightin\Theta}
    \refltransindex{\omega}
    \tight\circ
    \redparaindex{\omega+n+1}
    \tight\circ
    \refltransindex{\omega[+n]}
    \circ
    \antirefltransindex{\omega}
  \replpar{u}{q}{w_1/q}{q\tightin\Theta}
  \tightequal
  \\
  \replpar{\replpar{u}{p}{r_{1,p}\mu_{1,p}}{p\tightin\Pi_1}}
          {q}{w_1/q}{q\tightin\Theta}
  \tightequal
  \replpar{w_1}{q}{w_1/q}{q\tightin\Theta}
  \tightequal
  w_1
.}

\noindent
Therefore we are left with the following two cases for \math{q\in\Theta}:

\pagebreak

\yestop
\yestop
\noindent
\underline{\underline{\underline{\math{q\tightnotin\Pi_1}:}}}
Then \bigmath{q\tightequal \bar p_0.}
Define \math{\Pi_1':=\setwith{p}{q p\tightin\Pi_1}}.
We have two cases:

\noindent
\underline{\underline{``The variable overlap (if any) case'':
\math{
  \forall p\tightin\Pi_1'\tightcap\TPOS{l_{0,q}}\stopq
    l_{0,q}/p\tightin\V
}:}}
\begin{diagram}
l_{0,q}\mu_{0,q}&&&\rredparaindex{\omega+n+1}&&&w_1/q
\\&&&&&&\drefltransindex{\omega}
\\\dredindex{\omega,\,\emptyset}&&&&&&l_{0,q}\nu
\\&&&&&&\dredindex{\omega}
\\w_0/q&\requal&r_{0,q}\mu_{0,q}&\rredparaindex{\omega+n+1}&\circ
&\rrefltransindex{\omega}&r_{0,q}\nu
\end{diagram}
Define a function \math\Gamma\ on \V\ by (\math{x\tightin\V}):
\bigmath{
  \Gamma(x):=
  \setwith{(p',p'')}
          {l_{0,q}/p'\tightequal x\ \wedge\ p' p''\in\Pi_1'}
.}

\noindent
\underline{Claim~7:}
There is some \math{\nu\in\Xsubst} with
\\\LINEmath{
  \forall x\in\V\stopq
    \inparenthesesoplist{
       x\mu_{0,q}
       \redparaindex{\omega+n+1}
       \tight\circ
       \refltransindex\omega
       x\nu
    \oplistund
       \forall p'\tightin\DOM{\Gamma(x)}\stopq
         x\nu
         \antirefltransindex\omega
         \replpar
           {x\mu_{0,q}}
           {p''}
           {r_{1,q p' p''}\mu_{1,q p' p''}}
           {(p',p'')\tightin\Gamma(x)}
    }
.}
\\
\underline{Proof of Claim~7:}
\\
In case of \bigmath{\DOM{\Gamma(x)}\tightequal\emptyset} we define
\bigmath{x\nu:=x\mu_{0,q}.}
If there is some \math{p'} such that 
\bigmath{\DOM{\Gamma(x)}\tightequal\{p'\}}
we define 
\bigmath{
  x\nu
  :=
  \replpar
    {x\mu_{0,q}}{p''}{r_{1,q p' p''}\mu_{1,q p' p''}}{(p',p'')\tightin\Gamma(x)}
.}
This is appropriate since due to 
\bigmath{
  \forall(p',p'')\tightin\Gamma(x)\stopq
    x\mu_{0,q}/p''
    \tightequal 
    l_{0,q}\mu_{0,q}/p' p''
    \tightequal 
    u/q p' p''
    \tightequal 
    l_{1,q p' p''}\mu_{1,q p' p''}
}
we have
\\\LINEmath{
  \begin{array}{l@{}l@{}l}
  x\mu_{0,q}&
  \tightequal&
  \replpar
    {x\mu_{0,q}}
    {p''}
    {l_{1,q p' p''}\mu_{1,q p' p''}}
    {(p',p'')\tightin\Gamma(x)}
  \redparaindex{\omega+n+1}\\&&
  \replpar
     {x\mu_{0,q}}
     {p''}
     {r_{1,q p' p''}\mu_{1,q p' p''}}
     {(p',p'')\tightin\Gamma(x)}
  \tightequal
  x\nu.  
  \end{array}
}
\\
Finally, in case of \bigmath{\CARD{\DOM{\Gamma(x)}}\succ1,} 
\bigmath{l_{0,q}\tightequal l_{0,\bar p_0}\tightin\tcs}
is not linear in \math x. By the conditions of our lemma, this implies
\bigmath{x\tightin\Vcons.}
Therefore 
\bigmath{x\mu_{0,q}\tightin\tcc.}
Together with
\\ 
\bigmath{
  \forall p'\tightin\DOM{\Gamma(x)}\stopq
  x\mu_{0,q}
  \refltransindex{\omega+n+1}
  \replpar
    {x\mu_{0,q}}{p''}{r_{1,q p' p''}\mu_{1,q p' p''}}{(p',p'')\tightin\Gamma(x)}
}
this implies
\\
\bigmath{
  \forall p'\tightin\DOM{\Gamma(x)}\stopq
  x\mu_{0,q}
  \refltransindex{\omega}
  \replpar
    {x\mu_{0,q}}{p''}{r_{1,q p' p''}\mu_{1,q p' p''}}{(p',p'')\tightin\Gamma(x)}
  \in\tcc
}
\\
by \lemmaconskeeping.
By confluence of \redindex{\omega} 
and \lemmaconskeeping\ again, there is some \math{t\in\tcc} with
\\
\bigmath{
  \forall p'\tightin\DOM{\Gamma(x)}\stopq
  \replpar
    {x\mu_{0,q}}{p''}{r_{1,q p' p''}\mu_{1,q p' p''}}{(p',p'')\tightin\Gamma(x)}
  \refltransindex{\omega}
  t
.}
Therefore we can define \bigmath{x\nu:=t} in this case.
This is appropriate since by 
\bigmath{
  \exists p'\tightin\DOM{\Gamma(x)}\stopq
  x\mu_{0,q}
  \refltransindex{\omega}
  \penalty-1
  \replpar
    {x\mu_{0,q}}{p''}{r_{1,q p' p''}\mu_{1,q p' p''}}{(p',p'')\tightin\Gamma(x)}
  \refltransindex{\omega}
  x\nu
}
we have
\bigmath{
  x\mu_{0,q}
  \refltransindex{\omega}
  x\nu  
.}\QED{Claim~7}

\noindent
\underline{Claim~8:}
\bigmath{
  l_{0,q}\nu
  \antirefltransindex{\omega}
  w_1/q
.}
\\
\underline{Proof of Claim~8:}
\\
By Claim~7 we get
\bigmath{
  w_1/q
  \tightequal
  \replpar 
    {u/q}
    {p' p''}
    {r_{1,q p' p''}\mu_{1,q p' p''}}
    {\exists x\tightin\V\stopq(p',p'')\tightin\Gamma(x)}
  \tightequal\\
  \replpar
    {\replpar
       {l_{0,q}}
       {p'}
       {x\mu_{0,q}}
       {l_{0,q}/p'\tightequal x\tightin\V}
    }
    {p' p''}
    {r_{1,q p' p''}\mu_{1,q p' p''}}
    {\exists x\tightin\V\stopq(p',p'')\tightin\Gamma(x)}
  \tightequal\\
  \replpar
    {l_{0,q}}
    {p'}
    {\replpar
       {x\mu_{0,q}}
       {p''}
       {r_{1,q p' p''}\mu_{1,q p' p''}}
       {(p',p'')\tightin\Gamma(x)}}
    {l_{0,q}/p'\tightequal x\tightin\V}
  \refltransindex\omega\\
  \replpar
    {l_{0,q}}
    {p'}
    {x\nu}
    {l_{0,q}/p'\tightequal x\tightin\V}
  \tightequal
  l_{0,q}\nu
.}\QED{Claim~8}

\noindent
\underline{Claim~9:}
\bigmath{
  w_0/q
  \redparaindex{\omega+n+1}
  \tight\circ
  \refltransindex\omega
  r_{0,q}\nu
.}
\\
\underline{Proof of Claim~9:} 
Since 
\bigmath{
  w_0/q
  \tightequal 
  r_{0,q}\mu_{0,q}
,} 
this follows from Claim~7.%
\QED{Claim~9}

\noindent
By claims 8 and 9 it now suffices to show
\bigmath{
  r_{0,q}\nu
  \antiredindex{\omega}
  l_{0,q}\nu
,}
which again follows from 
Lemma~\ref{lemma invariance of fulfilledness two}
since 
\bigmath{\forall x\tightin\V\stopq x\mu_{0,q}\refltransindex{\omega+n+1}x\nu}
by Claim~7 and Corollary~\ref{corollary parallel one}.%
\QEDdouble{``The variable overlap (if any) case''}

\pagebreak

\yestop
\noindent
\underline{\underline{``The critical peak case'':
There is some \math{p\in \Pi_1'\tightcap\TPOS{l_{0,q}}}
with \math{l_{0,q}/p\tightnotin\V}:}}
\begin{diagram}
l_{0,q}\mu_{0,q}&\rredindex{\omega+n+1,p}&u'
&&\rredparaindex{\omega+n+1,\,\Pi_1'\setminus\{p\}}&&&&w_1/q
\\&&&&&&&
\\\dredindex{\omega,\emptyset}&&\donlyonceindex{\omega}
&&&&&&\drefltransindex{\omega}
\\&&&&&&&&
\\w_0/q&\rrefltransindex{\omega}
&v&\rrefltransindex{\omega}&\circ&\rredparaindex{\omega+n+1}
&\circ&\rrefltransindex{\omega[+n]}&\circ
\end{diagram}
\underline{Claim~10:}
\bigmath{p\tightnotequal\emptyset.}
\\
\underline{Proof of Claim~10:}
If \bigmath{p\tightequal\emptyset,} then
\bigmath{\emptyset\tightin\Pi_1',} then
\bigmath{q\tightin\Pi_1,} which contradicts our global case assumption.%
\QED{Claim~10}

\noindent
Let \math{\xi\in\SUBST\V\V} be a bijection with 
\bigmath{
  \xi[\VAR{\kurzregelindex{1,q p}}]\cap\VAR{\kurzregelindex{0,q}}
  =
  \emptyset
.}
\\
Define
\bigmath{
  \Y
  :=
  \xi[\VAR{\kurzregelindex{1,q p}}]\cup\VAR{\kurzregelindex{0,q}}
.}
\\
Let \math{\varrho\in\Xsubst} be given by
$\ x\varrho=
\left\{\begin{array}{@{}l@{}l@{}}
  x\mu_{0,q}        &\mbox{ if }x\in\VAR{\kurzregelindex{0,q}}\\
  x\xi^{-1}\mu_{1,q p}&\mbox{ else}\\
\end{array}\right\}
\:(x\tightin\V)$.
\\
By
\math{
  l_{1,q p}\xi\varrho
  \tightequal 
  l_{1,q p}\xi\xi^{-1}\mu_{1,q p}
  \tightequal 
  u/q p
  \tightequal 
  l_{0,q}\mu_{0,q}/p  
  \tightequal 
  l_{0,q}\varrho/p
  \tightequal 
  (l_{0,q}/p)\varrho
}
\\
let
\math{
  \sigma:=\minmgu{\{(l_{1,q p}\xi}{l_{0,q}/p)\},\Y}
}
and
\math{\varphi\in\Xsubst}
with
\math{
  \domres{\inpit{\sigma\varphi}}\Y
  \tightequal
  \domres\varrho\Y
}.
\\
Define 
\math{
  u':=  
  \repl{l_{0,q}\mu_{0,q}}
       {p}
       {r_{1,q p}\mu_{1,q p}}
}.
We get
\\\LINEmath{
  \arr{{l@{}l}
  u'
  \tightequal&  
  \repl
    {\replpar
       {u/q}{p'}{l_{1,q p'}\mu_{1,q p'}}{p'\tightin\Pi_1'\tightsetminus\{p\}}}
    {p}
    {r_{1,q p}\mu_{1,q p}}
  \redparaindex{\omega+n+1,\Pi_1'\setminus\{p\}}
  \\&
  \replpar{u/q}{p'}{r_{1,q p'}\mu_{1,q p'}}{p'\tightin\Pi_1'}
  \tightequal  
  w_1/q
  .
  }
}
\\
If 
\bigmath{
  \repl{l_{0,q}}{p}{r_{1,q p}\xi}\sigma
  \tightequal
  r_{0,q}\sigma
,}
then the proof is finished due to 
\\\LINEmath{
  w_0/q
  \tightequal
  r_{0,q}\mu_{0,q}
  \tightequal
  r_{0,q}\sigma\varphi
  \tightequal
  \repl{l_{0,q}}{p}{r_{1,q p}\xi}\sigma\varphi
  \tightequal
  u'
  \redparaindex{\omega+n+1,\Pi_1'\setminus\{p\}}
  w_1/q
.}
\\
Otherwise 
we have
\bigmath{
  (\,
   (\repl{l_{0,q}}{p}{r_{1,q p}\xi}\sigma,
    C_{1,q p}\xi\sigma,    
    1),\penalty-1\,
   (r_{0,q}\sigma,
    C_{0,q}\sigma,
    0),\penalty-1\,
    l_{0,q}\sigma,\penalty-1\,
    p\,)
  \in{\rm CP}(\R)
}
(due to Claim~5);
\bigmath{p\tightnotequal\emptyset}
(due to Claim~10);
\bigmath{C_{1,q p}\xi\sigma\varphi=C_{1,q p}\mu_{1,q p}}
is fulfilled \wrt\ \redindex{\omega+n};
\bigmath{C_{0,q}\sigma\varphi=C_{0,q}\mu_{0,q}}
is fulfilled \wrt\ \redindex{\omega}.
Since \RX\ is
\math\omega-shallow confluent up to \math{n}
(by our induction hypothesis),
due to our assumed \math\omega-shallow closedness up to \math\omega\ 
(matching the definition's \math{n_0} to our \math{n\tight+1} 
                  and its \math{n_1} to \math{0})
we have
\bigmath{
  u'
  \tightequal
  \repl{l_{0,q}}{p}{r_{1,q p}\xi}\sigma\varphi
  \onlyonceindex\omega
  v
  \antirefltransindex{\omega}
  r_{0,q}\sigma\varphi
  \tightequal
  r_{0,q}\mu_{0,q}
  \tightequal
  w_0/q
}
for some \math{v}.
We then have
\bigmath{
  v
  \antionlyonceindex\omega
  u'
  \redparaindex{\omega+n+1,\Pi_1'\setminus\{p\}}
  w_1/q
.}
We can finish the proof in this case due to our second induction level since
\bigmath{
  \CARD{\Pi_1'\tightsetminus\{p\}}
  \prec
  \CARD{\Pi_1'}
  \preceq
  \CARD{\Pi_1}
.}
\\
\Qeddouble{``The critical peak case''}\QEDtriple{``\math{q\tightnotin\Pi_1}''}

\pagebreak

\noindent
\underline{\underline{\underline{\math{q\tightin\Pi_1}:}}}
If there is no \math{\bar p_0'} with 
\bigmath{q \bar p_0'\tightequal \bar p_0,}
then the proof is finished due to
\bigmath{
  w_0/q
  \tightequal
  u/q
  \redindex{\omega+n+1}
  w_1/q
.}
Otherwise, we can
define \math{\bar p_0'} by 
\bigmath{q \bar p_0'\tightequal \bar p_0.}
We have two cases:

\yestop
\noindent
\underline{\underline{``The second variable overlap case'':}}
\\\LINEnomath{\underline{\underline{There are 
\math{x\in\V} and \math{p'}, \math{p''} such that 
\bigmath{
  l_{1,q}/p'\tightequal x
} 
and
\bigmath{
  p' p''\tightequal \bar p_0'
}:}}}
\begin{diagram}
l_{1,q}\mu_{1,q}&&&\rredindex{\omega+n+1,\,\emptyset}&&&w_1/q
\\&&&&&&\dequal
\\\dredindex{\omega,\,\bar p_0'}&&&&&&r_{1,q}\mu_{1,q}
\\&&&&&&\dredparaindex{\omega}
\\w_0/q&\rredparaindex{\omega}&l_{1,q}\nu&&\rredindex{\omega+n+1}&&r_{1,q}\nu
\end{diagram}
\noindent
\underline{Claim~11:}
For \math{\nu\in\Xsubst} defined by
\bigmath{
         x\nu
         \tightequal
         \repl
           {x\mu_{1,q}}
           {p''}
           {r_{0,\bar p_0}\mu_{0,\bar p_0}}
}
and
\bigmath{
  \forall y\tightin\V\tightsetminus\{x\}\stopq
         y\nu
         \tightequal
         y\mu_{1,q}
}
we get
\bigmath{
  \forall y\tightin\V\stopq
       y\mu_{1,q}
       \redparaindex\omega
       y\nu
.} 
\\
\underline{Proof of Claim~11:}
\\
Due to 
\bigmath{
    x\mu_{1,q}/p''
    \tightequal 
    l_{1,q}\mu_{1,q}/p' p''
    \tightequal 
    u/q p' p''
    \tightequal 
    u/q \bar p_0'
    \tightequal 
    u/\bar p_0
    \tightequal 
    l_{0,\bar p_0}\mu_{0,\bar p_0}
}
we have
\bigmath{
  x\mu_{1,q}
  \tightequal
  \repl{x\mu_{1,q}}{p''}{l_{0,\bar p_0}\mu_{0,\bar p_0}}
  \redindex{\omega}
  \repl{x\mu_{1,q}}{p''}{r_{0,\bar p_0}\mu_{0,\bar p_0}}
  \tightequal
  x\nu  
.}\QED{Claim~11}

\noindent
\underline{Claim~12:}
\bigmath{
  w_0/q
  \redparaindex{\omega} 
  l_{1,q}\nu
.}
\\
\underline{Proof of Claim~12:}
\\
By Claim~11 we get
\bigmath{
  w_0/q
  \tightequal
  \repl
    {u/q}
    {p' p''}
    {r_{0,\bar p_0}\mu_{0,\bar p_0}}
  \tightequal\\
  \repl
    {\replpar
       {l_{1,q}}
       {p'''}
       {y\mu_{1,q}}
       {l_{1,q}/p'''\tightequal y\tightin\V}
    }
    {p' p''}
    {r_{0,\bar p_0}\mu_{0,\bar p_0}}
  \tightequal\\
    {\replpar
       {\replpar
          {l_{1,q}}
          {p'''}
          {y\mu_{1,q}}
          {l_{1,q}/p'''\tightequal y\tightin\V\und x\tightnotequal y}
       }
       {p'''}
       {x\mu_{1,q}}
       {l_{1,q}/p'''\tightequal x\und p'''\tightnotequal p'}
     }
     \\\replsuffix
     {p'}
     {\repl
        {x\mu_{1,q}}
        {p''}
        {r_{0,\bar p_0}\mu_{0,\bar p_0}}}
  \tightequal\\
    {\replpar
       {\replpar
          {l_{1,q}}
          {p'''}
          {y\nu}
          {l_{1,q}/p'''\tightequal y\tightin\V\und x\tightnotequal y}
       }
       {p'''}
       {x\mu_{1,q}}
       {l_{1,q}/p'''\tightequal x\und p'''\tightnotequal p'}
     }
     \replsuffix
     {p'}
     {x\nu}
  \redparaindex{\omega}\\
    {\replpar
       {\replpar
          {l_{1,q}}
          {p'''}
          {y\nu}
          {l_{1,q}/p'''\tightequal y\tightin\V\und x\tightnotequal y}
       }
       {p'''}
       {x\nu}
       {l_{1,q}/p'''\tightequal x\und p'''\tightnotequal p'}
     }
     \replsuffix
     {p'}
     {x\nu}
  \tightequal\\
  \replpar
    {l_{1,q}}
    {p'''}
    {y\nu}
    {l_{1,q}/p'''\tightequal y\tightin\V}
  \tightequal
  l_{1,q}\nu
.}\QED{Claim~12}

\noindent
\underline{Claim~13:}
\bigmath{
  r_{1,q}\nu
  \antiredparaindex\omega
  w_1/q
.}
\\
\underline{Proof of Claim~13:} 
Since \bigmath{r_{1,q}\mu_{1,q}\tightequal w_1/q,} 
this follows directly from Claim~11.%
\QED{Claim~13}

\noindent
By claims 12 and 13 
using Corollary~\ref{corollary parallel one} 
it now suffices to show
\bigmath{
  l_{1,q}\nu
  \redindex{\omega+n+1}
  r_{1,q}\nu
,}
which again follows from 
Claim~11,
Lemma~\ref{lemma invariance of fulfilledness}
(matching its \math{n_0} to \math{0} and its \math{n_1} to our \math{n})
and our induction hypothesis that \RX\ is \math\omega-shallow confluent up
to \math n.%
\QEDdouble{``The second variable overlap case''}

\pagebreak

\yestop
\noindent
\underline{\underline{``The second critical peak case'':
\math{\bar p_0'\in\TPOS{l_{1,q}}}
with \math{l_{1,q}/\bar p_0'\tightnotin\V}:}}
\begin{diagram}
l_{1,q}\mu_{1,q}&&&\rredindex{\omega+n+1,\emptyset}&&&w_1/q
\\
\\\dredindex{\omega,\,\bar p_0'}&&&&&&\drefltransindex{\omega}
\\
\\w_0/q
&\rrefltransindex{\omega}
&\circ
&\rredparaindex{\omega+n+1}
&\circ
&\rrefltransindex{\omega[+n]}&\circ
\end{diagram}
Let \math{\xi\in\SUBST\V\V} be a bijection with 
\bigmath{
  \xi[\VAR{\kurzregelindex{0,\bar p_0}}]\cap\VAR{\kurzregelindex{1,q}}
  =
  \emptyset
.}
\\
Define
\bigmath{
  \Y
  :=
  \xi[\VAR{\kurzregelindex{0,\bar p_0}}]\cup\VAR{\kurzregelindex{1,q}}
.}
\\
Let \math{\varrho\in\Xsubst} be given by
$\ x\varrho=
\left\{\begin{array}{@{}l@{}l@{}}
  x\mu_{1,q}        &\mbox{ if }x\in\VAR{\kurzregelindex{1,q}}\\
  x\xi^{-1}\mu_{0,\bar p_0}&\mbox{ else}\\
\end{array}\right\}
\:(x\tightin\V)$.
\\
By
\math{
  l_{0,\bar p_0}\xi\varrho
  \tightequal 
  l_{0,\bar p_0}\xi\xi^{-1}\mu_{0,\bar p_0}
  \tightequal 
  u/\bar p_0
  \tightequal 
  l_{1,q}\mu_{1,q}/\bar p_0'  
  \tightequal 
  l_{1,q}\varrho/\bar p_0'
  \tightequal 
  (l_{1,q}/\bar p_0')\varrho
}
\\
let
\math{
  \sigma:=\minmgu{\{(l_{0,\bar p_0}\xi}{l_{1,q}/\bar p_0')\},\Y}
}
and
\math{\varphi\in\Xsubst}
with
\math{
  \domres{\inpit{\sigma\varphi}}\Y
  \tightequal
  \domres\varrho\Y
}.
\\
If 
\bigmath{
  \repl{l_{1,q}}{\bar p_0'}{r_{0,\bar p_0}\xi}\sigma
  \tightequal
  r_{1,q}\sigma
,}
then the proof is finished due to 
\\\LINEmath{
  w_0/q  
  \tightequal
  \repl{l_{1,q}\mu_{1,q}}{\bar p_0'}{r_{0,\bar p_0}\mu_{0,\bar p_0}}
  \tightequal
  \repl{l_{1,q}}{\bar p_0'}{r_{0,\bar p_0}\xi}\sigma\varphi
  \tightequal
  r_{1,q}\sigma\varphi
  \tightequal
  r_{1,q}\mu_{1,q}
  \tightequal
  w_1/q.
}
\\
Otherwise 
we have
\bigmath{
  (\,
   (\repl{l_{1,q}}{\bar p_0'}{r_{0,\bar p_0}\xi}\sigma,
    C_{0,\bar p_0}\xi\sigma,    
    0),\penalty-1\,
   (r_{1,q}\sigma,
    C_{1,q}\sigma,
    1),\penalty-1\,
    l_{1,q}\sigma,\penalty-1\,
    \bar p_0'\,)
  \in{\rm CP}(\R)
}
(due to Claim~5);
\bigmath{C_{0,\bar p_0}\xi\sigma\varphi=C_{0,\bar p_0}\mu_{0,\bar p_0}}
is fulfilled \wrt\ \redindex{\omega};
\bigmath{C_{1,q}\sigma\varphi=C_{1,q}\mu_{1,q}}
is fulfilled \wrt\ \redindex{\omega+n}.
Since \RX\  
\math\omega-shallow confluent up to \math{n}
(by our induction hypothesis),
due to our assumed \math\omega-shallow 
{[noisy]}
weak parallel joinability up to \math\omega\ 
(matching the definition's \math{n_0} to \math 0
                   and its \math{n_1} to our \math{n\tight+1})
we have
\bigmath{
  w_0/q
  \tightequal
  \repl{l_{1,q}}{\bar p_0'}{r_{0,\bar p_0}\xi}\sigma\varphi
  \refltransindex{\omega}
  \circ
  \redparaindex{\omega+n+1}
  \circ
  \refltransindex{\omega[+n]}
  \circ
  \antirefltransindex\omega
  r_{1,q}\sigma\varphi
  \tightequal
  r_{1,q}\mu_{1,q}
  \tightequal
  w_1/q
.}
\\\Qeddouble{``The second critical peak case''}
\end{proofparsepqed}

\pagebreak

\begin{proofparsepqed}{Lemma~\ref{lemma closed first level two}}
For \math{n\prec\omega} we are going to show by induction on \math n
the following property\footroom:
\\\LINEmath{
    w_0
    \antiredindex\omega
    u
    \redindex{\omega+n}
    w_1
  \quad\implies\quad
    w_0
    \onlyonceindex{\omega+n}
    \tight\circ
    \refltransindex{\omega[+(n\monus1)]}
    \circ
    \antirefltransindex\omega
    w_1
.}
\begin{diagram}
u&&\rredindex{\omega+n}&&w_1\\
\dredindex{\omega}&&&&\drefltransindex{\omega}\\
w_0&\ronlyonceindex{\omega+n}&\circ&
\rrefltransindex{\omega[+(n\monus1)]}&\circ\\
\end{diagram}

\noindent
\underline{Claim~1:}
If the above property holds for a fixed \math{n\prec\omega}, 
and 
\\\bigmath{
  \forall k\tightprec n\stopq
     (\RX\mbox{ is \math\omega-shallow confluent up to }k)
,} then
\bigmath{
    \redindex{\omega+n}
    \tight\circ
    \refltransindex{\omega[+(n\monus1)]}
}
strongly commutes over
\refltransindex\omega.
\\
\underline{Proof of Claim~1:}
By Lemma~\ref{lemma strong commutation one copy} it suffices to show that
\bigmath{
    \redindex{\omega+n}
    \tight\circ
    \refltransindex{\omega[+(n\monus1)]}
}
strongly commutes over
\redindex{\omega}.
Assume
\bigmath{
    w_0
    \antiredindex\omega
    u
    \redindex{\omega+n}
    w_1
    \refltransindex{\omega[+(n\monus1)]}
    w'
}
(\cf\ diagram below).
By the above property there is some \math{v'}
with
\bigmath{
    w_0
    \onlyonceindex{\omega+n}
    \tight\circ
    \refltransindex{\omega[+(n\monus1)]}
    v'
    \antirefltransindex\omega
    w_1
.}
We only have to show that we can close the peak
\bigmath{ 
    v'
    \antirefltransindex\omega
    w_1
    \refltransindex{\omega[+(n\monus1)]}
    w'
}
according to 
\bigmath{
    v'
    \refltransindex{\omega[+(n\monus1)]}
    \circ
    \antirefltransindex\omega
    w'
.}
{[In case of \bigmath{n\tightequal0:}]}
This is possible due to confluence of \redindex\omega.
{[Otherwise we have \bigmath{n\monus1\tightprec n} and due to the 
assumed \math\omega-shallow confluence up to \math{n\monus1}
this is possible again.]}
\begin{diagram}
u&&\rredindex{\omega+n}&&w_1&\rrefltransindex{\omega[+(n\monus1)]}&w'\\
\dredindex{\omega}&&&&\drefltransindex{\omega}
&&\drefltransindex{\omega}\\
w_0&\ronlyonceindex{\omega+n}&\circ&
\rrefltransindex{\omega[+(n\monus1)]}&v'
&\rrefltransindex{\omega[+(n\monus1)]}&\circ\\
\end{diagram}
\Qed{Claim~1}

\yestop
\noindent
\underline{Claim~2:}
If the above property holds for a fixed \math{n\prec\omega}, and 
\\\bigmath{
  \forall k\tightprec n\stopq
     (\RX\mbox{ is \math\omega-shallow confluent up to }k)
,} 
then
\redindex{\omega+n} and \redindex\omega\ are commuting.
\\
\underline{Proof of Claim~2:}
\bigmath{
    \redindex{\omega+n}
    \tight\circ
    \refltransindex{\omega[+(n\monus1)]}
}
and
\refltransindex\omega\
are commuting
by Lemma~\ref{lemma strong commutation one copy}
and Claim~1.
Since by \lemmamonotonicinbeta\
we have
\bigmath{
  \redindex{\omega+n}
  \subseteq
  \redindex{\omega+n}\tight\circ\refltransindex{\omega[+(n\monus1)]}
  \subseteq
  \refltransindex{\omega+n}
,}
now \redindex{\omega+n} and \redindex\omega\ are commuting, too.\QED{Claim~2}

\yestop
\noindent
\underline{Claim~3:}
If the above property holds for all 
\math{n\preceq m} for some \math{m\prec\omega}, then
\RX\ is \math\omega-shallow confluent up to \math{m}.
\\
\underline{Proof of Claim~3:}
By induction on \math{m} in \tightprec.
Assume
\bigmath{i\plusomega n\tightpreceq m}
and
\bigmath{
    w_0
    \antirefltransindex{\omega+i}
    u
    \refltransindex    {\omega+n}
    w_1
.}
By definition of `\plusomega' and 
\bigmath{i\plusomega n\tightprec\omega}
\wrog\ we have 
\bigmath{i\tightequal0}
and
\bigmath{n\tightpreceq m.}
By Claim~2 and our induction hypothesis we finally get
\bigmath{
  w_0\refltransindex{\omega+n}\circ\antirefltransindex{\omega} w_1
}
as desired.\QED{Claim~3}

\pagebreak

\yestop
\yestop
\noindent
Note that our property for is trivial for \bigmath{n\tightequal0} since
\redindex\omega\ is confluent.

The benefit of 
claims 1 and 3 is twofold: 
First, they say that our lemma is valid if the above property
holds for all \math{n\prec\omega}.
Second, they strengthen the property when used as induction hypothesis. Thus 
(writing \math{n\tight+1} instead of \math{n} since we may assume
 \math{0\tightprec n})
it
now suffices to show
for
\math{
  n\prec\omega
}
that 
\\\linemath{
    w_0
    \antiredindex{\omega,\bar p_0}
    u
    \redindex{\omega+n+1,\bar p_1}
    w_1
}
together with our induction hypothesis 
that\headroom
\\\linenomath{\headroom
  \RX\ is \math\omega-shallow confluent up to \math n
}
\headroom
implies
\\\LINEmath{
    w_0
    \onlyonceindex{\omega+n+1}
    \tight\circ
    \refltransindex{\omega[+n]}
    \circ
    \antirefltransindex\omega
    w_1
.}
\begin{diagram}
u&&\rredindex{\omega+n+1,\,\bar p_1}&&w_1
\\\dredindex{\omega,\,\bar p_0}&&&&\drefltransindex{\omega}
\\w_0&\ronlyonceindex{\omega+n+1}&\circ&\rrefltransindex{\omega[+n]}&\circ
\end{diagram}

\yestop
\noindent
Now for each \math{i\prec2} there are
\bigmath{\kurzregelindex{i}\in\R}
and
\bigmath{\mu_{i}\in\Xsubst}
with
\bigmath{u/\bar p_i\tightequal l_{i}\mu_{i},}
\bigmath{
  w_i\tightequal\repl{u}{\bar p_i}{r_{i}\mu_{i}}
,}
\bigmath{l_{0}\tightin\tcs,} 
\math{\ C_{0}\mu_{0}} fulfilled \wrt\ \redindex{\omega},
\math{C_{1}\mu_{1}} fulfilled \wrt\ \redindex{\omega+n}.

\yestop
\noindent
\underline{Claim~5:}
We may assume 
\bigmath{
    l_{1}\tightnotin\tcs
.}
\\
\underline{Proof of Claim~5:}
In case of \bigmath{l_{1}\tightin\tcs}
by Lemma~\ref{lemma invariance of fulfilledness two} 
(matching both its \math\mu\ and \math\nu\ to our \math{\mu_{1}})
we get
\bigmath{
   l_1\mu_1\redindex\omega r_1\mu_1
.}
Then the proof is finished
by confluence of \redindex\omega.\QED{Claim~5}

\yestop
\yestop
\noindent
In case of \bigmath{\neitherprefix{\bar p_0}{\bar p_1}} 
we have 
\bigmath{
  w_{i}/\bar p_{1-i}
  \tightequal
  \repl u{\bar p_{i}}{r_{i}\mu_{i}}/\bar p_{1-i}
  \tightequal
  u/\bar p_{1-i}
  \tightequal
  l_{1-i}\mu_{1-i}
} 
and
therefore
\bigmath{
  w_0
  \redindex{\omega+n+1}
  \replpar u{\bar p_{k}}{r_{k}\mu_{k}}{k\tightprec2}
  \antiredindex{\omega}
  w_1
,}
\ie\ our proof is finished.
Thus, according to whether \math{\bar p_0} is a prefix of \math{\bar p_1}
or vice versa, we have the following two cases left:

\pagebreak

\yestop
\yestop
\noindent
\underline{\underline{\underline{%
There is some \math{\bar p_1'} with
\bigmath{\bar p_0\bar p_1'\tightequal\bar p_1} and
\bigmath{\bar p_1'\tightnotequal\emptyset}:%
}}}

\noindent
We have two cases:

\noindent
\underline{\underline{``The variable overlap case'':}}
\\\LINEnomath{\underline{\underline{%
There are \math{x\in\V} and \math{p'}, \math{p''} such that
\math{
  l_{0}/p'\tightequal x
  \und
  p' p''\tightequal\bar p_1'
}:}}}
\begin{diagram}
l_{0}\mu_{0}&&&\rredindex{\omega+n+1,\,\bar p_1'}&&&w_1/\bar p_0
\\
&&&&&&\dequal
\\
\dredindex{\omega,\,\emptyset}&&&&&&l_{0}\nu
\\
&&&&&&\dredindex{\omega}
\\
w_0/\bar p_0&\requal&r_{0}\mu_{0}&&\ronlyonceindex{\omega+n+1}
&&r_{0}\nu
\end{diagram}
\noindent
\underline{Claim~6:}
We have 
\bigmath{x\mu_{0}/p''\tightequal l_{1}\mu_{1}}
and may assume
\bigmath{x\tightin\Vsig.}
\\\underline{Proof of Claim~6:}
We have 
\bigmath{
  x\mu_{0}/p''
  \tightequal
  l_{0}\mu_{0}/p' p''
  \tightequal
  u/\bar p_0 p' p''
  \tightequal
  u/\bar p_0\bar p_1'
  \tightequal
  u/\bar p_1
  \tightequal
  l_{1}\mu_{1}
.}
\\
If 
\bigmath{x\tightin\Vcons,}
then \bigmath{x\mu_{0}\tightin\tcc,}
then \bigmath{x\mu_{0}/p''\tightin\tcc,}
then 
\\
\bigmath{l_{1}\mu_{1}\tightin\tcc,}
and then \bigmath{l_{1}\tightin\tcs}
which we may assume not to be the case by Claim~5.%
\QED{Claim~6}

\noindent
\underline{Claim~7:}
We can define \math{\nu\in\Xsubst} by
\bigmath{
  x\nu
  \tightequal
  \repl
    {x\mu_{0}}
    {p''}
    {r_{1}\mu_{1}}
}
and
\bigmath{
  \forall y\tightin\V\tightsetminus\{x\}\stopq 
    y\nu\tightequal y\mu_{0}
.}
Then we have 
\bigmath{
  x\mu_{0}
  \redindex{\omega+n+1}
  x\nu
.}
\\
\underline{Proof of Claim~7:}
This follows directly from Claim~6.%
\QED{Claim~7}

\noindent
\underline{Claim~8:}
\bigmath{
  l_{0}\nu
  \tightequal
  w_1/\bar p_0
.}
\\
\underline{Proof of Claim~8:}
By the left-linearity assumption of our lemma and Claim~6 we may assume
\bigmath{
  \setwith
    {p'''}
    {l_{0}/p'''\tightequal x}
  =
  \{p'\}
.}
Thus, by Claim~7 we get
\bigmath{
  w_1/\bar p_0
  \tightequal
  \repl 
    {u/\bar p_0}
    {\bar p_1'}
    {r_{1}\mu_{1}}
  \tightequal\\
  \repl
    {\replpar
       {l_{0}}
       {p'''}
       {y\mu_{0}}
       {l_{0}/p'''\tightequal y\tightin\V}
    }
    {\bar p_1'}
    {r_{1}\mu_{1}}
  \tightequal\\
  \repl
    {\repl
       {\replpar
          {l_{0}}
          {p'''}
          {y\mu_{0}}
          {l_{0}/p'''\tightequal y\tightin\V\und y\tightnotequal x}
       }
       {p'}
       {x\mu_{0}}
    }
    {p' p''}
    {r_{1}\mu_{1}}
  \tightequal\\
  \repl
    {\replpar
       {l_{0}}
       {p'''}
       {y\nu}
       {l_{0}/p'''\tightequal y\tightin\V\und y\tightnotequal x}
    }
    {p'}
    {\repl
       {x\mu_{0}}
       {p''}
       {r_{1}\mu_{1}}
    }
  \tightequal\\
  \replpar
    {l_{0}}
    {p'''}
    {y\nu}
    {l_{0}/p'''\tightequal y\tightin\V}
  \tightequal
  l_{0}\nu
.}\QED{Claim~8}

\noindent
\underline{Claim~9:}
\bigmath{
  w_0/\bar p_0
  \onlyonceindex{\omega+n+1}
  r_{0}\nu
.}
\\
\underline{Proof of Claim~9:} 
By the right-linearity assumption of our lemma and Claim~6
we may assume 
\bigmath{\CARD{\setwith{p'''}{r_{0}/p'''\tightequal x}}\tightpreceq1.}
Thus by Claim~7 we get:
\bigmath{
  w_0/\bar p_0
  \tightequal 
  r_{0}\mu_{0}
  \tightequal
%  \replpar
%    {r_{0}}
%    {p'''}
%    {y\mu_{0}}
%    {r_{0}/p'''\tightequal y\tightin\V}
%  \tightequal
  \\
  \replpar
    {\replpar
       {r_{0}}
       {p'''}
       {y\mu_{0}}
       {r_{0}/p'''\tightequal y\tightin\V\tightsetminus\{x\}}
    }
    {p'''}
    {x\mu_{0}}
    {r_{0}/p'''\tightequal x}
  \onlyonceindex{\omega+n+1}
  \\
  \replpar
    {\replpar
       {r_{0}}
       {p'''}
       {y\mu_{0}}
       {r_{0}/p'''\tightequal y\tightin\V\tightsetminus\{x\}}
    }
    {p'''}
    {x\nu}
    {r_{0}/p'''\tightequal x}
  \tightequal
  \\
  \replpar
    {\replpar
       {r_{0}}
       {p'''}
       {y\nu}
       {r_{0}/p'''\tightequal y\tightin\V\tightsetminus\{x\}}
    }
    {p'''}
    {x\nu}
    {r_{0}/p'''\tightequal x}
  \tightequal
  r_{0}\nu
.}%
\QED{Claim~9}

\noindent
By claims 8 and 9 it now suffices to show
\bigmath{
  r_{0,q}\nu
  \antiredindex{\omega}
  l_{0,q}\nu
,}
which again follows from 
Lemma~\ref{lemma invariance of fulfilledness two}
since 
\bigmath{\forall y\tightin\V\stopq y\mu_{0,q}\refltransindex{\omega+n+1}y\nu}
by Claim~7.%
\QEDdouble{``The variable overlap case''}

\pagebreak

\yestop
\noindent
\underline{\underline{``The critical peak case'':
\math{
  \bar p_1'\tightin\TPOS{l_{0}}
  \und
  l_{0}/\bar p_1'\tightnotin\V
}:}}
\begin{diagram}
l_{0}\mu_{0}&&
&\rredindex{\omega+n+1,\,\bar p_1'}&&&w_1/\bar p_0
\\
\dredindex{\omega,\,\emptyset}&&
&&&&\drefltransindex{\omega}
\\
w_0/\bar p_0&&\ronlyonceindex{\omega+n+1}&&\circ
&\rrefltransindex{\omega{[+n]}}&\circ
\end{diagram}
Let \math{\xi\in\SUBST\V\V} be a bijection with 
\bigmath{
  \xi[\VAR{\kurzregelindex{1}}]\cap\VAR{\kurzregelindex{0}}
  =
  \emptyset
.}
\\
Define
\bigmath{
  \Y
  :=
  \xi[\VAR{\kurzregelindex{1}}]\cup\VAR{\kurzregelindex{0}}
.}
\\
Let \math{\varrho\in\Xsubst} be given by
$\ x\varrho=
\left\{\begin{array}{@{}l@{}l@{}}
  x\mu_{0}        &\mbox{ if }x\in\VAR{\kurzregelindex{0}}\\
  x\xi^{-1}\mu_{1}&\mbox{ else}\\
\end{array}\right\}
\:(x\tightin\V)$.
\\
By
\math{
  l_{1}\xi\varrho
  \tightequal 
  l_{1}\xi\xi^{-1}\mu_{1}
  \tightequal 
  u/\bar p_1
  \tightequal 
  u/\bar p_0\bar p_1'
  \tightequal 
  l_{0}\mu_{0}/\bar p_1'  
  \tightequal 
  l_{0}\varrho/\bar p_1'
  \tightequal 
  (l_{0}/\bar p_1')\varrho
}
\\
let
\math{
  \sigma:=\minmgu{\{(l_{1}\xi}{l_{0}/\bar p_1')\},\Y}
}
and
\math{\varphi\in\Xsubst}
with
\math{
  \domres{\inpit{\sigma\varphi}}\Y
  \tightequal
  \domres\varrho\Y
}.
\\
If 
\bigmath{
  \repl{l_{0}}{\bar p_1'}{r_{1}\xi}\sigma
  \tightequal
  r_{0}\sigma
,}
then the proof is finished due to 
\\\LINEmath{
  w_0/\bar p_0
  \tightequal
  r_{0}\mu_{0}
  \tightequal
  r_{0}\sigma\varphi
  \tightequal
  \repl{l_{0}}{\bar p_1'}{r_{1}\xi}\sigma\varphi
  \tightequal
  \repl{l_{0}\mu_{0}}{\bar p_1'}{r_{1}\mu_{1}}
  \tightequal
  w_1/\bar p_0
.}
\\
Otherwise 
we have
\bigmath{
  (\,
   (\repl{l_{0}}{\bar p_1'}{r_{1}\xi},
    C_{1}\xi,    
    1),\penalty-1\,
   (r_{0},
    C_{0},
    0),\penalty-1\,
    l_{0},\penalty-1\,
    \sigma,\penalty-1\,
    \bar p_1'\,)
  \in{\rm CP}(\R)
}
(due to Claim~5);
\bigmath{\bar p_1'\tightnotequal\emptyset}
(due the global case assumption);
\bigmath{C_{1}\xi\sigma\varphi=C_{1}\mu_{1}}
is fulfilled \wrt\ \redindex{\omega+n};
\bigmath{C_{0}\sigma\varphi=C_{0}\mu_{0}}
is fulfilled \wrt\ \redindex{\omega}.
Since \RX\ is
\math\omega-shallow confluent up to \math{n}
(by our induction hypothesis),
due to our assumed \math\omega-shallow {[noisy]}
anti-closedness up to \math\omega\ 
(matching the definition's \math{n_0} to our \math{n\tight+1} 
                  and its \math{n_1} to \math{0})
we have
\bigmath{
  w_1/q
  \tightequal
  \repl{l_0\mu_0}{\bar p_1'}{r_1\mu_1}
  \tightequal
  \repl{l_0}{\bar p_1'}{r_1\xi}\sigma\varphi
  \refltransindex\omega
  \circ
  \antirefltransindex{\omega[+n]}
  \tight\circ
  \antionlyonceindex{\omega+n+1}
  r_0\sigma\varphi
  \tightequal
  r_0\mu_0
  \tightequal
  w_0/q.
}
\\
\Qeddouble{``The critical peak case''}%
\QEDtriple{``There is some \math{\bar p_1'} with
\bigmath{\bar p_0\bar p_1'\tightequal\bar p_1} and
\bigmath{\bar p_1'\tightnotequal\emptyset}''}

\pagebreak

\yestop
\yestop
\noindent
\underline{\underline{\underline{%
There is some \math{\bar p_0'} with
\bigmath{\bar p_1\bar p_0'\tightequal\bar p_0}:%
}}}

\noindent
We have two cases:

\noindent
\underline{\underline{%
``The second variable overlap case'':%
}}
\\\LINEnomath{\underline{\underline{%
There are \math{x\tightin\V} and \math{p'}, \math{p''} such that
\math{
    l_{1}/p'
    \tightequal
    x
  \und
    p'p''
    \tightequal
    \bar p_0'
}:%
}}}
\begin{diagram}
l_{1}\mu_{1}&&&\rredindex{\omega+n+1,\,\emptyset}&&&w_1/\bar p_1
\\
&&&&&&\dequal
\\
\dredindex{\omega,\,\bar p_0'}&&&&&&r_{1}\mu_{1}
\\
&&&&&&\dredparaindex{\omega}
\\
w_0/\bar p_1&\requal&l_{1}\nu&&\rredindex{\omega+n+1}&&r_{1}\nu
\end{diagram}
We have 
\bigmath{
  x\mu_{1}/p''
  \tightequal
  l_{1}\mu_{1}/p' p''
  \tightequal
  u/\bar p_1 p' p''
  \tightequal
  u/\bar p_1\bar p_0'
  \tightequal
  u/\bar p_0
  \tightequal
  l_{0}\mu_{0}
.}

\noindent
\underline{Claim~11:}
We can define \math{\nu\in\Xsubst} by
\bigmath{
  x\nu
  \tightequal
  \repl
    {x\mu_{1}}
    {p''}
    {r_{0}\mu_{0}}
}
and
\bigmath{
  \forall y\tightin\V\tightsetminus\{x\}\stopq y\nu\tightequal y\mu_{1}
.}
Then we have 
\bigmath{
  x\mu_{1}\redindex{\omega}x\nu
.}
\\
\underline{Proof of Claim~11:}
This follows directly from the above equality and \lemmaconskeeping.%
\QED{Claim~11}

\noindent
\underline{Claim~12:}
\bigmath{
  w_0/\bar p_1
  \tightequal
  l_{1}\nu
.}
\\
\underline{Proof of Claim~12:}
\\
By the left-linearity assumption of our lemma and Claim~5 we may 
assume
\bigmath{
  \setwith
    {p'''}
    {l_{1}/p'''\tightequal x}
  =
  \{p'\}
.}
Thus, by Claim~11 we get
\bigmath{
  w_0/\bar p_1
  \tightequal
  \repl 
    {u/\bar p_1}
    {\bar p_0'}
    {r_{0}\mu_{0}}
  \tightequal\\
  \repl
    {\replpar
       {l_{1}}
       {p'''}
       {y\mu_{1}}
       {l_{1}/p'''\tightequal y\tightin\V}
    }
    {\bar p_0'}
    {r_{0}\mu_{0}}
  \tightequal\\
  \repl
    {\repl
       {\replpar
          {l_{1}}
          {p'''}
          {y\mu_{1}}
          {l_{1}/p'''\tightequal y\tightin\V\und y\tightnotequal x}
       }
       {p'}
       {x\mu_{1}}
    }
    {p' p''}
    {r_{0}\mu_{0}}
  \tightequal\\
  \repl
    {\replpar
       {l_{1}}
       {p'''}
       {y\nu}
       {l_{1}/p'''\tightequal y\tightin\V\und y\tightnotequal x}
    }
    {p'}
    {\repl
       {x\mu_{1}}
       {p''}
       {r_{0}\mu_{0}}
    }
  \tightequal\\
  \replpar
    {l_{1}}
    {p'''}
    {y\nu}
    {l_{1}/p'''\tightequal y\tightin\V}
  \tightequal
  l_{1}\nu
.}\QED{Claim~12}

\noindent
\underline{Claim~13:}
\bigmath{
  r_{1}\nu
  \antiredparaindex{\omega}
  w_1/\bar p_1
.}
\\
\underline{Proof of Claim~13:} 
Since \bigmath{r_{1}\mu_{1}\tightequal w_1/\bar p_1,} 
this follows directly from Claim~11.%
\QED{Claim~13}

\noindent
By claims 12 and 13 
using Corollary~\ref{corollary parallel one} 
it now suffices to show
\bigmath{
  l_{1,q}\nu
  \redindex{\omega+n+1}
  r_{1,q}\nu
,}
which again follows from 
Claim~11,
Lemma~\ref{lemma invariance of fulfilledness}
(matching its \math{n_0} to \math{0} and its \math{n_1} to our \math{n})
and our induction hypothesis that \RX\ is \math\omega-shallow confluent up
to \math n.%
\QEDdouble{``The second variable overlap (if any) case''}

\pagebreak

\yestop
\noindent
\underline{\underline{``The second critical peak case'':
\math{
    \bar p_0'\tightin\TPOS{l_{1}}
  \und
    l_{1}/\bar p_0'\tightnotin\V
}:%
}}
\begin{diagram}
l_{1}\mu_{1}
&&&\rredindex{\omega+n+1,\,\emptyset}&&&w_1/\bar p_1
\\
\dredindex{\omega,\,\bar p_0'}&&&&&&\drefltransindex{\omega}
\\
w_0/\bar p_1
&&\ronlyonceindex{\omega+n+1}&&\circ
&\rrefltransindex{\omega{[+n]}}&\circ
\end{diagram}
Let \math{\xi\in\SUBST\V\V} be a bijection with 
\bigmath{
  \xi[\VAR{\kurzregelindex{0}}]\cap\VAR{\kurzregelindex{1}}
  =
  \emptyset
.}
\\
Define
\bigmath{
  \Y
  :=
  \xi[\VAR{\kurzregelindex{0}}]\cup\VAR{\kurzregelindex{1}}
.}
\\
Let \math{\varrho\in\Xsubst} be given by
$\ x\varrho=
\left\{\begin{array}{@{}l@{}l@{}}
  x\mu_{1}        &\mbox{ if }x\in\VAR{\kurzregelindex{1}}\\
  x\xi^{-1}\mu_{0}&\mbox{ else}\\
\end{array}\right\}
\:(x\tightin\V)$.
\\
By
\math{
  l_{0}\xi\varrho
  \tightequal 
  l_{0}\xi\xi^{-1}\mu_{0}
  \tightequal 
  u/\bar p_0
  \tightequal 
  u/\bar p_1\bar p_0'
  \tightequal 
  l_{1}\mu_{1}/\bar p_0'  
  \tightequal 
  l_{1}\varrho/\bar p_0'
  \tightequal 
  (l_{1}/\bar p_0')\varrho
}
\\
let
\math{
  \sigma:=\minmgu{\{(l_{0}\xi}{l_{1}/\bar p_0')\},\Y}
}
and
\math{\varphi\in\Xsubst}
with
\math{
  \domres{\inpit{\sigma\varphi}}\Y
  \tightequal
  \domres\varrho\Y
}.
\\
If 
\bigmath{
  \repl{l_{1}}{\bar p_0'}{r_{0}\xi}\sigma
  \tightequal
  r_{1}\sigma
,}
then the proof is finished due to 
\\\linemath{
  w_0/\bar p_1
  \tightequal
  \repl{l_{1}\mu_{1}}{\bar p_0'}{r_{0}\mu_{0}}
  \tightequal
  \repl{l_{1}}{\bar p_0'}{r_{0}\xi}\sigma\varphi
  \tightequal
  r_{1}\sigma\varphi
  \tightequal
  r_{1}\mu_{1}
  \tightequal
  w_1/\bar p_1.
}
Otherwise 
we have
\bigmath{
  (\,
   (\repl{l_{1}}{\bar p_0'}{r_{0}\xi},
    C_{0}\xi,    
    0),\penalty-1\,
   (r_{1},
    C_{1},
    1),\penalty-1\,
    l_{1},\penalty-1\,
    \sigma,\penalty-1\,
    \bar p_0'\,)
  \in{\rm CP}(\R)
}
(due to Claim~5);
\bigmath{C_{0}\xi\sigma\varphi=C_{0}\mu_{0}}
is fulfilled \wrt\ \redindex{\omega};
\bigmath{C_{1}\sigma\varphi=C_{1}\mu_{1}}
is fulfilled \wrt\ \redindex{\omega+n}.
Since \RX\  
\math\omega-shallow confluent up to \math{n}
(by our induction hypothesis),
due to our assumed \math\omega-shallow [noisy]
strong joinability up to \math\omega\ 
(matching the definition's \math{n_0} to \math 0
                   and its \math{n_1} to our \math{n\tight+1})
we have
\bigmath{
  w_0/\bar p_1
  \tightequal
  \repl{l_{1}\mu_{1}}{\bar p_0'}{r_{0}\mu_{0}}
  \tightequal
  \repl{l_{1}}{\bar p_0'}{r_{0}\xi}\sigma\varphi
  \onlyonceindex{\omega+n+1}
  \tight\circ
  \refltransindex{\omega{[+n]}}
  \circ
  \antirefltransindex{\omega}
  \penalty-1
  r_{1}\sigma\varphi
  \tightequal
  r_{1}\mu_{1}
  \tightequal
  w_1/\bar p_1
.}
\\\Qeddouble{``The second critical peak case''}
\end{proofparsepqed}

\begin{proofparsepqed}{Lemma~\ref{lemma closed second level two}}
\noindent
\underline{Claim~0:}
\RX\ is \math\omega-shallow confluent up to \math\omega.
\\\underline{Proof of Claim~0:}
Directly by the assumed strong commutation
of   
\bigmath{
    \redindex{\RX,\omega+n}
    \tight\circ
    \refltransindex{\RX,\omega+(n\monus1)}
}
over \refltransindex{\RX,\omega},
\cf\ the proofs of the claims 2 and 3 of the proof of 
Lemma~\ref{lemma closed first level two}.\QED{Claim~0}

\yestop
\noindent
\underline{Claim~1:}
If 
\bigmath{
    \refltransindex{\omega}
    \tight\circ
    \redparaindex{\omega+n_1}
    \tight\circ
    \refltransindex{\omega+(n_1\monus1)}
}
strongly commutes over
\refltransindex{\omega+n_0},
then
\redindex{\omega+n_1} and \redindex{\omega+n_0} are commuting.
\\
\underline{Proof of Claim~1:}
\bigmath{
    \refltransindex{\omega}
    \tight\circ
    \redparaindex{\omega+n_1}
    \tight\circ
    \refltransindex{\omega+(n_1\monus1)}
}
and
\refltransindex{\omega+n_0}
are commuting
by Lemma~\ref{lemma strong commutation one copy}.
Since by Corollary~\ref{corollary parallel one}
and \lemmamonotonicinbeta\
we have
\bigmath{
  \redindex{\omega+n_1}
  \subseteq
  \refltransindex{\omega}
  \tight\circ
  \redparaindex{\omega+n_1}
  \tight\circ
  \refltransindex{\omega+(n_1\monus1)}
  \subseteq
  \refltransindex{\omega+n_1}
,}
now \redindex{\omega+n_1} 
and \redindex{\omega+n_0} are commuting, too.\QED{Claim~1}

\yestop
\yestop
\noindent
For \math{n_0\preceq n_1\prec\omega} 
we are going to show by induction on \math{n_0\plusomega n_1}
the following property\footroom:
\\\LINEmath{
    w_0
    \antiredindex{\omega+n_0}
    u
    \redparaindex{\omega+n_1}
    w_1
  \quad\implies\quad
    w_0
    \refltransindex{\omega}
    \tight\circ
    \redparaindex{\omega+n_1}
    \tight\circ
    \refltransindex{\omega+(n_1\monus1)}
    \circ
    \antirefltransindex{\omega+n_0}
    w_1
}
\begin{diagram}
u&&&\rredparaindex{\omega+n_1}&&&w_1
\\\dredindex{\omega+n_0}&&&&&&\drefltransindex{\omega+n_0}
\\w_0&\rrefltransindex{\omega}&\circ&\rredparaindex{\omega+n_1}
&\circ&\rrefltransindex{\omega+(n_1\monus1)}&\circ
\end{diagram}

\pagebreak

\noindent
\underline{Claim~2:}
Let \math{\delta\prec\omega\tight+\omega}.
If
\\
\linemath{
  \forall n_0,n_1\tightprec\omega\stopq
  \inparenthesesoplist{ 
    \inparenthesesoplist{
        n_0\tightpreceq n_1      
      \oplistund
        n_0\plusomega n_1\tightpreceq\delta
    }
    \oplistimplies
    \forall w_0,w_1,u\stopq
    \inparenthesesoplist{
        w_0
        \antiredindex{\omega+n_0}
        u
        \redparaindex{\omega+n_1}
        w_1
      \oplistimplies 
        w_0
        \refltransindex{\omega}
        \tight\circ
        \redparaindex{\omega+n_1}
        \tight\circ
        \refltransindex{\omega+(n_1\monus1)}
        \circ
        \antirefltransindex{\omega+n_0}
        w_1
    }
  }
,}
then
\\
\linemath{
  \forall n_0,n_1\tightprec\omega\stopq
  \inparenthesesoplist{ 
    \inparenthesesoplist{
        n_0\tightpreceq n_1      
      \oplistund
        n_0\plusomega n_1\tightpreceq\delta
    }
    \oplistimplies 
            \refltransindex{\omega}
            \tight\circ
            \redparaindex{\omega+n_1}
            \tight\circ
            \refltransindex{\omega+(n_1\monus1)}
      \mbox{ strongly commutes over }
            \refltransindex{\omega+n_0}
  }
,}
and 
\RX\ is \math\omega-shallow confluent up to \math{\delta}.
\\
\underline{Proof of Claim~2:}
By induction on \math{\delta} in \tightprec.
First we show the strong commutation.
Assume \math{n_0\preceq n_1\prec\omega} with
\math{n_0\plusomega n_1\tightpreceq\delta}. 
By Lemma~\ref{lemma strong commutation one copy} it suffices to show that
\bigmath{
    \refltransindex{\omega}
    \tight\circ
    \redparaindex{\omega+n_1}
    \tight\circ
    \refltransindex{\omega+(n_1\monus1)}
}
strongly commutes over
\redindex{\omega+n_0}.
Assume
\bigmath{
    u''
    \antiredindex{\omega+n_0}
    u'
    \refltransindex{\omega}
    u
    \redparaindex{\omega+n_1}
    w_1
    \refltransindex{\omega+(n_1\monus1)}
    w_2
}
(\cf\ diagram below).
By the strong commutation assumption of our lemma
there are \math{w_0} and \math{w_0'} with
\bigmath{
    u''
    \refltransindex{\omega}
    w_0'
    \antirefltransindex{\omega+(n_0\monus1)}
    w_0
    \antionlyonceindex{\omega+n_0}
    u
.}
By the above property there are some \math{w_3}, \math{w_1'}
with
\bigmath{
    w_0
    \refltransindex{\omega}
    w_3
    \redparaindex{\omega+n_1}
    \tight\circ
    \refltransindex{\omega+(n_1\monus1)}
    w_1'
    \antirefltransindex{\omega+n_0}
    w_1
.}
Next we show that we can close the peak
\bigmath{ 
    w_1'
    \antirefltransindex{\omega+n_0}
    w_1
    \refltransindex{\omega+(n_1\monus1)}
    w_2
}
according to 
\bigmath{
    w_1'
    \refltransindex{\omega+(n_1\monus1)}
    w_2'
    \antirefltransindex{\omega+n_0}
    w_2
}
for some \math{w_2'}.
In case of \bigmath{n_1\tightequal0}
this is possible due to the
\math\omega-shallow confluence up to \math{\omega}
given by Claim~0.
Otherwise we have 
\bigmath{
  n_0\plusomega(n_1\monus1)
  \tightprec 
  n_0\plusomega n_1
  \tightpreceq
  \delta
} 
and due to our induction hypothesis
(saying that \RX\ is \math\omega-shallow confluent up to all 
\math{\delta'\prec\delta
})
this is possible again.
By Claim~0 again,
we can close the peak
\bigmath{
  w_0'
  \antirefltransindex{\omega+(n_0\monus1)}
  w_0
  \refltransindex{\omega}
  w_3
}
according to 
\bigmath{
  w_0'
  \refltransindex{\omega}
  w_3'
  \antirefltransindex{\omega+(n_0\monus1)}
  w_3
}
for some \math{w_3'}.
To close the whole diagram, we only have to show that we can close the peak
\bigmath{
    w_3'
    \antirefltransindex{\omega+(n_0\monus1)}
    w_3
    \redparaindex{\omega+n_1}
    \tight\circ
    \refltransindex{\omega+(n_1\monus1)}
    w_2'
}
according to 
\bigmath{
    w_3'
    \refltransindex{\omega}
    \tight\circ
    \redparaindex{\omega+n_1}
    \tight\circ
    \refltransindex{\omega+(n_1\monus1)}
    \circ
    \antirefltransindex{\omega+(n_0\monus1)}
    w_2'
.}
In case of \bigmath{n_0\tightequal0}
this is possible since it is assumed for our lemma 
(below the strong commutation assumption).
Otherwise we have 
\bigmath{
  n_0\monus1
  \tightprec 
  n_0
  \tightpreceq
  n_1
} 
and
\bigmath{
  (n_0\monus1)\plusomega n_1
  \tightprec 
  n_0\plusomega n_1
  \tightpreceq
  \delta
,} 
and then due to our
induction hypothesis
this is possible again.
\begin{diagram}
u'&\rrefltransindex{\omega}&u
&&
&\rredparaindex{\omega+n_1}&&
&w_1&\rrefltransindex{\omega+(n_1\monus1)}&w_2
\\&&\donlyonceindex{\omega+n_0}
&&
&&&
&\drefltransindex{\omega+n_0}&&\drefltransindex{\omega+n_0}
\\\dredindex{\omega+n_0}&&w_0
&\rrefltransindex{\omega}&w_3
&\rredparaindex{\omega+n_1}&\circ&\rrefltransindex{\omega+(n_1\monus1)}
&w_1'&\rrefltransindex{\omega+(n_1\monus1)}&w_2'
\\&&\drefltransindex{\omega+(n_0\monus1)}
&&\drefltransindex{\omega+(n_0\monus1)}
&&&
&&&\drefltransindex{\omega+(n_0\monus1)}
\\u''&\rrefltransindex{\omega}&w_0'
&\rrefltransindex{\omega}&w_3'
&\rrefltransindex{\omega}&\circ&\rredparaindex{\omega+n_1}
&\circ&\rrefltransindex{\omega+(n_1\monus1)}&\circ
\end{diagram}
Finally we show \math\omega-shallow confluence up to \math\delta.
Assume
\bigmath{n_0\plusomega n_1\tightpreceq\delta}
and
\bigmath{
    w_0
    \antirefltransindex{\omega+n_0}
    u
    \refltransindex    {\omega+n_1}
    w_1
.}
Due to symmetry in \math{n_0} and \math{n_1} we may assume
\bigmath{n_0\tightpreceq n_1.}
Above we have shown that 
\bigmath{
    \refltransindex{\omega}
    \tight\circ
    \redparaindex{\omega+n_1}
    \tight\circ
    \refltransindex{\omega+(n_1\monus1)}
}
strongly commutes over
\refltransindex{\omega+n_0}. 
By Claim~1 we finally get
\bigmath{
  w_0\refltransindex{\omega+n_1}\circ\antirefltransindex{\omega+n_0} w_1
}
as desired.%
\QED{Claim~2}

\pagebreak

\yestop
\yestop
\noindent
Note that 
for \bigmath{n_0\tightequal0} 
our property follows 
from 
the assumption of our lemma 
(below the strong commutation assumption).

The benefit of 
Claim~2 is twofold: First, it says that our lemma is valid
if the above property
holds for all \math{n_0\preceq n_1\prec\omega}.
Second, it strengthens the property when used as induction hypothesis. Thus 
(writing \math{n_i\tight+1} instead of \math{n_i} since we may assume
 \math{0\tightprec n_0\tightpreceq n_1})
it
now suffices to show
for
\math{n_0\preceq n_1\prec\omega}
that 
\\\linemath{
    w_0
    \antiredindex{\omega+n_0+1,\bar p_0}
    u
    \redparaindex{\omega+n_1+1,\Pi_1}
    w_1
}
together with our induction hypotheses 
that\headroom
\\\linemath{\headroom
  \forall\delta\tightprec(n_0\tight+1)\plusomega(n_1\tight+1)\stopq
  \mbox{\RX\ is \math\omega-shallow confluent up to }\delta
}
\headroom
and (due to \math{n_0\tightpreceq n_1\tight+1}
and 
\math{
  n_0\plusomega(n_1\tight+1)\tightprec(n_0\tight+1)\plusomega(n_1\tight+1)
})
\\\linenomath{\headroom
  \bigmath{ 
    \refltransindex{\omega}
    \tight\circ
    \redparaindex{\omega+n_1+1}
    \tight\circ
    \refltransindex{\omega+n_1}
  }
  strongly commutes over
  \refltransindex{\omega+n_0}
}
\headroom
implies
\\\LINEmath{
    w_0
    \refltransindex{\omega}
    \tight\circ
    \redparaindex{\omega+n_1+1}
    \tight\circ
    \refltransindex{\omega+n_1}
    \circ
    \antirefltransindex{\omega+n_0+1}
    w_1
.}
\begin{diagram}
u&&&\rredparaindex{\omega+n_1+1,\,\Pi_1}
&&&w_1
\\\dredindex{\omega+n_0+1,\,\bar p_0}&&&
&&&\drefltransindex{\omega+n_0+1}
\\w_0&\rrefltransindex{\omega}&\circ&\rredparaindex{\omega+n_1+1}
&\circ&\rrefltransindex{\omega+n_1}&\circ
\end{diagram}

\yestop
\noindent
{Note that for the availability of our second induction hypothesis  
it is important that we have imposed the restriction
``\math{n_0\tightpreceq n_1}'' in opposition to the restriction
``\math{n_0\tightsucceq n_1}''.
In the latter case the availability of our second induction hypothesis would
require 
\bigmath{
  n_0\tight+1\tightsucceq n_1\tight+1
  \implies
  n_0\tightsucceq n_1\tight+1
}
which is not true for \bigmath{n_0\tightequal n_1.}
The additional hypothesis  
\\\linenomath{
  \bigmath{ 
    \refltransindex{\omega}
    \tight\circ
    \redparaindex{\omega+n_1}
    \tight\circ
    \refltransindex{\omega+(n_1\monus1)}
  }
  strongly commutes over
  \refltransindex{\omega+n_0+1}
}
of the latter restriction is useless for our proof.}

There are
\bigmath{\kurzregelindex{0,\bar p_0}\in\R}
and
\bigmath{\mu_{0,\bar p_0}\in\Xsubst}
with
\bigmath{u/p\tightequal l_{0,\bar p_0}\mu_{0,\bar p_0},}
\math{C_{0,\bar p_0}\mu_{0,\bar p_0}} fulfilled \wrt\ \redindex{\omega+n_0},
and
\bigmath{
  w_0
  \tightequal
  \repl{u}{p}{r_{0,\bar p_0}\mu_{0,\bar p_0}}
.}

\Wrog\ let the positions of \math{\Pi_1} be maximal
in the sense that for any \math{p\in\Pi_1} 
and \math{\Xi\subseteq \TPOS u\tightcap(p\N^+)}
we do not have 
\bigmath{
  u
  \redparaindex{\omega+n_1+1,(\Pi_1\setminus\{p\})\cup\Xi}
  w_1
}
anymore.
Then for each \math{p\in\Pi_1} there are
\bigmath{\kurzregelindex{1,p}\in\R}
and
\bigmath{\mu_{1,p}\in\Xsubst}
with
\bigmath{u/p\tightequal l_{1,p}\mu_{1,p},}
\bigmath{r_{1,p}\mu_{1,p}\tightequal w_1/p,}
\math{C_{1,p}\mu_{1,p}} fulfilled \wrt\ \redindex{\omega+n_1}.
Finally, 
\bigmath{
  w_1
  \tightequal
  \replpar{u}{p}{r_{1,p}\mu_{1,p}}{p\tightin\Pi_1}
.}

\pagebreak

\yestop
\noindent
\underline{Claim~5:}
\\
We may assume 
\bigmath{
    l_{0,\bar p_0}\tightnotin\tcs
}
and
\bigmath{
  \forall p\tightin\Pi_1\stopq
    l_{1,p}\tightnotin\tcs
.}
\\
\underline{Proof of Claim~5:}
In case of 
\bigmath{
    l_{0,\bar p_0}\tightin\tcs
}
we get 
\bigmath{w_0\antiredindex{\omega}u}
by Lemma~\ref{lemma invariance of fulfilledness two} 
(matching both its \math\mu\ and \math\nu\ to our \math{\mu_{0,\bar p_0}})
and then
our property follows 
from 
the assumption of our lemma 
(below the strong commutation assumption).
For the second restriction 
define \math{\Xi_1:=\setwith{p\tightin\Pi_1}{l_{1,p}\tightin\tcs}}
and 
\math{
  u_1':=\replpar{u}{p}{r_{1,p}\mu_{1,p}}{p\tightin\Pi_1\tightsetminus\Xi_1}
}.
If we have succeeded with our proof under the assumption of Claim~5,
then we have shown 
\bigmath{
  w_0
  \refltransindex{\omega}
  \tight\circ
  \redparaindex{\omega+n_1+1}
  \tight\circ
  \refltransindex{\omega+n_1}
  v_1
  \antirefltransindex{\omega+n_0+1}
  u_1'
}
for some \math{v_1}
(\cf\ diagram below).
By Lemma~\ref{lemma invariance of fulfilledness two} 
(matching both its \math\mu\ and \math\nu\ to our \math{\mu_{1,p}})
we get
\bigmath{
  \forall p\tightin\Xi_1\stopq
    l_{1,p}\mu_{1,p}\redindex\omega r_{1,p}\mu_{1,p}
}
and therefore
\bigmath{
  u_1'
  \refltransindex\omega
  w_1
.}
Thus from
\bigmath{
  v_1
  \penalty-1
  \antirefltransindex{\omega+n_0+1}
  \penalty-1
  u_1'
  \penalty-1
  \refltransindex\omega
  \penalty-1
  w_1
}
we get 
\bigmath{
  v_1
  \refltransindex\omega
  v_2
  \antirefltransindex{\omega+n_0+1}
  w_1
}
for some \math{v_2}
by \math\omega-shallow confluence up to \math{\omega} (\cf\ Claim~0).
\begin{diagram}
u&&&\rredparaindex{\omega+n_1+1,\,\Pi_1\setminus\Xi_1}
&&&u_1'&\rrefltransindex{\omega}&w_1
\\\dredindex{\omega+n_0+1,\,\bar p_0}&&&
&&&\drefltransindex{\omega+n_0+1}&&\drefltransindex{\omega+n_0+1}
\\w_0&\rrefltransindex{\omega}&\circ&\rredparaindex{\omega+n_1+1}
&\circ&\rrefltransindex{\omega+n_1}&v_1&\rrefltransindex{\omega}&v_2
\end{diagram}
\Qed{Claim~5}

\yestop
\noindent
Now we start a second level of induction on
\bigmath{
  \CARD{\Pi_1}
}
in \tightprec.

\noindent
Define the set of top positions by
\\\linemath{
  \displaystyle
  \Theta
  :=
      \setwith
      {p\tightin\{\bar p_0\}\tightcup\Pi_1}
      {\neg\exists q\tightin\{\bar p_0\}\tightcup\Pi_1\stopq
           \exists q'\tightin\N^+\stopq
             p\tightequal q q'
      }
.}
Since the prefix ordering is wellfounded we have
\bigmath{
  \forall p\tightin\{\bar p_0\}\tightcup\Pi_1\stopq
  \exists q\tightin\Theta\stopq
  \exists q'\tightin\N^\ast\stopq
    p\tightequal q q'
.}
It now suffices to show for all \math{q\in\Theta}
\\\linemath{\headroom\footroom
    w_0/q
    \refltransindex{\omega}
    \tight\circ
    \redparaindex{\omega+n_1+1}
    \tight\circ
    \refltransindex{\omega+n_1}
    \circ
    \antirefltransindex{\omega+n_0+1}
    w_1/q
}
because then we have 
\bigmath{
  w_0
  \tightequal
  \replpar{w_0}{q}{w_0/q}{q\tightin\Theta}
  \tightequal
  \replpar{\repl{u}{\bar p_0}{r_{0,\bar p_0}\mu_{0,\bar p_0}}}
          {q}{w_0/q}{q\tightin\Theta}
  \tightequal
  \replpar{u}{q}{w_0/q}{q\tightin\Theta}
    \refltransindex{\omega}
    \tight\circ
    \redparaindex{\omega+n_1+1}
    \tight\circ
    \refltransindex{\omega+n_1}
    \circ
    \antirefltransindex{\omega+n_0+1}
  \replpar{u}{q}{w_1/q}{q\tightin\Theta}
  \tightequal
  \\
  \replpar{\replpar{u}{p}{r_{1,p}\mu_{1,p}}{p\tightin\Pi_1}}
          {q}{w_1/q}{q\tightin\Theta}
  \tightequal
  \replpar{w_1}{q}{w_1/q}{q\tightin\Theta}
  \tightequal
  w_1
.}

\noindent
Therefore we are left with the following two cases for \math{q\in\Theta}:

\pagebreak

\yestop
\yestop
\noindent
\underline{\underline{\underline{\math{q\tightnotin\Pi_1}:}}}
Then \bigmath{q\tightequal\bar p_0.}
Define \math{\Pi_1':=\setwith{p}{q p\tightin\Pi_1}}.
\noindent
We have two cases:

\yestop
\noindent
\underline{\underline{``The variable overlap (if any) case'':
\math{
  \forall p\tightin\Pi_1'\tightcap\TPOS{l_{0,q}}\stopq
    l_{0,q}/p\tightin\V
}:}}
\begin{diagram}
l_{0,q}\mu_{0,q}&&\rredparaindex{\omega+n_1+1}&&&&w_1/q
\\&&&&&&\dequal
\\\dredindex{\omega+n_0+1,\,\emptyset}&&&&&&l_{0,q}\nu
\\&&&&&&\dredindex{\omega+n_0+1}
\\w_0/q&\requal&r_{0,q}\mu_{0,q}&&\rredparaindex{\omega+n_1+1}&&r_{0,q}\nu
\end{diagram}
\noindent
Define a function \math\Gamma\ on \V\ by (\math{x\tightin\V}):
\bigmath{
  \Gamma(x):=
  \setwith{(p',p'')}
          {l_{0,q}/p'\tightequal x\ \wedge\ p' p''\in\Pi_1'}
.}

\noindent
\underline{Claim~7:}
There is some \math{\nu\in\Xsubst} with
\\\LINEmath{
  \forall x\in\V\stopq
    \inparenthesesoplist{
       x\mu_{0,q}
       \redparaindex{\omega+n_1+1}
       x\nu
    \oplistund
       \forall p'\tightin\DOM{\Gamma(x)}\stopq
         x\nu
         \tightequal
         \replpar
           {x\mu_{0,q}}
           {p''}
           {r_{1,q p' p''}\mu_{1,q p' p''}}
           {(p',p'')\tightin\Gamma(x)}
    }
.}
\\
\underline{Proof of Claim~7:}
\\
In case of \bigmath{\DOM{\Gamma(x)}\tightequal\emptyset} we define
\bigmath{x\nu:=x\mu_{0,q}.}
If there is some \math{p'} such that 
\bigmath{\DOM{\Gamma(x)}\tightequal\{p'\}}
we define 
\bigmath{
  x\nu
  :=
  \replpar
    {x\mu_{0,q}}
    {p''}
    {r_{1,q p' p''}\mu_{1,q p' p''}}
    {(p',p'')\tightin\Gamma(x)}
.}
This is appropriate since due to 
\bigmath{
  \forall(p',p'')\tightin\Gamma(x)\stopq
    x\mu_{0,q}/p''
    \tightequal 
    l_{0,q}\mu_{0,q}/p' p''
    \tightequal 
    u/q p' p''
    \tightequal 
    l_{1,q p' p''}\mu_{1,q p' p''}
}
we have
\\\LINEmath{
  \begin{array}{l@{}l@{}l}
  x\mu_{0,q}&
  \tightequal&
  \replpar
    {x\mu_{0,q}}
    {p''}
    {l_{1,q p' p''}\mu_{1,q p' p''}}
    {(p',p'')\tightin\Gamma(x)}
  \redparaindex{\omega+n_1+1}\\&&
  \replpar
     {x\mu_{0,q}}
     {p''}
     {r_{1,q p' p''}\mu_{1,q p' p''}}
     {(p',p'')\tightin\Gamma(x)}
  \tightequal
  x\nu.  
  \end{array}
}
\\
Finally, in case of \bigmath{\CARD{\DOM{\Gamma(x)}}\succ1,} \math{l_{0,q}} is
not linear in \math x. By the conditions of our lemma and Claim~5 this implies
\bigmath{x\tightin\Vcons.}
Since there is some \math{(p',p'')\in\Gamma(x)} with
\bigmath{
    x\mu_{0,q}/p''
    \tightequal 
    l_{1,q p' p''}\mu_{1,q p' p''}
}
this implies 
\bigmath{
   l_{1,q p' p''}\mu_{1,q p' p''}\tightin\tcc
}
and then 
\bigmath{
   l_{1,q p' p''}\tightin\tcs
}
which contradicts Claim~5.%
\QED{Claim~7}

\noindent
\underline{Claim~8:}
\bigmath{
  l_{0,q}\nu
  \tightequal 
  w_1/q
.}
\\
\underline{Proof of Claim~8:}
\\
By Claim~7 we get
\bigmath{
  w_1/q
  \tightequal
  \replpar 
    {u/q}
    {p' p''}
    {r_{1,q p' p''}\mu_{1,q p' p''}}
    {\exists x\tightin\V\stopq(p',p'')\tightin\Gamma(x)}
  \tightequal\\
  \replpar
    {\replpar
       {l_{0,q}}
       {p'}
       {x\mu_{0,q}}
       {l_{0,q}/p'\tightequal x\tightin\V}
    }
    {p' p''}
    {r_{1,q p' p''}\mu_{1,q p' p''}}
    {\exists x\tightin\V\stopq(p',p'')\tightin\Gamma(x)}
  \tightequal\\
  \replpar
    {l_{0,q}}
    {p'}
    {\replpar
       {x\mu_{0,q}}
       {p''}
       {r_{1,q p' p''}\mu_{1,q p' p''}}
       {(p',p'')\tightin\Gamma(x)}}
    {l_{0,q}/p'\tightequal x\tightin\V}
  \tightequal\\
  \replpar
    {l_{0,q}}
    {p'}
    {x\nu}
    {l_{0,q}/p'\tightequal x\tightin\V}
  \tightequal
  l_{0,q}\nu
.}\QED{Claim~8}

\noindent
\underline{Claim~9:}
\bigmath{
  w_0/q
  \redparaindex{\omega+n_1+1}
  r_{0,q}\nu
.}
\\
\underline{Proof of Claim~9:} 
Since 
\bigmath{
  w_0/q
  \tightequal
  r_{0,q}\mu_{0,q}
,} 
this follows directly from Claim~7.%
\QED{Claim~9}

\noindent
By claims 8 and 9 it now suffices to show
\bigmath{
  l_{0,q}\nu
  \redindex{\omega+n_0+1}
  r_{0,q}\nu
,}
which again follows from 
Lemma~\ref{lemma invariance of fulfilledness}
since \RX\ is \math\omega-shallow confluent up to
\bigmath{(n_1\tight+1)\plusomega n_0}
by our induction hypothesis 
and since
\bigmath{\forall x\tightin\V\stopq x\mu_{0,q}\refltransindex{\omega+n_1+1}x\nu}
by Claim~7 and Corollary~\ref{corollary parallel one}.%
\\\Qeddouble{``The variable overlap (if any) case''}

\pagebreak

\yestop
\noindent
\underline{\underline{``The critical peak case'':
There is some \math{p\in \Pi_1'\tightcap\TPOS{l_{0,q}}}
with \math{l_{0,q}/p\tightnotin\V}:}}
\begin{diagram}
l_{0,q}\mu_{0,q}&\rredindex{\omega+n_1+1,\,p}&u'
&&&\rredparaindex{\omega+n_1+1,\,\Pi_1'\setminus\{p\}}
&&&w_1/q
\\&&\donlyonceindex{\omega+n_0+1}
&&&
&&&\drefltransindex{\omega+n_0+1}
\\\dredindex{\omega+n_0+1,\,\emptyset}&&v_1
&\rrefltransindex{\omega}&\circ&\rredparaindex{\omega+n_1+1}
&\circ&\rrefltransindex{\omega+n_1}&v_1'
\\&&\drefltransindex{\omega+n_0}
&&&
&&&\drefltransindex{\omega+n_0}
\\w_0/q&\rrefltransindex{\omega}&v_2
&\rrefltransindex{\omega}&\circ&\rredparaindex{\omega+n_1+1}
&\circ&\rrefltransindex{\omega+n_1}&\circ
\end{diagram}
\underline{Claim~10:}
\bigmath{p\tightnotequal\emptyset.}
\\
\underline{Proof of Claim~10:}
If \bigmath{p\tightequal\emptyset,} then
\bigmath{\emptyset\tightin\Pi_1',} then
\bigmath{q\tightin\Pi_1,} which contradicts our global case assumption.%
\QED{Claim~10}

\noindent
Let \math{\xi\in\SUBST\V\V} be a bijection with 
\bigmath{
  \xi[\VAR{\kurzregelindex{1,q p}}]\cap\VAR{\kurzregelindex{0,q}}
  =
  \emptyset
.}
\\
Define
\bigmath{
  \Y
  :=
  \xi[\VAR{\kurzregelindex{1,q p}}]\cup\VAR{\kurzregelindex{0,q}}
.}
\\
Let \math{\varrho\in\Xsubst} be given by
$\ x\varrho=
\left\{\begin{array}{@{}l@{}l@{}}
  x\mu_{0,q}        &\mbox{ if }x\in\VAR{\kurzregelindex{0,q}}\\
  x\xi^{-1}\mu_{1,q p}&\mbox{ else}\\
\end{array}\right\}
\:(x\tightin\V)$.
\\
By
\math{
  l_{1,q p}\xi\varrho
  \tightequal 
  l_{1,q p}\xi\xi^{-1}\mu_{1,q p}
  \tightequal 
  u/q p
  \tightequal 
  l_{0,q}\mu_{0,q}/p  
  \tightequal 
  l_{0,q}\varrho/p
  \tightequal 
  (l_{0,q}/p)\varrho
}
\\
let
\math{
  \sigma:=\minmgu{\{(l_{1,q p}\xi}{l_{0,q}/p)\},\Y}
}
and
\math{\varphi\in\Xsubst}
with
\math{
  \domres{\inpit{\sigma\varphi}}\Y
  \tightequal
  \domres\varrho\Y
}.
\\
Define 
\math{
  u':=  
  \repl{l_{0,q}\mu_{0,q}}
       {p}
       {r_{1,q p}\mu_{1,q p}}
}.
We get
\\\LINEmath{
  \arr{{l@{}l}
    u'\tightequal
    &
    \repl
      {\replpar
         {u/q}
         {p'}
         {l_{1,q p'}\mu_{1,q p'}}
         {p'\tightin\Pi_1'\tightsetminus\{p\}}}
      {p}
      {r_{1,q p}\mu_{1,q p}}
    \redparaindex{\omega+n_1+1,\Pi_1'\setminus\{p\}}
    \\&
    \replpar{u/q}{p'}{r_{1,q p'}\mu_{1,q p'}}{p'\tightin\Pi_1'}    
    \tightequal
    w_1/q
  .
  }
}
\\
If 
\bigmath{
  \repl{l_{0,q}}{p}{r_{1,q p}\xi}\sigma
  \tightequal
  r_{0,q}\sigma
,}
then the proof is finished due to 
\\\LINEmath{
  w_0/q
  \tightequal
  r_{0,q}\mu_{0,q}
  \tightequal
  r_{0,q}\sigma\varphi
  \tightequal
  \repl{l_{0,q}}{p}{r_{1,q p}\xi}\sigma\varphi
  \tightequal
  u'
  \redparaindex{\omega+n_1+1,\Pi_1'\setminus\{p\}}
  w_1/q
.}
\\
Otherwise 
we have
\bigmath{
  (\,
   (\repl{l_{0,q}}{p}{r_{1,q p}\xi}\sigma,
    C_{1,q p}\xi\sigma,    
    1),\penalty-1\,
   (r_{0,q}\sigma,
    C_{0,q}\sigma,
    1),\penalty-1\,
    l_{0,q}\sigma,\penalty-1\,
    p\,)
  \in{\rm CP}(\R)
}
(due to Claim~5);
\bigmath{p\tightnotequal\emptyset}
(due to Claim~10);
\bigmath{C_{1,q p}\xi\sigma\varphi=C_{1,q p}\mu_{1,q p}}
is fulfilled \wrt\ \redindex{\omega+n_1};
\bigmath{C_{0,q}\sigma\varphi=C_{0,q}\mu_{0,q}}
is fulfilled \wrt\ \redindex{\omega+n_0}.
Since 
\bigmath{
  \forall\delta\tightprec(n_1\tight+1)\plusomega(n_0\tight+1)\stopq
  \mbox{\RX\ is \math\omega-shallow confluent up to }\delta
}
(by our induction hypothesis) 
due to our assumed \math\omega-shallow closedness 
(matching the definition's \math{n_0} to our \math{n_1\tight+1}
                   and its \math{n_1} to our \math{n_0\tight+1})
we have
\bigmath{
  u'
  \tightequal
  \repl{l_{0,q}}{p}{r_{1,q p}\xi}\sigma\varphi
  \penalty-1
  \onlyonceindex{\omega+n_0+1}
  \penalty-1
  v_1
  \penalty-1
  \refltransindex{\omega+n_0}
  v_2
  \antirefltransindex{\omega}
  r_{0,q}\sigma\varphi
  \tightequal
  r_{0,q}\mu_{0,q}
  \tightequal
  w_0/q
}
for some \math{v_1}, \math{v_2}.
We then have
\bigmath{
  v_1
  \antionlyonceindex{\omega+n_0+1} 
  u'
  \redparaindex{\omega+n_1+1,\Pi_1'\setminus\{p\}}
  w_1/q
.}
By 
\bigmath{
  \CARD{\Pi_1'\tightsetminus\{p\}}
  \prec
  \CARD{\Pi_1'}
  \preceq
  \CARD{\Pi_1}
,}
due to our second induction level 
we get some \math{v_1'} with 
\bigmath{
  v_1
  \refltransindex{\omega}
  \tight\circ
  \redparaindex{\omega+n_1+1}
  \tight\circ
  \refltransindex{\omega+n_1}
  v_1'
  \antirefltransindex{\omega+n_0+1}
  w_1/q
.}
Finally by our induction hypothesis that
  \bigmath{ 
    \refltransindex{\omega}
    \tight\circ
    \redparaindex{\omega+n_1+1}
    \tight\circ
    \refltransindex{\omega+n_1}
  }
  strongly commutes over
  \refltransindex{\omega+n_0}
the peak at \math{v_1} can be closed according to 
\bigmath{
  v_2
  \refltransindex{\omega}
  \tight\circ
  \redparaindex{\omega+n_1}
  \tight\circ
  \refltransindex{\omega+n_1}
  \circ
  \antirefltransindex{\omega+n_0}
  v_1'
.}%
\\
\Qeddouble{``The critical peak case''}\QEDtriple{``\math{q\tightnotin\Pi_1}''}

\pagebreak

\noindent
\underline{\underline{\underline{\math{q\tightin\Pi_1}:}}}
If there is  no \math{\bar p_0'} with \bigmath{q\bar p_0'\tightequal\bar p_0,}
then the proof is finished due to 
\bigmath{
  w_0/q
  \tightequal
  u/q
  \tightequal
  l_{1,q}\mu_{1,q}
  \redindex{\omega+n_1+1}
  r_{1,q}\mu_{1,q}
  \tightequal
  w_1/q
.}
Otherwise, we can define \math{\bar p_0'} by
\bigmath{q\bar p_0'\tightequal\bar p_0.}
We have two cases:

\yestop
\noindent
\underline{\underline{%
``The second variable overlap case'':%
}}
\\\LINEnomath{\underline{\underline{%
There are \math{x\tightin\V} and \math{p'}, \math{p''} such that
\math{
    l_{1,q}/p'
    \tightequal
    x
  \und
    p'p''
    \tightequal
    \bar p_0'
}:%
}}}
\begin{diagram}
l_{1,q}\mu_{1,q}&&&\rredindex{\omega+n_1+1,\,\emptyset}&&&w_1/q
\\&&&&&&\dequal
\\\dredindex{\omega+n_0+1,\,\bar p_0'}&&&&&&r_{1,q}\mu_{1,q}
\\&&&&&&\dredparaindex{\omega+n_0+1}
\\w_0/q&\requal&l_{1,q}\nu&&\rredindex{\omega+n_1+1}&&r_{1,q}\nu
\end{diagram}
\noindent
\underline{Claim~11a:}
We have 
\bigmath{x\mu_{1,q}/p''\tightequal l_{0,\bar p_0}\mu_{0,\bar p_0}}
and may assume
\bigmath{x\tightin\Vsig.}
\\\underline{Proof of Claim~11a:}
We have 
\bigmath{
  x\mu_{1,q}/p''
  \tightequal
  l_{1,q}\mu_{1,q}/p' p''
  \tightequal
  u/q p' p''
  \tightequal
  u/q\bar p_0'
  \tightequal
  u/\bar p_0
  \tightequal
  l_{0,\bar p_0}\mu_{0,\bar p_0}
.}
If \bigmath{x\tightin\Vcons,}
then \bigmath{x\mu_{1,q}\tightin\tcc,}
then \bigmath{x\mu_{1,q}/p''\tightin\tcc,}
then \bigmath{l_{0,\bar p_0}\mu_{0,\bar p_0}\tightin\tcc,}
and then \bigmath{l_{0,\bar p_0}\tightin\tcs}
which we may assume not to be the case by Claim~5.%
\QED{Claim~11a}

\noindent
\underline{Claim~11b:}
We can define \math{\nu\in\Xsubst} by
\bigmath{
  x\nu
  \tightequal
  \repl
    {x\mu_{1,q}}
    {p''}
    {r_{0,\bar p_0}\mu_{0,\bar p_0}}
}
and
\bigmath{
  \forall y\tightin\V\tightsetminus\{x\}\stopq y\nu\tightequal y\mu_{1,q}
.}
Then we have 
\bigmath{
  x\mu_{1,q}\redindex{\omega+n_0+1}x\nu
.}
\\
\underline{Proof of Claim~11b:}
This follows directly from Claim~11a.%
\QED{Claim~11b}

\noindent
\underline{Claim~12:}
\bigmath{
  w_0/q
  \tightequal
  l_{1,q}\nu
.}
\\
\underline{Proof of Claim~12:}
By the left-linearity assumption of our lemma, Claim~5, and Claim~11a 
we may assume
\bigmath{
  \setwith
    {p'''}
    {l_{1,q}/p'''\tightequal x}
  =
  \{p'\}
.}
Thus, by Claim~11b we get
\bigmath{
  w_0/q
  \tightequal
  \repl 
    {u/q}
    {\bar p_0'}
    {r_{0,\bar p_0}\mu_{0,\bar p_0}}
  \tightequal\\
  \repl
    {\replpar
       {l_{1,q}}
       {p'''}
       {y\mu_{1,q}}
       {l_{1,q}/p'''\tightequal y\tightin\V}
    }
    {\bar p_0'}
    {r_{0,\bar p_0}\mu_{0,\bar p_0}}
  \tightequal\\
  \repl
    {\repl
       {\replpar
          {l_{1,q}}
          {p'''}
          {y\mu_{1,q}}
          {l_{1,q}/p'''\tightequal y\tightin\V\und y\tightnotequal x}
       }
       {p'}
       {x\mu_{1,q}}
    }
    {p' p''}
    {r_{0,\bar p_0}\mu_{0,\bar p_0}}
  \tightequal\\
  \repl
    {\replpar
       {l_{1,q}}
       {p'''}
       {y\nu}
       {l_{1,q}/p'''\tightequal y\tightin\V\und y\tightnotequal x}
    }
    {p'}
    {\repl
       {x\mu_{1,q}}
       {p''}
       {r_{0,\bar p_0}\mu_{0,\bar p_0}}
    }
  \tightequal\\
  \replpar
    {l_{1,q}}
    {p'''}
    {y\nu}
    {l_{1,q}/p'''\tightequal y\tightin\V}
  \tightequal
  l_{1,q}\nu
.}\QED{Claim~12}

\noindent
\underline{Claim~13:}
\bigmath{
  r_{1,q}\nu
  \antiredparaindex{\omega+n_0+1}
  w_1/q
.}
\\
\underline{Proof of Claim~13:} 
Since \bigmath{r_{1,q}\mu_{1,q}\tightequal w_1/q,} 
this follows directly from Claim~11b.%
\QED{Claim~13}

\noindent
By claims 12 and 13 
using Corollary~\ref{corollary parallel one} 
it now suffices to show
\bigmath{
  l_{1,q}\nu
  \redindex{\omega+n_1+1}
  r_{1,q}\nu
,}
which again follows from 
Claim~11b,
Lemma~\ref{lemma invariance of fulfilledness}
(matching 
 its \math{n_0} to our \math{n_0\tight+1} and 
 its \math{n_1} to our \math{n_1}),
and our induction hypothesis that \RX\ is \math\omega-shallow confluent up to
\bigmath{
  (n_0\tight+1)\plusomega n_1
.}%
\\\Qeddouble{``The second variable overlap case''}

\pagebreak

\yestop
\noindent
\underline{\underline{``The second critical peak case'':
\math{
    \bar p_0'\tightin\TPOS{l_{1,q}}
  \und
    l_{1,q}/\bar p_0'\tightnotin\V
}:%
}}
\begin{diagram}
l_{1,q}\mu_{1,q}
&&&
&\rredindex{\omega+n_1+1,\,\emptyset}&&
&&w_1/q
\\\dredindex{\omega+n_0+1,\,\bar p_0'}
&&&
&&&
&&\drefltransindex{\omega+n_0+1}
\\w_0/q
&\rrefltransindex\omega&\circ&
&\rredparaindex{\omega+n_1+1}&&\circ
&\rrefltransindex{\omega+n_1}&\circ
\end{diagram}
Let \math{\xi\in\SUBST\V\V} be a bijection with 
\bigmath{
  \xi[\VAR{\kurzregelindex{0,\bar p_0}}]\cap\VAR{\kurzregelindex{1,q}}
  =
  \emptyset
.}
\\
Define
\bigmath{
  \Y
  :=
  \xi[\VAR{\kurzregelindex{0,\bar p_0}}]\cup\VAR{\kurzregelindex{1,q}}
.}
\\
Let \math{\varrho\in\Xsubst} be given by
$\ x\varrho=
\left\{\begin{array}{@{}l@{}l@{}}
  x\mu_{1,q}        &\mbox{ if }x\in\VAR{\kurzregelindex{1,q}}\\
  x\xi^{-1}\mu_{0,\bar p_0}&\mbox{ else}\\
\end{array}\right\}
\:(x\tightin\V)$.
\\
By
\math{
  l_{0,\bar p_0}\xi\varrho
  \tightequal 
  l_{0,\bar p_0}\xi\xi^{-1}\mu_{0,\bar p_0}
  \tightequal 
  u/\bar p_0
  \tightequal 
  u/q\bar p_0'
  \tightequal 
  l_{1,q}\mu_{1,q}/\bar p_0'  
  \tightequal 
  l_{1,q}\varrho/\bar p_0'
  \tightequal 
  (l_{1,q}/\bar p_0')\varrho
}
\\
let
\math{
  \sigma:=\minmgu{\{(l_{0,\bar p_0}\xi}{l_{1,q}/\bar p_0')\},\Y}
}
and
\math{\varphi\in\Xsubst}
with
\math{
  \domres{\inpit{\sigma\varphi}}\Y
  \tightequal
  \domres\varrho\Y
}.
\\
If 
\bigmath{
  \repl{l_{1,q}}{\bar p_0'}{r_{0,\bar p_0}\xi}\sigma
  \tightequal
  r_{1,q}\sigma
,}
then the proof is finished due to 
\\\linemath{
  w_0/q
  \tightequal
  \repl{l_{1,q}\mu_{1,q}}{\bar p_0'}{r_{0,\bar p_0}\mu_{0,\bar p_0}}
  \tightequal
  \repl{l_{1,q}}{\bar p_0'}{r_{0,\bar p_0}\xi}\sigma\varphi
  \tightequal
  r_{1,q}\sigma\varphi
  \tightequal
  r_{1,q}\mu_{1,q}
  \tightequal
  w_1/q.
}
Otherwise 
we have
\bigmath{
  (\,
   (\repl{l_{1,q}}{\bar p_0'}{r_{0,\bar p_0}\xi}\sigma,
    C_{0,\bar p_0}\xi\sigma,    
    1),\penalty-1\,
   (r_{1,q}\sigma,
    C_{1,q}\sigma,
    1),\penalty-1\,
    l_{1,q}\sigma,\penalty-1\,
    \bar p_0'\,)
  \in{\rm CP}(\R)
}
(due to Claim~5);
\bigmath{C_{0,\bar p_0}\xi\sigma\varphi=C_{0,\bar p_0}\mu_{0,\bar p_0}}
is fulfilled \wrt\ \redindex{\omega+n_0};
\bigmath{C_{1,q}\sigma\varphi=C_{1,q}\mu_{1,q}}
is fulfilled \wrt\ \redindex{\omega+n_1}.
Since 
\bigmath{
  \forall\delta\tightprec(n_0\tight+1)\plusomega(n_1\tight+1)\stopq
  \mbox{\RX\ is \math\omega-shallow confluent up to }\delta
}
(by our induction hypothesis) 
due to our assumed \math\omega-shallow noisy weak parallel joinability 
(matching the definition's \math{n_0} to our \math{n_0\tight+1}
                   and its \math{n_1} to our \math{n_1\tight+1})
we have
\bigmath{
  w_0/q
  \tightequal
  \repl{l_{1,q}\mu_{1,q}}{\bar p_0'}{r_{0,\bar p_0}\mu_{0,\bar p_0}}
  \tightequal
  \repl{l_{1,q}}{\bar p_0'}{r_{0,\bar p_0}\xi}\sigma\varphi
  \refltransindex{\omega}
  \tight\circ
  \redparaindex{\omega+n_1+1}
  \tight\circ
  \refltransindex{\omega+n_1}
  \circ
  \antirefltransindex{\omega+n_0+1}
  \penalty-1
  r_{1,q}\sigma\varphi
  \tightequal
  r_{1,q}\mu_{1,q}
  \tightequal
  w_1/q
.}
\\\Qeddouble{``The second critical peak case''}
\end{proofparsepqed}

\begin{proofparsepqed}{Lemma~\ref{lemma strongly closed second level two}}
\underline{Claim~0:}
\RX\ is \math\omega-shallow confluent up to \math\omega.
\\\underline{Proof of Claim~0:}
Directly by the assumed strong commutation,
\cf\ the proofs of the claims 2 and 3 of the proof of 
Lemma~\ref{lemma parallel closed first level two}.\QED{Claim~0}

\yestop
\noindent
\underline{Claim~1:}
If 
\bigmath{
    \refltransindex{\omega}
    \tight\circ
    \redindex{\omega+n_1}
    \tight\circ
    \refltransindex{\omega+(n_1\monus1)}
}
strongly commutes over
\refltransindex{\omega+n_0},
then
\redindex{\omega+n_1} and \redindex{\omega+n_0} are commuting.
\\
\underline{Proof of Claim~1:}
\bigmath{
    \refltransindex{\omega}
    \tight\circ
    \redindex{\omega+n_1}
    \tight\circ
    \refltransindex{\omega+(n_1\monus1)}
}
and
\refltransindex{\omega+n_0}
are commuting
by Lemma~\ref{lemma strong commutation one copy}.
Since by \lemmamonotonicinbeta\
we have
\bigmath{
  \redindex{\omega+n_1}
  \subseteq
  \refltransindex{\omega}
  \tight\circ
  \redindex{\omega+n_1}
  \tight\circ
  \refltransindex{\omega+(n_1\monus1)}
  \subseteq
  \refltransindex{\omega+n_1}
,}
now \redindex{\omega+n_1} 
and \redindex{\omega+n_0} are commuting, too.\QED{Claim~1}

\yestop
\yestop
\noindent
For \math{n_0\preceq n_1\prec\omega} 
we are going to show by induction on \math{n_0\plusomega n_1}
the following property\footroom:
\\\LINEmath{
    w_0
    \antiredindex{\omega+n_0}
    u
    \redindex{\omega+n_1}
    w_1
  \quad\implies\quad
    w_0
    \refltransindex{\omega}
    \tight\circ
    \onlyonceindex{\omega+n_1}
    \tight\circ
    \refltransindex{\omega+(n_1\monus1)}
    \circ
    \antirefltransindex{\omega+n_0}
    w_1
.}
\begin{diagram}
u&&&\rredindex{\omega+n_1}&&&w_1
\\\dredindex{\omega+n_0}&&&&&&\drefltransindex{\omega+n_0}
\\w_0&\rrefltransindex{\omega}&\circ&\ronlyonceindex{\omega+n_1}
&\circ&\rrefltransindex{\omega+(n_1\monus1)}&\circ
\end{diagram}

\pagebreak

\noindent
\underline{Claim~2:}
Let \math{\delta\prec\omega\tight+\omega}.
If
\\
\linemath{
  \forall n_0,n_1\tightprec\omega\stopq
  \inparenthesesoplist{ 
    \inparenthesesoplist{
        n_0\tightpreceq n_1      
      \oplistund
        n_0\plusomega n_1\tightpreceq\delta
    }
    \oplistimplies
    \forall w_0,w_1,u\stopq
    \inparenthesesoplist{
        w_0
        \antiredindex{\omega+n_0}
        u
        \redindex{\omega+n_1}
        w_1
      \oplistimplies 
        w_0
        \refltransindex{\omega}
        \tight\circ
        \onlyonceindex{\omega+n_1}
        \tight\circ
        \refltransindex{\omega+(n_1\monus1)}
        \circ
        \antirefltransindex{\omega+n_0}
        w_1
    }
  }
,}
then
\\
\linemath{
  \forall n_0,n_1\tightprec\omega\stopq
  \inparenthesesoplist{ 
    \inparenthesesoplist{
        n_0\tightpreceq n_1      
      \oplistund
        n_0\plusomega n_1\tightpreceq\delta
    }
    \oplistimplies 
            \refltransindex{\omega}
            \tight\circ
            \redindex{\omega+n_1}
            \tight\circ
            \refltransindex{\omega+(n_1\monus1)}
      \mbox{ strongly commutes over }
            \refltransindex{\omega+n_0}
  }
,}
and 
\RX\ is \math\omega-shallow confluent up to \math{\delta}.
\\
\underline{Proof of Claim~2:}
By induction on \math{\delta} in \tightprec.
First we show the strong commutation.
Assume \math{n_0\preceq n_1\prec\omega} with
\math{n_0\plusomega n_1\tightpreceq\delta}. 
By Lemma~\ref{lemma strong commutation one copy} it suffices to show that
\bigmath{
    \refltransindex{\omega}
    \tight\circ
    \redindex{\omega+n_1}
    \tight\circ
    \refltransindex{\omega+(n_1\monus1)}
}
strongly commutes over
\redindex{\omega+n_0}.
Assume
\bigmath{
    u''
    \antiredindex{\omega+n_0}
    u'
    \refltransindex{\omega}
    u
    \redindex{\omega+n_1}
    w_1
    \refltransindex{\omega+(n_1\monus1)}
    w_2
}
(\cf\ diagram below).
By the strong commutation assumed for our lemma,
there are \math{w_0} and \math{w_0'} with
\bigmath{
    u''
    \refltransindex{\omega}
    w_0'
    \antirefltransindex{\omega+(n_0\monus1)}
    w_0
    \antionlyonceindex{\omega+n_0}
    u
.}
By the above property there are some \math{w_3}, \math{w_1'}
with
\bigmath{
    w_0
    \refltransindex{\omega}
    w_3
    \onlyonceindex{\omega+n_1}
    \tight\circ
    \refltransindex{\omega+(n_1\monus1)}
    w_1'
    \antirefltransindex{\omega+n_0}
    w_1
.}
Next we show that we can close the peak
\bigmath{ 
    w_1'
    \antirefltransindex{\omega+n_0}
    w_1
    \refltransindex{\omega+(n_1\monus1)}
    w_2
}
according to 
\bigmath{
    w_1'
    \refltransindex{\omega+(n_1\monus1)}
    w_2'
    \antirefltransindex{\omega+n_0}
    w_2
}
for some \math{w_2'}.
In case of \bigmath{n_1\tightequal0}
this is possible due to the
\math\omega-shallow confluence up to \math{\omega}
given by Claim~0.
Otherwise we have 
\bigmath{
  n_0\plusomega(n_1\monus1)
  \tightprec 
  n_0\plusomega n_1
  \tightpreceq
  \delta
} 
and due to our induction hypothesis
(saying that \RX\ is \math\omega-shallow confluent up to all 
\math{\delta'\prec\delta
})
this is possible again.
By Claim~0 again,
we can close the peak
\bigmath{
  w_0'
  \antirefltransindex{\omega+(n_0\monus1)}
  w_0
  \refltransindex{\omega}
  w_3
}
according to 
\bigmath{
  w_0'
  \refltransindex{\omega}
  w_3'
  \antirefltransindex{\omega+(n_0\monus1)}
  w_3
}
for some \math{w_3'}.
To close the whole diagram, we only have to show that we can close the peak
\bigmath{
    w_3'
    \antirefltransindex{\omega+(n_0\monus1)}
    w_3
    \onlyonceindex{\omega+n_1}
    \tight\circ
    \refltransindex{\omega+(n_1\monus1)}
    w_2'
}
according to 
\bigmath{
    w_3'
    \refltransindex{\omega}
    \tight\circ
    \onlyonceindex{\omega+n_1}
    \tight\circ
    \refltransindex{\omega+(n_1\monus1)}
    \circ
    \antirefltransindex{\omega+(n_0\monus1)}
    w_2'
.}
In case of \bigmath{n_0\tightequal0}
this is possible due to the strong commutation assumed for our lemma.
Otherwise we have 
\bigmath{
  n_0\monus1
  \tightprec 
  n_0
  \tightpreceq
  n_1
} 
and
\bigmath{
  (n_0\monus1)\plusomega n_1
  \tightprec 
  n_0\plusomega n_1
  \tightpreceq
  \delta
,} 
and then due to our
induction hypothesis
this is possible again.
\begin{diagram}
u'&\rrefltransindex{\omega}&u
&&
&\rredindex{\omega+n_1}&&
&w_1&\rrefltransindex{\omega+(n_1\monus1)}&w_2
\\&&\donlyonceindex{\omega+n_0}
&&
&&&
&\drefltransindex{\omega+n_0}&&\drefltransindex{\omega+n_0}
\\\dredindex{\omega+n_0}&&w_0
&\rrefltransindex{\omega}&w_3
&\ronlyonceindex{\omega+n_1}&\circ&\rrefltransindex{\omega+(n_1\monus1)}
&w_1'&\rrefltransindex{\omega+(n_1\monus1)}&w_2'
\\&&\drefltransindex{\omega+(n_0\monus1)}
&&\drefltransindex{\omega+(n_0\monus1)}
&&&
&&&\drefltransindex{\omega+(n_0\monus1)}
\\u''&\rrefltransindex{\omega}&w_0'
&\rrefltransindex{\omega}&w_3'
&\rrefltransindex{\omega}&\circ&\ronlyonceindex{\omega+n_1}
&\circ&\rrefltransindex{\omega+(n_1\monus1)}&\circ
\end{diagram}
Finally we show \math\omega-shallow confluence up to \math\delta.
Assume
\bigmath{n_0\plusomega n_1\tightpreceq\delta}
and
\bigmath{
    w_0
    \antirefltransindex{\omega+n_0}
    u
    \refltransindex    {\omega+n_1}
    w_1
.}
Due to symmetry in \math{n_0} and \math{n_1} we may assume
\bigmath{n_0\tightpreceq n_1.}
Above we have shown that 
\bigmath{
    \refltransindex{\omega}
    \tight\circ
    \redindex{\omega+n_1}
    \tight\circ
    \refltransindex{\omega+(n_1\monus1)}
}
strongly commutes over
\refltransindex{\omega+n_0}. 
By Claim~1 we finally get
\bigmath{
  w_0\refltransindex{\omega+n_1}\circ\antirefltransindex{\omega+n_0} w_1
}
as desired.%
\QED{Claim~2}

\pagebreak

\yestop
\yestop
\noindent
Note that 
for \bigmath{n_0\tightequal0} 
our property follows 
from 
the strong commutation assumption of our lemma.

The benefit of 
Claim~2 is twofold: First, it says that our lemma is valid
if the above property
holds for all \math{n_0\preceq n_1\prec\omega}.
Second, it strengthens the property when used as induction hypothesis. Thus 
(writing \math{n_i\tight+1} instead of \math{n_i} since we may assume
 \math{0\tightprec n_0\tightpreceq n_1})
it
now suffices to show
for
\math{n_0\preceq n_1\prec\omega}
that 
\\\linemath{
    w_0
    \antiredindex{\omega+n_0+1,\bar p_0}
    u
    \redindex    {\omega+n_1+1,\bar p_1}
    w_1
}
together with our induction hypotheses 
that\headroom
\\\linemath{\headroom
  \forall\delta\tightprec(n_0\tight+1)\plusomega(n_1\tight+1)\stopq
  \mbox{\RX\ is \math\omega-shallow confluent up to }\delta
}
\headroom
implies
\\\LINEmath{
    w_0
    \refltransindex{\omega}
    \tight\circ
    \onlyonceindex{\omega+n_1+1}
    \tight\circ
    \refltransindex{\omega+n_1}
    \circ
    \antirefltransindex{\omega+n_0+1}
    w_1
.}
\begin{diagram}
u&&&\rredindex{\omega+n_1+1,\,\bar p_1}
&&&w_1
\\
\dredindex{\omega+n_0+1,\,\bar p_0}&&&
&&&\drefltransindex{\omega+n_0+1}
\\
w_0&\rrefltransindex{\omega}&\circ&\ronlyonceindex{\omega+n_1+1}
&\circ&\rrefltransindex{\omega+n_1}&\circ
\end{diagram}

\yestop
\noindent
Now for each \math{i\prec2} there are
\bigmath{\kurzregelindex{i}\in\R}
and
\bigmath{\mu_{i}\in\Xsubst}
with
\bigmath{u/\bar p_i\tightequal l_{i}\mu_{i},}
\bigmath{
  w_i\tightequal\repl{u}{\bar p_i}{r_{i}\mu_{i}}
,}
and
\math{C_{i}\mu_{i}} fulfilled \wrt\ \redindex{\omega+n_i}.

\yestop
\noindent
\underline{Claim~5:}
We may assume 
\bigmath{
  \forall i\tightprec2\stopq
    l_{i}\tightnotin\tcs
.}
\\
\underline{Proof of Claim~5:}
In case of \bigmath{l_i\tightin\tcs} we get \bigmath{u\redindex\omega w_i}
by Lemma~\ref{lemma invariance of fulfilledness two}
(matching both its \math\mu\ and \math\nu\ to our \math{\mu_i}).
In case of ``\math{i\tightequal0}'' our property follows from the
strong commutation assumption of our lemma. 
In case of ``\math{i\tightequal1}'' our property follows from Claim~0.%
\QED{Claim~5}

\yestop
\yestop
\noindent
In case of \bigmath{\neitherprefix{\bar p_0}{\bar p_1}} 
we have 
\bigmath{
  w_{i}/\bar p_{1-i}
  \tightequal
  \repl u{\bar p_{i}}{r_{i}\mu_{i}}/\bar p_{1-i}
  \tightequal
  u/\bar p_{1-i}
  \tightequal
  l_{1-i}\mu_{1-i}
} 
and
therefore
\bigmath{
  w_{i}
  \redindex{\omega+n_{i}+1}
  \replpar u{\bar p_{k}}{r_{k}\mu_{k}}{k\tightprec2}
,}
\ie\ our proof is finished.
Thus, according to whether \math{\bar p_0} is a prefix of \math{\bar p_1}
or vice versa, we have the following two cases left:
\pagebreak

\yestop
\yestop
\noindent
\underline{\underline{\underline{%
There is some \math{\bar p_1'} with
\bigmath{\bar p_0\bar p_1'\tightequal\bar p_1} and
\bigmath{\bar p_1'\tightnotequal\emptyset}:%
}}}

\noindent
We have two cases:

\noindent
\underline{\underline{``The variable overlap case'':}}
\\\LINEnomath{\underline{\underline{%
There are \math{x\in\V} and \math{p'}, \math{p''} such that
\math{
  l_{0}/p'\tightequal x
  \und
  p' p''\tightequal\bar p_1'
}:}}}
\begin{diagram}
l_{0}\mu_{0}&&&\rredindex{\omega+n_1+1,\,\bar p_1'}&&&w_1/\bar p_0
\\
&&&&&&\dequal
\\
\dredindex{\omega+n_0+1,\,\emptyset}&&&&&&l_{0}\nu
\\
&&&&&&\dredindex{\omega+n_0+1}
\\
w_0/\bar p_0&\requal&r_{0}\mu_{0}&&\ronlyonceindex{\omega+n_1+1}
&&r_{0}\nu
\end{diagram}
\noindent
\underline{Claim~6:}
We have 
\bigmath{x\mu_{0}/p''\tightequal l_{1}\mu_{1}}
and may assume
\bigmath{x\tightin\Vsig.}
\\\underline{Proof of Claim~6:}
We have 
\bigmath{
  x\mu_{0}/p''
  \tightequal
  l_{0}\mu_{0}/p' p''
  \tightequal
  u/\bar p_0 p' p''
  \tightequal
  u/\bar p_0\bar p_1'
  \tightequal
  u/\bar p_1
  \tightequal
  l_{1}\mu_{1}
.}
\\
If 
\bigmath{x\tightin\Vcons,}
then \bigmath{x\mu_{0}\tightin\tcc,}
then \bigmath{x\mu_{0}/p''\tightin\tcc,}
then 
\\
\bigmath{l_{1}\mu_{1}\tightin\tcc,}
and then \bigmath{l_{1}\tightin\tcs}
which we may assume not to be the case by Claim~5.%
\QED{Claim~6}

\noindent
\underline{Claim~7:}
We can define \math{\nu\in\Xsubst} by
\bigmath{
  x\nu
  \tightequal
  \repl
    {x\mu_{0}}
    {p''}
    {r_{1}\mu_{1}}
}
and
\bigmath{
  \forall y\tightin\V\tightsetminus\{x\}\stopq 
    y\nu\tightequal y\mu_{0}
.}
Then we have 
\bigmath{
  x\mu_{0}
  \redindex{\omega+n_1+1}
  x\nu
.}
\\
\underline{Proof of Claim~7:}
This follows directly from Claim~6.%
\QED{Claim~7}

\noindent
\underline{Claim~8:}
\bigmath{
  l_{0}\nu
  \tightequal
  w_1/\bar p_0
.}
\\
\underline{Proof of Claim~8:}
By the left-linearity assumption of our lemma and claims 5 and 6 we may assume
\bigmath{
  \setwith
    {p'''}
    {l_{0}/p'''\tightequal x}
  =
  \{p'\}
.}
Thus, by Claim~7 we get
\bigmath{
  w_1/\bar p_0
  \tightequal
  \repl 
    {u/\bar p_0}
    {\bar p_1'}
    {r_{1}\mu_{1}}
  \tightequal\\
  \repl
    {\replpar
       {l_{0}}
       {p'''}
       {y\mu_{0}}
       {l_{0}/p'''\tightequal y\tightin\V}
    }
    {\bar p_1'}
    {r_{1}\mu_{1}}
  \tightequal\\
  \repl
    {\repl
       {\replpar
          {l_{0}}
          {p'''}
          {y\mu_{0}}
          {l_{0}/p'''\tightequal y\tightin\V\und y\tightnotequal x}
       }
       {p'}
       {x\mu_{0}}
    }
    {p' p''}
    {r_{1}\mu_{1}}
  \tightequal\\
  \repl
    {\replpar
       {l_{0}}
       {p'''}
       {y\nu}
       {l_{0}/p'''\tightequal y\tightin\V\und y\tightnotequal x}
    }
    {p'}
    {\repl
       {x\mu_{0}}
       {p''}
       {r_{1}\mu_{1}}
    }
  \tightequal\\
  \replpar
    {l_{0}}
    {p'''}
    {y\nu}
    {l_{0}/p'''\tightequal y\tightin\V}
  \tightequal
  l_{0}\nu
.}\QED{Claim~8}

\noindent
\underline{Claim~9:}
\bigmath{
  w_0/\bar p_0
  \onlyonceindex{\omega+n_1+1}
  r_{0}\nu
.}
\\
\underline{Proof of Claim~9:} 
By the right-linearity assumption of our lemma and claims 5 and 6
we may assume 
\bigmath{\CARD{\setwith{p'''}{r_{0}/p'''\tightequal x}}\tightpreceq1.}
Thus by Claim~7 we get:
\bigmath{
  w_0/\bar p_0
  \tightequal 
  r_{0}\mu_{0}
  \tightequal
%  \replpar
%    {r_{0}}
%    {p'''}
%    {y\mu_{0}}
%    {r_{0}/p'''\tightequal y\tightin\V}
%  \tightequal
  \\
  \replpar
    {\replpar
       {r_{0}}
       {p'''}
       {y\mu_{0}}
       {r_{0}/p'''\tightequal y\tightin\V\tightsetminus\{x\}}
    }
    {p'''}
    {x\mu_{0}}
    {r_{0}/p'''\tightequal x}
  \onlyonceindex{\omega+n_1+1}
  \\
  \replpar
    {\replpar
       {r_{0}}
       {p'''}
       {y\mu_{0}}
       {r_{0}/p'''\tightequal y\tightin\V\tightsetminus\{x\}}
    }
    {p'''}
    {x\nu}
    {r_{0}/p'''\tightequal x}
  \tightequal
  \\
  \replpar
    {\replpar
       {r_{0}}
       {p'''}
       {y\nu}
       {r_{0}/p'''\tightequal y\tightin\V\tightsetminus\{x\}}
    }
    {p'''}
    {x\nu}
    {r_{0}/p'''\tightequal x}
  \tightequal
  r_{0}\nu
.}%
\QED{Claim~9}

\noindent
By claims 8 and 9 it now suffices to show
\bigmath{
  l_{0}\nu
  \redindex{\omega+n_0+1}
  r_{0}\nu
,}
which again follows from 
Lemma~\ref{lemma invariance of fulfilledness}
(matching its \math{n_0} to our \math{n_1\tight+1}
      and its \math{n_1} to our \math{n_0})
since \RX\ is quasi-normal and \math\omega-shallow confluent up to
\bigmath{(n_1\tight+1)\plusomega n_0}
by our induction hypothesis,
and since
\bigmath{
  \forall y\tightin\V\stopq 
  y\mu_{0}
  \refltransindex{\omega+n_1+1}
  y\nu
}
by Claim~7.\QEDdouble{``The variable overlap case''}

\pagebreak

\yestop
\noindent
\underline{\underline{``The critical peak case'':
\math{
  \bar p_1'\tightin\TPOS{l_{0}}
  \und
  l_{0}/\bar p_1'\tightnotin\V
}:}}
\begin{diagram}
l_{0}\mu_{0}&&
&\rredindex{\omega+n_1+1,\,\bar p_1'}&&&w_1/\bar p_0
\\\dredindex{\omega+n_0+1,\,\emptyset}&&
&&&&\drefltransindex{\omega+n_0+1}
\\w_0/\bar p_0&\rrefltransindex\omega&\circ
&\ronlyonceindex{\omega+n_1+1}&\circ
&\rrefltransindex{\omega+n_1}&\circ
\end{diagram}
Let \math{\xi\in\SUBST\V\V} be a bijection with 
\bigmath{
  \xi[\VAR{\kurzregelindex{1}}]\cap\VAR{\kurzregelindex{0}}
  =
  \emptyset
.}
\\
Define
\bigmath{
  \Y
  :=
  \xi[\VAR{\kurzregelindex{1}}]\cup\VAR{\kurzregelindex{0}}
.}
\\
Let \math{\varrho\in\Xsubst} be given by
$\ x\varrho=
\left\{\begin{array}{@{}l@{}l@{}}
  x\mu_{0}        &\mbox{ if }x\in\VAR{\kurzregelindex{0}}\\
  x\xi^{-1}\mu_{1}&\mbox{ else}\\
\end{array}\right\}
\:(x\tightin\V)$.
\\
By
\math{
  l_{1}\xi\varrho
  \tightequal 
  l_{1}\xi\xi^{-1}\mu_{1}
  \tightequal 
  u/\bar p_1
  \tightequal 
  u/\bar p_0\bar p_1'
  \tightequal 
  l_{0}\mu_{0}/\bar p_1'  
  \tightequal 
  l_{0}\varrho/\bar p_1'
  \tightequal 
  (l_{0}/\bar p_1')\varrho
}
\\
let
\math{
  \sigma:=\minmgu{\{(l_{1}\xi}{l_{0}/\bar p_1')\},\Y}
}
and
\math{\varphi\in\Xsubst}
with
\math{
  \domres{\inpit{\sigma\varphi}}\Y
  \tightequal
  \domres\varrho\Y
}.
\\
If 
\bigmath{
  \repl{l_{0}}{\bar p_1'}{r_{1}\xi}\sigma
  \tightequal
  r_{0}\sigma
,}
then the proof is finished due to 
\\\LINEmath{
  w_0/\bar p_0
  \tightequal
  r_{0}\mu_{0}
  \tightequal
  r_{0}\sigma\varphi
  \tightequal
  \repl{l_{0}}{\bar p_1'}{r_{1}\xi}\sigma\varphi
  \tightequal
  \repl{l_{0}\mu_{0}}{\bar p_1'}{r_{1}\mu_{1}}
  \tightequal
  w_1/\bar p_0
.}
\\
Otherwise 
we have
\bigmath{
  (\,
   (\repl{l_{0}}{\bar p_1'}{r_{1}\xi},
    C_{1}\xi,    
    1),\penalty-1\,
   (r_{0},
    C_{0},
    1),\penalty-1\,
    l_{0},\penalty-1\,
    \sigma,\penalty-1\,
    \bar p_1'\,)
  \in{\rm CP}(\R)
}
(due to Claim~5);
\bigmath{\bar p_1'\tightnotequal\emptyset}
(due the global case assumption);
\bigmath{C_{1}\xi\sigma\varphi=C_{1}\mu_{1}}
is fulfilled \wrt\ \redindex{\omega+n_1};
\bigmath{C_{0}\sigma\varphi=C_{0}\mu_{0}}
is fulfilled \wrt\ \redindex{\omega+n_0}.
Since 
\bigmath{
  \forall\delta\tightprec(n_1\tight+1)\plusomega(n_0\tight+1)\stopq
    \RX\mbox{ is \math\omega-shallow confluent up to }\delta
}
(by our induction hypothesis),
due to our assumed \math\omega-shallow noisy  anti-closedness
(matching the definition's \math{n_0} to our \math{n_1\tight+1} 
                  and its \math{n_1} to \math{n_0\tight+1})
we have
\bigmath{
  w_1/\bar p_0
  \tightequal
  \repl{l_{0}\mu_{0}}{\bar p_1'}{r_{1}\mu_{1}}
  \tightequal
  \repl{l_{0}}{\bar p_1'}{r_{1}\xi}\sigma\varphi
  \refltransindex{\omega+n_0+1}
  \circ
  \antirefltransindex{\omega+n_1}
  \tight\circ
  \antionlyonceindex{\omega+n_1+1}
  \tight\circ
  \antirefltransindex{\omega}
  r_{0}\sigma\varphi
  \tightequal
  r_{0}\mu_{0}
  \tightequal
  w_0/\bar p_0.
}
\\
\Qeddouble{``The critical peak case''}%
\QEDtriple{``There is some \math{\bar p_1'} with
\bigmath{\bar p_0\bar p_1'\tightequal\bar p_1} and
\bigmath{\bar p_1'\tightnotequal\emptyset}''}

\pagebreak

\yestop
\yestop
\noindent
\underline{\underline{\underline{%
There is some \math{\bar p_0'} with
\bigmath{\bar p_1\bar p_0'\tightequal\bar p_0}:%
}}}

\noindent
We have two cases:

\noindent
\underline{\underline{%
``The second variable overlap case'':%
}}
\\\LINEnomath{\underline{\underline{%
There are \math{x\tightin\V} and \math{p'}, \math{p''} such that
\math{
    l_{1}/p'
    \tightequal
    x
  \und
    p'p''
    \tightequal
    \bar p_0'
}:%
}}}
\begin{diagram}
l_{1}\mu_{1}&&&\rredindex{\omega+n_1+1,\,\emptyset}&&&w_1/\bar p_1
\\&&&&&&\dequal
\\\dredindex{\omega+n_0+1,\,\bar p_0'}&&&&&&r_{1}\mu_{1}
\\&&&&&&\dredparaindex{\omega+n_0+1}
\\w_0/\bar p_1&\requal&l_{1}\nu&&\rredindex{\omega+n_1+1}&&r_{1}\nu
\end{diagram}
\noindent
\underline{Claim~11a:}
We have 
\bigmath{x\mu_{1}/p''\tightequal l_{0}\mu_{0}}
and may assume
\bigmath{x\tightin\Vsig.}
\\\underline{Proof of Claim~11a:}
We have 
\bigmath{
  x\mu_{1}/p''
  \tightequal
  l_{1}\mu_{1}/p' p''
  \tightequal
  u/\bar p_1 p' p''
  \tightequal
  u/\bar p_1\bar p_0'
  \tightequal
  u/\bar p_0
  \tightequal
  l_{0}\mu_{0}
.}
\\
If \bigmath{x\tightin\Vcons,}
then \bigmath{x\mu_{1}\tightin\tcc,}
then \bigmath{x\mu_{1}/p''\tightin\tcc,}
then 
\\
\bigmath{l_{0}\mu_{0}\tightin\tcc,}
and then \bigmath{l_{0}\tightin\tcs}
which we may assume not to be the case by Claim~5.%
\QED{Claim~11a}

\noindent
\underline{Claim~11b:}
We can define \math{\nu\in\Xsubst} by
\bigmath{
  x\nu
  \tightequal
  \repl
    {x\mu_{1}}
    {p''}
    {r_{0}\mu_{0}}
}
and
\bigmath{
  \forall y\tightin\V\tightsetminus\{x\}\stopq y\nu\tightequal y\mu_{1}
.}
Then we have 
\bigmath{
  x\mu_{1}\redindex{\omega+n_0+1}x\nu
.}
\\
\underline{Proof of Claim~11b:}
This follows directly from Claim~11a.%
\QED{Claim~11b}

\noindent
\underline{Claim~12:}
\bigmath{
  w_0/\bar p_1
  \tightequal
  l_{1}\nu
.}
\\
\underline{Proof of Claim~12:}
\\
By the left-linearity assumption of our lemma and claims 5 and 11a 
we may 
assume
\bigmath{
  \setwith
    {p'''}
    {l_{1}/p'''\tightequal x}
  =
  \{p'\}
.}
Thus, by Claim~11b we get
\bigmath{
  w_0/\bar p_1
  \tightequal
  \repl 
    {u/\bar p_1}
    {\bar p_0'}
    {r_{0}\mu_{0}}
  \tightequal\\
  \repl
    {\replpar
       {l_{1}}
       {p'''}
       {y\mu_{1}}
       {l_{1}/p'''\tightequal y\tightin\V}
    }
    {\bar p_0'}
    {r_{0}\mu_{0}}
  \tightequal\\
  \repl
    {\repl
       {\replpar
          {l_{1}}
          {p'''}
          {y\mu_{1}}
          {l_{1}/p'''\tightequal y\tightin\V\und y\tightnotequal x}
       }
       {p'}
       {x\mu_{1}}
    }
    {p' p''}
    {r_{0}\mu_{0}}
  \tightequal\\
  \repl
    {\replpar
       {l_{1}}
       {p'''}
       {y\nu}
       {l_{1}/p'''\tightequal y\tightin\V\und y\tightnotequal x}
    }
    {p'}
    {\repl
       {x\mu_{1}}
       {p''}
       {r_{0}\mu_{0}}
    }
  \tightequal\\
  \replpar
    {l_{1}}
    {p'''}
    {y\nu}
    {l_{1}/p'''\tightequal y\tightin\V}
  \tightequal
  l_{1}\nu
.}\QED{Claim~12}

\noindent
\underline{Claim~13:}
\bigmath{
  r_{1}\nu
  \antiredparaindex{\omega+n_0+1}
  w_1/\bar p_1
.}
\\
\underline{Proof of Claim~13:} 
Since \bigmath{r_{1}\mu_{1}\tightequal w_1/\bar p_1,} 
this follows directly from Claim~11b.%
\QED{Claim~13}

\noindent
By claims 12 and 13 
using Corollary~\ref{corollary parallel one} 
it now suffices to show
\bigmath{
  l_{1}\nu
  \redindex{\omega+n_1+1}
  r_{1}\nu
,}
which again follows from 
Claim~11b,
Lemma~\ref{lemma invariance of fulfilledness}
(matching 
 its \math{n_0} to our \math{n_0\tight+1} and 
 its \math{n_1} to our \math{n_1}),
and our induction hypothesis that \RX\ is \math\omega-shallow confluent up to
\bigmath{
  (n_0\tight+1)\plusomega n_1
.}%
\\\Qeddouble{``The second variable overlap case''}

\pagebreak

\yestop
\noindent
\underline{\underline{``The second critical peak case'':
\math{
    \bar p_0'\tightin\TPOS{l_{1}}
  \und
    l_{1}/\bar p_0'\tightnotin\V
}:%
}}
\begin{diagram}
l_{1}\mu_{1}
&&&\rredindex{\omega+n_1+1,\,\emptyset}&&&w_1/\bar p_1
\\
\dredindex{\omega+n_0+1,\,\bar p_0'}
&&
&&
&&\drefltransindex{\omega+n_0+1}
\\
w_0/\bar p_1
&\rrefltransindex\omega&\circ
&\ronlyonceindex{\omega+n_1+1}&\circ
&\rrefltransindex{\omega+n_1}&\circ
\end{diagram}
Let \math{\xi\in\SUBST\V\V} be a bijection with 
\bigmath{
  \xi[\VAR{\kurzregelindex{0}}]\cap\VAR{\kurzregelindex{1}}
  =
  \emptyset
.}
\\
Define
\bigmath{
  \Y
  :=
  \xi[\VAR{\kurzregelindex{0}}]\cup\VAR{\kurzregelindex{1}}
.}
\\
Let \math{\varrho\in\Xsubst} be given by
$\ x\varrho=
\left\{\begin{array}{@{}l@{}l@{}}
  x\mu_{1}        &\mbox{ if }x\in\VAR{\kurzregelindex{1}}\\
  x\xi^{-1}\mu_{0}&\mbox{ else}\\
\end{array}\right\}
\:(x\tightin\V)$.
\\
By
\math{
  l_{0}\xi\varrho
  \tightequal 
  l_{0}\xi\xi^{-1}\mu_{0}
  \tightequal 
  u/\bar p_0
  \tightequal 
  u/\bar p_1\bar p_0'
  \tightequal 
  l_{1}\mu_{1}/\bar p_0'  
  \tightequal 
  l_{1}\varrho/\bar p_0'
  \tightequal 
  (l_{1}/\bar p_0')\varrho
}
\\
let
\math{
  \sigma:=\minmgu{\{(l_{0}\xi}{l_{1}/\bar p_0')\},\Y}
}
and
\math{\varphi\in\Xsubst}
with
\math{
  \domres{\inpit{\sigma\varphi}}\Y
  \tightequal
  \domres\varrho\Y
}.
\\
If 
\bigmath{
  \repl{l_{1}}{\bar p_0'}{r_{0}\xi}\sigma
  \tightequal
  r_{1}\sigma
,}
then the proof is finished due to 
\\\linemath{
  w_0/\bar p_1
  \tightequal
  \repl{l_{1}\mu_{1}}{\bar p_0'}{r_{0}\mu_{0}}
  \tightequal
  \repl{l_{1}}{\bar p_0'}{r_{0}\xi}\sigma\varphi
  \tightequal
  r_{1}\sigma\varphi
  \tightequal
  r_{1}\mu_{1}
  \tightequal
  w_1/\bar p_1.
}
Otherwise 
we have
\bigmath{
  (\,
   (\repl{l_{1}}{\bar p_0'}{r_{0}\xi},
    C_{0}\xi,    
    1),\penalty-1\,
   (r_{1},
    C_{1},
    1),\penalty-1\,
    l_{1},\penalty-1\,
    \sigma,\penalty-1\,
    \bar p_0'\,)
  \in{\rm CP}(\R)
}
(due to Claim~5);
\bigmath{C_{0}\xi\sigma\varphi=C_{0}\mu_{0}}
is fulfilled \wrt\ \redindex{\omega+n_0};
\bigmath{C_{1}\sigma\varphi=C_{1}\mu_{1}}
is fulfilled \wrt\ \redindex{\omega+n_1}.
Since 
\bigmath{
  \forall\delta\tightprec(n_0\tight+1)\plusomega(n_1\tight+1)\stopq
  \mbox{\RX\ is \math\omega-shallow confluent up to }\delta
}
(by our induction hypothesis) 
due to our assumed \math\omega-shallow 
noisy  strong joinability 
(matching the definition's \math{n_0} to our \math{n_0\tight+1}
                   and its \math{n_1} to our \math{n_1\tight+1})
we have
\bigmath{
  w_0/\bar p_1
  \tightequal
  \repl{l_{1}\mu_{1}}{\bar p_0'}{r_{0}\mu_{0}}
  \tightequal
  \repl{l_{1}}{\bar p_0'}{r_{0}\xi}\sigma\varphi
  \refltransindex{\omega}
  \tight\circ
  \onlyonceindex{\omega+n_1+1}
  \tight\circ
  \refltransindex{\omega+n_1}
  \circ
  \antirefltransindex{\omega+n_0+1}
  \penalty-1
  r_{1}\sigma\varphi
  \tightequal
  r_{1}\mu_{1}
  \tightequal
  w_1/\bar p_1
.}
\\\Qeddouble{``The second critical peak case''}
\end{proofparsepqed}

\begin{proofqed}{Lemma~\ref{lemma invariance of fulfilledness level}}
For each literal \math L in \math C we have to show that
\math{L\nu} is fulfilled \wrt\ \redindex{\RX,\omega+n_1}.
Note that we already know that 
\math{L\mu} is fulfilled \wrt\ \redindex{\RX,\omega+n_1}.
If \bigmath{\VAR C\tightsubseteq\Vcons,}
then for all \math x in \VAR C we have \bigmath{x\mu\tightin\tcc}
and then by \lemmaconskeeping\ \bigmath{x\mu\refltransindex{\RX,\omega+0}y\mu.}
Thus, by the disjunctive assumption of our lemma we may assume
\bigmath{n_0\tightpreceq n_1.}
\\
\underline{\math{L=(s_0\boldequal s_1)}:}
We have
\bigmath{
  s_0\nu
  \antirefltransindex{\RX,\omega+n_0}
  s_0\mu
  \refltransindex    {\RX,\omega+n_1}\penalty-1
  t_0
  \antirefltransindex{\RX,\omega+n_1}\penalty-1
  s_1\mu
  \refltransindex    {\RX,\omega+n_0}
  s_1\nu
}
for some \math{t_0.}
By our \math\omega-level confluence up to \math{n_1} and
\bigmath{n_0\tightpreceq n_1,}
we get some \math v with
\bigmath{
  s_0\nu
  \refltransindex    {\RX,\omega+n_1}
  v
  \antirefltransindex{\RX,\omega+n_1}
  t_0
} 
and then 
(due to
\bigmath{
  v
  \antirefltransindex{\RX,\omega+n_1}
  s_1\mu
  \refltransindex    {\RX,\omega+n_0}
  s_1\nu
)}
\bigmath{
  v
  \refltransindex    {\RX,\omega+n_1}
  \circ
  \antirefltransindex{\RX,\omega+n_1}
  s_1\nu
.}
\\\underline{$L=(\DEF s)$:}
We know the existence of
\math{t\in\tgcons}
with
\bigmath{
  s\nu
  \antirefltransindex{\RX,\omega+n_0}
  s\mu
  \refltransindex    {\RX,\omega+n_1}
  t
.}
By our \math\omega-level confluence up to \math{n_1} and
\bigmath{n_0\tightpreceq n_1,}
there is some
\math{t'} with
\bigmath{
  s\nu
  \refltransindex    {\RX,\omega+n_1}
  t'
  \antirefltransindex{\RX,\omega+n_1}
  t
.}
By \lemmaconskeeping\ we get \bigmath{t'\tightin\tgcons.}
\\
\underline{$L=(s_0\boldunequal s_1)$:}
There exist some \math{t_0,t_1\in\tgcons}
with
\bigmath{
  \forall i\tightprec2\stopq
    s_i\nu
    \antirefltransindex{\RX,\omega+n_0}
    s_i\mu
    \refltransindex{\RX,\omega+n_1}
    t_i
}
and 
\bigmath{
  t_0
  \notconfluindex{\RX,\omega+n_1}  
  t_1
.}
Just like above
we get \math{t_0',\ t_1'\in\tgcons} with
\bigmath{
  \forall i\tightprec2\stopq
    s_i\nu
    \refltransindex    {\RX,\omega+n_1}
    \penalty-1
    t_i'
    \penalty-1
    \antirefltransindex{\RX,\omega+n_1}
    t_i
.}
Finally 
\bigmath{
  t_0'
  \antirefltransindex{\RX,\omega+n_1}
  t_0
  \notconfluindex{\RX,\omega+n_1}  
  t_1
  \refltransindex{\RX,\omega+n_1}
  t_1'
}
implies
\bigmath{
  t_0'
  \notconfluindex{\RX,\omega+n_1}  
  t_1'
.}
\end{proofqed}

\pagebreak

\begin{proofparsepqed}
{Lemma~\ref{lemma level parallel closed second level two}}
\underline{Claim~0:}
\RX\ is \math\omega-shallow confluent up to \math\omega.
\\\underline{Proof of Claim~0:}
Directly by the assumed strong commutation,
\cf\ the proofs of the claims 2 and 3 of the proof of 
Lemma~\ref{lemma parallel closed first level two}.\QED{Claim~0}

\yestop
\noindent
\underline{Claim~1:}
If 
\bigmath{
    \refltransindex{\omega}
    \tight\circ
    \redparaindex{\omega+n}
    \tight\circ
    \refltransindex{\omega}
}
strongly commutes over
\refltransindex{\omega+n},
then
\redindex{\omega+n} and \redindex{\omega+n} are commuting.
\\
\underline{Proof of Claim~1:}
\bigmath{
    \refltransindex{\omega}
    \tight\circ
    \redparaindex{\omega+n}
    \tight\circ
    \refltransindex{\omega}
}
and
\refltransindex{\omega+n}
are commuting
by Lemma~\ref{lemma strong commutation one copy}.
Since by Corollary~\ref{corollary parallel one}
and \lemmamonotonicinbeta\
we have
\bigmath{
  \redindex{\omega+n}
  \subseteq
  \refltransindex{\omega}
  \tight\circ
  \redparaindex{\omega+n}
  \tight\circ
  \refltransindex{\omega}
  \subseteq
  \refltransindex{\omega+n}
,}
now \redindex{\omega+n} 
and \redindex{\omega+n} are commuting, too.\QED{Claim~1}

\yestop
\yestop
\noindent
For \math{n\prec\omega} 
we are going to show by induction on \math n 
the following property\footroom:
\\\LINEmath{
    w_0
    \antiredparaindex{\omega+n}
    u
    \redparaindex{\omega+n}
    w_1
  \quad\implies\quad
    w_0
    \refltransindex{\omega}
    \tight\circ
    \redparaindex{\omega+n}
    \tight\circ
    \refltransindex{\omega}
    \circ
    \antirefltransindex{\omega+n}
    w_1
.}
\begin{diagram}
u&&&\rredparaindex{\omega+n}&&&w_1
\\\dredparaindex{\omega+n}&&&&&&\drefltransindex{\omega+n}
\\w_0&\rrefltransindex{\omega}&\circ&\rredparaindex{\omega+n}
&\circ&\rrefltransindex{\omega}&\circ
\end{diagram}

\noindent
\underline{Claim~2:}
Let \math{\delta\prec\omega}.
If
\bigmath{
  \forall 
        n\tightpreceq\delta
  \stopq
    \forall w_0,w_1,u\stopq
    \inparenthesesoplist{
        w_0
        \antiredparaindex{\omega+n}
        u
        \redparaindex    {\omega+n}
        w_1
      \oplistimplies 
        w_0
        \refltransindex{\omega}
        \tight\circ
        \redparaindex{\omega+n}
        \tight\circ
        \refltransindex{\omega}
        \circ
        \antirefltransindex{\omega+n}
        w_1
    }
,}
then
\bigmath{
  \forall n\tightpreceq\delta\stopq
  \inparentheses{
            \refltransindex{\omega}
            \tight\circ
            \redparaindex{\omega+n}
            \tight\circ
            \refltransindex{\omega}
      \mbox{ strongly commutes over }
            \refltransindex{\omega+n}
  }
,}
and 
\RX\ is \math\omega-level confluent up to \math{\delta}.
\\
\underline{Proof of Claim~2:}
First we show the strong commutation.
Assume \math{n\tightpreceq\delta}. 
By Lemma~\ref{lemma strong commutation one copy} it suffices to show that
\bigmath{
    \refltransindex{\omega}
    \tight\circ
    \redparaindex{\omega+n}
    \tight\circ
    \refltransindex{\omega}
}
strongly commutes over
\redindex{\omega+n}.
Assume
\bigmath{
    u''
    \antiredindex{\omega+n}
    u'
    \refltransindex{\omega}
    u
    \redparaindex{\omega+n}
    w_1
    \refltransindex{\omega}
    w_2
}
(\cf\ diagram below).
By the strong commutation assumed for our lemma
and Corollary~\ref{corollary parallel one},
there are \math{w_0} and \math{w_0'} with
\bigmath{
    u''
    \refltransindex{\omega}
    w_0'
    \antirefltransindex{\omega}
    w_0
    \antiredparaindex{\omega+n}
    u
.}
By the above property there are some \math{w_3}, \math{w_1'}
with
\bigmath{
    w_0
    \refltransindex{\omega}
    w_3
    \redparaindex{\omega+n}
    \tight\circ
    \refltransindex{\omega}
    w_1'
    \antirefltransindex{\omega+n}
    w_1
.}
By Claim~0 we can close the peak
\bigmath{ 
    w_1'
    \antirefltransindex{\omega+n}
    w_1
    \refltransindex{\omega}
    w_2
}
according to 
\bigmath{
    w_1'
    \refltransindex{\omega}
    w_2'
    \antirefltransindex{\omega+n}
    w_2
}
for some \math{w_2'}.
By Claim~0 again,
we can close the peak
\bigmath{
  w_0'
  \antirefltransindex{\omega}
  w_0
  \refltransindex{\omega}
  w_3
}
according to 
\bigmath{
  w_0'
  \refltransindex{\omega}
  w_3'
  \antirefltransindex{\omega}
  w_3
}
for some \math{w_3'}.
To close the whole diagram, we only have to show that we can close the peak
\bigmath{
    w_3'
    \antirefltransindex{\omega}
    w_3
    \redparaindex{\omega+n}
    \tight\circ
    \refltransindex{\omega}
    w_2'
}
according to 
\bigmath{
    w_3'
    \redparaindex{\omega+n}
    \tight\circ
    \refltransindex{\omega}
    \circ
    \antirefltransindex{\omega}
    w_2'
,}
which is possible due to the strong commutation assumed for our lemma.
\begin{diagram}
u'&\rrefltransindex{\omega}&u
&&
&\rredparaindex{\omega+n}&&
&w_1&\rrefltransindex{\omega}&w_2
\\&&\dredparaindex{\omega+n}
&&
&&&
&\drefltransindex{\omega+n}&&\drefltransindex{\omega+n}
\\\dredindex{\omega+n}&&w_0
&\rrefltransindex{\omega}&w_3
&\rredparaindex{\omega+n}&\circ&\rrefltransindex{\omega}
&w_1'&\rrefltransindex{\omega}&w_2'
\\&&\drefltransindex{\omega}
&&\drefltransindex{\omega}
&&&
&&&\drefltransindex{\omega}
\\u''&\rrefltransindex{\omega}&w_0'
&\rrefltransindex{\omega}&w_3'
&\rredparaindex{\omega+n}
&\circ&&\rrefltransindex{\omega}&&\circ
\end{diagram}

\pagebreak

\noindent
Finally we show \math\omega-level confluence up to \math\delta.
Assume
\math{n_0,n_1\prec\omega} with
\bigmath{\maxoftwo{n_0}{n_1}\tightpreceq\delta}
and
\bigmath{
    w_0
    \antirefltransindex{\omega+n_0}
    u
    \refltransindex    {\omega+n_1}
    w_1
.}
By \lemmamonotonicinbeta\ we get
\bigmath{
    w_0
    \antirefltransindex{\omega+\maxoftwo{n_0}{n_1}}
    u
    \refltransindex    {\omega+\maxoftwo{n_0}{n_1}}
    w_1
.}
Since \bigmath{\maxoftwo{n_0}{n_1}\tightpreceq\delta,}
above we have shown that 
\bigmath{
    \refltransindex{\omega}
    \tight\circ
    \redparaindex{\omega+\maxoftwo{n_0}{n_1}}
    \tight\circ
    \refltransindex{\omega}
}
strongly commutes over
\refltransindex{\omega+\maxoftwo{n_0}{n_1}}. 
By Claim~1 we finally get
\bigmath{
  w_0
  \refltransindex{\omega+\maxoftwo{n_0}{n_1}}
  \circ
  \antirefltransindex{\omega+\maxoftwo{n_0}{n_1}}
  w_1
}
as desired.%
\QED{Claim~2}

\yestop
\yestop
\yestop
\noindent
Note that 
for \bigmath{n\tightequal0} 
our property follows 
from 
\bigmath{\antiredparaindex\omega\subseteq\antirefltransindex\omega}
(by Corollary~\ref{corollary parallel one})
and Claim~0.

The benefit of 
Claim~2 is twofold: First, it says that our lemma is valid
if the above property
holds for all \math{n\prec\omega}.
Second, it strengthens the property when used as induction hypothesis. Thus 
(writing \math{n\tight+1} instead of \math{n} since we may assume
 \math{0\tightprec n})
it
now suffices to show
for
\math{n\prec\omega}
that 
\\\linemath{
    w_0
    \antiredparaindex{\omega+n+1,\Pi_0}
    u
    \redparaindex    {\omega+n+1,\Pi_1}
    w_1
}
together with our induction hypotheses 
that\headroom
\\\linemath{\headroom
  \mbox{\RX\ is \math\omega-level confluent up to }n
}
\headroom
implies
\\\LINEmath{
    w_0
    \refltransindex{\omega}
    \tight\circ
    \redparaindex{\omega+n+1}
    \tight\circ
    \refltransindex{\omega}
    \circ
    \antirefltransindex{\omega+n_1+1}
    w_1
.}
\begin{diagram}
u&&&\rredparaindex{\omega+n+1,\,\Pi_1}
&&&w_1
\\\dredparaindex{\omega+n+1,\,\Pi_0}&&&
&&&\drefltransindex{\omega+n+1}
\\w_0&\rrefltransindex{\omega}&\circ&\rredparaindex{\omega+n+1}
&\circ&\rrefltransindex{\omega}&\circ
\end{diagram}

\yestop
\noindent
\Wrog\ let the positions of \math{\Pi_i} be maximal
in the sense that for any \math{p\in\Pi_i} 
and \math{\Xi\subseteq \TPOS u\tightcap(p\N^+)}
we do not have 
\bigmath{
  u
  \redparaindex{\omega+n+1,(\Pi_i\setminus\{p\})\cup\Xi}
  w_i
}
anymore.
Then for each \math{i\prec2} and
\math{p\in\Pi_i} there are
\bigmath{\kurzregelindex{i,p}\in\R}
and
\bigmath{\mu_{i,p}\in\Xsubst}
with
\bigmath{u/p\tightequal l_{i,p}\mu_{i,p},}
\bigmath{r_{i,p}\mu_{i,p}\tightequal w_i/p,}
\math{C_{i,p}\mu_{i,p}} fulfilled \wrt\ \redindex{\omega+n}.
Finally, for each \math{i\prec2}:
\bigmath{
  w_i\tightequal\replpar{u}{p}{r_{i,p}\mu_{i,p}}{p\tightin\Pi_i}
.}

\pagebreak

\yestop
\noindent
\underline{Claim~5:}
We may assume 
\bigmath{
  \forall i\tightprec2\stopq
  \forall p\tightin\Pi_i\stopq
    l_{i,p}\tightnotin\tcs
.}
\\
\underline{Proof of Claim~5:}
Define \math{\Xi_i:=\setwith{p\tightin\Pi_i}{l_{i,p}\tightin\tcs}}
and 
\math{
  u_i':=\replpar{u}{p}{r_{i,p}\mu_{i,p}}{p\tightin\Pi_i\tightsetminus\Xi_i}
}.
If we have succeeded with our proof under the assumption of Claim~5,
then we have shown 
\bigmath{
  u_0'
  \refltransindex{\omega}
  v_0
  \redparaindex{\omega+n+1}
  \tight\circ
  \refltransindex{\omega}
  v_1
  \antirefltransindex{\omega+n+1}
  u_1'
}
for some \math{v_0}, \math{v_1}
(\cf\ diagram below).
By Lemma~\ref{lemma invariance of fulfilledness two} 
(matching both its \math\mu\ and \math\nu\ to our \math{\mu_{i,p}})
we get
\bigmath{
  \forall i\tightprec2\stopq
  \forall p\tightin\Xi_i\stopq
    l_{i,p}\mu_{i,p}\redindex\omega r_{i,p}\mu_{i,p}
}
and therefore
\bigmath{
  \forall i\tightprec2\stopq
  u_i'
  \redparaindex{\omega,\Xi_i}
  w_i
.}
Thus from
\bigmath{
  v_1
  \penalty-1
  \antirefltransindex{\omega+n_0+1}
  \penalty-1
  u_1'
  \penalty-1
  \refltransindex\omega
  \penalty-1
  w_1
}
we get 
\bigmath{
  v_1
  \refltransindex\omega
  v_2
  \antirefltransindex{\omega+n_0+1}
  w_1
}
for some \math{v_2}
by \math\omega-shallow confluence up to \math{\omega} (\cf\ Claim~0).
For the same reason 
we can close the peak 
\bigmath{
  w_0
  \antirefltransindex{\omega}
  u_0'
  \refltransindex{\omega}
  v_0
}
according to 
\bigmath{
  w_0
  \refltransindex{\omega}
  v_0'
  \antirefltransindex{\omega}
  v_0
}
for some \math{v_0'}.
By the assumption of our lemma that
\bigmath{
  \redparaindex{\RX,\omega+n_1+1}
  \tight\circ
  \refltransindex{\RX,\omega}
}
strongly commutes over \refltransindex{\omega},
from
\bigmath{
  v_0'
  \antirefltransindex\omega
  v_0
  \redparaindex{\omega+n_1+1}
  \tight\circ
  \refltransindex{\omega}
  v_1
  \refltransindex{\omega}
  v_2
}
we can finally conclude
\bigmath{
  v_0'
  \redparaindex{\omega+n_1+1}
  \tight\circ
  \refltransindex{\omega}
  \circ
  \penalty-1
  \antirefltransindex{\omega}
  v_2
%  \antirefltransindex{\omega+n_0+1}
%  w_1
.}
\begin{diagram}
u&&&\rredparaindex{\omega+n+1,\,\Pi_1\setminus\Xi_1}
&&&u_1'&\rredparaindex{\omega,\,\Xi_1}&w_1
\\
\dredparaindex{\omega+n+1,\,\Pi_0\setminus\Xi_0}&&&
&&&\drefltransindex{\omega+n+1}&&\drefltransindex{\omega+n+1}
\\
u_0'&\rrefltransindex{\omega}&v_0&\rredparaindex{\omega+n+1}
&\circ&\rrefltransindex{\omega}&v_1&\rrefltransindex{\omega}&v_2
\\
\dredparaindex{\omega,\,\Xi_0}&&\drefltransindex{\omega}&
&&&&&\drefltransindex{\omega}
\\
w_0&\rrefltransindex{\omega}&v_0'&\rredparaindex{\omega+n+1}
&\circ&&\rrefltransindex{\omega}&&\circ
\end{diagram}
\Qed{Claim~5}

\yestop
\noindent
Define
the set of inner overlapping positions by
\\\linemath{
  \displaystyle
  \Omega(\Pi_0,\Pi_1)
  :=
  \bigcup_{i\prec2}
    \setwith
      {p\tightin\Pi_{1-i}}
      {\exists q\tightin\Pi_i\stopq\exists q'\tightin\N^\ast\stopq
        p\tightequal q q'
      }
,}
and the length of a term by
\bigmath{\lambda(\anonymousfpp{t_0}{t_{m-1}}):=1+\sum_{j\prec m}\lambda(t_j).}

\yestop
\noindent
Now we start a second level of induction on
\bigmath{  \displaystyle
  \sum_{p'\in\Omega(\Pi_0,\Pi_1)}\lambda(u/p')
}
in \tightprec.

\yestop
\noindent
Define the set of top positions by
\\\linemath{
  \displaystyle
  \Theta
  :=
      \setwith
      {p\tightin\Pi_0\tightcup\Pi_1}
      {\neg\exists q\tightin\Pi_0\tightcup\Pi_1\stopq
           \exists q'\tightin\N^+\stopq
             p\tightequal q q'
      }
.}
Since the prefix ordering is wellfounded we have
\bigmath{
  \forall i\tightprec2\stopq
  \forall p\tightin\Pi_i\stopq
  \exists q\tightin\Theta\stopq
  \exists q'\tightin\N^\ast\stopq
    p\tightequal q q'
.}
Then
\bigmath{
  \forall i\tightprec2\stopq
  w_i
  \tightequal
  \replpar{w_i}{q}{w_i/q}{q\tightin\Theta}
  \tightequal
  \replpar{\replpar{u}{p}{r_{i,p}\mu_{i,p}}{p\tightin\Pi_i}}
          {q}{w_i/q}{q\tightin\Theta}
  \tightequal
  \replpar{u}{q}{w_i/q}{q\tightin\Theta}
.}
Thus, it now suffices to show for all \math{q\in\Theta}
\\\linemath{\headroom\footroom
    w_0/q
    \refltransindex{\omega}
    \tight\circ
    \redparaindex{\omega+n+1}
    \tight\circ
    \refltransindex{\omega}
    \circ
    \antirefltransindex{\omega+n+1}
    w_1/q
}
because then we have 
\\\LINEmath{
  w_0
  \tightequal
  \replpar{u}{q}{w_0/q}{q\tightin\Theta}
    \refltransindex{\omega}
    \tight\circ
    \redparaindex{\omega+n+1}
    \tight\circ
    \refltransindex{\omega}
    \circ
    \antirefltransindex{\omega+n+1}
  \replpar{u}{q}{w_1/q}{q\tightin\Theta}
  \tightequal
  w_1
.}

\noindent
Therefore we are left with the following two cases for \math{q\in\Theta}:

\pagebreak

\yestop
\yestop
\noindent
\underline{\underline{\underline{\math{q\tightnotin\Pi_1}:}}}
Then \bigmath{q\tightin\Pi_0.}
Define \math{\Pi_1':=\setwith{p}{q p\tightin\Pi_1}}.
\noindent
We have two cases:

\yestop
\noindent
\underline{\underline{``The variable overlap (if any) case'':
\math{
  \forall p\tightin\Pi_1'\tightcap\TPOS{l_{0,q}}\stopq
    l_{0,q}/p\tightin\V
}:}}
\begin{diagram}
l_{0,q}\mu_{0,q}&&\rredparaindex{\omega+n+1,\,\Pi_1'}&&&&w_1/q
\\&&&&&&\dequal
\\\dredindex{\omega+n+1,\,\emptyset}&&&&&&l_{0,q}\nu
\\&&&&&&\dredindex{\omega+n+1}
\\w_0/q&\requal&r_{0,q}\mu_{0,q}&&\rredparaindex{\omega+n+1}&&r_{0,q}\nu
\end{diagram}
\noindent
Define a function \math\Gamma\ on \V\ by (\math{x\tightin\V}):
\bigmath{
  \Gamma(x):=
  \setwith{(p',p'')}
          {l_{0,q}/p'\tightequal x\ \wedge\ p' p''\in\Pi_1'}
.}

\noindent
\underline{Claim~7:}
There is some \math{\nu\in\Xsubst} with
\\\LINEmath{
  \forall x\in\V\stopq
    \inparenthesesoplist{
       x\mu_{0,q}
       \redparaindex{\omega+n+1}
       x\nu
    \oplistund
       \forall p'\tightin\DOM{\Gamma(x)}\stopq
         x\nu
         \tightequal
         \replpar
           {x\mu_{0,q}}
           {p''}
           {r_{1,q p' p''}\mu_{1,q p' p''}}
           {(p',p'')\tightin\Gamma(x)}
    }
.}
\\
\underline{Proof of Claim~7:}
\\
In case of \bigmath{\DOM{\Gamma(x)}\tightequal\emptyset} we define
\bigmath{x\nu:=x\mu_{0,q}.}
If there is some \math{p'} such that 
\bigmath{\DOM{\Gamma(x)}\tightequal\{p'\}}
we define 
\bigmath{
  x\nu
  :=
  \replpar
    {x\mu_{0,q}}
    {p''}
    {r_{1,q p' p''}\mu_{1,q p' p''}}
    {(p',p'')\tightin\Gamma(x)}
.}
This is appropriate since due to 
\bigmath{
  \forall(p',p'')\tightin\Gamma(x)\stopq
    x\mu_{0,q}/p''
    \tightequal 
    l_{0,q}\mu_{0,q}/p' p''
    \tightequal 
    u/q p' p''
    \tightequal 
    l_{1,q p' p''}\mu_{1,q p' p''}
}
we have
\\\LINEmath{
  \begin{array}{l@{}l@{}l}
  x\mu_{0,q}&
  \tightequal&
  \replpar
    {x\mu_{0,q}}
    {p''}
    {l_{1,q p' p''}\mu_{1,q p' p''}}
    {(p',p'')\tightin\Gamma(x)}
  \redparaindex{\omega+n+1}\\&&
  \replpar
     {x\mu_{0,q}}
     {p''}
     {r_{1,q p' p''}\mu_{1,q p' p''}}
     {(p',p'')\tightin\Gamma(x)}
  \tightequal
  x\nu.  
  \end{array}
}
\\
Finally, in case of \bigmath{\CARD{\DOM{\Gamma(x)}}\succ1,} \math{l_{0,q}} is
not linear in \math x. By the conditions of our lemma and Claim~5 this implies
\bigmath{x\tightin\Vcons.}
Since there is some \math{(p',p'')\in\Gamma(x)} with
\bigmath{
    x\mu_{0,q}/p''
    \tightequal 
    l_{1,q p' p''}\mu_{1,q p' p''}
}
this implies 
\bigmath{
   l_{1,q p' p''}\mu_{1,q p' p''}\tightin\tcc
}
and then 
\bigmath{
   l_{1,q p' p''}\tightin\tcs
}
which contradicts Claim~5.%
\QED{Claim~7}

\noindent
\underline{Claim~8:}
\bigmath{
  l_{0,q}\nu
  \tightequal 
  w_1/q
.}
\\
\underline{Proof of Claim~8:}
\\
By Claim~7 we get
\bigmath{
  w_1/q
  \tightequal
  \replpar 
    {u/q}
    {p' p''}
    {r_{1,q p' p''}\mu_{1,q p' p''}}
    {\exists x\tightin\V\stopq(p',p'')\tightin\Gamma(x)}
  \tightequal\\
  \replpar
    {\replpar
       {l_{0,q}}
       {p'}
       {x\mu_{0,q}}
       {l_{0,q}/p'\tightequal x\tightin\V}
    }
    {p' p''}
    {r_{1,q p' p''}\mu_{1,q p' p''}}
    {\exists x\tightin\V\stopq(p',p'')\tightin\Gamma(x)}
  \tightequal\\
  \replpar
    {l_{0,q}}
    {p'}
    {\replpar
       {x\mu_{0,q}}
       {p''}
       {r_{1,q p' p''}\mu_{1,q p' p''}}
       {(p',p'')\tightin\Gamma(x)}}
    {l_{0,q}/p'\tightequal x\tightin\V}
  \tightequal\\
  \replpar
    {l_{0,q}}
    {p'}
    {x\nu}
    {l_{0,q}/p'\tightequal x\tightin\V}
  \tightequal
  l_{0,q}\nu
.}\QED{Claim~8}

\noindent
\underline{Claim~9:}
\bigmath{
  w_0/q
  \redparaindex{\omega+n+1}
  r_{0,q}\nu
.}
\\
\underline{Proof of Claim~9:} 
Since 
\bigmath{
  w_0/q
  \tightequal
  r_{0,q}\mu_{0,q}
,} 
this follows directly from Claim~7.%
\QED{Claim~9}

\noindent
By claims 8 and 9 it now suffices to show
\bigmath{
  l_{0,q}\nu
  \redindex{\omega+n+1}
  r_{0,q}\nu
,}
which again follows from 
Lemma~\ref{lemma invariance of fulfilledness level}
(matching its \math{n_0} to our \math{n\tight+1}
      and its \math{n_1} to our \math{n})
since \RX\ is \math\omega-level confluent up to
\math{n}
by our induction hypothesis 
and since
\bigmath{\forall x\tightin\V\stopq x\mu_{0,q}\refltransindex{\omega+n+1}x\nu}
by Claim~7 and Corollary~\ref{corollary parallel one}.%
\\\Qeddouble{``The variable overlap (if any) case''}

\pagebreak

\yestop
\noindent
\underline{\underline{``The critical peak case'':
There is some \math{p\in \Pi_1'\tightcap\TPOS{l_{0,q}}}
with \math{l_{0,q}/p\tightnotin\V}:}}
\begin{diagram}
l_{0,q}\mu_{0,q}&\rredindex{\omega+n+1,\,p}&u'
&&&\rredparaindex{\omega+n+1,\,\Pi_1'\setminus\{p\}}
&&&w_1/q
\\&&\dredparaindex{\omega+n+1}
&&&
&&&\drefltransindex{\omega+n+1}
\\\dredindex{\omega+n+1,\,\emptyset}&&v_1
&\rrefltransindex{\omega}&v_3&\rredparaindex{\omega+n+1}
&\circ&\rrefltransindex{\omega}&v_1'
\\&&\drefltransindex{\omega}
&&\drefltransindex{\omega}&
&&&\drefltransindex{\omega}
\\w_0/q&\rrefltransindex{\omega}&v_2
&\rrefltransindex{\omega}&v_4&\rredparaindex{\omega+n+1}
&\circ&\rrefltransindex{\omega}&\circ
\end{diagram}
\underline{Claim~10:}
\bigmath{p\tightnotequal\emptyset.}
\\
\underline{Proof of Claim~10:}
If \bigmath{p\tightequal\emptyset,} then
\bigmath{\emptyset\tightin\Pi_1',} then
\bigmath{q\tightin\Pi_1,} which contradicts our global case assumption.%
\QED{Claim~10}

\noindent
Let \math{\xi\in\SUBST\V\V} be a bijection with 
\bigmath{
  \xi[\VAR{\kurzregelindex{1,q p}}]\cap\VAR{\kurzregelindex{0,q}}
  =
  \emptyset
.}
\\
Define
\bigmath{
  \Y
  :=
  \xi[\VAR{\kurzregelindex{1,q p}}]\cup\VAR{\kurzregelindex{0,q}}
.}
\\
Let \math{\varrho\in\Xsubst} be given by
$\ x\varrho=
\left\{\begin{array}{@{}l@{}l@{}}
  x\mu_{0,q}        &\mbox{ if }x\in\VAR{\kurzregelindex{0,q}}\\
  x\xi^{-1}\mu_{1,q p}&\mbox{ else}\\
\end{array}\right\}
\:(x\tightin\V)$.
\\
By
\math{
  l_{1,q p}\xi\varrho
  \tightequal 
  l_{1,q p}\xi\xi^{-1}\mu_{1,q p}
  \tightequal 
  u/q p
  \tightequal 
  l_{0,q}\mu_{0,q}/p  
  \tightequal 
  l_{0,q}\varrho/p
  \tightequal 
  (l_{0,q}/p)\varrho
}
\\
let
\math{
  \sigma:=\minmgu{\{(l_{1,q p}\xi}{l_{0,q}/p)\},\Y}
}
and
\math{\varphi\in\Xsubst}
with
\math{
  \domres{\inpit{\sigma\varphi}}\Y
  \tightequal
  \domres\varrho\Y
}.
\\
Define 
\math{
  u':=  
  \repl{l_{0,q}\mu_{0,q}}
       {p}
       {r_{1,q p}\mu_{1,q p}}
}.
We get
\\\LINEmath{
  \arr{{l@{}l}
    u'\tightequal
    &
    \repl
      {\replpar
         {u/q}
         {p'}
         {l_{1,q p'}\mu_{1,q p'}}
         {p'\tightin\Pi_1'\tightsetminus\{p\}}}
      {p}
      {r_{1,q p}\mu_{1,q p}}
    \redparaindex{\omega+n+1,\Pi_1'\setminus\{p\}}
    \\&
    \replpar{u/q}{p'}{r_{1,q p'}\mu_{1,q p'}}{p'\tightin\Pi_1'}    
    \tightequal
    w_1/q
  .
  }
}
\\
If 
\bigmath{
  \repl{l_{0,q}}{p}{r_{1,q p}\xi}\sigma
  \tightequal
  r_{0,q}\sigma
,}
then the proof is finished due to 
\\\LINEmath{
  w_0/q
  \tightequal
  r_{0,q}\mu_{0,q}
  \tightequal
  r_{0,q}\sigma\varphi
  \tightequal
  \repl{l_{0,q}}{p}{r_{1,q p}\xi}\sigma\varphi
  \tightequal
  u'
  \redparaindex{\omega+n+1,\Pi_1'\setminus\{p\}}
  w_1/q
.}
\\
Otherwise 
we have
\bigmath{
  (\,
   (\repl{l_{0,q}}{p}{r_{1,q p}\xi}\sigma,
    C_{1,q p}\xi\sigma,    
    1),\penalty-1\,
   (r_{0,q}\sigma,
    C_{0,q}\sigma,
    1),\penalty-1\,
    l_{0,q}\sigma,\penalty-1\,
    p\,)
  \in{\rm CP}(\R)
}
(due to Claim~5);
\bigmath{p\tightnotequal\emptyset}
(due to Claim~10);
\bigmath{C_{1,q p}\xi\sigma\varphi=C_{1,q p}\mu_{1,q p}}
is fulfilled \wrt\ \redindex{\omega+n};
\bigmath{C_{0,q}\sigma\varphi=C_{0,q}\mu_{0,q}}
is fulfilled \wrt\ \redindex{\omega+n}.
Since \RX\ is \math\omega-level confluent up to \math n
(by our induction hypothesis) 
and \math\omega-shallow confluent up to \math\omega\
(by Claim~0)
due to our assumed \math\omega-level parallel closedness 
(matching the definition's \math{n} to our \math{n\tight+1})
we have
\bigmath{
  u'
  \tightequal
  \repl{l_{0,q}}{p}{r_{1,q p}\xi}\sigma\varphi
  \penalty-1
  \redparaindex{\omega+n+1}
  \penalty-1
  v_1
  \penalty-1
  \refltransindex{\omega}
  v_2
  \antirefltransindex{\omega}
  r_{0,q}\sigma\varphi
  \tightequal
  r_{0,q}\mu_{0,q}
  \tightequal
  w_0/q
}
for some \math{v_1}, \math{v_2}.
We then have
\bigmath{
  v_1
  \antiredparaindex{\omega+n+1,\Pi''} 
  u'
  \redparaindex{\omega+n+1,\Pi_1'\setminus\{p\}}
  w_1/q
}
for some \math{\Pi''}.
By 
\bigmath{
  \displaystyle
  \sum_{p''\in\Omega(\Pi'',\Pi_1'\setminus\{p\})}
  \lambda(u'/p'')
  \ \ \preceq
  \sum_{p''\in\Pi_1'\setminus\{p\}}
  \lambda(u'/p'')
  \ \ =
  \sum_{p''\in\Pi_1'\setminus\{p\}}
  \lambda(u/q p'')
  \ \ \prec
}\\\bigmath{\displaystyle
  \sum_{p''\in\Pi_1'}
  \lambda(u/q p'')
  \ \ =
  \sum_{p'\in q\Pi_1'}
  \lambda(u/p')
  \ \ =
  \sum_{p'\in\Omega(\{q\},\Pi_1)}
  \lambda(u/p')
  \ \ \preceq
  \sum_{p'\in\Omega(\Pi_0,\Pi_1)}
  \lambda(u/p')
,}
due to our second induction level 
we get some \math{v_1'}, \math{v_3} with 
\bigmath{
  v_1
  \refltransindex{\omega}
  v_3
  \redparaindex{\omega+n+1}
  \tight\circ
  \refltransindex{\omega}
  v_1'
  \antirefltransindex{\omega+n+1}
  w_1/q
.}
By Claim~0 we can close the peak at \math{v_1} according to
\bigmath{
  v_2
  \refltransindex\omega
  v_4
  \antirefltransindex\omega
  v_3
}
for some \math{v_4}.
Finally 
by the assumption of our lemma that
\bigmath{
  \redparaindex{\RX,\omega+n_1+1}
  \tight\circ
  \refltransindex{\RX,\omega}
}
strongly commutes over \refltransindex{\omega},
the peak at \math{v_3} can be closed according to 
\bigmath{
  v_4
  \redparaindex{\omega+n+1}
  \tight\circ
  \refltransindex{\omega}
  \circ
  \antirefltransindex{\omega}
  v_1'
.}%
\\
\Qeddouble{``The critical peak case''}\QEDtriple{``\math{q\tightnotin\Pi_1}''}

\pagebreak

\noindent
\underline{\underline{\underline{\math{q\tightin\Pi_1}:}}}
Define \math{\Pi_0':=\setwith{p}{q p\tightin\Pi_0}}.
We have two cases:

\yestop
\noindent
\underline{\underline{``The second variable overlap (if any) case'':
\math{
  \forall p\tightin\Pi_0'\tightcap\TPOS{l_{1,q}}\stopq
    l_{1,q}/p\tightin\V
}:}}
\begin{diagram}
l_{1,q}\mu_{1,q}&&&\rredindex{\omega+n+1,\,\emptyset}&&&w_1/q
\\&&&&&&\dequal
\\\dredparaindex{\omega+n+1\,\Pi_0'}&&&&&&r_{1,q}\mu_{1,q}
\\&&&&&&\dredparaindex{\omega+n+1}
\\w_0/q&\requal&l_{1,q}\nu&&\rredindex{\omega+n+1}&&r_{1,q}\nu
\end{diagram}
\noindent
Define a function \math\Gamma\ on \V\ by (\math{x\tightin\V}):
\bigmath{
  \Gamma(x):=
  \setwith{(p',p'')}
          {l_{1,q}/p'\tightequal x\ \wedge\ p' p''\in\Pi_0'}
.}

\noindent
\underline{Claim~11:}
There is some \math{\nu\in\Xsubst} with
\\\LINEmath{
  \forall x\in\V\stopq
    \inparenthesesoplist{
       x\nu
       \antiredparaindex{\omega+n+1}
       x\mu_{1,q}
    \oplistund
       \forall p'\tightin\DOM{\Gamma(x)}\stopq
         \replpar
           {x\mu_{1,q}}
           {p''}
           {r_{0,q p' p''}\mu_{0,q p' p''}}
           {(p',p'')\tightin\Gamma(x)}
         \tightequal
         x\nu
    }
.}
\\
\underline{Proof of Claim~11:}
\\
In case of \bigmath{\DOM{\Gamma(x)}\tightequal\emptyset} we define
\bigmath{x\nu:=x\mu_{1,q}.}
If there is some \math{p'} such that 
\bigmath{\DOM{\Gamma(x)}\tightequal\{p'\}}
we define 
\bigmath{
  x\nu
  :=
  \replpar
    {x\mu_{1,q}}{p''}{r_{0,q p' p''}\mu_{0,q p' p''}}{(p',p'')\tightin\Gamma(x)}
.}
This is appropriate since due to 
\bigmath{
  \forall(p',p'')\tightin\Gamma(x)\stopq
    x\mu_{1,q}/p''
    \tightequal 
    l_{1,q}\mu_{1,q}/p' p''
    \tightequal 
    u/q p' p''
    \tightequal 
    l_{0,q p' p''}\mu_{0,q p' p''}
}
we have
\\\LINEmath{
  \begin{array}{l@{}l@{}l}
  x\mu_{1,q}&
  \tightequal&
  \replpar
    {x\mu_{1,q}}
    {p''}
    {l_{0,q p' p''}\mu_{0,q p' p''}}
    {(p',p'')\tightin\Gamma(x)}
  \redparaindex{\omega+n+1}\\&&
  \replpar
     {x\mu_{1,q}}
     {p''}
     {r_{0,q p' p''}\mu_{0,q p' p''}}
     {(p',p'')\tightin\Gamma(x)}
  \tightequal
  x\nu.  
  \end{array}
}
\\
Finally, in case of \bigmath{\CARD{\DOM{\Gamma(x)}}\succ1,} \math{l_{1,q}} is
not linear in \math x. 
By the conditions of our lemma and Claim~5 this implies
\bigmath{x\tightin\Vcons.}
Since there is some \math{(p',p'')\in\Gamma(x)} with
\bigmath{
    x\mu_{1,q}/p''
    \tightequal 
    l_{0,q p' p''}\mu_{0,q p' p''}
}
this implies 
\bigmath{
   l_{0,q p' p''}\mu_{0,q p' p''}\tightin\tcc
}
and then 
\bigmath{
   l_{0,q p' p''}\tightin\tcs
}
which contradicts Claim~5.%
\QED{Claim~11}

\noindent
\underline{Claim~12:}
\bigmath{w_0/q\tightequal l_{1,q}\nu.}
\\
\underline{Proof of Claim~12:}
\\
By Claim~11 we get
\bigmath{
  w_0/q
  \tightequal
  \replpar 
    {u/q}
    {p' p''}
    {r_{0,q p' p''}\mu_{0,q p' p''}}
    {\exists x\tightin\V\stopq(p',p'')\tightin\Gamma(x)}
  \tightequal\\
  \replpar
    {\replpar
       {l_{1,q}}
       {p'}
       {x\mu_{1,q}}
       {l_{1,q}/p'\tightequal x\tightin\V}
    }
    {p' p''}
    {r_{0,q p' p''}\mu_{0,q p' p''}}
    {\exists x\tightin\V\stopq(p',p'')\tightin\Gamma(x)}
  \tightequal\\
  \replpar
    {l_{1,q}}
    {p'}
    {\replpar
       {x\mu_{1,q}}
       {p''}
       {r_{0,q p' p''}\mu_{0,q p' p''}}
       {(p',p'')\tightin\Gamma(x)}}
    {l_{1,q}/p'\tightequal x\tightin\V}
  \tightequal\\
  \replpar
    {l_{1,q}}
    {p'}
    {x\nu}
    {l_{1,q}/p'\tightequal x\tightin\V}
  \tightequal
  l_{1,q}\nu
.}\QED{Claim~12}

\noindent
\underline{Claim~13:}
\bigmath{
  r_{1,q}\nu
  \antiredparaindex{\omega+n+1}
  w_1/q
.}
\\
\underline{Proof of Claim~13:} 
Since \bigmath{r_{1,q}\mu_{1,q}\tightequal w_1/q,} 
this follows directly from Claim~11.%
\QED{Claim~13}

\noindent
By claims 12 and 13 
using Corollary~\ref{corollary parallel one} 
it now suffices to show
\bigmath{
  l_{1,q}\nu
  \redindex{\omega+n+1}
  r_{1,q}\nu
,}
which again follows from 
Lemma~\ref{lemma invariance of fulfilledness level}
(matching its \math{n_0} to our \math{n\tight+1}
      and its \math{n_1} to our \math{n})
since \RX\ is \math\omega-level confluent up to \math{n}
by our induction hypothesis 
and since
\bigmath{\forall x\tightin\V\stopq x\mu_{1,q}\refltransindex{\omega+n+1}x\nu}
by Claim~11 and Corollary~\ref{corollary parallel one}.%
\\\Qeddouble{``The second variable overlap (if any) case''}

\pagebreak

\yestop
\noindent
\underline{\underline{``The second critical peak case'':
There is some \math{p\in \Pi_0'\tightcap\TPOS{l_{1,q}}}
with \math{l_{1,q}/p\tightnotin\V}:}}
\begin{diagram}
l_{1,q}\mu_{1,q}
&&&
&\rredindex{\omega+n+1,\,\emptyset}&&
&&w_1/q
\\\dredindex{\omega+n+1,\,p}
&&&
&&&
&&\drefltransindex{\omega+n+1}
\\u'
&&&\rredparaindex{\omega+n+1}
&&&v_1
&\rrefltransindex{\omega}&v_2
\\\dredparaindex{\omega+n+1,\,\Pi_0'\setminus\{p\}}
&&&
&&&\drefltransindex{\omega+n+1}
&&\drefltransindex{\omega+n+1}
\\w_0/q
&\rrefltransindex{\omega}&\circ&\rredparaindex{\omega+n+1}
&\circ&\rrefltransindex{\omega}&v_1'
&\rrefltransindex{\omega}&\circ
\end{diagram}
Let \math{\xi\in\SUBST\V\V} be a bijection with 
\bigmath{
  \xi[\VAR{\kurzregelindex{0,q p}}]\cap\VAR{\kurzregelindex{1,q}}
  =
  \emptyset
.}
\\
Define
\bigmath{
  \Y
  :=
  \xi[\VAR{\kurzregelindex{0,q p}}]\cup\VAR{\kurzregelindex{1,q}}
.}
\\
Let \math{\varrho\in\Xsubst} be given by
$\ x\varrho=
\left\{\begin{array}{@{}l@{}l@{}}
  x\mu_{1,q}        &\mbox{ if }x\in\VAR{\kurzregelindex{1,q}}\\
  x\xi^{-1}\mu_{0,q p}&\mbox{ else}\\
\end{array}\right\}
\:(x\tightin\V)$.
\\
By
\math{
  l_{0,q p}\xi\varrho
  \tightequal 
  l_{0,q p}\xi\xi^{-1}\mu_{0,q p}
  \tightequal 
  u/q p
  \tightequal 
  l_{1,q}\mu_{1,q}/p  
  \tightequal 
  l_{1,q}\varrho/p
  \tightequal 
  (l_{1,q}/p)\varrho
}
\\
let
\math{
  \sigma:=\minmgu{\{(l_{0,q p}\xi}{l_{1,q}/p)\},\Y}
}
and
\math{\varphi\in\Xsubst}
with
\math{
  \domres{\inpit{\sigma\varphi}}\Y
  \tightequal
  \domres\varrho\Y
}.
\\
Define 
\math{
  u':=  
  \repl{l_{1,q}\mu_{1,q}}
       {p}
       {r_{0,q p}\mu_{0,q p}}
}. 
We get
\\\LINEmath{
  \begin{array}{l@{}l@{}l}
  w_0/q&
  \tightequal&
  \replpar{u/q}{p'}{r_{0,q p'}\mu_{0,q p'}}{p'\tightin\Pi_0'}
  \antiredparaindex{\omega+n+1,\Pi_0'\setminus\{p\}}
  \\&&
  \repl
    {\replpar
       {u/q}{p'}{l_{0,q p'}\mu_{0,q p'}}{p'\tightin\Pi_0'\tightsetminus\{p\}}}
    {p}
    {r_{0,q p}\mu_{0,q p}}
  \tightequal
  u'
  .  
  \end{array}
}  
\\
If 
\bigmath{
  \repl{l_{1,q}}{p}{r_{0,q p}\xi}\sigma
  \tightequal
  r_{1,q}\sigma
,}
then the proof is finished due to 
\\\linemath{
  w_0/q
  \antiredparaindex{\omega+n+1,\Pi_0'\setminus\{p\}}
  u'
  \tightequal
  \repl{l_{1,q}}{p}{r_{0,q p}\xi}\sigma\varphi
  \tightequal
  r_{1,q}\sigma\varphi
  \tightequal
  r_{1,q}\mu_{1,q}
  \tightequal
  w_1/q.
}
Otherwise 
we have
\bigmath{
  (\,
   (\repl{l_{1,q}}{p}{r_{0,q p}\xi}\sigma,
    C_{0,q p}\xi\sigma,    
    1),\penalty-1\,
   (r_{1,q}\sigma,
    C_{1,q}\sigma,
    1),\penalty-1\,
    l_{1,q}\sigma,\penalty-1\,
    p\,)
  \in{\rm CP}(\R)
}
(due to Claim~5);
\bigmath{C_{0,q p}\xi\sigma\varphi=C_{0,q p}\mu_{0,q p}}
is fulfilled \wrt\ \redindex{\omega+n};
\bigmath{C_{1,q}\sigma\varphi=C_{1,q}\mu_{1,q}}
is fulfilled \wrt\ \redindex{\omega+n}.
Since \RX\ is \math\omega-level confluent up to \math n
(by our induction hypothesis) 
and \math\omega-shallow confluent up to \math\omega\
(by Claim~0)
due to our assumed \math\omega-level parallel joinability 
(matching the definition's \math{n} to our \math{n\tight+1})
we have
\bigmath{
  u'
  \tightequal
  \repl{l_{1,q}}{p}{r_{0,q p}\xi}\sigma\varphi
  \redparaindex{\omega+n+1}
  \penalty-1
  v_1
  \refltransindex{\omega}
  \penalty-1
  v_2
  \antirefltransindex{\omega+n+1}
  \penalty-1
  r_{1,q}\sigma\varphi
  \tightequal
  r_{1,q}\mu_{1,q}
  \tightequal
  w_1/q
}
for some \math{v_1}, \math{v_2}.
We then have
\bigmath{
  w_0/q
  \antiredparaindex{\omega+n+1,\Pi_0'\setminus\{p\}}
  u'
  \redparaindex{\omega+n+1,\Pi''} 
  v_1
}
for some \math{\Pi''}.
Since
\bigmath{\displaystyle
  \sum_{p''\in\Omega(
    \Pi_0'\setminus\{p\}
    ,
    \Pi''
    )}
  \lambda(u'/p'')
  \ \ \preceq
  \sum_{p''\in\Pi_0'\setminus\{p\}}
  \lambda(u'/p'')
  \ \ =
  \sum_{p''\in\Pi_0'\setminus\{p\}}
  \lambda(u/q p'')
  \ \ \prec
  \sum_{p''\in\Pi_0'}
  \lambda(u/q p'')
  \ \ =
  \sum_{p'\in q\Pi_0'}
  \lambda(u/p')
  \ \ =
  \sum_{p'\in\Omega(
    \Pi_0
    ,
    \{q\}
    )}
  \lambda(u/p')
  \ \ \preceq
  \sum_{p'\in\Omega(\Pi_0,\Pi_1)}
  \lambda(u/p')
}
due to our second induction level 
we get some \math{v_1'} with 
\bigmath{
  w_0/q
  \refltransindex{\omega}
  \tight\circ
  \redparaindex{\omega+n+1}
  \tight\circ
  \refltransindex{\omega}
  v_1'
  \antirefltransindex{\omega+n+1}
  v_1
.} 
Finally the peak at \math{v_1} can be closed according to
\bigmath{
  v_1'
  \refltransindex{\omega}
  \circ
  \antirefltransindex{\omega+n+1}
  v_2
}
by Claim~0.%
\\\Qeddouble{``The second critical peak case''}
\end{proofparsepqed}

\pagebreak

\begin{proofparsepqed}
{Lemma~\ref{lemma level parallel closed second level three}}
\underline{Claim~0:}
\RX\ is \math\omega-shallow confluent up to \math\omega.
\\\underline{Proof of Claim~0:}
Directly by the assumed strong commutation,
\cf\ the proofs of the claims 2 and 3 of the proof of 
Lemma~\ref{lemma parallel closed first level two}.\QED{Claim~0}

\yestop
\noindent
\underline{Claim~1:}
If 
\bigmath{
    \refltransindex{\omega}
    \tight\circ
    \redparaindex{\omega+n}
    \tight\circ
    \refltransindex{\omega}
}
strongly commutes over
\refltransindex{\omega+n},
then
\redindex{\omega+n} and \redindex{\omega+n} are commuting.
\\
\underline{Proof of Claim~1:}
\bigmath{
    \refltransindex{\omega}
    \tight\circ
    \redparaindex{\omega+n}
    \tight\circ
    \refltransindex{\omega}
}
and
\refltransindex{\omega+n}
are commuting
by Lemma~\ref{lemma strong commutation one copy}.
Since by Corollary~\ref{corollary parallel one}
and \lemmamonotonicinbeta\
we have
\bigmath{
  \redindex{\omega+n}
  \subseteq
  \refltransindex{\omega}
  \tight\circ
  \redparaindex{\omega+n}
  \tight\circ
  \refltransindex{\omega}
  \subseteq
  \refltransindex{\omega+n}
,}
now \redindex{\omega+n} 
and \redindex{\omega+n} are commuting, too.\QED{Claim~1}

\yestop
\yestop
\noindent
For \math{n\prec\omega} 
we are going to show by induction on \math n 
the following property\footroom:
\\\LINEmath{
    w_0
    \antiredindex{\omega+n}
    u
    \redparaindex{\omega+n}
    w_1
  \quad\implies\quad
    w_0
    \refltransindex{\omega}
    \tight\circ
    \redparaindex{\omega+n}
    \tight\circ
    \refltransindex{\omega}
    \circ
    \antirefltransindex{\omega+n}
    w_1
.}
\begin{diagram}
u&&&\rredparaindex{\omega+n}&&&w_1
\\\dredindex{\omega+n}&&&&&&\drefltransindex{\omega+n}
\\w_0&\rrefltransindex{\omega}&\circ&\rredparaindex{\omega+n}
&\circ&\rrefltransindex{\omega}&\circ
\end{diagram}

\noindent
\underline{Claim~2:}
Let \math{\delta\prec\omega}.
If
\bigmath{
  \forall 
        n\tightpreceq\delta
  \stopq
    \forall w_0,w_1,u\stopq
    \inparenthesesoplist{
        w_0
        \antiredindex{\omega+n}
        u
        \redparaindex    {\omega+n}
        w_1
      \oplistimplies 
        w_0
        \refltransindex{\omega}
        \tight\circ
        \redparaindex{\omega+n}
        \tight\circ
        \refltransindex{\omega}
        \circ
        \antirefltransindex{\omega+n}
        w_1
    }
,}
then
\bigmath{
  \forall n\tightpreceq\delta\stopq
  \inparentheses{
            \refltransindex{\omega}
            \tight\circ
            \redparaindex{\omega+n}
            \tight\circ
            \refltransindex{\omega}
      \mbox{ strongly commutes over }
            \refltransindex{\omega+n}
  }
,}
and 
\RX\ is \math\omega-level confluent up to \math{\delta}.
\\
\underline{Proof of Claim~2:}
First we show the strong commutation.
Assume \math{n\tightpreceq\delta}. 
By Lemma~\ref{lemma strong commutation one copy} it suffices to show that
\bigmath{
    \refltransindex{\omega}
    \tight\circ
    \redparaindex{\omega+n}
    \tight\circ
    \refltransindex{\omega}
}
strongly commutes over
\redindex{\omega+n}.
Assume
\bigmath{
    u''
    \antiredindex{\omega+n}
    u'
    \refltransindex{\omega}
    u
    \redparaindex{\omega+n}
    w_1
    \refltransindex{\omega}
    w_2
}
(\cf\ diagram below).
By the strong commutation assumed for our lemma,
there are \math{w_0} and \math{w_0'} with
\bigmath{
    u''
    \refltransindex{\omega}
    w_0'
    \antirefltransindex{\omega}
    w_0
    \antionlyonceindex{\omega+n}
    u
.}
By the above property there are some \math{w_3}, \math{w_1'}
with
\bigmath{
    w_0
    \refltransindex{\omega}
    w_3
    \redparaindex{\omega+n}
    \tight\circ
    \refltransindex{\omega}
    w_1'
    \antirefltransindex{\omega+n}
    w_1
.}
By Claim~0 we can close the peak
\bigmath{ 
    w_1'
    \antirefltransindex{\omega+n}
    w_1
    \refltransindex{\omega}
    w_2
}
according to 
\bigmath{
    w_1'
    \refltransindex{\omega}
    w_2'
    \antirefltransindex{\omega+n}
    w_2
}
for some \math{w_2'}.
By Claim~0 again,
we can close the peak
\bigmath{
  w_0'
  \antirefltransindex{\omega}
  w_0
  \refltransindex{\omega}
  w_3
}
according to 
\bigmath{
  w_0'
  \refltransindex{\omega}
  w_3'
  \antirefltransindex{\omega}
  w_3
}
for some \math{w_3'}.
To close the whole diagram, we only have to show that we can close the peak
\bigmath{
    w_3'
    \antirefltransindex{\omega}
    w_3
    \redparaindex{\omega+n}
    \tight\circ
    \refltransindex{\omega}
    w_2'
}
according to 
\bigmath{
    w_3'
    \refltransindex{\omega}
    \tight\circ
    \redparaindex{\omega+n}
    \tight\circ
    \refltransindex{\omega}
    \circ
    \antirefltransindex{\omega}
    w_2'
,}
which is possible since it is assumed for our lemma 
(below the strong commutation assumption).
\begin{diagram}
u'&\rrefltransindex{\omega}&u
&&
&\rredparaindex{\omega+n}&&
&w_1&\rrefltransindex{\omega}&w_2
\\&&\donlyonceindex{\omega+n}
&&
&&&
&\drefltransindex{\omega+n}&&\drefltransindex{\omega+n}
\\\dredindex{\omega+n}&&w_0
&\rrefltransindex{\omega}&w_3
&\rredparaindex{\omega+n}&\circ&\rrefltransindex{\omega}
&w_1'&\rrefltransindex{\omega}&w_2'
\\&&\drefltransindex{\omega}
&&\drefltransindex{\omega}
&&&
&&&\drefltransindex{\omega}
\\u''&\rrefltransindex{\omega}&w_0'
&\rrefltransindex{\omega}&w_3'
&\rrefltransindex{\omega}&\circ&\rredparaindex{\omega+n}
&\circ&\rrefltransindex{\omega}&\circ
\end{diagram}

\pagebreak

\noindent
Finally we show \math\omega-level confluence up to \math\delta.
Assume
\math{n_0,n_1\prec\omega} with
\bigmath{\maxoftwo{n_0}{n_1}\tightpreceq\delta}
and
\bigmath{
    w_0
    \antirefltransindex{\omega+n_0}
    u
    \refltransindex    {\omega+n_1}
    w_1
.}
By \lemmamonotonicinbeta\ we get
\bigmath{
    w_0
    \antirefltransindex{\omega+\maxoftwo{n_0}{n_1}}
    u
    \refltransindex    {\omega+\maxoftwo{n_0}{n_1}}
    w_1
.}
Since \bigmath{\maxoftwo{n_0}{n_1}\tightpreceq\delta,}
above we have shown that 
\bigmath{
    \refltransindex{\omega}
    \tight\circ
    \redparaindex{\omega+\maxoftwo{n_0}{n_1}}
    \tight\circ
    \refltransindex{\omega}
}
strongly commutes over
\refltransindex{\omega+\maxoftwo{n_0}{n_1}}. 
By Claim~1 we finally get
\bigmath{
  w_0
  \refltransindex{\omega+\maxoftwo{n_0}{n_1}}
  \circ
  \antirefltransindex{\omega+\maxoftwo{n_0}{n_1}}
  w_1
}
as desired.%
\QED{Claim~2}

\yestop
\yestop
\noindent
Note that 
for \bigmath{n\tightequal0} 
our property follows 
from 
Corollary~\ref{corollary parallel one}
and 
Claim~0.

The benefit of 
Claim~2 is twofold: First, it says that our lemma is valid
if the above property
holds for all \math{n\prec\omega}.
Second, it strengthens the property when used as induction hypothesis. Thus 
(writing \math{n\tight+1} instead of \math{n} since we may assume
 \math{0\tightprec n})
it
now suffices to show
for
\math{n\prec\omega}
that 
\\\linemath{
    w_0
    \antiredindex{\omega+n+1,\bar p_0}
    u
    \redparaindex{\omega+n+1,\Pi_1}
    w_1
}
together with our induction hypotheses 
that\headroom
\\\linemath{\headroom
  \mbox{\RX\ is \math\omega-level confluent up to }n
}
\headroom
implies
\\\LINEmath{
    w_0
    \refltransindex{\omega}
    \tight\circ
    \redparaindex{\omega+n+1}
    \tight\circ
    \refltransindex{\omega}
    \circ
    \antirefltransindex{\omega+n+1}
    w_1
.}
\begin{diagram}
u&&&\rredparaindex{\omega+n_1+1,\,\Pi_1}
&&&w_1
\\\dredindex{\omega+n+1,\,\bar p_0}&&&
&&&\drefltransindex{\omega+n+1}
\\w_0&\rrefltransindex{\omega}&\circ&\rredparaindex{\omega+n+1}
&\circ&\rrefltransindex{\omega}&\circ
\end{diagram}

There are
\bigmath{\kurzregelindex{0,\bar p_0}\in\R}
and
\bigmath{\mu_{0,\bar p_0}\in\Xsubst}
with
\bigmath{u/p\tightequal l_{0,\bar p_0}\mu_{0,\bar p_0},}
\math{C_{0,\bar p_0}\mu_{0,\bar p_0}} fulfilled \wrt\ \redindex{\omega+n},
and
\bigmath{
  w_0
  \tightequal
  \repl{u}{p}{r_{0,\bar p_0}\mu_{0,\bar p_0}}
.}

\Wrog\ let the positions of \math{\Pi_1} be maximal
in the sense that for any \math{p\in\Pi_1} 
and \math{\Xi\subseteq \TPOS u\tightcap(p\N^+)}
we do not have 
\bigmath{
  u
  \redparaindex{\omega+n+1,(\Pi_1\setminus\{p\})\cup\Xi}
  w_1
}
anymore.
Then for each \math{p\in\Pi_1} there are
\bigmath{\kurzregelindex{1,p}\in\R}
and
\bigmath{\mu_{1,p}\in\Xsubst}
with
\bigmath{u/p\tightequal l_{1,p}\mu_{1,p},}
\bigmath{r_{1,p}\mu_{1,p}\tightequal w_1/p,}
\math{C_{1,p}\mu_{1,p}} fulfilled \wrt\ \redindex{\omega+n}.
Finally, 
\bigmath{
  w_1
  \tightequal
  \replpar{u}{p}{r_{1,p}\mu_{1,p}}{p\tightin\Pi_1}
.}

\pagebreak

\yestop
\noindent
\underline{Claim~5:}
\\
We may assume 
\bigmath{
    l_{0,\bar p_0}\tightnotin\tcs
}
and
\bigmath{
  \forall p\tightin\Pi_1\stopq
    l_{1,p}\tightnotin\tcs
.}
\\
\underline{Proof of Claim~5:}
In case of 
\bigmath{
    l_{0,\bar p_0}\tightin\tcs
}
we get 
\bigmath{w_0\antiredindex{\omega}u}
by Lemma~\ref{lemma invariance of fulfilledness two} 
(matching both its \math\mu\ and \math\nu\ to our \math{\mu_{0,\bar p_0}})
and then
our property follows 
from 
the assumption of our lemma 
(below the strong commutation assumption).
For the second restriction 
define \math{\Xi_1:=\setwith{p\tightin\Pi_1}{l_{1,p}\tightin\tcs}}
and 
\math{
  u_1':=\replpar{u}{p}{r_{1,p}\mu_{1,p}}{p\tightin\Pi_1\tightsetminus\Xi_1}
}.
If we have succeeded with our proof under the assumption of Claim~5,
then we have shown 
\bigmath{
  w_0
  \refltransindex{\omega}
  \tight\circ
  \redparaindex{\omega+n+1}
  \tight\circ
  \refltransindex{\omega}
  v_1
  \antirefltransindex{\omega+n+1}
  u_1'
}
for some \math{v_1}
(\cf\ diagram below).
By Lemma~\ref{lemma invariance of fulfilledness two} 
(matching both its \math\mu\ and \math\nu\ to our \math{\mu_{1,p}})
we get
\bigmath{
  \forall p\tightin\Xi_1\stopq
    l_{1,p}\mu_{1,p}\redindex\omega r_{1,p}\mu_{1,p}
}
and therefore
\bigmath{
  u_1'
  \redparaindex{\omega,\Xi_1}
  w_1
.}
Thus from
\bigmath{
  v_1
  \penalty-1
  \antirefltransindex{\omega+n_0+1}
  \penalty-1
  u_1'
  \penalty-1
  \refltransindex\omega
  \penalty-1
  w_1
}
we get 
\bigmath{
  v_1
  \refltransindex\omega
  v_2
  \antirefltransindex{\omega+n_0+1}
  w_1
}
for some \math{v_2}
by \math\omega-shallow confluence up to \math{\omega} (\cf\ Claim~0).
\begin{diagram}
u&&&\rredparaindex{\omega+n+1,\,\Pi_1\setminus\Xi_1}
&&&u_1'&\rredparaindex{\omega,\,\Xi_1}&w_1
\\
\dredindex{\omega+n+1,\,\bar p_0}&&&
&&&\drefltransindex{\omega+n+1}&&\drefltransindex{\omega+n+1}
\\w_0&\rrefltransindex{\omega}&\circ&\rredparaindex{\omega+n+1}
&\circ&\rrefltransindex{\omega}&v_1&\rrefltransindex{\omega}&v_2
\end{diagram}
\Qed{Claim~5}

\yestop
\noindent
Now we start a second level of induction on
\bigmath{
  \CARD{\Pi_1}
}
in \tightprec.

\noindent
Define the set of top positions by
\\\linemath{
  \displaystyle
  \Theta
  :=
      \setwith
      {p\tightin\{\bar p_0\}\tightcup\Pi_1}
      {\neg\exists q\tightin\{\bar p_0\}\tightcup\Pi_1\stopq
           \exists q'\tightin\N^+\stopq
             p\tightequal q q'
      }
.}
Since the prefix ordering is wellfounded we have
\bigmath{
  \forall p\tightin\{\bar p_0\}\tightcup\Pi_1\stopq
  \exists q\tightin\Theta\stopq
  \exists q'\tightin\N^\ast\stopq
    p\tightequal q q'
.}
It now suffices to show for all \math{q\in\Theta}
\\\linemath{\headroom\footroom
    w_0/q
    \refltransindex{\omega}
    \tight\circ
    \redparaindex{\omega+n+1}
    \tight\circ
    \refltransindex{\omega}
    \circ
    \antirefltransindex{\omega+n+1}
    w_1/q
}
because then we have 
\bigmath{
  w_0
  \tightequal
  \replpar{w_0}{q}{w_0/q}{q\tightin\Theta}
  \tightequal
  \replpar{\repl{u}{\bar p_0}{r_{0,\bar p_0}\mu_{0,\bar p_0}}}
          {q}{w_0/q}{q\tightin\Theta}
  \tightequal
  \replpar{u}{q}{w_0/q}{q\tightin\Theta}
    \refltransindex{\omega}
    \tight\circ
    \redparaindex{\omega+n+1}
    \tight\circ
    \refltransindex{\omega}
    \circ
    \antirefltransindex{\omega+n+1}
  \replpar{u}{q}{w_1/q}{q\tightin\Theta}
  \tightequal
  \\
  \replpar{\replpar{u}{p}{r_{1,p}\mu_{1,p}}{p\tightin\Pi_1}}
          {q}{w_1/q}{q\tightin\Theta}
  \tightequal
  \replpar{w_1}{q}{w_1/q}{q\tightin\Theta}
  \tightequal
  w_1
.}

\noindent
Therefore we are left with the following two cases for \math{q\in\Theta}:

\pagebreak

\yestop
\yestop
\noindent
\underline{\underline{\underline{\math{q\tightnotin\Pi_1}:}}}
Then \bigmath{q\tightequal\bar p_0.}
Define \math{\Pi_1':=\setwith{p}{q p\tightin\Pi_1}}.
\noindent
We have two cases:

\yestop
\noindent
\underline{\underline{``The variable overlap (if any) case'':
\math{
  \forall p\tightin\Pi_1'\tightcap\TPOS{l_{0,q}}\stopq
    l_{0,q}/p\tightin\V
}:}}
\begin{diagram}
l_{0,q}\mu_{0,q}&&\rredparaindex{\omega+n+1,\,\Pi_1'}&&&&w_1/q
\\&&&&&&\dequal
\\\dredindex{\omega+n+1,\,\emptyset}&&&&&&l_{0,q}\nu
\\&&&&&&\dredindex{\omega+n+1}
\\w_0/q&\requal&r_{0,q}\mu_{0,q}&&\rredparaindex{\omega+n+1}&&r_{0,q}\nu
\end{diagram}
\noindent
Define a function \math\Gamma\ on \V\ by (\math{x\tightin\V}):
\bigmath{
  \Gamma(x):=
  \setwith{(p',p'')}
          {l_{0,q}/p'\tightequal x\ \wedge\ p' p''\in\Pi_1'}
.}

\noindent
\underline{Claim~7:}
There is some \math{\nu\in\Xsubst} with
\\\LINEmath{
  \forall x\in\V\stopq
    \inparenthesesoplist{
       x\mu_{0,q}
       \redparaindex{\omega+n+1}
       x\nu
    \oplistund
       \forall p'\tightin\DOM{\Gamma(x)}\stopq
         x\nu
         \tightequal
         \replpar
           {x\mu_{0,q}}
           {p''}
           {r_{1,q p' p''}\mu_{1,q p' p''}}
           {(p',p'')\tightin\Gamma(x)}
    }
.}
\\
\underline{Proof of Claim~7:}
\\
In case of \bigmath{\DOM{\Gamma(x)}\tightequal\emptyset} we define
\bigmath{x\nu:=x\mu_{0,q}.}
If there is some \math{p'} such that 
\bigmath{\DOM{\Gamma(x)}\tightequal\{p'\}}
we define 
\bigmath{
  x\nu
  :=
  \replpar
    {x\mu_{0,q}}
    {p''}
    {r_{1,q p' p''}\mu_{1,q p' p''}}
    {(p',p'')\tightin\Gamma(x)}
.}
This is appropriate since due to 
\bigmath{
  \forall(p',p'')\tightin\Gamma(x)\stopq
    x\mu_{0,q}/p''
    \tightequal 
    l_{0,q}\mu_{0,q}/p' p''
    \tightequal 
    u/q p' p''
    \tightequal 
    l_{1,q p' p''}\mu_{1,q p' p''}
}
we have
\\\LINEmath{
  \begin{array}{l@{}l@{}l}
  x\mu_{0,q}&
  \tightequal&
  \replpar
    {x\mu_{0,q}}
    {p''}
    {l_{1,q p' p''}\mu_{1,q p' p''}}
    {(p',p'')\tightin\Gamma(x)}
  \redparaindex{\omega+n+1}\\&&
  \replpar
     {x\mu_{0,q}}
     {p''}
     {r_{1,q p' p''}\mu_{1,q p' p''}}
     {(p',p'')\tightin\Gamma(x)}
  \tightequal
  x\nu.  
  \end{array}
}
\\
Finally, in case of \bigmath{\CARD{\DOM{\Gamma(x)}}\succ1,} \math{l_{0,q}} is
not linear in \math x. By the conditions of our lemma and Claim~5 this implies
\bigmath{x\tightin\Vcons.}
Since there is some \math{(p',p'')\in\Gamma(x)} with
\bigmath{
    x\mu_{0,q}/p''
    \tightequal 
    l_{1,q p' p''}\mu_{1,q p' p''}
}
this implies 
\bigmath{
   l_{1,q p' p''}\mu_{1,q p' p''}\tightin\tcc
}
and then 
\bigmath{
   l_{1,q p' p''}\tightin\tcs
}
which contradicts Claim~5.%
\QED{Claim~7}

\noindent
\underline{Claim~8:}
\bigmath{
  l_{0,q}\nu
  \tightequal 
  w_1/q
.}
\\
\underline{Proof of Claim~8:}
\\
By Claim~7 we get
\bigmath{
  w_1/q
  \tightequal
  \replpar 
    {u/q}
    {p' p''}
    {r_{1,q p' p''}\mu_{1,q p' p''}}
    {\exists x\tightin\V\stopq(p',p'')\tightin\Gamma(x)}
  \tightequal\\
  \replpar
    {\replpar
       {l_{0,q}}
       {p'}
       {x\mu_{0,q}}
       {l_{0,q}/p'\tightequal x\tightin\V}
    }
    {p' p''}
    {r_{1,q p' p''}\mu_{1,q p' p''}}
    {\exists x\tightin\V\stopq(p',p'')\tightin\Gamma(x)}
  \tightequal\\
  \replpar
    {l_{0,q}}
    {p'}
    {\replpar
       {x\mu_{0,q}}
       {p''}
       {r_{1,q p' p''}\mu_{1,q p' p''}}
       {(p',p'')\tightin\Gamma(x)}}
    {l_{0,q}/p'\tightequal x\tightin\V}
  \tightequal\\
  \replpar
    {l_{0,q}}
    {p'}
    {x\nu}
    {l_{0,q}/p'\tightequal x\tightin\V}
  \tightequal
  l_{0,q}\nu
.}\QED{Claim~8}

\noindent
\underline{Claim~9:}
\bigmath{
  w_0/q
  \redparaindex{\omega+n+1}
  r_{0,q}\nu
.}
\\
\underline{Proof of Claim~9:} 
Since 
\bigmath{
  w_0/q
  \tightequal
  r_{0,q}\mu_{0,q}
,} 
this follows directly from Claim~7.%
\QED{Claim~9}

\noindent
By claims 8 and 9 it now suffices to show
\bigmath{
  l_{0,q}\nu
  \redindex{\omega+n+1}
  r_{0,q}\nu
,}
which again follows from 
Lemma~\ref{lemma invariance of fulfilledness level}
(matching its \math{n_0} to our \math{n\tight+1}
      and its \math{n_1} to our \math{n})
since \RX\ is \math\omega-level confluent up to
\math{n}
by our induction hypothesis 
and since
\bigmath{\forall x\tightin\V\stopq x\mu_{0,q}\refltransindex{\omega+n+1}x\nu}
by Claim~7 and Corollary~\ref{corollary parallel one}.%
\\\Qeddouble{``The variable overlap (if any) case''}

\pagebreak

\yestop
\noindent
\underline{\underline{``The critical peak case'':
There is some \math{p\in \Pi_1'\tightcap\TPOS{l_{0,q}}}
with \math{l_{0,q}/p\tightnotin\V}:}}
\begin{diagram}
l_{0,q}\mu_{0,q}&\rredindex{\omega+n+1,\,p}&u'
&&&\rredparaindex{\omega+n+1,\,\Pi_1'\setminus\{p\}}
&&&&&w_1/q
\\&&\donlyonceindex{\omega+n+1}
&&&
&&&&&\drefltransindex{\omega+n+1}
\\\dredindex{\omega+n+1,\,\emptyset}&&v_1
&\rrefltransindex{\omega}&v_3&&\rredparaindex{\omega+n+1}
&&\circ&\rrefltransindex{\omega}&v_1'
\\&&\drefltransindex{\omega}
&&\drefltransindex{\omega}&
&&&&&\drefltransindex{\omega}
\\w_0/q&\rrefltransindex{\omega}&v_2
&\rrefltransindex{\omega}&v_4&\rrefltransindex{\omega}&\circ
&\rredparaindex{\omega+n+1}
&\circ&\rrefltransindex{\omega}&\circ
\end{diagram}
\underline{Claim~10:}
\bigmath{p\tightnotequal\emptyset.}
\\
\underline{Proof of Claim~10:}
If \bigmath{p\tightequal\emptyset,} then
\bigmath{\emptyset\tightin\Pi_1',} then
\bigmath{q\tightin\Pi_1,} which contradicts our global case assumption.%
\QED{Claim~10}

\noindent
Let \math{\xi\in\SUBST\V\V} be a bijection with 
\bigmath{
  \xi[\VAR{\kurzregelindex{1,q p}}]\cap\VAR{\kurzregelindex{0,q}}
  =
  \emptyset
.}
\\
Define
\bigmath{
  \Y
  :=
  \xi[\VAR{\kurzregelindex{1,q p}}]\cup\VAR{\kurzregelindex{0,q}}
.}
\\
Let \math{\varrho\in\Xsubst} be given by
$\ x\varrho=
\left\{\begin{array}{@{}l@{}l@{}}
  x\mu_{0,q}        &\mbox{ if }x\in\VAR{\kurzregelindex{0,q}}\\
  x\xi^{-1}\mu_{1,q p}&\mbox{ else}\\
\end{array}\right\}
\:(x\tightin\V)$.
\\
By
\math{
  l_{1,q p}\xi\varrho
  \tightequal 
  l_{1,q p}\xi\xi^{-1}\mu_{1,q p}
  \tightequal 
  u/q p
  \tightequal 
  l_{0,q}\mu_{0,q}/p  
  \tightequal 
  l_{0,q}\varrho/p
  \tightequal 
  (l_{0,q}/p)\varrho
}
\\
let
\math{
  \sigma:=\minmgu{\{(l_{1,q p}\xi}{l_{0,q}/p)\},\Y}
}
and
\math{\varphi\in\Xsubst}
with
\math{
  \domres{\inpit{\sigma\varphi}}\Y
  \tightequal
  \domres\varrho\Y
}.
\\
Define 
\math{
  u':=  
  \repl{l_{0,q}\mu_{0,q}}
       {p}
       {r_{1,q p}\mu_{1,q p}}
}.
We get
\\\LINEmath{
  \arr{{l@{}l}
    u'\tightequal
    &
    \repl
      {\replpar
         {u/q}
         {p'}
         {l_{1,q p'}\mu_{1,q p'}}
         {p'\tightin\Pi_1'\tightsetminus\{p\}}}
      {p}
      {r_{1,q p}\mu_{1,q p}}
    \redparaindex{\omega+n+1,\Pi_1'\setminus\{p\}}
    \\&
    \replpar{u/q}{p'}{r_{1,q p'}\mu_{1,q p'}}{p'\tightin\Pi_1'}    
    \tightequal
    w_1/q
  .
  }
}
\\
If 
\bigmath{
  \repl{l_{0,q}}{p}{r_{1,q p}\xi}\sigma
  \tightequal
  r_{0,q}\sigma
,}
then the proof is finished due to 
\\\LINEmath{
  w_0/q
  \tightequal
  r_{0,q}\mu_{0,q}
  \tightequal
  r_{0,q}\sigma\varphi
  \tightequal
  \repl{l_{0,q}}{p}{r_{1,q p}\xi}\sigma\varphi
  \tightequal
  u'
  \redparaindex{\omega+n+1,\Pi_1'\setminus\{p\}}
  w_1/q
.}
\\
Otherwise 
we have
\bigmath{
  (\,
   (\repl{l_{0,q}}{p}{r_{1,q p}\xi}\sigma,
    C_{1,q p}\xi\sigma,    
    1),\penalty-1\,
   (r_{0,q}\sigma,
    C_{0,q}\sigma,
    1),\penalty-1\,
    l_{0,q}\sigma,\penalty-1\,
    p\,)
  \in{\rm CP}(\R)
}
(due to Claim~5);
\bigmath{p\tightnotequal\emptyset}
(due to Claim~10);
\bigmath{C_{1,q p}\xi\sigma\varphi=C_{1,q p}\mu_{1,q p}}
is fulfilled \wrt\ \redindex{\omega+n};
\bigmath{C_{0,q}\sigma\varphi=C_{0,q}\mu_{0,q}}
is fulfilled \wrt\ \redindex{\omega+n}.
Since \RX\ is \math\omega-level confluent up to \math n
(by our induction hypothesis) 
and \math\omega-shallow confluent up to \math\omega\
(by Claim~0)
due to our assumed \math\omega-level closedness 
(matching the definition's \math{n} to our \math{n\tight+1})
we have
\bigmath{
  u'
  \tightequal
  \repl{l_{0,q}}{p}{r_{1,q p}\xi}\sigma\varphi
  \penalty-1
  \onlyonceindex{\omega+n+1}
  \penalty-1
  v_1
  \penalty-1
  \refltransindex{\omega}
  v_2
  \antirefltransindex{\omega}
  r_{0,q}\sigma\varphi
  \tightequal
  r_{0,q}\mu_{0,q}
  \tightequal
  w_0/q
}
for some \math{v_1}, \math{v_2}.
We then have
\bigmath{
  v_1
  \antionlyonceindex{\omega+n+1} 
  u'
  \redparaindex{\omega+n+1,\Pi_1'\setminus\{p\}}
  w_1/q
.}
By 
\bigmath{
  \CARD{\Pi_1'\tightsetminus\{p\}}
  \prec
  \CARD{\Pi_1'}
  \preceq
  \CARD{\Pi_1}
,}
due to our second induction level 
we get some \math{v_1'} with 
\bigmath{
  v_1
  \refltransindex{\omega}
  \tight\circ
  \redparaindex{\omega+n+1}
  \tight\circ
  \refltransindex{\omega}
  v_1'
  \antirefltransindex{\omega+n+1}
  w_1/q
.}
By Claim~0 we can close the peak at \math{v_1} according to
\bigmath{
  v_2
  \refltransindex\omega
  v_4
  \antirefltransindex\omega
  v_3
}
for some \math{v_4}.
Finally by the assumption of our lemma
(below the strong commutation assumption)
the peak at \math{v_3} can be closed according to 
\bigmath{
  v_4
  \refltransindex{\omega}
  \tight\circ
  \redparaindex{\omega+n}
  \tight\circ
  \refltransindex{\omega}
  \circ
  \antirefltransindex{\omega+n}
  v_1'
.}%
\\
\Qeddouble{``The critical peak case''}\QEDtriple{``\math{q\tightnotin\Pi_1}''}

\pagebreak

\noindent
\underline{\underline{\underline{\math{q\tightin\Pi_1}:}}}
If there is  no \math{\bar p_0'} with \bigmath{q\bar p_0'\tightequal\bar p_0,}
then the proof is finished due to 
\bigmath{
  w_0/q
  \tightequal
  u/q
  \tightequal
  l_{1,q}\mu_{1,q}
  \redindex{\omega+n+1}
  r_{1,q}\mu_{1,q}
  \tightequal
  w_1/q
.}
Otherwise, we can define \math{\bar p_0'} by
\bigmath{q\bar p_0'\tightequal\bar p_0.}
We have two cases:

\yestop
\noindent
\underline{\underline{%
``The second variable overlap case'':%
}}
\\\LINEnomath{\underline{\underline{%
There are \math{x\tightin\V} and \math{p'}, \math{p''} such that
\math{
    l_{1,q}/p'
    \tightequal
    x
  \und
    p'p''
    \tightequal
    \bar p_0'
}:%
}}}
\begin{diagram}
l_{1,q}\mu_{1,q}&&&\rredindex{\omega+n+1,\,\emptyset}&&&w_1/q
\\&&&&&&\dequal
\\\dredindex{\omega+n+1,\,\bar p_0'}&&&&&&r_{1,q}\mu_{1,q}
\\&&&&&&\dredparaindex{\omega+n+1}
\\w_0/q&\requal&l_{1,q}\nu&&\rredindex{\omega+n+1}&&r_{1,q}\nu
\end{diagram}
\noindent
\underline{Claim~11a:}
We have 
\bigmath{x\mu_{1,q}/p''\tightequal l_{0,\bar p_0}\mu_{0,\bar p_0}}
and may assume
\bigmath{x\tightin\Vsig.}
\\\underline{Proof of Claim~11a:}
We have 
\bigmath{
  x\mu_{1,q}/p''
  \tightequal
  l_{1,q}\mu_{1,q}/p' p''
  \tightequal
  u/q p' p''
  \tightequal
  u/q\bar p_0'
  \tightequal
  u/\bar p_0
  \tightequal
  l_{0,\bar p_0}\mu_{0,\bar p_0}
.}
If \bigmath{x\tightin\Vcons,}
then \bigmath{x\mu_{1,q}\tightin\tcc,}
then \bigmath{x\mu_{1,q}/p''\tightin\tcc,}
then \bigmath{l_{0,\bar p_0}\mu_{0,\bar p_0}\tightin\tcc,}
and then \bigmath{l_{0,\bar p_0}\tightin\tcs}
which we may assume not to be the case by Claim~5.%
\QED{Claim~11a}

\noindent
\underline{Claim~11b:}
We can define \math{\nu\in\Xsubst} by
\bigmath{
  x\nu
  \tightequal
  \repl
    {x\mu_{1,q}}
    {p''}
    {r_{0,\bar p_0}\mu_{0,\bar p_0}}
}
and
\bigmath{
  \forall y\tightin\V\tightsetminus\{x\}\stopq y\nu\tightequal y\mu_{1,q}
.}
Then we have 
\bigmath{
  x\mu_{1,q}\redindex{\omega+n+1}x\nu
.}
\\
\underline{Proof of Claim~11b:}
This follows directly from Claim~11a.%
\QED{Claim~11b}

\noindent
\underline{Claim~12:}
\bigmath{
  w_0/q
  \tightequal
  l_{1,q}\nu
.}
\\
\underline{Proof of Claim~12:}
By the left-linearity assumption of our lemma, Claim~5, and Claim~11a 
we may assume
\bigmath{
  \setwith
    {p'''}
    {l_{1,q}/p'''\tightequal x}
  =
  \{p'\}
.}
Thus, by Claim~11b we get
\bigmath{
  w_0/q
  \tightequal
  \repl 
    {u/q}
    {\bar p_0'}
    {r_{0,\bar p_0}\mu_{0,\bar p_0}}
  \tightequal\\
  \repl
    {\replpar
       {l_{1,q}}
       {p'''}
       {y\mu_{1,q}}
       {l_{1,q}/p'''\tightequal y\tightin\V}
    }
    {\bar p_0'}
    {r_{0,\bar p_0}\mu_{0,\bar p_0}}
  \tightequal\\
  \repl
    {\repl
       {\replpar
          {l_{1,q}}
          {p'''}
          {y\mu_{1,q}}
          {l_{1,q}/p'''\tightequal y\tightin\V\und y\tightnotequal x}
       }
       {p'}
       {x\mu_{1,q}}
    }
    {p' p''}
    {r_{0,\bar p_0}\mu_{0,\bar p_0}}
  \tightequal\\
  \repl
    {\replpar
       {l_{1,q}}
       {p'''}
       {y\nu}
       {l_{1,q}/p'''\tightequal y\tightin\V\und y\tightnotequal x}
    }
    {p'}
    {\repl
       {x\mu_{1,q}}
       {p''}
       {r_{0,\bar p_0}\mu_{0,\bar p_0}}
    }
  \tightequal\\
  \replpar
    {l_{1,q}}
    {p'''}
    {y\nu}
    {l_{1,q}/p'''\tightequal y\tightin\V}
  \tightequal
  l_{1,q}\nu
.}\QED{Claim~12}

\noindent
\underline{Claim~13:}
\bigmath{
  r_{1,q}\nu
  \antiredparaindex{\omega+n+1}
  w_1/q
.}
\\
\underline{Proof of Claim~13:} 
Since \bigmath{r_{1,q}\mu_{1,q}\tightequal w_1/q,} 
this follows directly from Claim~11b.%
\QED{Claim~13}

\noindent
By claims 12 and 13 
using Corollary~\ref{corollary parallel one} 
it now suffices to show
\bigmath{
  l_{1,q}\nu
  \redindex{\omega+n+1}
  r_{1,q}\nu
,}
which again follows from 
Lemma~\ref{lemma invariance of fulfilledness level}
(matching its \math{n_0} to our \math{n\tight+1}
      and its \math{n_1} to our \math{n})
since \RX\ is \math\omega-level confluent up to \math{n}
by our induction hypothesis 
and since
\bigmath{\forall x\tightin\V\stopq x\mu_{1,q}\refltransindex{\omega+n+1}x\nu}
by Claim~11b.%
\\\Qeddouble{``The second variable overlap case''}

\pagebreak

\yestop
\noindent
\underline{\underline{``The second critical peak case'':
\math{
    \bar p_0'\tightin\TPOS{l_{1,q}}
  \und
    l_{1,q}/\bar p_0'\tightnotin\V
}:%
}}
\begin{diagram}
l_{1,q}\mu_{1,q}
&&&
&\rredindex{\omega+n+1,\,\emptyset}&&
&&w_1/q
\\\dredindex{\omega+n+1,\,\bar p_0'}
&&&
&&&
&&\drefltransindex{\omega+n+1}
\\w_0/q
&\rrefltransindex\omega&\circ&
&\rredparaindex{\omega+n+1}&&\circ
&\rrefltransindex{\omega}&\circ
\end{diagram}
Let \math{\xi\in\SUBST\V\V} be a bijection with 
\bigmath{
  \xi[\VAR{\kurzregelindex{0,\bar p_0}}]\cap\VAR{\kurzregelindex{1,q}}
  =
  \emptyset
.}
\\
Define
\bigmath{
  \Y
  :=
  \xi[\VAR{\kurzregelindex{0,\bar p_0}}]\cup\VAR{\kurzregelindex{1,q}}
.}
\\
Let \math{\varrho\in\Xsubst} be given by
$\ x\varrho=
\left\{\begin{array}{@{}l@{}l@{}}
  x\mu_{1,q}        &\mbox{ if }x\in\VAR{\kurzregelindex{1,q}}\\
  x\xi^{-1}\mu_{0,\bar p_0}&\mbox{ else}\\
\end{array}\right\}
\:(x\tightin\V)$.
\\
By
\math{
  l_{0,\bar p_0}\xi\varrho
  \tightequal 
  l_{0,\bar p_0}\xi\xi^{-1}\mu_{0,\bar p_0}
  \tightequal 
  u/\bar p_0
  \tightequal 
  u/q\bar p_0'
  \tightequal 
  l_{1,q}\mu_{1,q}/\bar p_0'  
  \tightequal 
  l_{1,q}\varrho/\bar p_0'
  \tightequal 
  (l_{1,q}/\bar p_0')\varrho
}
\\
let
\math{
  \sigma:=\minmgu{\{(l_{0,\bar p_0}\xi}{l_{1,q}/\bar p_0')\},\Y}
}
and
\math{\varphi\in\Xsubst}
with
\math{
  \domres{\inpit{\sigma\varphi}}\Y
  \tightequal
  \domres\varrho\Y
}.
\\
If 
\bigmath{
  \repl{l_{1,q}}{\bar p_0'}{r_{0,\bar p_0}\xi}\sigma
  \tightequal
  r_{1,q}\sigma
,}
then the proof is finished due to 
\\\linemath{
  w_0/q
  \tightequal
  \repl{l_{1,q}\mu_{1,q}}{\bar p_0'}{r_{0,\bar p_0}\mu_{0,\bar p_0}}
  \tightequal
  \repl{l_{1,q}}{\bar p_0'}{r_{0,\bar p_0}\xi}\sigma\varphi
  \tightequal
  r_{1,q}\sigma\varphi
  \tightequal
  r_{1,q}\mu_{1,q}
  \tightequal
  w_1/q.
}
Otherwise 
we have
\bigmath{
  (\,
   (\repl{l_{1,q}}{\bar p_0'}{r_{0,\bar p_0}\xi}\sigma,
    C_{0,\bar p_0}\xi\sigma,    
    1),\penalty-1\,
   (r_{1,q}\sigma,
    C_{1,q}\sigma,
    1),\penalty-1\,
    l_{1,q}\sigma,\penalty-1\,
    \bar p_0'\,)
  \in{\rm CP}(\R)
}
(due to Claim~5);
\bigmath{C_{0,\bar p_0}\xi\sigma\varphi=C_{0,\bar p_0}\mu_{0,\bar p_0}}
is fulfilled \wrt\ \redindex{\omega+n};
\bigmath{C_{1,q}\sigma\varphi=C_{1,q}\mu_{1,q}}
is fulfilled \wrt\ \redindex{\omega+n}.
Since \RX\ is \math\omega-level confluent up to \math n
(by our induction hypothesis) 
and \math\omega-shallow confluent up to \math\omega\
(by Claim~0)
due to our assumed \math\omega-level weak parallel joinability 
(matching the definition's \math{n} to our \math{n\tight+1})
we have
\bigmath{
  w_0/q
  \tightequal
  \repl{l_{1,q}\mu_{1,q}}{\bar p_0'}{r_{0,\bar p_0}\mu_{0,\bar p_0}}
  \tightequal
  \repl{l_{1,q}}{\bar p_0'}{r_{0,\bar p_0}\xi}\sigma\varphi
  \refltransindex{\omega}
  \tight\circ
  \redparaindex{\omega+n+1}
  \tight\circ
  \refltransindex{\omega}
  \circ
  \antirefltransindex{\omega+n+1}
  \penalty-1
  r_{1,q}\sigma\varphi
  \tightequal
  r_{1,q}\mu_{1,q}
  \tightequal
  w_1/q
.}
\\\Qeddouble{``The second critical peak case''}
\end{proofparsepqed}

\pagebreak

\begin{proofparsepqed}{Lemma~\ref
{lemma level strongly closed second level two}}
\underline{Claim~0:}
\RX\ is \math\omega-shallow confluent up to \math\omega.
\\\underline{Proof of Claim~0:}
Directly by the assumed strong commutation,
\cf\ the proofs of the claims 2 and 3 of the proof of 
Lemma~\ref{lemma parallel closed first level two}.\QED{Claim~0}

\yestop
\noindent
\underline{Claim~1:}
If 
\bigmath{
    \refltransindex{\omega}
    \tight\circ
    \redindex{\omega+n}
    \tight\circ
    \refltransindex{\omega}
}
strongly commutes over
\refltransindex{\omega+n},
then
\redindex{\omega+n} is confluent.
\\
\underline{Proof of Claim~1:}
\bigmath{
    \refltransindex{\omega}
    \tight\circ
    \redindex{\omega+n}
    \tight\circ
    \refltransindex{\omega}
}
and
\refltransindex{\omega+n}
are commuting
by Lemma~\ref{lemma strong commutation one copy}.
Since by \lemmamonotonicinbeta\
we have
\bigmath{
  \redindex{\omega+n}
  \subseteq
  \refltransindex{\omega}
  \tight\circ
  \redindex{\omega+n}
  \tight\circ
  \refltransindex{\omega}
  \subseteq
  \refltransindex{\omega+n}
,}
now \redindex{\omega+n} 
and \redindex{\omega+n} are commuting, too.\QED{Claim~1}

\yestop
\yestop
\noindent
For \math{n\prec\omega} 
we are going to show by induction on \math{n} the following property\footroom:
\\\LINEmath{
    w_0
    \antiredindex{\omega+n}
    u
    \redindex{\omega+n}
    w_1
  \quad\implies\quad
    w_0
    \refltransindex{\omega}
    \tight\circ
    \onlyonceindex{\omega+n}
    \tight\circ
    \refltransindex{\omega}
    \circ
    \antirefltransindex{\omega+n}
    w_1
.}
\begin{diagram}
u&&&\rredindex{\omega+n}&&&w_1
\\
\dredindex{\omega+n}&&&&&&\drefltransindex{\omega+n}
\\
w_0&\rrefltransindex{\omega}&\circ&\ronlyonceindex{\omega+n}
&\circ&\rrefltransindex{\omega}&\circ
\end{diagram}

\noindent
\underline{Claim~2:}
Let \math{\delta\prec\omega}.
If
\bigmath{
  \forall n\tightpreceq\delta\stopq
  \forall w_0,w_1,u\stopq
  \inparentheses{ 
    \inparenthesesoplist{
        w_0
        \antiredindex{\omega+n}
        u
        \redindex{\omega+n}
        w_1
      \oplistimplies 
        w_0
        \refltransindex{\omega}
        \tight\circ
        \onlyonceindex{\omega+n}
        \tight\circ
        \refltransindex{\omega}
        \circ
        \antirefltransindex{\omega+n}
        w_1
    }
  }
,}
then
\math{
  \forall n\tightpreceq\delta\stopq
  \inparentheses{ 
            \refltransindex{\omega}
            \tight\circ
            \redindex{\omega+n}
            \tight\circ
            \refltransindex{\omega}
      \mbox{ strongly commutes over }
            \refltransindex{\omega+n}
  }
,}
and 
\RX\ is \math\omega-level confluent up to \math{\delta}.
\\
\underline{Proof of Claim~2:}
First we show the strong commutation.
Assume \math{n\tightpreceq\delta}. 
By Lemma~\ref{lemma strong commutation one copy} it suffices to show that
\bigmath{
    \refltransindex{\omega}
    \tight\circ
    \redindex{\omega+n}
    \tight\circ
    \refltransindex{\omega}
}
strongly commutes over
\redindex{\omega+n}.
Assume
\bigmath{
    u''
    \antiredindex{\omega+n}
    u'
    \refltransindex{\omega}
    u
    \redindex{\omega+n}
    w_1
    \refltransindex{\omega}
    w_2
}
(\cf\ diagram below).
By the strong commutation assumed for our lemma,
there are \math{w_0} and \math{w_0'} with
\bigmath{
    u''
    \refltransindex{\omega}
    w_0'
    \antirefltransindex{\omega}
    w_0
    \antionlyonceindex{\omega+n}
    u
.}
By the above property there are some \math{w_3}, \math{w_1'}
with
\bigmath{
    w_0
    \refltransindex{\omega}
    w_3
    \onlyonceindex{\omega+n}
    \tight\circ
    \refltransindex{\omega}
    w_1'
    \antirefltransindex{\omega+n}
    w_1
.}
By Claim~0 we can close the peak
\bigmath{ 
    w_1'
    \antirefltransindex{\omega+n}
    w_1
    \refltransindex{\omega}
    w_2
}
according to 
\bigmath{
    w_1'
    \refltransindex{\omega}
    w_2'
    \antirefltransindex{\omega+n}
    w_2
}
for some \math{w_2'}.
By Claim~0 again,
we can close the peak
\bigmath{
  w_0'
  \antirefltransindex{\omega}
  w_0
  \refltransindex{\omega}
  w_3
}
according to 
\bigmath{
  w_0'
  \refltransindex{\omega}
  w_3'
  \antirefltransindex{\omega}
  w_3
}
for some \math{w_3'}.
To close the whole diagram, we only have to show that we can close the peak
\bigmath{
    w_3'
    \antirefltransindex{\omega}
    w_3
    \onlyonceindex{\omega+n}
    \tight\circ
    \refltransindex{\omega}
    w_2'
}
according to 
\bigmath{
    w_3'
    \onlyonceindex{\omega+n}
    \tight\circ
    \refltransindex{\omega}
    \circ
    \antirefltransindex{\omega}
    w_2'
,}
which is possible due to the strong commutation assumed for our lemma
or due to Claim~0.
\begin{diagram}
u'&\rrefltransindex{\omega}&u
&&
&\rredindex{\omega+n}&&
&w_1&\rrefltransindex{\omega}&w_2
\\&&\donlyonceindex{\omega+n}
&&
&&&
&\drefltransindex{\omega+n}&&\drefltransindex{\omega+n}
\\\dredindex{\omega+n}&&w_0
&\rrefltransindex{\omega}&w_3
&\ronlyonceindex{\omega+n}&\circ&\rrefltransindex{\omega}
&w_1'&\rrefltransindex{\omega}&w_2'
\\&&\drefltransindex{\omega}
&&\drefltransindex{\omega}
&&&
&&&\drefltransindex{\omega}
\\u''&\rrefltransindex{\omega}&w_0'
&\rrefltransindex{\omega}&w_3'
&\ronlyonceindex{\omega+n}
&\circ&&\rrefltransindex{\omega}&&\circ
\end{diagram}
\pagebreak

\noindent
Finally we show \math\omega-level confluence up to \math\delta.
Assume
\math{n_0,n_1\prec\omega} with
\bigmath{\maxoftwo{n_0}{n_1}\tightpreceq\delta}
and
\bigmath{
    w_0
    \antirefltransindex{\omega+n_0}
    u
    \refltransindex    {\omega+n_1}
    w_1
.}
By \lemmamonotonicinbeta\ we get
\bigmath{
    w_0
    \antirefltransindex{\omega+\maxoftwo{n_0}{n_1}}
    u
    \refltransindex    {\omega+\maxoftwo{n_0}{n_1}}
    w_1
.}
Since \bigmath{\maxoftwo{n_0}{n_1}\tightpreceq\delta,}
above we have shown that 
\bigmath{
    \refltransindex{\omega}
    \tight\circ
    \redindex{\omega+\maxoftwo{n_0}{n_1}}
    \tight\circ
    \refltransindex{\omega}
}
strongly commutes over
\refltransindex{\omega+\maxoftwo{n_0}{n_1}}. 
By Claim~1 we finally get
\bigmath{
  w_0
  \refltransindex{\omega+\maxoftwo{n_0}{n_1}}
  \circ
  \antirefltransindex{\omega+\maxoftwo{n_0}{n_1}}
  w_1
}
as desired.%
\QED{Claim~2}

\yestop
\yestop
\noindent
Note that 
for \bigmath{n\tightequal0} 
our property follows 
from Claim~0.

The benefit of 
Claim~2 is twofold: First, it says that our lemma is valid
if the above property
holds for all \math{n\prec\omega}.
Second, it strengthens the property when used as induction hypothesis. Thus 
(writing \math{n\tight+1} instead of \math{n} since we may assume
 \math{0\tightprec n})
it
now suffices to show
for
\math{n\prec\omega}
that 
\\\linemath{
    w_0
    \antiredindex{\omega+n+1,\bar p_0}
    u
    \redindex    {\omega+n+1,\bar p_1}
    w_1
}
together with our induction hypotheses 
that\headroom
\\\linemath{\headroom
  \mbox{\RX\ is \math\omega-level confluent up to }n
}
\headroom
implies
\\\LINEmath{
    w_0
    \refltransindex{\omega}
    \tight\circ
    \onlyonceindex{\omega+n+1}
    \tight\circ
    \refltransindex{\omega}
    \circ
    \antirefltransindex{\omega+n+1}
    w_1
.}
\begin{diagram}
u&&&\rredindex{\omega+n+1,\,\bar p_1}
&&&w_1
\\
\dredindex{\omega+n+1,\,\bar p_0}&&&
&&&\drefltransindex{\omega+n+1}
\\
w_0&\rrefltransindex{\omega}&\circ&\ronlyonceindex{\omega+n+1}
&\circ&\rrefltransindex{\omega}&\circ
\end{diagram}

\yestop
\noindent
Now for each \math{i\prec2} there are
\bigmath{\kurzregelindex{i}\in\R}
and
\bigmath{\mu_{i}\in\Xsubst}
with
\bigmath{u/\bar p_i\tightequal l_{i}\mu_{i},}
\bigmath{
  w_i\tightequal\repl{u}{\bar p_i}{r_{i}\mu_{i}}
,}
and
\math{C_{i}\mu_{i}} fulfilled \wrt\ \redindex{\omega+n}.

\yestop
\noindent
\underline{Claim~5:}
We may assume 
\bigmath{
  \forall i\tightprec2\stopq
    l_{i}\tightnotin\tcs
.}
\\
\underline{Proof of Claim~5:}
In case of \bigmath{l_i\tightin\tcs} we get \bigmath{u\redindex\omega w_i}
by Lemma~\ref{lemma invariance of fulfilledness two}
(matching both its \math\mu\ and \math\nu\ to our \math{\mu_i}).
In case of ``\math{i\tightequal0}'' our property follows from the
strong commutation assumption of our lemma. 
In case of ``\math{i\tightequal1}'' our property follows from Claim~0.%
\QED{Claim~5}

\yestop
\yestop
\noindent
In case of \bigmath{\neitherprefix{\bar p_0}{\bar p_1}} 
we have 
\bigmath{
  w_{i}/\bar p_{1-i}
  \tightequal
  \repl u{\bar p_{i}}{r_{i}\mu_{i}}/\bar p_{1-i}
  \tightequal
  u/\bar p_{1-i}
  \tightequal
  l_{1-i}\mu_{1-i}
} 
and
therefore
\bigmath{
  w_{i}
  \redindex{\omega+n+1}
  \replpar u{\bar p_{k}}{r_{k}\mu_{k}}{k\tightprec2}
,}
\ie\ our proof is finished.
Thus, according to whether \math{\bar p_0} is a prefix of \math{\bar p_1}
or vice versa, we have the following two cases left:
\pagebreak

\yestop
\yestop
\noindent
\underline{\underline{\underline{%
There is some \math{\bar p_1'} with
\bigmath{\bar p_0\bar p_1'\tightequal\bar p_1} and
\bigmath{\bar p_1'\tightnotequal\emptyset}:%
}}}

\noindent
We have two cases:

\noindent
\underline{\underline{``The variable overlap case'':}}
\\\LINEnomath{\underline{\underline{%
There are \math{x\in\V} and \math{p'}, \math{p''} such that
\math{
  l_{0}/p'\tightequal x
  \und
  p' p''\tightequal\bar p_1'
}:}}}
\begin{diagram}
l_{0}\mu_{0}&&&\rredindex{\omega+n+1,\,\bar p_1'}&&&w_1/\bar p_0
\\
&&&&&&\dequal
\\
\dredindex{\omega+n+1,\,\emptyset}&&&&&&l_{0}\nu
\\
&&&&&&\dredindex{\omega+n+1}
\\
w_0/\bar p_0&\requal&r_{0}\mu_{0}&&\ronlyonceindex{\omega+n+1}
&&r_{0}\nu
\end{diagram}
\noindent
\underline{Claim~6:}
We have 
\bigmath{x\mu_{0}/p''\tightequal l_{1}\mu_{1}}
and may assume
\bigmath{x\tightin\Vsig.}
\\\underline{Proof of Claim~6:}
We have 
\bigmath{
  x\mu_{0}/p''
  \tightequal
  l_{0}\mu_{0}/p' p''
  \tightequal
  u/\bar p_0 p' p''
  \tightequal
  u/\bar p_0\bar p_1'
  \tightequal
  u/\bar p_1
  \tightequal
  l_{1}\mu_{1}
.}
\\
If 
\bigmath{x\tightin\Vcons,}
then \bigmath{x\mu_{0}\tightin\tcc,}
then \bigmath{x\mu_{0}/p''\tightin\tcc,}
then 
\\
\bigmath{l_{1}\mu_{1}\tightin\tcc,}
and then \bigmath{l_{1}\tightin\tcs}
which we may assume not to be the case by Claim~5.%
\QED{Claim~6}

\noindent
\underline{Claim~7:}
We can define \math{\nu\in\Xsubst} by
\bigmath{
  x\nu
  \tightequal
  \repl
    {x\mu_{0}}
    {p''}
    {r_{1}\mu_{1}}
}
and
\bigmath{
  \forall y\tightin\V\tightsetminus\{x\}\stopq 
    y\nu\tightequal y\mu_{0}
.}
Then we have 
\bigmath{
  x\mu_{0}
  \redindex{\omega+n+1}
  x\nu
.}
\\
\underline{Proof of Claim~7:}
This follows directly from Claim~6.%
\QED{Claim~7}

\noindent
\underline{Claim~8:}
\bigmath{
  l_{0}\nu
  \tightequal
  w_1/\bar p_0
.}
\\
\underline{Proof of Claim~8:}
By the left-linearity assumption of our lemma and claims 5 and 6 we may assume
\bigmath{
  \setwith
    {p'''}
    {l_{0}/p'''\tightequal x}
  =
  \{p'\}
.}
Thus, by Claim~7 we get
\bigmath{
  w_1/\bar p_0
  \tightequal
  \repl 
    {u/\bar p_0}
    {\bar p_1'}
    {r_{1}\mu_{1}}
  \tightequal\\
  \repl
    {\replpar
       {l_{0}}
       {p'''}
       {y\mu_{0}}
       {l_{0}/p'''\tightequal y\tightin\V}
    }
    {\bar p_1'}
    {r_{1}\mu_{1}}
  \tightequal\\
  \repl
    {\repl
       {\replpar
          {l_{0}}
          {p'''}
          {y\mu_{0}}
          {l_{0}/p'''\tightequal y\tightin\V\und y\tightnotequal x}
       }
       {p'}
       {x\mu_{0}}
    }
    {p' p''}
    {r_{1}\mu_{1}}
  \tightequal\\
  \repl
    {\replpar
       {l_{0}}
       {p'''}
       {y\nu}
       {l_{0}/p'''\tightequal y\tightin\V\und y\tightnotequal x}
    }
    {p'}
    {\repl
       {x\mu_{0}}
       {p''}
       {r_{1}\mu_{1}}
    }
  \tightequal\\
  \replpar
    {l_{0}}
    {p'''}
    {y\nu}
    {l_{0}/p'''\tightequal y\tightin\V}
  \tightequal
  l_{0}\nu
.}\QED{Claim~8}

\noindent
\underline{Claim~9:}
\bigmath{
  w_0/\bar p_0
  \onlyonceindex{\omega+n+1}
  r_{0}\nu
.}
\\
\underline{Proof of Claim~9:} 
By the right-linearity assumption of our lemma and claims 5 and 6
we may assume 
\bigmath{\CARD{\setwith{p'''}{r_{0}/p'''\tightequal x}}\tightpreceq1.}
Thus by Claim~7 we get:
\bigmath{
  w_0/\bar p_0
  \tightequal 
  r_{0}\mu_{0}
  \tightequal
%  \replpar
%    {r_{0}}
%    {p'''}
%    {y\mu_{0}}
%    {r_{0}/p'''\tightequal y\tightin\V}
%  \tightequal
  \\
  \replpar
    {\replpar
       {r_{0}}
       {p'''}
       {y\mu_{0}}
       {r_{0}/p'''\tightequal y\tightin\V\tightsetminus\{x\}}
    }
    {p'''}
    {x\mu_{0}}
    {r_{0}/p'''\tightequal x}
  \onlyonceindex{\omega+n+1}
  \\
  \replpar
    {\replpar
       {r_{0}}
       {p'''}
       {y\mu_{0}}
       {r_{0}/p'''\tightequal y\tightin\V\tightsetminus\{x\}}
    }
    {p'''}
    {x\nu}
    {r_{0}/p'''\tightequal x}
  \tightequal
  \\
  \replpar
    {\replpar
       {r_{0}}
       {p'''}
       {y\nu}
       {r_{0}/p'''\tightequal y\tightin\V\tightsetminus\{x\}}
    }
    {p'''}
    {x\nu}
    {r_{0}/p'''\tightequal x}
  \tightequal
  r_{0}\nu
.}%
\QED{Claim~9}

\noindent
By claims 8 and 9 it now suffices to show
\bigmath{
  l_{0}\nu
  \redindex{\omega+n+1}
  r_{0}\nu
,}
which again follows from 
Lemma~\ref{lemma invariance of fulfilledness level}
since \RX\ is \math\omega-level confluent up to \math{n}
by our induction hypothesis 
and since
\bigmath{
  \forall y\tightin\V\stopq 
  y\mu_{0}
  \refltransindex{\omega+n+1}
  y\nu
}
by Claim~7.\QEDdouble{``The variable overlap case''}

\pagebreak

\yestop
\noindent
\underline{\underline{``The critical peak case'':
\math{
  \bar p_1'\tightin\TPOS{l_{0}}
  \und
  l_{0}/\bar p_1'\tightnotin\V
}:}}
\begin{diagram}
l_{0}\mu_{0}&&
&\rredindex{\omega+n+1,\,\bar p_1'}&&&w_1/\bar p_0
\\\dredindex{\omega+n+1,\,\emptyset}&&
&&&&\drefltransindex{\omega+n+1}
\\w_0/\bar p_0&\rrefltransindex\omega&\circ
&\ronlyonceindex{\omega+n+1}&\circ
&\rrefltransindex{\omega}&\circ
\end{diagram}
Let \math{\xi\in\SUBST\V\V} be a bijection with 
\bigmath{
  \xi[\VAR{\kurzregelindex{1}}]\cap\VAR{\kurzregelindex{0}}
  =
  \emptyset
.}
\\
Define
\bigmath{
  \Y
  :=
  \xi[\VAR{\kurzregelindex{1}}]\cup\VAR{\kurzregelindex{0}}
.}
\\
Let \math{\varrho\in\Xsubst} be given by
$\ x\varrho=
\left\{\begin{array}{@{}l@{}l@{}}
  x\mu_{0}        &\mbox{ if }x\in\VAR{\kurzregelindex{0}}\\
  x\xi^{-1}\mu_{1}&\mbox{ else}\\
\end{array}\right\}
\:(x\tightin\V)$.
\\
By
\math{
  l_{1}\xi\varrho
  \tightequal 
  l_{1}\xi\xi^{-1}\mu_{1}
  \tightequal 
  u/\bar p_1
  \tightequal 
  u/\bar p_0\bar p_1'
  \tightequal 
  l_{0}\mu_{0}/\bar p_1'  
  \tightequal 
  l_{0}\varrho/\bar p_1'
  \tightequal 
  (l_{0}/\bar p_1')\varrho
}
\\
let
\math{
  \sigma:=\minmgu{\{(l_{1}\xi}{l_{0}/\bar p_1')\},\Y}
}
and
\math{\varphi\in\Xsubst}
with
\math{
  \domres{\inpit{\sigma\varphi}}\Y
  \tightequal
  \domres\varrho\Y
}.
\\
If 
\bigmath{
  \repl{l_{0}}{\bar p_1'}{r_{1}\xi}\sigma
  \tightequal
  r_{0}\sigma
,}
then the proof is finished due to 
\\\LINEmath{
  w_0/\bar p_0
  \tightequal
  r_{0}\mu_{0}
  \tightequal
  r_{0}\sigma\varphi
  \tightequal
  \repl{l_{0}}{\bar p_1'}{r_{1}\xi}\sigma\varphi
  \tightequal
  \repl{l_{0}\mu_{0}}{\bar p_1'}{r_{1}\mu_{1}}
  \tightequal
  w_1/\bar p_0
.}
\\
Otherwise 
we have
\bigmath{
  (\,
   (\repl{l_{0}}{\bar p_1'}{r_{1}\xi},
    C_{1}\xi,    
    1),\penalty-1\,
   (r_{0},
    C_{0},
    1),\penalty-1\,
    l_{0},\penalty-1\,
    \sigma,\penalty-1\,
    \bar p_1'\,)
  \in{\rm CP}(\R)
}
(due to Claim~5);
\bigmath{\bar p_1'\tightnotequal\emptyset}
(due the global case assumption);
\bigmath{C_{1}\xi\sigma\varphi=C_{1}\mu_{1}}
is fulfilled \wrt\ \redindex{\omega+n};
\bigmath{C_{0}\sigma\varphi=C_{0}\mu_{0}}
is fulfilled \wrt\ \redindex{\omega+n}.
Since \RX\ is \math\omega-level confluent up to \math n
(by our induction hypothesis)
and \math\omega-shallow confluent up to \math\omega,
due to our assumed \math\omega-level anti-closedness
(matching the definition's \math{n} to our \math{n\tight+1})
we have
\bigmath{
  w_1/\bar p_0
  \tightequal
  \repl{l_{0}\mu_{0}}{\bar p_1'}{r_{1}\mu_{1}}
  \tightequal
  \repl{l_{0}}{\bar p_1'}{r_{1}\xi}\sigma\varphi
  \refltransindex{\omega+n+1}
  \circ
  \antirefltransindex{\omega}
  \tight\circ
  \antionlyonceindex{\omega+n+1}
  \tight\circ
  \antirefltransindex{\omega}
  r_{0}\sigma\varphi
  \tightequal
  r_{0}\mu_{0}
  \tightequal
  w_0/\bar p_0.
}
\\
\Qeddouble{``The critical peak case''}%
\QEDtriple{``There is some \math{\bar p_1'} with
\bigmath{\bar p_0\bar p_1'\tightequal\bar p_1} and
\bigmath{\bar p_1'\tightnotequal\emptyset}''}

\pagebreak

\yestop
\yestop
\noindent
\underline{\underline{\underline{%
There is some \math{\bar p_0'} with
\bigmath{\bar p_1\bar p_0'\tightequal\bar p_0}:%
}}}

\noindent
We have two cases:

\noindent
\underline{\underline{%
``The second variable overlap case'':%
}}
\\\LINEnomath{\underline{\underline{%
There are \math{x\tightin\V} and \math{p'}, \math{p''} such that
\math{
    l_{1}/p'
    \tightequal
    x
  \und
    p'p''
    \tightequal
    \bar p_0'
}:%
}}}
\begin{diagram}
l_{1}\mu_{1}&&&\rredindex{\omega+n+1,\,\emptyset}&&&w_1/\bar p_1
\\&&&&&&\dequal
\\\dredindex{\omega+n+1,\,\bar p_0'}&&&&&&r_{1}\mu_{1}
\\&&&&&&\dredparaindex{\omega+n+1}
\\w_0/\bar p_1&\requal&l_{1}\nu&&\rredindex{\omega+n+1}&&r_{1}\nu
\end{diagram}
\noindent
\underline{Claim~11a:}
We have 
\bigmath{x\mu_{1}/p''\tightequal l_{0}\mu_{0}}
and may assume
\bigmath{x\tightin\Vsig.}
\\\underline{Proof of Claim~11a:}
We have 
\bigmath{
  x\mu_{1}/p''
  \tightequal
  l_{1}\mu_{1}/p' p''
  \tightequal
  u/\bar p_1 p' p''
  \tightequal
  u/\bar p_1\bar p_0'
  \tightequal
  u/\bar p_0
  \tightequal
  l_{0}\mu_{0}
.}
\\
If \bigmath{x\tightin\Vcons,}
then \bigmath{x\mu_{1}\tightin\tcc,}
then \bigmath{x\mu_{1}/p''\tightin\tcc,}
then 
\\
\bigmath{l_{0}\mu_{0}\tightin\tcc,}
and then \bigmath{l_{0}\tightin\tcs}
which we may assume not to be the case by Claim~5.%
\QED{Claim~11a}

\noindent
\underline{Claim~11b:}
We can define \math{\nu\in\Xsubst} by
\bigmath{
  x\nu
  \tightequal
  \repl
    {x\mu_{1}}
    {p''}
    {r_{0}\mu_{0}}
}
and
\bigmath{
  \forall y\tightin\V\tightsetminus\{x\}\stopq y\nu\tightequal y\mu_{1}
.}
Then we have 
\bigmath{
  x\mu_{1}\redindex{\omega+n+1}x\nu
.}
\\
\underline{Proof of Claim~11b:}
This follows directly from Claim~11a.%
\QED{Claim~11b}

\noindent
\underline{Claim~12:}
\bigmath{
  w_0/\bar p_1
  \tightequal
  l_{1}\nu
.}
\\
\underline{Proof of Claim~12:}
\\
By the left-linearity assumption of our lemma and claims 5 and 11a 
we may 
assume
\bigmath{
  \setwith
    {p'''}
    {l_{1}/p'''\tightequal x}
  =
  \{p'\}
.}
Thus, by Claim~11b we get
\bigmath{
  w_0/\bar p_1
  \tightequal
  \repl 
    {u/\bar p_1}
    {\bar p_0'}
    {r_{0}\mu_{0}}
  \tightequal\\
  \repl
    {\replpar
       {l_{1}}
       {p'''}
       {y\mu_{1}}
       {l_{1}/p'''\tightequal y\tightin\V}
    }
    {\bar p_0'}
    {r_{0}\mu_{0}}
  \tightequal\\
  \repl
    {\repl
       {\replpar
          {l_{1}}
          {p'''}
          {y\mu_{1}}
          {l_{1}/p'''\tightequal y\tightin\V\und y\tightnotequal x}
       }
       {p'}
       {x\mu_{1}}
    }
    {p' p''}
    {r_{0}\mu_{0}}
  \tightequal\\
  \repl
    {\replpar
       {l_{1}}
       {p'''}
       {y\nu}
       {l_{1}/p'''\tightequal y\tightin\V\und y\tightnotequal x}
    }
    {p'}
    {\repl
       {x\mu_{1}}
       {p''}
       {r_{0}\mu_{0}}
    }
  \tightequal\\
  \replpar
    {l_{1}}
    {p'''}
    {y\nu}
    {l_{1}/p'''\tightequal y\tightin\V}
  \tightequal
  l_{1}\nu
.}\QED{Claim~12}

\noindent
\underline{Claim~13:}
\bigmath{
  r_{1}\nu
  \antiredparaindex{\omega+n+1}
  w_1/\bar p_1
.}
\\
\underline{Proof of Claim~13:} 
Since \bigmath{r_{1}\mu_{1}\tightequal w_1/\bar p_1,} 
this follows directly from Claim~11b.%
\QED{Claim~13}

\noindent
By claims 12 and 13 
using Corollary~\ref{corollary parallel one} 
it now suffices to show
\bigmath{
  l_{1}\nu
  \redindex{\omega+n+1}
  r_{1}\nu
,}
which again follows from 
Claim~11b,
Lemma~\ref{lemma invariance of fulfilledness level}
(matching 
 its \math{n_0} to our \math{n\tight+1} and 
 its \math{n_1} to our \math{n}),
and our induction hypothesis that \RX\ is \math\omega-level confluent up to
\math n.%
\\\Qeddouble{``The second variable overlap case''}

\pagebreak

\yestop
\noindent
\underline{\underline{``The second critical peak case'':
\math{
    \bar p_0'\tightin\TPOS{l_{1}}
  \und
    l_{1}/\bar p_0'\tightnotin\V
}:%
}}
\begin{diagram}
l_{1}\mu_{1}
&&&\rredindex{\omega+n+1,\,\emptyset}&&&w_1/\bar p_1
\\
\dredindex{\omega+n+1,\,\bar p_0'}
&&
&&
&&\drefltransindex{\omega+n+1}
\\
w_0/\bar p_1
&\rrefltransindex\omega&\circ
&\ronlyonceindex{\omega+n+1}&\circ
&\rrefltransindex{\omega}&\circ
\end{diagram}
Let \math{\xi\in\SUBST\V\V} be a bijection with 
\bigmath{
  \xi[\VAR{\kurzregelindex{0}}]\cap\VAR{\kurzregelindex{1}}
  =
  \emptyset
.}
\\
Define
\bigmath{
  \Y
  :=
  \xi[\VAR{\kurzregelindex{0}}]\cup\VAR{\kurzregelindex{1}}
.}
\\
Let \math{\varrho\in\Xsubst} be given by
$\ x\varrho=
\left\{\begin{array}{@{}l@{}l@{}}
  x\mu_{1}        &\mbox{ if }x\in\VAR{\kurzregelindex{1}}\\
  x\xi^{-1}\mu_{0}&\mbox{ else}\\
\end{array}\right\}
\:(x\tightin\V)$.
\\
By
\math{
  l_{0}\xi\varrho
  \tightequal 
  l_{0}\xi\xi^{-1}\mu_{0}
  \tightequal 
  u/\bar p_0
  \tightequal 
  u/\bar p_1\bar p_0'
  \tightequal 
  l_{1}\mu_{1}/\bar p_0'  
  \tightequal 
  l_{1}\varrho/\bar p_0'
  \tightequal 
  (l_{1}/\bar p_0')\varrho
}
\\
let
\math{
  \sigma:=\minmgu{\{(l_{0}\xi}{l_{1}/\bar p_0')\},\Y}
}
and
\math{\varphi\in\Xsubst}
with
\math{
  \domres{\inpit{\sigma\varphi}}\Y
  \tightequal
  \domres\varrho\Y
}.
\\
If 
\bigmath{
  \repl{l_{1}}{\bar p_0'}{r_{0}\xi}\sigma
  \tightequal
  r_{1}\sigma
,}
then the proof is finished due to 
\\\linemath{
  w_0/\bar p_1
  \tightequal
  \repl{l_{1}\mu_{1}}{\bar p_0'}{r_{0}\mu_{0}}
  \tightequal
  \repl{l_{1}}{\bar p_0'}{r_{0}\xi}\sigma\varphi
  \tightequal
  r_{1}\sigma\varphi
  \tightequal
  r_{1}\mu_{1}
  \tightequal
  w_1/\bar p_1.
}
Otherwise 
we have
\bigmath{
  (\,
   (\repl{l_{1}}{\bar p_0'}{r_{0}\xi},
    C_{0}\xi,    
    1),\penalty-1\,
   (r_{1},
    C_{1},
    1),\penalty-1\,
    l_{1},\penalty-1\,
    \sigma,\penalty-1\,
    \bar p_0'\,)
  \in{\rm CP}(\R)
}
(due to Claim~5);
\bigmath{C_{0}\xi\sigma\varphi=C_{0}\mu_{0}}
is fulfilled \wrt\ \redindex{\omega+n};
\bigmath{C_{1}\sigma\varphi=C_{1}\mu_{1}}
is fulfilled \wrt\ \redindex{\omega+n}.
Since \RX\ is \math\omega-level confluent up to \math n
(by our induction hypothesis) 
and \math\omega-shallow confluent up to \math\omega,
due to our assumed \math\omega-level strong joinability 
(matching the definition's \math{n} to our \math{n\tight+1})
we have
\bigmath{
  w_0/\bar p_1
  \tightequal
  \repl{l_{1}\mu_{1}}{\bar p_0'}{r_{0}\mu_{0}}
  \tightequal
  \repl{l_{1}}{\bar p_0'}{r_{0}\xi}\sigma\varphi
  \refltransindex{\omega}
  \tight\circ
  \onlyonceindex{\omega+n+1}
  \tight\circ
  \refltransindex{\omega}
  \circ
  \antirefltransindex{\omega+n+1}
  \penalty-1
  r_{1}\sigma\varphi
  \tightequal
  r_{1}\mu_{1}
  \tightequal
  w_1/\bar p_1
.}
\\\Qeddouble{``The second critical peak case''}
\end{proofparsepqed}

\begin{proof}{Lemma~\ref{lemma terminating reduction relation}}
Due to \Vmonotonicity\ of \math > and \bigmath{\tight >\subseteq\tight\rhd,}
it is easy to show by induction over \math\beta\ in \math\prec\ that
\bigmath{
  \forall\beta\tightpreceq\omega\tight+\alpha\stopq
    \redindex{\RX,\beta}\subseteq\tight\rhd
}
using \lemmamonotonicinbeta.
\end{proof}

\pagebreak

\begin{proofparsepqed}{Lemma~\ref
{lemma invariance of fulfilledness compatible}}
\underline{Claim~0:}
\bigmath{
  \forall u    \tightin\condterms C\stopq
  \forall\hat u\tightin\tsigX      \stopq
  \inparentheses{
      u\mu\refltrans\hat u
    \ \implies\ 
      u\nu\tight\downarrow\hat u
  }
.}
\\\underline{Proof of Claim~1:}
We get the following cases:
\\\underline{\math{l\mu\rhd u\mu}:}
\bigmath{
  u\nu\antirefltrans u\mu\refltrans\hat u
} 
implies
\bigmath{
  u\nu\tight\downarrow\hat u
} 
by the assumed confluence below \math{u\mu}.
\\\underline{\math{u\mu\tightnotin\DOM\red}:}
\bigmath{
  u\nu\antirefltrans u\mu\refltrans\hat u
} 
implies 
\bigmath{
  u\nu\tightequal u\mu\tightequal\hat u
.} 
\\{[\underline{\math{\VAR u\subseteq\Vcons}:}
By \lemmaconskeeping\ we get
\bigmath{
  \forall x\tightin\VAR u\stopq
    x\mu\refltransindex\omega x\nu
.}
Thus from 
\bigmath{
  u\nu
  \antirefltransindex\omega
  u\mu
  \refltrans
  \hat u
}
due to the assumed
\bigmath{
  \antirefltransindex{\RX,\omega}\circ\refltranssub
  \subseteq
  \tight\downarrow
}
we get
\bigmath{
  u\nu\tight\downarrow\hat u
.}]}%
\QED{Claim~0}

\noindent
By Lemma~\ref{lemma red is minimal}
it suffices to show that \math{C\nu} is fulfilled.
For each $L$ in \math C we have to show that
\math{L\nu}
is fulfilled.
Note that we already know that \math{L\mu} is fulfilled.
\\
\underline{$L=(u\boldequal v)$:}
There is some \math{\hat u} with
\bigmath{
  u\mu
  \refltrans
  \hat u
  \antirefltrans
  v\mu
.}
By Claim~0 there is some \math{\hat v}
with 
\bigmath{
  u\nu
  \refltrans
  \hat v
  \antirefltrans
  \hat u
  \antirefltrans
  v\mu
.}
Thus, by Claim~0 we get
\bigmath{
  u\nu
  \refltrans
  \hat v
  \tight\downarrow 
  v\nu
.}
\\\underline{$L=(\DEF u)$:}
We know the existence of
$\hat{u}\in\tgcons$
with
\bigmath{
  u\mu\refltrans\hat{u}
.}
By
Claim~0
we get
\bigmath{
  u\nu
  \refltrans
  u'
  \antirefltrans
  \hat u
} 
for some \math{u'}.
By 
Lemma~\ref{lemma about groundconskeeping}
we get \bigmath{u'\tightin\tgcons.}
\\
\underline{$L=(u\boldunequal v)$:}
We know the existence of
\bigmath{\hat{u},\hat{v}\in\tgcons}
with
\bigmath{ 
 u\mu
 \refltrans
 \hat u
 \notconflu
 \hat v
 \antirefltrans
 v\mu
.}
Just like above
we get
\math{u',v'\in\tgcons} with 
\bigmath{ 
 u\nu
 \refltrans
 u'
 \antirefltrans
 \hat u
}
and
\bigmath{
 \hat v
 \refltrans
 v'
 \antirefltrans
 v\nu
.}
Due to 
\bigmath{
 \hat u
 \notconflu
 \hat v
}
we finally get   
\bigmath{
 u'
 \notconflu
 v'
.}
\end{proofparsepqed}

\vfill

\begin{proofparsepqed}{Lemma~\ref{lemma quasi-free three}}
First notice that the usual modularization of the proof for the unconditional
analogue of the theorem (by showing first that local confluence is guaranteed
except for the cases that are matched by critical peaks 
(the so-called ``critical pair lemma'')) 
is not possible here because we need the 
confluence property to hold
for the condition terms even for the cases 
that are not matched by critical peaks.
Now to the proof:
For all \math{s\in\tsigX}
we are going to prove confluence below \math s by induction over \math s
in \math\lhd. 
Let $s$ be minimal in \math\lhd\ 
such that \red\ is not confluent below $s$.
Because of \bigmath{\red\subseteq\rhd} 
(by Lemma~\ref{lemma terminating reduction relation})
and minimality of \math s,
\red\ is not even locally confluent below \math s\@.
Let
\math{p,q\in\TPOS s};
\bigmath{t_{0}\antiredindex{\omega+\omega,p}s\redindex{\omega+\omega,q}t_{1};}
\bigmath{t_{0}\notconflu t_{1}.}
Now as one of $p,q$ must be a prefix of the other,
\wrog\ say that $q$ is a prefix of
$p$.
As
\bigmath{s\unrhd s/q,}
by the minimality of $s$ we have 
\bigmath{q\tightequal\emptyset.}
We start a second level of induction on 
\math p in \math{\lll_s}.
Thus assume that \math p is minimal
such that there are
\math{p\in\TPOS s}
and
\math{t_0,t_1\in\tsigX}
with 
\bigmath{
  t_{0}
  \antiredindex{\omega+\omega,p}
  s
  \redindex{\omega+\omega,\emptyset}
  t_{1}
}
and
\bigmath{t_{0}\notconflu t_{1}.}

Now for $k<2$ there must be
\math{((l_{k},r_{k}),C_{k})\in\R};
\math{\mu_{k}\in\Xsubst};
with
\bigmath{C_{k}\mu_{k}} fulfilled;
\bigmath{s\tightequal l_{1}\mu_{1};}
\bigmath{s/p\tightequal l_{0}\mu_{0};}
\bigmath{t_{0}\tightequal \repl{l_{1}\mu_{1}}{p}{r_{0}\mu_{0}};}
\bigmath{t_{1}\tightequal r_{1}\mu_{1}.}
Moreover, for $k<2$ we define
\bigmath{
  \Lambda_k
  :=
  \left\{\arr{{ll}
      0&\mbox{ if }l_k\tightin\tcs
    \\1&\mbox{ otherwise}
  }\right\}
.}

\pagebreak

\yestop
\noindent
\underline{Claim~0:}
We may assume that 
\bigmath{
  \forall q\tightin\TPOS s\stopq
  \inparentheses{
      \emptyset
      \tightnotequal
      {q}
      \lll_s
      p
    \ \implies\ 
      s\tightnotin\DOM{\redindex{\omega+\omega,q}}
  }
.}
\\
\underline{Proof of Claim~0:}
Otherwise there must be some
\math{q\in\TPOS s};
\math{\kurzregelindex2\in\R};
\math{\mu_2\in\Xsubst};
with
\math{C_2\mu_2} fulfilled;
\bigmath{s/q\tightequal l_2\mu_2;}
and
\bigmath{
      \emptyset
      \tightnotequal
      {q}
      \lll_s
      p
.}
By our second induction level we get
\bigmath{
  \repl
    {l_1\mu_1}
    {q}
    {r_2\mu_2}
  \refltrans
  w_1
  \antirefltrans
  r_1\mu_1
}
for some \math{w_1};
\cf\ the diagram below.
Next we are going to show that there is some \math{w_0} with
\bigmath{
  \repl{l_1\mu_1}{p}{r_0\mu_0}
  \refltrans
  w_0
  \antirefltrans
  \repl{l_1\mu_1}{q}{r_2\mu_2}
.}
Note that 
(since \bigmath{\red\subseteq\rhd} implies
\math{s\tight\rhd\repl{l_1\mu_1}{q}{r_2\mu_2}})
this finishes the proof of Claim~0 since then 
\bigmath{
  w_0
  \antirefltrans
  \repl{l_1\mu_1}{q}{r_2\mu_2}
  \refltrans
  w_1
}
by our first level of induction
implies the contradictory 
\bigmath{
  t_0
  \refltrans
  w_0
  \tight\downarrow
  w_1
  \antirefltrans
  t_1
.}
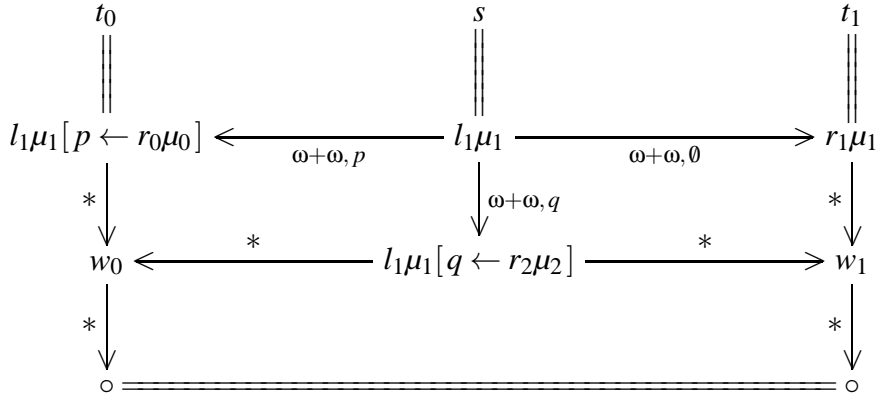
\begin{diagram}
t_0&&&&s&&&&t_1
\\
\dequal&&&&\dequal&&&&\dequal
\\
\repl{l_1\mu_1}{p}{r_0\mu_0}
&&\rantiredindex{\omega+\omega,\,p}
&&l_1\mu_1&&\rredindex{\omega+\omega,\,\emptyset}&&r_1\mu_1
\\
\drefltrans&&&&\dredindex{\omega+\omega,\,q}&&&&\drefltrans
\\
w_0&&\rantirefltrans&&\repl{l_1\mu_1}{q}{r_2\mu_2}&&\rrefltrans&&w_1
\\
\drefltrans&&&&&&&&\drefltrans
\\
\circ&&&&\requal&&&&\circ
\\
\end{diagram}
\noindent
In case of \bigmath{\neitherprefix{p}{q}} we simply can choose
\bigmath{
  w_0
  :=
  \repl
    {\repl{l_1\mu_1}{p}{r_0\mu_0}}
    {q}
    {r_2\mu_2}
.}
Otherwise, 
there must be some \math{\bar p}, \math{\hat p}, \math{\hat q},
with
\bigmath{
  p
  \tightequal
  \bar p\hat p
,}
\bigmath{
  q
  \tightequal
  \bar p\hat q
,}
and
\bigmath{
  \inparenthesesinlinetight{
      \hat p 
      \tightequal
      \emptyset
    \oder
      \hat q 
      \tightequal
      \emptyset
  }
.}
Now it suffices to show
\\\linemath{
  \repl
    {s/\bar p}
    {\hat p}
    {r_0\mu_0}
  \refltrans
  w_0'
  \antirefltrans
  \repl
    {s/\bar p}
    {\hat q}
    {r_2\mu_2}
}
for some \math{w_0'}, because by \monotonicity\ of \refltrans\ we then have
\\\math{
  \repl
    {l_1\mu_1}
    {p}
    {r_0\mu_0}
  \tightequal
  \repl
    {s}
    {\bar p\hat p}
    {r_0\mu_0}
  \tightequal
  \repl
    {\repl
       {s}
       {\bar p}
       {s/\bar p}
    }
    {\bar p\hat p}
    {r_0\mu_0}
  \tightequal
  \repl
    {s}
    {\bar p}
    {\repl
       {s/\bar p}
       {\hat p}
       {r_0\mu_0}
    }
  \refltrans\\
  \repl
    {s}
    {\bar p}
    {w_0'}
  \\\antirefltrans
  \repl
    {s}
    {\bar p}
    {\repl
       {s/\bar p}
       {\hat q}
       {r_2\mu_2}
    }
  \tightequal
  \repl
    {\repl
       {s}
       {\bar p}
       {s/\bar p}
    }
    {\bar p\hat q}
    {r_2\mu_2}
  \tightequal
  \repl
    {s}
    {\bar p\hat q}
    {r_2\mu_2}
  \tightequal
  \repl
    {l_1\mu_1}
    {q}
    {r_2\mu_2}
.}\\
Note that
\\\linemath{
  \repl
    {s/\bar p}
    {\hat p}
    {r_0\mu_0}
  \antiredindex{\omega+\omega,\hat p}
  s/\bar p
  \redindex{\omega+\omega,\hat q}
  \repl
    {s/\bar p}
    {\hat q}
    {r_2\mu_2}
.}
In case of \bigmath{\bar p\tightnotequal\emptyset}
(since then \bigmath{\superterm\subseteq\rhd} 
 implies \math{s\rhd s/\bar p})
we get some \math{w_0'} with
\bigmath{
  \repl
    {s/\bar p}
    {\hat p}
    {r_0\mu_0}
  \refltrans
  w_0'
  \antirefltrans
  \repl
    {s/\bar p}
    {\hat q}
    {r_2\mu_2}
}
by our first level of induction.
Otherwise, in case of \bigmath{\bar p\tightequal\emptyset,}
our disjunction from above means
\bigmath{
  \inparenthesesinlinetight{
      p 
      \tightequal
      \emptyset
    \oder
      q 
      \tightequal
      \emptyset
  }
.}
Since we have
\bigmath{
      \emptyset
      \tightnotequal
      q 
}
by our initial assumption, we may assume  
\bigmath{
      q
      \tightequal
      \hat q 
      \tightnotequal
      \emptyset
}
and
\bigmath{
      p 
      \tightequal
      \hat p 
      \tightequal
      \bar p
      \tightequal
      \emptyset
.}
Then the above divergence reads
\bigmath{
  \repl
    {s/\bar p}
    {\hat p}
    {r_0\mu_0}
  \antiredindex{\omega+\omega,\emptyset}
  s
  \redindex{\omega+\omega,q}
  \repl
    {s/\bar p}
    {\hat q}
    {r_2\mu_2}
}
and we get the required joinability by our second induction level due to 
\bigmath{
  q\lll_s p
.}%
\QED{Claim~0}

\vfill

\pagebreak

\yestop
\noindent
\underline{Claim~1:}
In case of 
\bigmath{
  \antiredindex\omega
  \circ
  \red
  \subseteq
  \tight\downarrow
}
we may assume \bigmath{s\tightnotin\DOM{\redindex\omega}.}
\\\underline{Proof of Claim~1:}
Assume 
\bigmath{
  \antiredindex\omega
  \circ
  \red
  \subseteq
  \tight\downarrow
.}
If there is a \math{t_2} with \bigmath{s\redindex\omega t_2} 
then we get some \math{t_0'}, \math{t_1'} with
\bigmath{
  t_0\refltrans t_0'\antirefltrans t_2\refltrans t_1'\antirefltrans t_1
.}
Due \bigmath{\redindex\omega\subseteq\red\subseteq\tight\rhd} 
by our first level
of induction we get the contradictory
\bigmath{
  t_0\refltrans t_0'\downarrow t_1'\antirefltrans t_1
.}%
\QED{Claim~1}

\vfill

\yestop
\noindent
\underline{Claim~2:}
In case of 
\bigmath{
  \antiredindex\omega\circ\red\subseteq\tight\downarrow
}
for each \math{k\prec 2} we may assume:
\\
\bigmath{l_k\mu_k\tightnotin\tcs}
and
\\\mbox{}\hfill\bigmath{
  \inparenthesesinline{
      l_k\tightnotin\tcs
    \oder
      \condterms{C_k\mu_k}\tightnotsubseteq\tcs
  }
.} 
\\\underline{Proof of Claim~2:}
By \lemmaconskeeping\ 
and
\bigmath{l_k\mu_k\redsimple r_k\mu_k,} 
\bigmath{l_k\mu_k\tightin\tcs}
implies 
\bigmath{
  {l_k\mu_k}\redindex{\omega}
  {r_k\mu_k}
} 
which we may assume not to be the case by Claim~1.
In case of 
\bigmath{l_k\tightin\tcs} and 
\bigmath{\condterms{C_k\mu_k}\tightsubseteq\tcs}
by \lemmaconskeeping\
\bigmath{C_k\mu_k} is fulfilled \wrt\ \redindex{\omega}
and then Corollary~\ref{corollary redsubomega is minimum} implies
\bigmath{
  {l_k\mu_k}
  \redindex{\omega}
  {r_k\mu_k}
}
again,
which we may assume not to be the case by Claim~1.%
\QED{Claim~2}

\vfill

\yestop
\noindent
Now we have two cases:

\vfill

%%%%%%%%%%%%%%%%%%%%%%%%%%%%%%%%%%%%%%%%%%%%%%%%%%%%%%%%%%%%%%%%%%%%%%%%%%%%%%%
\noindent
\underline{\underline{The variable overlap case:
\bigmath{p\tightequal q_{0}q_{1};\ l_{1}/q_{0}\tightequal x\in\V}:}}
\\
We have
\bigmath{
  x\mu_{1}/q_{1}
  \tightequal
  l_{1}\mu_{1}/q_{0}q_{1}
  \tightequal 
  s/p
  \tightequal
  l_{0}\mu_{0}
.}
%\\\headroom
%\underline{Claim~A:}
%In case of 
%\bigmath{
%  \antired\circ\redindex{\omega}
%  \subseteq
%  \tight\downarrow
%}
%we may assume \bigmath{x\in\Vsig}.
%\\\underline{Proof of Claim~A:}
%Otherwise we would have \math{x\tightin\Vcons}, which implies 
%\linebreak
%\math{x\mu_{1}\tightin\tcc} and then \math{l_0\mu_0\tightin\tcc}.
%We can assume that this is not the
%case by Claim~3.%
%\QED{Claim~A}
%
%\noindent
By
Lemma~\ref{lemma about sortkeeping}
(in case of $x\tightin\Vcons$), 
we can define $\nu\in\Xsubst$ by
($y\tightin\V$): 
\\\math{y\nu:=
 \left\{\begin{array}{ll}
 \repl{x\mu_{1}}{q_{1}}{r_{0}\mu_{0}}&\mbox{if } y\tightequal x\\
 y\mu_{1}&\mbox{otherwise}\\
 \end{array}\right\}
} and get 
\bigmath{y\mu_{1}\onlyonce y\nu}
for
\math{y\tightin\V}.
By Corollary~\ref{corollary monotonic}:\\
\bigmath{t_0\tightequal \repl{l_1\mu_1}{q_0q_1}{r_{0}\mu_{0}}\tightequal 
  \replpar{\repl{l_1}
               {q_0}
               {x\nu}}
          {q'}
          {y\mu_{1}}
          {l_{1}/q'\tightequal y\in\V\und q'\tightnotequal q_{0}}
  \ \refltrans
}\\
\bigmath{
  \replpar
    {l_{1}}
    {q'}
    {y\nu}
    {l_{1}/q'\tightequal y\tightin\V}
  \tightequal
  l_{1}\nu
;}\\
\bigmath{
  t_{1}\tightequal r_{1}\mu_{1}\refltrans r_{1}\nu
.}
It suffices to show
\bigmath{l_1\nu\redsimple r_1\nu,}
which follows from Lemma~\ref{lemma invariance of fulfilledness compatible}
because of 
{[\math{\antirefltransindex\omega\circ\refltrans\subseteq\tight\downarrow},]}
\bigmath{l_1\mu_1\tightequal s} and our first level of induction.%
\QEDdouble{The variable overlap case}

\vfill

\pagebreak

\noindent
\underline{\underline{The critical peak case:
\bigmath{p\in\TPOS{l_{1}};\ l_{1}/p\not\in\V}:}}
Let \math{\xi\in\SUBST\V\V} be a bijection 
with 
\\
\bigmath{\xi[\VAR{\sugarregelindex0}]\cap\VAR{\sugarregelindex1}\tightequal \emptyset.}
Define
\bigmath{\Y:=\VAR{(\sugarregelindex0)\xi,\sugarregelindex1}}\@.
\\
Let $\varrho$ be given by
$\ x\varrho\tightequal 
\left\{\begin{array}{@{}l@{}l@{}}
  x\mu_{1}        &\mbox{ if }x\in\VAR{\sugarregelindex1}\\
  x\xi^{-1}\mu_{0}&\mbox{ else}\\
\end{array}\right\}
\:(x\tightin\V)$.
By
\bigmath{l_{0}\xi\varrho\tightequal l_{0}\xi\xi^{-1}\mu_{0}\tightequal s/p
\tightequal l_{1}\mu_{1}/p\tightequal l_{1}\varrho/p\tightequal (l_{1}/p)\varrho}
let
\bigmath{
  \sigma:=\minmgu{\{(l_{0}\xi}{l_{1}/p)\},\Y}
}
and
\bigmath{\varphi\in\Xsubst}
\linebreak
with
\math{
  \domres{\inpit{\sigma\varphi}}\Y
    \tightequal\domres\varrho\Y
}\@.

\noindent
\underline{Claim~A:}
We may assume
\bigmath{
  \inparentheses{
      p\tightequal\emptyset
    \ \oder\ 
      \forall y\tightin\VAR{l_1}\stopq
         y\sigma\varphi\tightnotin\DOM\red
  }
.}
\\\underline{Proof of Claim~A:}
Otherwise, when 
\bigmath{ 
      p\tightnotequal\emptyset
} holds
but
\bigmath{
      \forall y\tightin\VAR{l_1}\stopq
         y\sigma\varphi\tightnotin\DOM\red
}
is not the case, there are some
\math{x\in\VAR{l_1},}
\math{\nu\tightin\Xsubst}  
with \bigmath{x\sigma\varphi\redsimple x\nu}
and 
\bigmath{
  \forall y\tightin\V\tightsetminus\{x\}\stopq
    y\mu_1\tightequal y\nu
.}
%In case of 
%\bigmath{
%  \antired\circ\redindex{\omega}
%  \subseteq
%  \tight\downarrow
%}
%we may assume \bigmath{x\tightin\Vsig}
%due to Claim~1 and \lemmaconskeeping.
Due to 
\bigmath{
  l_1\mu_1/p
  \lhd 
  l_1\mu_1
  \tightequal
  s
} 
by our first level of induction
from 
\bigmath{
  r_0\xi\sigma\varphi
  \antired
  l_0\xi\sigma\varphi
  \tightequal
  l_1\sigma\varphi/p
  \tightequal
  l_1\mu_1/p
  \refltrans
  l_1\nu/p
}
we know that there must be some \math u with
\bigmath{
  r_0\xi\sigma\varphi
  \refltrans
  u
  \antirefltrans
  l_1\nu/p
.}
Due to 
\bigmath{
  l_1\mu_1
  \trans
  l_1\nu
}
and
\bigmath{\red\nottight\subseteq\rhd} 
we get 
\bigmath{
  l_1\nu
  \lhd 
  l_1\mu_1
  \tightequal
  s
.}
Thus, by our first level of induction, from
\bigmath{
  \repl
    {l_1\nu}
    {p}
    {u}
  \antirefltrans
  l_1\nu
  \red
  r_1\nu
}
(which is due to Lemma~\ref{lemma invariance of fulfilledness compatible}, 
{[\math{\antirefltransindex\omega\circ\refltrans\subseteq\tight\downarrow},]}
\bigmath{l_1\mu_1\tightequal s} and our first level of induction)
we get
\bigmath{
  t_0
  \tightequal
  \repl
    {l_1\mu_1}
    {p}
    {r_0\xi\sigma\varphi}
  \refltrans
  \repl
    {l_1\nu}
    {p}
    {r_0\xi\sigma\varphi}
  \refltrans
  \repl
    {l_1\nu}
    {p}
    {u}
  \downarrow
  r_1\nu
  \antirefltrans
  r_1\mu_1
  \tightequal
  t_1
.}\QED{Claim~A}

\noindent
If 
\bigmath{
  \repl  
    {l_1}
    {p}
    {r_0\xi}
  \sigma
  \tightequal
  r_1 
  \sigma
,} 
then we are finished due to
\bigmath{
  t_0
  \tightequal
  \repl  
    {l_1}
    {p}
    {r_0\xi}
  \sigma
  \varphi
  \tightequal
  r_1 
  \sigma
  \varphi
  \tightequal
  t_1
.} 
Otherwise
\bigmath{
  ((\repl
      {l_1}
      {p}
      {r_0\xi}
    ,
    C_0\xi,
    \Lambda_0),\ 
   (r_1,
    C_1,
    \Lambda_1),\
   l_1,\
   \sigma,\
   p\
  )
}
is a critical peak in \math{{\rm CP}(\R)}.

\noindent
Now
\bigmath{
  (C_0\xi\,C_1)\sigma\varphi
  \tightequal
  C_0\mu_0\,C_1\mu_1
}
is fulfilled \wrt\ \red.
Due to  
\bigmath{
  l_1\sigma\varphi
  \tightequal 
  l_1\varrho
  \tightequal 
  l_1\mu_1
  \tightequal 
  s
,} 
by our first level of induction we get
\bigmath{
  \forall u\lhd l_1\sigma\varphi\stopq
    \inparenthesesinlinetight{\red\mbox{ is confluent below }u}
.}
{[By Claim~1 we get \math{l_1\sigma\varphi\tightnotin\DOM{\redindex\omega}.}]}
By Claim~0 we get
\bigmath{
  \forall q\tightin\TPOS{l_1\sigma\varphi}\stopq
  \inparentheses{
      \emptyset
      \tightnotequal
      {q}
      \lll_{l_1\sigma\varphi}
      p
    \ \implies\ 
      l_1\sigma\varphi\tightnotin\DOM{\redindex{\omega+\omega,q}}
  }
.}
This means \bigmath{l_1\sigma\varphi\tightnotin A(p).}
{[Define \math{D_0:=C_0\xi} and \math{D_1:=C_1}.
If \bigmath{\Lambda_k\tightequal 0} for some \math{k\prec 2}, then
\bigmath{l_k\tightin\tcs,} which by Claim~2 implies
\bigmath{\condterms{D_k\sigma\varphi}\tightnotsubseteq\tcs,}
and then 
\bigmath{\condterms{D_k\sigma}\tightnotsubseteq\tcc.}]}
Thus, in case of
\bigmath{
  \forall y\tightin\V\stopq
    y\varphi\tightnotin\DOM\red
,}
by Claim~A and
the assumed \math\rhd-weak joinability \wrt\ \RX\ besides \math A  
we get 
\bigmath{
  t_0
  \tightequal
  \repl  
    {l_1}
    {p}
    {r_0\xi}
  \sigma
  \varphi
  \downarrow
  r_1 
  \sigma
  \varphi
  \tightequal
  t_1
.}

\noindent
Otherwise, when \bigmath{
  \forall y\tightin\V\stopq
    y\varphi\tightnotin\DOM\red
}
is not the case,
by \bigmath{\red\subseteq\tight\rhd} and the Axiom of Choice
there is some \math{\varphi'\tightin\Xsubst}
with
\bigmath{
  \forall y\tightin\V\stopq
    y\varphi\refltrans y\varphi'\tightnotin\DOM\red
.}
Then, of course,
\bigmath{
  \forall y\tightin\V\stopq
    y\xi\sigma\varphi
    \refltrans
    y\xi\sigma\varphi'
}
and
\bigmath{
  \forall y\tightin\V\stopq
    y\sigma\varphi
    \refltrans
    y\sigma\varphi'
.} 
By Lemma~\ref{lemma invariance of fulfilledness compatible} 
(due to 
{[\math{\antirefltransindex\omega\circ\refltrans\subseteq\tight\downarrow};]}
\bigmath{l_0\xi\sigma\varphi,l_1\sigma\varphi\subtermeq s;} 
\bigmath{\subtermeq\subseteq\tight\trianglelefteq;}
and our first level of induction)
we know that
\bigmath{C_0\xi\sigma\varphi'}
and
\bigmath{C_1\sigma\varphi'}
are fulfilled.
Furthermore, we have 
\bigmath{
  \repl
    {l_1}
    {p}
    {r_0\xi}
  \sigma\varphi
  \refltrans
  \repl
    {l_1}
    {p}
    {r_0\xi}
  \sigma\varphi'
}
and 
\bigmath{
  r_1\sigma\varphi'
  \antirefltrans 
  r_1\sigma\varphi
.}
Therefore, in case of \bigmath{l_1\sigma\varphi\tightequal l_1\sigma\varphi'}
the proof succeeds like above with \math{\varphi'} 
instead of \math{\varphi}.
Otherwise we have \bigmath{l_1\sigma\varphi\trans l_1\sigma\varphi'.} 
Then due to \bigmath{\red\nottight\subseteq\rhd} we get
\bigmath{
  s
  \tightequal
  l_1\sigma\varphi
  \rhd
  l_1\sigma\varphi'
.}
Therefore, by our first level of induction,
from
\bigmath{
  \repl
    {l_1}
    {p}
    {r_0\xi}
  \sigma\varphi'
  \antired
  \repl
    {l_1}
    {p}
    {l_0\xi}
  \sigma\varphi'
  \tightequal
  l_1
  \sigma\varphi'
  \red
  r_1
  \sigma\varphi'
}
(which is due to 
{[\math{\antirefltransindex\omega\circ\refltrans\subseteq\tight\downarrow};]}
\bigmath{l_0\xi\sigma\varphi,l_1\sigma\varphi\subtermeq s;} 
\bigmath{\subtermeq\subseteq\tight\trianglelefteq;}
and our first level of induction)
we conclude
\bigmath{
  t_0
  \tightequal
  \repl
    {l_1}
    {p}
    {r_0\xi}
  \sigma\varphi
  \refltrans
  \repl
    {l_1}
    {p}
    {r_0\xi}
  \sigma\varphi'
  \downarrow
  r_1
  \sigma\varphi'
  \antirefltrans
  r_1
  \sigma\varphi
  \tightequal
  t_1
.}%
\QEDdouble{The critical peak case}
\end{proofparsepqed}

\pagebreak

\begin{proofparsepqed}
{Lemma~\ref{lemma quasi-free} and Lemma~\ref{lemma quasi-free two and zero}}
Since the proofs of the two lemmas are very similar, we treat them
together, indicating the differences where necessary and using `\math\alpha'
to denote \math\omega\ in the proof of Lemma~\ref{lemma quasi-free}.

\yestop
\noindent
For \math{(\delta,s)\preclhdeq(\beta,\hat s)} 
we are going to show that \RX\
is \math\alpha-shallow confluent up to \math\delta\ and \math{s} 
in \math\lhd\ 
by induction over \math{(\delta,s)} in \math\preclhd\@.
Suppose 
that for
\math{n_0,n_1\prec\omega}
we have
\bigmath{
  (n_0\plusalpha n_1,s)
  \preclhdeq
  (\beta,\hat s)
}
and
\bigmath{
 t_0'
 \antirefltransindex{\alpha+n_0}
 s
 \refltransindex    {\alpha+n_1}
 t_1'
.}
We have to show 
\bigmath{
  t_0'
  \refltransindex    {\alpha+n_1}
  \circ
  \antirefltransindex{\alpha+n_0}
  t_1'
.}

In case of \bigmath{\exists i\tightprec 2\stopq t_i'\tightequal s}
this is trivially true. 

Thus, 
for
\linemath{
 t_0'
 \antirefltransindex{\alpha+n_0}
 t_0
 \antiredindex      {\alpha+n_0,p}
 s
 \redindex          {\alpha+n_1,q}
 t_1
 \refltransindex    {\alpha+n_1}
 t_1'
}
using the induction hypothesis that
\\\LINEnomath{
  \headroom\footroom
  \math{\forall(\delta,w')\preclhd(n_0\plusalpha n_1,s)\stopq}
  \RX\ is 
  \math\alpha-shallow confluent up to \math\delta\ and \math{w'} in \math\lhd
}
\\
we have to show 
\\\LINEmath{
  t_0'
  \refltransindex    {\alpha+n_1}
  \circ
  \antirefltransindex{\alpha+n_0}
  t_1'
.}

\yestop
\noindent
Note that due to Lemma~\ref{lemma terminating reduction relation} 
we have \bigmath{\redindex{\omega+\alpha}\subseteq\tight\rhd.}

\yestop
\noindent
\underline{Claim~0:}
Now it is sufficient to show
\bigmath{
  t_0
  \refltransindex    {\alpha+n_1}
  u
  \antirefltransindex{\alpha+n_0}
  t_1
}
for some \math u.
\\\underline{Proof of Claim~0:}
Due to
\bigmath{\redindex{\omega+\alpha}\subseteq\tight\rhd}
we have 
\bigmath{
  s\rhd t_0, t_1
.}
Thus by our induction hypotheses 
\bigmath{
  u
  \antirefltransindex{\alpha+n_0}
  t_1
  \refltransindex    {\alpha+n_1}
  t_1'
}
(\cf\ diagram below)
implies
the existence of some \math v with
\bigmath{
  u
  \refltransindex    {\alpha+n_1}
  v
  \antirefltransindex{\alpha+n_0}
  t_1'
} 
and then
\bigmath{
  t_0'
  \antirefltransindex{\alpha+n_0}
  t_0
  \refltransindex    {\alpha+n_1}
  v
}
implies
\bigmath{
  t_0'
  \refltransindex    {\alpha+n_1}
  \circ
  \antirefltransindex{\alpha+n_0}
  v
.}
\notop
\begin{diagram}
  s&\rredindex{\alpha+n_1}&t_1
&\rrefltransindex{\alpha+n_1}&t_1'
\\\dredindex{\alpha+n_0}&&\drefltransindex{\alpha+n_0}
&&\drefltransindex{\alpha+n_0}
\\t_0&\rrefltransindex{\alpha+n_1}&u
&\rrefltransindex{\alpha+n_1}&v
\\\drefltransindex{\alpha+n_0}&&
&&\drefltransindex{\alpha+n_0}
\\t_0'&&\rrefltransindex{\alpha+n_1}
&&\circ
\end{diagram}
\notop
\notop
\notop
\notop

\mbox{}\QED{Claim~0}

\yestop
\noindent
In case of \bigmath{\neitherprefix p q} we have
\bigmath{
  t_0
  /
  q
  =
  \repl{s}
       {p}
       {t_0/p}
  /
  q
  =
  s
  /
  q
}
and
\bigmath{
  t_1
  /
  p
  =
  \repl{s}
       {q}
       {t_1/q}
  /
  p
  =
  s
  /
  p
}
and therefore
\bigmath{
  t_0
  \redindex{\alpha+n_1,q}
  \repl
    {\repl s p{t_0/p}}
    {q}   
    {t_1/q}   
  \antiredindex{\alpha+n_0,p}
  t_1
,}
\ie\ our proof is finished.
Otherwise one of \math{p,q} must be a prefix of the other, \wrog\ say that
\math q is a prefix of \math p.
In case of \bigmath{q\tightnotequal\emptyset}
due to \bigmath{\superterm\subseteq\tight\rhd} we get 
\bigmath{s/q\lhd s} and the proof finished by our induction hypothesis
and \monotonicity\ of \refltransindex{\alpha+n_k}.
Thus we may assume \bigmath{q\tightequal\emptyset.}
We start a second level of induction on 
\math p in \math{\lll_s}.
Thus we may assume 
the following induction hypothesis:
\\\math{
  \forall q\tightin\TPOS s\stopq
  \forall t_0',t_1'      \stopq
  \forall n_0',n_1'      \stopq
}\\\LINEmath{
  \inparentheses{
      \inparenthesesoplist{
           t_0'
           \antiredindex{\alpha+n_0',q}
           s
           \redindex{\alpha+n_1',\emptyset}
           t_1'
         \oplistund
           n_0'\plusalpha n_1'
           \preceq
           n_0\plusalpha n_1
         \oplistund
           q\lll_s p
      }
    \ \implies\ 
      t_0'
      \refltransindex{\alpha+n_1'}
      \circ
      \antirefltransindex{\alpha+n_0'}
      t_1'
  }
}

\yestop
\noindent
Now for \math{k\prec 2} there must be
\math{((l_{k},r_{k}),C_{k})\in\R};
\math{\mu_{k}\in\Xsubst};
with
\bigmath{C_{k}\mu_{k}} fulfilled \wrt\ \redindex{\alpha+(n_k\monus 1)};
\bigmath{s\tightequal l_{1}\mu_{1};}
\bigmath{s/p\tightequal l_{0}\mu_{0};}
\bigmath{t_{0}\tightequal \repl{l_{1}\mu_{1}}{p}{r_{0}\mu_{0}};}
\bigmath{t_{1}\tightequal r_{1}\mu_{1};}
and
\bigmath{
  \Lambda_k
  \tightpreceq
  n_k
}
and
\bigmath{
  \alpha\tightequal 0
  \implies
  \inparenthesesoplist{
      1\tightpreceq n_k
    \oplistund
      \Lambda_k\tightequal 0
  }
}
for 
\bigmath{
  \Lambda_k
  :=
  \left\{\arr{{ll}
      0&\mbox{ if }l_k\tightin\tcs
    \\1&\mbox{ otherwise}
  }\right\}
.}

\pagebreak

\yestop
\noindent
\underline{Claim~1:}
We may assume that 
\math{
  \ 
  \forall q\tightin\TPOS s\stopq
  \inparentheses{
          \emptyset
          \tightnotequal
          {q}
          \lll_s
          p
    \ \implies\ 
      s\tightnotin\DOM{\redindex{\alpha+\min\{n_0,n_1\},q}}
  }
.}
\\
\underline{Proof of Claim~1:}
Otherwise there must be some
\math{q\in\TPOS s};
\math{\kurzregelindex2\in\R};
\math{\mu_2\in\Xsubst};
with
\math{C_2\mu_2} fulfilled \wrt\ \redindex{\alpha+(\min\{n_0,n_1\}\monus 1)};
\bigmath{s/q\tightequal l_2\mu_2;}
\bigmath{
      \emptyset
      \tightnotequal
      {q}
      \lll_s
      p
.}
By our second induction level we get
\bigmath{
  \repl
    {l_1\mu_1}
    {q}
    {r_2\mu_2}
  \refltransindex{\alpha+n_1}
  w_1
  \antirefltransindex{\alpha+\min\{n_0,n_1\}}
  r_1\mu_1
}
for some \math{w_1};
\cf\ the diagram below.
Next we are going to show that there is some \math{w_0} with
\bigmath{
  \repl{l_1\mu_1}{p}{r_0\mu_0}
  \refltransindex{\alpha+\min\{n_0,n_1\}}
  w_0
  \antirefltransindex{\alpha+n_0}
  \repl{l_1\mu_1}{q}{r_2\mu_2}
.}
Note that 
(since \bigmath{\redindex{\omega+\alpha}\subseteq\tight\rhd} implies
\math{s\tight\rhd\repl{l_1\mu_1}{q}{r_2\mu_2}})
this finishes the proof since then 
\bigmath{
  w_0
  \antirefltransindex{\alpha+n_0}
  \repl{l_1\mu_1}{q}{r_2\mu_2}
  \refltransindex{\alpha+n_1}
  w_1
}
by our first level of induction
implies 
\\\LINEmath{
  t_0
  \refltransindex{\alpha+\min\{n_0,n_1\}}
  w_0
  \refltransindex{\alpha+n_1}
  \circ
  \antirefltransindex{\alpha+n_0}
  w_1
  \antirefltransindex{\alpha+\min\{n_0,n_1\}}
  t_1
.}
\begin{diagram}
t_0&&&&s&&&&t_1
\\
\dequal&&&&\dequal&&&&\dequal
\\
\repl{l_1\mu_1}{p}{r_0\mu_0}
&&\rantiredindex{\alpha+n_0,\,p}
&&l_1\mu_1&&\rredindex{\alpha+n_1,\,\emptyset}&&r_1\mu_1
\\
\drefltransindex{\alpha+\min\{n_0,n_1\}}
&&&&\dredindex{\alpha+\min\{n_0,n_1\},\,q}
&&&&\drefltransindex{\alpha+\min\{n_0,n_1\}}
\\
w_0&&\rantirefltransindex{\alpha+n_0}&&\repl{l_1\mu_1}{q}{r_2\mu_2}
&&\rrefltransindex{\alpha+n_1}&&w_1
\\
\drefltransindex{\alpha+n_1}&&&&&&&&\drefltransindex{\alpha+n_0}
\\
\circ&&&&\requal&&&&\circ
\\
\end{diagram}
\noindent
In case of \bigmath{\neitherprefix{p}{q}} we simply can choose
\bigmath{
  w_0
  :=
  \repl
    {\repl{l_1\mu_1}{p}{r_0\mu_0}}
    {q}
    {r_2\mu_2}
.}
Otherwise, 
there must be some \math{\bar p}, \math{\hat p}, \math{\hat q},
with
\bigmath{
  p
  \tightequal
  \bar p\hat p
,}
\bigmath{
  q
  \tightequal
  \bar p\hat q
,}
and
\bigmath{
  \inparenthesesinlinetight{
      \hat p 
      \tightequal
      \emptyset
    \oder
      \hat q 
      \tightequal
      \emptyset
  }
.}
Now it suffices to show
\\\linemath{
  \repl
    {s/\bar p}
    {\hat p}
    {r_0\mu_0}
  \refltransindex{\alpha+\min\{n_0,n_1\}}
  w_0'
  \antirefltransindex{\alpha+n_0}
  \repl
    {s/\bar p}
    {\hat q}
    {r_2\mu_2}
}
for some \math{w_0'}, because by \monotonicity\ 
of \refltransindex{\alpha+n'}\ we then have
\\\math{
  \repl
    {l_1\mu_1}
    {p}
    {r_0\mu_0}
  \tightequal
  \repl
    {s}
    {\bar p\hat p}
    {r_0\mu_0}
  \tightequal
  \repl
    {\repl
       {s}
       {\bar p}
       {s/\bar p}
    }
    {\bar p\hat p}
    {r_0\mu_0}
  \tightequal
  \\
  \repl
    {s}
    {\bar p}
    {\repl
       {s/\bar p}
       {\hat p}
       {r_0\mu_0}
    }
  \refltransindex{\alpha+\min\{n_0,n_1\}}
  \repl
    {s}
    {\bar p}
    {w_0'}
  \antirefltransindex{\alpha+n_0}
  \repl
    {s}
    {\bar p}
    {\repl
       {s/\bar p}
       {\hat q}
       {r_2\mu_2}
    }
  \tightequal
  \\
  \repl
    {\repl
       {s}
       {\bar p}
       {s/\bar p}
    }
    {\bar p\hat q}
    {r_2\mu_2}
  \tightequal
  \repl
    {s}
    {\bar p\hat q}
    {r_2\mu_2}
  \tightequal
  \repl
    {l_1\mu_1}
    {q}
    {r_2\mu_2}
.}\\
Note that
\\\linemath{\footroom
  \repl
    {s/\bar p}
    {\hat p}
    {r_0\mu_0}
  \antiredindex{\alpha+n_0,\hat p}
  s/\bar p
  \redindex{\alpha+\min\{n_0,n_1\},\hat q}
  \repl
    {s/\bar p}
    {\hat q}
    {r_2\mu_2}
.}
In case of \bigmath{\bar p\tightnotequal\emptyset}
(since then \bigmath{\superterm\subseteq\rhd} 
 implies \math{s\rhd s/\bar p})
we get some \math{w_0'} with
\bigmath{
  \repl
    {s/\bar p}
    {\hat p}
    {r_0\mu_0}
  \refltransindex{\alpha+\min\{n_0,n_1\}}
  w_0'
  \antirefltransindex{\alpha+n_0}
  \repl
    {s/\bar p}
    {\hat q}
    {r_2\mu_2}
}
by our first level of induction.
Otherwise, in case of \bigmath{\bar p\tightequal\emptyset,}
our disjunction from above means
\bigmath{
  \inparenthesesinlinetight{
      p 
      \tightequal
      \emptyset
    \oder
      q 
      \tightequal
      \emptyset
  }
.}
Since we have
\bigmath{
      \emptyset
      \tightnotequal
      q 
}
by our initial assumption, we may assume  
\bigmath{
      q
      \tightequal
      \hat q 
      \tightnotequal
      \emptyset
}
and
\bigmath{
      p 
      \tightequal
      \hat p 
      \tightequal
      \bar p
      \tightequal
      \emptyset
.}
Then the above divergence reads
\bigmath{
  \repl
    {s/\bar p}
    {\hat p}
    {r_0\mu_0}
  \antiredindex{\alpha+n_0,\emptyset}
  s
  \redindex{\alpha+\min\{n_0,n_1\},q}
  \repl
    {s/\bar p}
    {\hat q}
    {r_2\mu_2}
}
and we get the required joinability by our second induction level due to 
\bigmath{
  q\lll_s p
.}%
\QED{Claim~1}

\yestop
\yestop
\noindent
\underline{Claim~2 of the proof of Lemma~\ref{lemma quasi-free}:}
We may assume that for some \math{i\prec 2}:
\\
\bigmath{n_i\tightequal 0\tightprec n_{1-i};}
\bigmath{l_i\tightin\tcs;}
\bigmath{l_{1-i}\mu_{1-i}\tightnotin\tcs;}
\\\mbox{}\hfill
and 
\bigmath{
  \inparenthesesinline{
      l_{1-i}\tightnotin\tcs
    \oder
      \condterms{C_{1-i}\mu_{1-i}}\tightnotsubseteq\tcs
  }
.} 
\\\underline{Proof of Claim~2 of the proof of Lemma~\ref{lemma quasi-free}:}
If \bigmath{\forall i\tightprec 2\stopq s\redindex\omega t_{1-i},}
then the whole proof is finished by confluence of \redindex{\omega}.
Thus there is some \math{i\prec 2} with \bigmath{s\notredindex\omega t_{1-i}.}
Then we get \bigmath{0\tightprec n_{1-i}.}
The case of \bigmath{0\tightprec n_i} is empty,
since then due to
\bigmath{
  \beta
  \preceq
  \omega
  \prec
  n_0\plusomega n_1
}
the globally supposed ordering property 
\bigmath{
  (n_0\plusomega n_1,s)
  \preclhdeq
  (\beta,\hat s)
}
cannot hold.
Thus we get \bigmath{n_i\tightequal 0\tightprec n_{1-i}.}
\pagebreak
Due \bigmath{\Lambda_i\tightpreceq n_i\tightequal 0} we 
get \bigmath{l_i\tightin\tcs.}
By \lemmaconskeeping\ 
and
\bigmath{l_{1-i}\mu_{1-i}\redindex{\omega+n_{1-i}}r_{1-i}\mu_{1-i},} 
\bigmath{l_{1-i}\mu_{1-i}\tightin\tcs} would imply 
the contradictory
\bigmath{
  {l_{1-i}\mu_{1-i}}\redindex{\omega}
  {r_{1-i}\mu_{1-i}}
.}
Finally,
\bigmath{l_{1-i}\tightin\tcs} and 
\bigmath{\condterms{C_{1-i}\mu_{1-i}}\tightsubseteq\tcs}
by \lemmaconskeeping\
would imply that
\bigmath{C_{1-i}\mu_{1-i}} is fulfilled \wrt\ \redindex{\omega}
and then Corollary~\ref{corollary redsubomega is minimum} would 
imply the contradictory
\bigmath{
  {l_{1-i}\mu_{1-i}}
  \redindex{\omega}
  {r_{1-i}\mu_{1-i}}
}
again.\QED{Claim~2 of the proof of Lemma~\ref{lemma quasi-free}}

\yestop
\noindent
\underline{Claim~2 of the proof of Lemma~\ref{lemma quasi-free two and zero}:}
For each \math{k\prec 2} we may assume:
\bigmath{0\tightprec n_k;}
\\
\bigmath{
  \alpha\tightequal 0
  \ \implies\
  l_k\tightin\tcs
;}
and
\\
\bigmath{
  \alpha\tightequal\omega
  \ \implies\
  \inparenthesesoplist{
      l_k\mu_k\tightnotin\tcs
    \oplistund
      \inparenthesesoplist{
          l_k\tightnotin\tcs
        \oplistoder
          \condterms{C_k\mu_k}\tightnotsubseteq\tcs
     }
  }
.}      
\\
\underline{Proof of Claim~2 of the proof of Lemma~\ref{lemma quasi-free two and zero}:}
In case of \bigmath{\alpha\tightequal 0} we
have 
\bigmath{0\tightprec n_k} due to \bigmath{1\tightpreceq n_k}
and have
\bigmath{l_k\tightin\tcs} due to \bigmath{\Lambda_k\tightequal 0.}
Now we treat the case of \bigmath{\alpha\tightequal\omega:}
We may assume \bigmath{\forall k\tightprec 2\stopq s\notredindex\omega t_k,}
since otherwise the whole proof is finished by
\math\omega-shallow confluence up to \math\omega.
Thus we have 
\bigmath{
  0\tightprec n_0, n_1
.}
By \lemmaconskeeping\ 
and
\bigmath{l_k\mu_k\redindex{\omega+n_k}r_k\mu_k,} 
\bigmath{l_k\mu_k\tightin\tcs}
would imply the contradictory 
\bigmath{
  {l_k\mu_k}\redindex{\omega}
  {r_k\mu_k}
.}
Finally,
\bigmath{l_k\tightin\tcs} and 
\bigmath{\condterms{C_k\mu_k}\tightsubseteq\tcs}
by \lemmaconskeeping\
would imply that
\bigmath{C_k\mu_k} is fulfilled \wrt\ \redindex{\omega}
and then Corollary~\ref{corollary redsubomega is minimum} would imply
the contradictory
\bigmath{
  {l_k\mu_k}
  \redindex{\omega}
  {r_k\mu_k}
}
again.\QED{Claim~2 of the proof of Lemma~\ref{lemma quasi-free two and zero}}

\yestop
\noindent
\underline{Claim~3:}
For all \math{k\tightprec 2} we may assume:\\
\inparentheses{
  \alpha\tightequal 0
  \ \implies\ 
  l_k\tightin\tcs
};
\\
\inparentheses{
    \min\{n_0,n_1\}\tightpreceq(n_k\monus 1)
  \ \oder\ 
    \kurzregelindex k\mbox{ is }\alpha\mbox{-quasi-normal \wrt\ }\RX
};
\\
and
\RX\ is \math\alpha-shallow confluent up to 
\math{\min\{n_0,n_1\}\plusalpha(n_k\monus 1)}.
\\\underline{Proof of Claim~3 of the proof of Lemma~\ref{lemma quasi-free}:}
The first property is trivial due to \bigmath{\alpha\tightequal\omega.}
By Claim~2 we get 
\bigmath{
    \min\{n_0,n_1\}
    \tightequal
    0
    \tightpreceq
    (n_k\monus 1)
}
as well as
\math{
  \min\{n_0,n_1\}\plusomega(n_k\monus 1)
  \tightequal
  0\plusomega(n_k\monus 1)
  \tightequal
  (n_k\monus 1)
  \tightprec
  \max\{1,n_k\}
  \tightpreceq
  \max\{n_0,n_1\}
  \tightequal
  n_0\plusomega n_1
.} 
Thus \RX\ is \math\omega-shallow confluent up to 
\math{\min\{n_0,n_1\}\plusomega(n_k\monus 1)} by our first level of induction.%
\QED{Claim~3 of the proof of Lemma~\ref{lemma quasi-free}}
\\
\underline{Proof of Claim~3 of the proof of Lemma~\ref{lemma quasi-free two and zero}:}
The first property follows from Claim~2.
Since \RX\ is \math\alpha-quasi-normal, 
\kurzregelindex k is 
\math\alpha-quasi-normal \wrt\ \RX.
By Claim~2 we have 
\math{
  \min\{n_0,n_1\}\plusalpha(n_k\monus 1)
  \tightprec
  \min\{n_0,n_1\}\plusalpha n_k
  \tightpreceq
  n_0\plusalpha n_1
.} 
Thus Claim~3 follows from our first level of induction.%
\QED{Claim~3 of the proof of Lemma~\ref{lemma quasi-free two and zero}}

\yestop
\noindent
\underline{Claim~4:}
For any \math{k\prec 2} and \math{\nu\in\Xsubst},
if \math{C_k\nu} is fulfilled \wrt\ \redindex{\alpha+(n_k\monus 1)},
then \bigmath{l_k\nu\redindex{\alpha+n_k}r_k\nu.}
\\\underline{Proof of Claim~4 of the proof of Lemma~\ref{lemma quasi-free}:}
By Claim~2 we have 
\bigmath{0\tightprec n_k} 
or 
\\
\bigmath{n_k\tightequal 0\ \und\ l_k\tightin\tcs.}
In the first case
Claim~4 is trivial due to \bigmath{(n_k\monus 1)+1\tightequal n_k.}
In the second case 
\math{C_k\nu} is fulfilled \wrt\ \redindex{\omega} and
\bigmath{l_k\tightin\tcs.}
Thus, by Corollary~\ref{corollary redsubomega is minimum},
we get 
\bigmath{l_k\nu\redindex{\omega}r_k\nu,} 
which completes the proof of Claim~4 due to \bigmath{n_k\tightequal 0}
in this case.%
\QED{Claim~4 of the proof of Lemma~\ref{lemma quasi-free}}
\\
\underline{Proof of Claim~4 of the proof of Lemma~\ref{lemma quasi-free two and zero}:}
By Claim~2 we have 
\bigmath{
  0\tightprec n_k
}
and 
\bigmath{
  \alpha\tightequal 0
  \implies
  l_k\tightin\tcs
.} 
Thus Claim~4 is trivial due to \bigmath{(n_k\monus 1)+1\tightequal n_k.}%
\QED{Claim~4 of the proof of Lemma~\ref{lemma quasi-free two and zero}}

\yestop
\noindent
Two cases:

\yestop
\noindent
\underline{\underline{The variable-overlap case:
There are \math{q_0'}, \math{q_1'} such that
\bigmath{p\tightequal q_0'q_1';}
\bigmath{l_1/q_0'\tightequal x\tightin\V:}}}
\begin{diagram}
l_1\mu_1
&&&&\rredindex{\alpha+n_1}
&&&&r_1\mu_1
\\
\dredindex{\alpha+n_0}
&&&&
&&&&\drefltransindex{\alpha+n_0}
\\
\repl{l_1\mu_1}p{r_0\mu_0}
&&\rrefltransindex{\alpha+n_1}&&l_1\nu
&&\rredindex{\alpha+n_1}&&r_1\nu
\end{diagram}
We have 
\bigmath{
  x\mu_1/q_1'
  \tightequal
  l_1\mu_1/q_0'q_1'
  \tightequal
  s/p
  \tightequal
  l_0\mu_0
.}

\yestop
\noindent
\underline{Claim~A of the proof of Lemma~\ref{lemma quasi-free}:}
\\
In case of ``\math{i\tightequal 1}'' for the `\math i' of Claim~2
we may assume 
\bigmath{x\in\Vsig.}
\\\underline{Proof of Claim~A of the proof of Lemma~\ref{lemma quasi-free}:}
Otherwise we would have \math{x\tightin\Vcons}, which implies 
\math{x\mu_1\tightin\tcc} and then \math{l_0\mu_0\tightin\tcc}.
We may assume \math{l_{1-i}\mu_{1-i}\tightnotin\tcc}
for the \math i of Claim~2.%
\QED{Claim~A of the proof of Lemma~\ref{lemma quasi-free}}

\yestop
\noindent
\underline{Claim~A of the proof of Lemma~\ref{lemma quasi-free two and zero}:}
\\
We may assume 
\bigmath{
              \inparenthesesoplist{
                  \inparentheses{ 
                      \alpha\tightequal 0
                    \ \implies\
                      l_1\tightin\tcs
                  }
                \oplistund
                  \inparentheses{ 
                      \alpha\tightequal\omega
                    \ \implies\
                      x\tightin\Vsig
                  } 
              }
.}
\\
\underline{Proof of Claim~A of the proof of Lemma~\ref{lemma quasi-free two and zero}:}
The first statement follows from Claim~2. The second is show by contradiction:
Suppose we would have \math{x\tightin\Vcons}, which implies 
\math{x\mu_{1}\tightin\tcc} and then \math{l_0\mu_0\tightin\tcc}.
By Claim~2 we can assume that this is not the
case for \bigmath{\alpha\tightequal\omega.}%
\QED{Claim~A of the proof of Lemma~\ref{lemma quasi-free two and zero}}

\yestop
\noindent
By \lemmaconskeeping\ (in case of \math{x\tightin\Vcons}), 
we can define $\nu\in\Xsubst$ by
($y\tightin\V$):
\\ 
\math{y\nu:=
 \left\{\begin{array}{ll}
   \repl{x\mu_1}
        {q_1'}
        {r_0\mu_0}
   &
   \mbox{if } y=x
   \\
   y\mu_1
   &
   \mbox{otherwise}
   \\
 \end{array}\right\}
} 
and get 
\bigmath{y\mu_1\onlyonceindex{\alpha+n_0} y\nu}
for
\bigmath{y\in\V.}
\\
By \monotonicity\ of \redindex{\alpha+n_0} we get
\bigmath{
  r_1\mu_1
  \refltransindex{\alpha+n_0}
  r_1\nu
}
and
\\\arr{{@{}l@{}}
  \repl{l_1\mu_1}{q_0' q_1'}{r_0\mu_0}
  =
\\
  \replpar{\repl{l_1}
                 {q_0'}
                 {x\nu}}
          {q''}
          {y\mu_1}
          {l_1/q''\tightequal y\in\V\ \wedge\ q''\tightnotequal q_0'}
  =
\\
  \replpar{
           \replpar{\repl{l_1}
                         {q_0'}
                         {x\nu}}
                   {q''}
                   {x\mu_1}
                   {l_1/q''\boldequal x\ \wedge\ q''\boldunequal q_0'}
          }
          {q''}
          {y\nu}
          {x\boldunequal l_1/q''\boldequal y\tightin\V
           \ \wedge\ 
           q''\tightnotequal q_0'}
\\
  \refltransindex{\alpha+n_0}\ \ 
  \replpar{l_1}{q''}{y\nu}{l_1/q''=y\in\V}
  =l_1\nu
.}

\yestop
\noindent
\underline{Claim~B:}
  \bigmath{
      \repl{l_1\mu_1}p{r_0\mu_0}
      \refltransindex{\alpha+n_1}
      l_1\nu  
.}
\\\underline{Proof of Claim~B of the proof of Lemma~\ref{lemma quasi-free}:}
By case distinction over the `\math i' of Claim~2:
\\\underline{``\math{i\tightequal0}'':}
\bigmath{
  n_0\tightequal 0\tightprec n_1
}
implies
\bigmath{
  \redindex{\omega+n_0}
  \subseteq
  \redindex{\omega+n_1}
}
by \lemmamonotonicinbeta.
\\\underline{``\math{i\tightequal1}'':}
\mbox{}
In this case we have 
\bigmath{l_1\tightin\tcs.} 
By Claim~A we may assume \math{x\tightin\Vsig}.
Then \math{l_1} is linear in \math x.
Thus
\bigmath{
  \setwith
    {q''}
    {l_1/q''\boldequal x\ \wedge\ q''\boldunequal q_0'}
  =
  \emptyset
,}
which means that the above reduction takes \math0 steps, \ie\
\bigmath{ 
    \repl{l_1\mu_1}p{r_0\mu_0}
    \tightequal
    l_1\nu
.}%
\QED{Claim~B of the proof of Lemma~\ref{lemma quasi-free}}
\\
\underline{Proof of Claim~B of the proof of Lemma~\ref{lemma quasi-free two and zero}:}
By Claim~A and the assumption of our lemma we know that 
\math{l_0} is linear in \math x.
Thus
\bigmath{
  \setwith
    {q''}
    {l_{1}/q''\boldequal x\ \wedge\ q''\boldunequal q_0'}
  =
  \emptyset
,}
which means that the above reduction takes \math0 steps, \ie\
\bigmath{ 
    \repl{l_{1}\mu_{1}}p{r_0\mu_0}
    =
    l_{1}\nu
.}%
\QED{Claim~B of the proof of Lemma~\ref{lemma quasi-free two and zero}}

\pagebreak

\yestop
\noindent
\underline{Claim~C:}
\bigmath{
  l_1\nu\redindex{\alpha+n_1}r_1\nu
.}
\\
\underline{Proof of Claim~C of the proof of Lemma~\ref{lemma quasi-free}:}
By case distinction over the `\math i' of Claim~2:
\\\underline{``\math{i\tightequal0}'':}
\mbox{}
Due to 
\bigmath{
  n_0\tightequal 0\tightprec n_1
}
this follows directly from Lemma~\ref{lemma invariance of fulfilledness}
(matching its \math{n_0} to our \math{n_0\tightequal0} and its 
              \math{n_1} to our \math{n_1\monus 1})
(since \bigmath{0\tightpreceq n_1\monus 1}
and \RX\ is \math\omega-shallow confluent up to \math{n_1\monus 1}
by our induction hypothesis).
\\\underline{``\math{i\tightequal1}'':}
\mbox{}
In this case we have \bigmath{n_1\tightequal 0}
and \bigmath{l_1\tightin\tcs.}
Thus, since \math{C_1\mu_1} is fulfilled \wrt\ \redindex{\omega},
by assumption of the lemma we know that \kurzregelindex 1 is 
quasi-normal \wrt\ \RX\
and that for all \math{u\in\condterms{C_1}} we have
\bigmath{
  l_1\mu_1
  \rhd\,
  u\mu_1
}
or
\bigmath{
   u\mu_1
   \tightnotin
   \DOM\red
}
or
\bigmath{
   \VAR u
   \subseteq
   \Vcons
.}
In the latter case,
since we may assume \bigmath{x\tightin\Vsig} by Claim~A,
we get
\bigmath{\forall y\tightin\VAR u\stopq y\mu_1\tightequal y\nu}
and, moreover, 
\bigmath{
  \forall\delta\tightprec n_0\plusomega n_1\stopq
  \RX\mbox{ is \math\omega-shallow confluent up to }\delta
}
by our induction hypothesis.
In the first case,
due to \bigmath{l_1\mu_1\tightequal s} our induction hypothesis
even implies that 
\RX\ is \math\omega-shallow confluent up to \math{n_0\plusomega n_1}
and \math{u\mu_1} in \tight\lhd.
Thus  Lemma~\ref{lemma invariance of fulfilledness}
(matching its \math{n_0} to our \math{n_0} and its 
              \math{n_1} to our \math{n_1})
implies that \bigmath{C_1\nu} is fulfilled \wrt\ \redindex{\omega+n_1}.
Now since \bigmath{n_1\tightequal 0,}
Corollary~\ref{corollary redsubomega is minimum}
implies 
\bigmath{l_1\nu\redindex{\omega+n_1}r_1\nu.}%
\QED{Claim~C of the proof of Lemma~\ref{lemma quasi-free}}%
\\
\underline{Proof of Claim~C of the proof of Lemma~\ref{lemma quasi-free two and zero}:}
Directly Lemma~\ref{lemma invariance of fulfilledness}
(matching its \math{n_0} to our \math{n_0} and its 
              \math{n_1} to our \math{n_1\monus 1})
(by Claim~2 and since  
\RX\ is 
\math\alpha-quasi-normal and
\math\alpha-shallow confluent up to \math{n_0\plusalpha(n_1\monus 1)}
by our first level of induction 
due to \math{n_1\monus 1\preceq n_1} by Claim~2).%
\QED{Claim~C of the proof of Lemma~\ref{lemma quasi-free two and zero}}

\noindent
\Qeddouble{The variable-overlap case}

\vfill

\yestop
\noindent
\underline{\underline{The critical peak case:
\math{p\tightin\TPOS{l_{1}};} \math{l_1/p\tightnotin\V}:}}
\ 
Let \math{\xi_0\in\SUBST\V\V} be a bijection with 
\bigmath{
  \xi_0[\VAR{\sugarregelindex{0}}]\cap\VAR{\sugarregelindex{1}}
  =
  \emptyset
.}
Define
\bigmath{
  \Y
  :=
  \VAR{(\sugarregelindex{0})\xi_0,
        \sugarregelindex{1}
  }
.}
Define \math{\xi_1:=\domres\id\V}.
Let $\varrho$ be given by
$\ x\varrho=
\left\{\begin{array}{@{}l@{}l@{}}
  x\mu_{{1}}        &\mbox{ if }x\in\VAR{\sugarregelindex{1}}\\
  x\xi_0^{-1}\mu_{{0}}&\mbox{ else}\\
\end{array}\right\}
\:(x\tightin\V)$.
\\
By
\bigmath{
  l_{{0}}\xi_0\varrho
  \tightequal
  l_0\xi_0\xi_0^{-1}\mu_0
  \tightequal
  s/p
  \tightequal
  l_1\mu_1/p
  \tightequal
  l_1\varrho/p
  \tightequal
  (l_1/p)\varrho
}
let
\bigmath{
  \sigma:=\minmgu{\{(l_0\xi_0}{l_1/p)\},\Y}
}
and
\bigmath{\varphi\in\Xsubst}
with
\math{
  \domres{\inpit{\sigma\varphi}}\Y
    \!=\!\domres\varrho\Y
.}

\yestop
\noindent
\underline{Claim~A:}
We may assume
\bigmath{
  \inparentheses{
      p\tightequal\emptyset
    \ \oder\ 
      \forall y\tightin\VAR{l_1}\stopq
         y\sigma\varphi
         \tightnotin
         \DOM{\redindex{\alpha+\min\{n_0,n_1\}}}
  }
.}
\\\underline{Proof of Claim~A:}
Otherwise, when 
\bigmath{ 
      p\tightnotequal\emptyset
} 
holds but
\bigmath{
      \forall y\tightin\VAR{l_1}\stopq
         y\sigma\varphi
         \tightnotin
         \DOM{\redindex{\alpha+\min\{n_0,n_1\}}}
}
is not the case, there are some
\math{x\in\VAR{l_1},}
\math{\nu\in\Xsubst}  
with 
\bigmath{
  x\mu_1
  \redindex{\alpha+\min\{n_0,n_1\}}
  x\nu
}
and 
\bigmath{
  \forall y\tightin\V\tightsetminus\{x\}\stopq
    y\mu_1\tightequal y\nu
.}
Due to 
\bigmath{
  l_1\mu_1/p
  \lhd 
  l_1\mu_1
  \tightequal
  s
} 
by our first level of induction
from 
\bigmath{
  r_0\xi_0\sigma\varphi
  \antiredindex{\alpha+n_0}
  l_0\xi_0\sigma\varphi
  \tightequal
  l_1\sigma\varphi/p
  \tightequal
  l_1\mu_1/p
  \refltransindex{\alpha+\min\{n_0,n_1\}}
  l_1\nu/p
}
we know that there must be some \math u with
\bigmath{
  r_0\xi_0\sigma\varphi
  \refltransindex{\alpha+\min\{n_0,n_1\}}
  u
  \antirefltransindex{\alpha+n_0}
  l_1\nu/p
.}
Due to Claim~3,
by Lemma~\ref{lemma invariance of fulfilledness}
(matching its \math{n_0} to our \math{\min\{n_0,n_1\}} 
      and its \math{n_1} to our \math{(n_1\monus 1)})
\bigmath{C_1\nu} is fulfilled \wrt\ \redindex{\alpha+(n_1\monus 1)}.
Then Claim~4 implies
\bigmath{
  l_1\nu
  \redindex{\alpha+n_1}
  r_1\nu
.}
Due to 
\bigmath{
  l_1\mu_1
  \transindex{\alpha+\min\{n_0,n_1\}}
  l_1\nu
}
and
\bigmath{\redindex{\omega+\alpha}\nottight\subseteq\rhd} 
we get 
\bigmath{
  l_1\nu
  \lhd 
  l_1\mu_1
  \tightequal
  s
.}
Thus, by our first level of induction, from
\bigmath{
  \repl
    {l_1\nu}
    {p}
    {u}
  \antirefltransindex{\alpha+n_0}
  l_1\nu
  \redindex{\alpha+n_1}
  r_1\nu
}
we get
\bigmath{
  t_0
  \tightequal
  \repl{l_1\mu_1}p{r_0\xi_0\sigma\varphi}
  \refltransindex{\alpha+\min\{n_0,n_1\}}
  \repl{l_1\nu}  p{r_0\xi_0\sigma\varphi}
  \refltransindex{\alpha+\min\{n_0,n_1\}}
  \repl{l_1\nu}  p{u}
  \refltransindex{\alpha+n_1}
  \circ
  \antirefltransindex{\alpha+n_0}
  r_1\nu
  \antirefltransindex{\alpha+\min\{n_0,n_1\}}
  r_1\mu_1
  \tightequal
  t_1
.}\QED{Claim~A}

\pagebreak

\yestop
\noindent
If 
\bigmath{
  \repl{l_1}p{r_0\xi_0}
  \sigma
  \tightequal
  r_1
  \sigma
,}
then we are finished due to
\bigmath{
  t_0
  \tightequal
  \repl{l_1}p{r_0\xi_0}
  \sigma
  \varphi
  \tightequal
  r_1
  \sigma
  \varphi
  \tightequal
  t_1
.}
Otherwise 
we have
\bigmath{
  (\,
   (\repl{l_1}p{r_0\xi_0},
    C_{0}\xi_0,    
    \Lambda_{0}),\,
   (r_{1},
    C_{1}\xi_1,
    \Lambda_{1}),\,
    l_{1},\,
    \sigma,\,
    p\,)
  \in{\rm CP}(\R)
}
with the following additional structure:
\\\underline{In the proof of Lemma~\ref{lemma quasi-free}:}
By Claim~2 the critical peak cannot be of the form \math{(1,1)}.
Moreover, if it is of the form \math{(0,0)}, then we have
\bigmath{\forall k\tightprec 2\stopq l_k\tightin\tcs,} which by Claim~2 
for some \math{i\prec 2} implies
\bigmath{\condterms{C_{1-i}\xi_{1-i}\sigma\varphi}\tightnotsubseteq\tcs,}
and then 
\bigmath{\condterms{C_{1-i}\xi_{1-i}\sigma}\tightnotsubseteq\tcc,}
\ie\
\bigmath{\condterms{C_0\xi_0\sigma\,C_1\xi_1\sigma}\tightnotsubseteq\tcc.}
\\\underline{In the proof of Lemma~\ref{lemma quasi-free two and zero}:}
For all \math{k\prec 2} we have:
\bigmath{\alpha\tightequal 0\implies\Lambda_k\tightequal 0.}
If  \bigmath{\alpha\tightequal\omega} and 
\bigmath{\Lambda_k\tightequal 0} for some \math{k\prec 2}, then
\bigmath{l_k\tightin\tcs,} which by Claim~2 implies
\bigmath{\condterms{C_k\xi_k\sigma\varphi}\tightnotsubseteq\tcs,}
and then 
\bigmath{\condterms{C_k\xi_k\sigma}\tightnotsubseteq\tcc.}

\yestop
\noindent
Now
\bigmath{C_{0}\xi_0\sigma\varphi=C_{0}\mu_{0}}
is fulfilled \wrt\ \redindex{\alpha+(n_{0  }\monus 1)};
\bigmath{C_{1}\xi_1\sigma\varphi=C_{1}\mu_{1}}
is fulfilled \wrt\ \redindex{\alpha+(n_{1}\monus 1)}.
Since 
\bigmath{
  l_1\sigma\varphi
  \tightequal 
  l_1\mu_1
  \tightequal
  s
,}
by our induction hypothesis
we have
\math{
  \ \forall(\delta,s')\preclhd(n_0\plusalpha n_1,l_{1}\sigma\varphi)\stopq
} 
(\RX\ is \math\alpha-shallow confluent up to
\math{\delta} and \math{s'} in \math\lhd).
By Claim~1 we get
\bigmath{
  \forall q\tightin\TPOS{l_1\sigma\varphi}\stopq
  \inparentheses{
      \emptyset
      \tightnotequal
      {q}
      \lll_{l_1\sigma\varphi}
      p
    \ \implies\ 
      l_1\sigma\varphi\tightnotin\DOM{\redindex{\alpha+\min\{n_0,n_1\},q}}
  }
.}
This means \bigmath{l_1\sigma\varphi\tightnotin A(p,\min\{n_0,n_1\}).}
Furthermore, 
\bigmath{
  (n_0\plusalpha n_1,l_{1}\sigma\varphi) 
  =
  (n_0\plusalpha n_1,s)
  \preclhdeq
  (\beta,\hat s)
.}
Therefore,
in case of
\bigmath{
  \forall y\tightin\V\stopq
    y\varphi\tightnotin\DOM{\redindex{\alpha+\min\{n_0,n_1\}}}
,}
by Claim~A and
by the assumed form of \math\alpha-shallow joinability up to 
\math{\beta} and \math{\hat s} \wrt\ \RX\ and \math\lhd\ 
{[besides \math A]},
we get
\bigmath{
  t_0
  \tightequal
  \repl{l_1}p{r_0\xi_0}\sigma\varphi
  \refltransindex    {\alpha+n_1}
  \circ
  \antirefltransindex{\alpha+n_0}
  r_1\sigma\varphi
  \tightequal
  t_1
.}

\yestop
\noindent
Otherwise, when \bigmath{
  \forall y\tightin\V\stopq
    y\varphi\tightnotin\DOM{\redindex{\alpha+\min\{n_0,n_1\}}}
}
is not the case,
by \bigmath{\redindex{\omega+\alpha}\subseteq\tight\rhd} 
and the Axiom of Choice
there is some \math{\varphi'\tightin\Xsubst}
with
\bigmath{
  \forall y\tightin\V\stopq
    y\varphi
    \refltransindex{\alpha+\min\{n_0,n_1\}}
    y\varphi'
    \tightnotin
    \linebreak
    \DOM{\redindex{\alpha+\min\{n_0,n_1\}}}
.}
Then, of course,
\bigmath{
  \forall i\tightprec 2\stopq
  \forall y\tightin\V\stopq
    y\xi_i\sigma\varphi
    \refltransindex{\alpha+\min\{n_0,n_1\}}
    y\xi_i\sigma\varphi'
.}
Due to Claim~3,
by Lemma~\ref{lemma invariance of fulfilledness}
(matching its \math{n_0} to our \math{\min\{n_0,n_1\}} 
      and its \math{n_1} to our \math{(n_i\monus 1)})
we know that
\bigmath{
  \forall i\tightprec 2\stopq
      C_i\xi_i\sigma\varphi'\mbox{ is fulfilled \wrt\ }
      \redindex{\alpha+(n_i\monus 1)}
.}
Then Claim~4 implies
\bigmath{
  \forall i\tightprec 2\stopq
      l_i\xi_i\sigma\varphi'  
      \redindex{\alpha+n_i}
      r_i\xi_i\sigma\varphi'
.} 
Furthermore, we have 
\bigmath{
  \repl
    {l_1}
    {p}
    {r_0\xi_0}
  \sigma\varphi
  \refltransindex{\alpha+\min\{n_0,n_1\}}
  \repl
    {l_1}
    {p}
    {r_0\xi_0}
  \sigma\varphi'
}
and 
\bigmath{
  r_1\sigma\varphi'
  \antirefltransindex{\alpha+\min\{n_0,n_1\}}
  r_1\sigma\varphi
,}
\cf\ the diagram below.
Therefore, in case of 
\bigmath{
  l_1\sigma\varphi
  \tightequal 
  l_1\sigma\varphi'
}
the proof succeeds like above with \math{\varphi'} 
instead of \math{\varphi}.
Otherwise we have 
\bigmath{
  l_1\sigma\varphi
  \transindex{\omega+\alpha}
  l_1\sigma\varphi'
.} 
Then due to \bigmath{\redindex{\omega+\alpha}\nottight\subseteq\rhd} we get
\bigmath{
  s
  \tightequal
  l_1\sigma\varphi
  \rhd
  l_1\sigma\varphi'
.}
Therefore, by our first level of induction,
from
\bigmath{
  \repl{l_1}p{r_0\xi_0}\sigma\varphi'
  \antiredindex{\alpha+n_0,p}
  \repl{l_1}p{l_0\xi_0}\sigma\varphi'
  \tightequal
  l_1\sigma\varphi'
  \redindex{\alpha+n_1,\emptyset}
  r_1\sigma\varphi'
}
we conclude
\bigmath{
%  t_0
%  \tightequal
%  \repl{l_1}p{r_0\xi_0}\sigma\varphi
%  \refltransindex    {\alpha+\min\{n_0,n_1\}}
  \repl{l_1}p{r_0\xi_0}\sigma\varphi'
  \refltransindex    {\alpha+n_1}
  \circ
  \antirefltransindex{\alpha+n_0}
  r_1\sigma\varphi'
%  \antirefltransindex{\alpha+\min\{n_0,n_1\}}
%  r_1\sigma\varphi
%  \tightequal
%  t_1
.}
\begin{diagram}
t_0&&&&s&&&&t_1
\\
\dequal&&&&\dequal&&&&\dequal
\\
\repl{l_1}{p}{r_0\xi_0}\sigma\varphi
&&\rantiredindex{\alpha+n_0,\,p}
&&l_1\sigma\varphi&&\rredindex{\alpha+n_1,\,\emptyset}&&r_1\sigma\varphi
\\
\drefltransindex{\alpha+\min\{n_0,n_1\}}
&&&&
\drefltransindex{\alpha+\min\{n_0,n_1\}}
&&&&
\drefltransindex{\alpha+\min\{n_0,n_1\}}
\\
\repl{l_1}{p}{r_0\xi_0}\sigma\varphi'
&&\rantirefltransindex{\alpha+n_0,\,p}&&l_1\sigma\varphi'
&&\rrefltransindex{\alpha+n_1,\,\emptyset}&&r_1\sigma\varphi'
\\
\drefltransindex{\alpha+n_1}&&&&&&&&\drefltransindex{\alpha+n_0}
\\
\circ&&&&\requal&&&&\circ
\\
\end{diagram}
\QEDdouble{The critical peak case}
\end{proofparsepqed}

\pagebreak

\begin{proofparsepqed}
{Lemma~\ref{lemma level one}}
For \math{(\delta,s)\preclhdeq(\beta,\hat s)} 
we are going to show that \RX\
is \math\omega-level confluent up to \math\delta\ and \math{s} 
in \math\lhd\ 
by induction over \math{(\delta,s)} in \math\preclhd\@.
Suppose 
that for
\math{\bar n_0,\bar n_1\prec\omega}
we have
\bigmath{
  (\max\{\bar n_0,\bar n_1\},s)
  \preclhdeq
  (\beta,\hat s)
}
and
\bigmath{
 t_0'
 \antirefltransindex{\omega+\bar n_0}
 s
 \refltransindex    {\omega+\bar n_1}
 t_1'
.}
We have to show 
\bigmath{
  t_0'
  \refltransindex    {\omega+\max\{\bar n_0,\bar n_1\}}
  \circ
  \antirefltransindex{\omega+\max\{\bar n_0,\bar n_1\}}
  t_1'
.}

In case of \bigmath{\exists i\tightprec 2\stopq t_i'\tightequal s}
this is trivially true by \lemmamonotonicinbeta. 
In case of \bigmath{\bar n_0\tightequal \bar n_1\tightequal 0} this is true by 
confluence of \redindex{\omega}.
Using symmetry in \math 0 and \math 1,
\wrog\ we may assume \bigmath{\bar n_0\tightpreceq \bar n_1.}

Thus, assuming \bigmath{\bar n_0\tightpreceq \bar n_1\tightsucc 0,}
for
\linemath{
 t_0'
 \antirefltransindex{\omega+\bar n_0}
 t_0
 \antiredindex      {\omega+\bar n_0}
 s
 \redindex          {\omega+\bar n_1}
 t_1
 \refltransindex    {\omega+\bar n_1}
 t_1'
}
using the induction hypothesis that
\\\LINEnomath{
  \headroom\footroom
  \math{\forall (m,w')\preclhd(\max\{\bar n_0,\bar n_1\},s)\stopq}
  \RX\ is 
  \math\omega-level confluent up to \math m and \math{w'} in \math\lhd
}
\\
we have to show 
\\\LINEmath{
  t_0'
  \refltransindex    {\omega+\bar n_1}
  \circ
  \antirefltransindex{\omega+\bar n_1}
  t_1'
.}

\yestop
\noindent
\underline{Claim~0:}
Now it is sufficient to show
\bigmath{
  t_0
  \refltransindex    {\omega+\bar n_1}
  u
  \antirefltransindex{\omega+\bar n_1}
  t_1
}
for some \math u.
\\\underline{Proof of Claim~0:}
By Lemma~\ref{lemma terminating reduction relation} we have 
\bigmath{
  s\rhd t_0, t_1
.}
Thus,
due to%
\footnote{Note that it is this change from \math{\bar n_0}
to \math{\bar n_1} in \math{\max\{\bar n_0,\bar n_1\}} that makes a two level treatment
similar to that for \math\omega-shallow confluence (\ie\ considering
\math{\bar n_0\plusomega \bar n_1} instead of  \math{\bar n_0\tight+\bar n_1}) impossible
because then for \bigmath{\bar n_0\tightequal 0\tightprec \bar n_1}
we would get 
\bigmath{
  \max_{_{\omega}}\{\bar n_0,\bar n_1\}
  \tightprec
  \omega
  \tightpreceq
  \max_{_{\omega}}\{\bar n_1,\bar n_1\}
}
and thus would not be allowed to apply our induction hypothesis here.}
\bigmath{
  (\max\{\bar n_1,\bar n_1\},t_1)\preclhd\penalty-1(\max\{\bar n_0,\bar n_1\},s)
,}
by our induction hypotheses 
\bigmath{
  u
  \antirefltransindex{\omega+\bar n_1}
  t_1
  \refltransindex    {\omega+\bar n_1}
  t_1'
}
(\cf\ diagram below)
implies
the existence of some \math v with
\bigmath{
  u
  \refltransindex    {\omega+\bar n_1}
  v
  \antirefltransindex{\omega+\bar n_1}
  t_1'
} 
and then
\bigmath{
  t_0'
  \antirefltransindex{\omega+\bar n_0}
  t_0
  \refltransindex    {\omega+\bar n_1}
  v
}
implies
\bigmath{
  t_0'
  \refltransindex    {\omega+\bar n_1}
  \circ
  \antirefltransindex{\omega+\bar n_1}
  v
.}
\notop
\begin{diagram}
  s&\rredindex{\omega+\bar n_1}&t_1
&\rrefltransindex{\omega+\bar n_1}&t_1'
\\\dredindex{\omega+\bar n_0}&&\drefltransindex{\omega+\bar n_1}
&&\drefltransindex{\omega+\bar n_1}
\\t_0&\rrefltransindex{\omega+\bar n_1}&u
&\rrefltransindex{\omega+\bar n_1}&v
\\\drefltransindex{\omega+\bar n_0}&&
&&\drefltransindex{\omega+\bar n_1}
\\t_0'&&\rrefltransindex{\omega+\bar n_1}
&&\circ
\end{diagram}\notop\notop\notop\notop\halftop

\mbox{}\QED{Claim~0}

\yestop
\noindent
Defining \math{n:=\bar n_1\monus 1} and using \lemmamonotonicinbeta\ we can now
restate our proof task in the following symmetric way:

\noindent
For \math{n\prec\omega}, 
\bigmath{t_0\antiredindex{\omega+n+1,p}s\redindex{\omega+n+1,q}t_1}
using the induction hypothesis that
\\\LINEnomath{
  \headroom\footroom
  \math{\forall (m,w')\preclhd(n\tight+1,s)\stopq}
  \RX\ is 
  \math\omega-level confluent up to \math m and \math{w'} in \math\lhd
}
\\
we have to show 
\\\LINEmath{
  t_0
  \refltransindex    {\omega+n+1}
  \circ
  \antirefltransindex{\omega+n+1}
  t_1
.}

\yestop
\noindent
In case of \neitherprefix p q this is trivial.
Otherwise one of \math{p,q} must be a prefix of the other, \wrog\ say that
\math q is a prefix of \math p.
In case of \bigmath{q\tightnotequal\emptyset}
due to \bigmath{\superterm\subseteq\tight\rhd} we get 
\bigmath{s/q\lhd s} and the proof finished by our induction hypothesis
and \monotonicity\ of \refltransindex{\omega+n+1}.
Thus we may assume \bigmath{q\tightequal\emptyset.}
We start a second level of induction on 
\math p in \math{\lll_s}.
Thus we may assume 
the following induction hypothesis:
\\\LINEmath{
  \forall q\tightin\TPOS s\stopq
  \forall t_0',t_1'      \stopq
  \inparentheses{
      \inparenthesesoplist{
           q\lll_s p
         \oplistund
           t_0'
           \antiredindex{\omega+n+1,q}
           s
           \redindex{\omega+n+1,\emptyset}
           t_1'
      }
    \ \implies\ 
      t_0'\downarrowindex{\omega+n+1}t_1'
  }
}

\pagebreak

\yestop
\noindent
Now for $k<2$ there must be
\math{((l_{k},r_{k}),C_{k})\in\R};
\math{\mu_{k}\in\Xsubst};
with
\bigmath{C_{k}\mu_{k}} fulfilled \wrt\ \redindex{\omega+n};
\bigmath{s\tightequal l_{1}\mu_{1};}
\bigmath{s/p\tightequal l_{0}\mu_{0};}
\bigmath{t_{0}\tightequal \repl{l_{1}\mu_{1}}{p}{r_{0}\mu_{0}};}
\bigmath{t_{1}\tightequal r_{1}\mu_{1}.}
Moreover, for $k<2$ we define
\bigmath{
  \Lambda_k
  :=
  \left\{\arr{{ll}
      0&\mbox{ if }l_k\tightin\tcs
    \\1&\mbox{ otherwise}
  }\right\}
.}

\yestop
\noindent
\underline{Claim~1:}
We may assume that 
\bigmath{
  \forall q\tightin\TPOS s\stopq
  \inparentheses{
      \emptyset
      \tightnotequal
      {q}
      \lll_s
      p
    \ \implies\ 
      s\tightnotin\DOM{\redindex{\omega+n+1,q}}
  }
.}
\\
\underline{Proof of Claim~1:}
Otherwise there must be some
\math{q\in\TPOS s};
\math{\kurzregelindex2\in\R};
\math{\mu_2\in\Xsubst};
with
\math{C_2\mu_2} fulfilled \wrt\ \redindex{\omega+n};
\bigmath{s/q\tightequal l_2\mu_2;}
and
\bigmath{
      \emptyset
      \tightnotequal
      {q}
      \lll_s
      p
.}
By our second induction level we get
\bigmath{
  \repl
    {l_1\mu_1}
    {q}
    {r_2\mu_2}
  \refltransindex{\omega+n+1}
  w_1
  \antirefltransindex{\omega+n+1}
  r_1\mu_1
}
for some \math{w_1}; \cf\ the diagram below.
Next we are going to show that there is some \math{w_0} with
\bigmath{
  \repl{l_1\mu_1}{p}{r_0\mu_0}
  \refltransindex{\omega+n+1}
  w_0
  \antirefltransindex{\omega+n+1}
  \repl{l_1\mu_1}{q}{r_2\mu_2}
.}
Note that 
(since \bigmath{\red\subseteq\rhd} implies
\math{s\tight\rhd\repl{l_1\mu_1}{q}{r_2\mu_2}})
this finishes the proof since then 
\bigmath{
  w_0
  \antirefltransindex{\omega+n+1}
  \repl{l_1\mu_1}{q}{r_2\mu_2}
  \refltransindex{\omega+n+1}
  w_1
}
by our first level of induction
implies 
\bigmath{
  t_0
  \refltransindex{\omega+n+1}
  w_0
  \downarrowindex{\omega+n+1}
  w_1
  \antirefltransindex{\omega+n+1}
  t_1
.}
\begin{diagram}
t_0&&&&s&&&&t_1
\\
\dequal&&&&\dequal&&&&\dequal
\\
\repl{l_1\mu_1}{p}{r_0\mu_0}
&&\rantiredindex{\omega+n+1,\,p}
&&l_1\mu_1&&\rredindex{\omega+n+1,\,\emptyset}&&r_1\mu_1
\\
\drefltransindex{\omega+n+1}&&&&\dredindex{\omega+n+1,\,q}&&&&\drefltransindex{\omega+n+1}
\\
w_0&&\rantirefltransindex{\omega+n+1}&&\repl{l_1\mu_1}{q}{r_2\mu_2}&&\rrefltransindex{\omega+n+1}&&w_1
\\
\drefltransindex{\omega+n+1}&&&&&&&&\drefltransindex{\omega+n+1}
\\
\circ&&&&\requal&&&&\circ
\\
\end{diagram}
\noindent
In case of \bigmath{\neitherprefix{p}{q}} we simply can choose
\bigmath{
  w_0
  :=
  \repl
    {\repl{l_1\mu_1}{p}{r_0\mu_0}}
    {q}
    {r_2\mu_2}
.}
Otherwise, 
there must be some \math{\bar p}, \math{\hat p}, \math{\hat q},
with
\bigmath{
  p
  \tightequal
  \bar p\hat p
,}
\bigmath{
  q
  \tightequal
  \bar p\hat q
,}
and
\bigmath{
  \inparenthesesinlinetight{
      \hat p 
      \tightequal
      \emptyset
    \oder
      \hat q 
      \tightequal
      \emptyset
  }
.}
Now it suffices to show
\\\linemath{
  \repl
    {s/\bar p}
    {\hat p}
    {r_0\mu_0}
  \refltransindex{\omega+n+1}
  w_0'
  \antirefltransindex{\omega+n+1}
  \repl
    {s/\bar p}
    {\hat q}
    {r_2\mu_2}
}
for some \math{w_0'}, because by \monotonicity\ 
of \refltransindex{\omega+n+1}\ we then have
\\\math{
  \repl
    {l_1\mu_1}
    {p}
    {r_0\mu_0}
  \tightequal
  \repl
    {s}
    {\bar p\hat p}
    {r_0\mu_0}
  \tightequal
  \repl
    {\repl
       {s}
       {\bar p}
       {s/\bar p}
    }
    {\bar p\hat p}
    {r_0\mu_0}
  \tightequal
  \\
  \repl
    {s}
    {\bar p}
    {\repl
       {s/\bar p}
       {\hat p}
       {r_0\mu_0}
    }
  \refltransindex{\omega+n+1}
  \repl
    {s}
    {\bar p}
    {w_0'}
  \antirefltransindex{\omega+n+1}
  \repl
    {s}
    {\bar p}
    {\repl
       {s/\bar p}
       {\hat q}
       {r_2\mu_2}
    }
  \tightequal
  \\
  \repl
    {\repl
       {s}
       {\bar p}
       {s/\bar p}
    }
    {\bar p\hat q}
    {r_2\mu_2}
  \tightequal
  \repl
    {s}
    {\bar p\hat q}
    {r_2\mu_2}
  \tightequal
  \repl
    {l_1\mu_1}
    {q}
    {r_2\mu_2}
.}\\
Note that
\\\linemath{
  \repl
    {s/\bar p}
    {\hat p}
    {r_0\mu_0}
  \antiredindex{\omega+n+1,\hat p}
  s/\bar p
  \redindex{\omega+n+1,\hat q}
  \repl
    {s/\bar p}
    {\hat q}
    {r_2\mu_2}
.}
In case of \bigmath{\bar p\tightnotequal\emptyset}
(since then \bigmath{\superterm\subseteq\rhd} 
 implies \math{s\rhd s/\bar p})
we get some \math{w_0'} with
\bigmath{
  \repl
    {s/\bar p}
    {\hat p}
    {r_0\mu_0}
  \refltransindex{\omega+n+1}
  w_0'
  \antirefltransindex{\omega+n+1}
  \repl
    {s/\bar p}
    {\hat q}
    {r_2\mu_2}
}
by our first level of induction.
Otherwise, in case of \bigmath{\bar p\tightequal\emptyset,}
our disjunction from above means
\bigmath{
  \inparenthesesinlinetight{
      p 
      \tightequal
      \emptyset
    \oder
      q 
      \tightequal
      \emptyset
  }
.}
Since we have
\bigmath{
      \emptyset
      \tightnotequal
      q 
}
by our initial assumption, we may assume  
\bigmath{
      q
      \tightequal
      \hat q 
      \tightnotequal
      \emptyset
}
and
\bigmath{
      p 
      \tightequal
      \hat p 
      \tightequal
      \bar p
      \tightequal
      \emptyset
.}
Then the above divergence reads
\bigmath{
  \repl
    {s/\bar p}
    {\hat p}
    {r_0\mu_0}
  \antiredindex{\omega+n+1,\emptyset}
  s
  \redindex{\omega+n+1,q}
  \repl
    {s/\bar p}
    {\hat q}
    {r_2\mu_2}
}
and we get the required joinability by our second induction level due to 
\bigmath{
  q\lll_s p
.}%
\QED{Claim~1}

\yestop
\noindent
\underline{Claim~2:}
We may assume:
\bigmath{
  \exists i\tightprec 2\stopq
      l_i\tightnotin\tcs
.} 
\\\underline{Proof of Claim~2:}
Since \math{C_i\mu_i} is fulfilled \wrt\ \redindex{\omega+n},
by Lemma~\ref{lemma invariance of fulfilledness two} (matching both its
\math\mu\ and \math\nu\ to our \math{\mu_1})
\bigmath{l_i\tightin\tcs}
implies 
\bigmath{
  {l_i\mu_i}\redindex{\omega}
  {r_i\mu_i}
}
and then 
\bigmath{s\redindex{\omega}t_i.}
Thus, if the claim does not hold,
we have
\bigmath{
  t_0
  \antiredindex{\omega}
  s
  \redindex    {\omega}
  t_1
}
and
the proof is finished by confluence of 
\redindex{\omega}.
\QED{Claim~2}

\pagebreak

\yestop
\noindent
Now we have two cases:

\vfill

%%%%%%%%%%%%%%%%%%%%%%%%%%%%%%%%%%%%%%%%%%%%%%%%%%%%%%%%%%%%%%%%%%%%%%%%%%%%%%%
\noindent
\underline{\underline{The variable overlap case:
\bigmath{p\tightequal q_{0}q_{1};\ l_{1}/q_{0}\tightequal x\in\V}:}}
\\
We have
\bigmath{
  x\mu_{1}/q_{1}
  \tightequal
  l_{1}\mu_{1}/q_{0}q_{1}
  \tightequal 
  s/p
  \tightequal
  l_{0}\mu_{0}
.}
By
Lemma~\ref{lemma about sortkeeping}
(in case of $x\tightin\Vcons$), 
we can define $\nu\in\Xsubst$ by
($y\tightin\V$): 
\\\math{y\nu:=
 \left\{\begin{array}{ll}
 \repl{x\mu_{1}}{q_{1}}{r_{0}\mu_{0}}&\mbox{if } y\tightequal x\\
 y\mu_{1}&\mbox{otherwise}\\
 \end{array}\right\}
} and get 
\bigmath{y\mu_{1}\onlyonceindex{\omega+n+1} y\nu}
for
\math{y\tightin\V}.
By Corollary~\ref{corollary monotonic}:\\
\bigmath{t_0\tightequal \repl{l_1\mu_1}{q_0q_1}{r_{0}\mu_{0}}\tightequal 
  \replpar{\repl{l_1}
               {q_0}
               {x\nu}}
          {q'}
          {y\mu_{1}}
          {l_{1}/q'\tightequal y\in\V\und q'\tightnotequal q_{0}}
  \ \refltransindex{\omega+n+1}
}\\
\bigmath{
  \replpar
    {l_{1}}
    {q'}
    {y\nu}
    {l_{1}/q'\tightequal y\tightin\V}
  \tightequal
  l_{1}\nu
;}\\
\bigmath{
  t_{1}\tightequal r_{1}\mu_{1}\refltransindex{\omega+n+1} r_{1}\nu
.}
It suffices to show
\bigmath{l_1\nu\redindex{\omega+n+1} r_1\nu,}
which follows from our first level of induction 
saying that \RX\ is \math\omega-level confluent up to \math n
by Lemma~\ref{lemma invariance of fulfilledness level}
(matching its \math{n_0} to our \math{n\tight+1}  
      and its \math{n_1} to our \math{n}).%
\QEDdouble{The variable overlap case}

\vfill

\noindent
\underline{\underline{The critical peak case:
\bigmath{p\in\TPOS{l_{1}};\ l_{1}/p\not\in\V}:}}
Let \math{\xi\in\SUBST\V\V} be a bijection 
with 
\\
\bigmath{\xi[\VAR{\sugarregelindex0}]\cap\VAR{\sugarregelindex1}\tightequal \emptyset.}
Define
\bigmath{\Y:=\VAR{(\sugarregelindex0)\xi,\sugarregelindex1}}\@.
\\
Let $\varrho$ be given by
$\ x\varrho\tightequal 
\left\{\begin{array}{@{}l@{}l@{}}
  x\mu_{1}        &\mbox{ if }x\in\VAR{\sugarregelindex1}\\
  x\xi^{-1}\mu_{0}&\mbox{ else}\\
\end{array}\right\}
\:(x\tightin\V)$.
By
\bigmath{l_{0}\xi\varrho\tightequal l_{0}\xi\xi^{-1}\mu_{0}\tightequal s/p
\tightequal l_{1}\mu_{1}/p\tightequal l_{1}\varrho/p\tightequal (l_{1}/p)\varrho}
let
\bigmath{
  \sigma:=\minmgu{\{(l_{0}\xi}{l_{1}/p)\},\Y}
}
and
\bigmath{\varphi\in\Xsubst}
\linebreak
with
\math{
  \domres{\inpit{\sigma\varphi}}\Y
    \tightequal\domres\varrho\Y
}\@.

\noindent
\underline{Claim~A:}
We may assume
\bigmath{
  \inparentheses{
      p\tightequal\emptyset
    \ \oder\ 
      \forall y\tightin\VAR{l_1}\stopq
         y\sigma\varphi\tightnotin\DOM{\redindex{\omega+n+1}}
  }
.}
\\\underline{Proof of Claim~A:}
Otherwise, when 
\bigmath{ 
      p\tightnotequal\emptyset
} holds
but
\bigmath{
      \forall y\tightin\VAR{l_1}\stopq
         y\sigma\varphi\tightnotin\DOM{\redindex{\omega+n+1}}
}
is not the case, there are some
\math{x\in\VAR{l_1},}
\math{\nu\tightin\Xsubst}  
with \bigmath{x\sigma\varphi\redindex{\omega+n+1} x\nu}
and 
\bigmath{
  \forall y\tightin\V\tightsetminus\{x\}\stopq
    y\mu_1\tightequal y\nu
.}
Due to 
\bigmath{
  l_1\mu_1/p
  \lhd 
  l_1\mu_1
  \tightequal
  s
} 
by our first level of induction
from 
\bigmath{
  r_0\xi\sigma\varphi
  \antiredindex{\omega+n+1}
  l_0\xi\sigma\varphi
  \tightequal
  l_1\sigma\varphi/p
  \tightequal
  l_1\mu_1/p
  \refltransindex{\omega+n+1}
  l_1\nu/p
}
we know that there must be some \math u with
\bigmath{
  r_0\xi\sigma\varphi
  \refltransindex{\omega+n+1}
  u
  \antirefltransindex{\omega+n+1}
  l_1\nu/p
.}
Due to 
\bigmath{
  l_1\mu_1
  \transindex{\omega+n+1}
  l_1\nu
}
and
\bigmath{\red\nottight\subseteq\rhd} 
we get 
\bigmath{
  l_1\nu
  \lhd 
  l_1\mu_1
  \tightequal
  s
.}
Thus, by our first level of induction, from
\bigmath{
  \repl
    {l_1\nu}
    {p}
    {u}
  \antirefltransindex{\omega+n+1}
  l_1\nu
  \redindex{\omega+n+1}
  r_1\nu
}
(which is due to Lemma~\ref{lemma invariance of fulfilledness level} 
and our first level of induction
saying that \RX\ is \math\omega-level confluent up to \math n)
we get
\bigmath{
  t_0
  \tightequal
  \repl
    {l_1\mu_1}
    {p}
    {r_0\xi\sigma\varphi}
  \refltransindex{\omega+n+1}
  \repl
    {l_1\nu}
    {p}
    {r_0\xi\sigma\varphi}
  \refltransindex{\omega+n+1}
  \repl
    {l_1\nu}
    {p}
    {u}
  \downarrowindex{\omega+n+1}
  r_1\nu
  \antirefltransindex{\omega+n+1}
  r_1\mu_1
  \tightequal
  t_1
.}\QED{Claim~A}

\noindent
If 
\bigmath{
  \repl  
    {l_1}
    {p}
    {r_0\xi}
  \sigma
  \tightequal
  r_1 
  \sigma
,} 
then we are finished due to
\bigmath{
  t_0
  \tightequal
  \repl  
    {l_1}
    {p}
    {r_0\xi}
  \sigma
  \varphi
  \tightequal
  r_1 
  \sigma
  \varphi
  \tightequal
  t_1
.} 
Otherwise
\bigmath{
  ((\repl
      {l_1}
      {p}
      {r_0\xi}
    ,
    C_0\xi,
    \Lambda_0),\ 
   (r_1,
    C_1,
    \Lambda_1),\
   l_1,\
   \sigma,\
   p\
  )
}
is a critical peak in \math{{\rm CP}(\R)}.
Furthermore, due to Claim~2, 
this critical peak is not of the form \math{(0,0)}.

\noindent
Now
\bigmath{
  (C_0\xi\,C_1)\sigma\varphi
  \tightequal
  C_0\mu_0\,C_1\mu_1
}
is fulfilled \wrt\ \redindex{\omega+n}.
Due to  
\bigmath{
  l_1\sigma\varphi
  \tightequal 
  l_1\varrho
  \tightequal 
  l_1\mu_1
  \tightequal 
  s
,} 
by our first level of induction we get
\math{
  \ 
  \forall
  (\delta,s')
  \preclhd
  (n\tight+1,l_1\sigma\varphi)\stopq
} 
(\RX\ is \math\omega-level confluent up to
\math{\delta} and \math{s'} in \math\lhd).
By Claim~1 we get
\bigmath{
  \forall q\tightin\TPOS{l_1\sigma\varphi}\stopq
  \inparentheses{
      \emptyset
      \tightnotequal
      {q}
      \lll_{l_1\sigma\varphi}
      p
    \ \implies\ 
      l_1\sigma\varphi\tightnotin\DOM{\redindex{\omega+n+1,q}}
  }
.}
This means \bigmath{l_1\sigma\varphi\tightnotin A(p,n\tight+1).}
Furthermore, 
\bigmath{
  (n\tight+1,l_1\sigma\varphi) 
  \tightequal
  (\max\{n_0,n_1\},s)
  \preclhdeq
  (\beta,\hat s)
.}
Thus, in case of
\bigmath{
  \forall y\tightin\V\stopq
    y\varphi\tightnotin\DOM{\redindex{\omega+n+1}}
,}
by Claim~A and
the assumed 
by \math\omega-level joinability up to 
\math{\beta} and \math{\hat s} \wrt\ \RX\ and \math\lhd\ 
{[besides \math A]}  
(matching the definition's \math{n_0} and \math{n_1} to our \math{n\tight+1})
we get 
\bigmath{
  t_0
  \tightequal
  \repl  
    {l_1}
    {p}
    {r_0\xi}
  \sigma
  \varphi
  \downarrowindex{\omega+n+1}
  r_1 
  \sigma
  \varphi
  \tightequal
  t_1
.}

\pagebreak

\noindent
Otherwise, when \bigmath{
  \forall y\tightin\V\stopq
    y\varphi\tightnotin\DOM{\redindex{\omega+n+1}}
}
is not the case,
by \bigmath{\red\subseteq\tight\rhd} and the Axiom of Choice
there is some \math{\varphi'\tightin\Xsubst}
with
\bigmath{
  \forall y\tightin\V\stopq
    y\varphi
    \refltransindex{\omega+n+1}
    y\varphi'
    \tightnotin
    \DOM{\redindex{\omega+n+1}}
.}
Then, of course,
\bigmath{
  \forall y\tightin\V\stopq
    y\xi\sigma\varphi
    \refltransindex{\omega+n+1}
    y\xi\sigma\varphi'
}
and
\bigmath{
  \forall y\tightin\V\stopq
    y\sigma\varphi
    \refltransindex{\omega+n+1}
    y\sigma\varphi'
.} 
By Lemma~\ref{lemma invariance of fulfilledness level} 
(due to our first level of induction
saying that \RX\ is \math\omega-level confluent up to \math n)
we know that
\bigmath{C_0\xi\sigma\varphi'}
and
\bigmath{C_1\sigma\varphi'}
are fulfilled \wrt\ \redindex{\omega+n}.
Furthermore, we have 
\bigmath{
  \repl
    {l_1}
    {p}
    {r_0\xi}
  \sigma\varphi
  \refltransindex{\omega+n+1}
  \repl
    {l_1}
    {p}
    {r_0\xi}
  \sigma\varphi'
}
and 
\bigmath{
  r_1\sigma\varphi'
  \antirefltransindex{\omega+n+1}
  r_1\sigma\varphi
.}
Therefore, in case of 
\bigmath{
  l_1\sigma\varphi
  \tightequal 
  l_1\sigma\varphi'
}
the proof succeeds like above with \math{\varphi'} 
instead of \math{\varphi}.
Otherwise we have \bigmath{l_1\sigma\varphi\trans l_1\sigma\varphi'.} 
Then due to \bigmath{\red\nottight\subseteq\rhd} we get
\bigmath{
  s
  \tightequal
  l_1\sigma\varphi
  \rhd
  l_1\sigma\varphi'
.}
Therefore, by our first level of induction,
from
\bigmath{
  \repl
    {l_1}
    {p}
    {r_0\xi}
  \sigma\varphi'
  \antiredindex{\omega+n+1}
  \repl
    {l_1}
    {p}
    {l_0\xi}
  \sigma\varphi'
  \tightequal
  l_1
  \sigma\varphi'
  \redindex{\omega+n+1}
  r_1
  \sigma\varphi'
}
(which is due to Lemma~\ref{lemma invariance of fulfilledness level} 
and our first level of induction
saying that \RX\ is \math\omega-level confluent up to \math n)
we conclude
\bigmath{
  t_0
  \tightequal
  \repl
    {l_1}
    {p}
    {r_0\xi}
  \sigma\varphi
  \refltransindex{\omega+n+1}
  \repl
    {l_1}
    {p}
    {r_0\xi}
  \sigma\varphi'
  \downarrowindex{\omega+n+1}
  r_1
  \sigma\varphi'
  \antirefltransindex{\omega+n+1}
  r_1
  \sigma\varphi
  \tightequal
  t_1
.}%
\QEDdouble{The critical peak case}
\end{proofparsepqed}

\vfill

\begin{proofparsepqed}{Lemma~\ref{lemmaa}}
\underline{\underline{\ref{lemmaa iteme}.:}}
Since the direction ``\tight\supseteq'' 
is trivial we only have to show ``\tight\subseteq'' and begin with the first
equation.
For \math{t'\in\supertermeq{[{\rm T}]}} there are some \math{t\in{\rm T}}
and \math{p\in\TPOS t} with \bigmath{t/p\tightequal t'.}
Now, in case of \bigmath{t'\rightrightarrows t''} by sort-invariance and
{\rm T}-monotonicity of \math\rightrightarrows\ we get 
\bigmath{
  t
  \tightequal
  \repl t p{t'}
  \rightrightarrows
  \repl t p{t''}
  \tightin
  {\rm T}
,}
which implies 
\bigmath{
  t''
  \tightin
  \supertermeq{[{\rm T}]}
.}
Thus we have shown
\bigmath{
    \tight{\domres\id{\supertermeq{[{\rm T}]}}}
    \nottight\circ
    \tight\rightrightarrows
    {\nottight{\nottight\subseteq}}
    \tight{\domres\id{\supertermeq{[{\rm T}]}}}
    \nottight\circ
    \tight\rightrightarrows
    \nottight\circ
    \tight{\domres\id{\supertermeq{[{\rm T}]}}}
.}
In case of \bigmath{t'\tightin{\rm T}} we can choose
\bigmath{p\tightequal\emptyset} and get \bigmath{t''\tightin{\rm T},}
which proves
\bigmath{
    \tight{\domres\id{{\rm T}}}
    \nottight\circ
    \tight\rightrightarrows
    {\nottight{\nottight\subseteq}}
    \tight{\domres\id{{\rm T}}}
    \nottight\circ
    \tight\rightrightarrows
    \nottight\circ
    \tight{\domres\id{{\rm T}}}
.}

\noindent
\underline{\underline{\ref{lemmaa itema}.:}}
For \bigmath{{\rm T}\ni t\superterm t'\rightrightarrows t''} there is a 
\math{p\in\TPOS{t}};
\math{p\tightnotequal\emptyset} with 
\bigmath{t'\tightequal t/p.} 
By sort-invariance and {\rm T}-monotonicity\ of $\rightrightarrows$ we get 
\bigmath{t=\repl{t}{p}{t'}\rightrightarrows\repl{t}{p}{t''}\superterm t''}
and \bigmath{\repl{t}{p}{t''}\tightin{\rm T}.}

\noindent
\underline{\underline{\ref{lemmaa itemb}.:}}
The subset relationship is simple:
\\\linemath{
    \tight{\domres\id{\supertermeq{[{\rm T}]}}}
    \nottight\circ
    \transclosureinline{(\rightrightarrows\cup\superterm)}
    \nottight{\nottight{\nottight\subseteq}}
    \tight\subtermeq
    \nottight\circ
    \tight{\domres\id{{\rm T}}}
    \nottight\circ
    \tight\supertermeq
    \nottight\circ
    \transclosureinline{(\rightrightarrows\cup\superterm)}
    \nottight{\nottight{\nottight\subseteq}}
    \tight\subtermeq
    \nottight\circ
    \tight{\domres\id{{\rm T}}}
    \nottight\circ
    \transclosureinline{(\rightrightarrows\cup\superterm)}
.}
The first equality follows from
(\ref{lemmaa iteme})
and 
\bigmath{
  \nottight{
    \tight{\domres\id{\supertermeq{[{\rm T}]}}}
    \nottight\circ
    \tight\superterm
  {\nottight{\nottight{\nottight=}}}
    \tight{\domres\id{\supertermeq{[{\rm T}]}}}
    \nottight\circ
    \tight\superterm
    \nottight\circ
    \tight{\domres\id{\supertermeq{[{\rm T}]}}}
  }
.}
For the second equality
consider the following subset relationships as a 
word rewriting system over the alphabet 
\bigmath{
  \{
     \tight{\domres\id{{\rm T}}},
     \tight\rightrightarrows,
     \tight\superterm
  \}
} (containing three letters):
\\\indent
\bigmath{
  \begin{array}{l@{$\nottight{\nottight{\nottight\subseteq}}$}l l}
  \tight{\domres\id{{\rm T}}}
  \nottight\circ
  \tight\superterm
  \nottight\circ
  \tight\rightrightarrows
 &
  \tight{\domres\id{{\rm T}}}
  \nottight\circ
  \tight\rightrightarrows
  \nottight\circ
  \tight{\domres\id{{\rm T}}}
  \nottight\circ
  \tight\superterm
 &;
 \\
  \tight\superterm
  \nottight\circ
  \tight\superterm
 &
  \tight\superterm
 &;  
 \\
  \tight{\domres\id{{\rm T}}}
  \nottight\circ
  \tight\rightrightarrows
  \nottight\circ
  \tight\superterm
 &
  \tight{\domres\id{{\rm T}}}
  \nottight\circ
  \tight\rightrightarrows
  \nottight\circ
  \tight{\domres\id{{\rm T}}}
  \nottight\circ
  \tight\superterm
 &;  
 \\
  \tight{\domres\id{{\rm T}}}
  \nottight\circ
  \tight\rightrightarrows
  \nottight\circ
  \tight\rightrightarrows
 &
  \tight{\domres\id{{\rm T}}}
  \nottight\circ
  \tight\rightrightarrows
  \nottight\circ
  \tight{\domres\id{{\rm T}}}
  \nottight\circ
  \tight\rightrightarrows
 &.
 \\
  \end{array}
}  
\\
First note that the system is sound: The first rule was proved in
(\ref{lemmaa itema}). The second is transitivity of \tight\superterm.
The third and fourth are implied by (\ref{lemmaa iteme}).
Since the number of substrings from
\bigmath{
  \{
    \tight\rightrightarrows,
    \tight\superterm
  \}^2
}
is decreased by 1 by each of the rules, the word rewriting system 
is terminating. Thus, since all normal forms from
\bigmath{
    \tight{\domres\id{{\rm T}}}
  \{
    \tight\rightrightarrows,
    \tight\superterm
  \}^+
}
are in
\bigmath{
  \{
    \tight{\domres\id{{\rm T}}}
    \tight\superterm
  \}
  \cup
  \{
    \tight{\domres\id{{\rm T}}}
    \,
    \tight\rightrightarrows
  \}^+
  [\{
    \tight{\domres\id{{\rm T}}}
    \tight\superterm
  \}]
,}
we get
\bigmath{
    \tight{\domres\id{{\rm T}}}
    \nottight\circ
    \transclosureinline{
      \inparenthesesinline{
        \tight\rightrightarrows
        \cup
        \tight\superterm
      }
    }
    \nottight{\nottight\subseteq}  
    \inparentheses{
      \tight{\domres\id{{\rm T}}}
      \nottight\circ
      \tight\superterm
    }
    \nottight\cup
    \inparentheses{
      \transclosureinline{
        \inparenthesesinline{
           \tight{\domres\id{{\rm T}}}
           \nottight\circ
           \tight\rightrightarrows
        }
      }
      \nottight\circ
      \reflclosureinline{ 
        \inparenthesesinline{
          \tight{\domres\id{{\rm T}}}
          \nottight\circ
          \tight\superterm
        }
      }
    }
.}
Using
(\ref{lemmaa iteme})
again as well as
\bigmath{
  \reflclosureinline\superterm
  \nottight\subseteq
  \supertermeq
,}
this implies the one direction; the other direction as well as the special
case are trivial.

\pagebreak

\noindent
\underline{\underline{\ref{lemmaa itemc}.:}}
By the first equation of (\ref{lemmaa itemb}) we conclude 
\bigmath{
  \tight\rhd
  \nottight{\nottight\subseteq}
  \supertermeq{[{\rm T}]}
  \nottight\times
  \supertermeq{[{\rm T}]}
}
as well as transitivity of \math\rhd\@.
Suppose that 
\math\rhd\
is not terminating. 
By the first equation of 
(\ref{lemmaa itemb}) there is some \FUNDEF r\N{\supertermeq{[{\rm T}]}}
with
\bigmath{
  \forall i\tightin\N\stopq 
    \inparenthesesinline{
        r_i\tight\rightrightarrows r_{i+1}
      \oder
        r_i\superterm r_{i+1}
    }
.}
There is some \math{t_0\in{\rm T}} and some \math{p_0\in\TPOS{t_0}} with
\bigmath{t_0/p_0\tightequal r_0.}
Moreover, there is also some 
\FUNDEF p{\N_+}{\N^\ast}
such that 
\\\linemath{
  \forall i\tightin\N\stopq 
    \inparentheses{
        \inparenthesesoplist{
            r_i\rightrightarrows r_{i+1}
          \oplistund
            p_{i+1}\tightequal\emptyset
        }
      \ \oder\ 
        \inparenthesesoplist{
            r_i\superterm r_{i+1}
          \oplistund
            r_i/p_{i+1}\tightequal r_{i+1}
        }
    }
.}
Define \math{(t_n)_{n\in\N}} inductively by
\bigmath{
   t_{n+1}
   :=
   \repl{t_n}{p_0\ldots p_{n+1}}{r_{n+1}}
.}
%\\\underline{Claim~1:}
%For each \math{n\in\N} we get
%\bigmath{
%     t_n/p_0\ldots p_{n}
%     \tightequal
%     r_{n}
%.}
%\\\underline{Proof of Claim~1:} 
%\bigmath{
%  t_0/p_0
%  \tightequal 
%  r_0
%.}
%\\
%\bigmath{
%  t_{n+1}/p_0\ldots p_{n+1} 
%  \tightequal 
%  \repl{t_n}{p_0\ldots p_{n+1}}{r_{n+1}}/p_0\ldots p_{n+1} 
%  \tightequal
%  r_{n+1}
%.}\QED{Claim~1}
\\\underline{Claim~2:}
For each \math{n\in\N} we get
\bigmath{
  \inparenthesesoplist{
      t_n,t_{n+1}\tightin{\rm T}
    \oplistund
      t_n/p_0\ldots p_n\tightequal r_n
    \oplistund
      t_{n+1}/p_0\ldots p_{n+1}\tightequal r_{n+1}
    \oplistund
      \inparentheses{
          t_n\tight\rightrightarrows t_{n+1}
        \ \oder\ 
          \inparenthesesoplist{
              t_n\tightequal t_{n+1}
            \oplistund
              r_n\superterm r_{n+1}
        }
    }
  }
.}
\\\underline{Proof of Claim~2:}
We have \bigmath{t_n\tightin{\rm T}} 
and \bigmath{t_n/p_0\ldots p_n\tightequal r_n}
in case of 
\bigmath{n\tightequal 0} by our choice above 
and otherwise inductively by Claim~2.
In case of 
        \bigmath{
            r_n\tight\rightrightarrows r_{n+1}
          \und
            p_{n+1}\tightequal\emptyset
        ,}
since \math\rightrightarrows\ is sort-invariant and {\rm T}-monotonic,
we thus get:
\bigmath{
   t_n
   \tightequal
   \repl{t_n}{p_0\ldots p_{n}}{r_n}
   \rightrightarrows
   \repl
     {t_n}
     {p_0\ldots p_{n}}
     {r_{n+1}}
   \tightequal
   \repl
     {t_n}
     {p_0\ldots p_{n}p_{n+1}}
     {r_{n+1}}
   \tightequal
   t_{n+1}
   \tightin
   {\rm T}
.}
Otherwise we have
\bigmath{r_n\superterm r_{n+1}}
and
\bigmath{r_n/p_{n+1}\tightequal r_{n+1}}
and get:
\bigmath{
  {\rm T}
  \tightni
  t_n
  \tightequal
  \repl
    {t_n}
    {p_0\ldots p_n}
    {r_n}
  \tightequal
  \repl
    {t_n}
    {p_0\ldots p_n}
    {\repl
      {r_n}
      {p_{n+1}}
      {r_{n+1}}}
  \tightequal
  \\
  \repl
    {\repl
       {t_n}
       {p_0\ldots p_n}
       {r_n}}
    {p_0\ldots p_n p_{n+1}}
    {r_{n+1}}
  \tightequal
  \repl
    {t_n}
    {p_0\ldots p_n p_{n+1}}
    {r_{n+1}}
  \tightequal
  t_{n+1}
.}
In both cases we have 
\bigmath{
  t_{n+1}/p_0\ldots p_{n+1} 
  \tightequal 
  \repl{t_n}{p_0\ldots p_{n+1}}{r_{n+1}}/p_0\ldots p_{n+1} 
  \tightequal
  r_{n+1}
.}\QED{Claim~2}
\\
Since \superterm\ is terminating,
Claim~2 contradicts \math\rightrightarrows\ being terminating 
(below all \math{t\in{\rm T}}).
\\
If \math\rightrightarrows\ and {\rm T} are \stable, additionally, 
then $\rhd$ is \stable\ too, because 
\math{\supertermeq{[{\rm T}]}},
\math{\domres\id{\supertermeq{[{\rm T}]}}},
and
\superterm\ are.
\\
Here is an example for $\rhd$ not sort-invariant and not 
{\rm  T}-monotonic:
Let $A,B$ be two different sorts. Let 
\bigmath{\sigarity(\appzero)=A},
\bigmath{\sigarity(\fsymbol)=A\aritysugar B},
\bigmath{\sigarity(\gsymbol)=A\aritysugar A}\@.
Define $\rightrightarrows:=\emptyset$
and \math{{\rm T}:=\vt}\@.
Then we have $\rhd\,=\superterm$ and therefrom: 
\bigmath{\fppeins\appzero\rhd\appzero}
(hence not sort-invariant);
and
\bigmath{\gppeins\appzero\rhd\appzero} but
\bigmath{\fppeins{\gppeins\appzero}\ntriangleright\fppeins\appzero} 
(hence not {\rm  T}-monotonic).

\noindent
\underline{\underline{\ref{lemmaa itemd}.:}}
Take the signature from the example in the proof of (\ref{lemmaa itemc}). 
Define
$\rightrightarrows\,:=\{(\appzero,\fppeins\appzero)\}$ 
and \math{{\rm T}:=\vt}\@.
Now \math\rightrightarrows\
is a {\rm  T}-monotonic\ (indeed!),
terminating relation on \vt\ that is not sort-invariant;
whereas $\rhd$ is not irreflexive: 
\bigmath{\appzero\ \rightrightarrows\ 
         \fppeins\appzero\ \superterm\ \appzero
}\@.
If one changes $\sigarity(\fsymbol)$ to be 
\bigmath{\sigarity(\fsymbol)=A\aritysugar A},
then $\rightrightarrows$ is a sort-invariant, terminating relation on 
\vt\ that is not {\rm  T}-monotonic\ but \math\emptyset-monotonic;
whereas neither \math\rhd\ nor 
\transclosureinline{
  \inparenthesesinlinetight{
    \rightrightarrows\cup\superterm
  }
}
(in contrast to 
\math{
  \domres\id{\supertermeq{[\emptyset]}}
  \circ
  \transclosureinline{
    \inparenthesesinlinetight{
      \rightrightarrows\cup\superterm
    }
  }
}) 
are irreflexive.
\end{proofparsepqed}
%%%%%%%%%%%%%%%%%%%%%%%%%%%%%%%%%%%%%%%%%%%%%%%%%%%%%%%%%%%%%%%%%%%%%%%%%%%%%%%

\pagebreak

\begin{proofparsepqed}{Lemma~\ref{lemma for theorem quasi overlay joinable}}
For 
the proof of Claim~3 below,
we \signatureenlarge\ 
the signatures by a new sort $s_{\rm new}$ and 
new
constructor 
symbols
\eqindexsymbol{\bar s}
for each old sort $\bar{s}\in\sigsorts$
with arity $\bar{s}\bar{s}\aritysugar s_{\rm new}$
and $\bot$ with arity $s_{\rm new}$.
We take (in addition to \R) the following set of new rules
(with \math{X_{\bar{s}}\in\Vsigindex{\bar{s}}} for 
\math{\bar{s}\in\sigsorts}):\\ 
\linemath{
  \R'
  :=
  \setwith{\eqindexpp{\bar s}{X_{\bar s}}{X_{\bar s}}=\bot}
          {\bar{s}\in\sigsorts}
.}
Since the sort restrictions
do not allow \redindex{\R\cup\R',\X,\beta} 
to make any use of terms of the sort 
$s_{\rm new}$ when rewriting terms of an 
``old''
sort, we get\\
\linemath{
  \forall\beta\preceq\omega\tightplus\omega\stopq
  \ 
  \redindex{\R\cup\R',\X,\beta}\cap(\tsigX\tighttimes\vt)
  \ =\ 
  \redindex{\R,\X,\beta/\sig/\cons}
} 
(the latter being defined over the non-enriched signatures).
Thus,
\bigmath{
  {\rm T}
  :=
  \refltrans{[\{\hat s\}]}
,} 
\bigmath{\domres\redsimple{{\rm T}},} 
and
\bigmath{\domres\redsimple{\trianglerighteq{[{\rm T}]}}}
do not change when we exchange the one \red\ with the other.
We use `\superterm' to denote the subterm ordering over 
the \signatureenlarged\ signature.
For keeping the assumptions of our lemma valid for this subterm ordering
(instead of the subterm ordering on the non-\signatureenlarged\ signature)
we have to extend \math\rhd\ with
\bigmath{
  \eqindexpp{\bar s}{t_0}{t_1}
  \rhd
  t'
}
if
\bigmath{
  \exists i\tightprec 2\stopq
     t_i
     \supertermeq
     t'
} 
for some \math{\bar s\in\sigsorts} and \math{t_0, t_1\in\tss_{\bar s}}.
This extension neither changes
\bigmath{
  \tight\trianglerighteq{[{\rm T}]}
}
nor 
\bigmath{\domres\redsimple{\trianglerighteq{[{\rm T}]}}.}
Thus, since \math{\domres\redsimple{\trianglerighteq{[{\rm T}]}}} is not
changed by any of the extensions, it now suffices to show its confluence
after the extensions.
Since the sort restrictions do not allow a term of the sort \math{s_{\rm new}}
to be a proper subterm of any other term, it is obvious that 
after the extension of \math\rhd\ we still may assume either that\/ 
\bigmath{\domres{\redindex{\R\cup\R',\X}}{{\rm T}}}
is terminating 
and \bigmath{\tight\rhd=\superterm}
or that
\bigmath{
    \domres{\redindex{\R\cup\R',\X}}{\trianglerighteq{[{\rm T}]}}
    {\nottight\subseteq}
    \math\rhd
,}
\bigmath{
    \superterm
    {\nottight\subseteq}
    \math\rhd
,}
and \tight\rhd\ is a wellfounded ordering on \vt.
Moreover, again due to the sort 
restrictions not allowing a term of the sort \math{s_{\rm new}}
to be a proper subterm of any other term,
if 
\bigmath{
  w\,\transclosureinline{(\antired\tightcup\,\lhd)}\,\,
  (\hat t/p )\sigma\varphi
}
holds for the extended \red\ and \math\rhd\ and 
if \math{\hat t} is an old term,
then this also holds for the non-extended \red\ and \math\rhd.
Therefore, (as no new critical peaks occur)
the critical peaks keep being \math\rhd-quasi overlay joinable.

We define 
\bigmath{\redindex{\beta}:=\redindex{\R\cup\R',\X,\beta}}
for any ordinal $\beta$ with \bigmath{\beta\prec\omega\tight+\omega;} and
\bigmath{\red:=\redindex{\omega+\omega}:=\redindex{\R\cup\R',\X}.}

Since
\red\ is sort-invariant, {\rm T}-monotonic
(\cf\ Corollary~\ref{corollary monotonic}),
and terminating below all \math{t\in{\rm T}}, 
by Lemma~\ref{lemmaa}(\ref{lemmaa itemc}), 
\bigmath{
  \tight\rhd'
  \nottight{\nottight{:=}}
  \domres\id{\supertermeq{[{\rm T}]}}
  \circ
  \transclosureinline{
    \inparenthesesinlinetight{
      \red
      \cup
      \superterm
    }
  }
}
is a wellfounded ordering on \bigmath{\supertermeq{[{\rm T}]}.} 
In case of \bigmath{\tight\rhd\tightequal\superterm,}
we define \bigmath{\tight>:=\tight\rhd'}.
Otherwise, in case that
\bigmath{
    \domres\redsub{\trianglerighteq{[{\rm T}]}}
    {\nottight\subseteq}
    \math\rhd
,}
\bigmath{
    \superterm
    {\nottight\subseteq}
    \math\rhd
,}
and \tight\rhd\ is a wellfounded ordering,
we define
\bigmath{\tight>\nottight{:=}\tight\rhd\cap
 ({\tight\trianglerighteq{[{\rm T}]}}
  \times
  {\tight\trianglerighteq{[{\rm T}]}}
 )
}.
In any case, \tight> is a wellfounded ordering
on \bigmath{\tight\trianglerighteq{[{\rm T}]}} containing 
\bigmath{
  \domres\id{\trianglerighteq{[{\rm T}]}}
  \circ
  \transclosureinline{
    \inparenthesesinlinetight{
      \red
      \cup
      \superterm
      \cup
      \tight\rhd
    }
  }
.}
This means in particular that \bigmath{\tight\trianglerighteq{[{\rm T}]}}
is closed under \red, \superterm, and \tight\rhd.
 
We say that $P(v,u,s,t,\Pi)$ holds \udiff\ 
for  
\math{v,u,t\in\tsigX}
and
\math{s\in\tight\trianglerighteq{[{\rm T}]}}
with 
\bigmath{v\antirefltrans u;} 
and
\bigmath{
  s
  \refltrans 
  t
;}
\bigmath{\Pi\subseteq\TPOS{u}} with 
\bigmath{
  \forall p,q\tightin\Pi\stopq
    \inparenthesesinline{
     p\tightnotequal q
    \implies
     \neitherprefix p q
    }
} and
\bigmath{
  \forall o\tightin\Pi\stopq
    u/o
    \tightequal
    s
;}
we have
\bigmath{v\downarrow\replpar{u}{o}{t}{o\tightin\Pi}.}
Now (by $\Pi:=\{\emptyset\}$) 
it suffices to show that $P(v,u,s,t,\Pi)$
holds for all appropriate $v,u,s,t,\Pi$.
We will show this by terminating induction over the lexicographic
combination of the following orderings:
\\\indent\math{\begin{array}{@{}lll}
 1.&>
   &\\
 2.&\succ                                                     
   &          \\
 3.&\succ                                    
   &          \\
\end{array}}
\\
using the following measure on $(v,u,s,t,\Pi)$:
\\\indent\math{\begin{array}{@{}ll}
 1.&s                                                                    \\
 2.&\mbox{the smallest ordinal }
    \beta\preceq\omega\tightplus\omega\mbox{ for which }
    v\antirefltransindex{\beta}u                                         \\
 3.&\mbox{the smallest }n\in\N\mbox{ for which }
    v\antiredindexn{n}{\beta}u\mbox{ for the }\beta\mbox{ of (2)}  
   \hspace{8.5em}\mbox{}\\
\end{array}}

\pagebreak

\yestop
\noindent
For the limit ordinals $0$, $\omega$, $\omega\tightplus\omega$ 
in the second position of 
the measure, 
the induction step is trivial
(\bigmath{\antirefltransindex{0}\subseteq\id}; \ 
 \bigmath{\antirefltransindex{\omega}\subseteq
  \bigcup_{i\in\N}\antirefltransindex{i}
 }; \ 
 \bigmath{\antirefltransindex{\omega+\omega}\subseteq
  \bigcup_{i\in\N}\antirefltransindex{\omega+i}
 }).
Thus,
as we now suppose a smallest $(v,u,s,t,\Pi)$ with 
$P(v,u,s,t,\Pi)$
not holding for,
the second position of the measure must be a non-limit ordinal 
\math{\beta\tightplus1.}
\nopagebreak

As $P(v,u,s,t,\Pi)$ holds trivially for 
\bigmath{u=v} or 
\bigmath{s=t}
we have some $u',s'$
with
\\
\bigmath{v\antiredindexn{n}{\beta+1}u'\antiredindex{\beta+1}u}
$(n\tightin\N)$
(with 
\math{
  \forall m\tightin\N\stopq
     (v\antiredindexn m{\beta+1}u\nottight
      \implies 
      m\tightsucc n)
})
and
\bigmath{s\redsimple s'\refltrans t.}
Now for a contradiction it is sufficient to show

\noindent
\LINEnomath{
\underline{\underline{{\bf Claim}:}} \mbox{} 
There is some $z$ with 
$v\refltrans z\antirefltrans\replpar{u}{o}{s'}{o\in\Pi}.$
}

\noindent
because then we have $z\downarrow\replpar{u}{o}{t}{o\in\Pi}$
by 
\math{P(z,\replpar{u}{o}{s'}{o\in\Pi},s',t,\Pi)},
which is smaller than $(v,u,s,t,\Pi)$
in the first position 
of the measure 
by
$s\redsimple s'$.

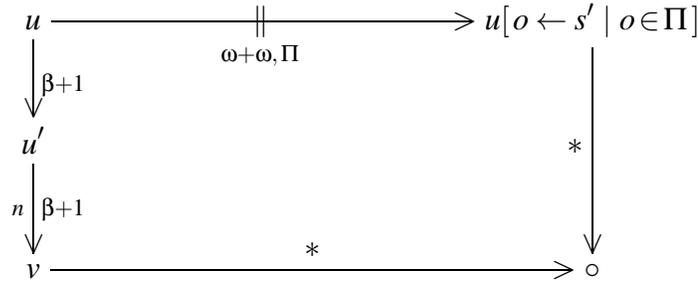
\yestop
\begin{diagram}
u&&&\rredparaindex{\omega+\omega,\,\Pi}&&&\replpar u{o}{s'}{o\tightin\Pi}
\\
\dredindex{\beta+1}&&&&&&
\\
u'&&&&&&\drefltrans
\\
\dredindexn n{\beta+1}&&&&&&
\\
v&&&\rrefltrans&&&\circ
\\
\end{diagram}

\yestop
\noindent
\underline{Claim~0:}
We may assume 
\bigmath{
  \forall p''\tightin\TPOS s\tightsetminus\{\emptyset\}\stopq 
    s/p''\tightnotin\DOM\red
.}
\\
\underline{Proof of Claim~0:}
Otherwise there are some \math{p''\in\TPOS s\tightsetminus\{\emptyset\}}
and some \math{s''} with \bigmath{s/p''\redsimple s''.}
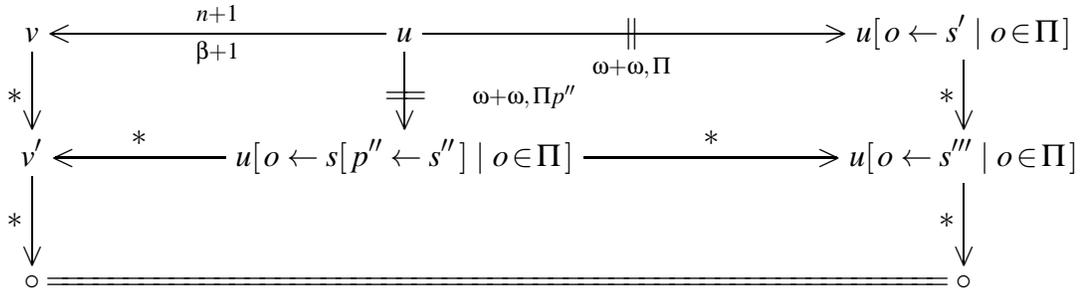
\begin{diagram}
v&&\rantiredindexn{n+1}{\beta+1}&&u&&&\rredparaindex{\omega+\omega,\,\Pi}
&&&\replpar u{o}{s'  }{o\tightin\Pi}
\\
\drefltrans&&&&\dredparaindex{\omega+\omega,\,\Pi p''}&&&&&&\drefltrans
\\
v'&&\rantirefltrans&&\replpar{u}{o}{\repl s{p''}{s''}}{o\tightin\Pi}
&&&\rrefltrans&&&\replpar u{o}{s'''}{o\tightin\Pi}
\\
\drefltrans&&&&&&&&&&\drefltrans
\\
\circ&&&&&\requal&&&&&\circ
\\
\end{diagram}
Then,
by
\math{P(s',s,s/p'',s'',\{p''\})},
which is smaller in the first position of the measure by
\math{s\superterm s/p''},
we get
\bigmath{s'\refltrans s'''\antirefltrans\repl s{p''}{s''}} 
for some \math{s'''}.
Similarly,
by
\math{P(v,u,s/p'',s'',\Pi p'')}
we get
\bigmath{
  v
  \refltrans 
  v'
  \antirefltrans
  \replpar{u}{p} {s''}              {p\tightin\Pi p''}
  \tightequal
  \replpar{u}{o}{\repl s{p''}{s''}}{o\tightin\Pi}
} 
for some \math{v'}.
Finally, 
by
\bigmath{P(v',\ 
        \replpar{u}{o}{\repl s{p''}{s''}}{o\tightin\Pi},\ 
        \repl s{p''}{s''},\ 
        s''',\ 
        \Pi\ )
,}
which is smaller in the first position of the measure by
\bigmath{
  s
  \red
  \repl s{p''}{s''}
,}
we get
\bigmath{
  v'
  \downarrow
  \replpar u{o}{s'''}{o\tightin\Pi}
  \antirefltrans
  \replpar u{o}{s'  }{o\tightin\Pi}
.}\QED{Claim~0}

\yestop
\noindent
By Claim~0
there are some 
$\kurzregelindex0\in\R\cup\R'$;
$\mu_{0}\in\Xsubst$; with
$s=l_{0}\mu_{0}$;
$s'=r_{0}\mu_{0}$;
and $C_0\mu_0$ is fulfilled \wrt\ \red.
Furthermore, we have some $q\in\TPOS u$; $\kurzregelindex1\in\R\cup\R'$;
\math{\mu_1\in\Xsubst}; with
\bigmath{
  u/q
  \tightequal
  l_1\mu_1
;}
\bigmath{
  u'
  \tightequal
  \repl u q{r_1\mu_1}
;}
$C_1\mu_1$ fulfilled \wrt\ \redindex\beta;
and 
if \math{C_1} contains some inequality \math{(u\boldunequal v)} then 
\bigmath{\omega\tightpreceq\beta.}
By Claim~0 we may assume that 
\math{q} is not strictly below any \math{p\in\Pi}, \ie\
that there are no \math{p}, \math{p'}
with \bigmath{p p'\tightequal q,} 
\bigmath{p'\tightnotequal\emptyset,} and \bigmath{p\tightin\Pi.}

\yestop
\noindent
Define 
\LINEmath{
  \begin{array}[t]{lrll}
    \Xi
    &:=
    &\Pi\setminus(q\N^\ast)
    &;
    \\
    \Pi'
    &:=
    &\setwith
       {p'}
       {\ q p'\tightin\Pi
        \ \und\ 
        \inparenthesesinlinetight{
            p'\tightin\TPOS{l_1}
          \implies
            l_1/p'\tightin\V
        }
       }
    &;
    \\
    \Pi''
    &:=
    &\setwith
       {p'}
       {qp'\tightin\Pi\tightsetminus(q\Pi')}
    &.
    \\
  \end{array}
}

\noindent
Define a function \math\Gamma\ on \V\ by ($x\tightin\V$):
\bigmath{
  \Gamma(x):=
  \setwith{p''}
          {\exists p'\stopq(l_{1}/p'=x\ \wedge\ p' p''\in\Pi')}
.}
Since for $p''\in\Gamma(x)$ we always have some $p'$ with 
\bigmath{
  l_{1}/p'
  \tightequal
  x
;} 
\bigmath{
  x\mu_{1}/p''
  \tightequal
  l_{1}\mu_{1}/p' p''
  \tightequal
  u/q p' p''
  \tightequal
  s
;}
we have
\\\LINEmath{
  \forall x \tightin\V       \stopq
  \forall p''\tightin\Gamma(x)\stopq
    x\mu_1/p''\tightequal s
.}(\#0)
\\
Since the proper subterm ordering is irreflexive we cannot have
\bigmath{s\superterm s,}
and therefore get
\\\LINEmath{
  \forall x\tightin\V\stopq
  \forall p',p''\tightin\Gamma(x)\stopq
    \inparenthesesinline{
        p'\tightequal p''
      \oder
        \neitherprefix{p'}{p''}
    }
.}(\#1)
\\
Due to (\#0) and (\#1) we can define $\mu_{1}'$ by ($x\tightin\V$):
\\\LINEmath{x\mu_{1}':=\replpar{x\mu_{1}}{p''}{s'}{p''\in\Gamma(x)}.}
\\
Define for \math{\bar w\in\vt}:
\\\LINEmath{
  \Theta_{\bar w}
  :=
     \setwithstart
       {p' p''}
     \setwithmark
        \exists x\stopq
        \inparenthesesinlinetight{
            \bar w/p'\tightequal x
          \und
            p''\tightin\Gamma(x)
        } 
     \setwithstop
.}
\\
By (\#0) we get 
\\\LINEmath{
  \forall\bar w \tightin\vt            \stopq  
  \forall p'    \tightin\Theta_{\bar w}\stopq
     \bar w\mu_1/p'
     \tightequal
     s
}(\#\tight\Theta1)
\\
and by (\#1)
\\\LINEmath{
  \forall\bar w \tightin\vt\stopq  
  \forall p',p''\tightin\Theta_{\bar w}\stopq
    \inparenthesesinline{
        p'\tightequal p''
      \ \oder
        \neitherprefix{p'}{p''}
    }
}(\#\tight\Theta2)
\\
and
\\\LINEmath{
  \forall\bar w\tightin\vt\stopq  
    \bar w\mu_1'
    \tightequal
    \replpar
      {\bar w\mu_1}
      {p'}
      {s'}
      {p'\tightin\Theta_{\bar w}}
.}(\#\tight\Theta3)
\\
Note that 
for
\math{
    \Lambda
    :=
    \Theta_{l_1}\tightsetminus\Pi'
}
we have 
\\\LINEmath{
  \Theta_{l_1}
  =
  \Pi'
  \uplus
  \Lambda
.}(\#2)
\\
By (\#\tight\Theta1) and (\#2) we get
\\\LINEmath{
  \forall p'\in\Pi'\cup\Lambda\cup\Pi''\stopq
      l_1\mu_1/p'\tightequal s
}(\#3)
\\
and
by (\#\tight\Theta2) 
and 
(\#2) 
\\\LINEmath{
 \forall p',p''\tightin\Pi'\tightcup\Lambda\stopq
 \inparenthesesinline{
     p'\tightequal p''
   \ \oder\ 
     \neitherprefix{p'}{p''}
 }
.}(\#4)
\\
Since
\\\LINEmath{
  \forall p'\tightin\Pi'\tightcup\Lambda\stopq
      \inparenthesesinlinetight{
         p'\tightin\TPOS{l_1}
       \ \implies\ 
         l_1/p'\in\V 
      }
;}
\\\LINEmath{
  \forall p''\tightin\Pi''\stopq
  \inparenthesesinline{
      p''\tightin\TPOS{l_1}
    \ \und\ 
      l_1/p''\notin\V 
  }
}(\#5)
\\
we get by (\#3) 
\\\LINEmath{
  \forall p'\tightin\Pi'\tightcup\Lambda\stopq
  \forall p''\tightin\Pi''\stopq
    \neitherprefix{p''}{p'}
}(\#6)
\\
and then together with (\#2) and (\#4)
\\\LINEmath{
  \forall p',p''\in\Pi'\uplus\Lambda\uplus\Pi''\stopq
  \inparenthesesinline{
      p'\tightequal p''
    \ \oder\
      \neitherprefix{p'}{p''}
  }
.}(\#7)
\\
Now due to (\#2) and (\#\tight\Theta3) we have
\\\LINEmath{
  l_1\mu_1'
  \tightequal
  \replpar
    {l_1\mu_1}
    {p'}
    {s}
    {p'\tightin\Pi'\tightcup\Lambda}
}(\#8)\\
and then by (\#6) and (\#3)
\\\LINEmath{
  \forall p''\tightin\Pi''\stopq
    l_1\mu_1'/p''
    \tightequal
    s
.}(\#9)

\pagebreak

\yestop
\noindent
Summing up and defining we have:
\\\LINEmath{
  \headroom
  \begin{array}[t]{@{}l@{}lrl@{}ll}
    &\check u_0
    &:=
    &\repl u q{\replpar{l_1\mu_1}{p'}{s'}{p'\tightin\Pi'}}
    &\replparsuffix{o}{s'}{o\in\Xi}
    &;
    \\
    &\check u_1
    &:=
    &\repl u q{\replpar{l_1\mu_1}{p'}{s'}{p'\tightin\Pi'\tightcup\Lambda}}
    &\replparsuffix{o}{s'}{o\in\Xi}
    &
    \\
    \mbox{(by (\#8))}
    &
    &=
    &\repl u q{l_1\mu_1'}
    &\replparsuffix{o}{s'}{o\in\Xi}
    &;
    \\
    &\check u_2
    &:=
    &\repl u q{\replpar{l_1\mu_1}{p'}{s'}{p'\tightin\Pi'\tightcup\Pi''}}
    &\replparsuffix{o}{s'}{o\in\Xi}
    &;
    \\
    &\check u_3
    &:=
    &\repl 
      u 
      q
      {\replpar
        {l_1\mu_1}
        {p'}
        {s'}
        {p'\tightin\Pi'\tightcup\Lambda\tightcup\Pi''}
      }
    &\replparsuffix{o}{s'}{o\in\Xi}
    &
    \\
    \mbox{(by (\#6), (\#8))}
    &
    &=
    &\repl u q{\replpar{l_1\mu_1'}{p'}{s'}{p'\tightin\Pi''}}
    &\replparsuffix{o}{s'}{o\in\Xi}
    &;
    \\
    &u'
    &=
    &\repl u q{r_1\mu_1}
    &
    &;
    \\
    &\hat u_0
    &:=
    &\repl u q{r_1\mu_1'}
    &\replparsuffix{o}{s'}{o\in\Xi}
    &
    \\
    \mbox{(by Claim~2)}
    &
    &=
    &\repl 
      u 
      q
      {\bar u_0}
    &\replparsuffix{o}{s'}{o\in\Xi}
    &;
    \\
    &\hat u_{i+1}
    &:=
    &\repl 
      u 
      q
      {\bar u_{i+1}}
    &\replparsuffix{o}{s'}{o\in\Xi}
    &.
    \\
  \end{array}
}
\begin{diagram}
u&&&&\rredindex{\beta+1,\,q}&&&&u'
&&\rredindexn n{\beta+1}&&v
\\
\dredparaindex{\omega+\omega,\,\Xi\cup(q\Pi')}
&&&&&&&&\dredparaindex{\omega+\omega,\,\,\Xi\,\cup\,(q\Theta_{r_1})}
&&
&&\drefltrans
\\
\check u_0
&&\rredparaindex{\omega+\omega,\,(q\Lambda)}
&&\check u_1
&&\rredindex{\omega+\omega,\,q}
&&\hat u_0
&&\rrefltrans
&&w_0
\\
&&
&&
&&
&&\dredindex{\omega+\omega,\,q\bar p_0}
&&
&&\drefltrans
\\
\dredparaindex{\omega+\omega,\,(q\Pi'')}
&&
&&\dredparaindex{\omega+\omega,\,(q\Pi'')}
&&
&&\hat u_1
&&\rrefltrans
&&w_1
\\
&&
&&
&&
&&\vdots
&&
&&\vdots
\\
\check u_2
&&\rredparaindex{\omega+\omega,\,(q\Lambda)}
&&\check u_3
&&\rrefltrans
&&\hat u_n
&&\rrefltrans
&&w_n
\end{diagram}
Due to (\#3) we have
\bigmath{
  \check u_2  
  \antiredparaindex{\omega+\omega,(q\Pi'')}
  \check u_0
  \redparaindex{\omega+\omega,(q\Lambda)}
  \check u_1
.}
\\
Thus by (\#6):
\bigmath{
  \check u_2  
  \redparaindex{\omega+\omega,(q\Lambda)}
  \check u_3
  \antiredparaindex{\omega+\omega,(q\Pi'')}
  \check u_1
.}

\noindent
We get 
\math{
  \check u_1
  \redindex{\omega+\omega,q}
  \hat u_0
}
by Lemma~\ref{lemma red is minimal} and 

\noindent
\underline{Claim~3:} \bigmath{C_{1}\mu_{1}'} is fulfilled.

\noindent
Moreover,
we get 
\bigmath{
   \hat u_0
   \refltrans
   w_0
   \antirefltrans
   v
} 
for some \math{w_0}
by 
(\#\tight\Theta1),
(\#\tight\Theta2),
(\#\tight\Theta3),
and
\math{
  P\inparenthesesinlinetight{
     v,u',s,s',
     \Xi
     \cup
     (q\Theta_{r_1})
   }
,}
which is smaller in the second or third position of the measure.

\noindent
\underline{Claim~1:}
We may assume that there is some 
\math{p\tightin\Pi''} with 
\bigmath{
  \repl
    {l_1\mu_1'}
    {p}
    {s'}
  \tightnotequal
  r_1\mu_1'
.}

\noindent
\underline{Claim~2:}
There are some 
\math{\bar n\tightin\N};
\bigmath{\FUNDEF{\bar p}{\{0,\ldots,\bar n\tight-1\}}{\N^\ast};}
\bigmath{\FUNDEF{\bar u}{\{0,\ldots,\bar n\}}        \tsigX      ;}
such that
\bigmath{
              \replpar{l_1\mu_1'}{p''}{s'}{p''\tightin\Pi''}
              \refltrans
              \bar u_n
;}
\\
\bigmath{
              \forall i\tightprec n\stopq
              \inparenthesesoplist{
                  \bar u_{i+1}
                  \tightequal           
                  \repl{\bar u_i}{\bar p_i}{\bar u_{i+1}/\bar p_i}
                \oplistund
                  \bar u_{i+1}/\bar p_i
                  \antirefltrans
                  \bar u_i/\bar p_i
                  <
                  s
              }
;}
and
\bigmath{
              \bar u_0
              \tightequal
              r_1\mu_1'
.}

\yestop
\noindent
Inductively for \math{i\prec n} we now get some
\math{w_{i+1}} with 
\bigmath{
  \hat u_{i+1}
  \refltrans
  w_{i+1}
  \antirefltrans
  w_i
}
due to 
Claim~2
and 
\math{
  P(w_i,\hat u_i,\bar u_i/\bar p_i,\bar u_{i+1}/\bar p_i,\{q\bar p_i\})
}
which is smaller in the first position of the measure by Claim~2.
Finally by Claim~2 
we get
\bigmath{
  \check u_3
  \refltrans
  \repl 
    u 
    q
    {\bar u_n}
  \replparsuffix{o}{s'}{o\in\Xi}
  \tightequal
  \hat u_n
.}
This completes the proof of {\bf Claim} due to 
\bigmath{
  \replpar{u}{o}{s'}{o\in\Pi}
  \tightequal
  \check u_2
  \refltrans 
  w_n
  \antirefltrans
  v
.}

\pagebreak

\yestop
\noindent
\underline{Proof of Claim~1:}
In case of 
\math{p,p'\tightin\Pi''}
with 
\bigmath{
  \repl
    {l_1\mu_1'}
    {p}
    {s'}
  \tightequal
  r_1\mu_1'
}
and
\bigmath{
  \repl
    {l_1\mu_1'}
    {p'}
    {s'}
  \tightequal
  r_1\mu_1'
}
we cannot have \bigmath{\neitherprefix{p}{p'}} because then 
by (\#9) 
we would get the contradiction
\bigmath{
  s
  \tightequal
  l_1\mu_1'/p
  \tightequal
  \repl
    {l_1\mu_1'}
    {p'}
    {s'}
  /p
  \tightequal
  r_1\mu_1'/p
  \tightequal
  \repl
    {l_1\mu_1'}
    {p}
    {s'}
  /p
  \tightequal
  s'
  \tight<
  s
.}
Therefore, 
if Claim~1 does not hold, 
\ie\ if
\bigmath{
  \forall p''\tightin\Pi''\stopq
    \repl
      {l_1\mu_1'}
      {p''}
      {s'}
    \tightequal
    r_1\mu_1'
,}
by (\#7) 
must have 
we have \bigmath{\CARD{\Pi''}\tightpreceq 1.}
In case \bigmath{\Pi''\tightequal\emptyset,} 
we have 
\bigmath{
  \check u_3
  \tightequal
  \check u_1
  \refltrans
  w_0
.}
Otherwise, in case of 
\bigmath{
  \Pi''
  \tightequal
  \{p\}
}
and 
\bigmath{
  \repl
    {l_1\mu_1'}
    {p}
    {s'}
  \tightequal
  r_1\mu_1'
,}
we have
\bigmath{
  \replpar
    {l_1\mu_1'}
    {p'}
    {s'}
    {p'\tightin\Pi''}
  \tightequal
  \repl
    {l_1\mu_1'}
    {p}
    {s'}
  \tightequal
  r_1\mu_1'
,}
and then
\bigmath{
  \check u_3
  \tightequal
  \hat u_0
  \refltrans
  w_0
.}
In both cases we have shown {\bf Claim} due to 
\bigmath{
  \replpar{u}{o}{s'}{o\in\Pi}
  \tightequal
  \check u_2
  \refltrans 
  \check u_3
  \refltrans 
  w_0
  \antirefltrans
  v
.}\QED{Claim~1}

\noindent
\underline{Proof of Claim~2}: 
Let \math{\xi\in\SUBST\V\V} be a bijection
with 
\bigmath{\xi[
\VAR{\sugarregelindex0}
]\cap
\VAR{\sugarregelindex1}
=\emptyset.}
Let $\varrho$ be given by (\math{x\in\V}): 
$x\varrho:=
 \left\{\begin{array}[c]{@{}l@{}l@{}}
  x\mu_1'       &\mbox{ if }x\in\VAR{\sugarregelindex1}\\
  x\xi^{-1}\mu_0&\mbox{ otherwise}                         \\
  \end{array}\right\}
$.\\
By (\#9) and (\#5) 
for the \math p of Claim~1
we have
\bigmath{
  l_{0}\xi\varrho
  \tightequal
  l_{0}\xi\xi^{-1}\mu_{0}
  \tightequal
  s
  \tightequal
    l_1\mu_1'
    /
    p
  \tightequal
  (l_{1}/p)\varrho
}
and 
\bigmath{
  l_1/p\tightnotin\V
.}
Thus,
let 
\bigmath{\Y:=\VAR{(\sugarregelindex0)\xi,\sugarregelindex1};}
\bigmath{
  \sigma
  :=
  \minmgu
    {\{(l_{0}\xi,l_{1}/p)\}}
    {\Y}
;} 
and
$\varphi\in\Xsubst$
with 
\bigmath{\domres{\inpit{\sigma\varphi}}\Y=\domres\varrho\Y.}
Let \math{t_0:=\repl{l_{1}}{p}{r_{0}\xi}}
and \math{t_1:=r_{1}}.
By Claim~1 we may assume 
\bigmath{
  t_0\sigma
  \tightnotequal
  t_1\sigma
}
(since otherwise
\math{
  \repl
    {l_1\mu_1'}
    {p}
    {s'}
  \tightequal
  \repl
    {l_1\mu_1'}
    {p}
    {r_0\mu_0}
  \tightequal
  t_0\sigma\varphi
  \tightequal
  t_1\sigma\varphi
  \tightequal
  r_1\mu_1'
}). 
Thus
\bigmath{
  ((t_0,
    C_0\xi,
    \dots
   ),\ 
   (t_1,
    C_1,
    \ldots
   ),\ 
   l_1,\ 
   \sigma,\ 
   p)
}
is a critical peak.
By \lemmamonotonicinbeta,
\bigmath{(C_0\xi\,C_1)\sigma\varphi}
is fulfilled \wrt\ \redindex{\omega+\omega}\@.
Since \bigmath{(l_1/p)\sigma\varphi\tightequal s} it makes sense to 
define 
\bigmath{
  \Delta
  :=
  \setwith
    {p'\tightin\TPOS{l_1}\tightsetminus\{p\}}
    {  l_1/p'\tightnotin\V
     \und
       (l_1/p')\sigma\varphi
       \tightequal
       s
    }
.}
Then by (\#5) and (\#9) we get 
\bigmath{\Pi''\subseteq\{p\}\tightcup\Delta.}
Thus by \bigmath{p\tightin\Pi''} we get 
\bigmath{\Pi''\tightcup\Delta=\{p\}\tightcup\Delta}
and therefore
\bigmath{
  \begin{array}[t]{lcl@{}l@{}ll}
  \replpar
    {l_1\mu_1'}
    {p''}
    {s'}
    {p''\tightin\Pi''}
  &\tightequal
  & {l_1\sigma\varphi}
  \replparsuffix
    {p''}
    {s'}
    {p''\tightin\Pi''}
  &\replparsuffix
    {p''}
    {s}
    {p''\tightin\Delta\tightsetminus\Pi''}
  \\
  &\refltrans
  & {l_1\sigma\varphi}
  \replparsuffix
    {p''}
    {s'}
    {p''\tightin\Pi''}
  &\replparsuffix
    {p''}
    {s'}
    {p''\tightin\Delta\tightsetminus\Pi''}
  \\
  &\tightequal
  & {l_1\sigma\varphi}
  \replparsuffix
    {p''}
    {s'}
    {p''\tightin\{p\}\tightcup\Delta}
  \\
  &\tightequal
  &\repl
        {l_1}
        {p}
        {r_0\xi}
  &\replparsuffix
      {p''}
      {r_0\xi}
      {p''\tightin\Delta}
  &\sigma\varphi
  \\
  &\tightequal
  &{t_0}
  &\replparsuffix
      {p''}
      {t_0/p}
      {p''\tightin\Delta}
  &\sigma\varphi
  .
  \\
  \end{array}
}
\\
Moreover, for \math{w}
with
\bigmath{
  w\,\transclosureinline{(\antired\tightcup\,\lhd)}\,\,
  (l_1/p)\sigma\varphi
}
due to 
\bigmath{
  (l_1/p)\sigma\varphi
  \tightequal
  s
}
we have \bigmath{w\tight<s}
and therefore \red\ is confluent
below \math w
due to 
\math{P\inparenthesesinlinetight{?,w,w,?,\{\emptyset\}}}
which is smaller in the first position of the measure.
Finally, by Claim~0 we get
\bigmath{
              \forall p''
                \tightin
                \TPOS{(l_1/p)\sigma\varphi}\tightsetminus\{\emptyset\}
                \stopq
                (l_1/p )\sigma\varphi/p''\tightnotin\DOM\redsub
.}
Thus, by \tight\rhd-quasi overlay joinability, 
there are some
\math{\bar n\tightin\N;}
\bigmath{\FUNDEF{\bar p}{\{0,\ldots,\bar n\tight-1\}}{\N^\ast};}
\bigmath{\FUNDEF{\bar u}{\{0,\ldots,\bar n\}}        \vt      ;}
with
\bigmath{
  t_0\replparsuffix
      {p''}
      {t_0/p}
      {p''\tightin\Delta}
  \sigma\varphi
  \refltrans
  \bar u_{\bar n}
;}
\\
\bigmath{
                  \forall i\tightprec\bar n\stopq
                  \inparenthesesoplist{
                      \bar u_{i+1}
                      \tightequal           
                      \repl{\bar u_i}{\bar p_i}{\bar u_{i+1}/\bar p_i}
                    \oplistund
                      \bar u_{i+1}/\bar p_i
                      \antirefltrans
                      \bar u_i/\bar p_i
                      \;\transclosureinline{(\antired\cup\tight\lhd)}\; 
                      (l_1/p)\sigma\varphi
                      \tightequal
                      s
                  }
}
and 
\bigmath{
                  \bar u_0
                  \tightequal
                  t_1\sigma\varphi
                  \tightequal
                  r_1\mu_1'
.}%
Finally, for all \math{\bar v}
due to \bigmath{s\tightin\tight\trianglerighteq{[{\rm T}]}}
we know that
\bigmath{s\transclosureinline{(\red\cup\tight\rhd)}\bar v}
implies
\bigmath{s\tight>\bar v.}%
\QED{Claim~2}

\noindent
\underline{Proof of Claim~3:}
For $(\bar{u}\boldequal\bar{v})$ in $C_{1}$ we have  
\bigmath{
  \bar{u}\mu_{1}\downarrowindex{\beta}\bar{v}\mu_{1}
.}
In case of \bigmath{\beta\tightequal0,}
due to 
(\#\tight\Theta1),
(\#\tight\Theta2),
and
(\#\tight\Theta3),
we have 
\bigmath{
  \replpar
    {\bar u\mu_1}  
    {p'}  
    {s'}  
    {p'\tightin\Theta_{\bar u}}  
  \antiredparaindex{\Theta_{\bar u}}
  \penalty-1
  \bar u\mu_1
  \tightequal
  \bar v\mu_1
  \redparaindex{\Theta_{\bar v}}
  \penalty-1
  \replpar
    {\bar v\mu_1}  
    {p'}  
    {s'}  
    {p'\tightin\Theta_{\bar v}}  
}
and then
\bigmath{
  \bar u\mu_1'
  \tightequal
  \replpar
    {\bar u\mu_1}  
    {p'}  
    {s'}  
    {p'\tightin\Theta_{\bar u}}  
  \redparaindex{\Theta_{\bar v}\setminus\Theta_{\bar u}}
}
\\\mbox{\math{
  \replpar
    {\bar u\mu_1}  
    {p'}  
    {s'}  
    {p'\tightin\Theta_{\bar u}\tightcup\Theta_{\bar v}}  
  \tightequal
  \replpar
    {\bar v\mu_1}  
    {p'}  
    {s'}  
    {p'\tightin\Theta_{\bar v}\tightcup\Theta_{\bar u}}  
  \antiredparaindex{\Theta_{\bar u}\setminus\Theta_{\bar v}}
  \replpar
    {\bar v\mu_1}  
    {p'}  
    {s'}  
    {p'\tightin\Theta_{\bar v}}  
  \tightequal
  \bar v\mu_1
}.}
\\
Otherwise, in case of \bigmath{0\tightprec\beta,}
we have for the sort 
$\bar{s}\in\sigsorts$ of $\bar{u}$:
$
 \bot
 \antitransindex{\beta}
 (\eqindexpp{\bar s}{\bar u}{\bar v})\mu_1
$.
We get 
\bigmath{
  \bot\downarrow(\eqindexpp{\bar s}{\bar u}{\bar v})\mu_1'
}
by 
\math{P\inparenthesesinlinetight{
          \bot,\ (\eqindexpp{\bar s}{\bar u}{\bar v})\mu_1,\ s,\ s'\ , 
          \Theta_{\eqindexpp{\bar s}{\bar u}{\bar v}}
       }
}
which is smaller in the second position.
Since there are 
no rules for $\bot$ and only one for \eqindexsymbol{\bar s}, 
this means 
\bigmath{\bar{u}\mu_{1}'\downarrow\bar{v}\mu_{1}'.}
\ 
For 
$(\Def\,\bar{u})$ in $C_{1}$ we know the existence of some
$\vec{\bar{u}}\in\tgcons$ with 
$\vec{\bar{u}}
 \antirefltransindex{\beta}
 \bar{u}\mu_{1}
$.
We get some $\hat{\bar{u}}$ with
$\vec{\bar{u}}\refltrans\hat{\bar{u}}\antirefltrans\bar{u}\mu_{1}'$ 
by 
\math{P\inparenthesesinlinetight{
          \vec{\bar{u}},\bar{u}\mu_{1},s,s',
          \Theta_{\bar u}
       }
}
which is smaller in the second position. 
By \lemmaconskeeping\
we get $\hat{\bar{u}}\in\tgcons$.
Finally, for $(\bar{u}\boldunequal\bar{v})$ in $C_{1}$
we have some 
$\vec{\bar{u}},\vec{\bar{v}}\in\tgcons$
with
$\bar{u}\mu_{1}\refltransindex{\beta}\vec{\bar{u}}\notconflu\vec{\bar{v}}
 \antirefltransindex{\beta}\bar{v}\mu_{1}$
(by \lemmaaboutconflu\ and \math{\omega\tightpreceq\beta}).
By applying the same procedure as before twice we get 
\bigmath{\hat{\bar{u}},\hat{\bar{v}}\in\tgcons} with
$\bar{u}\mu_{1}'\refltrans\hat{\bar{u}}\antirefltrans\vec{\bar{u}}
 \notconflu
 \vec{\bar{v}}\refltrans\hat{\bar{v}}\antirefltrans\bar{v}\mu_{1}'
$,
\ie\ 
$\bar{u}\mu_{1}'\refltrans\hat{\bar{u}}\notconflu\hat{\bar{v}}\antirefltrans\bar{v}\mu_{1}'
$.%
\QED{Claim~3}
\mycomment{%%%%%%%%%%%%%%%%%%%%%%%%%%%%%%%%%%%%%%%%%%%%%%%%%%%%%%%%%%%%%%%%%%%%%
\yestop
\noindent
\Qedtriple{``$q$ not strictly below any $p\in\Pi$''}

\pagebreak

\yestop
%%%%%%%%%%%%%%%%%%%%%%%%%%%%%%%%%%%%%%%%%%%%%%%%%%%%%%%%%%%%%%%%%%%%%%%%
\noindent
\underline{\underline{\underline{``$q$ strictly below $p\in\Pi$'':
There are some \math p, \math{p'} with \bigmath{p p'\tightequal q,} 
\bigmath{p'\tightnotequal\emptyset,} and 
\bigmath{p\tightin\Pi}:}}}

\noindent
\underline{Claim~4:}
For \bigmath{o\in\Pi\tightsetminus\{p\}} we have 
\bigmath{
  u'/o
  \tightequal 
  s
  \tightequal 
  l_{0}\mu_{0}
.}
\\\underline{Proof of Claim~4:}
Since \bigmath{\neitherprefix o p,} 
we have 
\bigmath{
  u'/o
  \tightequal 
  \repl{u}{pp'}{\ldots}/o
  \tightequal 
  u/o
  \tightequal 
  s
  \tightequal 
  l_{0}\mu_{0}
.}\QED{Claim~4}

\noindent
\underline{\underline{The (second) variable overlap case:
There are 
$\check{p},\hat{p},x$ with \bigmath{p'\tightequal \check{p}\hat{p}} and
\bigmath{l_{0}/\check{p}\tightequal x\in\V}:}}
\\
Define $\mu_0'$ by ($y\tightin\V$):
\math{
 y\mu_{0}':=
 \left\{\begin{array}[c]{@{}l@{}l@{}}
   \repl{x\mu_{0}}{\hat{p}}{r_{1}\mu_{1}}&\mbox{ if }y\tightequal x    \\
   y\mu_{0}                              &\mbox{ else}\\
 \end{array}\right\}
.} 
Define

\noindent\LINEmath{
  \begin{array}[t]{lrll}
    \check{u}
    &:=
    &\replpar{u}{o}{l_{0}\mu_{0}'}{o\tightin\Pi}
    &;
    \\
    \hat{u}
    &:=
    &\replpar u o{r_0\mu_0'}{o\tightin\Pi}
    &.
    \\
  \end{array}
}

\noindent
Since
(by Claim~4)

\noindent\LINEmath{
 u'
 \tightequal
 \repl{u}{p\check{p}\hat{p}}{r_{1}\mu_{1}}
 \tightequal
 \replpar{\repl u p{\repl{l_0\mu_0}{\check p}{x\mu_0'}}}
          o
         {l_0\mu_0}
         {o\in\Pi\setminus\!\{p\}}
} 
\begin{diagram}
  u&&&&\rredparaindex{\omega+\omega,\,\Pi}&&&&\replpar{u}{o}{s'}{o\tightin\Pi}
\\\dredindex{\beta+1}&&&&&&&&\drefltransindex{\beta+1}
\\u'&&\rredparaindex{\arr{{@{}l@{}}\scriptstyle\beta+1,\,\,
   \setwithstart
     o\check o\hat p
   \setwithmark 
     o
     \in
     \Pi
   \und
   \vspace{-.8ex}\\\scriptstyle
     l_{0}/\check{o}
     =
     x
   \und
     o\check o 
     \not=
     p\check p
 \setwithstop
}}
&&\check u&&\rredparaindex{\omega+\omega,\,\Pi}&&\hat u
\\\dredindexn{n}{\beta+1}&&&&\drefltrans&&&&\drefltrans
\\v&&\rrefltrans&&w&&\rrefltrans&&\circ
\end{diagram}
and 
$
 x\mu_{0}/\hat{p}
 \tightequal
 l_{0}\mu_{0}/\check{p}\hat{p}
 \tightequal
 s/p'
 \tightequal
 u/pp'
 \tightequal
 u/q
 \tightequal
 l_{1}\mu_{1}
 \redindex{\beta+1}
 r_{1}\mu_{1}
$,
we get some $w$ with
$v\refltrans w\antirefltrans\check{u}$
by
$P(v,u',l_{1}\mu_{1},r_{1}\mu_{1},
 \setwithstart
     o\check o\hat p
 \setwithmark 
     o
     \tightin
     \Pi
   \und
     l_{0}/\check{o}
     \tightequal
     x
   \und
     o\check o 
     \tightnotequal
     p\check p
 \setwithstop
 )
$,
which is smaller in the first position by
\bigmath{
 s
 \superterm
 s/p'
 \tightequal
 u/pp'
 \tightequal
 u/q
 \tightequal
 l_{1}\mu_{1}
.}
Now by
\bigmath{
 \hat{u}
 \antirefltransindex{\beta+1}
 \replpar{u}{o}{r_{0}\mu_{0}}{o\tightin\Pi}
 \tightequal
 \replpar{u}{o}{s'}{o\tightin\Pi}
}
for {\bf Claim}
we only have to show 
\bigmath{
  w
  \tight\downarrow
  \hat u 
.}
But this is given by Claim~6 below 
(which implies \math{l_0\mu_0'\redsimple r_0\mu_0'} by 
 Lemma~\ref{lemma red is minimal})
and
$P(w,\check{u},l_{0}\mu_{0}',r_{0}\mu_{0}',\Pi)$,
which is smaller in the first position of the measure by 
\bigmath{
 s
 \tightequal
 l_{0}\mu_{0}
 \trans
 l_{0}\mu_{0}'
.}
\\
\underline{Claim~6:}
\bigmath{C_{0}\mu_{0}'} is fulfilled.
\\
\underline{Proof of Claim~6:}
For $(\bar{u}\boldequal\bar{v})$ in $C_{0}$ we have  
some $w'$ with $\bar{u}\mu_{0}\refltrans w'\antirefltrans\bar{v}\mu_{0}$.
We get some $w''$ with $w'\refltrans w''\antirefltrans\bar{u}\mu_{0}'$
by
$P(w',\bar{u}\mu_{0},l_{1}\mu_{1},r_{1}\mu_{1},
   \setwith{o\hat{p}}{\bar{u}/o=x})$, 
and then \bigmath{w''\tight\downarrow\bar v \mu_0'}
by
\bigmath{
 P(w'',\bar{v}\mu_{0},l_{1}\mu_{1},r_{1}\mu_{1},\setwith{o\hat p}{\bar v/o=x})
,}
which are smaller in the first position of the measure by 
$s\superterm l_{1}\mu_{1}$ (shown above).
For 
$(\Def\,\bar{u})$ in $C_{0}$ we know the existence of some
$\vec{\bar{u}}\in\tgcons$ with 
$
 \vec{\bar{u}}
 \antirefltrans 
 \bar{u}\mu_{0}
$.
We get some $\hat{\bar{u}}$ with
$\vec{\bar{u}}\refltrans\hat{\bar{u}}\antirefltrans\bar{u}\mu_{0}'$ 
by 
$P(\vec{\bar{u}},\bar{u}\mu_{0},l_{1}\mu_{1},r_{1}\mu_{1},
   \setwith{o\hat{p}}{\bar{u}/o=x})$, 
which is smaller in the first position (as before). 
By \lemmaconskeeping\ 
\nopagebreak we get $\hat{\bar{u}}\in\tgcons$.
\nopagebreak Finally, for $(\bar{u}\boldunequal\bar{v})$ in $C_{0}$
\nopagebreak we have some 
\nopagebreak $\vec{\bar{u}},\vec{\bar{v}}\in\tgcons$
\nopagebreak with
\nopagebreak $\bar{u}\mu_{0}\refltrans \vec{\bar{u}}\notconflu\vec{\bar{v}}
 \antirefltrans \bar{v}\mu_{0}$.
\nopagebreak By applying the same procedure
\nopagebreak as before twice we get
\nopagebreak \bigmath{\hat{\bar{u}},\hat{\bar{v}}\in\tgcons} with 
\nopagebreak $\bar{u}\mu_{0}'\refltrans\hat{\bar{u}}\antirefltrans\vec{\bar{u}}
 \notconflu
 \vec{\bar{v}}\refltrans\hat{\bar{v}}\antirefltrans\bar{v}\mu_{0}'
$,
\nopagebreak\ie\ 
\nopagebreak$\bar{u}\mu_{0}'\refltrans\hat{\bar{u}}
 \notconflu
 \hat{\bar{v}}\antirefltrans\bar{v}\mu_{0}'
$.\\\Qed{Claim~6}\QEDdouble{The (second) variable overlap case}

\pagebreak

\noindent
\underline{\underline{The (second) critical peak case:
\bigmath{p'\in\TPOS{l_{0}}\ \wedge\ l_{0}/p'\not\in\V}:}}
\\\underline{Claim~5:}
There is some $w$ with 
\bigmath{
  s
  >
  \repl s{p'}{r_1\mu_1}
  \refltrans
  w
  \antirefltrans
  s'
.}
\\
\underline{Proof of Claim~5}:
Let \math{\xi\in\SUBST\V\V} be a bijection 
with 
\bigmath{\xi[
\VAR{\sugarregelindex1}
]\cap
\VAR{\sugarregelindex0}
\tightequal\emptyset.}
Let $\varrho$ be given by
$(x\tightin\V)$:
$x\varrho\!:=\!\!\left\{\!\begin{array}[c]{@{}l@{}l@{}}
  x\mu_{0}        &\mbox{ if }x\tightin\VAR{\sugarregelindex0}\\
  x\xi^{-1}\!\mu_{1}&\mbox{ otherwise}                         \\
  \end{array}\!\right\}\!
$\@.
By
\math{
 l_{1}\xi\varrho
  \tightequal 
 l_{1}\xi\xi^{-1}\mu_{1}
  \tightequal 
 u/q
  \tightequal 
 u/pp'
  \tightequal 
 l_{0}\mu_{0}/p'
  \tightequal 
 (l_{0}/p')\varrho
}
let 
\math{\Y\!:=\!\VAR{(\sugarregelindex1)\xi,\sugarregelindex0}},
\math{
  \sigma
  :=
  \minmgu
    {\{(l_{1}\xi,l_{0}/p')\}}
    {\Y}
} 
and
$\varphi\in\Xsubst$
with 
\bigmath{\domres{\inpit{\sigma\varphi}}\Y=\domres\varrho\Y.}
Let
\math{
  t_0
  :=
  \repl
    {l_0}
    {p'}
    {r_1\xi}
}
and
\math{t_1:=r_0}.
Since 
\bigmath{
  s/p' 
  \tightequal
  u/p p'
  \tightequal
  u/q
  \tightequal
  l_1\mu_1
}
we have 
\bigmath{
  s
  \red
  \repl
    {s}
    {p'}
    {r_1\mu_1}
.} 
Because
\bigmath{
  \repl
    {s}
    {p'}
    {r_1\mu_1}
  \tightequal 
  \repl
    {l_0\mu_0}
    {p'}
    {r_1\mu_1}
  \tightequal 
  t_0\sigma\varphi
}
and
\bigmath{
  t_1\sigma\varphi
  \tightequal 
  r_0\mu_0
  \tightequal 
  s'
}
we 
now
only have 
to show
\bigmath{
  t_0\sigma\varphi
  \tight\downarrow 
  t_1\sigma\varphi
.}
If 
\bigmath{
  t_0\sigma
  \tightequal
  t_1\sigma
,}
this is trivial.
Otherwise, if 
\bigmath{
  t_0\sigma
  \tightnotequal
  t_1\sigma
,} 
\bigmath{
  ((t_0,
    C_1\xi,
    \dots
   ),\ 
   (t_1,
    C_0,
    \ldots
   ),\ 
   l_0,\ 
   \sigma,\ 
   p')
}
is a critical peak.
By \lemmamonotonicinbeta,
\bigmath{(C_1\xi\,C_0)\sigma\varphi}
is fulfilled \wrt\ \red\@.
Moreover, for \math{w}
with
\bigmath{
  w\,\transclosureinline{(\antired\tightcup\,\lhd)}\,\,
  (l_0/p')\sigma\varphi
}
due to 
\bigmath{
  (l_0/p')\sigma\varphi
  \tightequal
  s/p'
  \subterm 
  s
}
we have \bigmath{w\tight<s}
and therefore \red\ is confluent
below \math w
due to 
\math{P\inparenthesesinlinetight{?,w,w,?,\{\emptyset\}}}
which is smaller in the first position of the measure.
Thus,
by \math\rhd-quasi overlay joinability,
choosing \math{\Delta:=\emptyset},
there are some
\math{\bar n\tightin\N;}
\bigmath{\FUNDEF{\bar u}{\{0,\ldots,\bar n\}}        \tsigX      ;}
with
\bigmath{
  t_0\sigma\varphi
  \refltrans
  \bar u_{\bar n}
  \antirefltrans
  \bar u_{\bar n-1}
  \antirefltrans
  \ldots
  \antirefltrans
  \bar u_0
  \tightequal 
  t_1\sigma\varphi
.}\QED{Claim~5}

\noindent
We get
\bigmath{
 u'
 \tightequal 
 \repl{u}{p p'}{r_1\mu_1}
 \tightequal 
 \repl{u}{p}{\repl s{p'}{r_1\mu_1}}
.}
Summing up and defining 
we have 
\\\LINEmath{
  \headroom
  \begin{array}[t]{lrl@{}ll}
      u'
      &=
      &\repl{u}{p}{\repl s{p'}{r_1\mu_1}}
      &&;
      \\
      u''
      &:=
      &\repl{u}{p}{\repl s{p'}{r_1\mu_1}}
      &\replparsuffix
         {o}
         {s'}
         {o\tightin\Pi\tightsetminus\{p\}}
      &;
      \\
      \hat u
      &:=
      &\repl u p w
      &\replparsuffix
         {o}
         {s'}
         {o\tightin\Pi\tightsetminus\{p\}}
      &;
      \\
      \check u
      &:=
      &\repl u p{s'}
      &\replparsuffix
         {o}
         {s'}
         {o\tightin\Pi\tightsetminus\{q\}}
      &.
      \\
  \end{array}
}
\begin{diagram}
  u&&&&\rredparaindex{\omega+\omega,\,\Pi}&&&&\check u
\\\dredindex{\beta+1}&&&&&&&&\drefltrans
\\u'&&\rredparaindex{\omega+\omega,\,\,\Pi\setminus\{p\}}
&&u''&&\rredindex{\omega+\omega,\,p}&&\hat u
\\\dredindexn n{\beta+1}&&&&\drefltrans&&&&\drefltrans
\\v&&\rrefltrans&&w'&&\rrefltrans&&\circ
\end{diagram}
We get 
\bigmath{
 \hat{u}
 \antirefltrans
 \check u
 \tightequal 
 \replpar{u}{o}{s'}{o\tightin\Pi}
}
by 
\bigmath{w\antirefltrans s'} (by Claim~5).
Thus,
for {\bf Claim} we only have to show 
$v\tight\downarrow\hat{u}$.
There is some $w'$ with $v\refltrans w'\antirefltrans u''$ by Claim~4
and $P(v,u',s,s',\Pi\tightsetminus\{p\})$, 
which is smaller in the second or third
position.
Finally, 
we get $w'\tight\downarrow\hat{u}$ 
by 
Claim~5
and 
$P(w',u'',\repl{s}{p'}{r_{1}\mu_{1}},w,\{p\})$, 
which is smaller in the first position
by Claim~5.\QEDdouble{The (second) critical peak case}
\\\Qedtriple{``$q$ strictly below $p\in\Pi$''}
}%comment%%%%%%%%%%%%%%%%%%%%%%%%%%%%%%%%%%%%%%%%%%%%%%%%%%%%%%%%%%%%%%%%%%%%%%
\end{proofparsepqed}

\pagebreak

\begin{proofparsepqed}
{Lemma~\ref{lemma level parallel closed second coarse level two}}
\underline{Claim~0:}
\red\ and \redindex\omega\ are commuting.
\\\underline{Proof of Claim~0:}
By the assumed strong commutation assumption 
and Lemma~\ref{lemma strong commutation one copy}
\bigmath{\redpara\circ\refltransindex\omega}
and \refltransindex{\omega} are commuting.
Since by Corollary~\ref{corollary parallel one}
we have
\bigmath{
  \red
  \subseteq
  \redpara
  \tight\circ
  \refltransindex{\omega}
  \subseteq
  \refltrans
,}
now \red\ 
and \redindex{\omega} are commuting, too.\QED{Claim~0}

\yestop
\noindent
\underline{Claim~1:}
If 
\bigmath{
    \refltransindex{\omega}
    \tight\circ
    \redpara
    \tight\circ
    \refltransindex{\omega}
}
strongly commutes over
\refltrans,
then
\red\ is confluent.
\\
\underline{Proof of Claim~1:}
\bigmath{
    \refltransindex{\omega}
    \tight\circ
    \redpara
    \tight\circ
    \refltransindex{\omega}
}
and
\refltrans\
are commuting
by Lemma~\ref{lemma strong commutation one copy}.
Since by Corollary~\ref{corollary parallel one}
we have
\bigmath{
  \red
  \subseteq
  \refltransindex{\omega}
  \tight\circ
  \redpara
  \tight\circ
  \refltransindex{\omega}
  \subseteq
  \refltrans
,}
now \red\
and \red\ are commuting, too.\QED{Claim~1}

\yestop
\yestop
\noindent
We are going to show the following property\footroom:
\\\LINEmath{
    w_0
    \antiredparaindex{\omega+\omega,\,\Pi_0}
    u
    \redparaindex{\omega+\omega,\,\Pi_1}
    w_1
  \quad\implies\quad
    w_0
    \refltransindex{\omega}
    \tight\circ
    \redpara
    \tight\circ
    \refltransindex{\omega}
    \circ
    \antirefltrans
    w_1
.}
\begin{diagram}
u&&&\rredparaindex{\omega+\omega,\,\Pi_1}
&&&w_1
\\
\dredparaindex{\omega+\omega,\,\Pi_0}
&&&&&&\drefltrans
\\
w_0&\rrefltransindex{\omega}&\circ&\rredpara
&\circ&\rrefltransindex{\omega}&\circ
\end{diagram}

\noindent
\underline{Claim~2:}
The above property implies
that
\bigmath{
            \refltransindex{\omega}
            \tight\circ
            \redpara
            \tight\circ
            \refltransindex{\omega}
}
strongly commutes over 
\refltrans\
and that \red\ is confluent.
\\
\underline{Proof of Claim~2:}
First we show the strong commutation.
By Lemma~\ref{lemma strong commutation one copy} it suffices to show that
\bigmath{
    \refltransindex{\omega}
    \tight\circ
    \redpara
    \tight\circ
    \refltransindex{\omega}
}
strongly commutes over
\red.
Assume
\bigmath{
    u''
    \antired
    u'
    \refltransindex{\omega}
    u
    \redpara
    w_1
    \refltransindex{\omega}
    w_2
}
(\cf\ diagram below).
By the strong commutation assumed for our lemma
and Corollary~\ref{corollary parallel one},
there are \math{w_0} and \math{w_0'} with
\bigmath{
    u''
    \refltransindex{\omega}
    w_0'
    \antirefltransindex{\omega}
    w_0
    \antiredpara
    u
.}
By the above property there are some \math{w_3}, \math{w_1'}
with
\bigmath{
    w_0
    \refltransindex{\omega}
    w_3
    \redpara
    \tight\circ
    \refltransindex{\omega}
    w_1'
    \antirefltrans
    w_1
.}
By Claim~0 we can close the peak
\bigmath{ 
    w_1'
    \antirefltrans
    w_1
    \refltransindex{\omega}
    w_2
}
according to 
\bigmath{
    w_1'
    \refltransindex{\omega}
    w_2'
    \antirefltrans
    w_2
}
for some \math{w_2'}.
By the assumed confluence of \redindex\omega,
we can close the peak
\bigmath{
  w_0'
  \antirefltransindex{\omega}
  w_0
  \refltransindex{\omega}
  w_3
}
according to 
\bigmath{
  w_0'
  \refltransindex{\omega}
  w_3'
  \antirefltransindex{\omega}
  w_3
}
for some \math{w_3'}.
To close the whole diagram, we only have to show that we can close the peak
\bigmath{
    w_3'
    \antirefltransindex{\omega}
    w_3
    \redpara
    \tight\circ
    \refltransindex{\omega}
    w_2'
}
according to 
\bigmath{
    w_3'
    \redpara
    \tight\circ
    \refltransindex{\omega}
    \circ
    \antirefltransindex{\omega}
    w_2'
,}
which is possible due to the strong commutation assumed for our lemma.
\begin{diagram}
u'&\rrefltransindex{\omega}&u
&&
&\rredpara
&&
&w_1&\rrefltransindex{\omega}&w_2
\\
&&\dredpara
&&
&&&
&\drefltrans
&&\drefltrans
\\
\dred
&&w_0
&\rrefltransindex{\omega}&w_3
&\rredpara
&\circ&\rrefltransindex{\omega}
&w_1'&\rrefltransindex{\omega}&w_2'
\\
&&\drefltransindex{\omega}
&&\drefltransindex{\omega}
&&&
&&&\drefltransindex{\omega}
\\
u''&\rrefltransindex{\omega}&w_0'
&\rrefltransindex{\omega}&w_3'
&\rredpara
&\circ&&\rrefltransindex{\omega}&&\circ
\end{diagram}
Finally, confluence of \red\ follows from Claim~1.\QED{Claim~2}

\yestop
\yestop
\yestop
\yestop
\noindent
\Wrog\ let the positions of \math{\Pi_i} be maximal
in the sense that for any \math{p\in\Pi_i} 
and \math{\Xi\subseteq \TPOS u\tightcap(p\N^+)}
we do not have 
\bigmath{
  u
  \redparaindex{\omega+\omega,(\Pi_i\setminus\{p\})\cup\Xi}
  w_i
}
anymore.
Then for each \math{i\prec2} and
\math{p\in\Pi_i} there are
\bigmath{\kurzregelindex{i,p}\in\R}
and
\bigmath{\mu_{i,p}\in\Xsubst}
with
\bigmath{u/p\tightequal l_{i,p}\mu_{i,p},}
\bigmath{r_{i,p}\mu_{i,p}\tightequal w_i/p,}
\math{C_{i,p}\mu_{i,p}} fulfilled \wrt\ \red.
Finally, for each \math{i\prec2}:
\bigmath{
  w_i\tightequal\replpar{u}{p}{r_{i,p}\mu_{i,p}}{p\tightin\Pi_i}
.}

\pagebreak

\yestop
\noindent
\underline{Claim~5:}
We may assume 
\bigmath{
  \forall i\tightprec2\stopq
  \forall p\tightin\Pi_i\stopq
    l_{i,p}\tightnotin\tcs
.}
\\
\underline{Proof of Claim~5:}
Define \math{\Xi_i:=\setwith{p\tightin\Pi_i}{l_{i,p}\tightin\tcs}}
and 
\math{
  u_i':=\replpar{u}{p}{r_{i,p}\mu_{i,p}}{p\tightin\Pi_i\tightsetminus\Xi_i}
}.
If we have succeeded with our proof under the assumption of Claim~5,
then we have shown 
\bigmath{
  u_0'
  \refltransindex{\omega}
  v_0
  \redpara
  \tight\circ
  \refltransindex{\omega}
  v_1
  \antirefltrans
  u_1'
}
for some \math{v_0}, \math{v_1}
(\cf\ diagram below).
By Lemma~\ref{lemma invariance of fulfilledness two} 
(matching both its \math\mu\ and \math\nu\ to our \math{\mu_{i,p}})
we get
\bigmath{
  \forall i\tightprec2\stopq
  \forall p\tightin\Xi_i\stopq
    l_{i,p}\mu_{i,p}\redindex\omega r_{i,p}\mu_{i,p}
}
and therefore
\bigmath{
  \forall i\tightprec2\stopq
  u_i'
  \refltransindex\omega
  w_i
.}
Thus from
\bigmath{
  v_1
  \penalty-1
  \antirefltrans
  \penalty-1
  u_1'
  \penalty-1
  \refltransindex\omega
  \penalty-1
  w_1
}
we get 
\bigmath{
  v_1
  \refltransindex\omega
  v_2
  \antirefltrans
  w_1
}
for some \math{v_2}
by Claim~0.
Due to the assumed confluence of \redindex\omega,
we can close the peak 
\bigmath{
  w_0
  \antirefltransindex{\omega}
  u_0'
  \refltransindex{\omega}
  v_0
}
according to 
\bigmath{
  w_0
  \refltransindex{\omega}
  v_0'
  \antirefltransindex{\omega}
  v_0
}
for some \math{v_0'}.
By the strong commutation assumption of our lemma,
from
\bigmath{
  v_0'
  \antirefltransindex\omega
  v_0
  \redpara
  \tight\circ
  \refltransindex{\omega}
  v_1
  \refltransindex{\omega}
  v_2
}
we can finally conclude
\bigmath{
  v_0'
  \redpara
  \tight\circ
  \refltransindex{\omega}
  \circ
  \penalty-1
  \antirefltransindex{\omega}
  v_2
.}
\begin{diagram}
u&&&\rredparaindex{\omega+\omega,\,\Pi_1\setminus\Xi_1}
&&&u_1'&\rrefltransindex{\omega}&w_1
\\
\dredparaindex{\omega+\omega,\,\Pi_0\setminus\Xi_0}&&&
&&&\drefltrans&&\drefltrans
\\
u_0'&\rrefltransindex{\omega}&v_0&\rredpara
&\circ&\rrefltransindex{\omega}&v_1&\rrefltransindex{\omega}&v_2
\\
\drefltransindex{\omega}&&\drefltransindex{\omega}&
&&&&&\drefltransindex{\omega}
\\w_0&\rrefltransindex{\omega}&v_0'&\rredpara
&\circ&&\rrefltransindex{\omega}&&\circ
\end{diagram}
\Qed{Claim~5}

\yestop
\noindent
Define
the set of inner overlapping positions by
\\\linemath{
  \displaystyle
  \Omega(\Pi_0,\Pi_1)
  :=
  \bigcup_{i\prec2}
    \setwith
      {p\tightin\Pi_{1-i}}
      {\exists q\tightin\Pi_i\stopq\exists q'\tightin\N^\ast\stopq
        p\tightequal q q'
      }
,}
and the length of a term by
\bigmath{\lambda(\anonymousfpp{t_0}{t_{m-1}}):=1+\sum_{j\prec m}\lambda(t_j).}

\yestop
\noindent
Now we start an induction on
\bigmath{  \displaystyle
  \sum_{p'\in\Omega(\Pi_0,\Pi_1)}\lambda(u/p')
}
in \tightprec.

\yestop
\noindent
Define the set of top positions by
\\\linemath{
  \displaystyle
  \Theta
  :=
      \setwith
      {p\tightin\Pi_0\tightcup\Pi_1}
      {\neg\exists q\tightin\Pi_0\tightcup\Pi_1\stopq
           \exists q'\tightin\N^+\stopq
             p\tightequal q q'
      }
.}
Since the prefix ordering is wellfounded we have
\bigmath{
  \forall i\tightprec2\stopq
  \forall p\tightin\Pi_i\stopq
  \exists q\tightin\Theta\stopq
  \exists q'\tightin\N^\ast\stopq
    p\tightequal q q'
.}
Then
\bigmath{
  \forall i\tightprec2\stopq
  w_i
  \tightequal
  \replpar{w_i}{q}{w_i/q}{q\tightin\Theta}
  \tightequal
  \replpar{\replpar{u}{p}{r_{i,p}\mu_{i,p}}{p\tightin\Pi_i}}
          {q}{w_i/q}{q\tightin\Theta}
  \tightequal
  \replpar{u}{q}{w_i/q}{q\tightin\Theta}
.}
Thus, it now suffices to show for all \math{q\in\Theta}
\\\linemath{\headroom\footroom
    w_0/q
    \refltransindex{\omega}
    \tight\circ
    \redpara
    \tight\circ
    \refltransindex{\omega}
    \circ
    \antirefltrans
    w_1/q
}
because then we have 
\\\LINEmath{
  w_0
  \tightequal
  \replpar{u}{q}{w_0/q}{q\tightin\Theta}
    \refltransindex{\omega}
    \tight\circ
    \redpara
    \tight\circ
    \refltransindex{\omega}
    \circ
    \antirefltrans
  \replpar{u}{q}{w_1/q}{q\tightin\Theta}
  \tightequal
  w_1
.}

\noindent
Therefore we are left with the following two cases for \math{q\in\Theta}:

\pagebreak

\yestop
\yestop
\noindent
\underline{\underline{\underline{\math{q\tightnotin\Pi_1}:}}}
Then \bigmath{q\tightin\Pi_0.}
Define \math{\Pi_1':=\setwith{p}{q p\tightin\Pi_1}}.
\noindent
We have two cases:

\yestop
\noindent
\underline{\underline{``The variable overlap (if any) case'':
\math{
  \forall p\tightin\Pi_1'\tightcap\TPOS{l_{0,q}}\stopq
    l_{0,q}/p\tightin\V
}:}}
\begin{diagram}
l_{0,q}\mu_{0,q}&&\rredparaindex{\omega+\omega,\,\Pi_1'}&&&&w_1/q
\\&&&&&&
\dequal
\\\dredindex{\omega+\omega,\,\emptyset}&&&&&&l_{0,q}\nu
\\&&&&&&\dred
\\w_0/q&\requal&r_{0,q}\mu_{0,q}&&\rredpara
&&r_{0,q}\nu
\end{diagram}
\noindent
Define a function \math\Gamma\ on \V\ by (\math{x\tightin\V}):
\bigmath{
  \Gamma(x):=
  \setwith{(p',p'')}
          {l_{0,q}/p'\tightequal x\ \wedge\ p' p''\in\Pi_1'}
.}

\noindent
\underline{Claim~7:}
There is some \math{\nu\in\Xsubst} with
\\\LINEmath{
  \forall x\in\V\stopq
    \inparenthesesoplist{
       x\mu_{0,q}
       \redpara
       x\nu
    \oplistund
       \forall p'\tightin\DOM{\Gamma(x)}\stopq
         x\nu
         \tightequal
         \replpar
           {x\mu_{0,q}}
           {p''}
           {r_{1,q p' p''}\mu_{1,q p' p''}}
           {(p',p'')\tightin\Gamma(x)}
    }
.}
\\
\underline{Proof of Claim~7:}
\\
In case of \bigmath{\DOM{\Gamma(x)}\tightequal\emptyset} we define
\bigmath{x\nu:=x\mu_{0,q}.}
If there is some \math{p'} such that 
\bigmath{\DOM{\Gamma(x)}\tightequal\{p'\}}
we define 
\bigmath{
  x\nu
  :=
  \replpar
    {x\mu_{0,q}}
    {p''}
    {r_{1,q p' p''}\mu_{1,q p' p''}}
    {(p',p'')\tightin\Gamma(x)}
.}
This is appropriate since due to 
\bigmath{
  \forall(p',p'')\tightin\Gamma(x)\stopq
    x\mu_{0,q}/p''
    \tightequal 
    l_{0,q}\mu_{0,q}/p' p''
    \tightequal 
    u/q p' p''
    \tightequal 
    l_{1,q p' p''}\mu_{1,q p' p''}
}
we have
\\\LINEmath{
  \begin{array}{l@{}l@{}l}
  x\mu_{0,q}&
  \tightequal&
  \replpar
    {x\mu_{0,q}}
    {p''}
    {l_{1,q p' p''}\mu_{1,q p' p''}}
    {(p',p'')\tightin\Gamma(x)}
  \redpara\\&&
  \replpar
     {x\mu_{0,q}}
     {p''}
     {r_{1,q p' p''}\mu_{1,q p' p''}}
     {(p',p'')\tightin\Gamma(x)}
  \tightequal
  x\nu.  
  \end{array}
}
\\
Finally, in case of \bigmath{\CARD{\DOM{\Gamma(x)}}\succ1,} \math{l_{0,q}} is
not linear in \math x. By the conditions of our lemma and Claim~5 this implies
\bigmath{x\tightin\Vcons.}
Since there is some \math{(p',p'')\in\Gamma(x)} with
\bigmath{
    x\mu_{0,q}/p''
    \tightequal 
    l_{1,q p' p''}\mu_{1,q p' p''}
}
this implies 
\bigmath{
   l_{1,q p' p''}\mu_{1,q p' p''}\tightin\tcc
}
and then 
\bigmath{
   l_{1,q p' p''}\tightin\tcs
}
which contradicts Claim~5.%
\QED{Claim~7}

\noindent
\underline{Claim~8:}
\bigmath{
  l_{0,q}\nu
  \tightequal 
  w_1/q
.}
\\
\underline{Proof of Claim~8:}
\\
By Claim~7 we get
\bigmath{
  w_1/q
  \tightequal
  \replpar 
    {u/q}
    {p' p''}
    {r_{1,q p' p''}\mu_{1,q p' p''}}
    {\exists x\tightin\V\stopq(p',p'')\tightin\Gamma(x)}
  \tightequal\\
  \replpar
    {\replpar
       {l_{0,q}}
       {p'}
       {x\mu_{0,q}}
       {l_{0,q}/p'\tightequal x\tightin\V}
    }
    {p' p''}
    {r_{1,q p' p''}\mu_{1,q p' p''}}
    {\exists x\tightin\V\stopq(p',p'')\tightin\Gamma(x)}
  \tightequal\\
  \replpar
    {l_{0,q}}
    {p'}
    {\replpar
       {x\mu_{0,q}}
       {p''}
       {r_{1,q p' p''}\mu_{1,q p' p''}}
       {(p',p'')\tightin\Gamma(x)}}
    {l_{0,q}/p'\tightequal x\tightin\V}
  \tightequal\\
  \replpar
    {l_{0,q}}
    {p'}
    {x\nu}
    {l_{0,q}/p'\tightequal x\tightin\V}
  \tightequal
  l_{0,q}\nu
.}\QED{Claim~8}

\noindent
\underline{Claim~9:}
\bigmath{
  w_0/q
  \redpara
  r_{0,q}\nu
.}
\\
\underline{Proof of Claim~9:} 
Since 
\bigmath{
  w_0/q
  \tightequal
  r_{0,q}\mu_{0,q}
,} 
this follows directly from Claim~7.%
\QED{Claim~9}

\noindent
By claims 8 and 9 it now suffices to show
\bigmath{
  l_{0,q}\nu
  \red
  r_{0,q}\nu
,}
which again follows from 
Lemma~\ref{lemma invariance of fulfilledness coarse level}
since \red\ and \redindex\omega\ are commuting by Claim~0
and since
\bigmath{
  \forall x\tightin\V\stopq 
    x\mu_{0,q}
    \refltrans
    x\nu
}
by Claim~7 and Corollary~\ref{corollary parallel one}.%
\\\Qeddouble{``The variable overlap (if any) case''}

\pagebreak

\yestop
\noindent
\underline{\underline{``The critical peak case'':
There is some \math{p\in \Pi_1'\tightcap\TPOS{l_{0,q}}}
with \math{l_{0,q}/p\tightnotin\V}:}}
\begin{diagram}
l_{0,q}\mu_{0,q}&\rredindex{\omega+\omega,\,p}&u'
&&&\rredparaindex{\omega+\omega,\,\Pi_1'\setminus\{p\}}
&&&w_1/q
\\
&&\dredparaindex{\omega+\omega,\,\Pi''}
&&&
&&&\drefltrans
\\
\dredindex{\omega+\omega,\,\emptyset}&&v_1
&\rrefltransindex{\omega}&v_3&\rredpara
&\circ&\rrefltransindex{\omega}&v_1'
\\
&&\drefltransindex{\omega}
&&\drefltransindex{\omega}&
&&&\drefltransindex{\omega}
\\
w_0/q&\rrefltransindex{\omega}&v_2
&\rrefltransindex{\omega}&v_4&\rredpara
&\circ&\rrefltransindex{\omega}&\circ
\end{diagram}
\underline{Claim~10:}
\bigmath{p\tightnotequal\emptyset.}
\\
\underline{Proof of Claim~10:}
If \bigmath{p\tightequal\emptyset,} then
\bigmath{\emptyset\tightin\Pi_1',} then
\bigmath{q\tightin\Pi_1,} which contradicts our global case assumption.%
\QED{Claim~10}

\noindent
Let \math{\xi\in\SUBST\V\V} be a bijection with 
\bigmath{
  \xi[\VAR{\kurzregelindex{1,q p}}]\cap\VAR{\kurzregelindex{0,q}}
  =
  \emptyset
.}
\\
Define
\bigmath{
  \Y
  :=
  \xi[\VAR{\kurzregelindex{1,q p}}]\cup\VAR{\kurzregelindex{0,q}}
.}
\\
Let \math{\varrho\in\Xsubst} be given by
$\ x\varrho=
\left\{\begin{array}{@{}l@{}l@{}}
  x\mu_{0,q}        &\mbox{ if }x\in\VAR{\kurzregelindex{0,q}}\\
  x\xi^{-1}\mu_{1,q p}&\mbox{ else}\\
\end{array}\right\}
\:(x\tightin\V)$.
\\
By
\math{
  l_{1,q p}\xi\varrho
  \tightequal 
  l_{1,q p}\xi\xi^{-1}\mu_{1,q p}
  \tightequal 
  u/q p
  \tightequal 
  l_{0,q}\mu_{0,q}/p  
  \tightequal 
  l_{0,q}\varrho/p
  \tightequal 
  (l_{0,q}/p)\varrho
}
\\
let
\math{
  \sigma:=\minmgu{\{(l_{1,q p}\xi}{l_{0,q}/p)\},\Y}
}
and
\math{\varphi\in\Xsubst}
with
\math{
  \domres{\inpit{\sigma\varphi}}\Y
  \tightequal
  \domres\varrho\Y
}.
\\
Define 
\math{
  u':=  
  \repl{l_{0,q}\mu_{0,q}}
       {p}
       {r_{1,q p}\mu_{1,q p}}
}.
We get
\\\LINEmath{
  \arr{{l@{}l}
    u'\tightequal
    &
    \repl
      {\replpar
         {u/q}
         {p'}
         {l_{1,q p'}\mu_{1,q p'}}
         {p'\tightin\Pi_1'\tightsetminus\{p\}}}
      {p}
      {r_{1,q p}\mu_{1,q p}}
    \redparaindex{\omega+\omega,\Pi_1'\setminus\{p\}}
    \\&
    \replpar{u/q}{p'}{r_{1,q p'}\mu_{1,q p'}}{p'\tightin\Pi_1'}    
    \tightequal
    w_1/q
  .
  }
}
\\
If 
\bigmath{
  \repl{l_{0,q}}{p}{r_{1,q p}\xi}\sigma
  \tightequal
  r_{0,q}\sigma
,}
then the proof is finished due to 
\\\LINEmath{
  w_0/q
  \tightequal
  r_{0,q}\mu_{0,q}
  \tightequal
  r_{0,q}\sigma\varphi
  \tightequal
  \repl{l_{0,q}}{p}{r_{1,q p}\xi}\sigma\varphi
  \tightequal
  u'
  \redparaindex{\omega+\omega,\Pi_1'\setminus\{p\}}
  w_1/q
.}
\\
Otherwise 
we have
\bigmath{
  (\,
   (\repl{l_{0,q}}{p}{r_{1,q p}\xi}\sigma,
    C_{1,q p}\xi\sigma,    
    1),\penalty-1\,
   (r_{0,q}\sigma,
    C_{0,q}\sigma,
    1),\penalty-1\,
    l_{0,q}\sigma,\penalty-1\,
    p\,)
  \in{\rm CP}(\R)
}
(due to Claim~5);
\bigmath{p\tightnotequal\emptyset}
(due to Claim~10);
\bigmath{C_{1,q p}\xi\sigma\varphi=C_{1,q p}\mu_{1,q p}}
is fulfilled \wrt\ \red;
\bigmath{C_{0,q}\sigma\varphi=C_{0,q}\mu_{0,q}}
is fulfilled \wrt\ \red.
Due to Claim~0 and our assumed \math\omega-coarse level parallel closedness 
we have
\bigmath{
  u'
  \tightequal
  \repl{l_{0,q}}{p}{r_{1,q p}\xi}\sigma\varphi
  \penalty-1
  \redpara
  \penalty-1
  v_1
  \penalty-1
  \refltransindex{\omega}
  v_2
  \antirefltransindex{\omega}
  r_{0,q}\sigma\varphi
  \tightequal
  r_{0,q}\mu_{0,q}
  \tightequal
  w_0/q
}
for some \math{v_1}, \math{v_2}.
We then have
\bigmath{
  v_1
  \antiredparaindex{\omega+\omega,\Pi''} 
  u'
  \redparaindex{\omega+\omega,\Pi_1'\setminus\{p\}}
  w_1/q
}
for some \math{\Pi''}.
By 
\bigmath{
  \displaystyle
  \sum_{p''\in\Omega(\Pi'',\Pi_1'\setminus\{p\})}
  \lambda(u'/p'')
  \ \ \preceq
  \sum_{p''\in\Pi_1'\setminus\{p\}}
  \lambda(u'/p'')
  \ \ =
  \sum_{p''\in\Pi_1'\setminus\{p\}}
  \lambda(u/q p'')
  \ \ \prec
}\\\bigmath{\displaystyle
  \sum_{p''\in\Pi_1'}
  \lambda(u/q p'')
  \ \ =
  \sum_{p'\in q\Pi_1'}
  \lambda(u/p')
  \ \ =
  \sum_{p'\in\Omega(\{q\},\Pi_1)}
  \lambda(u/p')
  \ \ \preceq
  \sum_{p'\in\Omega(\Pi_0,\Pi_1)}
  \lambda(u/p')
,}
due to our induction hypothesis
we get some \math{v_1'}, \math{v_3} with 
\bigmath{
  v_1
  \refltransindex{\omega}
  v_3
  \redpara
  \tight\circ
  \refltransindex{\omega}
  v_1'
  \antirefltrans
  w_1/q
.}
By confluence of \redindex\omega\
we can close the peak at \math{v_1} according to
\bigmath{
  v_2
  \refltransindex\omega
  v_4
  \antirefltransindex\omega
  v_3
}
for some \math{v_4}.
Finally 
by the strong commutation assumption of our lemma,
the peak at \math{v_3} can be closed according to 
\bigmath{
  v_4
  \redpara
  \tight\circ
  \refltransindex{\omega}
  \circ
  \antirefltransindex{\omega}
  v_1'
.}%
\\
\Qeddouble{``The critical peak case''}\QEDtriple{``\math{q\tightnotin\Pi_1}''}

\pagebreak

\noindent
\underline{\underline{\underline{\math{q\tightin\Pi_1}:}}}
Define \math{\Pi_0':=\setwith{p}{q p\tightin\Pi_0}}.
We have two cases:

\yestop
\noindent
\underline{\underline{``The second variable overlap (if any) case'':
\math{
  \forall p\tightin\Pi_0'\tightcap\TPOS{l_{1,q}}\stopq
    l_{1,q}/p\tightin\V
}:}}
\begin{diagram}
l_{1,q}\mu_{1,q}&&&\rredindex{\omega+\omega,\,\emptyset}&&&w_1/q
\\
&&&&&&\dequal
\\
\dredparaindex{\omega+\omega,\,\Pi_0'}
&&&&&&r_{1,q}\mu_{1,q}
\\
&&&&&&\dredpara
\\
w_0/q&\requal&l_{1,q}\nu&&\rred
&&r_{1,q}\nu
\end{diagram}
\noindent
Define a function \math\Gamma\ on \V\ by (\math{x\tightin\V}):
\bigmath{
  \Gamma(x):=
  \setwith{(p',p'')}
          {l_{1,q}/p'\tightequal x\ \wedge\ p' p''\in\Pi_0'}
.}

\noindent
\underline{Claim~11:}
There is some \math{\nu\in\Xsubst} with
\\\LINEmath{
  \forall x\in\V\stopq
    \inparenthesesoplist{
       x\nu
       \antiredpara
       x\mu_{1,q}
    \oplistund
       \forall p'\tightin\DOM{\Gamma(x)}\stopq
         \replpar
           {x\mu_{1,q}}
           {p''}
           {r_{0,q p' p''}\mu_{0,q p' p''}}
           {(p',p'')\tightin\Gamma(x)}
         \tightequal
         x\nu
    }
.}
\\
\underline{Proof of Claim~11:}
\\
In case of \bigmath{\DOM{\Gamma(x)}\tightequal\emptyset} we define
\bigmath{x\nu:=x\mu_{1,q}.}
If there is some \math{p'} such that 
\bigmath{\DOM{\Gamma(x)}\tightequal\{p'\}}
we define 
\bigmath{
  x\nu
  :=
  \replpar
    {x\mu_{1,q}}{p''}{r_{0,q p' p''}\mu_{0,q p' p''}}{(p',p'')\tightin\Gamma(x)}
.}
This is appropriate since due to 
\bigmath{
  \forall(p',p'')\tightin\Gamma(x)\stopq
    x\mu_{1,q}/p''
    \tightequal 
    l_{1,q}\mu_{1,q}/p' p''
    \tightequal 
    u/q p' p''
    \tightequal 
    l_{0,q p' p''}\mu_{0,q p' p''}
}
we have
\\\LINEmath{
  \begin{array}{l@{}l@{}l}
  x\mu_{1,q}&
  \tightequal&
  \replpar
    {x\mu_{1,q}}
    {p''}
    {l_{0,q p' p''}\mu_{0,q p' p''}}
    {(p',p'')\tightin\Gamma(x)}
  \redpara\\&&
  \replpar
     {x\mu_{1,q}}
     {p''}
     {r_{0,q p' p''}\mu_{0,q p' p''}}
     {(p',p'')\tightin\Gamma(x)}
  \tightequal
  x\nu.  
  \end{array}
}
\\
Finally, in case of \bigmath{\CARD{\DOM{\Gamma(x)}}\succ1,} \math{l_{1,q}} is
not linear in \math x. 
By the conditions of our lemma and Claim~5 this implies
\bigmath{x\tightin\Vcons.}
Since there is some \math{(p',p'')\in\Gamma(x)} with
\bigmath{
    x\mu_{1,q}/p''
    \tightequal 
    l_{0,q p' p''}\mu_{0,q p' p''}
}
this implies 
\bigmath{
   l_{0,q p' p''}\mu_{0,q p' p''}\tightin\tcc
}
and then 
\bigmath{
   l_{0,q p' p''}\tightin\tcs
}
which contradicts Claim~5.%
\QED{Claim~11}

\noindent
\underline{Claim~12:}
\bigmath{w_0/q\tightequal l_{1,q}\nu.}
\\
\underline{Proof of Claim~12:}
\\
By Claim~11 we get
\bigmath{
  w_0/q
  \tightequal
  \replpar 
    {u/q}
    {p' p''}
    {r_{0,q p' p''}\mu_{0,q p' p''}}
    {\exists x\tightin\V\stopq(p',p'')\tightin\Gamma(x)}
  \tightequal\\
  \replpar
    {\replpar
       {l_{1,q}}
       {p'}
       {x\mu_{1,q}}
       {l_{1,q}/p'\tightequal x\tightin\V}
    }
    {p' p''}
    {r_{0,q p' p''}\mu_{0,q p' p''}}
    {\exists x\tightin\V\stopq(p',p'')\tightin\Gamma(x)}
  \tightequal\\
  \replpar
    {l_{1,q}}
    {p'}
    {\replpar
       {x\mu_{1,q}}
       {p''}
       {r_{0,q p' p''}\mu_{0,q p' p''}}
       {(p',p'')\tightin\Gamma(x)}}
    {l_{1,q}/p'\tightequal x\tightin\V}
  \tightequal\\
  \replpar
    {l_{1,q}}
    {p'}
    {x\nu}
    {l_{1,q}/p'\tightequal x\tightin\V}
  \tightequal
  l_{1,q}\nu
.}\QED{Claim~12}

\noindent
\underline{Claim~13:}
\bigmath{
  r_{1,q}\nu
  \antiredpara
  w_1/q
.}
\\
\underline{Proof of Claim~13:} 
Since \bigmath{r_{1,q}\mu_{1,q}\tightequal w_1/q,} 
this follows directly from Claim~11.%
\QED{Claim~13}

\noindent
By claims 12 and 13 
using Corollary~\ref{corollary parallel one} 
it now suffices to show
\bigmath{
  l_{1,q}\nu
  \red
  r_{1,q}\nu
,}
which again follows from 
Lemma~\ref{lemma invariance of fulfilledness coarse level}
since \red\ and \redindex\omega\ are commuting by Claim~0 
and since
\bigmath{
  \forall x\tightin\V\stopq 
    x\mu_{1,q}
    \refltrans
    x\nu
}
by Claim~11 and Corollary~\ref{corollary parallel one}.%
\\\Qeddouble{``The second variable overlap (if any) case''}

\pagebreak

\yestop
\noindent
\underline{\underline{``The second critical peak case'':
There is some \math{p\in \Pi_0'\tightcap\TPOS{l_{1,q}}}
with \math{l_{1,q}/p\tightnotin\V}:}}
\begin{diagram}
l_{1,q}\mu_{1,q}
&&&
&\rredindex{\omega+\omega,\,\emptyset}&&
&&w_1/q
\\
\dredindex{\omega+\omega,\,p}
&&&
&&&
&&\drefltrans
\\
u'
&&&\rredparaindex{\omega+\omega,\,\Pi''}
&&&v_1
&\rrefltransindex{\omega}&v_2
\\
\dredparaindex{\omega+\omega,\,\Pi_0'\setminus\{p\}}
&&&
&&&\drefltrans
&&\drefltrans
\\
w_0/q
&\rrefltransindex{\omega}&\circ&\rredpara
&\circ&\rrefltransindex{\omega}&v_1'
&\rrefltransindex{\omega}&\circ
\end{diagram}
Let \math{\xi\in\SUBST\V\V} be a bijection with 
\bigmath{
  \xi[\VAR{\kurzregelindex{0,q p}}]\cap\VAR{\kurzregelindex{1,q}}
  =
  \emptyset
.}
\\
Define
\bigmath{
  \Y
  :=
  \xi[\VAR{\kurzregelindex{0,q p}}]\cup\VAR{\kurzregelindex{1,q}}
.}
\\
Let \math{\varrho\in\Xsubst} be given by
$\ x\varrho=
\left\{\begin{array}{@{}l@{}l@{}}
  x\mu_{1,q}        &\mbox{ if }x\in\VAR{\kurzregelindex{1,q}}\\
  x\xi^{-1}\mu_{0,q p}&\mbox{ else}\\
\end{array}\right\}
\:(x\tightin\V)$.
\\
By
\math{
  l_{0,q p}\xi\varrho
  \tightequal 
  l_{0,q p}\xi\xi^{-1}\mu_{0,q p}
  \tightequal 
  u/q p
  \tightequal 
  l_{1,q}\mu_{1,q}/p  
  \tightequal 
  l_{1,q}\varrho/p
  \tightequal 
  (l_{1,q}/p)\varrho
}
\\
let
\math{
  \sigma:=\minmgu{\{(l_{0,q p}\xi}{l_{1,q}/p)\},\Y}
}
and
\math{\varphi\in\Xsubst}
with
\math{
  \domres{\inpit{\sigma\varphi}}\Y
  \tightequal
  \domres\varrho\Y
}.
\\
Define 
\math{
  u':=  
  \repl{l_{1,q}\mu_{1,q}}
       {p}
       {r_{0,q p}\mu_{0,q p}}
}. 
We get
\\\LINEmath{
  \begin{array}{l@{}l@{}l}
  w_0/q&
  \tightequal&
  \replpar{u/q}{p'}{r_{0,q p'}\mu_{0,q p'}}{p'\tightin\Pi_0'}
  \antiredparaindex{\omega+\omega,\Pi_0'\setminus\{p\}}
  \\&&
  \repl
    {\replpar
       {u/q}{p'}{l_{0,q p'}\mu_{0,q p'}}{p'\tightin\Pi_0'\tightsetminus\{p\}}}
    {p}
    {r_{0,q p}\mu_{0,q p}}
  \tightequal
  u'
  .  
  \end{array}
}  
\\
If 
\bigmath{
  \repl{l_{1,q}}{p}{r_{0,q p}\xi}\sigma
  \tightequal
  r_{1,q}\sigma
,}
then the proof is finished due to 
\\\linemath{
  w_0/q
  \antiredparaindex{\omega+\omega,\Pi_0'\setminus\{p\}}
  u'
  \tightequal
  \repl{l_{1,q}}{p}{r_{0,q p}\xi}\sigma\varphi
  \tightequal
  r_{1,q}\sigma\varphi
  \tightequal
  r_{1,q}\mu_{1,q}
  \tightequal
  w_1/q.
}
Otherwise 
we have
\bigmath{
  (\,
   (\repl{l_{1,q}}{p}{r_{0,q p}\xi}\sigma,
    C_{0,q p}\xi\sigma,    
    1),\penalty-1\,
   (r_{1,q}\sigma,
    C_{1,q}\sigma,
    1),\penalty-1\,
    l_{1,q}\sigma,\penalty-1\,
    p\,)
  \in{\rm CP}(\R)
}
(due to Claim~5);
\bigmath{C_{0,q p}\xi\sigma\varphi=C_{0,q p}\mu_{0,q p}}
is fulfilled \wrt\ \red;
\bigmath{C_{1,q}\sigma\varphi=C_{1,q}\mu_{1,q}}
is fulfilled \wrt\ \red.
Due to Claim~0 and our assumed \math\omega-coarse level parallel joinability 
we have
\bigmath{
  u'
  \tightequal
  \repl{l_{1,q}}{p}{r_{0,q p}\xi}\sigma\varphi
  \redpara
  \penalty-1
  v_1
  \refltransindex{\omega}
  \penalty-1
  v_2
  \antirefltrans
  \penalty-1
  r_{1,q}\sigma\varphi
  \tightequal
  r_{1,q}\mu_{1,q}
  \tightequal
  w_1/q
}
for some \math{v_1}, \math{v_2}.
We then have
\bigmath{
  w_0/q
  \antiredparaindex{\omega+\omega,\Pi_0'\setminus\{p\}}
  u'
  \redparaindex{\omega+\omega,\Pi''} 
  v_1
}
for some \math{\Pi''}.
Since
\bigmath{\displaystyle
  \sum_{p''\in\Omega(
    \Pi_0'\setminus\{p\}
    ,
    \Pi''
    )}
  \lambda(u'/p'')
  \ \ \preceq
  \sum_{p''\in\Pi_0'\setminus\{p\}}
  \lambda(u'/p'')
  \ \ =
  \sum_{p''\in\Pi_0'\setminus\{p\}}
  \lambda(u/q p'')
  \ \ \prec
  \sum_{p''\in\Pi_0'}
  \lambda(u/q p'')
  \ \ =
  \sum_{p'\in q\Pi_0'}
  \lambda(u/p')
  \ \ =
  \sum_{p'\in\Omega(
    \Pi_0
    ,
    \{q\}
    )}
  \lambda(u/p')
  \ \ \preceq
  \sum_{p'\in\Omega(\Pi_0,\Pi_1)}
  \lambda(u/p')
}
due to our induction hypothesis
we get some \math{v_1'} with 
\bigmath{
  w_0/q
  \refltransindex{\omega}
  \tight\circ
  \redpara
  \tight\circ
  \refltransindex{\omega}
  v_1'
  \antirefltrans
  v_1
.} 
Finally the peak at \math{v_1} can be closed according to
\bigmath{
  v_1'
  \refltransindex{\omega}
  \circ
  \antirefltrans
  v_2
}
by Claim~0.%
\\\Qeddouble{``The second critical peak case''}
\end{proofparsepqed}

\pagebreak
\begin{proof}{Lemma~\ref{lemma invariance of fulfilledness coarse level}}
By Lemma~\ref{lemma red is minimal} it suffices to show 
for each literal \math L in \math C that
\math{L\nu} is fulfilled \wrt\ \redsub.
Note that we already know that 
\math{L\mu} is fulfilled \wrt\ \redsub.
Since \bigmath{\VAR C\tightsubseteq\Vcons,}
for all \math x in \VAR C we have \bigmath{x\mu\tightin\tcc}
and then by \lemmaconskeeping\ \bigmath{x\mu\refltransindex{\RX,\omega}y\mu.}
\\
\underline{\math{L=(s_0\boldequal s_1)}:}
We have
\bigmath{
  s_0\nu
  \antirefltransindex{\RX,\omega}
  s_0\mu
  \refltranssub
  t_0
  \antirefltranssub
  s_1\mu
  \refltransindex    {\RX,\omega}
  s_1\nu
}
for some \math{t_0.}
By the inclusion assumption of the lemma we get some \math v with
\bigmath{
  s_0\nu
  \refltranssub
  v
  \antirefltranssub
  t_0
} 
and then 
(due to
\bigmath{
  v
  \antirefltranssub
  s_1\mu
  \refltransindex    {\RX,\omega}
  s_1\nu
)}
\bigmath{
  v
  \refltranssub
  \circ
  \antirefltranssub
  s_1\nu
.}
\\\underline{$L=(\DEF s)$:}
We know the existence of
\math{t\in\tgcons}
with
\bigmath{
  s\nu
  \antirefltransindex{\RX,\omega}
  s\mu
  \refltranssub
  t
.}
By the above inclusion property again,
there is some
\math{t'} with
\bigmath{
  s\nu
  \refltranssub
  t'
  \antirefltranssub
  t
.}
By \lemmaconskeeping\ we get \bigmath{t'\tightin\tgcons.}
\\
\underline{$L=(s_0\boldunequal s_1)$:}
There exist some \math{t_0,t_1\in\tgcons}
with
\bigmath{
  \forall i\tightprec2\stopq
    s_i\nu
    \antirefltransindex{\RX,\omega}
    s_i\mu
    \refltranssub
    t_i
}
and 
\bigmath{
  t_0
  \notconflusub
  t_1
.}
Just like above
we get \math{t_0',\ t_1'\in\tgcons} with
\bigmath{
  \forall i\tightprec2\stopq
    s_i\nu
    \refltranssub
    \penalty-1
    t_i'
    \penalty-1
    \antirefltranssub
    t_i
.}
Finally 
\bigmath{
  t_0'
  \antirefltranssub
  t_0
  \notconflusub
  t_1
  \refltranssub
  t_1'
}
implies
\bigmath{
  t_0'
  \notconflusub
  t_1'
.}
\end{proof}

\vfill
\end{document}